\PassOptionsToPackage{dvipsnames,svgnames}{xcolor}
\documentclass[sigconf, screen]{acmart}

\setcopyright{none}
\renewcommand\footnotetextcopyrightpermission[1]{}
\acmConference[]{}{}{}
\usepackage{amssymb}
\usepackage{multirow}
\usepackage{enumitem}
\AtBeginDocument{%
  }

\usepackage{booktabs}
\usepackage{tabularx}
\usepackage{array}[=2016-10-06]
\usepackage{longtable}
\newcolumntype{Y}{>{\raggedright\arraybackslash}X}
\usepackage{listings}
\usepackage{pgfplots}
\pgfplotsset{compat=1.18}
\usepackage{pgfplotstable}

\definecolor{myblue}{RGB}{100,149,237}
\definecolor{mygreen}{RGB}{60,179,113}
\definecolor{mysalmon}{RGB}{250,128,114}
\definecolor{mypurple}{RGB}{147,112,219}
\definecolor{myorange}{RGB}{210,140,80}

\AtBeginDocument{\microtypesetup{expansion=false}}

\renewcommand{\dblfloatpagefraction}{0.7}
\begin{document}

\title[FashionMV]{FashionMV: Product-Level Composed Image Retrieval\\with Multi-View Fashion Data}

\author{Peng Yuan}
\email{yp24@mails.tsinghua.edu.cn}
\affiliation{%
  \institution{Tsinghua University}
  \city{Beijing}
  \country{China}
}

\author{Bingyin Mei}
\email{meiby25@mails.tsinghua.edu.cn}
\affiliation{%
  \institution{Tsinghua University}
  \city{Beijing}
  \country{China}
}

\author{Hui Zhang}
\email{huizhang@tsinghua.edu.cn}
\affiliation{%
  \institution{Tsinghua University}
  \city{Beijing}
  \country{China}
}

\renewcommand{\shortauthors}{Yuan, Mei, and Zhang}

\begin{abstract}
Composed Image Retrieval (CIR) retrieves target images using a reference image paired with modification text.
Despite rapid advances, all existing methods and datasets operate at the \emph{image level}---a single reference image plus modification text in, a single target image out---while real e-commerce users reason about \emph{products} shown from multiple viewpoints.
We term this mismatch \textbf{View Incompleteness} and formally define a new \textbf{Multi-View CIR} task that generalizes standard CIR from image-level to product-level retrieval.
To support this task, we construct \textbf{FashionMV}, the first large-scale multi-view fashion dataset for product-level CIR, comprising \textbf{127K} products, \textbf{472K} multi-view images, and over \textbf{220K} CIR triplets, built through a fully automated pipeline leveraging large multimodal models.
We further propose \textbf{ProCIR} (\textbf{Pro}duct-level \textbf{C}omposed \textbf{I}mage \textbf{R}etrieval), a modeling framework built upon a multimodal large language model that employs three complementary mechanisms---two-stage dialogue, caption-based alignment, and chain-of-thought guidance---together with an optional supervised fine-tuning (SFT) stage that injects structured product knowledge prior to contrastive training.
Systematic ablation across 16 configurations (8 mechanism variants $\times$ 2 initializations) on three fashion benchmarks reveals that: (1)~alignment is the single most critical mechanism; (2)~the two-stage dialogue architecture is a prerequisite for effective alignment; and (3)~SFT and chain-of-thought serve as partially redundant knowledge injection paths---once the base model has internalized product semantics through SFT, chain-of-thought becomes unnecessary.
Our best 0.8B-parameter model outperforms all baselines, including general-purpose embedding models 10$\times$ its size.
The dataset, model, and code are publicly available at \url{https://github.com/yuandaxia2001/FashionMV}.
\end{abstract}

\maketitle
\pagestyle{plain}
\thispagestyle{plain}

% ================================================================
\section{Introduction}
\label{sec:intro}

Composed Image Retrieval (CIR)~\cite{TIRG} retrieves target images using a reference image paired with modification text. The field has advanced rapidly in fusion strategies~\cite{SAC,CLIP4CIR}, training paradigms~\cite{LinCIR,CompoDiff,GenZSCIR}, LLM-based methods~\cite{CoLLM}, multi-turn interaction~\cite{MAI}, fine-grained semantics~\cite{FineCIR}, and video domains~\cite{EgoCVR}. Yet from TIRG (2019) to CoLLM (2025), \emph{all methods and datasets assume a single reference image}---the visual input side has never been challenged.

\begin{figure}[h]
  \centering
  \includegraphics[width=\columnwidth]{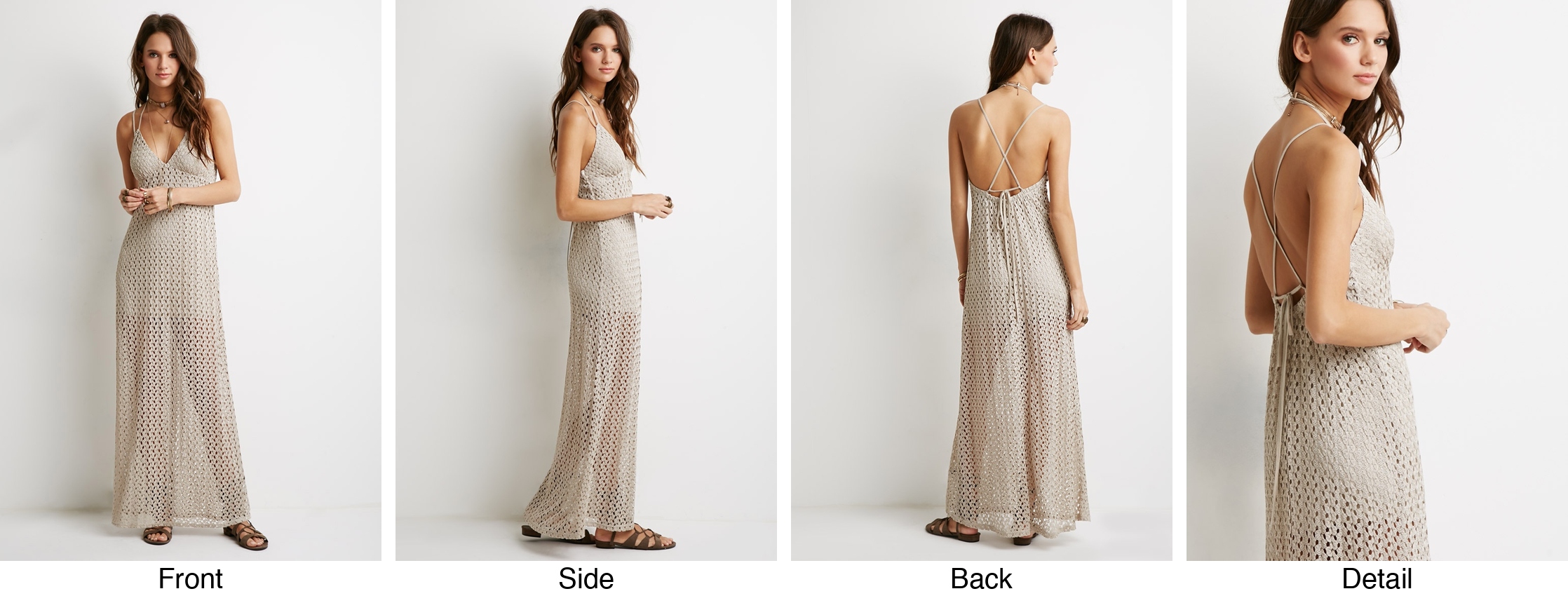}
  \caption{A fashion product displayed from four viewpoints. The deep V-neckline is only visible from the front, while the open-back criss-cross straps are only visible from the back---no single image can capture both defining features, illustrating \emph{View Incompleteness}.}
  \label{fig:multiview_example}
\end{figure}

This single-image assumption causes a fundamental problem we term \textbf{View Incompleteness}. In fashion e-commerce, products are displayed from multiple viewpoints (front, side, back, detail, etc.), and users reason holistically. Figure~\ref{fig:multiview_example} shows a concrete example: a deep V-neckline is only visible from the front, while the open-back criss-cross strap design is only visible from the back---no single image can capture both defining features simultaneously. When a modification involves such unobserved regions (e.g., requesting a ``backless design'' from a frontal view), retrieval becomes inherently unreliable. Both mainstream directions are constrained: (i)~General CIR methods~\cite{TIRG,CLIP4CIR,Pic2Word,LinCIR,CoLLM} operate on a single visual vector and cannot recover unobserved viewpoint details. (ii)~Fashion vision-language methods~\cite{FashionViL,FashionVLP,FashionFINE} recognize multi-view data's value but do not aggregate views into a unified product-level embedding. For instance, FashionViL~\cite{FashionViL} proposes Multi-View Contrastive Learning to encourage per-image semantic consistency across views, yet retrieval still operates on individual images---the gap from image-level to product-level remains unbridged.

Moreover, existing datasets (e.g., FashionIQ~\cite{FashionIQ}) and evaluation protocols are confined to image-level retrieval, ignoring the ``product entity'' concept central to e-commerce.

To address this gap, we propose a unified solution spanning task definition, dataset construction, and method design:

\textbf{(i)~Task.} We formally define \textbf{Multi-View CIR}, where both query and gallery operate at the product level. Each product aggregates its multi-view image set $\mathcal{V}^P{=}\{I_1^P, \ldots, I_{N_P}^P\}$ into a unified embedding combined with modification text to form the query. The formulation degenerates to standard CIR when $N_P{=}1$.

\textbf{(ii)~Data.} We construct \textbf{FashionMV}, the first large-scale multi-view fashion dataset for product-level CIR, comprising \textbf{127K} products, \textbf{472K} images, and over \textbf{220K} CIR triplets via a fully automated pipeline.

\textbf{(iii)~Method.} We propose \textbf{ProCIR}, a framework built upon a multimodal large language model with three complementary mechanisms: a \emph{two-stage dialogue architecture} that decouples visual perception from modification reasoning, producing a pure visual embedding for cross-modal alignment; \emph{caption-based alignment} that injects product-level semantics into the embedding space; and \emph{chain-of-thought (CoT) guidance} that progressively injects and removes product captions during reasoning. We further introduce an optional \emph{supervised fine-tuning (SFT)} stage that injects structured product knowledge prior to contrastive training; ablation reveals that SFT and CoT serve as partially redundant knowledge injection paths, and CoT becomes unnecessary once the base model has been fine-tuned.

Our main contributions are:
\begin{itemize}
  \item We formally define \textbf{Multi-View CIR}, generalizing CIR from image-level to product-level retrieval.
  \item We construct \textbf{FashionMV}, the first large-scale multi-view fashion dataset for product-level CIR (127K products, 472K images, 220K+ triplets) built through a fully automated pipeline.
  \item We propose \textbf{ProCIR}, which adapts a multimodal LLM for product-level CIR through three complementary mechanisms---two-stage dialogue, caption-based alignment, and chain-of-thought guidance---together with an optional SFT stage; systematic ablation across 16 configurations reveals that SFT and CoT are partially redundant knowledge injection paths.
\end{itemize}

% ================================================================
\section{Related Work}
\label{sec:related}

\begin{figure*}[t]
  \centering
  \includegraphics[width=\textwidth]{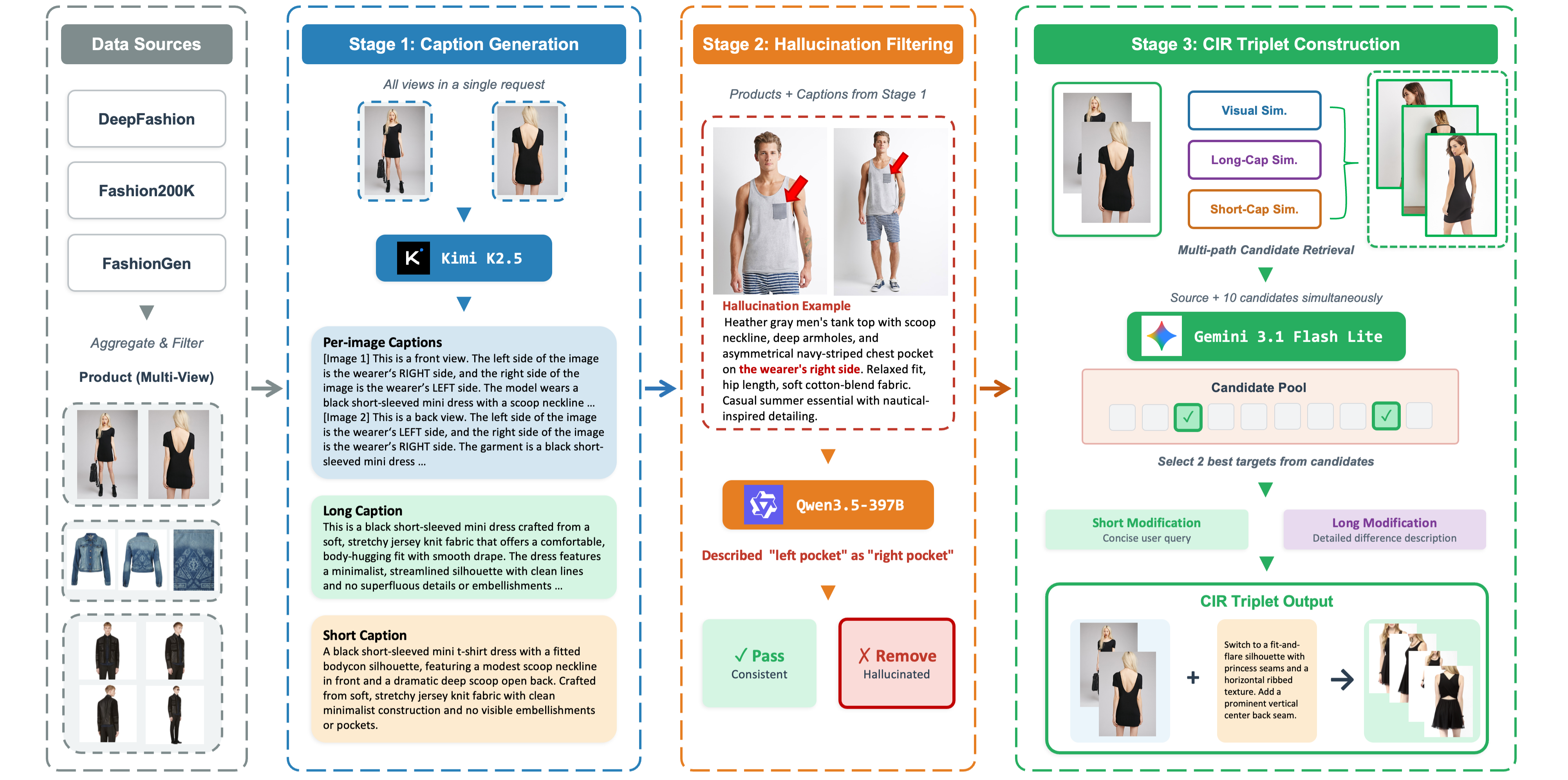}
  \caption{Overview of the FashionMV dataset construction pipeline. \textbf{Stage~1}: multi-view images from three data sources are fed to Kimi K2.5 in a single request to generate per-image captions, long captions, and short captions. \textbf{Stage~2}: Qwen3.5-397B cross-checks directional descriptions against visual evidence, removing products with confirmed hallucinations. \textbf{Stage~3}: multi-path candidate retrieval (visual, long-caption, and short-caption similarity) produces a candidate pool; Gemini 3.1 Flash Lite selects up to 2 best targets from 10 candidates and generates short and long modification texts.}
  \label{fig:data_pipeline}
\end{figure*}

\subsection{Composed Image Retrieval}
\label{sec:cir}

CIR has progressed from supervised feature fusion to zero-shot and LLM-based methods.
Early supervised work composes image and text features through residual gating~\cite{TIRG}, compositional autoencoders~\cite{ComposeAE}, dual composition networks with composition--correction learning~\cite{DCNet}, and cross-attention driven shift encoding~\cite{CASE}.

CLIP~\cite{CLIP} shifted the field toward pre-trained alignment spaces. CLIP4CIR~\cite{CLIP4CIR} introduced a combiner network in the CLIP space and remains a widely used baseline. Zero-shot CIR (ZS-CIR) methods avoid expensive triplet annotations: Pic2Word~\cite{Pic2Word} and SEARLE~\cite{SEARLE} map images to pseudo-word tokens; LinCIR~\cite{LinCIR} trains only on text via self-masking projection; SLERP+TAT~\cite{SLERPTTAT} merges bimodal embeddings through spherical linear interpolation; and CompoDiff~\cite{CompoDiff} leverages latent diffusion models for generative ZS-CIR. More recently, CoLLM~\cite{CoLLM} uses LLMs as the backbone for joint image--text embeddings. FineCIR~\cite{FineCIR}, MAI~\cite{MAI}, and EgoCVR~\cite{EgoCVR} extend CIR to fine-grained semantics, multi-turn interaction, and egocentric video, respectively.

Throughout this line of work, the visual input side has remained unchanged: every method takes exactly one reference image, leaving product-level retrieval unaddressed.

\subsection{Visual-Language Representation in Fashion}
\label{sec:fashion}

Fashion data involves fine-grained attributes (material, cut, style) that place heavy demands on visual-language representations. FashionVLP~\cite{FashionVLP} fuses multi-level visual context from a single image (whole image, cropped clothing, landmark regions, and regions of interest) through an asymmetric VLP transformer. FashionFINE~\cite{FashionFINE} combines global and patch-level embeddings with a modality-agnostic adapter and hard negative mining to align fine-grained details on FashionGen and FashionIQ. FAME-ViL~\cite{FAMEViL} processes multiple heterogeneous fashion tasks with a single model via cross-attention and task-specific adapters, saving 61.5\% parameters. FaD-VLP~\cite{FaDVLP} proposes a unified fashion vision-language pre-training framework that supports both retrieval and captioning. ADDE~\cite{ADDER} learns attribute-driven disentangled representations, enabling controllable modification of individual attributes without affecting others.

On the multi-view side, FashionViL~\cite{FashionViL} observes that fashion e-commerce data associates ``\emph{more than one image with a given text}'' and proposes Multi-View Contrastive Learning (MVC) as a pre-training task. MVC aligns each single view's visual representation with a cross-view multimodal representation (another view combined with text), encouraging per-image semantic consistency rather than aggregating multiple views into one product-level embedding. Existing methods therefore treat multi-view images as a form of data augmentation for improving single-image robustness; at inference time, retrieval still operates on individual images and no method performs true multi-view aggregation at the product level.

\subsection{Fashion and CIR Datasets}
\label{sec:datasets}

FashionIQ~\cite{FashionIQ} collects relative descriptions from user feedback and is the most widely used fashion CIR benchmark. CIRR~\cite{CIRR} extends CIR to open-domain images; CIRCO~\cite{SEARLE} introduces multiple ground truths to mitigate false negatives. GeneCIS~\cite{GeneCIS} tests generalization across diverse similarity conditions, and FACap~\cite{FACap} constructs over 227K fine-grained fashion CIR triplets via an automated VLM+LLM pipeline.

All of these datasets enforce a \textbf{single-view paradigm}: every triplet $(I_{\text{ref}}, T_{\text{mod}}, I_{\text{target}})$ is defined at the image level.
A telling example is FACap~\cite{FACap}, whose source images originate from Fashion200K~\cite{Fashion200K} and DeepFashion-MultiModal~\cite{Text2Human}---both containing multiple images per product---yet FACap explicitly filters out all intra-product pairs and operates entirely at the single-image level, discarding the multi-view structure that could otherwise provide complementary cross-viewpoint information.
No existing CIR dataset binds multi-view images into product-level groups or provides product-level descriptions; FashionMV is the first to do so.

% ================================================================
\section{The FashionMV Dataset}
\label{sec:dataset}

We construct FashionMV, the first large-scale multi-view fashion dataset for product-level CIR, through a fully automated three-stage pipeline powered by large multimodal models. Figure~\ref{fig:data_pipeline} illustrates the complete pipeline.

\subsection{Data Sources}
\label{sec:data_source}

We aggregate products from DeepFashion~\cite{DeepFashion}, Fashion200K~\cite{Fashion200K}, and FashionGen~\cite{FashionGen} (train and validation splits), each containing $2$--$5$ display images per product from different viewpoints. After filtering non-clothing items, the raw collection contains \textbf{144,396} clothing products with \textbf{535K} images.

\subsection{Stage 1: Multi-View Caption Generation}
\label{sec:stage1}

For each product, we feed all its multi-view images to Kimi K2.5~\cite{KimiK2.5} in a single request, generating: (i)~\emph{per-image captions} ($50$--$200$ words) identifying the viewpoint and view-specific details; (ii)~a \emph{long caption} ($200$--$400$ words) synthesizing all viewpoints into a comprehensive product description; and (iii)~a \emph{short caption} ($\sim$50 words) highlighting the garment type, style, color, and distinctive features. These captions later serve as supervision signals for caption-based alignment (\S\ref{sec:align}).

\subsection{Stage 2: Directional Hallucination Filtering}
\label{sec:stage2}

Multi-view fashion images require correct left--right directional reasoning (e.g., mapping ``left side of image'' to ``wearer's right'' in a front view). We find that Kimi K2.5 occasionally hallucinates directional descriptions, especially for asymmetric features such as single chest pockets and off-center logos.

\textbf{Human evaluation.}
We manually annotate 1,000 randomly sampled products. Among them, 6.2\% contain \emph{severe hallucinations}---including left--right errors (5.0\%) and other large-scale description errors (1.2\%). Left--right errors account for over 80\% of all severe hallucinations, making them the most critical type to address.

\textbf{Automated directional filtering.}
We employ Qwen3.5-397B-A17B~\cite{Qwen3_5} as a cross-view verifier that checks whether directional descriptions are consistent with visual evidence. On the human-annotated set, this detector achieves 42.9\% precision and 36.0\% recall for left--right hallucinations, confirming it can identify a substantial portion of errors. Products with confirmed errors are removed, filtering \textbf{6,307} products (4.37\%) and leaving \textbf{138,089} products with \textbf{510,877} images.

\subsection{Stage 3: CIR Triplet Construction}
\label{sec:stage3}

We construct CIR triplets $(P_\text{src}, T_\text{mod}, P_\text{tgt})$ through a multi-candidate selection mechanism. For each source product, we retrieve candidate targets via three complementary paths---visual similarity, long-caption similarity, and short-caption similarity---taking the union of top-20 results per path as the candidate pool. We then randomly sample 10 candidates and present them alongside the source to Gemini 3.1 Flash Lite~\cite{Gemini3_1FlashLite}.

The model selects \textbf{up to 2 best candidates}, optimizing for: (i)~differences spanning at least 2 viewpoints; (ii)~clear distinguishability; and (iii)~specific construction details. For each selected pair, the model generates:
\begin{itemize}
  \item A \emph{short modification text} (16--32 words), which reflects the concise, natural-language queries that real users would issue in a search box;
  \item A \emph{long modification text} (64--128 words), which provides detailed descriptions of the differences between source and target.
\end{itemize}

\textbf{Comparison with pairwise annotation.}
Existing CIR datasets---whether human-annotated (e.g., FashionIQ~\cite{FashionIQ}) or model-generated (e.g., FACap~\cite{FACap})---pair each source with a single similar product via a single retrieval modality and generate modification text from this isolated pair. This suffers from two limitations: (1)~without cross-modal verification, categorically mismatched products (e.g., a men's jacket paired with a women's skirt) cannot be filtered out; (2)~observing only one pair, the modification text describes category-level commonalities rather than target-specific distinctions, causing \emph{multi-positive ambiguity}.

Our mechanism addresses both issues by presenting 10 candidates simultaneously: the model can reject incompatible pairs before generating text, and contrast the selected target against 9 competitors, focusing on attributes \emph{uniquely} characteristic of the target.

The short version represents the realistic user scenario and serves as the primary evaluation metric; the long version provides richer training signal that helps the model learn cross-product reasoning transferable to short-text inference. Both versions are sampled with equal probability during training (\S\ref{sec:baseline}).

This stage yields \textbf{220,733} CIR triplets covering \textbf{127,231} products, with 99.83\% involving $\geq$2 viewpoints. The dominant view combinations are back+front (81.6\%) and front+side (12.5\%).

\begin{figure*}[t]
  \centering
  \includegraphics[width=\textwidth]{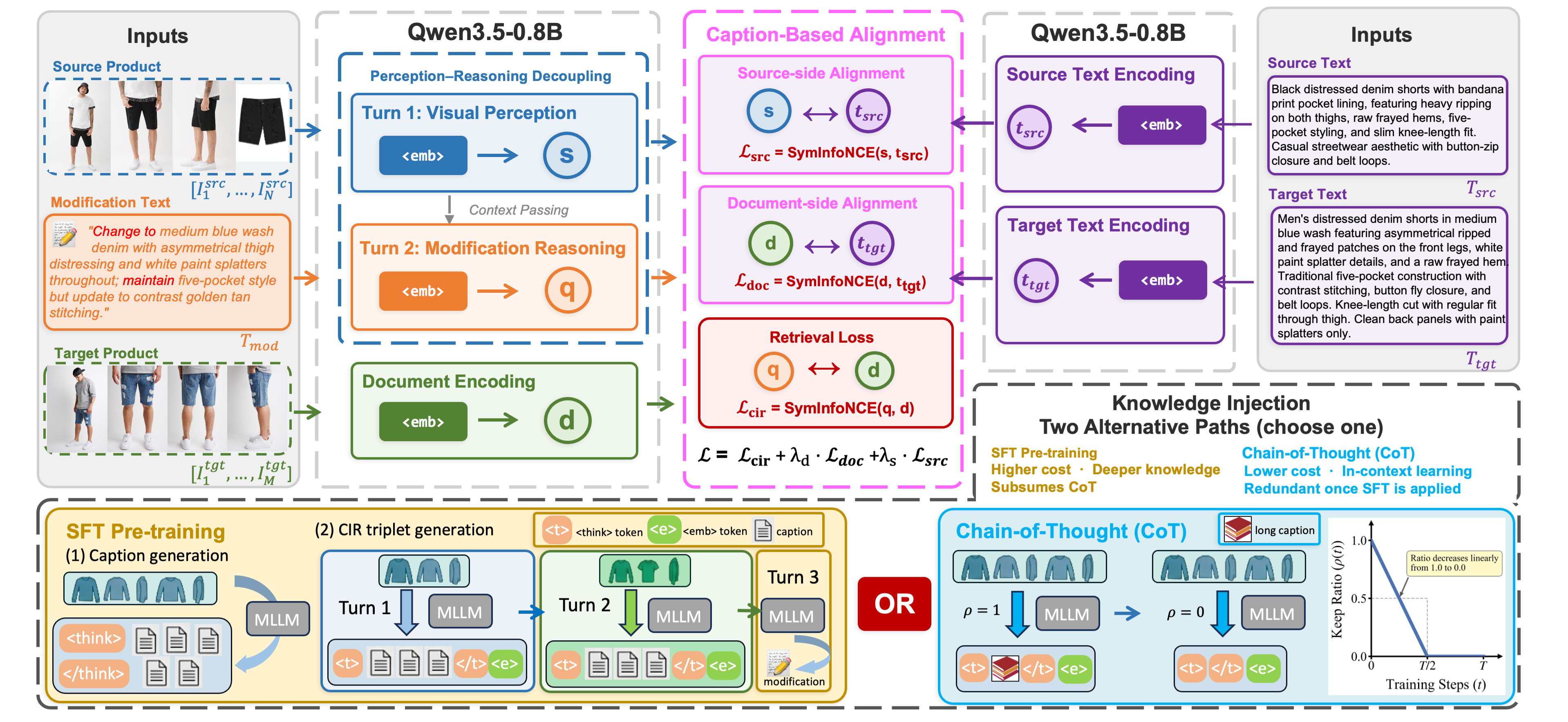}
  \caption{Overview of the ProCIR training architecture. The two-stage dialogue decomposes the query into Turn~1 (visual perception $\rightarrow$ source embedding $\mathbf{s}$) and Turn~2 (modification reasoning $\rightarrow$ query embedding $\mathbf{q}$); Turn~2 inherits the full dialogue context from Turn~1, enabling the model to reason about modifications conditioned on the perceived visual content. The document encoder produces target embedding $\mathbf{d}$ from multi-view target images. Caption-based alignment encodes source and target captions into $\mathbf{t}_\text{src}$ and $\mathbf{t}_\text{tgt}$. The training loss combines $\mathcal{L}_\text{cir}$, $\mathcal{L}_\text{src}$, and $\mathcal{L}_\text{doc}$, all based on SymInfoNCE. An optional SFT stage injects structured product knowledge prior to contrastive training. Two alternative knowledge injection paths are shown at the bottom: SFT pre-training (left) and Chain-of-Thought with progressive removal (right).}
  \label{fig:method}
\end{figure*}

\subsection{Dataset Splits and Statistics}
\label{sec:splits}

We partition products into training and validation sets at the product level with zero overlap. FashionGen uses its native train/val split (42,612 / 5,292 products); DeepFashion and Fashion200K are randomly split (8,856 / 2,791 and 56,960 / 10,720 products, respectively). CIR triplets are constructed independently within each partition---neighbor retrieval, candidate selection, and modification text generation all operate within the same split---yielding 188,015 training and 32,718 validation triplets. Table~\ref{tab:dataset_stats} provides the per-dataset breakdown.

\begin{table}[!h]
  \caption{FashionMV CIR triplet splits.}
  \label{tab:dataset_stats}
  \normalsize
  \begin{tabular}{lrrrr}
    \toprule
    Dataset & Split & Triplets & Products \\
    \midrule
    \multirow{2}{*}{DeepFashion}
    & Train & 16,399 & 8,856 \\
    & Val   & 5,188  & 2,791 \\
    \midrule
    \multirow{2}{*}{Fashion200K}
    & Train & 98,800 & 56,960 \\
    & Val   & 18,499 & 10,720 \\
    \midrule
    \multirow{2}{*}{FashionGen}
    & Train & 72,816 & 42,612 \\
    & Val   & 9,031  & 5,292 \\
    \midrule
    \multirow{2}{*}{\textbf{Total}}
    & Train & \textbf{188,015} & \textbf{108,428} \\
    & Val   & \textbf{32,718}  & \textbf{18,803} \\
    \bottomrule
  \end{tabular}
\end{table}

% ================================================================
\section{Method}
\label{sec:method}

We propose \textbf{ProCIR} (\textbf{Pro}duct-level \textbf{CIR}), a framework built upon a multimodal large language model (MLLM) with three complementary mechanisms: a two-stage dialogue architecture (\textbf{MT}) that decouples visual perception from modification reasoning, caption-based alignment (\textbf{Align}) that injects product-level semantics, and chain-of-thought (\textbf{CoT}) guidance that elicits structured product understanding. The product knowledge that CoT injects can also be internalized more efficiently through supervised fine-tuning (SFT); our ablation reveals that SFT subsumes CoT, rendering it redundant once the base model has been fine-tuned (\S\ref{sec:ablation}). Figure~\ref{fig:method} illustrates the overall training architecture.

\subsection{Task Formulation}
\label{sec:formulation}

In Multi-View CIR, each product $P$ is represented by multi-view images $\mathcal{V}^P = \{I_1^P, \ldots, I_{N_P}^P\}$ where $N_P \in [2, 5]$. Given a source product $P_\text{src}$ and modification text $T_\text{mod}$, the goal is to retrieve the target product $P_\text{tgt}$ from a gallery of product-level embeddings, requiring the model to aggregate all views into a single embedding.

\subsection{Embedding Extraction via Native MLLM}
\label{sec:backbone}

We build upon Qwen3.5-0.8B~\cite{Qwen3_5}, a multimodal LLM that processes interleaved image and text tokens through a unified transformer. To extract fixed-dimensional embeddings, we append a special token \texttt{<emb>} to each assistant response. The hidden state at this position serves as the product-level embedding:
\begin{equation}
  \mathbf{e} = h_{\text{LLM}}[\texttt{<emb>}] \in \mathbb{R}^d
  \label{eq:emb}
\end{equation}
where $d{=}1024$. All multi-view images are fed as visual tokens in a single request; the model's causal self-attention naturally aggregates information across views without explicit pooling. Every assistant turn includes a \texttt{<think>...</think>} block (empty when CoT is not used).

\subsection{Baseline: Single-Turn CIR}
\label{sec:baseline}

The baseline variant processes CIR in a single dialogue turn. On the \textbf{query side}, source images and modification text are concatenated in a single user message:
\begin{equation}
  \text{Query:}\quad \underbrace{[I_1^{\text{src}}, \ldots, I_N^{\text{src}}]}_{\text{multi-view images}} \oplus T_\text{mod} \;\rightarrow\; \mathbf{q}
  \label{eq:query_baseline}
\end{equation}
On the \textbf{document side}, target images are encoded without any text:
\begin{equation}
  \text{Doc:}\quad [I_1^{\text{tgt}}, \ldots, I_M^{\text{tgt}}] \;\rightarrow\; \mathbf{d}
  \label{eq:doc_baseline}
\end{equation}
Both $\mathbf{q}$ and $\mathbf{d}$ are extracted at the \texttt{<emb>} position. The training loss is a symmetric InfoNCE~\cite{InfoNCE}:
\begin{equation}
  \mathcal{L}_\text{cir} = \text{SymInfoNCE}(\mathbf{q}, \mathbf{d}, \tau)
  \label{eq:lcir}
\end{equation}
which averages the query-to-document and document-to-query cross-entropy losses:
\begin{equation}
  \text{SymInfoNCE} = -\frac{1}{2B}\sum_{i=1}^{B}\left[\log\frac{e^{\mathbf{q}_i^\top \mathbf{d}_i/\tau}}{\sum_j e^{\mathbf{q}_i^\top \mathbf{d}_j/\tau}} + \log\frac{e^{\mathbf{d}_i^\top \mathbf{q}_i/\tau}}{\sum_j e^{\mathbf{d}_i^\top \mathbf{q}_j/\tau}}\right]
  \label{eq:syminfonce}
\end{equation}
where $B$ is the batch size and $\tau{=}0.07$ is the temperature. Negatives are drawn from all other in-batch samples, with DDP all-gather across GPUs.

\textbf{Training-time text augmentation.}
The modification text is randomly sampled from the long (64--128 words) or short (16--32 words) version with equal probability. The long version carries richer inter-product difference information, helping the model learn fine-grained reasoning; the short version ensures robustness to concise inference-time queries. Evaluation uses only the short modification text.

\subsection{Two-Stage Dialogue Architecture}
\label{sec:multiturn}

The single-turn baseline entangles visual perception and modification reasoning in one forward pass: the modification text may compete with visual content for attention, and the resulting query embedding mixes visual and textual features, precluding source-side image--text alignment.

We propose a \textbf{two-stage dialogue} design that addresses both issues by decomposing the query into two dialogue turns:
\begin{align}
  &\text{Turn 1 (Perception):} \quad [I_1^{\text{src}}, \ldots, I_N^{\text{src}}] \;\rightarrow\; \mathbf{s} \label{eq:s}\\
  &\text{Turn 2 (Reasoning):} \quad T_\text{mod} \;\rightarrow\; \mathbf{q} \label{eq:q_mt}
\end{align}
The first turn processes only source images, producing a \emph{source embedding} $\mathbf{s}$ at the first \texttt{<emb>}. The second turn receives the modification text and produces the query embedding $\mathbf{q}$ at the second \texttt{<emb>}. Due to causal attention, $\mathbf{q}$ attends to the full context including Turn~1, while $\mathbf{s}$ remains unaffected by Turn~2. This decoupling provides four advantages: (i)~$\mathbf{s}$ is a \emph{pure} visual representation; (ii)~this enables source-side image--text alignment (\S\ref{sec:align}), impossible in the single-turn baseline; (iii)~$\mathbf{s}$ is a zero-cost byproduct directly compatible with gallery embeddings, enabling the source product to be indexed without extra encoding for similar-product retrieval; and (iv)~the Turn~1 key--value states can be pre-computed and cached before any modification text arrives, so that Turn~2 only requires incremental inference over the modification tokens---offering a practical path to low-latency interactive retrieval in industrial deployments.

\subsection{Caption-Based Alignment}
\label{sec:align}

To inject product knowledge into the embedding space, we introduce \textbf{caption-based alignment}. Every product in FashionMV has long and short captions from Stage~1 (\S\ref{sec:stage1}). We encode them through text-only forward passes:
\begin{equation}
  \mathbf{t}_\text{cap} = h_\text{LLM}[T_\text{cap}; \texttt{<emb>}] \in \mathbb{R}^d
  \label{eq:text_emb}
\end{equation}
where $T_\text{cap}$ is randomly selected from the long or short caption with equal probability at each training step.

For the \textbf{document side}, we align the target visual embedding with its caption:
\begin{equation}
  \mathcal{L}_\text{doc} = \text{SymInfoNCE}(\mathbf{d}, \mathbf{t}_\text{tgt}, \tau)
  \label{eq:ldoc}
\end{equation}
For \textbf{two-stage dialogue variants} where $\mathbf{s}$ is a pure visual embedding, we additionally align it with the source caption:
\begin{equation}
  \mathcal{L}_\text{src} = \text{SymInfoNCE}(\mathbf{s}, \mathbf{t}_\text{src}, \tau)
  \label{eq:lsrc}
\end{equation}
Note that in single-turn variants, $\mathbf{q}$ is a mixed image--text representation (containing the modification text), so source-side alignment is not applicable.

\subsection{Chain-of-Thought with Progressive Removal}
\label{sec:cot}

While alignment injects knowledge through an auxiliary loss, we also explore directly injecting textual knowledge into the model's \emph{reasoning process}. We place the product's long caption inside the \texttt{<think>} block: \texttt{assistant: <think>\{long caption (subsampled)\}</think> <emb>}. This lets the model ``read'' a product description before producing the embedding. CoT is applied to both the document side (target) and, in two-stage variants, to the first query turn (source); the second turn does not use CoT.

\textbf{Progressive removal.}
To prevent dependence on captions at inference time, a keep ratio $\rho(t)$ controls the fraction of caption tokens retained:
\begin{equation}
  \rho(t) = \max\!\left(0,\; 1 - \frac{t}{0.5 \cdot T}\right)
  \label{eq:keepratio}
\end{equation}
where $T$ is the total training steps. The ratio decreases linearly from 1.0 to 0.0 over the first half of training; we randomly retain $\lceil \rho \cdot |\text{tokens}| \rceil$ tokens. This transitions the model from full caption availability to inference mode (no caption), enabling it to internalize semantic knowledge.

\subsection{Training Objective}
\label{sec:objective}

The full training objective combines three losses:
\begin{equation}
  \mathcal{L} = \mathcal{L}_\text{cir} + \lambda_d \cdot \mathcal{L}_\text{doc} + \lambda_s \cdot \mathcal{L}_\text{src}
  \label{eq:total}
\end{equation}
where $\lambda_d = \lambda_s = 0.25$. The alignment losses $\mathcal{L}_\text{doc}$ and $\mathcal{L}_\text{src}$ are activated only in alignment variants; $\mathcal{L}_\text{src}$ is further restricted to two-stage dialogue variants where $\mathbf{s}$ exists. Table~\ref{tab:cir_main} summarizes all eight ablation variants and their active components.

\subsection{Supervised Fine-Tuning}
\label{sec:sft}

Before contrastive training, we optionally perform supervised fine-tuning (SFT) on Qwen3.5-0.8B to inject structured product understanding. The SFT stage uses two data types from the FashionMV training set:

\textbf{(1)~Caption generation.}
A single-turn dialogue where the user provides all multi-view images and the assistant generates per-image captions inside a \texttt{<think>} block followed by a JSON object with long and short captions. This teaches the model to identify views, extract attributes, and synthesize cross-view information.

\textbf{(2)~CIR triplet generation.}
A three-turn dialogue from a CIR triplet. Turn~1: the user provides source images; the assistant generates the source caption in a \texttt{<think>} block followed by \texttt{<emb>}. Turn~2: same format for the target. Turn~3: the assistant generates modification text in a JSON object. This familiarizes the model with multi-turn product reasoning and embedding token semantics.

We randomly sample 20\% of the available data from both types and train the full model for one epoch. The training objective is the standard autoregressive language modeling loss over assistant tokens:
\begin{equation}
  \mathcal{L}_\text{sft} = -\sum_{t \in \mathcal{A}} \log\, p_\theta(x_t \mid x_{<t})
  \label{eq:sft}
\end{equation}
where $\mathcal{A}$ denotes the set of assistant token positions and $x_{<t}$ is the preceding context. The resulting SFT checkpoint serves as an alternative initialization for all eight ablation variants (\S\ref{sec:ablation}), enabling us to disentangle the contribution of knowledge injection through SFT from the in-context injection provided by CoT.

% ================================================================
\section{Experiments}
\label{sec:experiments}

\setcounter{dbltopnumber}{1}
\renewcommand{\dblfloatpagefraction}{0.35}
\begin{table*}[t]
  \caption{Comparison with existing models on multi-view CIR under three encoding strategies. \textbf{Joint}: all views encoded jointly; \textbf{MeanPool}: per-view embeddings averaged; \textbf{MaxSim}: per-view retrieval with max-score selection. ``--'': unsupported or unavailable. \textbf{Bold}: best per column; \underline{underline}: second best.}
  \label{tab:baselines}
  \large
  \centering
  \setlength{\tabcolsep}{1.5pt}
  \begin{tabular}{l r l cc cc cc | cc c}
    \toprule
    & & & \multicolumn{2}{c}{\textbf{DeepFashion}} & \multicolumn{2}{c}{\textbf{F200K}} & \multicolumn{2}{c|}{\textbf{FashionGen}} & \multicolumn{3}{c}{\textbf{Average}} \\
    Model & Params & Encoding & R@5 & R@10 & R@5 & R@10 & R@5 & R@10 & R@5 & R@10 & Avg \\
    \midrule
    \multirow{3}{*}{CLIP4CIR} & \multirow{3}{*}{0.25B}
      & Joint & -- & -- & -- & -- & -- & -- & -- & -- & -- \\
    & & MeanPool & 28.0 & 39.3 & 13.3 & 19.3 & 16.2 & 23.6 & 19.2 & 27.4 & 23.3 \\
    & & MaxSim & 25.7 & 36.6 & 11.9 & 17.7 & 17.1 & 25.0 & 18.2 & 26.4 & 22.3 \\
    \midrule
    \multirow{3}{*}{SPRC} & \multirow{3}{*}{1.2B}
      & Joint & -- & -- & -- & -- & -- & -- & -- & -- & -- \\
    & & MeanPool & 53.4 & 65.1 & 32.9 & 42.3 & 38.5 & 48.7 & 41.6 & 52.0 & 46.8 \\
    & & MaxSim & 55.8 & 67.7 & 34.4 & 43.7 & 42.7 & 53.0 & 44.3 & 54.8 & 49.6 \\
    \midrule
    \multirow{3}{*}{Qwen3-V-2B} & \multirow{3}{*}{2B}
      & Joint & 75.7 & 86.4 & 61.1 & 72.3 & 63.0 & 74.1 & 66.6 & 77.6 & 72.1 \\
    & & MeanPool & 76.8 & 87.5 & 60.3 & 72.0 & 56.3 & 68.3 & 64.5 & 75.9 & 70.2 \\
    & & MaxSim & 73.9 & 85.4 & 57.3 & 70.3 & 58.2 & 69.9 & 63.1 & 75.2 & 69.2 \\
    \midrule
    \multirow{3}{*}{RzenEmbed} & \multirow{3}{*}{8B}
      & Joint & 24.3 & 32.0 & 12.7 & 17.0 & 18.9 & 25.6 & 18.6 & 24.9 & 21.8 \\
    & & MeanPool & 47.5 & 58.0 & 28.0 & 36.5 & 32.5 & 42.2 & 36.0 & 45.6 & 40.8 \\
    & & MaxSim & 29.6 & 38.8 & 16.0 & 21.8 & 22.1 & 28.8 & 22.6 & 29.8 & 26.2 \\
    \midrule
    \multirow{3}{*}{Qwen3-V-8B} & \multirow{3}{*}{8B}
      & Joint & \underline{87.4} & \underline{93.2} & \underline{73.8} & \underline{82.1} & \underline{74.7} & \underline{83.5} & \underline{78.6} & \underline{86.3} & \underline{82.5} \\
    & & MeanPool & 85.3 & 92.6 & 68.1 & 77.8 & 67.9 & 78.4 & 73.8 & 82.9 & 78.4 \\
    & & MaxSim & 85.6 & 92.3 & 68.8 & 78.2 & 70.4 & 79.6 & 74.9 & 83.4 & 79.2 \\
    \midrule
    \multirow{3}{*}{Doubao-E-V} & \multirow{3}{*}{--}
      & Joint & 67.8 & 84.0 & 50.1 & 64.0 & 56.1 & 70.4 & 58.0 & 72.8 & 65.4 \\
    & & MeanPool & 82.4 & 90.5 & 61.2 & 71.8 & 61.4 & 72.6 & 68.3 & 78.3 & 73.3 \\
    & & MaxSim & 82.3 & 90.8 & 62.2 & 72.6 & 67.2 & 77.1 & 70.6 & 80.2 & 75.4 \\
    \midrule
    Ours & 0.8B & Joint & \textbf{89.2} & \textbf{94.9} & \textbf{77.6} & \textbf{86.6} & \textbf{75.0} & \textbf{85.3} & \textbf{80.6} & \textbf{88.9} & \textbf{84.8} \\
    \bottomrule
  \end{tabular}
\end{table*}
\setcounter{dbltopnumber}{4}
\renewcommand{\dblfloatpagefraction}{0.7}

\subsection{Experimental Setup}
\label{sec:setup}

\textbf{Backbone.}
We use Qwen3.5-0.8B~\cite{Qwen3_5} as the base MLLM. It has 24 transformer layers with a hidden dimension of 1024. We add one special token \texttt{<emb>} and resize the embedding layer accordingly.

\textbf{Training.}
All variants are trained for 1 epoch on the full training set (188K CIR triplets) with AdamW (lr=$10^{-5}$, weight decay=$0.01$), cosine learning rate schedule without warmup, gradient clipping at 1.0, and bfloat16 mixed precision. The batch size is 16 per GPU on 10$\times$RTX 3090 GPUs, yielding an effective batch size of 160 and a total of 1,175 training steps. We cap the maximum pixel count per image at $262{,}144$ ($= 512{\times}512$); images exceeding this limit are proportionally downscaled while preserving the aspect ratio. Each product contains up to 5 views. The temperature $\tau{=}0.07$ and alignment loss weights $\lambda_d {=} \lambda_s {=} 0.25$.

\textbf{Evaluation.}
We evaluate on three fashion datasets independently: \textbf{DeepFashion} (DF)~\cite{DeepFashion}, \textbf{Fashion200K} (F200K)~\cite{Fashion200K}, and \textbf{FashionGen-val} (FG)~\cite{FashionGen}. Each dataset has its own document gallery constructed from its validation products. We evaluate using \emph{short modification text only}, as it reflects the realistic user query scenario (\S\ref{sec:stage3}), and report Recall@5 and Recall@10.

\textbf{Model initialization.}
Each of the 8 ablation variants is trained from two initializations: (i)~\textbf{Pretrained}: the public Qwen3.5-0.8B checkpoint; (ii)~\textbf{SFT}: a checkpoint obtained by supervised fine-tuning on caption generation and CIR triplet generation tasks (\S\ref{sec:sft}). This yields 16 configurations in total, isolating the effect of each mechanism and the base model quality.

\subsection{Comparison with Existing Models}
\label{sec:baselines}

Since no prior method directly addresses multi-view product-level CIR, we adapt representative embedding models to our task using three multi-view encoding strategies. Let $\{\mathbf{e}_1, \ldots, \mathbf{e}_N\}$ denote the per-view embeddings of a product with $N$ views.
\begin{itemize}[nosep,leftmargin=*]
  \item \textbf{Joint}---all $N$ views are fed into the model simultaneously, producing a single product-level embedding $\mathbf{e}_\text{prod}$ directly. This requires the model to natively accept multi-image input. The retrieval score between query product $Q$ and document product $D$ is:
  \begin{equation}
    s_\text{Joint}(Q, D) = \cos(\mathbf{e}_\text{prod}^Q,\; \mathbf{e}_\text{prod}^D)
  \end{equation}
  \item \textbf{MeanPool}---each view is encoded independently, and the $N$ embeddings are averaged into a single product-level representation:
  \begin{equation}
    s_\text{MeanPool}(Q, D) = \cos\!\left(\frac{1}{N_Q}\sum_{i=1}^{N_Q}\mathbf{e}_i^Q,\;\; \frac{1}{N_D}\sum_{j=1}^{N_D}\mathbf{e}_j^D\right)
  \end{equation}
  \item \textbf{MaxSim}---each view is encoded independently, yielding $N_Q$ query embeddings and $N_D$ document embeddings. The retrieval score is the maximum pairwise cosine similarity across all view combinations:
  \begin{equation}
    s_\text{MaxSim}(Q, D) = \max_{i \in [N_Q],\, j \in [N_D]} \cos(\mathbf{e}_i^Q,\; \mathbf{e}_j^D)
  \end{equation}
\end{itemize}
Traditional vision--language models (CLIP~\cite{CLIP}, BLIP~\cite{BLIP}) lack composed-image retrieval capability entirely.
CLIP4CIR~\cite{CLIP4CIR} and SPRC~\cite{SPRC} support CIR but only via single-image queries, limiting them to MeanPool and MaxSim.
Open-source VLM embedding models (Qwen3-VL-Embedding~\cite{Qwen3VLEmb} at 2B and 8B, abbreviated as Qwen3-V-2B and Qwen3-V-8B; RzenEmbed~\cite{RzenEmbed}) and the closed-source Doubao-Embedding-Vision (Doubao-E-V) natively accept multi-image input and thus support all three strategies.

Table~\ref{tab:baselines} reports CIR results on the validation set under all three encoding strategies.
Our 0.8B model outperforms all baselines on every dataset, including the 10$\times$ larger Qwen3-V-8B.
Under the same Joint encoding, our model surpasses Qwen3-V-8B by +1.8/+3.8/+0.3pp R@5 on DF/F200K/FG while using only one-tenth the parameters.
Single-image CIR methods (CLIP4CIR, SPRC) perform substantially worse, confirming that architectures designed for single-image queries are fundamentally inadequate for multi-view product retrieval. Among them, SPRC---which leverages sentence-level prompts from BLIP-2---considerably outperforms CLIP4CIR, yet still lags far behind VLM embedding models that can natively process multiple images.

Comparing the two per-view strategies, MeanPool outperforms MaxSim on the majority of models (CLIP4CIR, Qwen3-V-2B, RzenEmbed), with MaxSim taking the lead only on stronger models (SPRC, Qwen3-V-8B, Doubao-E-V). Each strategy has inherent limitations: MeanPool fuses information across views but the averaging operation may dilute view-specific details; MaxSim preserves the most discriminative single-view match but fails when the modification text involves attributes spread across multiple views, since the best-matching view for one attribute may not match another.
The more revealing contrast, however, is between Joint and per-view strategies.
Among models with strong multi-image comprehension (e.g., Qwen3-V-8B), Joint encoding surpasses both MeanPool and MaxSim, indicating that explicit cross-view reasoning is more effective than post-hoc aggregation. In contrast, weaker models show the opposite pattern: RzenEmbed's Joint encoding (18.6 Avg R@5) drops substantially below its MeanPool result (36.0), losing nearly half its accuracy when processing all views simultaneously. Doubao-E-V shows a similar trend, with Joint (58.0 Avg R@5) lagging behind MeanPool (68.3) and MaxSim (70.6) by a considerable margin. For such models, independently encoding each view and aggregating afterwards consistently yields better results.

\subsection{Ablation Results}
\label{sec:ablation}

Table~\ref{tab:cir_main} presents the complete ablation results on the validation set. All 16 configurations (8 variants $\times$ 2 initializations) are reported. We organize the analysis around three key findings.

\begin{table*}[t]
  \caption{Ablation study on the validation set. MT = two-stage dialogue, Align = caption-based alignment, CoT = chain-of-thought. Cost = relative wall-clock time with absolute duration in parentheses; 1.00$\times$ corresponds to the fastest variant; SFT rows include SFT pre-training overhead (4h\,29m). \textbf{Bold}: best in each section; \underline{underline}: second best.}
  \label{tab:cir_main}
  \large
  \centering
  \setlength{\tabcolsep}{1.2pt}
  \begin{tabular}{cccc|cc|cc|cc|ccc|l}
    \toprule
    & & & & \multicolumn{2}{c|}{\textbf{DeepFashion}} & \multicolumn{2}{c|}{\textbf{F200K}} & \multicolumn{2}{c|}{\textbf{FashionGen}} & \multicolumn{3}{c|}{\textbf{Average}} & \\
    MT & Align & CoT & SFT & R@5 & R@10 & R@5 & R@10 & R@5 & R@10 & R@5 & R@10 & Avg & \multicolumn{1}{c}{Cost} \\
    \midrule
    \multicolumn{14}{l}{\emph{Pretrained base}} \\
     & & & & 77.0 & 87.8 & 61.4 & 73.6 & 59.7 & 72.5 & 66.0 & 78.0 & 72.0 & 1.01$\times$\,(7h25m) \\
     & $\surd$ & & & 78.7 & 88.7 & 61.5 & \underline{74.1} & 62.6 & 74.5 & 67.6 & 79.1 & 73.4 & 1.12$\times$\,(8h13m) \\
     & & $\surd$ & & 77.9 & 87.9 & 60.5 & 71.6 & 56.3 & 68.6 & 64.9 & 76.0 & 70.5 & 1.16$\times$\,(8h31m) \\
     & $\surd$ & $\surd$ & & 77.5 & 87.4 & \underline{61.8} & 73.1 & 57.4 & 69.6 & 65.6 & 76.7 & 71.2 & 1.27$\times$\,(9h20m) \\
    $\surd$ & & & & 77.8 & 88.3 & 60.7 & 73.0 & 59.4 & 72.3 & 66.0 & 77.9 & 72.0 & 1.00$\times$\,(7h20m) \\
    $\surd$ & $\surd$ & & & \underline{79.5} & \underline{89.0} & \underline{61.8} & 73.8 & \underline{62.7} & \underline{74.9} & \underline{68.0} & \underline{79.2} & \underline{73.6} & 1.21$\times$\,(8h53m) \\
    $\surd$ & & $\surd$ & & 72.3 & 84.2 & 59.1 & 71.2 & 54.4 & 67.1 & 61.9 & 74.2 & 68.1 & 1.15$\times$\,(8h27m) \\
    $\surd$ & $\surd$ & $\surd$ & & \textbf{84.1} & \textbf{91.5} & \textbf{68.6} & \textbf{79.0} & \textbf{67.7} & \textbf{77.8} & \textbf{73.5} & \textbf{82.8} & \textbf{78.2} & 1.37$\times$\,(10h04m) \\
    \midrule
    \multicolumn{14}{l}{\emph{SFT base}} \\
     & & & $\surd$ & 84.5 & 92.7 & 69.5 & 81.7 & 65.5 & 78.4 & 73.2 & 84.3 & 78.8 & 1.62$\times$\,(11h54m) \\
     & $\surd$ & & $\surd$ & \underline{88.2} & \underline{94.2} & \underline{76.8} & \underline{86.1} & \underline{74.5} & \underline{84.5} & \underline{79.8} & \underline{88.3} & \underline{84.1} & 1.73$\times$\,(12h42m) \\
     & & $\surd$ & $\surd$ & 83.4 & 91.1 & 68.1 & 78.2 & 62.4 & 73.2 & 71.3 & 80.8 & 76.1 & 1.77$\times$\,(13h00m) \\
     & $\surd$ & $\surd$ & $\surd$ & 83.7 & 91.3 & 69.1 & 78.9 & 63.8 & 74.7 & 72.2 & 81.6 & 76.9 & 1.88$\times$\,(13h49m) \\
    $\surd$ & & & $\surd$ & 82.7 & 92.3 & 68.2 & 81.0 & 65.6 & 78.7 & 72.2 & 84.0 & 78.1 & 1.61$\times$\,(11h49m) \\
    $\surd$ & $\surd$ & & $\surd$ & \textbf{89.2} & \textbf{94.9} & \textbf{77.6} & \textbf{86.6} & \textbf{75.0} & \textbf{85.3} & \textbf{80.6} & \textbf{88.9} & \textbf{84.8} & 1.82$\times$\,(13h22m) \\
    $\surd$ & & $\surd$ & $\surd$ & 83.4 & 91.4 & 70.1 & 80.0 & 67.0 & 77.6 & 73.5 & 83.0 & 78.3 & 1.76$\times$\,(12h56m) \\
    $\surd$ & $\surd$ & $\surd$ & $\surd$ & \underline{88.2} & 94.0 & 75.6 & 84.3 & 73.3 & 82.4 & 79.0 & 86.9 & 83.0 & 1.98$\times$\,(14h33m) \\
    \bottomrule
  \end{tabular}
\end{table*}
\setcounter{dbltopnumber}{4}
\renewcommand{\dblfloatpagefraction}{0.7}

\textbf{Finding 1: MT + Align + CoT maximizes knowledge injection without SFT.}
Among all Pretrained-base variants, the full combination of all three mechanisms achieves the best performance across every dataset, substantially outperforming variants that lack any single mechanism.
The three mechanisms contribute complementary knowledge injection: MT decouples perception from reasoning and enables source-side alignment; Align anchors the visual--textual embedding space; CoT injects product captions into the reasoning process, bridging visual features and semantic understanding.
When all three are active, the model receives the richest supervision signal from the training data.

\textbf{Finding 2: Alignment is the single most critical mechanism.}
Across both initializations, alignment consistently provides the largest individual gain.
The effect is especially dramatic in the two-stage setting: under the Pretrained base, adding alignment to the MT+CoT variant yields +11.8/+9.5/+13.3pp R@5 on DF/F200K/FG.
Without alignment, the two-stage architecture actually underperforms the single-turn baseline (e.g., R@5 72.3 vs.\ 77.0 on DF), demonstrating that the decoupled architecture \emph{requires} explicit cross-modal anchoring to be effective.
In contrast, CoT without alignment shows limited or even negative impact.

\textbf{Finding 3: CoT becomes unnecessary after SFT.}
Both CoT and SFT inject product understanding into the model---CoT through in-context caption injection during training, SFT through prior multimodal fine-tuning.
When neither is present, adding CoT to the MT+Align variant improves R@5 by +4.6/+6.8/+5.0pp (Pretrained base).
However, after SFT, CoT not only becomes unnecessary but can even hurt: adding CoT to the MT+Align+SFT variant yields $-$1.0/$-$2.0/$-$1.7pp R@5 on DF/F200K/FG.
As a result, the best overall configuration is \textbf{MT+Align+SFT} (without CoT), which achieves the highest R@5 across all three datasets in the SFT group.
This reveals that SFT and CoT are alternative knowledge injection paths serving overlapping functions: once the base model has internalized product semantics through SFT, the additional in-context caption knowledge from CoT provides diminishing returns and may even introduce noise.

\subsection{Analysis}
\label{sec:analysis}

\textbf{Dataset difficulty.}
The three evaluation datasets pose distinct challenges.
\textbf{DeepFashion} is the easiest: it has the smallest gallery (2,791 products) and high image resolution ($>$512$\times$512), providing both a reduced search space and rich visual detail.
\textbf{Fashion200K} raises the difficulty through the largest gallery (10,720 products); although its images are also high-resolution, its average views per product is the lowest (${\sim}$3.3), limiting visual coverage.
\textbf{FashionGen-val} is consistently the hardest (lowest recall across all variants) due to its low resolution (256$\times$256), despite having the most views (${\sim}$4.1).
These complementary characteristics allow the three benchmarks to evaluate different aspects of model capability.
The improvement from alignment is largest on FashionGen-val (e.g., +13.3pp R@5 when adding Align to MT+CoT), suggesting that caption-based alignment is especially beneficial in challenging retrieval scenarios.

\textbf{SFT universally improves all variants.}
Every variant benefits from SFT initialization, with an average improvement of +8.5pp R@5 across all 8 variants and 3 datasets. The gain is particularly large for simpler variants (e.g., the baseline improves by +7.5/+8.1/+5.8pp R@5 on DF/F200K/FG).
This suggests that, for pre-trained MLLMs, injecting domain-specific knowledge through supervised fine-tuning is more efficient than contrastive learning alone.
However, this efficiency comes at a cost: SFT requires rich multi-granularity supervision---per-image captions, product-level captions, and modification texts---essentially all intermediate byproducts of the dataset construction pipeline. Such data is difficult to obtain without an automated generation pipeline, and SFT introduces additional training overhead.
Finally, an SFT-trained model alone has no embedding capability; contrastive training remains indispensable for projecting internalized knowledge into the retrieval embedding space.

\textbf{Training cost.}
The fastest Pretrained variant (MT-only) completes in ${\sim}$7h\,20m on 10$\times$RTX\,3090 GPUs, while the most expensive configuration (MT+Align+CoT+SFT) requires ${\sim}$14h\,33m.
The SFT pre-training stage alone takes 4h\,29m; this overhead is included in every SFT variant's reported time.
The cost differences among mechanisms stem from the number of forward passes per training step.
Alignment adds a text-only forward pass to encode product captions; without the two-stage dialogue, alignment operates on the document side only (one extra forward pass), whereas with it, both source-side and document-side alignment are active (two extra forward passes), leading to a larger time increment.
CoT increases token sequence lengths due to the injected caption text, which proportionally increases the compute per forward pass.
These overheads are additive, so the full combination of all mechanisms incurs the highest wall-clock time.

% ================================================================
\section{Conclusion}
\label{sec:conclusion}

We have identified \emph{View Incompleteness} as a fundamental limitation of existing CIR methods and datasets, and addressed it by formally defining the Multi-View CIR task, constructing the FashionMV dataset, and proposing ProCIR, a modeling framework with three complementary mechanisms.

\textbf{FashionMV} is the first large-scale multi-view fashion dataset for product-level CIR, containing 127K products, 472K images, and 220K+ triplets with dual-granularity captions and modification texts, constructed through a fully automated three-stage pipeline.

Our \textbf{ProCIR} framework demonstrates that three complementary mechanisms---two-stage dialogue, caption-based alignment, and chain-of-thought guidance---each contribute to transferring a pre-trained MLLM's generative capabilities toward product-level composed retrieval.
Systematic ablation across 16 configurations (8 variants $\times$ 2 initializations) reveals that alignment is the single most critical mechanism, and that CoT and SFT serve as partially redundant knowledge injection paths: when the base model has already internalized product semantics through SFT, the marginal benefit of CoT diminishes.
The two-stage dialogue architecture is a prerequisite for effective alignment, as it produces a pure visual embedding that enables source-side image--text alignment impossible in single-turn designs.

\textbf{Limitations.}
Our current experiments use a 0.8B-parameter model due to computational constraints; scaling to larger MLLMs may yield further improvements. The dataset is limited to fashion products; extending to other multi-view e-commerce domains (furniture, electronics) is a natural next step.

\bibliographystyle{ACM-Reference-Format}
\bibliography{FashionMV}

\clearpage
\appendix

\twocolumn[{%
\begin{center}
  {\fontsize{22}{26}\bfseries\selectfont Supplementary Material}\\[8pt]
  {\Large\bfseries FashionMV: Product-Level Composed Image Retrieval\\[2pt]with Multi-View Fashion Data}
\end{center}
\vspace{4pt}
}]

\begin{quote}
\small
\textbf{Abstract.}
This document provides supplementary material for the main paper \emph{FashionMV: Product-Level Composed Image Retrieval with Multi-View Fashion Data}.
It is organised into three parts.

\textbf{Part~I — Dataset Construction} (\S\ref{sec:prompts}--\S\ref{sec:dataset_statistics}) details every step of the three-stage data pipeline.
Section~\ref{sec:prompts} reproduces the complete LLM prompts for multi-view caption generation, directional hallucination filtering, and CIR triplet construction.
Section~\ref{sec:dataset_details} reports per-dataset hallucination detection rates, caption word-count statistics, and CIR triplet view-combination distributions.
Section~\ref{sec:dataset_statistics} presents quantitative dataset statistics including multi-view image distributions and text-length histograms for captions and modification texts.

\textbf{Part~II — Model and Experiments} (\S\ref{sec:implementation}--\S\ref{sec:additional_results}) provides full implementation and experimental details.
Section~\ref{sec:implementation} covers training hyperparameters, evaluation protocol, and SFT pre-training data formats and statistics.
Section~\ref{sec:additional_results} reports complete Image-to-Text (I2T) and Text-to-Image (T2I) retrieval results for all 16 model configurations.

\textbf{Part~III — Visual Galleries} (\S\ref{sec:sample_gallery}--\S\ref{sec:retrieval_gallery}) provides visual illustrations of the dataset and model outputs.
Section~\ref{sec:sample_gallery} shows a curated gallery of 24 product samples with their multi-view images.
Section~\ref{sec:triplet_gallery} shows selected CIR triplets (source product $\to$ modification text $\to$ target product).
Section~\ref{sec:retrieval_gallery} presents 9 retrieval case studies comparing our model against four baselines.
\end{quote}

% ================================================================

% ================================================================
% Part I: Dataset Construction
% ================================================================
\section{Dataset Construction Prompts}
\label{sec:prompts}

This section presents the complete prompts used in the three-stage FashionMV construction pipeline. All prompts are reproduced verbatim.

% ---------------------------------------------------------------
\subsection{Stage 1: Multi-View Caption Generation}
\label{sec:prompt_caption}

\textbf{Model:} kimi-k2.5 \quad \textbf{Max tokens:} 8192 \quad \textbf{Temperature:} 1.0 \quad \textbf{Thinking:} enabled

Each product's multi-view images (up to 5) are provided as base64-inlined content with text labels (\texttt{[Image~1]}, \texttt{[Image~2]}, etc.). The system prompt is:

\begin{lstlisting}
# Fashion Product Caption Generation

You are a professional fashion product analyst with expertise in detailed
garment description. You will be given multiple images of the same fashion
product from different angles/views.

**Image Numbering**: Each image is preceded by a text label indicating its
number (e.g., "[Image 1]", "[Image 2]", etc.). When generating captions,
you MUST reference these image numbers.

## Your Task

Analyze all provided images to extract the complete characteristics of
this fashion product. Generate clear, accurate descriptions for each
image and synthesize into comprehensive product captions.

**IMPORTANT**:
- When describing the final product, do NOT mention which image shows what
- Treat all images as different perspectives of ONE product and describe
  the PRODUCT itself

Provide:

1. **is_clothing**: Whether this is a clothing item (wearable garments:
   shirts, blouses, t-shirts, sweaters, hoodies, jackets, coats, blazers,
   vests, pants, trousers, jeans, shorts, skirts, dresses, jumpsuits,
   rompers, suits, underwear, sleepwear. EXCLUDED: jewelry, watches,
   bags, purses, shoes, boots, sandals, hats, caps, scarves, belts,
   glasses, sunglasses, gloves, socks)

2. **image_captions**: An array of description strings for each input
   image (in the same order). Each 50-200 words and MUST start with
   the image number (e.g., "[Image 1] I can see...").

3. **long_caption**: Comprehensive 200-400 word product description
   synthesizing all information from all views. Describe the PRODUCT
   itself, not which image shows what.

4. **short_caption**: Concise ~50 word summary highlighting garment type,
   key style, main color, and distinctive features.

## Description Guidelines

Each image description should follow this strict output order:

### Step 1: Determine View Type and Left/Right Orientation

**Important: Images are NOT mirrored - directly captured by camera.**

1. Determine whether FRONT VIEW, BACK VIEW, or SIDE VIEW.
2. Apply left/right rules:
   - FRONT VIEW: left of image = wearer's RIGHT; right = wearer's LEFT
   - BACK VIEW: left of image = wearer's LEFT; right = wearer's RIGHT
   - SIDE VIEW: carefully observe which side of the wearer is shown

### Step 2: Describe Garment Details
- Overall: type, style, color, silhouette, fit
- Details: buttons, pockets, slits/vents, pleats, ruffles (exact count,
  position, and color)
- Positions: from wearer's perspective
- Material: apparent fabric type and texture

### Step 3: Describing Asymmetrical Features

**CRITICAL**: If garment has left-right asymmetry, describe in detail.
For any asymmetrical feature, mention BOTH:
1. Which side of the image the feature appears on
2. Which side of the wearer it corresponds to

## Response Format
JSON format ONLY:
{
    "is_clothing": true/false,
    "image_captions": ["[Image 1] ...", "[Image 2] ...", ...],
    "long_caption": "...",
    "short_caption": "..."
}
\end{lstlisting}

% ---------------------------------------------------------------
\subsection{Stage 2: Directional Hallucination Filtering}
\label{sec:prompt_hallucination}

\textbf{Model:} qwen3.5-397b-a17b \quad \textbf{Max tokens:} 16384 \quad \textbf{Temperature:} 1.0 \quad \textbf{Thinking:} enabled

For each product, all images are horizontally stitched into a panoramic composite. Each sub-image's bounding box (normalized 0--1000 coordinates) is provided. Five products are batched per request.

\begin{lstlisting}
# Fashion Caption Left/Right Error Detector
  (Per-Image Reasoning with Bounding Box)

You are a meticulous fashion product analyst. Your ONLY task is to detect
**left/right direction errors** in fashion product captions.

**IMPORTANT: You should ONLY analyze FRONT VIEW and BACK VIEW images.
Completely SKIP and IGNORE any SIDE VIEW images.**

You will be given a list of fashion products (usually 5). Each includes:
- A stitched panoramic image (multiple views combined horizontally,
  WITHOUT text labels)
- Bounding box of each sub-image (0-1000 normalized coordinates)
- Per-image captions, long caption, and short caption

## Your ONLY Task

1. ONLY check FRONT VIEW and BACK VIEW for left/right direction errors.
2. Completely IGNORE SIDE VIEW images.
3. Ignore ALL other types of issues (missing features, counting errors,
   color errors, etc.).

Check whether captions correctly describe which side of the garment a
feature is on (patches, logos, pockets, labels, zippers, buttons,
asymmetric designs, etc.).

## Bounding Box Requirement (CRITICAL)

For EVERY asymmetric feature, provide its bounding box in the stitched
panoramic image: [x_min, y_min, x_max, y_max] as integers 0-1000.

## Left/Right Mapping Rules

### FRONT VIEW (person FACING camera)
- Left of image = wearer's RIGHT side
- Right of image = wearer's LEFT side

### BACK VIEW (person's BACK facing camera)
- Left of image = wearer's LEFT side
- Right of image = wearer's RIGHT side

## Response Format

JSON array with one element per product:
[
  {
    "product_index": 1,
    "product_id": "<product_id>",
    "caption_analyses": [
      {
        "caption_name": "image_caption_1",
        "bounding_boxes": [
          {
            "element": "<description>",
            "bbox": ["x_min", "y_min", "x_max", "y_max"],
            "position_in_image": "<which side>"
          }
        ],
        "reasoning": "<100-400 words>",
        "has_error": false
      }
    ]
  }
]

## CRITICAL Requirements
- Output ONLY a JSON array. No extra text.
- has_error = true ONLY for confirmed left/right errors.
- Do NOT force-find errors.
\end{lstlisting}

% ---------------------------------------------------------------
\subsection{Stage 3: CIR Triplet Construction}
\label{sec:prompt_cir}

\textbf{Model:} gemini-3.1-flash-lite \quad \textbf{Max tokens:} 4096 \quad \textbf{Temperature:} 1.0

For each source product, 10 same-category candidates (from the union of visual, long-caption, and short-caption top-20 neighbors) are provided with their stitched images and captions.

\begin{lstlisting}
You are an expert fashion analyst specializing in Composed Image
Retrieval (CIR).

You will be given:
1. A source product with its composite image (multiple views stitched
   horizontally), per-image descriptions, and long caption
2. Multiple candidate products (each with composite image, per-image
   descriptions, and long caption), identified by IDs like
   [Product 1], [Product 2], etc.

Your task:
1. Examine the source and ALL candidates from every available view
2. Select the 2 BEST candidates for high-quality modification text

## What Makes a Good Selection

A good (source, modification_text, target) triplet:
- Modification text describes specific, concrete differences
- Differences distributed across at least 2 views
- Differences clearly identifiable and unambiguous
- Differences involve garment construction details (stitching, seams,
  pockets, collars, hems, closures, panels)

## Selection Criteria (ALL must be met):

### SAME CATEGORY & SAME GENDER (Mandatory)
- Exact same sub-category as source
- Same gender as source

### MULTI-VIEW REQUIREMENT (Core)
- Differences MUST span at least 2 different views
- Each view MUST contribute distinct information

### CLEAR DISTINGUISHABILITY (Critical)
- Target MUST be clearly different from source
- Do NOT select candidates too similar to source

### DETAIL-ORIENTED (Quality Amplifier)
- Prefer specific construction details: stitching patterns, seam
  types, pocket configurations, collar styles, hem treatments

## Output Format

{
  "selections": [
    {
      "target_id": <int>,
      "same_category_check": "<explain>",
      "views_involved": ["front", "back"],
      "difficulty_reasoning": "<explain per-view contributions>",
      "modification_text_long": "Detailed (64-128 words). MUST
        reference at least 2 named views.",
      "modification_text_short": "Concise (16-32 words) summary
        with key changes from each view."
    },
    { ... }
  ]
}

Return EXACTLY 2 selections.
\end{lstlisting}

% ================================================================
\section{Dataset Construction Details}
\label{sec:dataset_details}

This section provides detailed statistics collected during the three-stage FashionMV construction pipeline, including hallucination detection outcomes, caption word-count distributions, and CIR triplet structural breakdowns.

\subsection{Per-Dataset Hallucination Detection}

Table~\ref{tab:hallucination} reports the directional hallucination detection results for each dataset. Fashion200K has the highest error rate (4.68\%), likely because its images have more diverse and challenging compositions.

\begin{table}[t]
  \caption{Directional hallucination detection results per dataset.}
  \label{tab:hallucination}
  \small
  \begin{tabular}{lrrrr}
    \toprule
    Dataset & Checked & Errors & Retained & Error Rate \\
    \midrule
    DeepFashion & 12,711 & 487 & 12,224 & 3.83\% \\
    Fashion200K & 77,106 & 3,607 & 73,499 & 4.68\% \\
    FashionGen-train & 48,476 & 1,942 & 46,534 & 4.01\% \\
    FashionGen-val & 6,086 & 271 & 5,815 & 4.45\% \\
    \midrule
    \textbf{Total} & \textbf{144,379} & \textbf{6,307} & \textbf{138,072} & \textbf{4.37\%} \\
    \bottomrule
  \end{tabular}
\end{table}

\subsection{Caption Statistics}

Table~\ref{tab:caption_stats} shows the average word counts for the three types of captions generated in Stage~1.

\begin{table}[t]
  \caption{Average caption word counts per dataset.}
  \label{tab:caption_stats}
  \small
  \begin{tabular}{lrrr}
    \toprule
    Dataset & Per-Image & Long Cap. & Short Cap. \\
    \midrule
    DeepFashion & 115 & 219 & 39 \\
    Fashion200K & 115 & 220 & 39 \\
    FashionGen-train & 115 & 222 & 40 \\
    FashionGen-val & 115 & 223 & 40 \\
    \bottomrule
  \end{tabular}
\end{table}

\subsection{CIR Triplet View Combinations}

Table~\ref{tab:view_combo} shows the distribution of view combinations across CIR triplets. The dominant combination is back+front (81.6\%), reflecting the typical multi-view display in fashion e-commerce.

\begin{table}[t]
  \caption{View combination distribution in CIR triplets.}
  \label{tab:view_combo}
  \small
  \begin{tabular}{lrr}
    \toprule
    View Combination & Count & Percentage \\
    \midrule
    back + front & 180,040 & 81.6\% \\
    front + side & 27,556 & 12.5\% \\
    back + front + side & 6,390 & 2.9\% \\
    back + side & 4,636 & 2.1\% \\
    back + detail + front & 416 & 0.2\% \\
    Other combinations & 1,695 & 0.8\% \\
    \midrule
    \textbf{Total} & \textbf{220,733} & 100\% \\
    \bottomrule
  \end{tabular}
\end{table}

% ================================================================

\section{Dataset Statistics}
\label{sec:dataset_statistics}

This section presents quantitative statistics of the FashionMV dataset, covering product counts, image view distributions, CIR triplet distributions, and text length distributions for captions and modification texts.

\subsection{Views-per-Product Distribution}

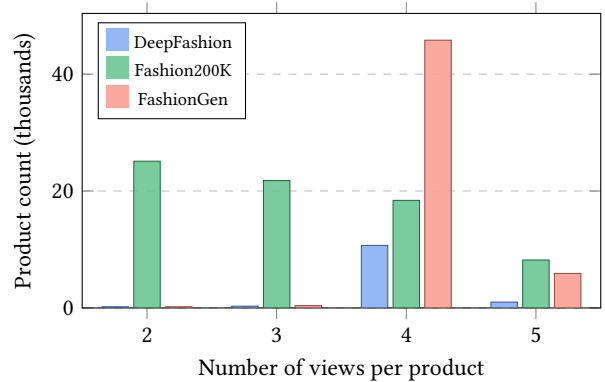
\begin{figure}[tp]
\centering
\begin{tikzpicture}
\begin{axis}[
  ybar,
  bar width=10pt,
  width=\columnwidth,
  height=5.5cm,
  xlabel={Number of views per product},
  ylabel={Product count (thousands)},
  xtick={2,3,4,5},
  xticklabels={2,3,4,5},
  xmin=1.5, xmax=5.5,
  ymin=0,
  ymajorgrids=true,
  grid style=dashed,
  legend style={at={(0.03,0.97)},anchor=north west,font=\footnotesize},
  every axis plot/.append style={fill opacity=0.85},
  legend image code/.code={\draw[#1,draw=none] (0cm,-0.1cm) rectangle (0.3cm,0.2cm);},
]
% DeepFashion: 2->0, 3->0, 4->10774, 5->1451 (approx from avg 4.04, std 0.64)
% f200k:       2->25k, 3->22k, 4->18k, 5->8k (approx from avg 3.33)
% fashiongen:  2->0, 3->0, 4->46k, 5->6k (approx from avg 4.13, std 0.35)
% Using exact per-dataset stats from views_per_dataset.csv
% deepfashion: min=2, p25=4, median=4, p75=4, max=5 -> mostly 4-view
% f200k: min=2, p25=2, median=3, p75=4, max=5 -> spread
% fashiongen: min=2, p25=4, median=4, p75=4, max=5 -> mostly 4-view
\addplot[fill=myblue!80, draw=myblue!60!black]
  coordinates {(2,0.2) (3,0.3) (4,10.7) (5,1.0)};
\addplot[fill=mygreen!80, draw=mygreen!60!black]
  coordinates {(2,25.1) (3,21.8) (4,18.4) (5,8.2)};
\addplot[fill=mysalmon!80, draw=mysalmon!60!black]
  coordinates {(2,0.2) (3,0.4) (4,45.8) (5,5.9)};
\legend{DeepFashion, Fashion200K, FashionGen}
\end{axis}
\end{tikzpicture}
\caption{Distribution of the number of multi-view images per product, broken down by sub-dataset.}
\label{fig:views_per_product}
\end{figure}

Figure~\ref{fig:views_per_product} shows the distribution of the number of multi-view images per product across all three sub-datasets.

\subsection{CIR Triplet Distribution}

The FashionMV dataset contains a total of \textbf{220{,}733} CIR triplets.
The dominant view combination is \texttt{back+front} (180.0K, 81.6\%), followed by \texttt{front+side} (27.6K) and \texttt{back+front+side} (6.4K).

\subsection{Text Length Distributions}

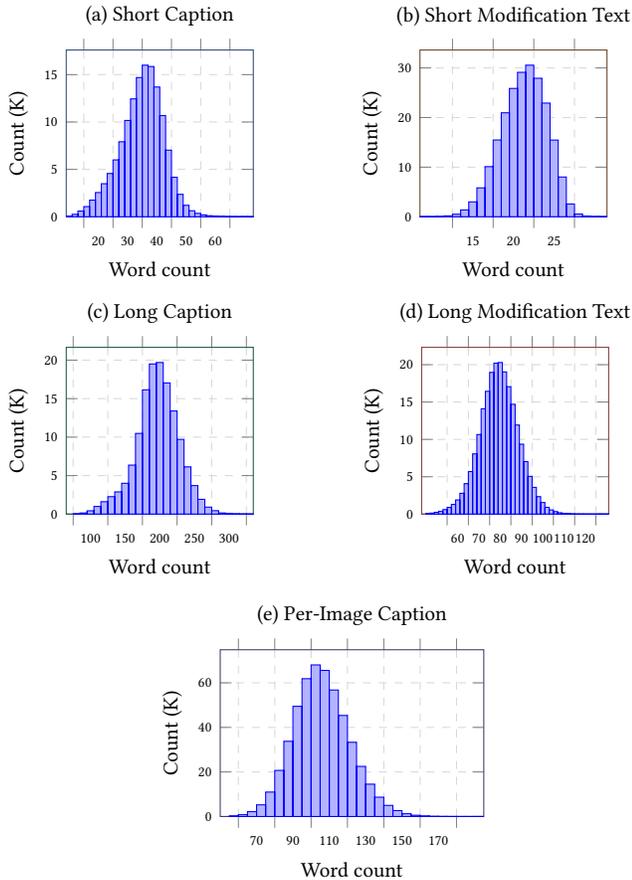
\begin{figure}[tp]
\centering
% Row 1: Short Caption | Short Modification Text
\begin{tikzpicture}
\begin{axis}[
  ybar interval,
  width=0.48\columnwidth,
  height=3.8cm,
  title={\small (a) Short Caption},
  xlabel={\small Word count},
  ylabel={\small Count (K)},
  xmin=14, xmax=78,
  ymin=0,
  ymajorgrids=true,
  grid style={dashed,gray!30},
  fill=myblue!70,
  draw=myblue!50!black,
  xtick={20,30,40,50,60,70},
  tick label style={font=\tiny},
  title style={font=\small},
  xlabel style={font=\tiny},
  ylabel style={font=\tiny},
]
\addplot coordinates {
(14,0.065)(16,0.256)(18,0.603)(20,1.073)(22,1.756)(24,2.555)
(26,3.487)(28,4.566)(30,6.003)(32,7.916)(34,10.164)(36,12.443)
(38,14.695)(40,16.000)(42,15.860)(44,13.695)(46,10.679)(48,7.029)
(50,4.170)(52,2.369)(54,1.210)(56,0.635)(58,0.359)(60,0.198)
(62,0.111)(64,0.050)(66,0.034)(68,0.011)(70,0.008)(72,0.002)
(74,0.002)(76,0.001)(78,0)
};
\end{axis}
\end{tikzpicture}%
\hfill%
\begin{tikzpicture}
\begin{axis}[
  ybar interval,
  width=0.48\columnwidth,
  height=3.8cm,
  title={\small (b) Short Modification Text},
  xlabel={\small Word count},
  ylabel={\small Count (K)},
  xmin=11, xmax=34,
  ymin=0,
  ymajorgrids=true,
  grid style={dashed,gray!30},
  fill=myorange!70,
  draw=myorange!50!black,
  xtick={15,20,25,30},
  tick label style={font=\tiny},
  title style={font=\small},
  xlabel style={font=\tiny},
  ylabel style={font=\tiny},
]
\addplot coordinates {
(11,0.005)(12,0.012)(13,0.050)(14,0.179)(15,0.512)(16,1.369)
(17,3.001)(18,5.824)(19,10.103)(20,15.477)(21,20.957)(22,25.866)
(23,29.112)(24,30.554)(25,27.917)(26,22.913)(27,15.457)(28,8.016)
(29,2.573)(30,0.541)(31,0.095)(32,0.024)(33,0.003)(34,0)
};
\end{axis}
\end{tikzpicture}

\vspace{3pt}

% Row 2: Long Caption | Long Modification Text
\begin{tikzpicture}
\begin{axis}[
  ybar interval,
  width=0.48\columnwidth,
  height=3.8cm,
  title={\small (c) Long Caption},
  xlabel={\small Word count},
  ylabel={\small Count (K)},
  xmin=90, xmax=360,
  ymin=0,
  ymajorgrids=true,
  grid style={dashed,gray!30},
  fill=mygreen!70,
  draw=mygreen!50!black,
  xtick={100,150,200,250,300,350},
  tick label style={font=\tiny},
  title style={font=\small},
  xlabel style={font=\tiny},
  ylabel style={font=\tiny},
]
\addplot coordinates {
(100,0.032)(110,0.145)(120,0.428)(130,0.992)(140,1.609)(150,2.262)
(160,2.885)(170,4.001)(180,6.363)(190,10.481)(200,16.144)(210,19.505)
(220,19.695)(230,17.040)(240,13.397)(250,9.705)(260,6.136)(270,3.705)
(280,1.916)(290,0.903)(300,0.443)(310,0.145)(320,0.065)(330,0.022)
(340,0.005)(350,0.001)(360,0)
};
\end{axis}
\end{tikzpicture}%
\hfill%
\begin{tikzpicture}
\begin{axis}[
  ybar interval,
  width=0.48\columnwidth,
  height=3.8cm,
  title={\small (d) Long Modification Text},
  xlabel={\small Word count},
  ylabel={\small Count (K)},
  xmin=48, xmax=136,
  ymin=0,
  ymajorgrids=true,
  grid style={dashed,gray!30},
  fill=mysalmon!70,
  draw=mysalmon!50!black,
  xtick={60,70,80,90,100,110,120,130},
  tick label style={font=\tiny},
  title style={font=\small},
  xlabel style={font=\tiny},
  ylabel style={font=\tiny},
]
\addplot coordinates {
(50,0.095)(52,0.131)(54,0.237)(56,0.357)(58,0.602)(60,0.895)
(62,1.293)(64,1.893)(66,2.784)(68,4.076)(70,5.691)(72,8.061)
(74,10.648)(76,14.101)(78,16.439)(80,18.972)(82,20.205)(84,20.294)
(86,19.032)(88,17.054)(90,14.701)(92,11.923)(94,9.356)(96,7.031)
(98,5.047)(100,3.580)(102,2.336)(104,1.551)(106,0.906)(108,0.557)
(110,0.335)(112,0.180)(114,0.075)(116,0.049)(118,0.016)(120,0.007)
(122,0.002)(124,0.002)(132,0.001)(136,0)
};
\end{axis}
\end{tikzpicture}

\vspace{3pt}

% Row 3: Per-Image Caption (centered, wider)
\begin{center}
\begin{tikzpicture}
\begin{axis}[
  ybar interval,
  width=0.60\columnwidth,
  height=3.8cm,
  title={\small (e) Per-Image Caption},
  xlabel={\small Word count},
  ylabel={\small Count (K)},
  xmin=60, xmax=205,
  ymin=0,
  ymajorgrids=true,
  grid style={dashed,gray!30},
  fill=mypurple!70,
  draw=mypurple!50!black,
  xtick={70,90,110,130,150,170,190},
  tick label style={font=\tiny},
  title style={font=\small},
  xlabel style={font=\tiny},
  ylabel style={font=\tiny},
]
\addplot coordinates {
(65,0.289)(70,0.808)(75,2.173)(80,5.235)(85,10.972)(90,20.693)
(95,33.762)(100,49.469)(105,61.857)(110,68.042)(115,65.576)(120,56.771)
(125,45.361)(130,33.320)(135,22.496)(140,14.517)(145,8.678)(150,4.992)
(155,2.663)(160,1.250)(165,0.567)(170,0.257)(175,0.098)(180,0.039)
(185,0.015)(190,0.003)(195,0.003)(200,0.002)(205,0)
};
\end{axis}
\end{tikzpicture}
\end{center}
\caption{Word-count distributions for all five text types in FashionMV.
(a) Short captions peak around 40 words (median=40).
(b) Short modification texts are tightly concentrated at 23--24 words (median=23).
(c) Long captions are normally distributed around 220 words (median=222).
(d) Long modification texts peak around 82--84 words (median=84).
(e) Per-image captions concentrate at 105--115 words (median=115).}
\label{fig:text_length_dist}
\end{figure}

Figure~\ref{fig:text_length_dist} shows the word-count distributions for the five types of text in FashionMV: short captions, long captions, per-image captions, short modification texts, and long modification texts.

% ================================================================
% Part II: Model and Experiments
% ================================================================
\section{Implementation Details}
\label{sec:implementation}

This section provides full implementation details, including training hyperparameters, evaluation protocol, and SFT pre-training data formats and statistics.

\subsection{Training Configuration}

Table~\ref{tab:hyperparams} lists the full set of hyperparameters used for contrastive embedding training.

\begin{table}[t]
  \caption{Full training hyperparameters.}
  \label{tab:hyperparams}
  \small
  \begin{tabular}{lr}
    \toprule
    Parameter & Value \\
    \midrule
    Backbone & Qwen3.5-0.8B \\
    Hidden dimension & 1024 \\
    Transformer layers & 24 \\
    Image resolution & $128{\times}128$ -- $512{\times}512$ \\
    Max views per product & 5 \\
    Optimizer & AdamW \\
    Learning rate & $1 \times 10^{-5}$ \\
    Weight decay & 0.01 \\
    LR schedule & Cosine (no warmup) \\
    Gradient clipping & 1.0 \\
    Precision & bfloat16 \\
    Batch size per GPU & 16 \\
    GPUs & 10 $\times$ 3090 \\
    Effective batch size & 160 \\
    Training epochs & 1 \\
    Total training steps & 1,175 \\
    Temperature $\tau$ & 0.07 \\
    $\lambda_d$ (doc alignment) & 0.25 \\
    $\lambda_s$ (src alignment) & 0.25 \\
    CoT keep ratio schedule & Linear $1.0 \to 0.0$ over first 50\% steps \\
    Long/short text sampling & 50\%/50\% per step \\
    \bottomrule
  \end{tabular}
\end{table}

\subsection{Evaluation Protocol}

\begin{itemize}
  \item \textbf{Gallery construction}: Each dataset's validation products form an independent gallery. Product-level embeddings are computed by processing all multi-view images through the model.
  \item \textbf{CIR evaluation}: For each query (source product + modification text), we compute cosine similarity against all gallery embeddings and rank by similarity.
  \item \textbf{I2T/T2I evaluation}: We compute cosine similarity between product visual embeddings and caption text embeddings. For I2T, each visual embedding queries the text gallery; for T2I, each text embedding queries the visual gallery.
\end{itemize}

\subsection{SFT Pre-training Details}
\label{sec:sft_details}

The backbone model (Qwen3.5-0.8B) is first fine-tuned via supervised fine-tuning (SFT) on two fashion-domain tasks before the contrastive embedding training. This SFT stage teaches the model fashion-specific visual understanding and structured output generation.

\paragraph{Task 1 — Multi-View Caption Generation.}
Given up to 5 images of a product, the model generates per-image descriptions (in a \texttt{<think>} block) and then a structured JSON with \texttt{long\_caption} and \texttt{short\_caption} fields. The conversation format is:

\begin{small}
\begin{verbatim}
[System]: <caption_system_prompt>
[User]:   <img_1>...<img_N>
[Asst]:   <think>
              [Image 1] front view...
              [Image 2] back view...
          </think>
          {"long_caption":"...", "short_caption":"..."}
\end{verbatim}
\end{small}

\paragraph{Task 2 — CIR Triplet Generation.}
Given source and target product images, the model generates the modification text describing how to transform the source into the target. This is a 3-turn dialogue:

\begin{small}
\begin{verbatim}
[User]:   <src_img_1>...<src_img_N>
[Asst]:   <think>{source_long_caption}</think> <emb_all>
[User]:   <tgt_img_1>...<tgt_img_M>
[Asst]:   <think>{target_long_caption}</think> <emb_all>
[User]:   Generate modification text...
[Asst]:   <think></think>
          {"views_involved": [...],
           "modification_text_long": "...",
           "modification_text_short": "..."}
\end{verbatim}
\end{small}

\paragraph{SFT Data Statistics.}
Table~\ref{tab:sft_stats} summarises the SFT training data. The caption data covers all three training splits; the CIR data covers all training triplets. Both datasets are sampled at 20\% for the SFT run used in our experiments.

\begin{table}[tp]
  \centering
  \caption{SFT training data statistics (full dataset; 20\% sampled for training).}
  \label{tab:sft_stats}
  \small
  \begin{tabular}{lrr}
    \toprule
    \textbf{Task} & \textbf{Full Size} & \textbf{Used (20\%)} \\
    \midrule
    Caption Generation & 587,050 & 117,610 \\
    CIR Triplet Generation & 939,700 & 187,940 \\
    \midrule
    \textbf{Total} & \textbf{1,526,750} & \textbf{305,550} \\
    \bottomrule
  \end{tabular}
\end{table}

\paragraph{SFT Hyperparameters.}
The SFT uses LLaMA-Factory with DeepSpeed ZeRO-3, full-parameter fine-tuning, cosine LR schedule ($\text{lr}=10^{-5}$), batch size 10 (1 per GPU $\times$ 10 gradient accumulation steps), 1 epoch, and a context length of 4,096 tokens. The \texttt{<emb\_all>} special token is added and the vocabulary is resized accordingly.

\section{Additional Experimental Results}
\label{sec:additional_results}

This section supplements the main paper with complete Image-to-Text (I2T) and Text-to-Image (T2I) retrieval results for all 16 model configurations.
As short captions better reflect real-world query scenarios (concise user descriptions), we report results using \textbf{short captions} throughout this section.
Table~\ref{tab:i2t_short} reports I2T retrieval performance, and Table~\ref{tab:t2i_short} reports T2I retrieval performance.
The alignment mechanism (\textbf{Align}) produces strong image--text retrieval capabilities as a byproduct of CIR training; in particular, the two-stage dialogue (\textbf{MT}) combined with alignment achieves the best results across all three datasets.

\subsection{Image-to-Text (I2T) Retrieval}
\label{sec:i2t_sub}

Table~\ref{tab:i2t_short} presents I2T retrieval R@1/R@5 (\%) for all 16 configurations with short captions.
MT = two-stage dialogue, Align = caption-based alignment, CoT = chain-of-thought, SFT = supervised fine-tuning pre-training.
A $\surd$ indicates the corresponding mechanism is active.
\textbf{Bold}: best per column; \underline{underline}: second best.

\begin{table}[tp]
  \caption{I2T retrieval R@1/R@5 (\%) with \textbf{short captions}. MT = two-stage dialogue, Align = caption-based alignment, CoT = chain-of-thought, SFT = supervised fine-tuning. \textbf{Bold}: best per column; \underline{underline}: second best.}
  \label{tab:i2t_short}
  \small
  \centering
  \setlength{\tabcolsep}{3pt}
  \begin{tabular}{cccc|cc|cc|cc}
    \toprule
    & & & & \multicolumn{2}{c|}{\textbf{DeepFashion}} & \multicolumn{2}{c|}{\textbf{F200K}} & \multicolumn{2}{c}{\textbf{FashionGen}} \\
    MT & Align & CoT & SFT & R@1 & R@5 & R@1 & R@5 & R@1 & R@5 \\
    \midrule
    \multicolumn{10}{l}{\emph{Pretrained base}} \\
     &  &  &  & 60.2 & 89.6 & 57.8 & 81.8 & 53.1 & 78.6 \\
     & $\surd$ &  &  & 71.1 & 94.4 & 64.1 & 85.9 & 52.7 & 78.1 \\
     &  & $\surd$ &  & 44.2 & 73.3 & 28.9 & 53.5 & 21.7 & 43.1 \\
     & $\surd$ & $\surd$ &  & 50.9 & 79.7 & 35.8 & 61.7 & 23.8 & 46.9 \\
    $\surd$ &  &  &  & 62.7 & 90.6 & 56.4 & 81.9 & \underline{53.6} & 78.4 \\
    $\surd$ & $\surd$ &  &  & \underline{75.4} & \underline{95.6} & \underline{66.1} & \underline{87.6} & \textbf{57.0} & \textbf{81.0} \\
    $\surd$ &  & $\surd$ &  & 35.6 & 67.7 & 29.0 & 54.1 & 26.2 & 50.5 \\
    $\surd$ & $\surd$ & $\surd$ &  & \textbf{78.3} & \textbf{97.1} & \textbf{67.9} & \textbf{88.6} & \textbf{57.0} & \underline{80.7} \\
    \midrule
    \multicolumn{10}{l}{\emph{SFT base}} \\
     &  &  & $\surd$ & 53.3 & 84.7 & 66.9 & 87.6 & 57.1 & 82.9 \\
     & $\surd$ &  & $\surd$ & 80.3 & 97.4 & \underline{77.4} & \underline{93.2} & \underline{66.5} & \underline{88.8} \\
     &  & $\surd$ & $\surd$ & 45.2 & 75.5 & 35.3 & 61.0 & 29.0 & 52.7 \\
     & $\surd$ & $\surd$ & $\surd$ & 50.9 & 80.2 & 42.8 & 68.7 & 32.1 & 56.3 \\
    $\surd$ &  &  & $\surd$ & 45.9 & 78.3 & 61.8 & 84.6 & 54.2 & 80.7 \\
    $\surd$ & $\surd$ &  & $\surd$ & \underline{81.9} & \textbf{97.9} & \textbf{78.9} & \textbf{94.1} & \textbf{68.3} & \textbf{90.0} \\
    $\surd$ &  & $\surd$ & $\surd$ & 45.3 & 76.6 & 37.7 & 65.4 & 32.3 & 55.6 \\
    $\surd$ & $\surd$ & $\surd$ & $\surd$ & \textbf{82.7} & \underline{97.7} & 75.6 & 91.9 & 65.9 & 87.3 \\
    \bottomrule
  \end{tabular}
\end{table}

\subsection{Text-to-Image (T2I) Retrieval}
\label{sec:t2i_sub}

Table~\ref{tab:t2i_short} presents T2I retrieval R@1/R@5 (\%) for all 16 configurations with short captions.
The alignment mechanism consistently provides the largest performance gain in both directions, confirming that the auxiliary alignment loss effectively calibrates the visual--textual embedding space.
Furthermore, the two-stage dialogue architecture enables source-side alignment ($\mathcal{L}_\text{src}$), which further improves performance: \texttt{MT+Align} outperforms \texttt{Align}-only variants across all three datasets.

\begin{table}[tp]
  \caption{T2I retrieval R@1/R@5 (\%) with \textbf{short captions}. MT = two-stage dialogue, Align = caption-based alignment, CoT = chain-of-thought, SFT = supervised fine-tuning. \textbf{Bold}: best per column; \underline{underline}: second best.}
  \label{tab:t2i_short}
  \small
  \centering
  \setlength{\tabcolsep}{3pt}
  \begin{tabular}{cccc|cc|cc|cc}
    \toprule
    & & & & \multicolumn{2}{c|}{\textbf{DeepFashion}} & \multicolumn{2}{c|}{\textbf{F200K}} & \multicolumn{2}{c}{\textbf{FashionGen}} \\
    MT & Align & CoT & SFT & R@1 & R@5 & R@1 & R@5 & R@1 & R@5 \\
    \midrule
    \multicolumn{10}{l}{\emph{Pretrained base}} \\
     &  &  &  & 71.2 & 93.5 & 60.3 & 83.5 & 57.3 & 81.5 \\
     & $\surd$ &  &  & 75.9 & 95.6 & 66.3 & 87.0 & 56.8 & 81.3 \\
     &  & $\surd$ &  & 56.6 & 85.0 & 41.3 & 67.4 & 38.5 & 64.9 \\
     & $\surd$ & $\surd$ &  & 65.6 & 90.9 & 54.6 & 78.9 & 45.8 & 72.4 \\
    $\surd$ &  &  &  & 65.7 & 90.5 & 55.2 & 81.6 & 56.6 & 81.4 \\
    $\surd$ & $\surd$ &  &  & \underline{78.1} & \underline{96.2} & \underline{70.4} & \underline{89.5} & \underline{60.6} & \underline{83.9} \\
    $\surd$ &  & $\surd$ &  & 50.6 & 81.2 & 33.0 & 58.9 & 38.7 & 63.9 \\
    $\surd$ & $\surd$ & $\surd$ &  & \textbf{81.2} & \textbf{97.4} & \textbf{73.0} & \textbf{90.6} & \textbf{62.9} & \textbf{85.1} \\
    \midrule
    \multicolumn{10}{l}{\emph{SFT base}} \\
     &  &  & $\surd$ & 67.4 & 92.4 & 68.9 & 88.5 & 60.3 & 84.1 \\
     & $\surd$ &  & $\surd$ & \underline{86.0} & \underline{98.3} & \underline{79.5} & \underline{94.0} & \underline{69.3} & \underline{89.6} \\
     &  & $\surd$ & $\surd$ & 65.2 & 90.6 & 51.4 & 77.3 & 43.3 & 67.7 \\
     & $\surd$ & $\surd$ & $\surd$ & 73.7 & 94.6 & 63.1 & 85.5 & 48.5 & 74.6 \\
    $\surd$ &  &  & $\surd$ & 60.8 & 90.5 & 64.8 & 86.4 & 56.8 & 80.4 \\
    $\surd$ & $\surd$ &  & $\surd$ & \textbf{87.1} & \textbf{98.6} & \textbf{81.5} & \textbf{95.0} & \textbf{71.8} & \textbf{91.0} \\
    $\surd$ &  & $\surd$ & $\surd$ & 66.9 & 92.6 & 56.0 & 81.4 & 51.7 & 76.6 \\
    $\surd$ & $\surd$ & $\surd$ & $\surd$ & 85.3 & 98.1 & 79.4 & 93.6 & 69.1 & 89.1 \\
    \bottomrule
  \end{tabular}
\end{table}

\textbf{Key observations:}
\begin{itemize}
  \item The alignment mechanism (\textbf{Align}) provides the largest single improvement in both I2T and T2I performance. For example, adding Align to MT+SFT yields +36.0pp I2T R@1 on DeepFashion (45.9\%$\to$81.9\%).
  \item The two-stage dialogue (\textbf{MT}) combined with alignment consistently achieves the best or second-best results in both retrieval directions, confirming the complementary nature of these two mechanisms.
  \item CoT without alignment degrades I2T/T2I performance, consistent with the CIR finding that CoT requires alignment to be beneficial.
  \item SFT pre-training significantly boosts performance for alignment variants, with \texttt{MT+Align+SFT} achieving the best results overall.
\end{itemize}

% ================================================================

% ================================================================
% Part III: Visual Galleries
% ================================================================
% \clearpage
\onecolumn
\section{Dataset Sample Gallery}
\label{sec:sample_gallery}

This section presents representative product samples from the FashionMV dataset. Each entry shows up to five multi-view images of a garment (front, side, back, full, and additional views) along with its garment type and automatically generated short caption.

\noindent\hfill%
  \begin{minipage}[t][4.1cm][t]{7.75cm}
    \makebox[7.75cm][l]{\includegraphics[width=1.55cm,height=2.7cm,keepaspectratio]{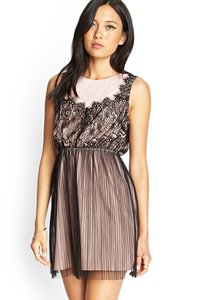}%
    \includegraphics[width=1.55cm,height=2.7cm,keepaspectratio]{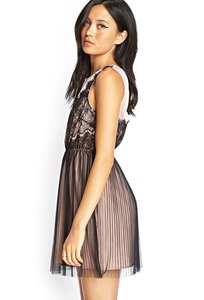}%
    \includegraphics[width=1.55cm,height=2.7cm,keepaspectratio]{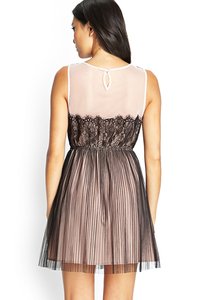}%
    \includegraphics[width=1.55cm,height=2.7cm,keepaspectratio]{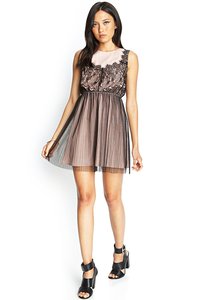}%
    \phantom{\rule{1.55cm}{2.7cm}}}
    \par\vspace{1pt}
    {\scriptsize\textbf{Women's Dresses}}
    \par\vspace{0pt}
    \begin{minipage}[t][1.1cm][t]{7.75cm}
      {\tiny\raggedright Sleeveless mini dress featuring a black floral lace bodice with scalloped edges over blush pink underlay, elasticized waist, and pleated tulle skirt. Round neckline with solid chest panel, keyhole back closure with button, and A-line silhouette. Feminine cocktail style in black and nude tones.}
    \end{minipage}
  \end{minipage}%
\hfill%
  \begin{minipage}[t][4.1cm][t]{7.75cm}
    \makebox[7.75cm][l]{\includegraphics[width=1.55cm,height=2.7cm,keepaspectratio]{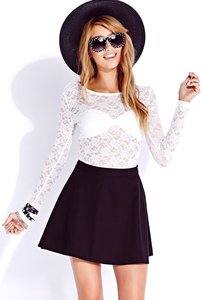}%
    \includegraphics[width=1.55cm,height=2.7cm,keepaspectratio]{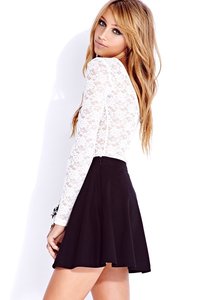}%
    \includegraphics[width=1.55cm,height=2.7cm,keepaspectratio]{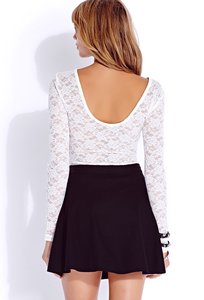}%
    \includegraphics[width=1.55cm,height=2.7cm,keepaspectratio]{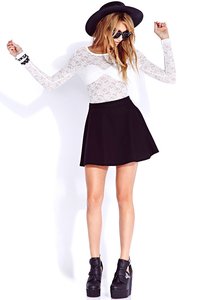}%
    \includegraphics[width=1.55cm,height=2.7cm,keepaspectratio]{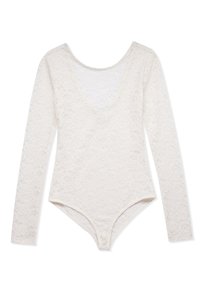}}
    \par\vspace{1pt}
    {\scriptsize\textbf{Women's Tees Tanks}}
    \par\vspace{0pt}
    \begin{minipage}[t][1.1cm][t]{7.75cm}
      {\tiny\raggedright A white floral lace long-sleeve bodysuit featuring a solid white bust panel, sheer lace sleeves, and a deep scoop back. The fitted silhouette includes a round front neckline and high-cut leg openings, combining feminine lace details with practical bodysuit construction for seamless styling.}
    \end{minipage}
  \end{minipage}%
\hfill
\par\vspace{8pt}

\noindent\hfill%
  \begin{minipage}[t][4.1cm][t]{7.75cm}
    \makebox[7.75cm][l]{\includegraphics[width=1.55cm,height=2.7cm,keepaspectratio]{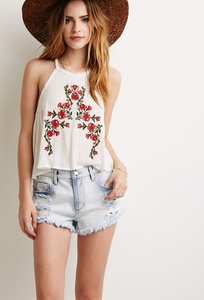}%
    \includegraphics[width=1.55cm,height=2.7cm,keepaspectratio]{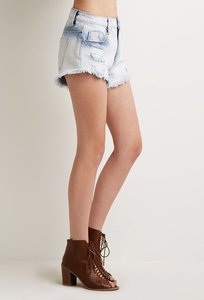}%
    \includegraphics[width=1.55cm,height=2.7cm,keepaspectratio]{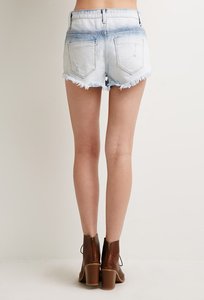}%
    \includegraphics[width=1.55cm,height=2.7cm,keepaspectratio]{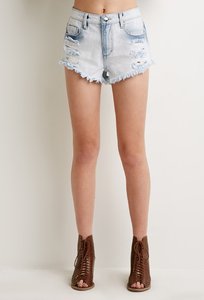}%
    \phantom{\rule{1.55cm}{2.7cm}}}
    \par\vspace{1pt}
    {\scriptsize\textbf{Women's Shorts}}
    \par\vspace{0pt}
    \begin{minipage}[t][1.1cm][t]{7.75cm}
      {\tiny\raggedright Light blue high-waisted denim shorts featuring symmetrical distressed thigh rips, frayed raw hem edges, and a bleached vintage wash. Classic five-pocket design with metal button/zipper closure, belt loops, and subtle back pocket distressing. Slim fit casual summer essential with bohemian-inspired worn-in character.}
    \end{minipage}
  \end{minipage}%
\hfill%
  \begin{minipage}[t][4.1cm][t]{7.75cm}
    \makebox[7.75cm][l]{\includegraphics[width=1.55cm,height=2.7cm,keepaspectratio]{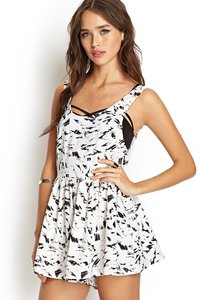}%
    \includegraphics[width=1.55cm,height=2.7cm,keepaspectratio]{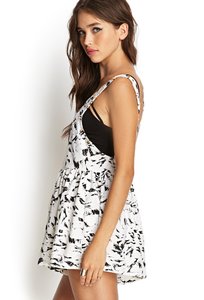}%
    \includegraphics[width=1.55cm,height=2.7cm,keepaspectratio]{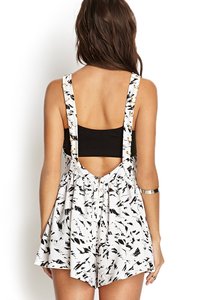}%
    \includegraphics[width=1.55cm,height=2.7cm,keepaspectratio]{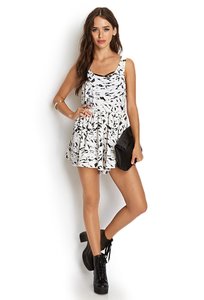}%
    \phantom{\rule{1.55cm}{2.7cm}}}
    \par\vspace{1pt}
    {\scriptsize\textbf{Women's Rompers Jumpsuits}}
    \par\vspace{0pt}
    \begin{minipage}[t][1.1cm][t]{7.75cm}
      {\tiny\raggedright A sleeveless white romper with an abstract black brushstroke print, featuring wide straps with gold grommet hardware, side cutouts, a low V-back, and flared skater-style shorts. Made from lightweight woven fabric with a gathered elastic waist for a comfortable, playful fit.}
    \end{minipage}
  \end{minipage}%
\hfill
\par\vspace{8pt}

\noindent\hfill%
  \begin{minipage}[t][4.1cm][t]{7.75cm}
    \makebox[7.75cm][l]{\includegraphics[width=1.55cm,height=2.7cm,keepaspectratio]{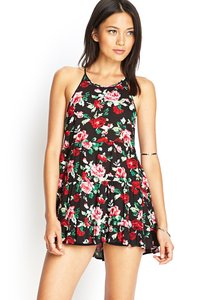}%
    \includegraphics[width=1.55cm,height=2.7cm,keepaspectratio]{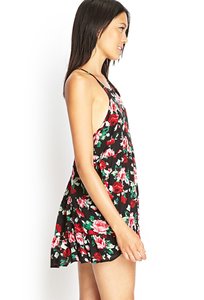}%
    \includegraphics[width=1.55cm,height=2.7cm,keepaspectratio]{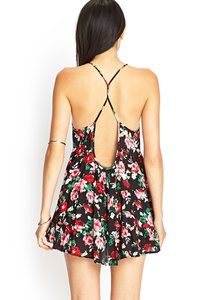}%
    \includegraphics[width=1.55cm,height=2.7cm,keepaspectratio]{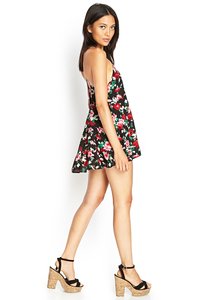}%
    \includegraphics[width=1.55cm,height=2.7cm,keepaspectratio]{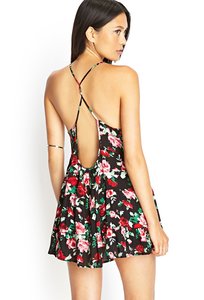}}
    \par\vspace{1pt}
    {\scriptsize\textbf{Women's Dresses}}
    \par\vspace{0pt}
    \begin{minipage}[t][1.1cm][t]{7.75cm}
      {\tiny\raggedright A sleeveless black halter-neck mini dress with a vibrant red and pink rose floral print, featuring a tiered ruffled skirt and dramatic crisscross open back design. Lightweight and flowy with a relaxed fit.}
    \end{minipage}
  \end{minipage}%
\hfill%
  \begin{minipage}[t][4.1cm][t]{7.75cm}
    \makebox[7.75cm][l]{\includegraphics[width=1.55cm,height=2.7cm,keepaspectratio]{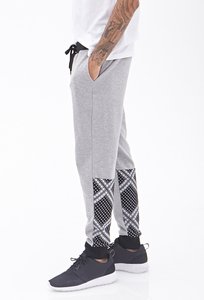}%
    \includegraphics[width=1.55cm,height=2.7cm,keepaspectratio]{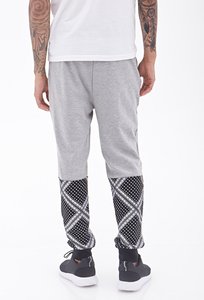}%
    \includegraphics[width=1.55cm,height=2.7cm,keepaspectratio]{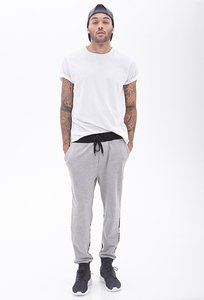}%
    \includegraphics[width=1.55cm,height=2.7cm,keepaspectratio]{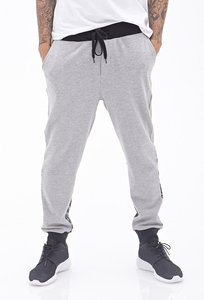}%
    \phantom{\rule{1.55cm}{2.7cm}}}
    \par\vspace{1pt}
    {\scriptsize\textbf{Men's Pants}}
    \par\vspace{0pt}
    \begin{minipage}[t][1.1cm][t]{7.75cm}
      {\tiny\raggedright Heather gray men's jogger sweatpants featuring a contrasting black elastic waistband with drawstring closure, side slash pockets, and distinctive black-and-white bandana print panels on the lower legs. Designed with a relaxed fit through the thighs that tapers to elasticized ankle cuffs for a modern casual silhouette.}
    \end{minipage}
  \end{minipage}%
\hfill
\par\vspace{8pt}

\noindent\hfill%
  \begin{minipage}[t][4.1cm][t]{7.75cm}
    \makebox[7.75cm][l]{\includegraphics[width=1.55cm,height=2.7cm,keepaspectratio]{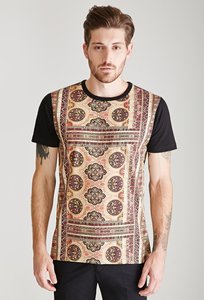}%
    \includegraphics[width=1.55cm,height=2.7cm,keepaspectratio]{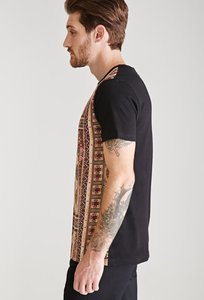}%
    \includegraphics[width=1.55cm,height=2.7cm,keepaspectratio]{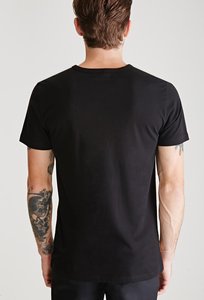}%
    \includegraphics[width=1.55cm,height=2.7cm,keepaspectratio]{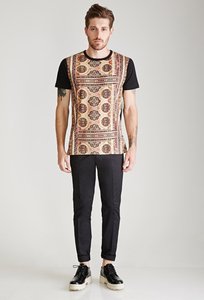}%
    \includegraphics[width=1.55cm,height=2.7cm,keepaspectratio]{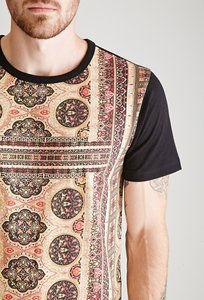}}
    \par\vspace{1pt}
    {\scriptsize\textbf{Men's Tees Tanks}}
    \par\vspace{0pt}
    \begin{minipage}[t][1.1cm][t]{7.75cm}
      {\tiny\raggedright Men's slim-fit short-sleeve t-shirt featuring an ornate tapestry-inspired medallion print on the front panel with solid black back and sleeves, crew neckline with ribbed trim, and bold ethnic-inspired geometric patterns in beige, black, and red tones creating a striking contrast.}
    \end{minipage}
  \end{minipage}%
\hfill%
  \begin{minipage}[t][4.1cm][t]{7.75cm}
    \makebox[7.75cm][l]{\includegraphics[width=1.55cm,height=2.7cm,keepaspectratio]{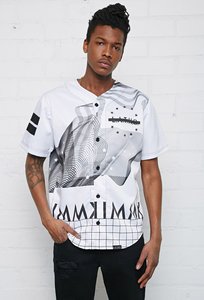}%
    \includegraphics[width=1.55cm,height=2.7cm,keepaspectratio]{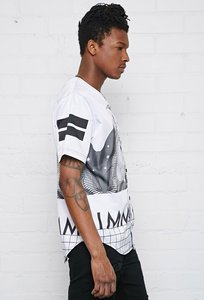}%
    \includegraphics[width=1.55cm,height=2.7cm,keepaspectratio]{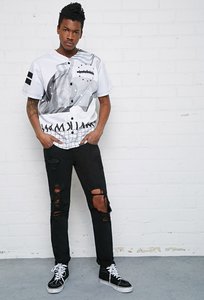}%
    \includegraphics[width=1.55cm,height=2.7cm,keepaspectratio]{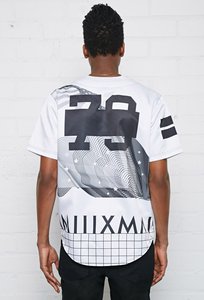}%
    \phantom{\rule{1.55cm}{2.7cm}}}
    \par\vspace{1pt}
    {\scriptsize\textbf{Men's Tees Tanks}}
    \par\vspace{0pt}
    \begin{minipage}[t][1.1cm][t]{7.75cm}
      {\tiny\raggedright White baseball jersey with black abstract wave prints, featuring asymmetrical striped sleeve, "79" back graphic, "LATHC" front branding, button-front closure, and curved hem with grid pattern. Relaxed fit streetwear style.}
    \end{minipage}
  \end{minipage}%
\hfill
\par\vspace{8pt}

\noindent\hfill%
  \begin{minipage}[t][4.1cm][t]{7.75cm}
    \makebox[7.75cm][l]{\includegraphics[width=1.55cm,height=2.7cm,keepaspectratio]{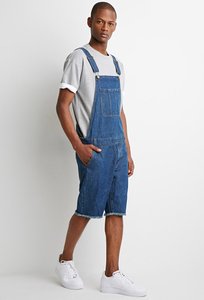}%
    \includegraphics[width=1.55cm,height=2.7cm,keepaspectratio]{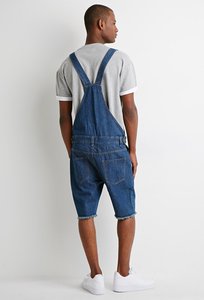}%
    \includegraphics[width=1.55cm,height=2.7cm,keepaspectratio]{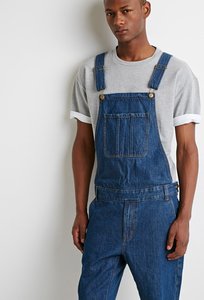}%
    \phantom{\rule{1.55cm}{2.7cm}}%
    \phantom{\rule{1.55cm}{2.7cm}}}
    \par\vspace{1pt}
    {\scriptsize\textbf{Men's Shorts}}
    \par\vspace{0pt}
    \begin{minipage}[t][1.1cm][t]{7.75cm}
      {\tiny\raggedright Men's medium blue denim short overalls featuring adjustable shoulder straps with metal hardware, a chest bib pocket with contrast stitching, and frayed raw-edge hems at knee length. Includes side hip pockets, dual back pockets, belt loops, and a relaxed fit perfect for casual summer layering.}
    \end{minipage}
  \end{minipage}%
\hfill%
  \begin{minipage}[t][4.1cm][t]{7.75cm}
    \makebox[7.75cm][l]{\includegraphics[width=1.55cm,height=2.7cm,keepaspectratio]{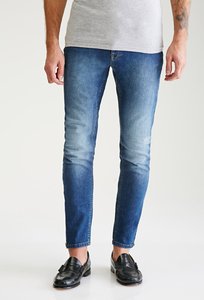}%
    \includegraphics[width=1.55cm,height=2.7cm,keepaspectratio]{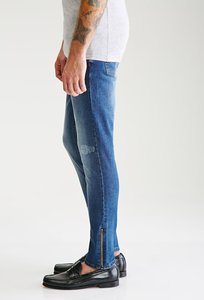}%
    \includegraphics[width=1.55cm,height=2.7cm,keepaspectratio]{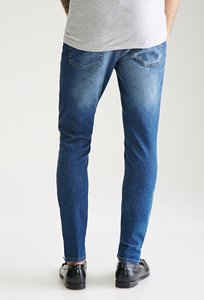}%
    \includegraphics[width=1.55cm,height=2.7cm,keepaspectratio]{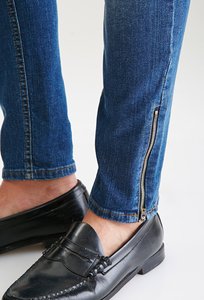}%
    \includegraphics[width=1.55cm,height=2.7cm,keepaspectratio]{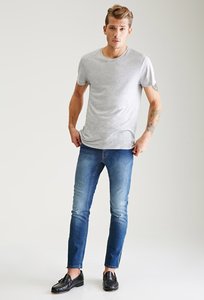}}
    \par\vspace{1pt}
    {\scriptsize\textbf{Men's Denim}}
    \par\vspace{0pt}
    \begin{minipage}[t][1.1cm][t]{7.75cm}
      {\tiny\raggedright Men's slim-fit ankle jeans in medium blue wash with thigh whiskering, contrast stitching, and a distinctive metal zipper detail on the wearer's left outer ankle. Features five-pocket styling, subtle knee distressing, and a modern tapered silhouette.}
    \end{minipage}
  \end{minipage}%
\hfill
\par\vspace{8pt}

\noindent\hfill%
  \begin{minipage}[t][4.1cm][t]{7.75cm}
    \makebox[7.75cm][l]{\includegraphics[width=1.55cm,height=2.7cm,keepaspectratio]{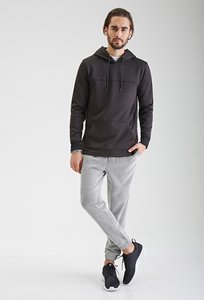}%
    \includegraphics[width=1.55cm,height=2.7cm,keepaspectratio]{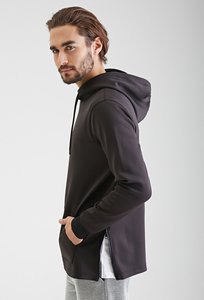}%
    \includegraphics[width=1.55cm,height=2.7cm,keepaspectratio]{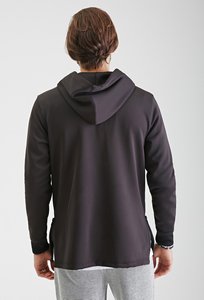}%
    \includegraphics[width=1.55cm,height=2.7cm,keepaspectratio]{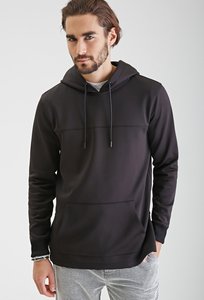}%
    \includegraphics[width=1.55cm,height=2.7cm,keepaspectratio]{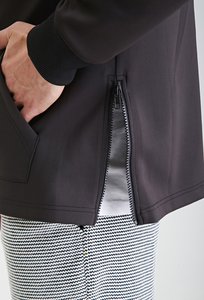}}
    \par\vspace{1pt}
    {\scriptsize\textbf{Men's Sweatshirts Hoodies}}
    \par\vspace{0pt}
    \begin{minipage}[t][1.1cm][t]{7.75cm}
      {\tiny\raggedright Black technical hoodie featuring a drawstring hood with metal-tipped cords, kangaroo front pocket, and distinctive side zip vents revealing silver mesh lining. Long sleeves with ribbed cuffs, horizontal chest seam, relaxed fit, and smooth neoprene-like fabric.}
    \end{minipage}
  \end{minipage}%
\hfill%
  \begin{minipage}[t][4.1cm][t]{7.75cm}
    \makebox[7.75cm][l]{\includegraphics[width=1.55cm,height=2.7cm,keepaspectratio]{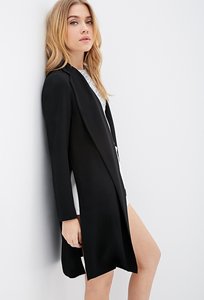}%
    \includegraphics[width=1.55cm,height=2.7cm,keepaspectratio]{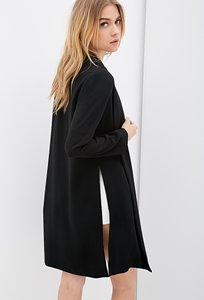}%
    \includegraphics[width=1.55cm,height=2.7cm,keepaspectratio]{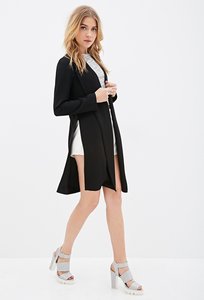}%
    \includegraphics[width=1.55cm,height=2.7cm,keepaspectratio]{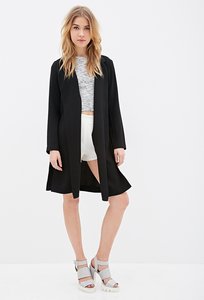}%
    \phantom{\rule{1.55cm}{2.7cm}}}
    \par\vspace{1pt}
    {\scriptsize\textbf{Women's Jackets Coats}}
    \par\vspace{0pt}
    \begin{minipage}[t][1.1cm][t]{7.75cm}
      {\tiny\raggedright A black longline open-front coat featuring notched lapels, long sleeves, and dramatic high side slits on both sides. The minimalist design has no closures or pockets, crafted from lightweight woven fabric in a relaxed straight fit that falls to mid-thigh.}
    \end{minipage}
  \end{minipage}%
\hfill
\par\vspace{8pt}

\noindent\hfill%
  \begin{minipage}[t][4.1cm][t]{7.75cm}
    \makebox[7.75cm][l]{\includegraphics[width=1.55cm,height=2.7cm,keepaspectratio]{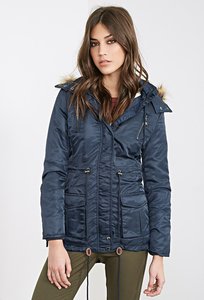}%
    \includegraphics[width=1.55cm,height=2.7cm,keepaspectratio]{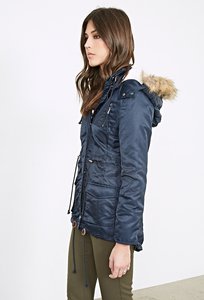}%
    \includegraphics[width=1.55cm,height=2.7cm,keepaspectratio]{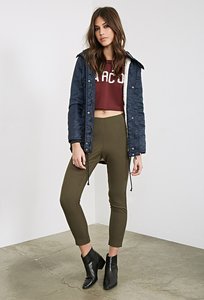}%
    \includegraphics[width=1.55cm,height=2.7cm,keepaspectratio]{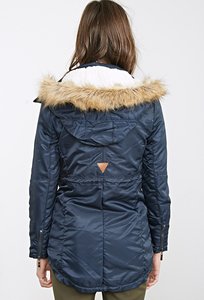}%
    \phantom{\rule{1.55cm}{2.7cm}}}
    \par\vspace{1pt}
    {\scriptsize\textbf{Women's Jackets Coats}}
    \par\vspace{0pt}
    \begin{minipage}[t][1.1cm][t]{7.75cm}
      {\tiny\raggedright Navy blue utility parka jacket featuring a faux fur-trimmed hood with white sherpa lining, adjustable drawstring waist, and multiple storage pockets. Constructed from shiny satin-finish nylon with distinctive triangular leather back detail, fishtail hem, and elasticized cuffs. Front zipper and snap button closure.}
    \end{minipage}
  \end{minipage}%
\hfill%
  \begin{minipage}[t][4.1cm][t]{7.75cm}
    \makebox[7.75cm][l]{\includegraphics[width=1.55cm,height=2.7cm,keepaspectratio]{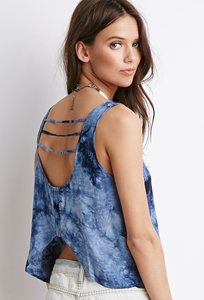}%
    \includegraphics[width=1.55cm,height=2.7cm,keepaspectratio]{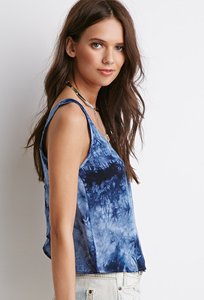}%
    \includegraphics[width=1.55cm,height=2.7cm,keepaspectratio]{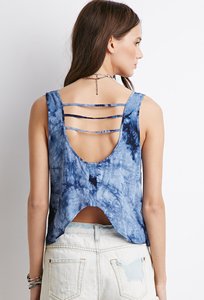}%
    \includegraphics[width=1.55cm,height=2.7cm,keepaspectratio]{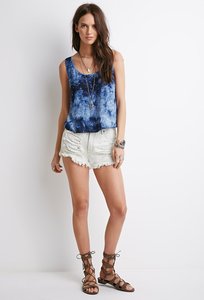}%
    \includegraphics[width=1.55cm,height=2.7cm,keepaspectratio]{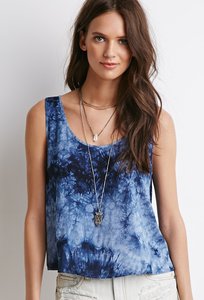}}
    \par\vspace{1pt}
    {\scriptsize\textbf{Women's Blouses Shirts}}
    \par\vspace{0pt}
    \begin{minipage}[t][1.1cm][t]{7.75cm}
      {\tiny\raggedright Blue tie-dye sleeveless tank top with relaxed cropped fit, scoop neckline, and distinctive open back featuring three horizontal ladder straps. Constructed from lightweight jersey knit fabric with wide armholes and a casual bohemian aesthetic.}
    \end{minipage}
  \end{minipage}%
\hfill
\par\vspace{8pt}

\noindent\hfill%
  \begin{minipage}[t][4.1cm][t]{7.75cm}
    \makebox[7.75cm][l]{\includegraphics[width=1.55cm,height=2.7cm,keepaspectratio]{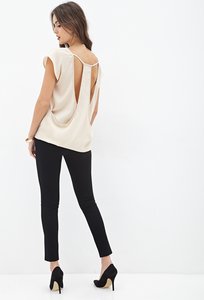}%
    \includegraphics[width=1.55cm,height=2.7cm,keepaspectratio]{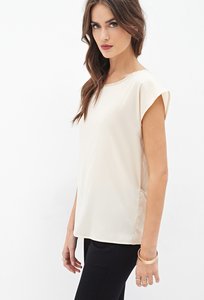}%
    \includegraphics[width=1.55cm,height=2.7cm,keepaspectratio]{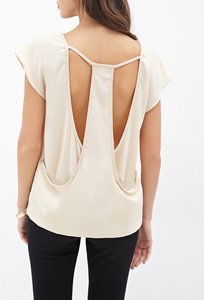}%
    \includegraphics[width=1.55cm,height=2.7cm,keepaspectratio]{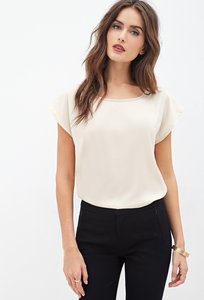}%
    \phantom{\rule{1.55cm}{2.7cm}}}
    \par\vspace{1pt}
    {\scriptsize\textbf{Women's Blouses Shirts}}
    \par\vspace{0pt}
    \begin{minipage}[t][1.1cm][t]{7.75cm}
      {\tiny\raggedright Cream chiffon blouse with dramatic open back featuring draped panel and crochet trim. Round neckline, cap sleeves, relaxed fit, and side vents. Elegant evening or dressy casual top.}
    \end{minipage}
  \end{minipage}%
\hfill%
  \begin{minipage}[t][4.1cm][t]{7.75cm}
    \makebox[7.75cm][l]{\includegraphics[width=1.55cm,height=2.7cm,keepaspectratio]{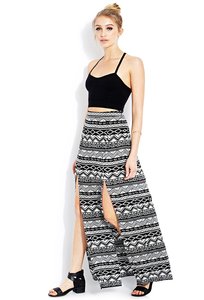}%
    \includegraphics[width=1.55cm,height=2.7cm,keepaspectratio]{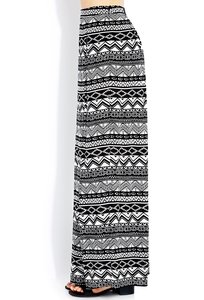}%
    \includegraphics[width=1.55cm,height=2.7cm,keepaspectratio]{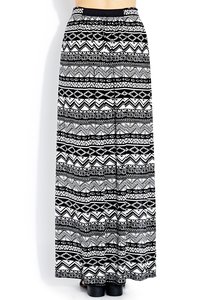}%
    \includegraphics[width=1.55cm,height=2.7cm,keepaspectratio]{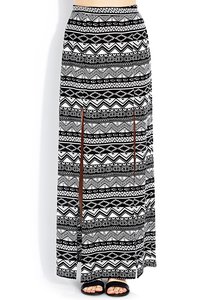}%
    \includegraphics[width=1.55cm,height=2.7cm,keepaspectratio]{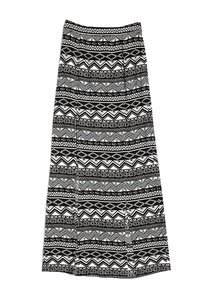}}
    \par\vspace{1pt}
    {\scriptsize\textbf{Women's Skirts}}
    \par\vspace{0pt}
    \begin{minipage}[t][1.1cm][t]{7.75cm}
      {\tiny\raggedright Black and white geometric tribal print maxi skirt with high-rise waistband, flowy A-line silhouette, and symmetrical dual front thigh-high slits. Crafted from lightweight fluid fabric with all-over ethnic zigzag and diamond patterns, ankle length, casual bohemian summer style.}
    \end{minipage}
  \end{minipage}%
\hfill
\par\vspace{8pt}

\noindent\hfill%
  \begin{minipage}[t][4.1cm][t]{7.75cm}
    \makebox[7.75cm][l]{\includegraphics[width=1.55cm,height=2.7cm,keepaspectratio]{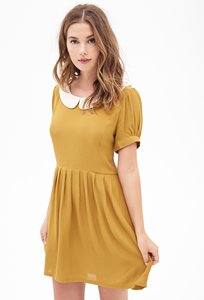}%
    \includegraphics[width=1.55cm,height=2.7cm,keepaspectratio]{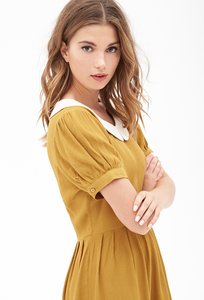}%
    \includegraphics[width=1.55cm,height=2.7cm,keepaspectratio]{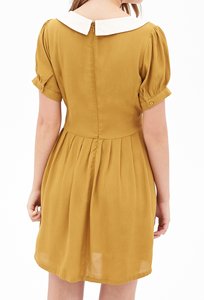}%
    \includegraphics[width=1.55cm,height=2.7cm,keepaspectratio]{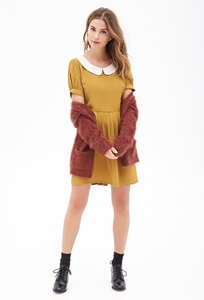}%
    \phantom{\rule{1.55cm}{2.7cm}}}
    \par\vspace{1pt}
    {\scriptsize\textbf{Women's Dresses}}
    \par\vspace{0pt}
    \begin{minipage}[t][1.1cm][t]{7.75cm}
      {\tiny\raggedright Mustard yellow short-sleeved dress with contrasting white Peter Pan collar, puff sleeves with functional button cuffs, fitted princess-seam bodice, and pleated above-knee skirt. Features back zipper closure and lightweight, semi-sheer textured fabric. Retro-inspired silhouette suitable for casual or semi-formal wear.}
    \end{minipage}
  \end{minipage}%
\hfill%
  \begin{minipage}[t][4.1cm][t]{7.75cm}
    \makebox[7.75cm][l]{\includegraphics[width=1.55cm,height=2.7cm,keepaspectratio]{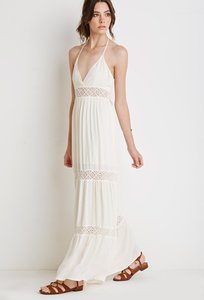}%
    \includegraphics[width=1.55cm,height=2.7cm,keepaspectratio]{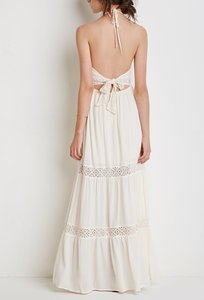}%
    \includegraphics[width=1.55cm,height=2.7cm,keepaspectratio]{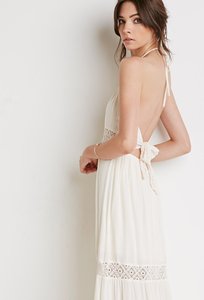}%
    \phantom{\rule{1.55cm}{2.7cm}}%
    \phantom{\rule{1.55cm}{2.7cm}}}
    \par\vspace{1pt}
    {\scriptsize\textbf{Women's Dresses}}
    \par\vspace{0pt}
    \begin{minipage}[t][1.1cm][t]{7.75cm}
      {\tiny\raggedright Cream halter maxi dress with deep V-neck, open back with tie details, three horizontal lace trim bands, empire waist, and flowy tiered skirt in lightweight fabric.}
    \end{minipage}
  \end{minipage}%
\hfill
\par\vspace{8pt}

\noindent\hfill%
  \begin{minipage}[t][4.1cm][t]{7.75cm}
    \makebox[7.75cm][l]{\includegraphics[width=1.55cm,height=2.7cm,keepaspectratio]{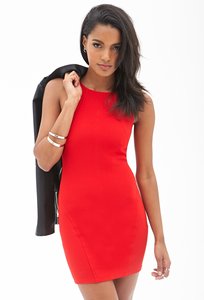}%
    \includegraphics[width=1.55cm,height=2.7cm,keepaspectratio]{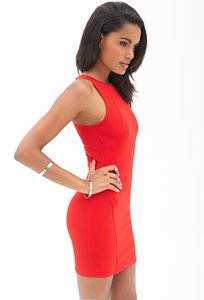}%
    \includegraphics[width=1.55cm,height=2.7cm,keepaspectratio]{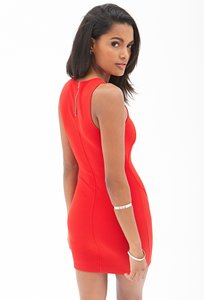}%
    \includegraphics[width=1.55cm,height=2.7cm,keepaspectratio]{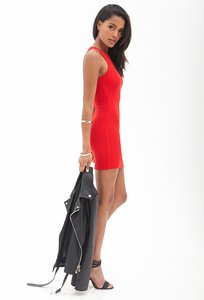}%
    \phantom{\rule{1.55cm}{2.7cm}}}
    \par\vspace{1pt}
    {\scriptsize\textbf{Women's Dresses}}
    \par\vspace{0pt}
    \begin{minipage}[t][1.1cm][t]{7.75cm}
      {\tiny\raggedright A vibrant red sleeveless bodycon dress featuring a round neckline, exposed back zipper closure, and strategic seaming for a fitted silhouette. The short-length garment is crafted from structured knit fabric with a sleek, minimalist design perfect for cocktail or evening wear.}
    \end{minipage}
  \end{minipage}%
\hfill%
  \begin{minipage}[t][4.1cm][t]{7.75cm}
    \makebox[7.75cm][l]{\includegraphics[width=1.55cm,height=2.7cm,keepaspectratio]{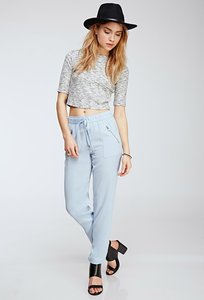}%
    \includegraphics[width=1.55cm,height=2.7cm,keepaspectratio]{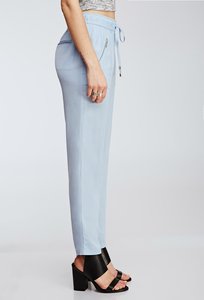}%
    \includegraphics[width=1.55cm,height=2.7cm,keepaspectratio]{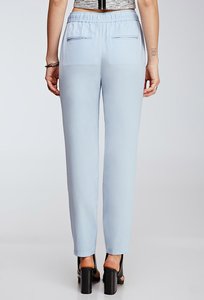}%
    \includegraphics[width=1.55cm,height=2.7cm,keepaspectratio]{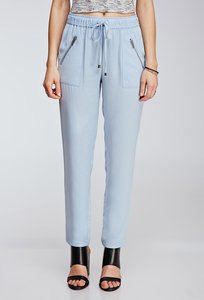}%
    \phantom{\rule{1.55cm}{2.7cm}}}
    \par\vspace{1pt}
    {\scriptsize\textbf{Women's Pants}}
    \par\vspace{0pt}
    \begin{minipage}[t][1.1cm][t]{7.75cm}
      {\tiny\raggedright Light blue relaxed-fit trousers featuring an elastic drawstring waist with metal aglets, diagonal zippered front pockets, and horizontal back pockets. These ankle-length straight-leg pants are crafted from lightweight woven fabric with a comfortable mid-rise fit and contemporary casual styling.}
    \end{minipage}
  \end{minipage}%
\hfill
\par\vspace{8pt}

\noindent\hfill%
  \begin{minipage}[t][4.1cm][t]{7.75cm}
    \makebox[7.75cm][l]{\includegraphics[width=1.55cm,height=2.7cm,keepaspectratio]{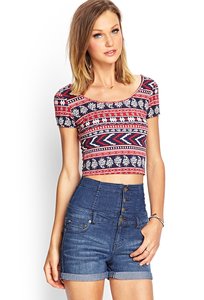}%
    \includegraphics[width=1.55cm,height=2.7cm,keepaspectratio]{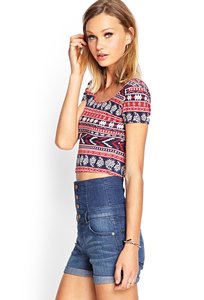}%
    \includegraphics[width=1.55cm,height=2.7cm,keepaspectratio]{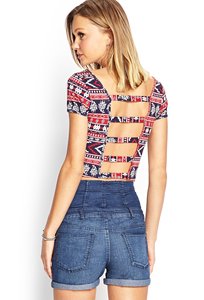}%
    \includegraphics[width=1.55cm,height=2.7cm,keepaspectratio]{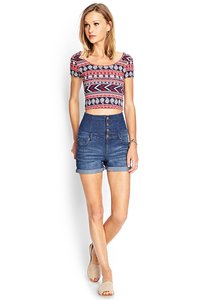}%
    \phantom{\rule{1.55cm}{2.7cm}}}
    \par\vspace{1pt}
    {\scriptsize\textbf{Women's Tees Tanks}}
    \par\vspace{0pt}
    \begin{minipage}[t][1.1cm][t]{7.75cm}
      {\tiny\raggedright A fitted short-sleeve crop top featuring a vibrant navy, red, and white tribal print with a distinctive ladder-back design of horizontal cut-out strips, scoop neckline, and bodycon silhouette in stretchy knit fabric.}
    \end{minipage}
  \end{minipage}%
\hfill%
  \begin{minipage}[t][4.1cm][t]{7.75cm}
    \makebox[7.75cm][l]{\includegraphics[width=1.55cm,height=2.7cm,keepaspectratio]{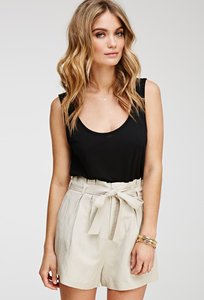}%
    \includegraphics[width=1.55cm,height=2.7cm,keepaspectratio]{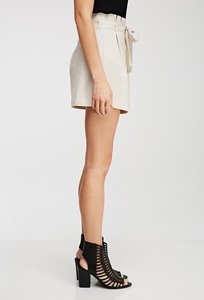}%
    \includegraphics[width=1.55cm,height=2.7cm,keepaspectratio]{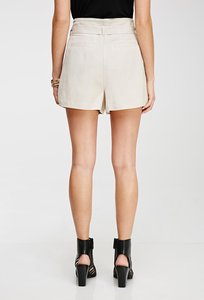}%
    \includegraphics[width=1.55cm,height=2.7cm,keepaspectratio]{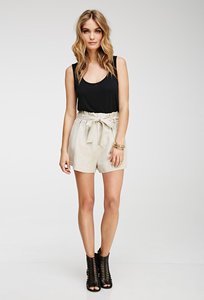}%
    \includegraphics[width=1.55cm,height=2.7cm,keepaspectratio]{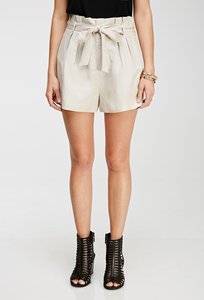}}
    \par\vspace{1pt}
    {\scriptsize\textbf{Women's Shorts}}
    \par\vspace{0pt}
    \begin{minipage}[t][1.1cm][t]{7.75cm}
      {\tiny\raggedright High-waisted paperbag shorts in cream linen-blend fabric featuring a gathered paperbag waist with self-tie belt, front pleats, and relaxed A-line silhouette. Includes side pockets and back welt pockets, with a mid-thigh length and clean straight hem for versatile summer styling.}
    \end{minipage}
  \end{minipage}%
\hfill
\par\vspace{8pt}

\noindent\hfill%
  \begin{minipage}[t][4.1cm][t]{7.75cm}
    \makebox[7.75cm][l]{\includegraphics[width=1.55cm,height=2.7cm,keepaspectratio]{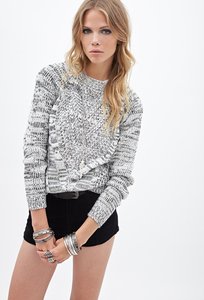}%
    \includegraphics[width=1.55cm,height=2.7cm,keepaspectratio]{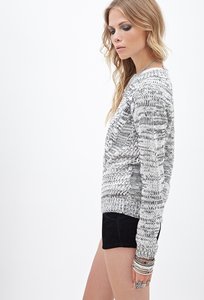}%
    \includegraphics[width=1.55cm,height=2.7cm,keepaspectratio]{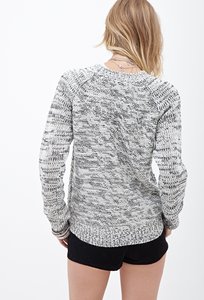}%
    \phantom{\rule{1.55cm}{2.7cm}}%
    \phantom{\rule{1.55cm}{2.7cm}}}
    \par\vspace{1pt}
    {\scriptsize\textbf{Women's Sweaters}}
    \par\vspace{0pt}
    \begin{minipage}[t][1.1cm][t]{7.75cm}
      {\tiny\raggedright Marled black and white cable-knit pullover sweater featuring textured cable front panel, plain stockinette back, and horizontally striped long sleeves. Classic crew neckline with ribbed trim, relaxed oversized fit, and chunky knit construction. Pullover style with no closures, pockets, or hardware.}
    \end{minipage}
  \end{minipage}%
\hfill%
  \begin{minipage}[t][4.1cm][t]{7.75cm}
    \makebox[7.75cm][l]{\includegraphics[width=1.55cm,height=2.7cm,keepaspectratio]{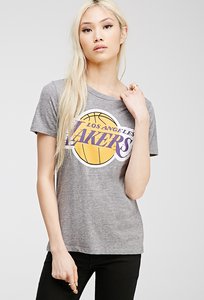}%
    \includegraphics[width=1.55cm,height=2.7cm,keepaspectratio]{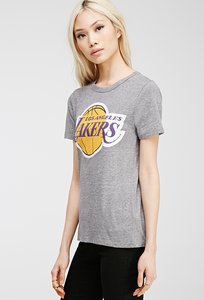}%
    \includegraphics[width=1.55cm,height=2.7cm,keepaspectratio]{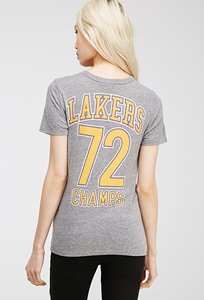}%
    \phantom{\rule{1.55cm}{2.7cm}}%
    \phantom{\rule{1.55cm}{2.7cm}}}
    \par\vspace{1pt}
    {\scriptsize\textbf{Women's Graphic Tees}}
    \par\vspace{0pt}
    \begin{minipage}[t][1.1cm][t]{7.75cm}
      {\tiny\raggedright This heather gray short-sleeve t-shirt features Los Angeles Lakers branding with a front basketball logo and back "LAKERS 72 CHAMPS" commemorative graphic. It has a classic crew neckline, relaxed casual fit, and soft cotton-blend jersey construction perfect for everyday fan wear.}
    \end{minipage}
  \end{minipage}%
\hfill
\par\vspace{8pt}

\twocolumn

% ================================================================
\clearpage
\onecolumn
\section{CIR Triplet Gallery}
\label{sec:triplet_gallery}

This section presents representative Composed Image Retrieval (CIR) triplets from the FashionMV dataset. Each row shows a source garment (left, up to five multi-view images), a short modification text above the arrow, and the corresponding target garment (right). Empty image slots indicate fewer than five available views.

\noindent\hfill%
  \begin{minipage}[t]{6.0cm}
    \noindent\hbox to 6.0cm{\includegraphics[width=1.20cm,height=2.0cm,keepaspectratio]{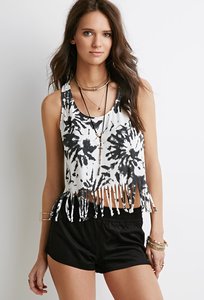}\hss\includegraphics[width=1.20cm,height=2.0cm,keepaspectratio]{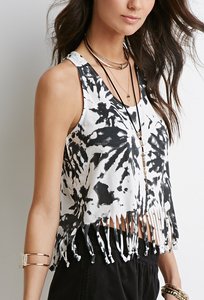}\hss\includegraphics[width=1.20cm,height=2.0cm,keepaspectratio]{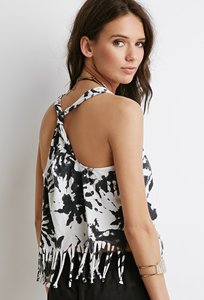}\hss\phantom{\rule{1.20cm}{2.0cm}}\hss\phantom{\rule{1.20cm}{2.0cm}}\hss}
    \par\vspace{1pt}
    {\scriptsize\textbf{}}
    \par\vspace{0pt}
    \parbox[t]{6.0cm}{\tiny\raggedright A sleeveless cropped tank top in black and white tie-dye with a distinctive fringe hem and twisted back strap detail. Features a scoop neckline, racerback-inspired armholes, and lightweight jersey fabric. Bohemian festival style with knotted tassel ends.}
  \end{minipage}%
\hfill%
  \begin{minipage}[t]{3.5cm}
    \centering
    \vspace{0.45cm}%
    \parbox{3.5cm}{\centering\tiny Switch to solid white, replace the back racerback knot with shoulder cut-outs and ties, and add vertical fringe trim down the side seams.}\\[2pt]
    {\normalsize$\longrightarrow$}
  \end{minipage}%
\hfill%
  \begin{minipage}[t]{6.0cm}
    \noindent\hbox to 6.0cm{\includegraphics[width=1.20cm,height=2.0cm,keepaspectratio]{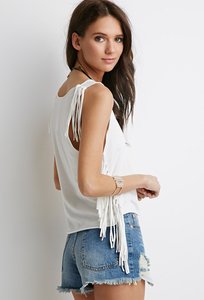}\hss\includegraphics[width=1.20cm,height=2.0cm,keepaspectratio]{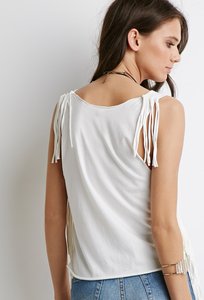}\hss\includegraphics[width=1.20cm,height=2.0cm,keepaspectratio]{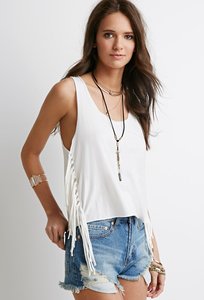}\hss\phantom{\rule{1.20cm}{2.0cm}}\hss\phantom{\rule{1.20cm}{2.0cm}}\hss}
    \par\vspace{1pt}
    {\scriptsize\textbf{}}
    \par\vspace{0pt}
    \parbox[t]{6.0cm}{\tiny\raggedright White sleeveless tank top featuring scoop neckline, open shoulder cut-outs with knotted fringe ties, and long side fringe trim. Relaxed, flowy fit in lightweight jersey fabric. Bohemian festival style with wide armholes and cropped length.}
  \end{minipage}%
\hfill
\par\vspace{2pt}

\noindent\hfill%
  \begin{minipage}[t]{6.0cm}
    \noindent\hbox to 6.0cm{\includegraphics[width=1.20cm,height=2.0cm,keepaspectratio]{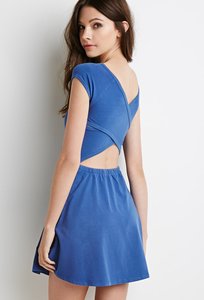}\hss\includegraphics[width=1.20cm,height=2.0cm,keepaspectratio]{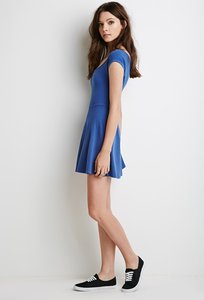}\hss\includegraphics[width=1.20cm,height=2.0cm,keepaspectratio]{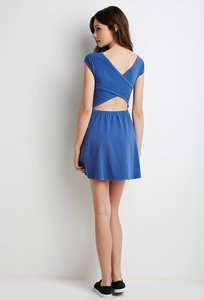}\hss\includegraphics[width=1.20cm,height=2.0cm,keepaspectratio]{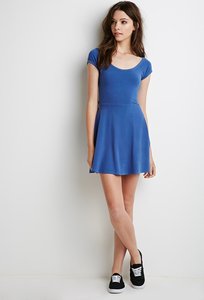}\hss\phantom{\rule{1.20cm}{2.0cm}}\hss}
    \par\vspace{1pt}
    {\scriptsize\textbf{}}
    \par\vspace{0pt}
    \parbox[t]{6.0cm}{\tiny\raggedright Royal blue short-sleeve fit-and-flare dress featuring a scoop neckline, cap sleeves, and distinctive crisscross open-back design with a bold waist cut-out. Crafted from soft jersey knit fabric with a flared A-line skirt falling above the knee.}
  \end{minipage}%
\hfill%
  \begin{minipage}[t]{3.5cm}
    \centering
    \vspace{0.45cm}%
    \parbox{3.5cm}{\centering\tiny Change to a red woven dress with a square neckline and wide straps; replace the open crisscross back with a full back panel featuring an exposed gold zipper.}\\[2pt]
    {\normalsize$\longrightarrow$}
  \end{minipage}%
\hfill%
  \begin{minipage}[t]{6.0cm}
    \noindent\hbox to 6.0cm{\includegraphics[width=1.20cm,height=2.0cm,keepaspectratio]{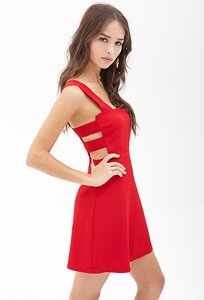}\hss\includegraphics[width=1.20cm,height=2.0cm,keepaspectratio]{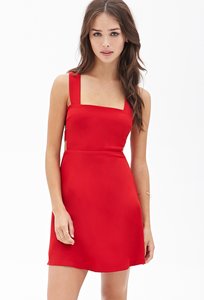}\hss\includegraphics[width=1.20cm,height=2.0cm,keepaspectratio]{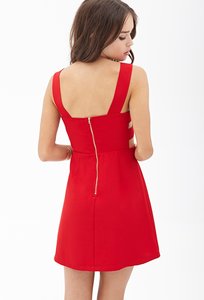}\hss\includegraphics[width=1.20cm,height=2.0cm,keepaspectratio]{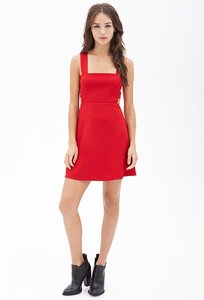}\hss\phantom{\rule{1.20cm}{2.0cm}}\hss}
    \par\vspace{1pt}
    {\scriptsize\textbf{}}
    \par\vspace{0pt}
    \parbox[t]{6.0cm}{\tiny\raggedright Vibrant red sleeveless fit-and-flare mini dress featuring a square neckline, wide shoulder straps, symmetrical side cut-outs at the waist, and a prominent exposed gold back zipper. Crafted from structured woven fabric with a fitted bodice and flared A-line skirt falling above the knee.}
  \end{minipage}%
\hfill
\par\vspace{2pt}

\noindent\hfill%
  \begin{minipage}[t]{6.0cm}
    \noindent\hbox to 6.0cm{\includegraphics[width=1.20cm,height=2.0cm,keepaspectratio]{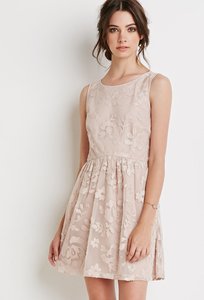}\hss\includegraphics[width=1.20cm,height=2.0cm,keepaspectratio]{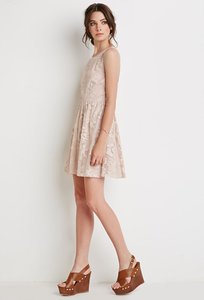}\hss\includegraphics[width=1.20cm,height=2.0cm,keepaspectratio]{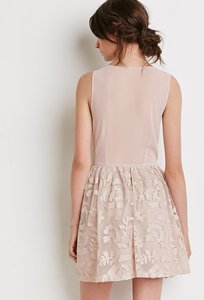}\hss\phantom{\rule{1.20cm}{2.0cm}}\hss\phantom{\rule{1.20cm}{2.0cm}}\hss}
    \par\vspace{1pt}
    {\scriptsize\textbf{}}
    \par\vspace{0pt}
    \parbox[t]{6.0cm}{\tiny\raggedright Blush pink sleeveless fit-and-flare mini dress featuring tonal floral embroidery on sheer organza overlay, with a distinctive solid back bodice contrasting the sheer front, round neckline, gathered waist seam, and flared A-line skirt falling above the knee.}
  \end{minipage}%
\hfill%
  \begin{minipage}[t]{3.5cm}
    \centering
    \vspace{0.45cm}%
    \parbox{3.5cm}{\centering\tiny Change blush organza floral appliqué to all-over white lace with an illusion yoke front; add princess seams and a buttoned keyhole to the back.}\\[2pt]
    {\normalsize$\longrightarrow$}
  \end{minipage}%
\hfill%
  \begin{minipage}[t]{6.0cm}
    \noindent\hbox to 6.0cm{\includegraphics[width=1.20cm,height=2.0cm,keepaspectratio]{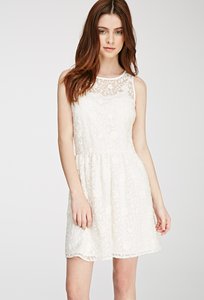}\hss\includegraphics[width=1.20cm,height=2.0cm,keepaspectratio]{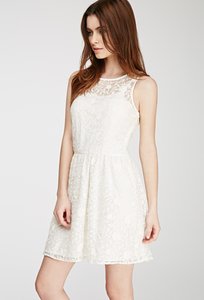}\hss\includegraphics[width=1.20cm,height=2.0cm,keepaspectratio]{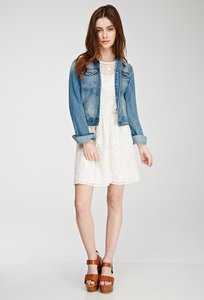}\hss\includegraphics[width=1.20cm,height=2.0cm,keepaspectratio]{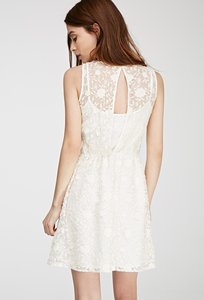}\hss\phantom{\rule{1.20cm}{2.0cm}}\hss}
    \par\vspace{1pt}
    {\scriptsize\textbf{}}
    \par\vspace{0pt}
    \parbox[t]{6.0cm}{\tiny\raggedright A sleeveless white floral lace fit-and-flare dress featuring a sheer illusion neckline, keyhole back with button closure, and scalloped hem. The fully lined design offers a flattering silhouette with princess-seamed bodice and flared skirt, perfect for spring and summer occasions.}
  \end{minipage}%
\hfill
\par\vspace{2pt}

\noindent\hfill%
  \begin{minipage}[t]{6.0cm}
    \noindent\hbox to 6.0cm{\includegraphics[width=1.20cm,height=2.0cm,keepaspectratio]{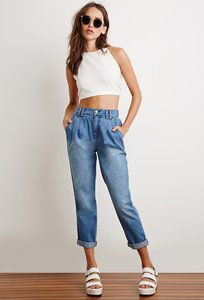}\hss\includegraphics[width=1.20cm,height=2.0cm,keepaspectratio]{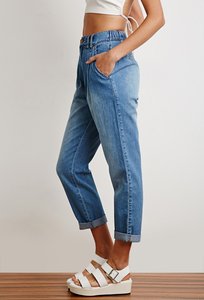}\hss\includegraphics[width=1.20cm,height=2.0cm,keepaspectratio]{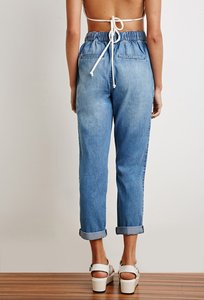}\hss\includegraphics[width=1.20cm,height=2.0cm,keepaspectratio]{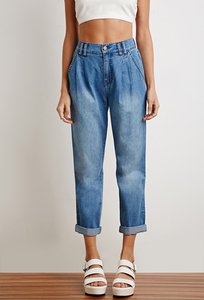}\hss\phantom{\rule{1.20cm}{2.0cm}}\hss}
    \par\vspace{1pt}
    {\scriptsize\textbf{}}
    \par\vspace{0pt}
    \parbox[t]{6.0cm}{\tiny\raggedright High-waisted pleated denim pants in a light blue vintage wash featuring a relaxed tapered fit, distinctive front waist pleats, comfortable elasticized back waistband, functional side pockets, and casually rolled cuffs for a casual yet polished aesthetic.}
  \end{minipage}%
\hfill%
  \begin{minipage}[t]{3.5cm}
    \centering
    \vspace{0.45cm}%
    \parbox{3.5cm}{\centering\tiny Front: Add heavy, asymmetrical shredded holes and knee blowouts; Back: Maintain a clean, undistressed rear panel with the original V-shaped yoke and patch pocket construction.}\\[2pt]
    {\normalsize$\longrightarrow$}
  \end{minipage}%
\hfill%
  \begin{minipage}[t]{6.0cm}
    \noindent\hbox to 6.0cm{\includegraphics[width=1.20cm,height=2.0cm,keepaspectratio]{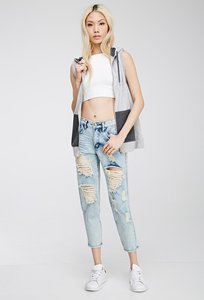}\hss\includegraphics[width=1.20cm,height=2.0cm,keepaspectratio]{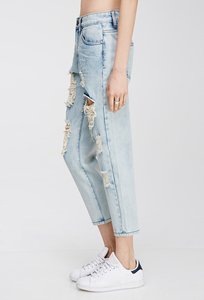}\hss\includegraphics[width=1.20cm,height=2.0cm,keepaspectratio]{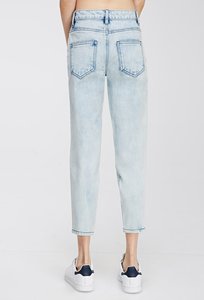}\hss\includegraphics[width=1.20cm,height=2.0cm,keepaspectratio]{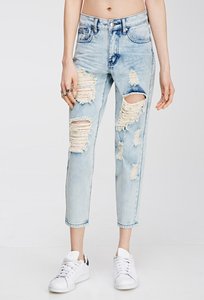}\hss\phantom{\rule{1.20cm}{2.0cm}}\hss}
    \par\vspace{1pt}
    {\scriptsize\textbf{}}
    \par\vspace{0pt}
    \parbox[t]{6.0cm}{\tiny\raggedright Light blue acid-wash cropped jeans with heavy asymmetrical distressing, featuring dual thigh rips on one leg and a prominent knee blowout on the other. Relaxed boyfriend fit with mid-rise waist, classic five-pocket styling, golden contrast stitching, and clean cropped hem. Medium-weight cotton denim with frayed destroyed details.}
  \end{minipage}%
\hfill
\par\vspace{2pt}

\noindent\hfill%
  \begin{minipage}[t]{6.0cm}
    \noindent\hbox to 6.0cm{\includegraphics[width=1.20cm,height=2.0cm,keepaspectratio]{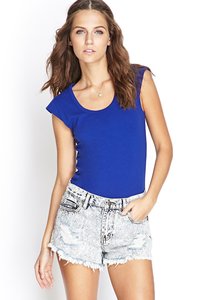}\hss\includegraphics[width=1.20cm,height=2.0cm,keepaspectratio]{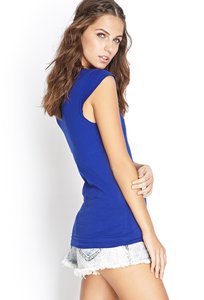}\hss\includegraphics[width=1.20cm,height=2.0cm,keepaspectratio]{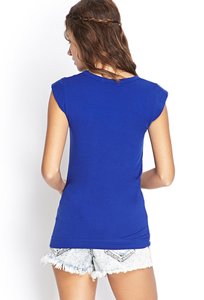}\hss\includegraphics[width=1.20cm,height=2.0cm,keepaspectratio]{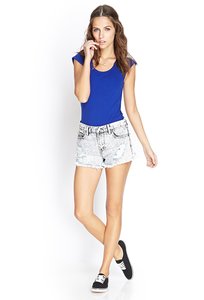}\hss\phantom{\rule{1.20cm}{2.0cm}}\hss}
    \par\vspace{1pt}
    {\scriptsize\textbf{}}
    \par\vspace{0pt}
    \parbox[t]{6.0cm}{\tiny\raggedright Royal blue cap-sleeve top featuring a scoop neckline front and back with form-fitting silhouette. Crafted from smooth jersey knit with extended shoulder coverage for minimal arm exposure. Minimalist design without logos, pockets, or embellishments. Versatile summer basic ideal for tucking into high-waisted bottoms.}
  \end{minipage}%
\hfill%
  \begin{minipage}[t]{3.5cm}
    \centering
    \vspace{0.45cm}%
    \parbox{3.5cm}{\centering\tiny Change to a high mock neckline and sleeveless construction; add an asymmetrical vertical slit to the wearer's right side hem only.}\\[2pt]
    {\normalsize$\longrightarrow$}
  \end{minipage}%
\hfill%
  \begin{minipage}[t]{6.0cm}
    \noindent\hbox to 6.0cm{\includegraphics[width=1.20cm,height=2.0cm,keepaspectratio]{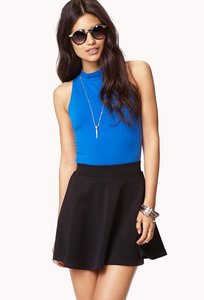}\hss\includegraphics[width=1.20cm,height=2.0cm,keepaspectratio]{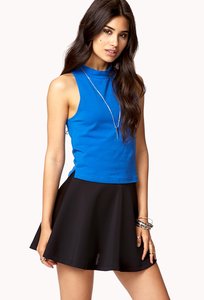}\hss\includegraphics[width=1.20cm,height=2.0cm,keepaspectratio]{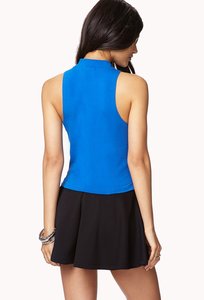}\hss\includegraphics[width=1.20cm,height=2.0cm,keepaspectratio]{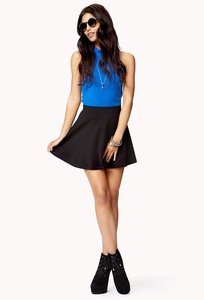}\hss\phantom{\rule{1.20cm}{2.0cm}}\hss}
    \par\vspace{1pt}
    {\scriptsize\textbf{}}
    \par\vspace{0pt}
    \parbox[t]{6.0cm}{\tiny\raggedright Royal blue sleeveless mock neck tank top featuring a fitted bodycon silhouette, distinctive asymmetrical side slit on the wearer's right side, and clean minimalist aesthetic. Crafted from soft stretch knit fabric with wide armholes, high neckline, and cropped waist length for versatile styling.}
  \end{minipage}%
\hfill
\par\vspace{2pt}

\noindent\hfill%
  \begin{minipage}[t]{6.0cm}
    \noindent\hbox to 6.0cm{\includegraphics[width=1.20cm,height=2.0cm,keepaspectratio]{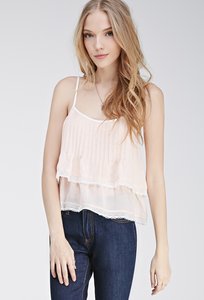}\hss\includegraphics[width=1.20cm,height=2.0cm,keepaspectratio]{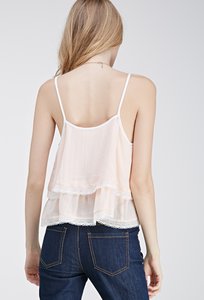}\hss\includegraphics[width=1.20cm,height=2.0cm,keepaspectratio]{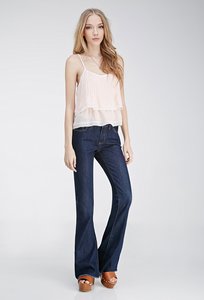}\hss\includegraphics[width=1.20cm,height=2.0cm,keepaspectratio]{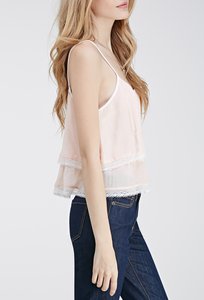}\hss\phantom{\rule{1.20cm}{2.0cm}}\hss}
    \par\vspace{1pt}
    {\scriptsize\textbf{}}
    \par\vspace{0pt}
    \parbox[t]{6.0cm}{\tiny\raggedright A blush pink tiered camisole top featuring delicate white lace trim, thin spaghetti straps, and vertical pintuck pleating on the bodice. Crafted from semi-sheer lightweight fabric with a relaxed cropped fit and scalloped lace hem. Romantic feminine style perfect for spring and summer casual wear.}
  \end{minipage}%
\hfill%
  \begin{minipage}[t]{3.5cm}
    \centering
    \vspace{0.45cm}%
    \parbox{3.5cm}{\centering\tiny Change from solid blush pink with pintuck pleats to a floral print with a flat bodice, and replace simple back straps with a crisscross X-back design.}\\[2pt]
    {\normalsize$\longrightarrow$}
  \end{minipage}%
\hfill%
  \begin{minipage}[t]{6.0cm}
    \noindent\hbox to 6.0cm{\includegraphics[width=1.20cm,height=2.0cm,keepaspectratio]{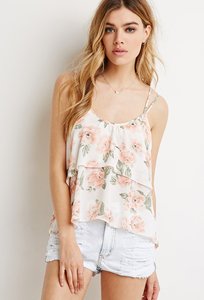}\hss\includegraphics[width=1.20cm,height=2.0cm,keepaspectratio]{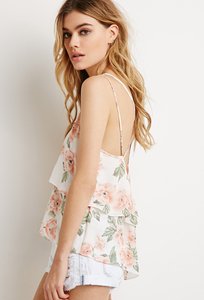}\hss\includegraphics[width=1.20cm,height=2.0cm,keepaspectratio]{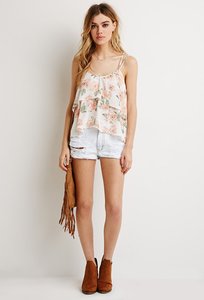}\hss\includegraphics[width=1.20cm,height=2.0cm,keepaspectratio]{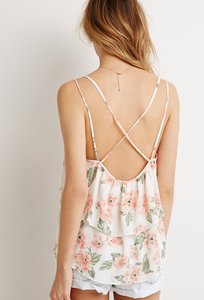}\hss\phantom{\rule{1.20cm}{2.0cm}}\hss}
    \par\vspace{1pt}
    {\scriptsize\textbf{}}
    \par\vspace{0pt}
    \parbox[t]{6.0cm}{\tiny\raggedright Women's sleeveless camisole top featuring a white base with peach floral print and sage green leaves. Designed with tiered ruffle layers, scoop neckline front, and dramatic crisscross back straps. Crafted from lightweight flowy fabric in a relaxed fit that hits at the hip. Pullover style with no closures.}
  \end{minipage}%
\hfill
\par\vspace{2pt}

\noindent\hfill%
  \begin{minipage}[t]{6.0cm}
    \noindent\hbox to 6.0cm{\includegraphics[width=1.20cm,height=2.0cm,keepaspectratio]{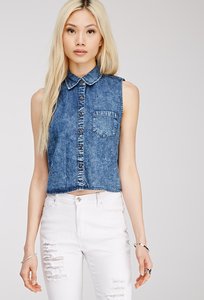}\hss\includegraphics[width=1.20cm,height=2.0cm,keepaspectratio]{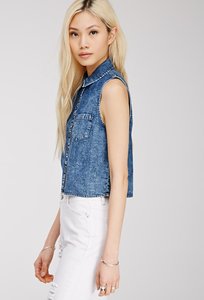}\hss\includegraphics[width=1.20cm,height=2.0cm,keepaspectratio]{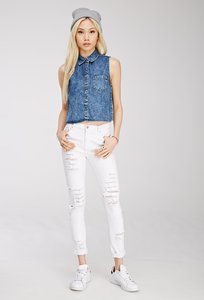}\hss\includegraphics[width=1.20cm,height=2.0cm,keepaspectratio]{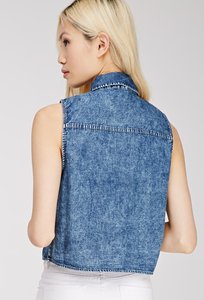}\hss\phantom{\rule{1.20cm}{2.0cm}}\hss}
    \par\vspace{1pt}
    {\scriptsize\textbf{}}
    \par\vspace{0pt}
    \parbox[t]{6.0cm}{\tiny\raggedright Sleeveless acid wash denim crop top with pointed collar, six-button front closure, single left chest pocket, and frayed raw hem. Features contrast white stitching and a boxy, cropped silhouette. Medium blue vintage-wash cotton denim construction.}
  \end{minipage}%
\hfill%
  \begin{minipage}[t]{3.5cm}
    \centering
    \vspace{0.45cm}%
    \parbox{3.5cm}{\centering\tiny Swap the acid-wash denim for blue/white/red plaid. Replace the single pocket and raw hem with dual button-flap pockets and a front self-tie knot.}\\[2pt]
    {\normalsize$\longrightarrow$}
  \end{minipage}%
\hfill%
  \begin{minipage}[t]{6.0cm}
    \noindent\hbox to 6.0cm{\includegraphics[width=1.20cm,height=2.0cm,keepaspectratio]{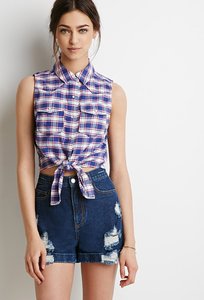}\hss\includegraphics[width=1.20cm,height=2.0cm,keepaspectratio]{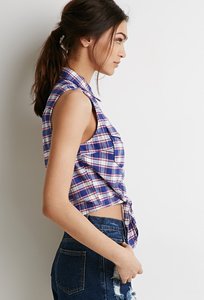}\hss\includegraphics[width=1.20cm,height=2.0cm,keepaspectratio]{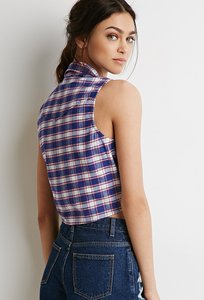}\hss\phantom{\rule{1.20cm}{2.0cm}}\hss\phantom{\rule{1.20cm}{2.0cm}}\hss}
    \par\vspace{1pt}
    {\scriptsize\textbf{}}
    \par\vspace{0pt}
    \parbox[t]{6.0cm}{\tiny\raggedright Sleeveless tie-front plaid shirt featuring a blue, white, and red check pattern. Designed with a pointed collar, full button-front closure, and two symmetrical chest pockets with buttoned flaps. The cropped hem includes a self-tie waist for adjustable fit. Crafted from lightweight woven cotton-blend fabric, this casual summer top pairs perfectly with high-waisted bottoms.}
  \end{minipage}%
\hfill
\par\vspace{2pt}

\noindent\hfill%
  \begin{minipage}[t]{6.0cm}
    \noindent\hbox to 6.0cm{\includegraphics[width=1.20cm,height=2.0cm,keepaspectratio]{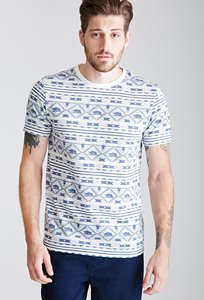}\hss\includegraphics[width=1.20cm,height=2.0cm,keepaspectratio]{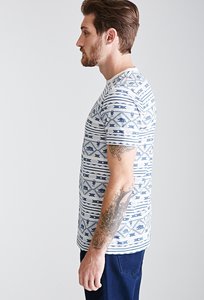}\hss\includegraphics[width=1.20cm,height=2.0cm,keepaspectratio]{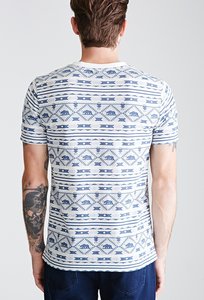}\hss\includegraphics[width=1.20cm,height=2.0cm,keepaspectratio]{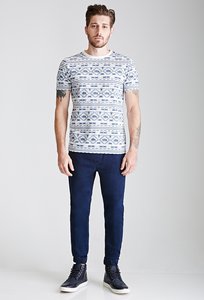}\hss\includegraphics[width=1.20cm,height=2.0cm,keepaspectratio]{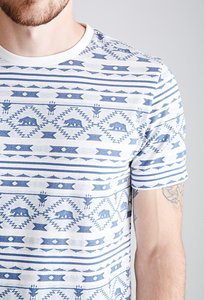}\hss}
    \par\vspace{1pt}
    {\scriptsize\textbf{}}
    \par\vspace{0pt}
    \parbox[t]{6.0cm}{\tiny\raggedright Men's casual white short-sleeve crew neck t-shirt featuring an all-over blue geometric Southwestern print with bear and deer motifs, regular fit, hip length.}
  \end{minipage}%
\hfill%
  \begin{minipage}[t]{3.5cm}
    \centering
    \vspace{0.45cm}%
    \parbox{3.5cm}{\centering\tiny Remove all-over print; switch to solid white base. Add a navy paisley chest pocket on the left and navy paisley interior linings to the sleeve cuffs.}\\[2pt]
    {\normalsize$\longrightarrow$}
  \end{minipage}%
\hfill%
  \begin{minipage}[t]{6.0cm}
    \noindent\hbox to 6.0cm{\includegraphics[width=1.20cm,height=2.0cm,keepaspectratio]{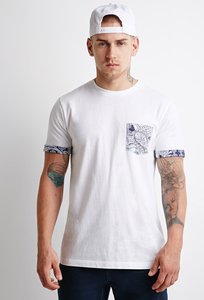}\hss\includegraphics[width=1.20cm,height=2.0cm,keepaspectratio]{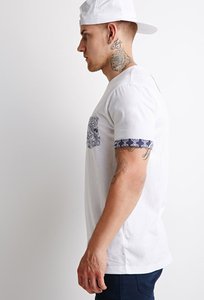}\hss\includegraphics[width=1.20cm,height=2.0cm,keepaspectratio]{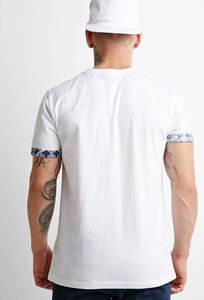}\hss\includegraphics[width=1.20cm,height=2.0cm,keepaspectratio]{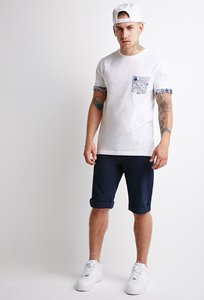}\hss\includegraphics[width=1.20cm,height=2.0cm,keepaspectratio]{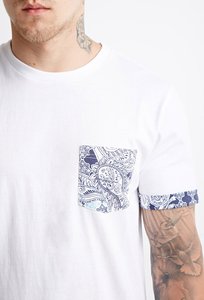}\hss}
    \par\vspace{1pt}
    {\scriptsize\textbf{}}
    \par\vspace{0pt}
    \parbox[t]{6.0cm}{\tiny\raggedright White crew neck t-shirt with navy paisley bandana print chest pocket and matching rolled sleeve cuffs. Regular fit short sleeve cotton tee featuring asymmetrical pocket placement on wearer's left chest. Casual style with western-inspired detailing and heathered fabric texture.}
  \end{minipage}%
\hfill
\par\vspace{2pt}

\noindent\hfill%
  \begin{minipage}[t]{6.0cm}
    \noindent\hbox to 6.0cm{\includegraphics[width=1.20cm,height=2.0cm,keepaspectratio]{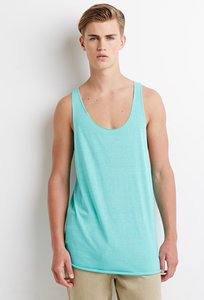}\hss\includegraphics[width=1.20cm,height=2.0cm,keepaspectratio]{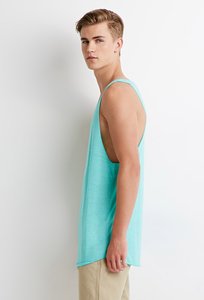}\hss\includegraphics[width=1.20cm,height=2.0cm,keepaspectratio]{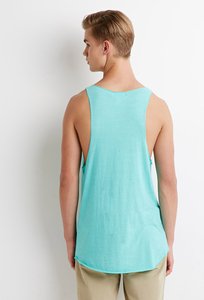}\hss\phantom{\rule{1.20cm}{2.0cm}}\hss\phantom{\rule{1.20cm}{2.0cm}}\hss}
    \par\vspace{1pt}
    {\scriptsize\textbf{}}
    \par\vspace{0pt}
    \parbox[t]{6.0cm}{\tiny\raggedright A men's relaxed-fit tank top in heathered turquoise featuring deep scoop necklines front and back, dramatically dropped armholes for a breezy aesthetic, and a high-low hem with extended back length. Crafted from lightweight jersey fabric with a casual, minimalist silhouette perfect for warm weather layering or athletic wear.}
  \end{minipage}%
\hfill%
  \begin{minipage}[t]{3.5cm}
    \centering
    \vspace{0.45cm}%
    \parbox{3.5cm}{\centering\tiny Switch to an optic white longline tank with side slits; replace the source's drop-tail curved hem with a straight, elongated hem and functional side vents.}\\[2pt]
    {\normalsize$\longrightarrow$}
  \end{minipage}%
\hfill%
  \begin{minipage}[t]{6.0cm}
    \noindent\hbox to 6.0cm{\includegraphics[width=1.20cm,height=2.0cm,keepaspectratio]{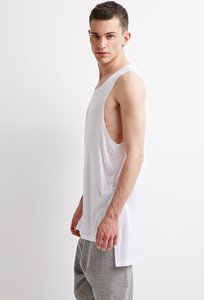}\hss\includegraphics[width=1.20cm,height=2.0cm,keepaspectratio]{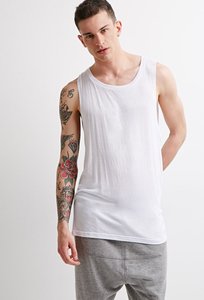}\hss\includegraphics[width=1.20cm,height=2.0cm,keepaspectratio]{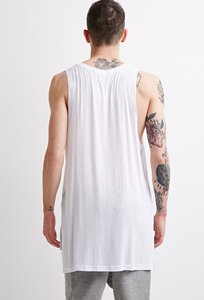}\hss\includegraphics[width=1.20cm,height=2.0cm,keepaspectratio]{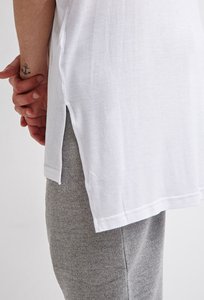}\hss\phantom{\rule{1.20cm}{2.0cm}}\hss}
    \par\vspace{1pt}
    {\scriptsize\textbf{}}
    \par\vspace{0pt}
    \parbox[t]{6.0cm}{\tiny\raggedright White longline muscle tank top with deep armholes, scoop neckline front and back, and side hem slits. Relaxed fit in lightweight cotton-blend jersey, perfect for casual or athletic wear.}
  \end{minipage}%
\hfill
\par\vspace{2pt}

\noindent\hfill%
  \begin{minipage}[t]{6.0cm}
    \noindent\hbox to 6.0cm{\includegraphics[width=1.20cm,height=2.0cm,keepaspectratio]{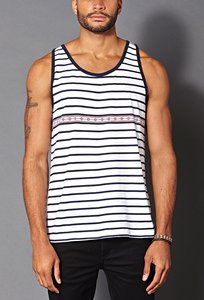}\hss\includegraphics[width=1.20cm,height=2.0cm,keepaspectratio]{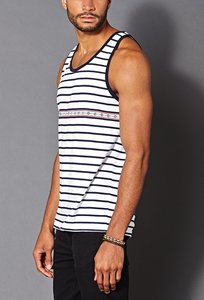}\hss\includegraphics[width=1.20cm,height=2.0cm,keepaspectratio]{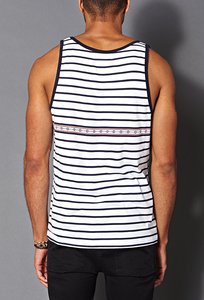}\hss\includegraphics[width=1.20cm,height=2.0cm,keepaspectratio]{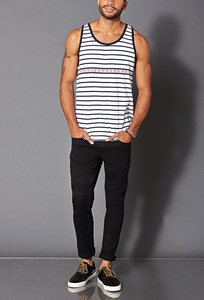}\hss\includegraphics[width=1.20cm,height=2.0cm,keepaspectratio]{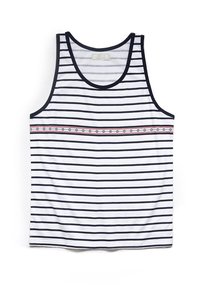}\hss}
    \par\vspace{1pt}
    {\scriptsize\textbf{}}
    \par\vspace{0pt}
    \parbox[t]{6.0cm}{\tiny\raggedright Men's casual sleeveless tank top featuring navy horizontal stripes on a white base with a distinctive decorative geometric chest band in red and navy. Designed with a scoop neckline, navy trim on neck and armholes, relaxed fit, and hip-length hem. Crafted from soft jersey knit fabric, this versatile summer garment offers breathability and comfort for everyday wear.}
  \end{minipage}%
\hfill%
  \begin{minipage}[t]{3.5cm}
    \centering
    \vspace{0.45cm}%
    \parbox{3.5cm}{\centering\tiny Transform the striped source into a tri-color red/white/navy block tank, adding a navy chest pocket and solid-colored panels replacing the original stripe pattern.}\\[2pt]
    {\normalsize$\longrightarrow$}
  \end{minipage}%
\hfill%
  \begin{minipage}[t]{6.0cm}
    \noindent\hbox to 6.0cm{\includegraphics[width=1.20cm,height=2.0cm,keepaspectratio]{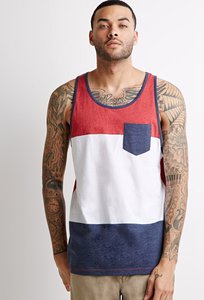}\hss\includegraphics[width=1.20cm,height=2.0cm,keepaspectratio]{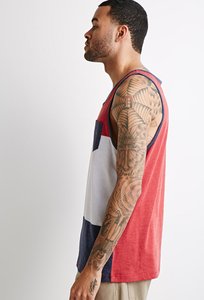}\hss\includegraphics[width=1.20cm,height=2.0cm,keepaspectratio]{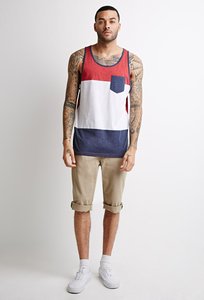}\hss\includegraphics[width=1.20cm,height=2.0cm,keepaspectratio]{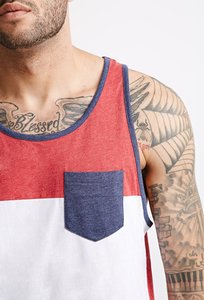}\hss\phantom{\rule{1.20cm}{2.0cm}}\hss}
    \par\vspace{1pt}
    {\scriptsize\textbf{}}
    \par\vspace{0pt}
    \parbox[t]{6.0cm}{\tiny\raggedright Men's sleeveless color-block tank top with horizontal heather red, white, and navy panels. Features scoop neckline with navy binding, deep armholes, and single chest pocket on wearer's left side. Made from soft heathered cotton jersey with relaxed fit and contrast hem stitching. Casual athletic style perfect for summer wear.}
  \end{minipage}%
\hfill
\par\vspace{2pt}

\noindent\hfill%
  \begin{minipage}[t]{6.0cm}
    \noindent\hbox to 6.0cm{\includegraphics[width=1.20cm,height=2.0cm,keepaspectratio]{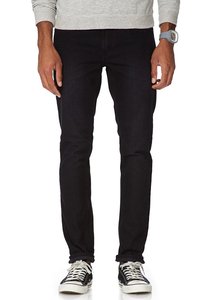}\hss\includegraphics[width=1.20cm,height=2.0cm,keepaspectratio]{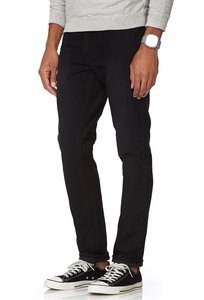}\hss\includegraphics[width=1.20cm,height=2.0cm,keepaspectratio]{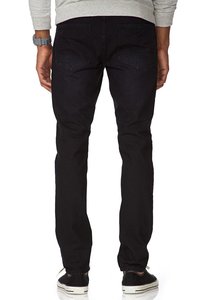}\hss\includegraphics[width=1.20cm,height=2.0cm,keepaspectratio]{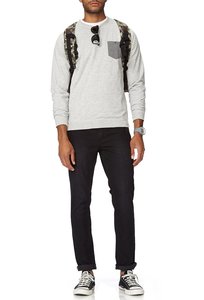}\hss\includegraphics[width=1.20cm,height=2.0cm,keepaspectratio]{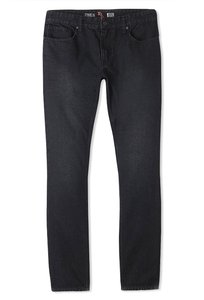}\hss}
    \par\vspace{1pt}
    {\scriptsize\textbf{}}
    \par\vspace{0pt}
    \parbox[t]{6.0cm}{\tiny\raggedright Men's slim straight-leg black denim jeans with classic five-pocket styling, button fly closure, and mid-rise waist. Features tonal stitching, slight whiskering detail, branded waistband label, and asymmetrical coin pocket placement on wearer's right side only.}
  \end{minipage}%
\hfill%
  \begin{minipage}[t]{3.5cm}
    \centering
    \vspace{0.45cm}%
    \parbox{3.5cm}{\centering\tiny Add asymmetrical front distressing: horizontal thigh rips on the right leg and a large frayed knee blowout on the left, while keeping the back clean and uniform.}\\[2pt]
    {\normalsize$\longrightarrow$}
  \end{minipage}%
\hfill%
  \begin{minipage}[t]{6.0cm}
    \noindent\hbox to 6.0cm{\includegraphics[width=1.20cm,height=2.0cm,keepaspectratio]{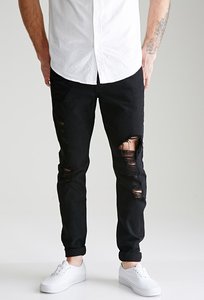}\hss\includegraphics[width=1.20cm,height=2.0cm,keepaspectratio]{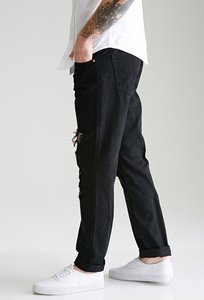}\hss\includegraphics[width=1.20cm,height=2.0cm,keepaspectratio]{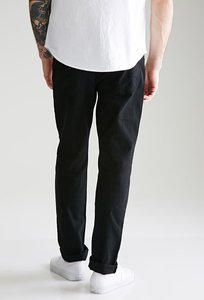}\hss\includegraphics[width=1.20cm,height=2.0cm,keepaspectratio]{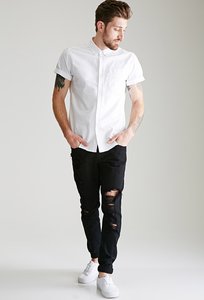}\hss\phantom{\rule{1.20cm}{2.0cm}}\hss}
    \par\vspace{1pt}
    {\scriptsize\textbf{}}
    \par\vspace{0pt}
    \parbox[t]{6.0cm}{\tiny\raggedright Men's black slim-fit denim jeans featuring asymmetrical heavy distressing with multiple thigh rips on the right leg and a large knee blowout on the left, cuffed hems, five-pocket styling with copper hardware, and a tapered silhouette. Casual streetwear aesthetic.}
  \end{minipage}%
\hfill
\par\vspace{2pt}

\noindent\hfill%
  \begin{minipage}[t]{6.0cm}
    \noindent\hbox to 6.0cm{\includegraphics[width=1.20cm,height=2.0cm,keepaspectratio]{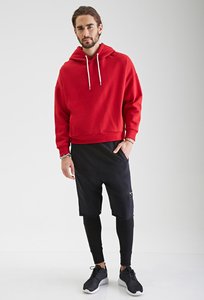}\hss\includegraphics[width=1.20cm,height=2.0cm,keepaspectratio]{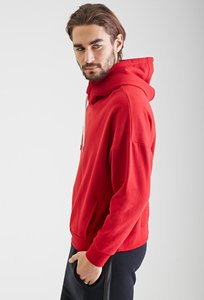}\hss\includegraphics[width=1.20cm,height=2.0cm,keepaspectratio]{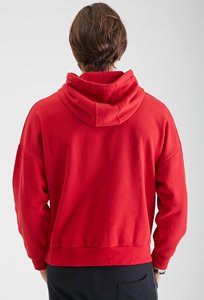}\hss\includegraphics[width=1.20cm,height=2.0cm,keepaspectratio]{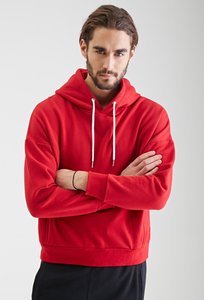}\hss\includegraphics[width=1.20cm,height=2.0cm,keepaspectratio]{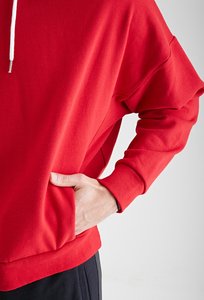}\hss}
    \par\vspace{1pt}
    {\scriptsize\textbf{}}
    \par\vspace{0pt}
    \parbox[t]{6.0cm}{\tiny\raggedright Men's vibrant red pullover hoodie featuring a relaxed boxy fit, dropped shoulders, and crossover hood with white contrast drawstrings and metal tips. Includes kangaroo front pocket, ribbed cuffs and hem band, and horizontal sleeve panel details. Clean, logo-free design crafted from soft medium-weight fleece fabric, ideal for casual streetwear.}
  \end{minipage}%
\hfill%
  \begin{minipage}[t]{3.5cm}
    \centering
    \vspace{0.45cm}%
    \parbox{3.5cm}{\centering\tiny Front: Change color to white and shorten sleeves with rolled cuffs. Side: Add an asymmetrical gold zipper vent to the wearer's left hem.}\\[2pt]
    {\normalsize$\longrightarrow$}
  \end{minipage}%
\hfill%
  \begin{minipage}[t]{6.0cm}
    \noindent\hbox to 6.0cm{\includegraphics[width=1.20cm,height=2.0cm,keepaspectratio]{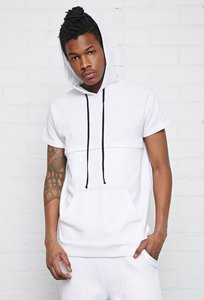}\hss\includegraphics[width=1.20cm,height=2.0cm,keepaspectratio]{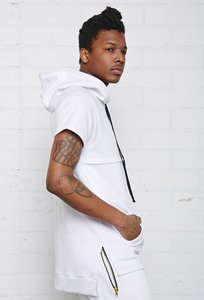}\hss\includegraphics[width=1.20cm,height=2.0cm,keepaspectratio]{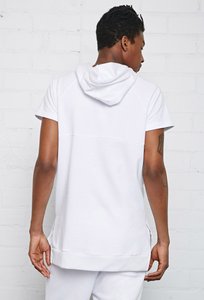}\hss\includegraphics[width=1.20cm,height=2.0cm,keepaspectratio]{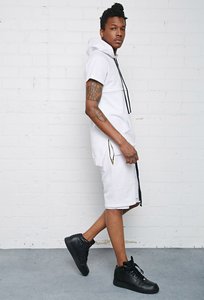}\hss\phantom{\rule{1.20cm}{2.0cm}}\hss}
    \par\vspace{1pt}
    {\scriptsize\textbf{}}
    \par\vspace{0pt}
    \parbox[t]{6.0cm}{\tiny\raggedright White short-sleeved hoodie with black drawstrings, kangaroo pocket, and asymmetrical gold side zipper on wearer's left. Features horizontal chest seam, rolled cuffs, and relaxed fit. Clean back design, soft cotton-blend fleece construction. Contemporary streetwear style.}
  \end{minipage}%
\hfill
\par\vspace{2pt}

\twocolumn

\clearpage
\onecolumn
\section{Retrieval Cases Gallery}
\label{sec:retrieval_gallery}

This section presents nine selected retrieval cases from the DeepFashion validation set, using \textbf{short modification texts} and the \textbf{joint encoding} strategy. For each case, the top panel shows the source garment (left) with its short caption and modification text, alongside the ground-truth target garment. The bottom panel shows the Top-10 retrieval results for each model; a \textcolor{green!60!black}{\textbf{green}} border indicates the correct target was retrieved, and a \textcolor{red!70!black}{\textbf{red}} border indicates an incorrect result. Each retrieved product shows up to five multi-view images.

\par\vspace{16pt}
\noindent\textbf{\large Example 1}
\par\vspace{4pt}
\noindent\rule{\linewidth}{1.2pt}
\par\vspace{4pt}
% ── Case 1: short::deepfashion::1363 ──
\noindent\hfill%
  \begin{minipage}[t]{6.0cm}
    \noindent\hbox to 6.0cm{\includegraphics[width=1.20cm,height=2.0cm,keepaspectratio]{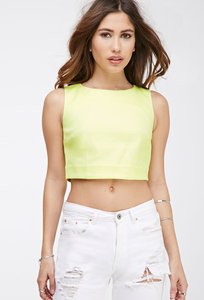}\hss\includegraphics[width=1.20cm,height=2.0cm,keepaspectratio]{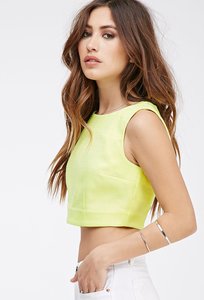}\hss\includegraphics[width=1.20cm,height=2.0cm,keepaspectratio]{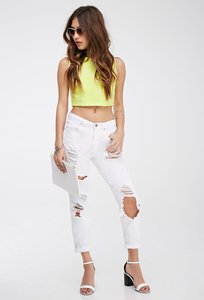}\hss\includegraphics[width=1.20cm,height=2.0cm,keepaspectratio]{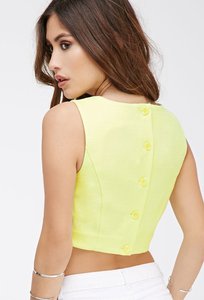}\hss\phantom{\rule{1.20cm}{2.0cm}}\hss}
    \par\vspace{1pt}
    {\scriptsize\textbf{}}
    \par\vspace{0pt}
    \parbox[t]{6.0cm}{\tiny\raggedright Bright yellow sleeveless crop top with high neckline and center back button closure. Textured woven fabric in a boxy, structured fit with cropped length.}
  \end{minipage}%
\hfill%
  \begin{minipage}[t]{3.5cm}
    \centering
    \vspace{0.45cm}%
    \parbox{3.5cm}{\centering\tiny Add a mandarin collar and dual chest patch pockets. Remove the back button placket, change to a high-low hem, and switch to a smooth, lightweight crepe fabric.}\\[2pt]
    {\normalsize$\longrightarrow$}
  \end{minipage}%
\hfill%
  \begin{minipage}[t]{6.0cm}
    \noindent\hbox to 6.0cm{\includegraphics[width=1.20cm,height=2.0cm,keepaspectratio]{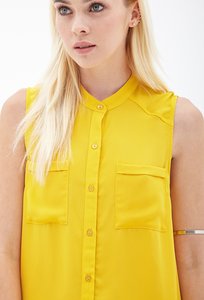}\hss\includegraphics[width=1.20cm,height=2.0cm,keepaspectratio]{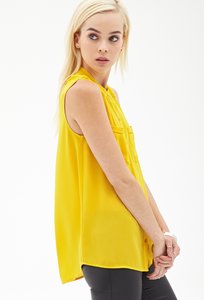}\hss\includegraphics[width=1.20cm,height=2.0cm,keepaspectratio]{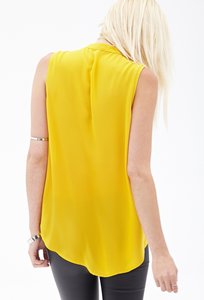}\hss\includegraphics[width=1.20cm,height=2.0cm,keepaspectratio]{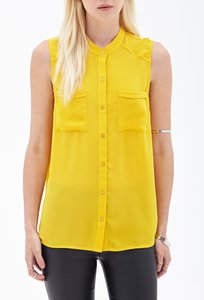}\hss\phantom{\rule{1.20cm}{2.0cm}}\hss}
    \par\vspace{1pt}
    {\scriptsize\textbf{Ground Truth}}
    \par\vspace{0pt}
    \parbox[t]{6.0cm}{\tiny\raggedright Bright yellow sleeveless blouse with mandarin collar, full button front, dual chest pockets, and high-low hem. Features shoulder gathering and lightweight flowy fabric in a relaxed tunic fit.}
  \end{minipage}%
\hfill
\par\vspace{4pt}
\par\vspace{4pt}
\noindent\rule{\linewidth}{0.4pt}
% -- mt_align --
{\small\textbf{\textbf{Ours}}}\quad{\small Rank~2}\\
\noindent
  \begin{minipage}[t]{5.55cm}
    \fcolorbox{red!70!black}{white}{
      \begin{minipage}[t]{5.25cm}
        {\tiny\textbf{\#1}}\\
        \noindent\hbox to 5.25cm{\includegraphics[width=1.10cm,height=1.8cm,keepaspectratio]{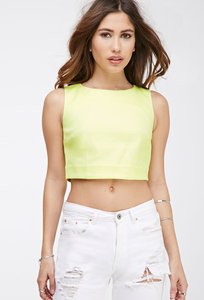}\hss\includegraphics[width=1.10cm,height=1.8cm,keepaspectratio]{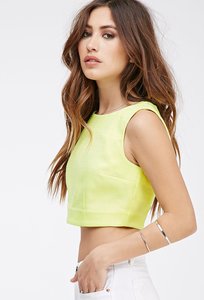}\hss\includegraphics[width=1.10cm,height=1.8cm,keepaspectratio]{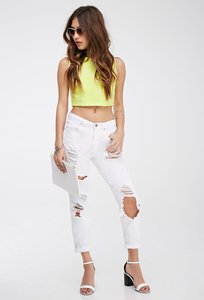}\hss\includegraphics[width=1.10cm,height=1.8cm,keepaspectratio]{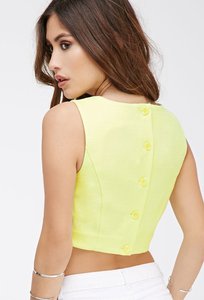}\hss\phantom{\rule{1.10cm}{1.8cm}}\hss}
      \end{minipage}
    }
  \end{minipage}
\hfill
  \begin{minipage}[t]{5.55cm}
    \fcolorbox{green!60!black}{white}{
      \begin{minipage}[t]{5.25cm}
        {\tiny\textbf{\#2}}\\
        \noindent\hbox to 5.25cm{\includegraphics[width=1.10cm,height=1.8cm,keepaspectratio]{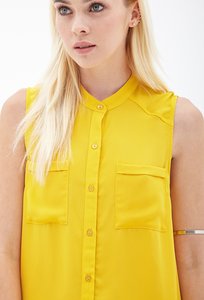}\hss\includegraphics[width=1.10cm,height=1.8cm,keepaspectratio]{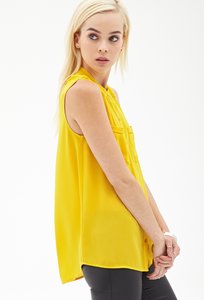}\hss\includegraphics[width=1.10cm,height=1.8cm,keepaspectratio]{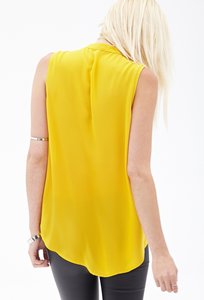}\hss\includegraphics[width=1.10cm,height=1.8cm,keepaspectratio]{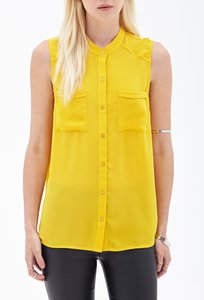}\hss\phantom{\rule{1.10cm}{1.8cm}}\hss}
      \end{minipage}
    }
  \end{minipage}
\hfill
  \begin{minipage}[t]{5.55cm}
    \fcolorbox{red!70!black}{white}{
      \begin{minipage}[t]{5.25cm}
        {\tiny\textbf{\#3}}\\
        \noindent\hbox to 5.25cm{\includegraphics[width=1.10cm,height=1.8cm,keepaspectratio]{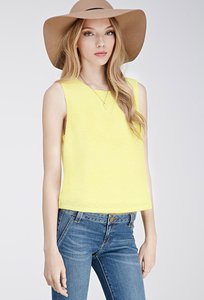}\hss\includegraphics[width=1.10cm,height=1.8cm,keepaspectratio]{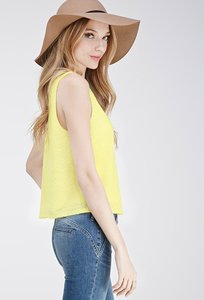}\hss\includegraphics[width=1.10cm,height=1.8cm,keepaspectratio]{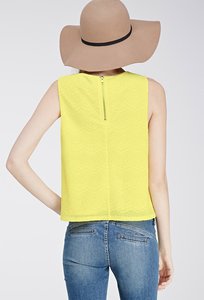}\hss\phantom{\rule{1.10cm}{1.8cm}}\hss\phantom{\rule{1.10cm}{1.8cm}}\hss}
      \end{minipage}
    }
  \end{minipage}
\par\vspace{2pt}
\noindent
  \begin{minipage}[t]{5.55cm}
    \fcolorbox{red!70!black}{white}{
      \begin{minipage}[t]{5.25cm}
        {\tiny\textbf{\#4}}\\
        \noindent\hbox to 5.25cm{\includegraphics[width=1.10cm,height=1.8cm,keepaspectratio]{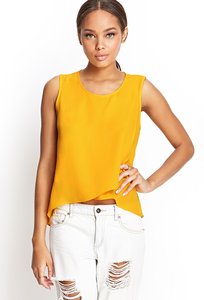}\hss\includegraphics[width=1.10cm,height=1.8cm,keepaspectratio]{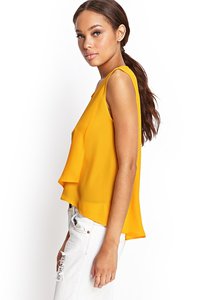}\hss\includegraphics[width=1.10cm,height=1.8cm,keepaspectratio]{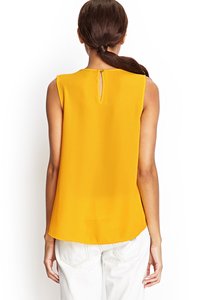}\hss\includegraphics[width=1.10cm,height=1.8cm,keepaspectratio]{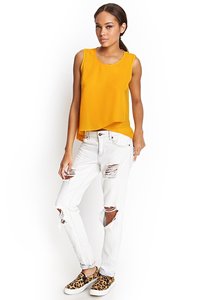}\hss\phantom{\rule{1.10cm}{1.8cm}}\hss}
      \end{minipage}
    }
  \end{minipage}
\hfill
  \begin{minipage}[t]{5.55cm}
    \fcolorbox{red!70!black}{white}{
      \begin{minipage}[t]{5.25cm}
        {\tiny\textbf{\#5}}\\
        \noindent\hbox to 5.25cm{\includegraphics[width=1.10cm,height=1.8cm,keepaspectratio]{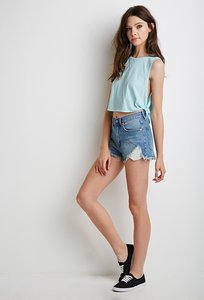}\hss\includegraphics[width=1.10cm,height=1.8cm,keepaspectratio]{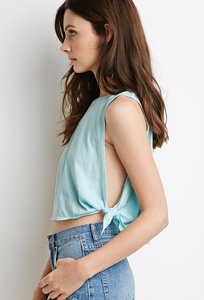}\hss\includegraphics[width=1.10cm,height=1.8cm,keepaspectratio]{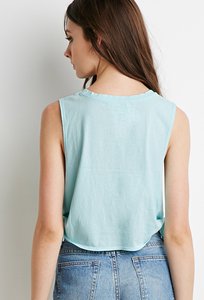}\hss\includegraphics[width=1.10cm,height=1.8cm,keepaspectratio]{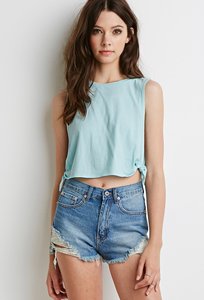}\hss\phantom{\rule{1.10cm}{1.8cm}}\hss}
      \end{minipage}
    }
  \end{minipage}
\hfill
  \begin{minipage}[t]{5.55cm}
    \fcolorbox{red!70!black}{white}{
      \begin{minipage}[t]{5.25cm}
        {\tiny\textbf{\#6}}\\
        \noindent\hbox to 5.25cm{\includegraphics[width=1.10cm,height=1.8cm,keepaspectratio]{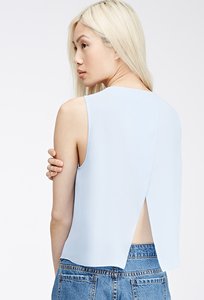}\hss\includegraphics[width=1.10cm,height=1.8cm,keepaspectratio]{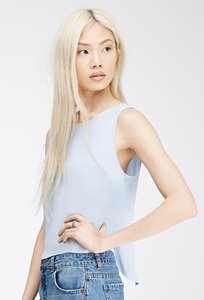}\hss\includegraphics[width=1.10cm,height=1.8cm,keepaspectratio]{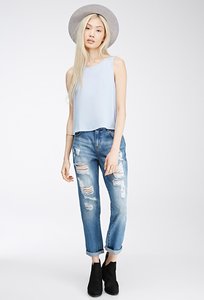}\hss\phantom{\rule{1.10cm}{1.8cm}}\hss\phantom{\rule{1.10cm}{1.8cm}}\hss}
      \end{minipage}
    }
  \end{minipage}
\par\vspace{2pt}
\noindent
  \begin{minipage}[t]{5.55cm}
    \fcolorbox{red!70!black}{white}{
      \begin{minipage}[t]{5.25cm}
        {\tiny\textbf{\#7}}\\
        \noindent\hbox to 5.25cm{\includegraphics[width=1.10cm,height=1.8cm,keepaspectratio]{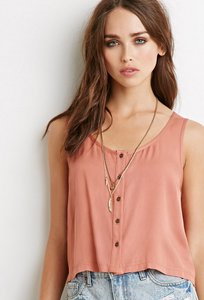}\hss\includegraphics[width=1.10cm,height=1.8cm,keepaspectratio]{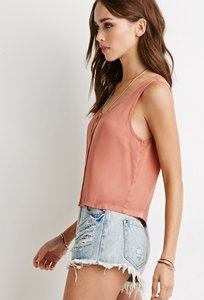}\hss\includegraphics[width=1.10cm,height=1.8cm,keepaspectratio]{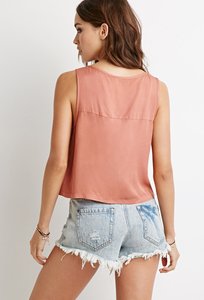}\hss\includegraphics[width=1.10cm,height=1.8cm,keepaspectratio]{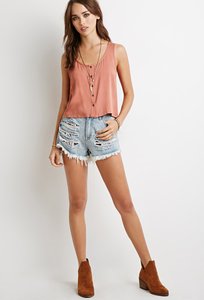}\hss\phantom{\rule{1.10cm}{1.8cm}}\hss}
      \end{minipage}
    }
  \end{minipage}
\hfill
  \begin{minipage}[t]{5.55cm}
    \fcolorbox{red!70!black}{white}{
      \begin{minipage}[t]{5.25cm}
        {\tiny\textbf{\#8}}\\
        \noindent\hbox to 5.25cm{\includegraphics[width=1.10cm,height=1.8cm,keepaspectratio]{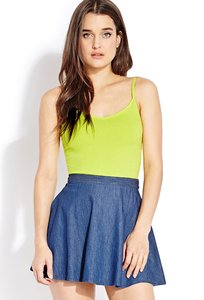}\hss\includegraphics[width=1.10cm,height=1.8cm,keepaspectratio]{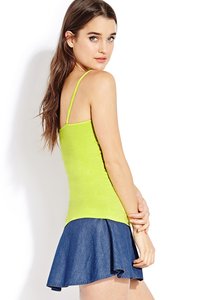}\hss\includegraphics[width=1.10cm,height=1.8cm,keepaspectratio]{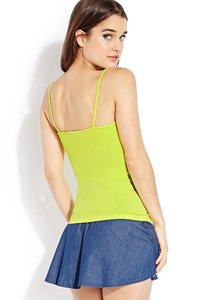}\hss\includegraphics[width=1.10cm,height=1.8cm,keepaspectratio]{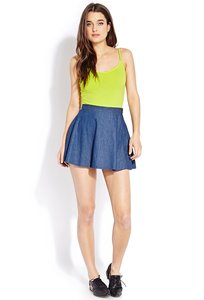}\hss\phantom{\rule{1.10cm}{1.8cm}}\hss}
      \end{minipage}
    }
  \end{minipage}
\hfill
  \begin{minipage}[t]{5.55cm}
    \fcolorbox{red!70!black}{white}{
      \begin{minipage}[t]{5.25cm}
        {\tiny\textbf{\#9}}\\
        \noindent\hbox to 5.25cm{\includegraphics[width=1.10cm,height=1.8cm,keepaspectratio]{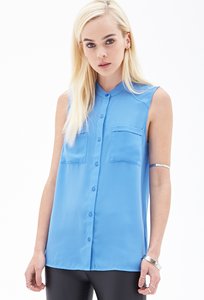}\hss\includegraphics[width=1.10cm,height=1.8cm,keepaspectratio]{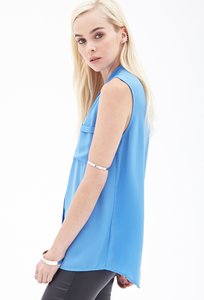}\hss\includegraphics[width=1.10cm,height=1.8cm,keepaspectratio]{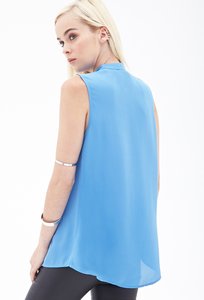}\hss\includegraphics[width=1.10cm,height=1.8cm,keepaspectratio]{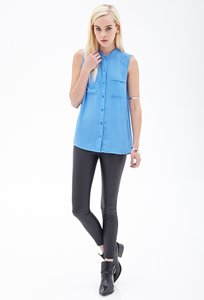}\hss\phantom{\rule{1.10cm}{1.8cm}}\hss}
      \end{minipage}
    }
  \end{minipage}
\par\vspace{2pt}
\noindent
  \begin{minipage}[t]{5.55cm}
    \fcolorbox{red!70!black}{white}{
      \begin{minipage}[t]{5.25cm}
        {\tiny\textbf{\#10}}\\
        \noindent\hbox to 5.25cm{\includegraphics[width=1.10cm,height=1.8cm,keepaspectratio]{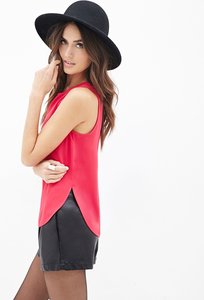}\hss\includegraphics[width=1.10cm,height=1.8cm,keepaspectratio]{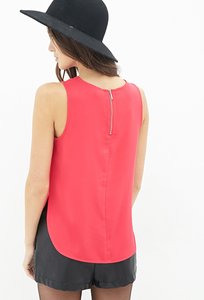}\hss\includegraphics[width=1.10cm,height=1.8cm,keepaspectratio]{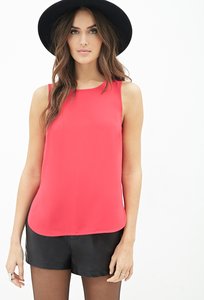}\hss\phantom{\rule{1.10cm}{1.8cm}}\hss\phantom{\rule{1.10cm}{1.8cm}}\hss}
      \end{minipage}
    }
  \end{minipage}
\hfill
  \begin{minipage}[t]{5.55cm}\end{minipage}
\hfill
  \begin{minipage}[t]{5.55cm}\end{minipage}
\par\vspace{2pt}
\noindent\rule{\linewidth}{0.4pt}
\par\vspace{2pt}
% -- qwen3_vl_2b --
{\small\textbf{Qwen3-VL-2B}}\quad{\small Rank~4}\\
\noindent
  \begin{minipage}[t]{5.55cm}
    \fcolorbox{red!70!black}{white}{
      \begin{minipage}[t]{5.25cm}
        {\tiny\textbf{\#1}}\\
        \noindent\hbox to 5.25cm{\includegraphics[width=1.10cm,height=1.8cm,keepaspectratio]{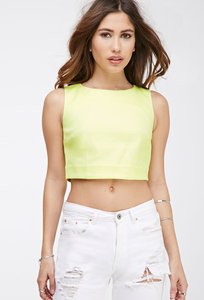}\hss\includegraphics[width=1.10cm,height=1.8cm,keepaspectratio]{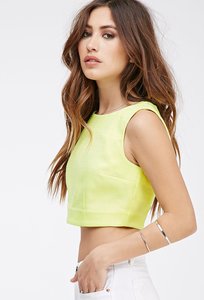}\hss\includegraphics[width=1.10cm,height=1.8cm,keepaspectratio]{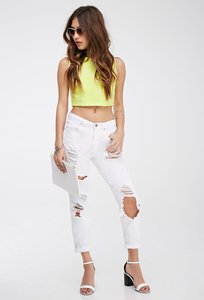}\hss\includegraphics[width=1.10cm,height=1.8cm,keepaspectratio]{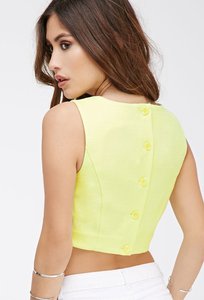}\hss\phantom{\rule{1.10cm}{1.8cm}}\hss}
      \end{minipage}
    }
  \end{minipage}
\hfill
  \begin{minipage}[t]{5.55cm}
    \fcolorbox{red!70!black}{white}{
      \begin{minipage}[t]{5.25cm}
        {\tiny\textbf{\#2}}\\
        \noindent\hbox to 5.25cm{\includegraphics[width=1.10cm,height=1.8cm,keepaspectratio]{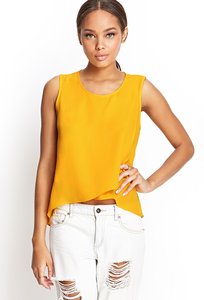}\hss\includegraphics[width=1.10cm,height=1.8cm,keepaspectratio]{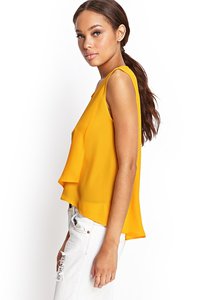}\hss\includegraphics[width=1.10cm,height=1.8cm,keepaspectratio]{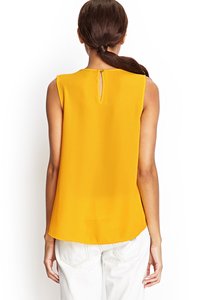}\hss\includegraphics[width=1.10cm,height=1.8cm,keepaspectratio]{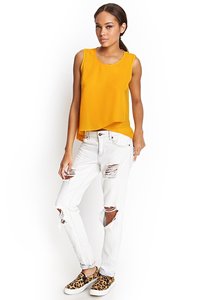}\hss\phantom{\rule{1.10cm}{1.8cm}}\hss}
      \end{minipage}
    }
  \end{minipage}
\hfill
  \begin{minipage}[t]{5.55cm}
    \fcolorbox{red!70!black}{white}{
      \begin{minipage}[t]{5.25cm}
        {\tiny\textbf{\#3}}\\
        \noindent\hbox to 5.25cm{\includegraphics[width=1.10cm,height=1.8cm,keepaspectratio]{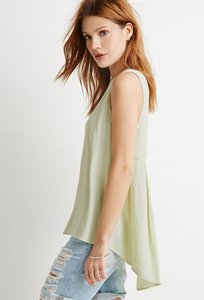}\hss\includegraphics[width=1.10cm,height=1.8cm,keepaspectratio]{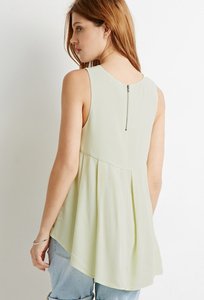}\hss\includegraphics[width=1.10cm,height=1.8cm,keepaspectratio]{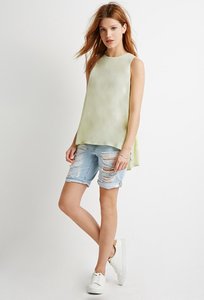}\hss\phantom{\rule{1.10cm}{1.8cm}}\hss\phantom{\rule{1.10cm}{1.8cm}}\hss}
      \end{minipage}
    }
  \end{minipage}
\par\vspace{2pt}
\noindent
  \begin{minipage}[t]{5.55cm}
    \fcolorbox{green!60!black}{white}{
      \begin{minipage}[t]{5.25cm}
        {\tiny\textbf{\#4}}\\
        \noindent\hbox to 5.25cm{\includegraphics[width=1.10cm,height=1.8cm,keepaspectratio]{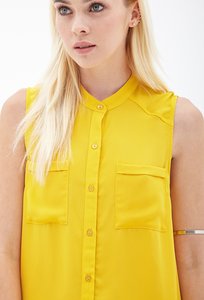}\hss\includegraphics[width=1.10cm,height=1.8cm,keepaspectratio]{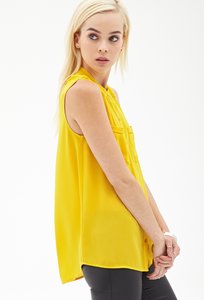}\hss\includegraphics[width=1.10cm,height=1.8cm,keepaspectratio]{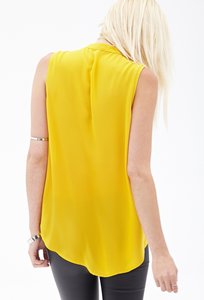}\hss\includegraphics[width=1.10cm,height=1.8cm,keepaspectratio]{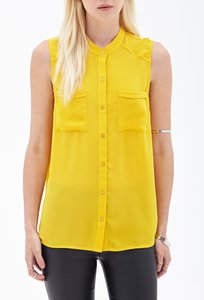}\hss\phantom{\rule{1.10cm}{1.8cm}}\hss}
      \end{minipage}
    }
  \end{minipage}
\hfill
  \begin{minipage}[t]{5.55cm}
    \fcolorbox{red!70!black}{white}{
      \begin{minipage}[t]{5.25cm}
        {\tiny\textbf{\#5}}\\
        \noindent\hbox to 5.25cm{\includegraphics[width=1.10cm,height=1.8cm,keepaspectratio]{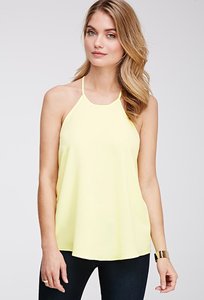}\hss\includegraphics[width=1.10cm,height=1.8cm,keepaspectratio]{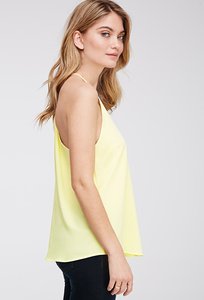}\hss\includegraphics[width=1.10cm,height=1.8cm,keepaspectratio]{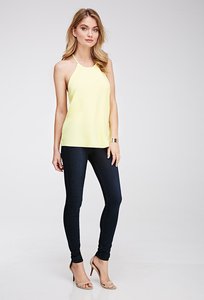}\hss\includegraphics[width=1.10cm,height=1.8cm,keepaspectratio]{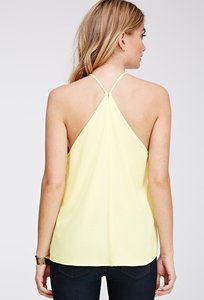}\hss\phantom{\rule{1.10cm}{1.8cm}}\hss}
      \end{minipage}
    }
  \end{minipage}
\hfill
  \begin{minipage}[t]{5.55cm}
    \fcolorbox{red!70!black}{white}{
      \begin{minipage}[t]{5.25cm}
        {\tiny\textbf{\#6}}\\
        \noindent\hbox to 5.25cm{\includegraphics[width=1.10cm,height=1.8cm,keepaspectratio]{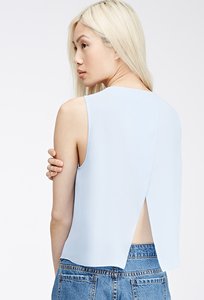}\hss\includegraphics[width=1.10cm,height=1.8cm,keepaspectratio]{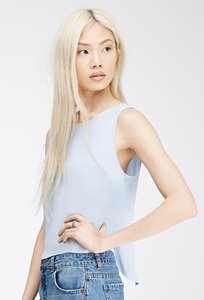}\hss\includegraphics[width=1.10cm,height=1.8cm,keepaspectratio]{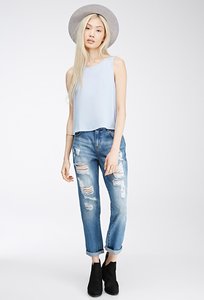}\hss\phantom{\rule{1.10cm}{1.8cm}}\hss\phantom{\rule{1.10cm}{1.8cm}}\hss}
      \end{minipage}
    }
  \end{minipage}
\par\vspace{2pt}
\noindent
  \begin{minipage}[t]{5.55cm}
    \fcolorbox{red!70!black}{white}{
      \begin{minipage}[t]{5.25cm}
        {\tiny\textbf{\#7}}\\
        \noindent\hbox to 5.25cm{\includegraphics[width=1.10cm,height=1.8cm,keepaspectratio]{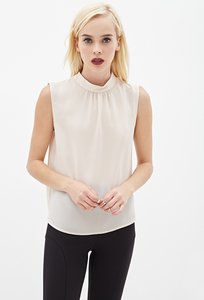}\hss\includegraphics[width=1.10cm,height=1.8cm,keepaspectratio]{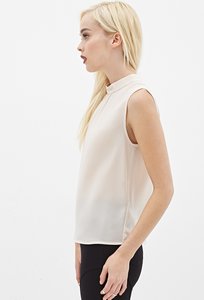}\hss\includegraphics[width=1.10cm,height=1.8cm,keepaspectratio]{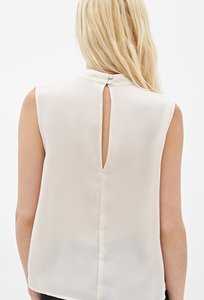}\hss\includegraphics[width=1.10cm,height=1.8cm,keepaspectratio]{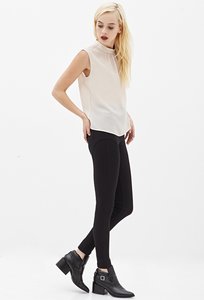}\hss\includegraphics[width=1.10cm,height=1.8cm,keepaspectratio]{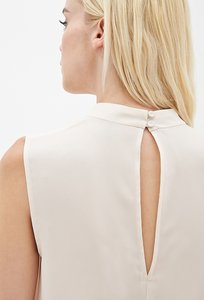}\hss}
      \end{minipage}
    }
  \end{minipage}
\hfill
  \begin{minipage}[t]{5.55cm}
    \fcolorbox{red!70!black}{white}{
      \begin{minipage}[t]{5.25cm}
        {\tiny\textbf{\#8}}\\
        \noindent\hbox to 5.25cm{\includegraphics[width=1.10cm,height=1.8cm,keepaspectratio]{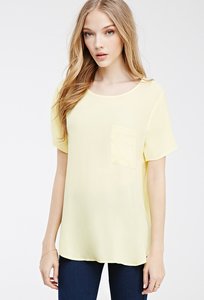}\hss\includegraphics[width=1.10cm,height=1.8cm,keepaspectratio]{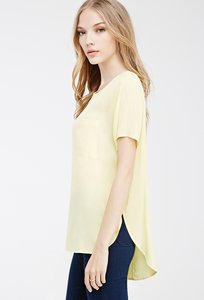}\hss\includegraphics[width=1.10cm,height=1.8cm,keepaspectratio]{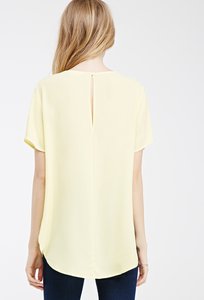}\hss\includegraphics[width=1.10cm,height=1.8cm,keepaspectratio]{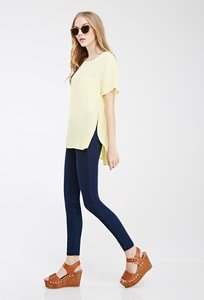}\hss\phantom{\rule{1.10cm}{1.8cm}}\hss}
      \end{minipage}
    }
  \end{minipage}
\hfill
  \begin{minipage}[t]{5.55cm}
    \fcolorbox{red!70!black}{white}{
      \begin{minipage}[t]{5.25cm}
        {\tiny\textbf{\#9}}\\
        \noindent\hbox to 5.25cm{\includegraphics[width=1.10cm,height=1.8cm,keepaspectratio]{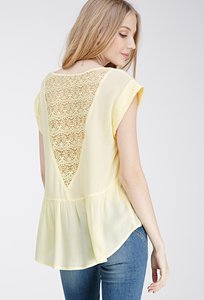}\hss\includegraphics[width=1.10cm,height=1.8cm,keepaspectratio]{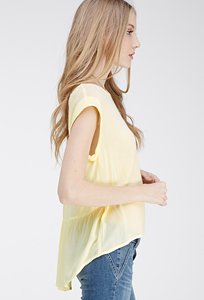}\hss\includegraphics[width=1.10cm,height=1.8cm,keepaspectratio]{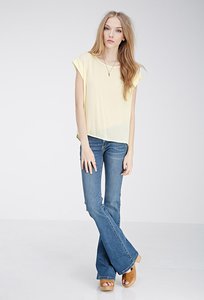}\hss\includegraphics[width=1.10cm,height=1.8cm,keepaspectratio]{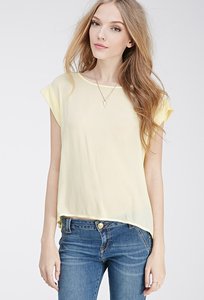}\hss\phantom{\rule{1.10cm}{1.8cm}}\hss}
      \end{minipage}
    }
  \end{minipage}
\par\vspace{2pt}
\noindent
  \begin{minipage}[t]{5.55cm}
    \fcolorbox{red!70!black}{white}{
      \begin{minipage}[t]{5.25cm}
        {\tiny\textbf{\#10}}\\
        \noindent\hbox to 5.25cm{\includegraphics[width=1.10cm,height=1.8cm,keepaspectratio]{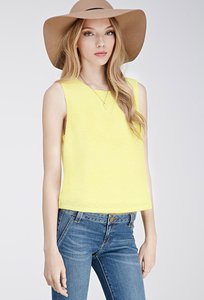}\hss\includegraphics[width=1.10cm,height=1.8cm,keepaspectratio]{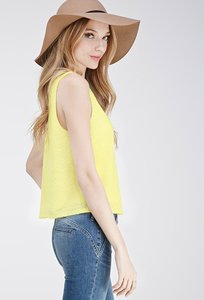}\hss\includegraphics[width=1.10cm,height=1.8cm,keepaspectratio]{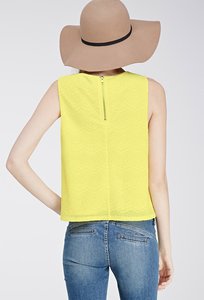}\hss\phantom{\rule{1.10cm}{1.8cm}}\hss\phantom{\rule{1.10cm}{1.8cm}}\hss}
      \end{minipage}
    }
  \end{minipage}
\hfill
  \begin{minipage}[t]{5.55cm}\end{minipage}
\hfill
  \begin{minipage}[t]{5.55cm}\end{minipage}
\par\vspace{2pt}
\noindent\rule{\linewidth}{0.4pt}
\par\vspace{2pt}
% -- qwen3_vl_8b --
{\small\textbf{Qwen3-VL-8B}}\quad{\small Rank~3}\\
\noindent
  \begin{minipage}[t]{5.55cm}
    \fcolorbox{red!70!black}{white}{
      \begin{minipage}[t]{5.25cm}
        {\tiny\textbf{\#1}}\\
        \noindent\hbox to 5.25cm{\includegraphics[width=1.10cm,height=1.8cm,keepaspectratio]{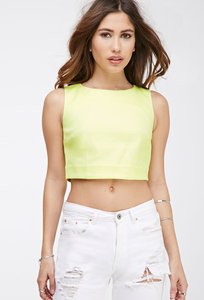}\hss\includegraphics[width=1.10cm,height=1.8cm,keepaspectratio]{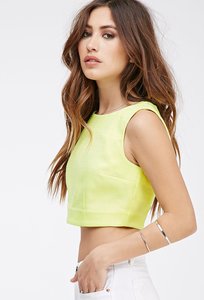}\hss\includegraphics[width=1.10cm,height=1.8cm,keepaspectratio]{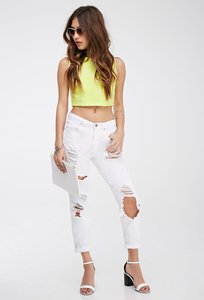}\hss\includegraphics[width=1.10cm,height=1.8cm,keepaspectratio]{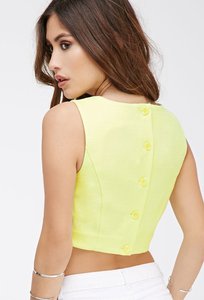}\hss\phantom{\rule{1.10cm}{1.8cm}}\hss}
      \end{minipage}
    }
  \end{minipage}
\hfill
  \begin{minipage}[t]{5.55cm}
    \fcolorbox{red!70!black}{white}{
      \begin{minipage}[t]{5.25cm}
        {\tiny\textbf{\#2}}\\
        \noindent\hbox to 5.25cm{\includegraphics[width=1.10cm,height=1.8cm,keepaspectratio]{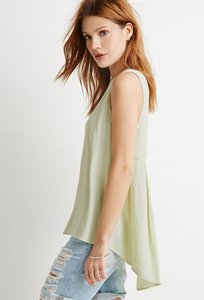}\hss\includegraphics[width=1.10cm,height=1.8cm,keepaspectratio]{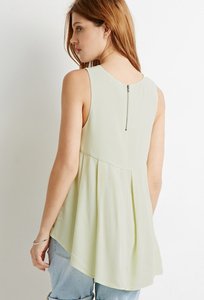}\hss\includegraphics[width=1.10cm,height=1.8cm,keepaspectratio]{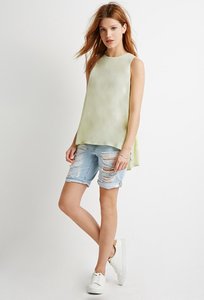}\hss\phantom{\rule{1.10cm}{1.8cm}}\hss\phantom{\rule{1.10cm}{1.8cm}}\hss}
      \end{minipage}
    }
  \end{minipage}
\hfill
  \begin{minipage}[t]{5.55cm}
    \fcolorbox{green!60!black}{white}{
      \begin{minipage}[t]{5.25cm}
        {\tiny\textbf{\#3}}\\
        \noindent\hbox to 5.25cm{\includegraphics[width=1.10cm,height=1.8cm,keepaspectratio]{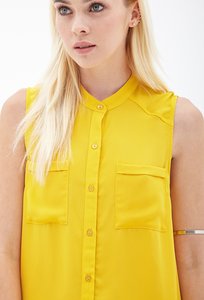}\hss\includegraphics[width=1.10cm,height=1.8cm,keepaspectratio]{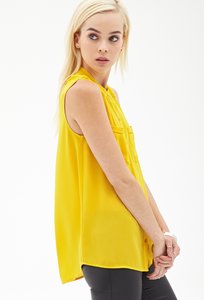}\hss\includegraphics[width=1.10cm,height=1.8cm,keepaspectratio]{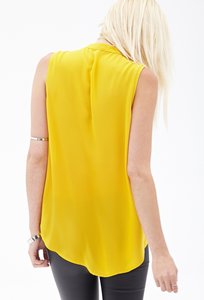}\hss\includegraphics[width=1.10cm,height=1.8cm,keepaspectratio]{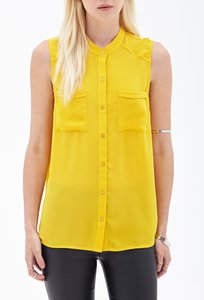}\hss\phantom{\rule{1.10cm}{1.8cm}}\hss}
      \end{minipage}
    }
  \end{minipage}
\par\vspace{2pt}
\noindent
  \begin{minipage}[t]{5.55cm}
    \fcolorbox{red!70!black}{white}{
      \begin{minipage}[t]{5.25cm}
        {\tiny\textbf{\#4}}\\
        \noindent\hbox to 5.25cm{\includegraphics[width=1.10cm,height=1.8cm,keepaspectratio]{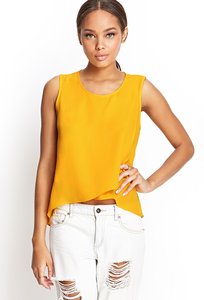}\hss\includegraphics[width=1.10cm,height=1.8cm,keepaspectratio]{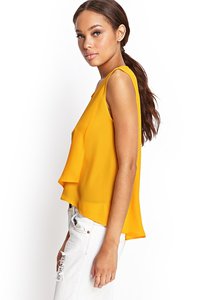}\hss\includegraphics[width=1.10cm,height=1.8cm,keepaspectratio]{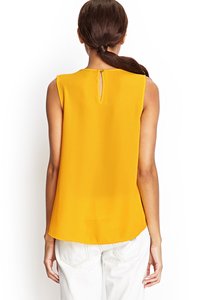}\hss\includegraphics[width=1.10cm,height=1.8cm,keepaspectratio]{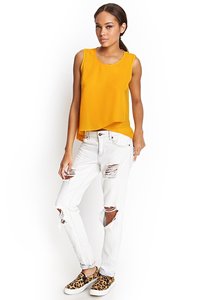}\hss\phantom{\rule{1.10cm}{1.8cm}}\hss}
      \end{minipage}
    }
  \end{minipage}
\hfill
  \begin{minipage}[t]{5.55cm}
    \fcolorbox{red!70!black}{white}{
      \begin{minipage}[t]{5.25cm}
        {\tiny\textbf{\#5}}\\
        \noindent\hbox to 5.25cm{\includegraphics[width=1.10cm,height=1.8cm,keepaspectratio]{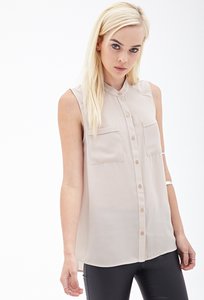}\hss\includegraphics[width=1.10cm,height=1.8cm,keepaspectratio]{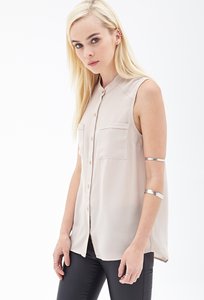}\hss\includegraphics[width=1.10cm,height=1.8cm,keepaspectratio]{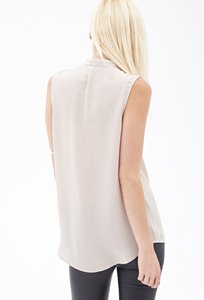}\hss\includegraphics[width=1.10cm,height=1.8cm,keepaspectratio]{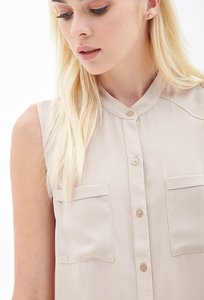}\hss\phantom{\rule{1.10cm}{1.8cm}}\hss}
      \end{minipage}
    }
  \end{minipage}
\hfill
  \begin{minipage}[t]{5.55cm}
    \fcolorbox{red!70!black}{white}{
      \begin{minipage}[t]{5.25cm}
        {\tiny\textbf{\#6}}\\
        \noindent\hbox to 5.25cm{\includegraphics[width=1.10cm,height=1.8cm,keepaspectratio]{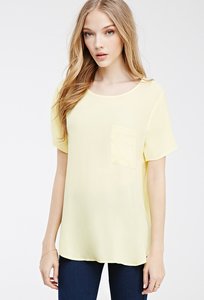}\hss\includegraphics[width=1.10cm,height=1.8cm,keepaspectratio]{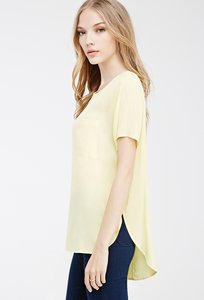}\hss\includegraphics[width=1.10cm,height=1.8cm,keepaspectratio]{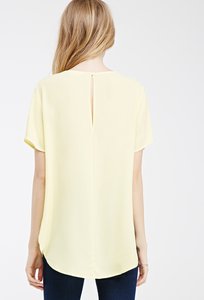}\hss\includegraphics[width=1.10cm,height=1.8cm,keepaspectratio]{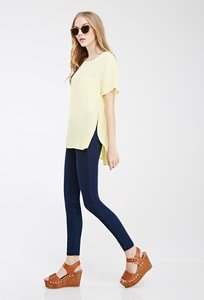}\hss\phantom{\rule{1.10cm}{1.8cm}}\hss}
      \end{minipage}
    }
  \end{minipage}
\par\vspace{2pt}
\noindent
  \begin{minipage}[t]{5.55cm}
    \fcolorbox{red!70!black}{white}{
      \begin{minipage}[t]{5.25cm}
        {\tiny\textbf{\#7}}\\
        \noindent\hbox to 5.25cm{\includegraphics[width=1.10cm,height=1.8cm,keepaspectratio]{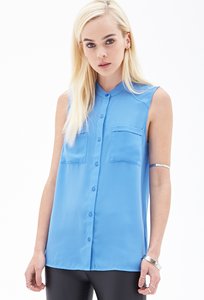}\hss\includegraphics[width=1.10cm,height=1.8cm,keepaspectratio]{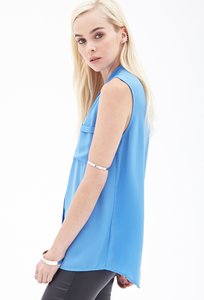}\hss\includegraphics[width=1.10cm,height=1.8cm,keepaspectratio]{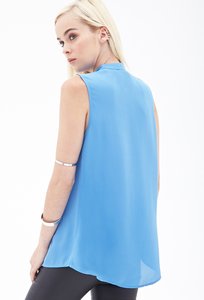}\hss\includegraphics[width=1.10cm,height=1.8cm,keepaspectratio]{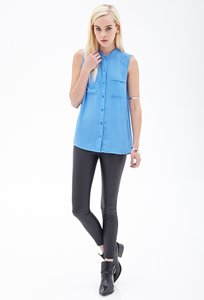}\hss\phantom{\rule{1.10cm}{1.8cm}}\hss}
      \end{minipage}
    }
  \end{minipage}
\hfill
  \begin{minipage}[t]{5.55cm}
    \fcolorbox{red!70!black}{white}{
      \begin{minipage}[t]{5.25cm}
        {\tiny\textbf{\#8}}\\
        \noindent\hbox to 5.25cm{\includegraphics[width=1.10cm,height=1.8cm,keepaspectratio]{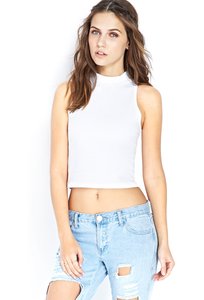}\hss\includegraphics[width=1.10cm,height=1.8cm,keepaspectratio]{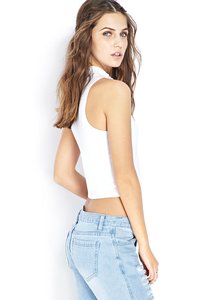}\hss\includegraphics[width=1.10cm,height=1.8cm,keepaspectratio]{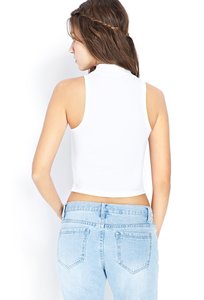}\hss\includegraphics[width=1.10cm,height=1.8cm,keepaspectratio]{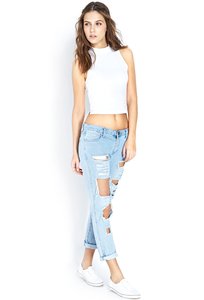}\hss\phantom{\rule{1.10cm}{1.8cm}}\hss}
      \end{minipage}
    }
  \end{minipage}
\hfill
  \begin{minipage}[t]{5.55cm}
    \fcolorbox{red!70!black}{white}{
      \begin{minipage}[t]{5.25cm}
        {\tiny\textbf{\#9}}\\
        \noindent\hbox to 5.25cm{\includegraphics[width=1.10cm,height=1.8cm,keepaspectratio]{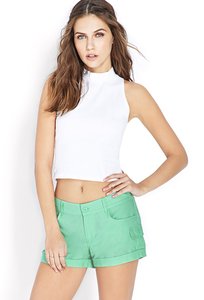}\hss\includegraphics[width=1.10cm,height=1.8cm,keepaspectratio]{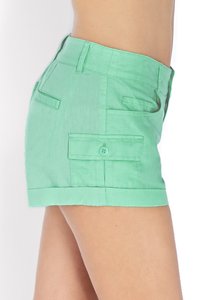}\hss\includegraphics[width=1.10cm,height=1.8cm,keepaspectratio]{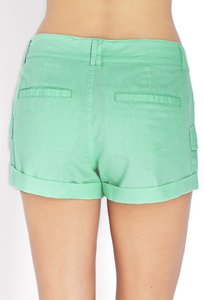}\hss\includegraphics[width=1.10cm,height=1.8cm,keepaspectratio]{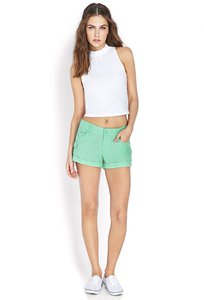}\hss\includegraphics[width=1.10cm,height=1.8cm,keepaspectratio]{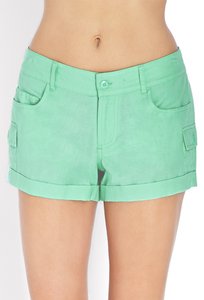}\hss}
      \end{minipage}
    }
  \end{minipage}
\par\vspace{2pt}
\noindent
  \begin{minipage}[t]{5.55cm}
    \fcolorbox{red!70!black}{white}{
      \begin{minipage}[t]{5.25cm}
        {\tiny\textbf{\#10}}\\
        \noindent\hbox to 5.25cm{\includegraphics[width=1.10cm,height=1.8cm,keepaspectratio]{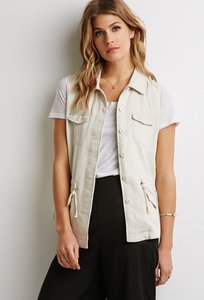}\hss\includegraphics[width=1.10cm,height=1.8cm,keepaspectratio]{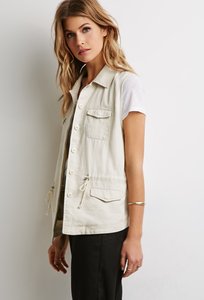}\hss\includegraphics[width=1.10cm,height=1.8cm,keepaspectratio]{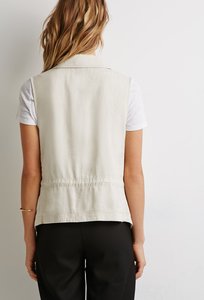}\hss\includegraphics[width=1.10cm,height=1.8cm,keepaspectratio]{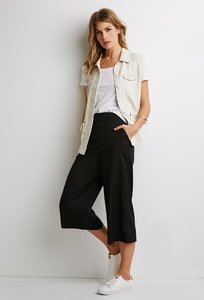}\hss\phantom{\rule{1.10cm}{1.8cm}}\hss}
      \end{minipage}
    }
  \end{minipage}
\hfill
  \begin{minipage}[t]{5.55cm}\end{minipage}
\hfill
  \begin{minipage}[t]{5.55cm}\end{minipage}
\par\vspace{2pt}
\noindent\rule{\linewidth}{0.4pt}
\par\vspace{2pt}
% -- reznembed --
{\small\textbf{RezNEmbed}}\quad{\small Rank~5}\\
\noindent
  \begin{minipage}[t]{5.55cm}
    \fcolorbox{red!70!black}{white}{
      \begin{minipage}[t]{5.25cm}
        {\tiny\textbf{\#1}}\\
        \noindent\hbox to 5.25cm{\includegraphics[width=1.10cm,height=1.8cm,keepaspectratio]{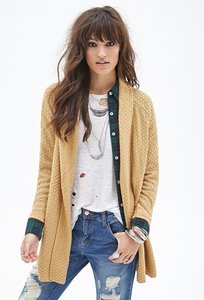}\hss\includegraphics[width=1.10cm,height=1.8cm,keepaspectratio]{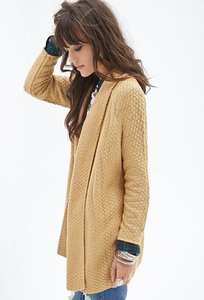}\hss\includegraphics[width=1.10cm,height=1.8cm,keepaspectratio]{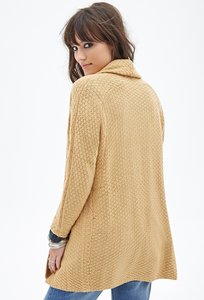}\hss\includegraphics[width=1.10cm,height=1.8cm,keepaspectratio]{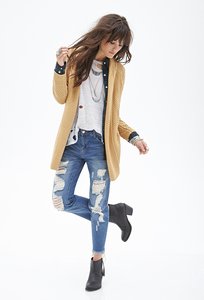}\hss\phantom{\rule{1.10cm}{1.8cm}}\hss}
      \end{minipage}
    }
  \end{minipage}
\hfill
  \begin{minipage}[t]{5.55cm}
    \fcolorbox{red!70!black}{white}{
      \begin{minipage}[t]{5.25cm}
        {\tiny\textbf{\#2}}\\
        \noindent\hbox to 5.25cm{\includegraphics[width=1.10cm,height=1.8cm,keepaspectratio]{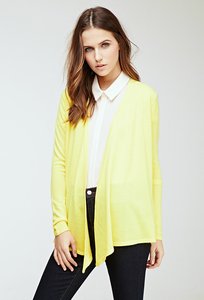}\hss\includegraphics[width=1.10cm,height=1.8cm,keepaspectratio]{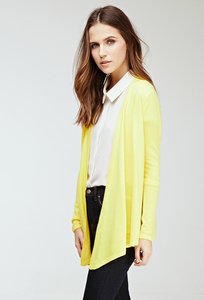}\hss\includegraphics[width=1.10cm,height=1.8cm,keepaspectratio]{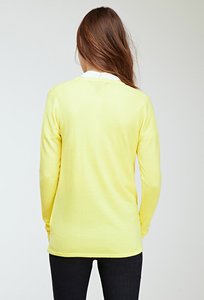}\hss\includegraphics[width=1.10cm,height=1.8cm,keepaspectratio]{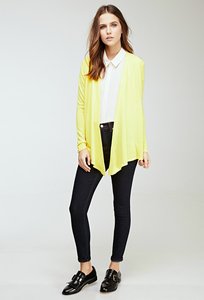}\hss\phantom{\rule{1.10cm}{1.8cm}}\hss}
      \end{minipage}
    }
  \end{minipage}
\hfill
  \begin{minipage}[t]{5.55cm}
    \fcolorbox{red!70!black}{white}{
      \begin{minipage}[t]{5.25cm}
        {\tiny\textbf{\#3}}\\
        \noindent\hbox to 5.25cm{\includegraphics[width=1.10cm,height=1.8cm,keepaspectratio]{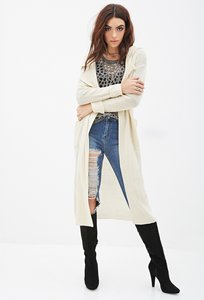}\hss\includegraphics[width=1.10cm,height=1.8cm,keepaspectratio]{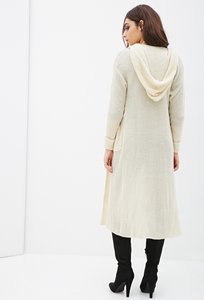}\hss\includegraphics[width=1.10cm,height=1.8cm,keepaspectratio]{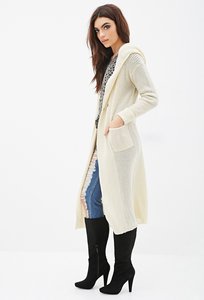}\hss\phantom{\rule{1.10cm}{1.8cm}}\hss\phantom{\rule{1.10cm}{1.8cm}}\hss}
      \end{minipage}
    }
  \end{minipage}
\par\vspace{2pt}
\noindent
  \begin{minipage}[t]{5.55cm}
    \fcolorbox{red!70!black}{white}{
      \begin{minipage}[t]{5.25cm}
        {\tiny\textbf{\#4}}\\
        \noindent\hbox to 5.25cm{\includegraphics[width=1.10cm,height=1.8cm,keepaspectratio]{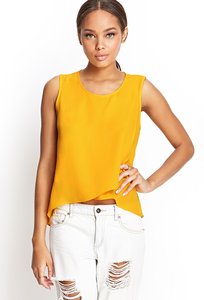}\hss\includegraphics[width=1.10cm,height=1.8cm,keepaspectratio]{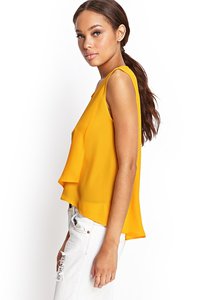}\hss\includegraphics[width=1.10cm,height=1.8cm,keepaspectratio]{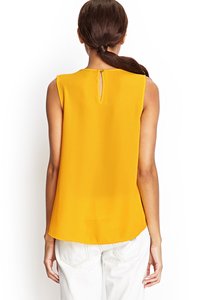}\hss\includegraphics[width=1.10cm,height=1.8cm,keepaspectratio]{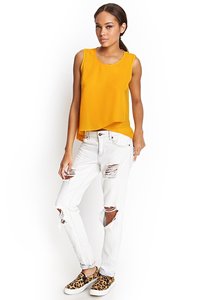}\hss\phantom{\rule{1.10cm}{1.8cm}}\hss}
      \end{minipage}
    }
  \end{minipage}
\hfill
  \begin{minipage}[t]{5.55cm}
    \fcolorbox{green!60!black}{white}{
      \begin{minipage}[t]{5.25cm}
        {\tiny\textbf{\#5}}\\
        \noindent\hbox to 5.25cm{\includegraphics[width=1.10cm,height=1.8cm,keepaspectratio]{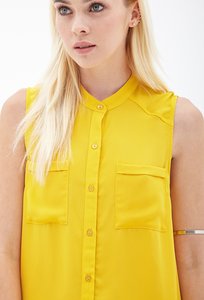}\hss\includegraphics[width=1.10cm,height=1.8cm,keepaspectratio]{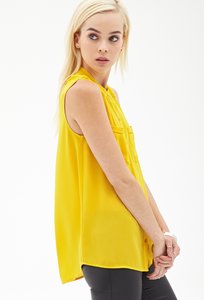}\hss\includegraphics[width=1.10cm,height=1.8cm,keepaspectratio]{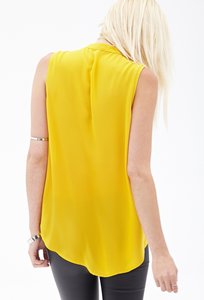}\hss\includegraphics[width=1.10cm,height=1.8cm,keepaspectratio]{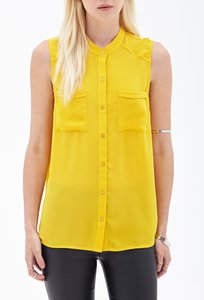}\hss\phantom{\rule{1.10cm}{1.8cm}}\hss}
      \end{minipage}
    }
  \end{minipage}
\hfill
  \begin{minipage}[t]{5.55cm}
    \fcolorbox{red!70!black}{white}{
      \begin{minipage}[t]{5.25cm}
        {\tiny\textbf{\#6}}\\
        \noindent\hbox to 5.25cm{\includegraphics[width=1.10cm,height=1.8cm,keepaspectratio]{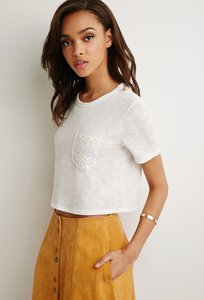}\hss\includegraphics[width=1.10cm,height=1.8cm,keepaspectratio]{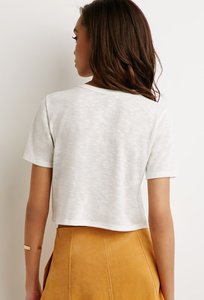}\hss\includegraphics[width=1.10cm,height=1.8cm,keepaspectratio]{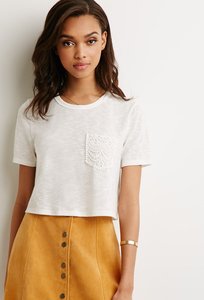}\hss\phantom{\rule{1.10cm}{1.8cm}}\hss\phantom{\rule{1.10cm}{1.8cm}}\hss}
      \end{minipage}
    }
  \end{minipage}
\par\vspace{2pt}
\noindent
  \begin{minipage}[t]{5.55cm}
    \fcolorbox{red!70!black}{white}{
      \begin{minipage}[t]{5.25cm}
        {\tiny\textbf{\#7}}\\
        \noindent\hbox to 5.25cm{\includegraphics[width=1.10cm,height=1.8cm,keepaspectratio]{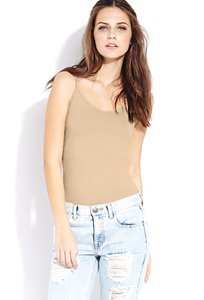}\hss\includegraphics[width=1.10cm,height=1.8cm,keepaspectratio]{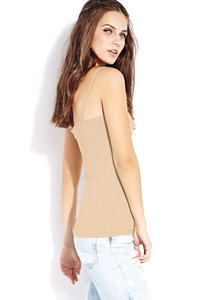}\hss\includegraphics[width=1.10cm,height=1.8cm,keepaspectratio]{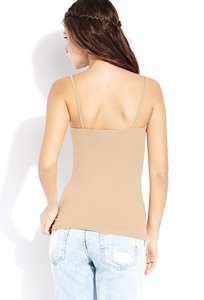}\hss\includegraphics[width=1.10cm,height=1.8cm,keepaspectratio]{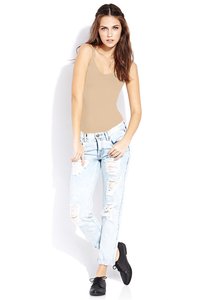}\hss\phantom{\rule{1.10cm}{1.8cm}}\hss}
      \end{minipage}
    }
  \end{minipage}
\hfill
  \begin{minipage}[t]{5.55cm}
    \fcolorbox{red!70!black}{white}{
      \begin{minipage}[t]{5.25cm}
        {\tiny\textbf{\#8}}\\
        \noindent\hbox to 5.25cm{\includegraphics[width=1.10cm,height=1.8cm,keepaspectratio]{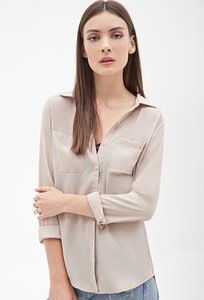}\hss\includegraphics[width=1.10cm,height=1.8cm,keepaspectratio]{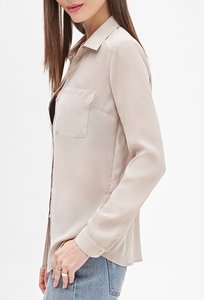}\hss\includegraphics[width=1.10cm,height=1.8cm,keepaspectratio]{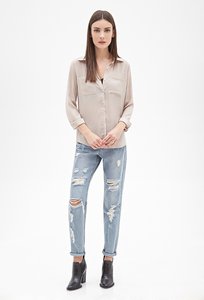}\hss\includegraphics[width=1.10cm,height=1.8cm,keepaspectratio]{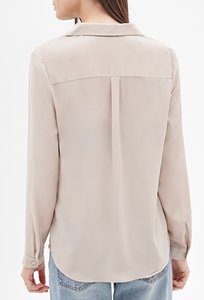}\hss\phantom{\rule{1.10cm}{1.8cm}}\hss}
      \end{minipage}
    }
  \end{minipage}
\hfill
  \begin{minipage}[t]{5.55cm}
    \fcolorbox{red!70!black}{white}{
      \begin{minipage}[t]{5.25cm}
        {\tiny\textbf{\#9}}\\
        \noindent\hbox to 5.25cm{\includegraphics[width=1.10cm,height=1.8cm,keepaspectratio]{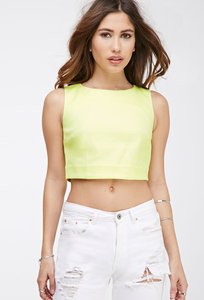}\hss\includegraphics[width=1.10cm,height=1.8cm,keepaspectratio]{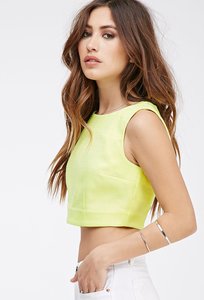}\hss\includegraphics[width=1.10cm,height=1.8cm,keepaspectratio]{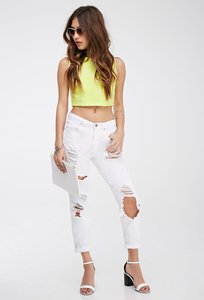}\hss\includegraphics[width=1.10cm,height=1.8cm,keepaspectratio]{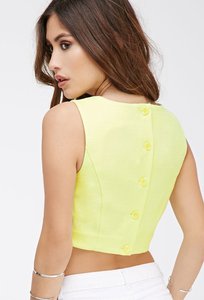}\hss\phantom{\rule{1.10cm}{1.8cm}}\hss}
      \end{minipage}
    }
  \end{minipage}
\par\vspace{2pt}
\noindent
  \begin{minipage}[t]{5.55cm}
    \fcolorbox{red!70!black}{white}{
      \begin{minipage}[t]{5.25cm}
        {\tiny\textbf{\#10}}\\
        \noindent\hbox to 5.25cm{\includegraphics[width=1.10cm,height=1.8cm,keepaspectratio]{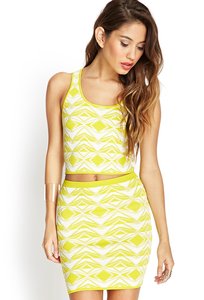}\hss\includegraphics[width=1.10cm,height=1.8cm,keepaspectratio]{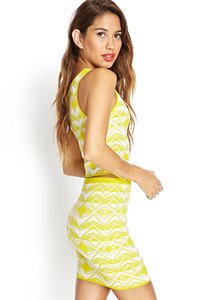}\hss\includegraphics[width=1.10cm,height=1.8cm,keepaspectratio]{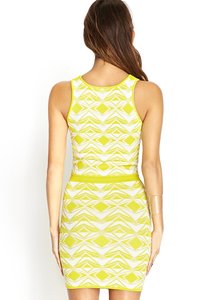}\hss\includegraphics[width=1.10cm,height=1.8cm,keepaspectratio]{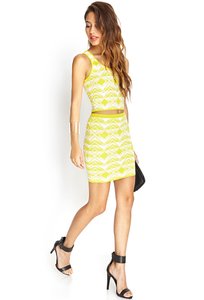}\hss\phantom{\rule{1.10cm}{1.8cm}}\hss}
      \end{minipage}
    }
  \end{minipage}
\hfill
  \begin{minipage}[t]{5.55cm}\end{minipage}
\hfill
  \begin{minipage}[t]{5.55cm}\end{minipage}
\par\vspace{2pt}
\noindent\rule{\linewidth}{0.4pt}
\par\vspace{2pt}
% -- doubao --
{\small\textbf{Doubao-E-V}}\quad{\small Rank~3}\\
\noindent
  \begin{minipage}[t]{5.55cm}
    \fcolorbox{red!70!black}{white}{
      \begin{minipage}[t]{5.25cm}
        {\tiny\textbf{\#1}}\\
        \noindent\hbox to 5.25cm{\includegraphics[width=1.10cm,height=1.8cm,keepaspectratio]{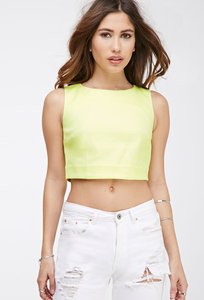}\hss\includegraphics[width=1.10cm,height=1.8cm,keepaspectratio]{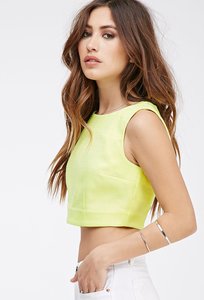}\hss\includegraphics[width=1.10cm,height=1.8cm,keepaspectratio]{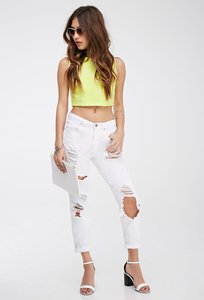}\hss\includegraphics[width=1.10cm,height=1.8cm,keepaspectratio]{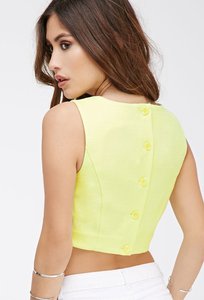}\hss\phantom{\rule{1.10cm}{1.8cm}}\hss}
      \end{minipage}
    }
  \end{minipage}
\hfill
  \begin{minipage}[t]{5.55cm}
    \fcolorbox{red!70!black}{white}{
      \begin{minipage}[t]{5.25cm}
        {\tiny\textbf{\#2}}\\
        \noindent\hbox to 5.25cm{\includegraphics[width=1.10cm,height=1.8cm,keepaspectratio]{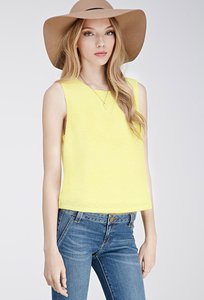}\hss\includegraphics[width=1.10cm,height=1.8cm,keepaspectratio]{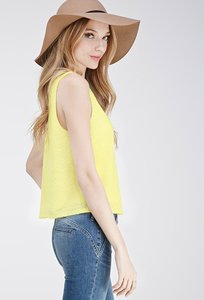}\hss\includegraphics[width=1.10cm,height=1.8cm,keepaspectratio]{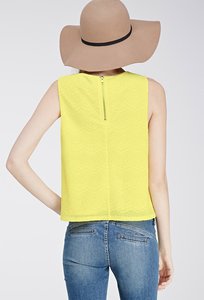}\hss\phantom{\rule{1.10cm}{1.8cm}}\hss\phantom{\rule{1.10cm}{1.8cm}}\hss}
      \end{minipage}
    }
  \end{minipage}
\hfill
  \begin{minipage}[t]{5.55cm}
    \fcolorbox{green!60!black}{white}{
      \begin{minipage}[t]{5.25cm}
        {\tiny\textbf{\#3}}\\
        \noindent\hbox to 5.25cm{\includegraphics[width=1.10cm,height=1.8cm,keepaspectratio]{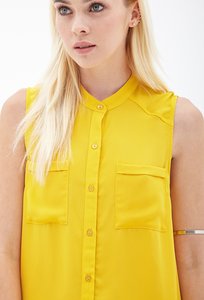}\hss\includegraphics[width=1.10cm,height=1.8cm,keepaspectratio]{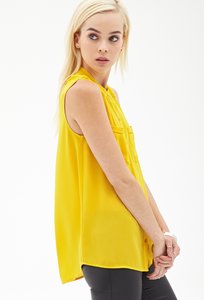}\hss\includegraphics[width=1.10cm,height=1.8cm,keepaspectratio]{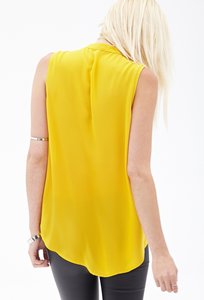}\hss\includegraphics[width=1.10cm,height=1.8cm,keepaspectratio]{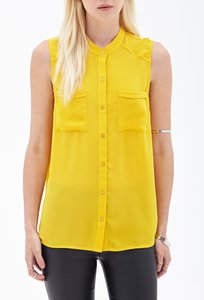}\hss\phantom{\rule{1.10cm}{1.8cm}}\hss}
      \end{minipage}
    }
  \end{minipage}
\par\vspace{2pt}
\noindent
  \begin{minipage}[t]{5.55cm}
    \fcolorbox{red!70!black}{white}{
      \begin{minipage}[t]{5.25cm}
        {\tiny\textbf{\#4}}\\
        \noindent\hbox to 5.25cm{\includegraphics[width=1.10cm,height=1.8cm,keepaspectratio]{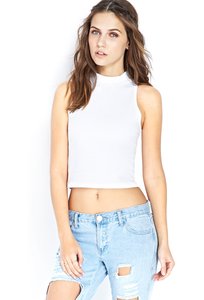}\hss\includegraphics[width=1.10cm,height=1.8cm,keepaspectratio]{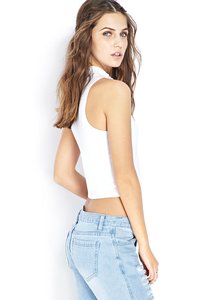}\hss\includegraphics[width=1.10cm,height=1.8cm,keepaspectratio]{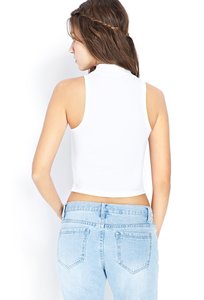}\hss\includegraphics[width=1.10cm,height=1.8cm,keepaspectratio]{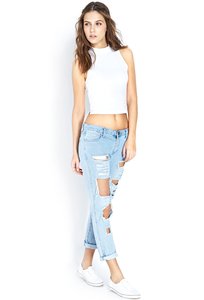}\hss\phantom{\rule{1.10cm}{1.8cm}}\hss}
      \end{minipage}
    }
  \end{minipage}
\hfill
  \begin{minipage}[t]{5.55cm}
    \fcolorbox{red!70!black}{white}{
      \begin{minipage}[t]{5.25cm}
        {\tiny\textbf{\#5}}\\
        \noindent\hbox to 5.25cm{\includegraphics[width=1.10cm,height=1.8cm,keepaspectratio]{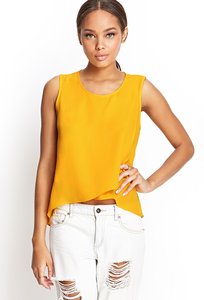}\hss\includegraphics[width=1.10cm,height=1.8cm,keepaspectratio]{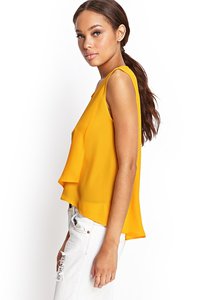}\hss\includegraphics[width=1.10cm,height=1.8cm,keepaspectratio]{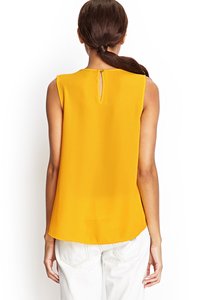}\hss\includegraphics[width=1.10cm,height=1.8cm,keepaspectratio]{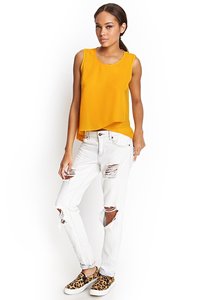}\hss\phantom{\rule{1.10cm}{1.8cm}}\hss}
      \end{minipage}
    }
  \end{minipage}
\hfill
  \begin{minipage}[t]{5.55cm}
    \fcolorbox{red!70!black}{white}{
      \begin{minipage}[t]{5.25cm}
        {\tiny\textbf{\#6}}\\
        \noindent\hbox to 5.25cm{\includegraphics[width=1.10cm,height=1.8cm,keepaspectratio]{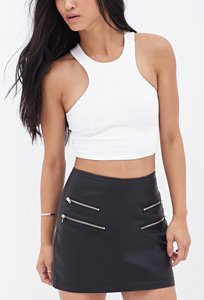}\hss\includegraphics[width=1.10cm,height=1.8cm,keepaspectratio]{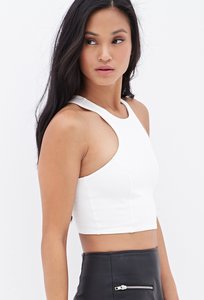}\hss\includegraphics[width=1.10cm,height=1.8cm,keepaspectratio]{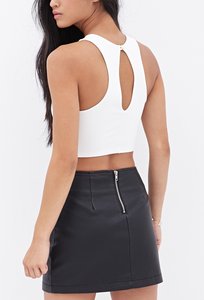}\hss\phantom{\rule{1.10cm}{1.8cm}}\hss\phantom{\rule{1.10cm}{1.8cm}}\hss}
      \end{minipage}
    }
  \end{minipage}
\par\vspace{2pt}
\noindent
  \begin{minipage}[t]{5.55cm}
    \fcolorbox{red!70!black}{white}{
      \begin{minipage}[t]{5.25cm}
        {\tiny\textbf{\#7}}\\
        \noindent\hbox to 5.25cm{\includegraphics[width=1.10cm,height=1.8cm,keepaspectratio]{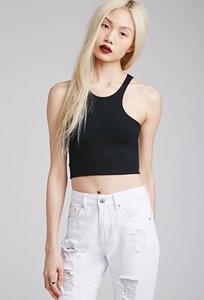}\hss\includegraphics[width=1.10cm,height=1.8cm,keepaspectratio]{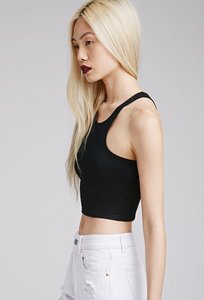}\hss\includegraphics[width=1.10cm,height=1.8cm,keepaspectratio]{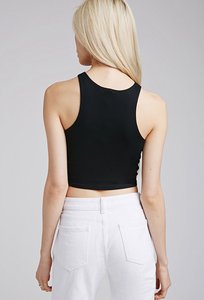}\hss\includegraphics[width=1.10cm,height=1.8cm,keepaspectratio]{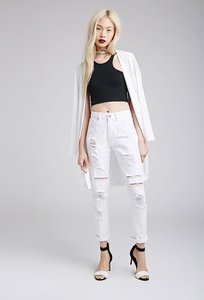}\hss\phantom{\rule{1.10cm}{1.8cm}}\hss}
      \end{minipage}
    }
  \end{minipage}
\hfill
  \begin{minipage}[t]{5.55cm}
    \fcolorbox{red!70!black}{white}{
      \begin{minipage}[t]{5.25cm}
        {\tiny\textbf{\#8}}\\
        \noindent\hbox to 5.25cm{\includegraphics[width=1.10cm,height=1.8cm,keepaspectratio]{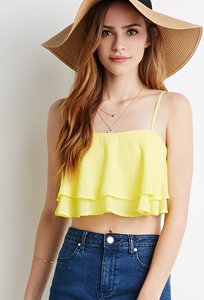}\hss\includegraphics[width=1.10cm,height=1.8cm,keepaspectratio]{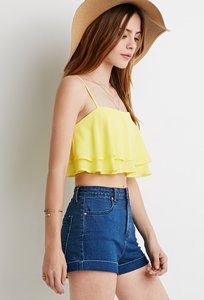}\hss\includegraphics[width=1.10cm,height=1.8cm,keepaspectratio]{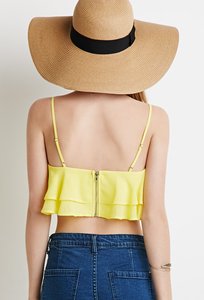}\hss\phantom{\rule{1.10cm}{1.8cm}}\hss\phantom{\rule{1.10cm}{1.8cm}}\hss}
      \end{minipage}
    }
  \end{minipage}
\hfill
  \begin{minipage}[t]{5.55cm}
    \fcolorbox{red!70!black}{white}{
      \begin{minipage}[t]{5.25cm}
        {\tiny\textbf{\#9}}\\
        \noindent\hbox to 5.25cm{\includegraphics[width=1.10cm,height=1.8cm,keepaspectratio]{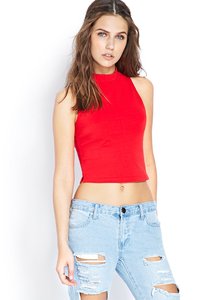}\hss\includegraphics[width=1.10cm,height=1.8cm,keepaspectratio]{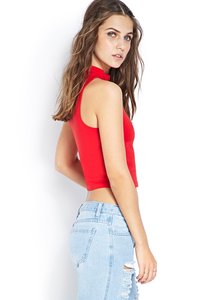}\hss\includegraphics[width=1.10cm,height=1.8cm,keepaspectratio]{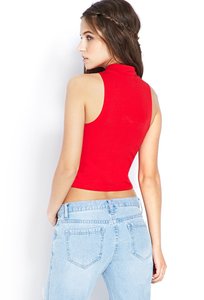}\hss\includegraphics[width=1.10cm,height=1.8cm,keepaspectratio]{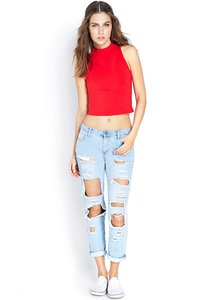}\hss\phantom{\rule{1.10cm}{1.8cm}}\hss}
      \end{minipage}
    }
  \end{minipage}
\par\vspace{2pt}
\noindent
  \begin{minipage}[t]{5.55cm}
    \fcolorbox{red!70!black}{white}{
      \begin{minipage}[t]{5.25cm}
        {\tiny\textbf{\#10}}\\
        \noindent\hbox to 5.25cm{\includegraphics[width=1.10cm,height=1.8cm,keepaspectratio]{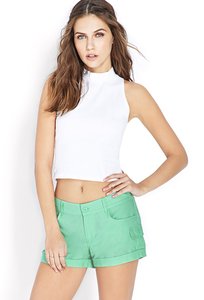}\hss\includegraphics[width=1.10cm,height=1.8cm,keepaspectratio]{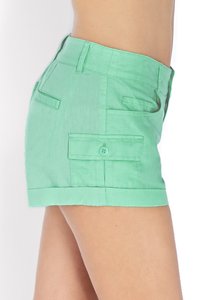}\hss\includegraphics[width=1.10cm,height=1.8cm,keepaspectratio]{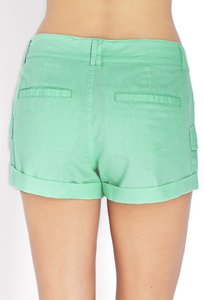}\hss\includegraphics[width=1.10cm,height=1.8cm,keepaspectratio]{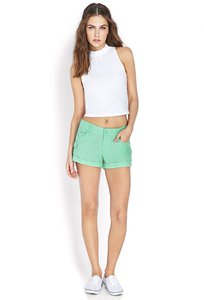}\hss\includegraphics[width=1.10cm,height=1.8cm,keepaspectratio]{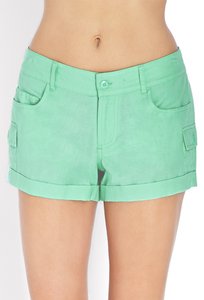}\hss}
      \end{minipage}
    }
  \end{minipage}
\hfill
  \begin{minipage}[t]{5.55cm}\end{minipage}
\hfill
  \begin{minipage}[t]{5.55cm}\end{minipage}
\par\vspace{2pt}
\noindent\rule{\linewidth}{0.4pt}
\par\vspace{20pt}

\par\vspace{16pt}
\noindent\textbf{\large Example 2}
\par\vspace{4pt}
\noindent\rule{\linewidth}{1.2pt}
\par\vspace{4pt}
% ── Case 2: short::deepfashion::1702 ──
\noindent\hfill%
  \begin{minipage}[t]{6.0cm}
    \noindent\hbox to 6.0cm{\includegraphics[width=1.20cm,height=2.0cm,keepaspectratio]{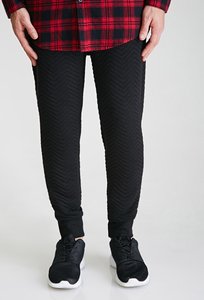}\hss\includegraphics[width=1.20cm,height=2.0cm,keepaspectratio]{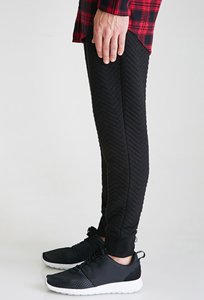}\hss\includegraphics[width=1.20cm,height=2.0cm,keepaspectratio]{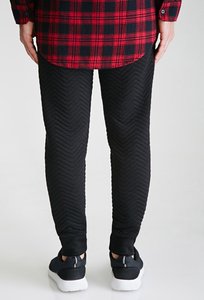}\hss\includegraphics[width=1.20cm,height=2.0cm,keepaspectratio]{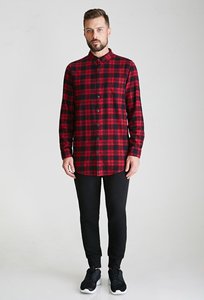}\hss\phantom{\rule{1.20cm}{2.0cm}}\hss}
    \par\vspace{1pt}
    {\scriptsize\textbf{}}
    \par\vspace{0pt}
    \parbox[t]{6.0cm}{\tiny\raggedright Black chevron-quilted jogger pants featuring elastic ankle cuffs, a slim tapered fit, and dimensional textured knit fabric. The all-over chevron pattern creates visual depth while the mid-rise waist and side pockets offer functionality. Styled with a red plaid flannel shirt and black sneakers for a casual, contemporary look.}
  \end{minipage}%
\hfill%
  \begin{minipage}[t]{3.5cm}
    \centering
    \vspace{0.45cm}%
    \parbox{3.5cm}{\centering\tiny Switch from chevron-quilted knit fabric to smooth technical woven fabric; add an asymmetrical rear welt pocket and a waistband snap closure.}\\[2pt]
    {\normalsize$\longrightarrow$}
  \end{minipage}%
\hfill%
  \begin{minipage}[t]{6.0cm}
    \noindent\hbox to 6.0cm{\includegraphics[width=1.20cm,height=2.0cm,keepaspectratio]{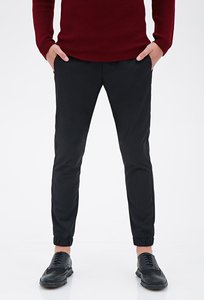}\hss\includegraphics[width=1.20cm,height=2.0cm,keepaspectratio]{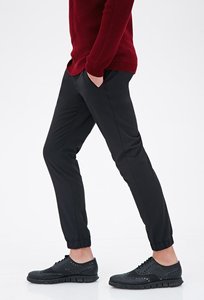}\hss\includegraphics[width=1.20cm,height=2.0cm,keepaspectratio]{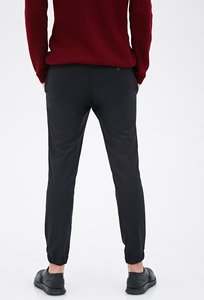}\hss\includegraphics[width=1.20cm,height=2.0cm,keepaspectratio]{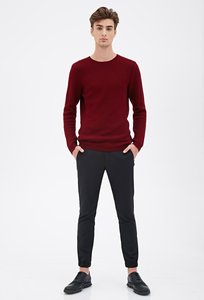}\hss\phantom{\rule{1.20cm}{2.0cm}}\hss}
    \par\vspace{1pt}
    {\scriptsize\textbf{Ground Truth}}
    \par\vspace{0pt}
    \parbox[t]{6.0cm}{\tiny\raggedright Men's black slim-fit jogger pants crafted from smooth woven technical fabric with a subtle sheen. Features elasticized ribbed ankle cuffs, slanted side pockets, and a tapered silhouette with mid-rise flat-front waist. The refined construction bridges athletic comfort and tailored sophistication, suitable for versatile smart-casual styling.}
  \end{minipage}%
\hfill
\par\vspace{4pt}
\par\vspace{4pt}
\noindent\rule{\linewidth}{0.4pt}
% -- mt_align --
{\small\textbf{\textbf{Ours}}}\quad{\small Rank~1}\\
\noindent
  \begin{minipage}[t]{5.55cm}
    \fcolorbox{green!60!black}{white}{
      \begin{minipage}[t]{5.25cm}
        {\tiny\textbf{\#1}}\\
        \noindent\hbox to 5.25cm{\includegraphics[width=1.10cm,height=1.8cm,keepaspectratio]{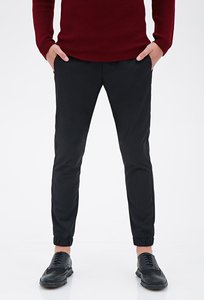}\hss\includegraphics[width=1.10cm,height=1.8cm,keepaspectratio]{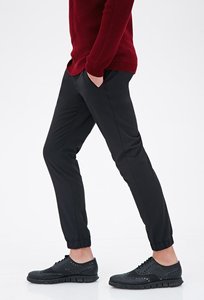}\hss\includegraphics[width=1.10cm,height=1.8cm,keepaspectratio]{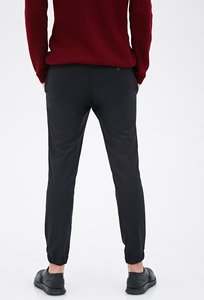}\hss\includegraphics[width=1.10cm,height=1.8cm,keepaspectratio]{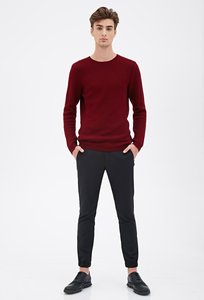}\hss\phantom{\rule{1.10cm}{1.8cm}}\hss}
      \end{minipage}
    }
  \end{minipage}
\hfill
  \begin{minipage}[t]{5.55cm}
    \fcolorbox{red!70!black}{white}{
      \begin{minipage}[t]{5.25cm}
        {\tiny\textbf{\#2}}\\
        \noindent\hbox to 5.25cm{\includegraphics[width=1.10cm,height=1.8cm,keepaspectratio]{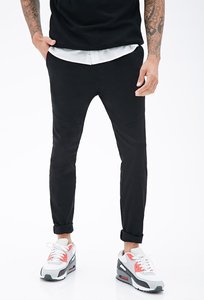}\hss\includegraphics[width=1.10cm,height=1.8cm,keepaspectratio]{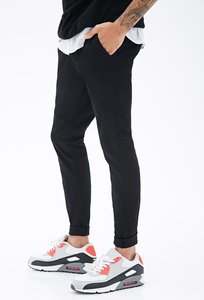}\hss\includegraphics[width=1.10cm,height=1.8cm,keepaspectratio]{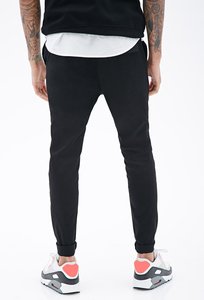}\hss\includegraphics[width=1.10cm,height=1.8cm,keepaspectratio]{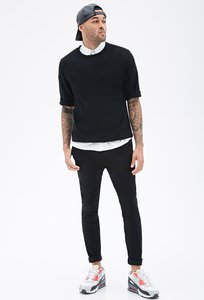}\hss\phantom{\rule{1.10cm}{1.8cm}}\hss}
      \end{minipage}
    }
  \end{minipage}
\hfill
  \begin{minipage}[t]{5.55cm}
    \fcolorbox{red!70!black}{white}{
      \begin{minipage}[t]{5.25cm}
        {\tiny\textbf{\#3}}\\
        \noindent\hbox to 5.25cm{\includegraphics[width=1.10cm,height=1.8cm,keepaspectratio]{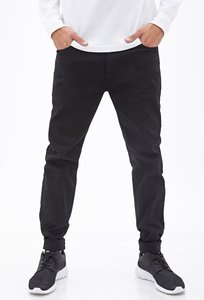}\hss\includegraphics[width=1.10cm,height=1.8cm,keepaspectratio]{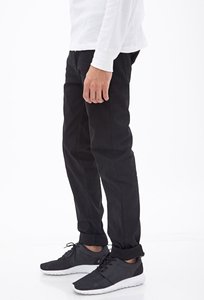}\hss\includegraphics[width=1.10cm,height=1.8cm,keepaspectratio]{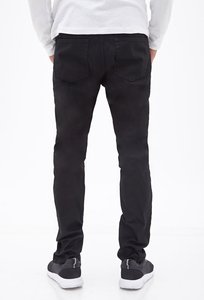}\hss\includegraphics[width=1.10cm,height=1.8cm,keepaspectratio]{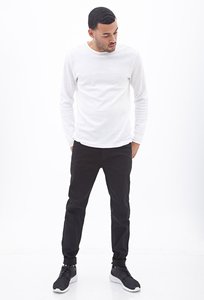}\hss\phantom{\rule{1.10cm}{1.8cm}}\hss}
      \end{minipage}
    }
  \end{minipage}
\par\vspace{2pt}
\noindent
  \begin{minipage}[t]{5.55cm}
    \fcolorbox{red!70!black}{white}{
      \begin{minipage}[t]{5.25cm}
        {\tiny\textbf{\#4}}\\
        \noindent\hbox to 5.25cm{\includegraphics[width=1.10cm,height=1.8cm,keepaspectratio]{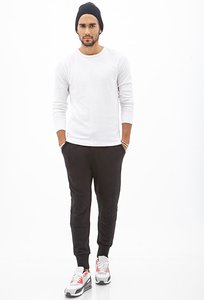}\hss\includegraphics[width=1.10cm,height=1.8cm,keepaspectratio]{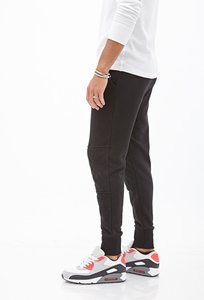}\hss\includegraphics[width=1.10cm,height=1.8cm,keepaspectratio]{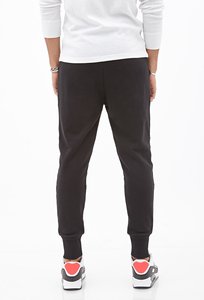}\hss\includegraphics[width=1.10cm,height=1.8cm,keepaspectratio]{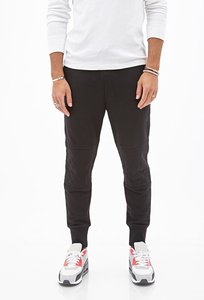}\hss\phantom{\rule{1.10cm}{1.8cm}}\hss}
      \end{minipage}
    }
  \end{minipage}
\hfill
  \begin{minipage}[t]{5.55cm}
    \fcolorbox{red!70!black}{white}{
      \begin{minipage}[t]{5.25cm}
        {\tiny\textbf{\#5}}\\
        \noindent\hbox to 5.25cm{\includegraphics[width=1.10cm,height=1.8cm,keepaspectratio]{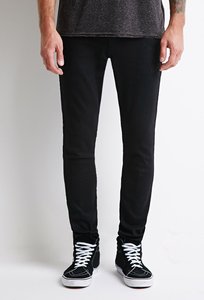}\hss\includegraphics[width=1.10cm,height=1.8cm,keepaspectratio]{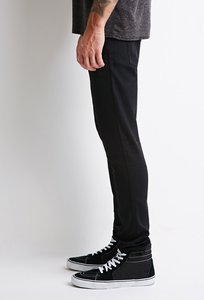}\hss\includegraphics[width=1.10cm,height=1.8cm,keepaspectratio]{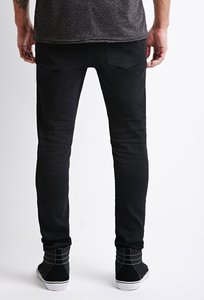}\hss\includegraphics[width=1.10cm,height=1.8cm,keepaspectratio]{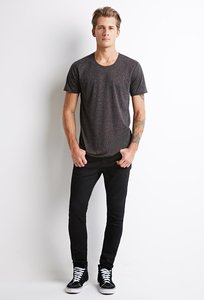}\hss\phantom{\rule{1.10cm}{1.8cm}}\hss}
      \end{minipage}
    }
  \end{minipage}
\hfill
  \begin{minipage}[t]{5.55cm}
    \fcolorbox{red!70!black}{white}{
      \begin{minipage}[t]{5.25cm}
        {\tiny\textbf{\#6}}\\
        \noindent\hbox to 5.25cm{\includegraphics[width=1.10cm,height=1.8cm,keepaspectratio]{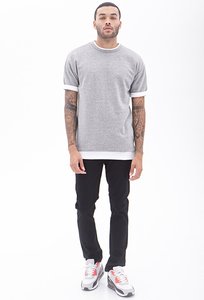}\hss\includegraphics[width=1.10cm,height=1.8cm,keepaspectratio]{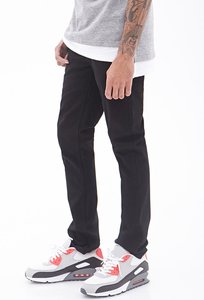}\hss\includegraphics[width=1.10cm,height=1.8cm,keepaspectratio]{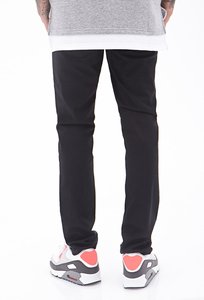}\hss\includegraphics[width=1.10cm,height=1.8cm,keepaspectratio]{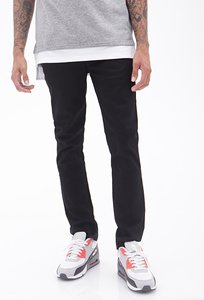}\hss\phantom{\rule{1.10cm}{1.8cm}}\hss}
      \end{minipage}
    }
  \end{minipage}
\par\vspace{2pt}
\noindent
  \begin{minipage}[t]{5.55cm}
    \fcolorbox{red!70!black}{white}{
      \begin{minipage}[t]{5.25cm}
        {\tiny\textbf{\#7}}\\
        \noindent\hbox to 5.25cm{\includegraphics[width=1.10cm,height=1.8cm,keepaspectratio]{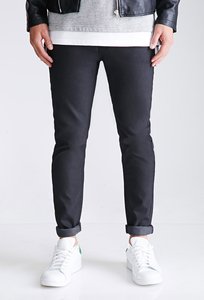}\hss\includegraphics[width=1.10cm,height=1.8cm,keepaspectratio]{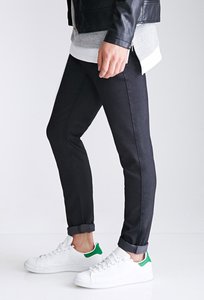}\hss\includegraphics[width=1.10cm,height=1.8cm,keepaspectratio]{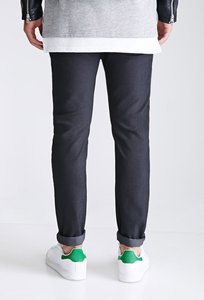}\hss\includegraphics[width=1.10cm,height=1.8cm,keepaspectratio]{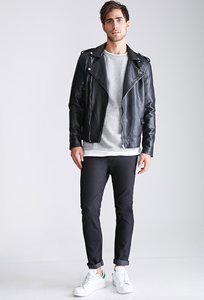}\hss\phantom{\rule{1.10cm}{1.8cm}}\hss}
      \end{minipage}
    }
  \end{minipage}
\hfill
  \begin{minipage}[t]{5.55cm}
    \fcolorbox{red!70!black}{white}{
      \begin{minipage}[t]{5.25cm}
        {\tiny\textbf{\#8}}\\
        \noindent\hbox to 5.25cm{\includegraphics[width=1.10cm,height=1.8cm,keepaspectratio]{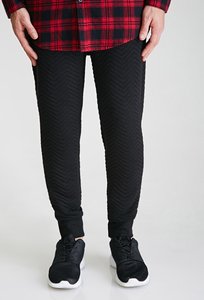}\hss\includegraphics[width=1.10cm,height=1.8cm,keepaspectratio]{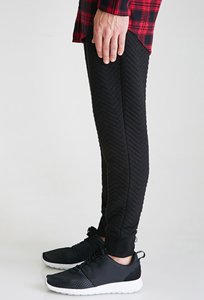}\hss\includegraphics[width=1.10cm,height=1.8cm,keepaspectratio]{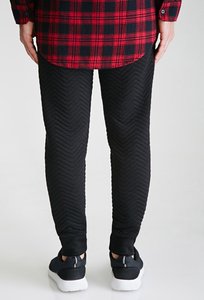}\hss\includegraphics[width=1.10cm,height=1.8cm,keepaspectratio]{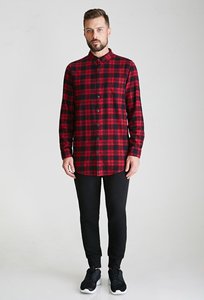}\hss\phantom{\rule{1.10cm}{1.8cm}}\hss}
      \end{minipage}
    }
  \end{minipage}
\hfill
  \begin{minipage}[t]{5.55cm}
    \fcolorbox{red!70!black}{white}{
      \begin{minipage}[t]{5.25cm}
        {\tiny\textbf{\#9}}\\
        \noindent\hbox to 5.25cm{\includegraphics[width=1.10cm,height=1.8cm,keepaspectratio]{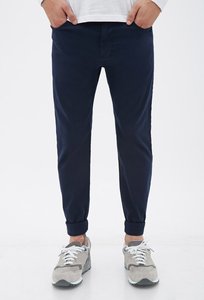}\hss\includegraphics[width=1.10cm,height=1.8cm,keepaspectratio]{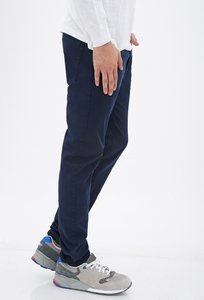}\hss\includegraphics[width=1.10cm,height=1.8cm,keepaspectratio]{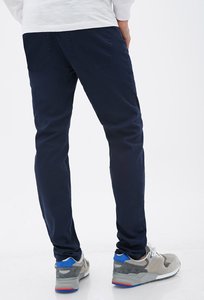}\hss\includegraphics[width=1.10cm,height=1.8cm,keepaspectratio]{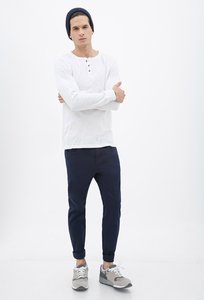}\hss\phantom{\rule{1.10cm}{1.8cm}}\hss}
      \end{minipage}
    }
  \end{minipage}
\par\vspace{2pt}
\noindent
  \begin{minipage}[t]{5.55cm}
    \fcolorbox{red!70!black}{white}{
      \begin{minipage}[t]{5.25cm}
        {\tiny\textbf{\#10}}\\
        \noindent\hbox to 5.25cm{\includegraphics[width=1.10cm,height=1.8cm,keepaspectratio]{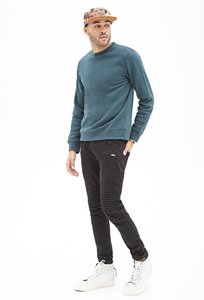}\hss\includegraphics[width=1.10cm,height=1.8cm,keepaspectratio]{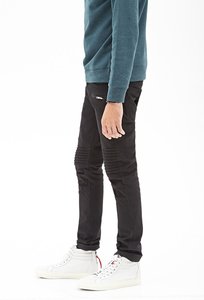}\hss\includegraphics[width=1.10cm,height=1.8cm,keepaspectratio]{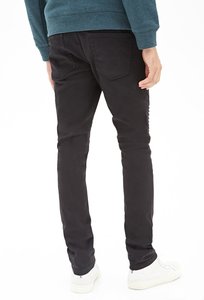}\hss\includegraphics[width=1.10cm,height=1.8cm,keepaspectratio]{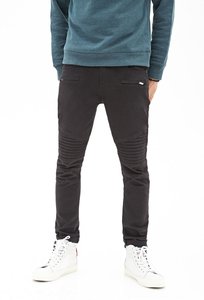}\hss\phantom{\rule{1.10cm}{1.8cm}}\hss}
      \end{minipage}
    }
  \end{minipage}
\hfill
  \begin{minipage}[t]{5.55cm}\end{minipage}
\hfill
  \begin{minipage}[t]{5.55cm}\end{minipage}
\par\vspace{2pt}
\noindent\rule{\linewidth}{0.4pt}
\par\vspace{2pt}
% -- qwen3_vl_2b --
{\small\textbf{Qwen3-VL-2B}}\quad{\small Rank~4}\\
\noindent
  \begin{minipage}[t]{5.55cm}
    \fcolorbox{red!70!black}{white}{
      \begin{minipage}[t]{5.25cm}
        {\tiny\textbf{\#1}}\\
        \noindent\hbox to 5.25cm{\includegraphics[width=1.10cm,height=1.8cm,keepaspectratio]{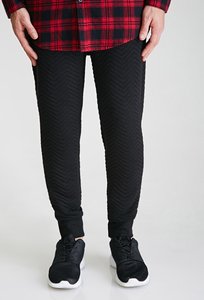}\hss\includegraphics[width=1.10cm,height=1.8cm,keepaspectratio]{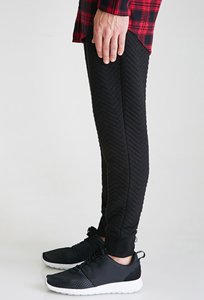}\hss\includegraphics[width=1.10cm,height=1.8cm,keepaspectratio]{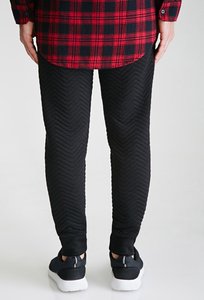}\hss\includegraphics[width=1.10cm,height=1.8cm,keepaspectratio]{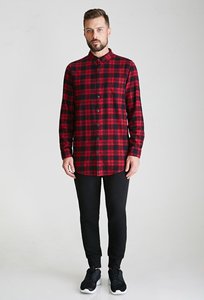}\hss\phantom{\rule{1.10cm}{1.8cm}}\hss}
      \end{minipage}
    }
  \end{minipage}
\hfill
  \begin{minipage}[t]{5.55cm}
    \fcolorbox{red!70!black}{white}{
      \begin{minipage}[t]{5.25cm}
        {\tiny\textbf{\#2}}\\
        \noindent\hbox to 5.25cm{\includegraphics[width=1.10cm,height=1.8cm,keepaspectratio]{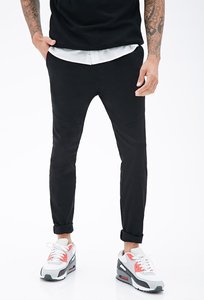}\hss\includegraphics[width=1.10cm,height=1.8cm,keepaspectratio]{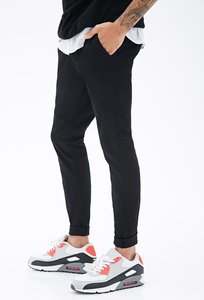}\hss\includegraphics[width=1.10cm,height=1.8cm,keepaspectratio]{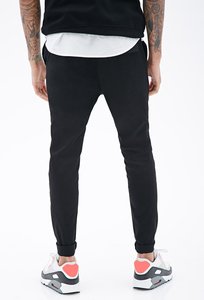}\hss\includegraphics[width=1.10cm,height=1.8cm,keepaspectratio]{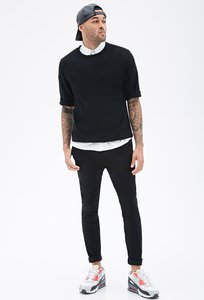}\hss\phantom{\rule{1.10cm}{1.8cm}}\hss}
      \end{minipage}
    }
  \end{minipage}
\hfill
  \begin{minipage}[t]{5.55cm}
    \fcolorbox{red!70!black}{white}{
      \begin{minipage}[t]{5.25cm}
        {\tiny\textbf{\#3}}\\
        \noindent\hbox to 5.25cm{\includegraphics[width=1.10cm,height=1.8cm,keepaspectratio]{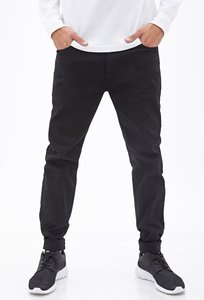}\hss\includegraphics[width=1.10cm,height=1.8cm,keepaspectratio]{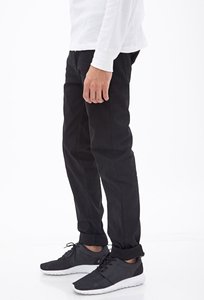}\hss\includegraphics[width=1.10cm,height=1.8cm,keepaspectratio]{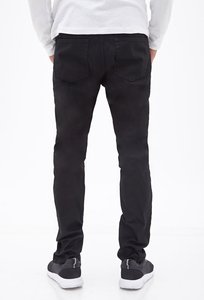}\hss\includegraphics[width=1.10cm,height=1.8cm,keepaspectratio]{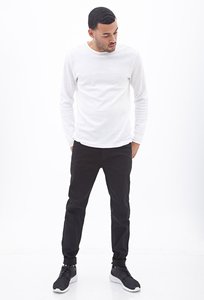}\hss\phantom{\rule{1.10cm}{1.8cm}}\hss}
      \end{minipage}
    }
  \end{minipage}
\par\vspace{2pt}
\noindent
  \begin{minipage}[t]{5.55cm}
    \fcolorbox{green!60!black}{white}{
      \begin{minipage}[t]{5.25cm}
        {\tiny\textbf{\#4}}\\
        \noindent\hbox to 5.25cm{\includegraphics[width=1.10cm,height=1.8cm,keepaspectratio]{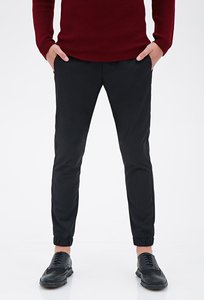}\hss\includegraphics[width=1.10cm,height=1.8cm,keepaspectratio]{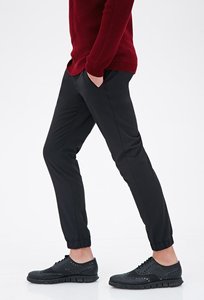}\hss\includegraphics[width=1.10cm,height=1.8cm,keepaspectratio]{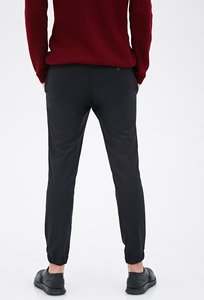}\hss\includegraphics[width=1.10cm,height=1.8cm,keepaspectratio]{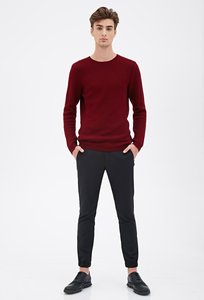}\hss\phantom{\rule{1.10cm}{1.8cm}}\hss}
      \end{minipage}
    }
  \end{minipage}
\hfill
  \begin{minipage}[t]{5.55cm}
    \fcolorbox{red!70!black}{white}{
      \begin{minipage}[t]{5.25cm}
        {\tiny\textbf{\#5}}\\
        \noindent\hbox to 5.25cm{\includegraphics[width=1.10cm,height=1.8cm,keepaspectratio]{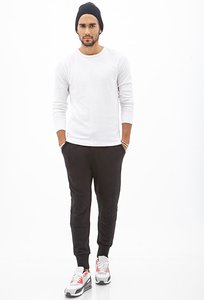}\hss\includegraphics[width=1.10cm,height=1.8cm,keepaspectratio]{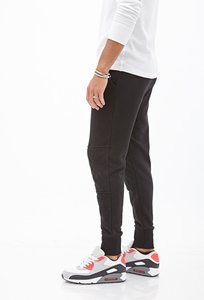}\hss\includegraphics[width=1.10cm,height=1.8cm,keepaspectratio]{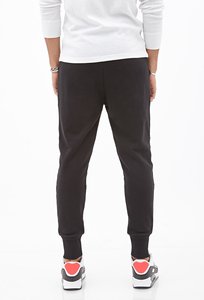}\hss\includegraphics[width=1.10cm,height=1.8cm,keepaspectratio]{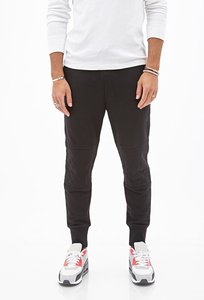}\hss\phantom{\rule{1.10cm}{1.8cm}}\hss}
      \end{minipage}
    }
  \end{minipage}
\hfill
  \begin{minipage}[t]{5.55cm}
    \fcolorbox{red!70!black}{white}{
      \begin{minipage}[t]{5.25cm}
        {\tiny\textbf{\#6}}\\
        \noindent\hbox to 5.25cm{\includegraphics[width=1.10cm,height=1.8cm,keepaspectratio]{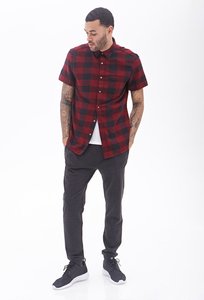}\hss\includegraphics[width=1.10cm,height=1.8cm,keepaspectratio]{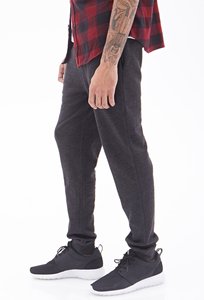}\hss\includegraphics[width=1.10cm,height=1.8cm,keepaspectratio]{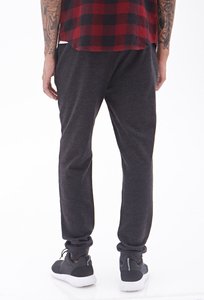}\hss\includegraphics[width=1.10cm,height=1.8cm,keepaspectratio]{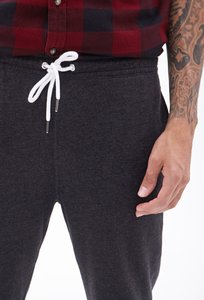}\hss\phantom{\rule{1.10cm}{1.8cm}}\hss}
      \end{minipage}
    }
  \end{minipage}
\par\vspace{2pt}
\noindent
  \begin{minipage}[t]{5.55cm}
    \fcolorbox{red!70!black}{white}{
      \begin{minipage}[t]{5.25cm}
        {\tiny\textbf{\#7}}\\
        \noindent\hbox to 5.25cm{\includegraphics[width=1.10cm,height=1.8cm,keepaspectratio]{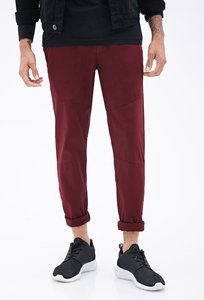}\hss\includegraphics[width=1.10cm,height=1.8cm,keepaspectratio]{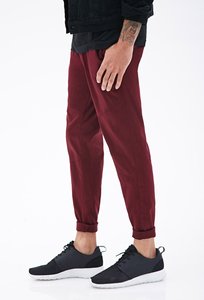}\hss\includegraphics[width=1.10cm,height=1.8cm,keepaspectratio]{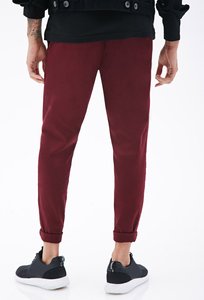}\hss\includegraphics[width=1.10cm,height=1.8cm,keepaspectratio]{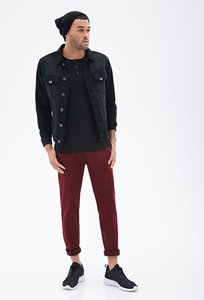}\hss\phantom{\rule{1.10cm}{1.8cm}}\hss}
      \end{minipage}
    }
  \end{minipage}
\hfill
  \begin{minipage}[t]{5.55cm}
    \fcolorbox{red!70!black}{white}{
      \begin{minipage}[t]{5.25cm}
        {\tiny\textbf{\#8}}\\
        \noindent\hbox to 5.25cm{\includegraphics[width=1.10cm,height=1.8cm,keepaspectratio]{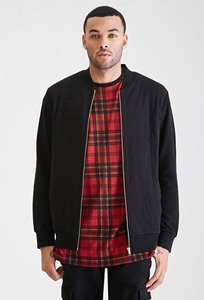}\hss\includegraphics[width=1.10cm,height=1.8cm,keepaspectratio]{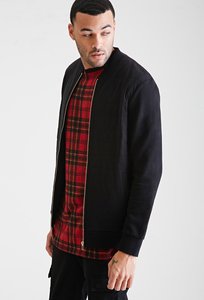}\hss\includegraphics[width=1.10cm,height=1.8cm,keepaspectratio]{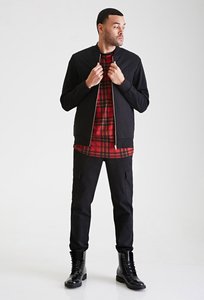}\hss\includegraphics[width=1.10cm,height=1.8cm,keepaspectratio]{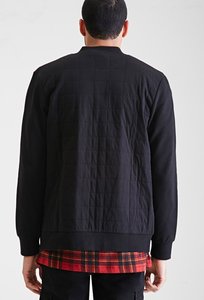}\hss\phantom{\rule{1.10cm}{1.8cm}}\hss}
      \end{minipage}
    }
  \end{minipage}
\hfill
  \begin{minipage}[t]{5.55cm}
    \fcolorbox{red!70!black}{white}{
      \begin{minipage}[t]{5.25cm}
        {\tiny\textbf{\#9}}\\
        \noindent\hbox to 5.25cm{\includegraphics[width=1.10cm,height=1.8cm,keepaspectratio]{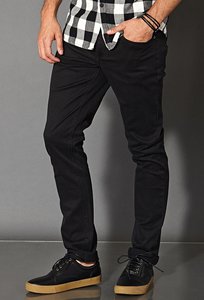}\hss\includegraphics[width=1.10cm,height=1.8cm,keepaspectratio]{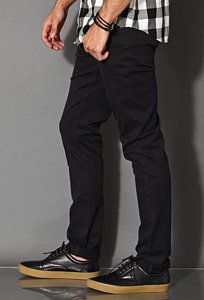}\hss\includegraphics[width=1.10cm,height=1.8cm,keepaspectratio]{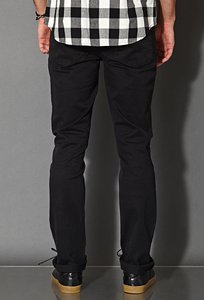}\hss\includegraphics[width=1.10cm,height=1.8cm,keepaspectratio]{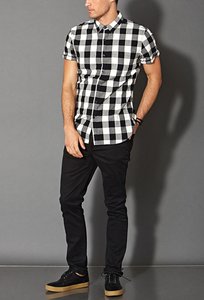}\hss\includegraphics[width=1.10cm,height=1.8cm,keepaspectratio]{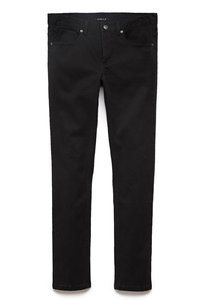}\hss}
      \end{minipage}
    }
  \end{minipage}
\par\vspace{2pt}
\noindent
  \begin{minipage}[t]{5.55cm}
    \fcolorbox{red!70!black}{white}{
      \begin{minipage}[t]{5.25cm}
        {\tiny\textbf{\#10}}\\
        \noindent\hbox to 5.25cm{\includegraphics[width=1.10cm,height=1.8cm,keepaspectratio]{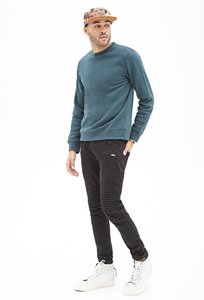}\hss\includegraphics[width=1.10cm,height=1.8cm,keepaspectratio]{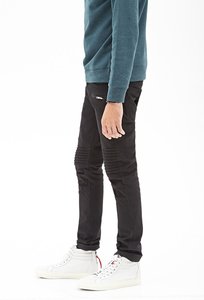}\hss\includegraphics[width=1.10cm,height=1.8cm,keepaspectratio]{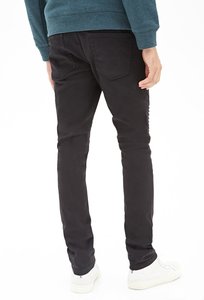}\hss\includegraphics[width=1.10cm,height=1.8cm,keepaspectratio]{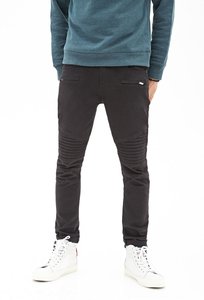}\hss\phantom{\rule{1.10cm}{1.8cm}}\hss}
      \end{minipage}
    }
  \end{minipage}
\hfill
  \begin{minipage}[t]{5.55cm}\end{minipage}
\hfill
  \begin{minipage}[t]{5.55cm}\end{minipage}
\par\vspace{2pt}
\noindent\rule{\linewidth}{0.4pt}
\par\vspace{2pt}
% -- qwen3_vl_8b --
{\small\textbf{Qwen3-VL-8B}}\quad{\small Rank~4}\\
\noindent
  \begin{minipage}[t]{5.55cm}
    \fcolorbox{red!70!black}{white}{
      \begin{minipage}[t]{5.25cm}
        {\tiny\textbf{\#1}}\\
        \noindent\hbox to 5.25cm{\includegraphics[width=1.10cm,height=1.8cm,keepaspectratio]{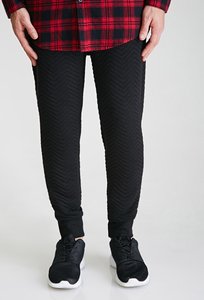}\hss\includegraphics[width=1.10cm,height=1.8cm,keepaspectratio]{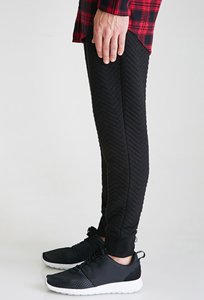}\hss\includegraphics[width=1.10cm,height=1.8cm,keepaspectratio]{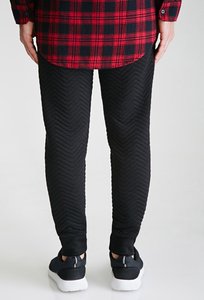}\hss\includegraphics[width=1.10cm,height=1.8cm,keepaspectratio]{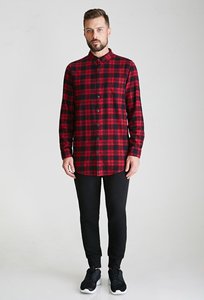}\hss\phantom{\rule{1.10cm}{1.8cm}}\hss}
      \end{minipage}
    }
  \end{minipage}
\hfill
  \begin{minipage}[t]{5.55cm}
    \fcolorbox{red!70!black}{white}{
      \begin{minipage}[t]{5.25cm}
        {\tiny\textbf{\#2}}\\
        \noindent\hbox to 5.25cm{\includegraphics[width=1.10cm,height=1.8cm,keepaspectratio]{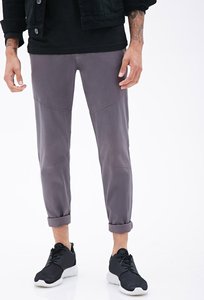}\hss\includegraphics[width=1.10cm,height=1.8cm,keepaspectratio]{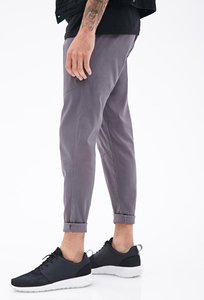}\hss\includegraphics[width=1.10cm,height=1.8cm,keepaspectratio]{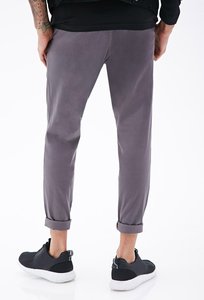}\hss\includegraphics[width=1.10cm,height=1.8cm,keepaspectratio]{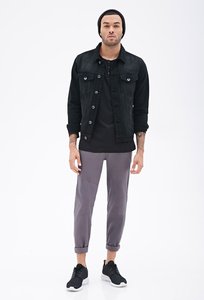}\hss\phantom{\rule{1.10cm}{1.8cm}}\hss}
      \end{minipage}
    }
  \end{minipage}
\hfill
  \begin{minipage}[t]{5.55cm}
    \fcolorbox{red!70!black}{white}{
      \begin{minipage}[t]{5.25cm}
        {\tiny\textbf{\#3}}\\
        \noindent\hbox to 5.25cm{\includegraphics[width=1.10cm,height=1.8cm,keepaspectratio]{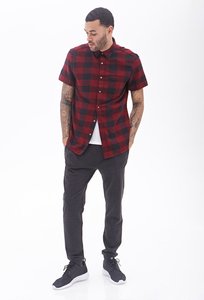}\hss\includegraphics[width=1.10cm,height=1.8cm,keepaspectratio]{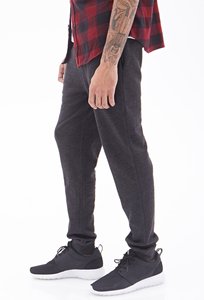}\hss\includegraphics[width=1.10cm,height=1.8cm,keepaspectratio]{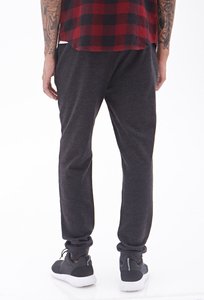}\hss\includegraphics[width=1.10cm,height=1.8cm,keepaspectratio]{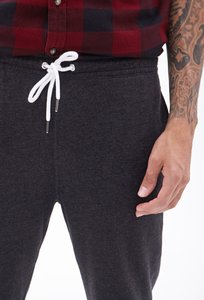}\hss\phantom{\rule{1.10cm}{1.8cm}}\hss}
      \end{minipage}
    }
  \end{minipage}
\par\vspace{2pt}
\noindent
  \begin{minipage}[t]{5.55cm}
    \fcolorbox{green!60!black}{white}{
      \begin{minipage}[t]{5.25cm}
        {\tiny\textbf{\#4}}\\
        \noindent\hbox to 5.25cm{\includegraphics[width=1.10cm,height=1.8cm,keepaspectratio]{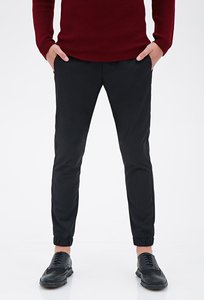}\hss\includegraphics[width=1.10cm,height=1.8cm,keepaspectratio]{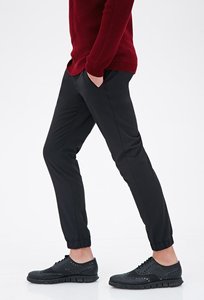}\hss\includegraphics[width=1.10cm,height=1.8cm,keepaspectratio]{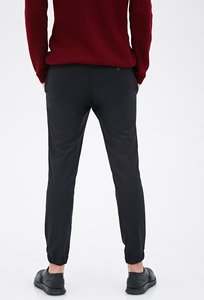}\hss\includegraphics[width=1.10cm,height=1.8cm,keepaspectratio]{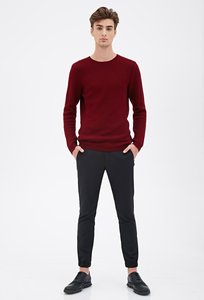}\hss\phantom{\rule{1.10cm}{1.8cm}}\hss}
      \end{minipage}
    }
  \end{minipage}
\hfill
  \begin{minipage}[t]{5.55cm}
    \fcolorbox{red!70!black}{white}{
      \begin{minipage}[t]{5.25cm}
        {\tiny\textbf{\#5}}\\
        \noindent\hbox to 5.25cm{\includegraphics[width=1.10cm,height=1.8cm,keepaspectratio]{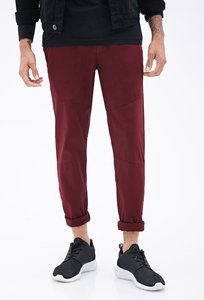}\hss\includegraphics[width=1.10cm,height=1.8cm,keepaspectratio]{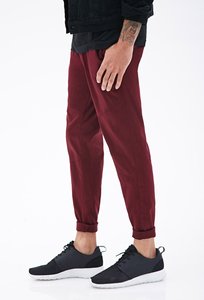}\hss\includegraphics[width=1.10cm,height=1.8cm,keepaspectratio]{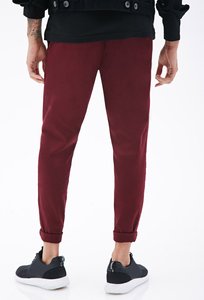}\hss\includegraphics[width=1.10cm,height=1.8cm,keepaspectratio]{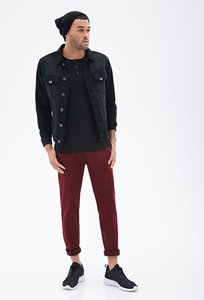}\hss\phantom{\rule{1.10cm}{1.8cm}}\hss}
      \end{minipage}
    }
  \end{minipage}
\hfill
  \begin{minipage}[t]{5.55cm}
    \fcolorbox{red!70!black}{white}{
      \begin{minipage}[t]{5.25cm}
        {\tiny\textbf{\#6}}\\
        \noindent\hbox to 5.25cm{\includegraphics[width=1.10cm,height=1.8cm,keepaspectratio]{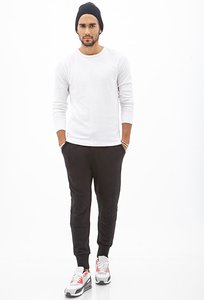}\hss\includegraphics[width=1.10cm,height=1.8cm,keepaspectratio]{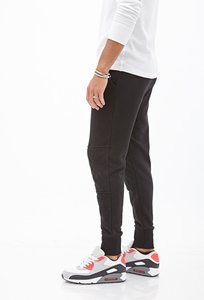}\hss\includegraphics[width=1.10cm,height=1.8cm,keepaspectratio]{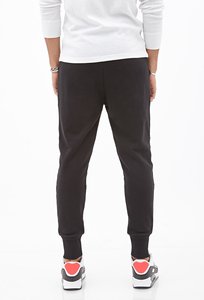}\hss\includegraphics[width=1.10cm,height=1.8cm,keepaspectratio]{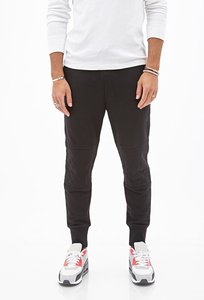}\hss\phantom{\rule{1.10cm}{1.8cm}}\hss}
      \end{minipage}
    }
  \end{minipage}
\par\vspace{2pt}
\noindent
  \begin{minipage}[t]{5.55cm}
    \fcolorbox{red!70!black}{white}{
      \begin{minipage}[t]{5.25cm}
        {\tiny\textbf{\#7}}\\
        \noindent\hbox to 5.25cm{\includegraphics[width=1.10cm,height=1.8cm,keepaspectratio]{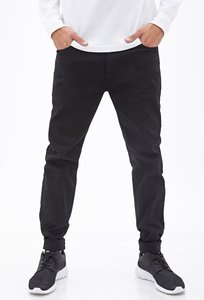}\hss\includegraphics[width=1.10cm,height=1.8cm,keepaspectratio]{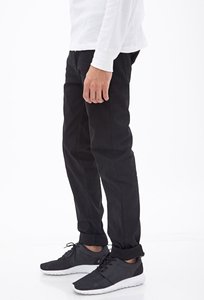}\hss\includegraphics[width=1.10cm,height=1.8cm,keepaspectratio]{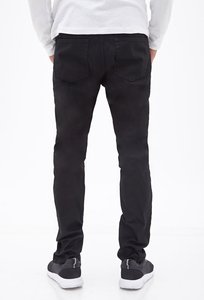}\hss\includegraphics[width=1.10cm,height=1.8cm,keepaspectratio]{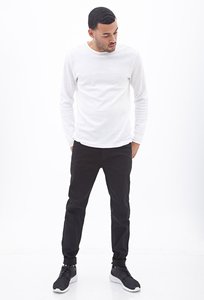}\hss\phantom{\rule{1.10cm}{1.8cm}}\hss}
      \end{minipage}
    }
  \end{minipage}
\hfill
  \begin{minipage}[t]{5.55cm}
    \fcolorbox{red!70!black}{white}{
      \begin{minipage}[t]{5.25cm}
        {\tiny\textbf{\#8}}\\
        \noindent\hbox to 5.25cm{\includegraphics[width=1.10cm,height=1.8cm,keepaspectratio]{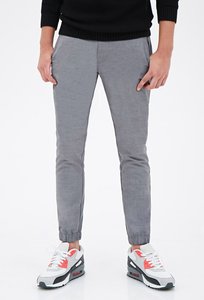}\hss\includegraphics[width=1.10cm,height=1.8cm,keepaspectratio]{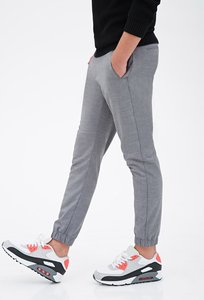}\hss\includegraphics[width=1.10cm,height=1.8cm,keepaspectratio]{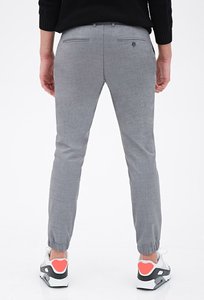}\hss\includegraphics[width=1.10cm,height=1.8cm,keepaspectratio]{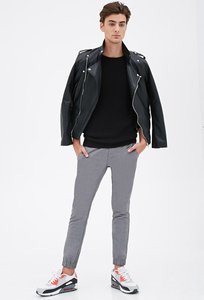}\hss\phantom{\rule{1.10cm}{1.8cm}}\hss}
      \end{minipage}
    }
  \end{minipage}
\hfill
  \begin{minipage}[t]{5.55cm}
    \fcolorbox{red!70!black}{white}{
      \begin{minipage}[t]{5.25cm}
        {\tiny\textbf{\#9}}\\
        \noindent\hbox to 5.25cm{\includegraphics[width=1.10cm,height=1.8cm,keepaspectratio]{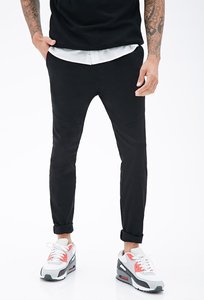}\hss\includegraphics[width=1.10cm,height=1.8cm,keepaspectratio]{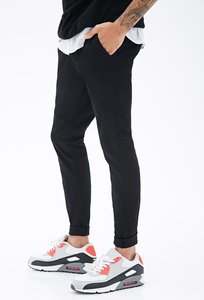}\hss\includegraphics[width=1.10cm,height=1.8cm,keepaspectratio]{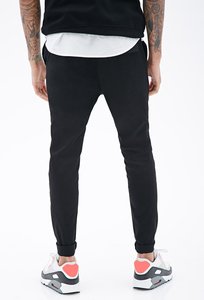}\hss\includegraphics[width=1.10cm,height=1.8cm,keepaspectratio]{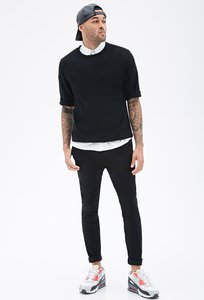}\hss\phantom{\rule{1.10cm}{1.8cm}}\hss}
      \end{minipage}
    }
  \end{minipage}
\par\vspace{2pt}
\noindent
  \begin{minipage}[t]{5.55cm}
    \fcolorbox{red!70!black}{white}{
      \begin{minipage}[t]{5.25cm}
        {\tiny\textbf{\#10}}\\
        \noindent\hbox to 5.25cm{\includegraphics[width=1.10cm,height=1.8cm,keepaspectratio]{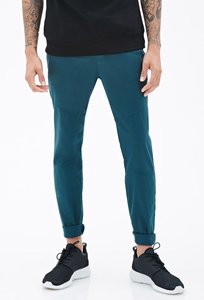}\hss\includegraphics[width=1.10cm,height=1.8cm,keepaspectratio]{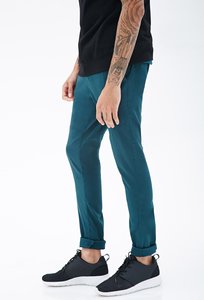}\hss\includegraphics[width=1.10cm,height=1.8cm,keepaspectratio]{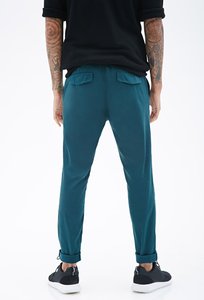}\hss\includegraphics[width=1.10cm,height=1.8cm,keepaspectratio]{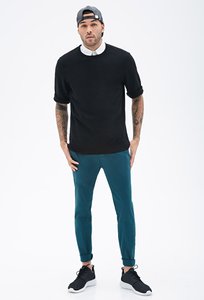}\hss\phantom{\rule{1.10cm}{1.8cm}}\hss}
      \end{minipage}
    }
  \end{minipage}
\hfill
  \begin{minipage}[t]{5.55cm}\end{minipage}
\hfill
  \begin{minipage}[t]{5.55cm}\end{minipage}
\par\vspace{2pt}
\noindent\rule{\linewidth}{0.4pt}
\par\vspace{2pt}
% -- reznembed --
{\small\textbf{RezNEmbed}}\quad{\small Rank~2}\\
\noindent
  \begin{minipage}[t]{5.55cm}
    \fcolorbox{red!70!black}{white}{
      \begin{minipage}[t]{5.25cm}
        {\tiny\textbf{\#1}}\\
        \noindent\hbox to 5.25cm{\includegraphics[width=1.10cm,height=1.8cm,keepaspectratio]{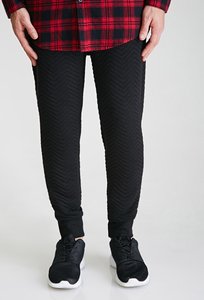}\hss\includegraphics[width=1.10cm,height=1.8cm,keepaspectratio]{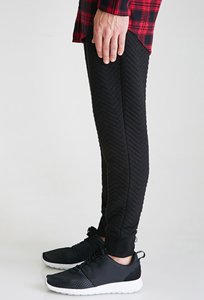}\hss\includegraphics[width=1.10cm,height=1.8cm,keepaspectratio]{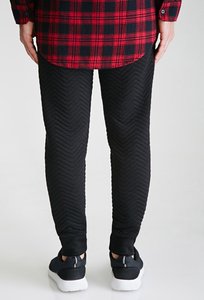}\hss\includegraphics[width=1.10cm,height=1.8cm,keepaspectratio]{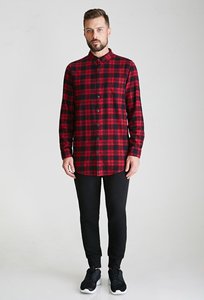}\hss\phantom{\rule{1.10cm}{1.8cm}}\hss}
      \end{minipage}
    }
  \end{minipage}
\hfill
  \begin{minipage}[t]{5.55cm}
    \fcolorbox{green!60!black}{white}{
      \begin{minipage}[t]{5.25cm}
        {\tiny\textbf{\#2}}\\
        \noindent\hbox to 5.25cm{\includegraphics[width=1.10cm,height=1.8cm,keepaspectratio]{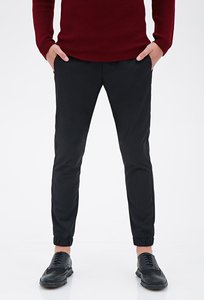}\hss\includegraphics[width=1.10cm,height=1.8cm,keepaspectratio]{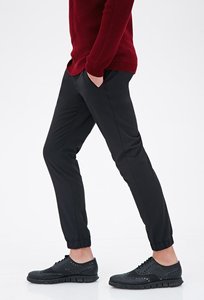}\hss\includegraphics[width=1.10cm,height=1.8cm,keepaspectratio]{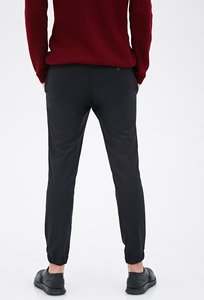}\hss\includegraphics[width=1.10cm,height=1.8cm,keepaspectratio]{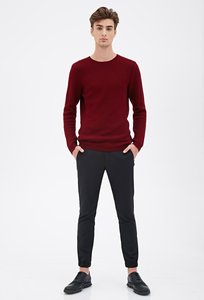}\hss\phantom{\rule{1.10cm}{1.8cm}}\hss}
      \end{minipage}
    }
  \end{minipage}
\hfill
  \begin{minipage}[t]{5.55cm}
    \fcolorbox{red!70!black}{white}{
      \begin{minipage}[t]{5.25cm}
        {\tiny\textbf{\#3}}\\
        \noindent\hbox to 5.25cm{\includegraphics[width=1.10cm,height=1.8cm,keepaspectratio]{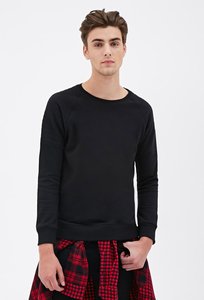}\hss\includegraphics[width=1.10cm,height=1.8cm,keepaspectratio]{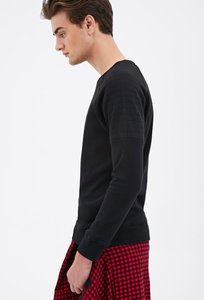}\hss\includegraphics[width=1.10cm,height=1.8cm,keepaspectratio]{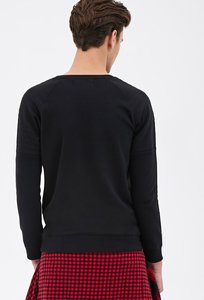}\hss\phantom{\rule{1.10cm}{1.8cm}}\hss\phantom{\rule{1.10cm}{1.8cm}}\hss}
      \end{minipage}
    }
  \end{minipage}
\par\vspace{2pt}
\noindent
  \begin{minipage}[t]{5.55cm}
    \fcolorbox{red!70!black}{white}{
      \begin{minipage}[t]{5.25cm}
        {\tiny\textbf{\#4}}\\
        \noindent\hbox to 5.25cm{\includegraphics[width=1.10cm,height=1.8cm,keepaspectratio]{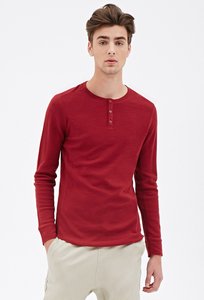}\hss\includegraphics[width=1.10cm,height=1.8cm,keepaspectratio]{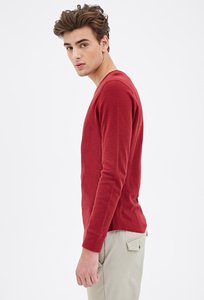}\hss\includegraphics[width=1.10cm,height=1.8cm,keepaspectratio]{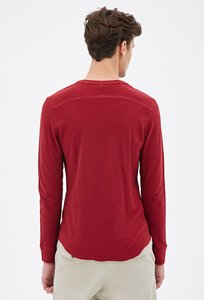}\hss\includegraphics[width=1.10cm,height=1.8cm,keepaspectratio]{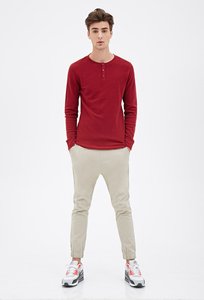}\hss\phantom{\rule{1.10cm}{1.8cm}}\hss}
      \end{minipage}
    }
  \end{minipage}
\hfill
  \begin{minipage}[t]{5.55cm}
    \fcolorbox{red!70!black}{white}{
      \begin{minipage}[t]{5.25cm}
        {\tiny\textbf{\#5}}\\
        \noindent\hbox to 5.25cm{\includegraphics[width=1.10cm,height=1.8cm,keepaspectratio]{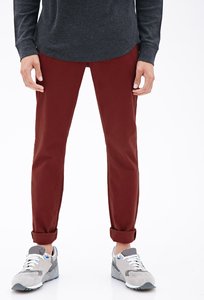}\hss\includegraphics[width=1.10cm,height=1.8cm,keepaspectratio]{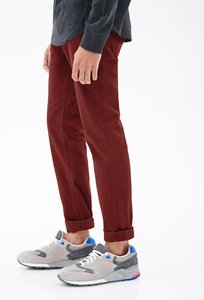}\hss\includegraphics[width=1.10cm,height=1.8cm,keepaspectratio]{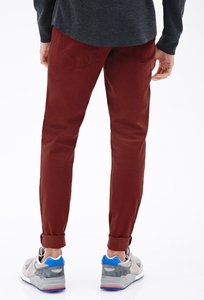}\hss\includegraphics[width=1.10cm,height=1.8cm,keepaspectratio]{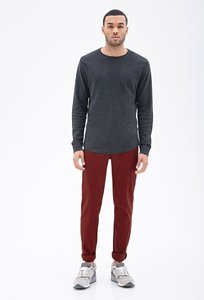}\hss\phantom{\rule{1.10cm}{1.8cm}}\hss}
      \end{minipage}
    }
  \end{minipage}
\hfill
  \begin{minipage}[t]{5.55cm}
    \fcolorbox{red!70!black}{white}{
      \begin{minipage}[t]{5.25cm}
        {\tiny\textbf{\#6}}\\
        \noindent\hbox to 5.25cm{\includegraphics[width=1.10cm,height=1.8cm,keepaspectratio]{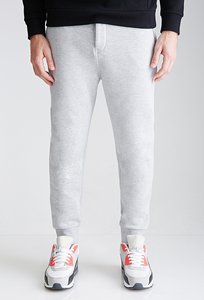}\hss\includegraphics[width=1.10cm,height=1.8cm,keepaspectratio]{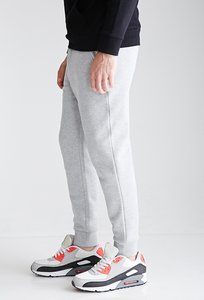}\hss\includegraphics[width=1.10cm,height=1.8cm,keepaspectratio]{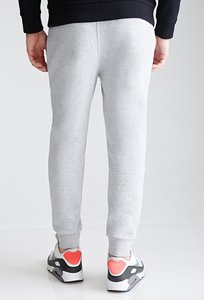}\hss\includegraphics[width=1.10cm,height=1.8cm,keepaspectratio]{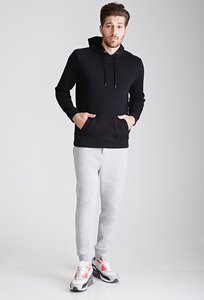}\hss\phantom{\rule{1.10cm}{1.8cm}}\hss}
      \end{minipage}
    }
  \end{minipage}
\par\vspace{2pt}
\noindent
  \begin{minipage}[t]{5.55cm}
    \fcolorbox{red!70!black}{white}{
      \begin{minipage}[t]{5.25cm}
        {\tiny\textbf{\#7}}\\
        \noindent\hbox to 5.25cm{\includegraphics[width=1.10cm,height=1.8cm,keepaspectratio]{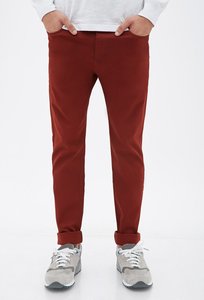}\hss\includegraphics[width=1.10cm,height=1.8cm,keepaspectratio]{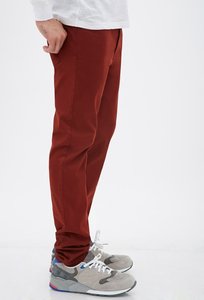}\hss\includegraphics[width=1.10cm,height=1.8cm,keepaspectratio]{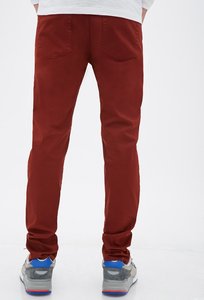}\hss\includegraphics[width=1.10cm,height=1.8cm,keepaspectratio]{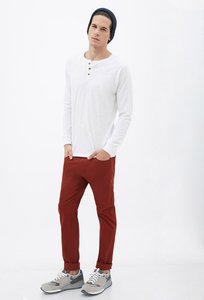}\hss\phantom{\rule{1.10cm}{1.8cm}}\hss}
      \end{minipage}
    }
  \end{minipage}
\hfill
  \begin{minipage}[t]{5.55cm}
    \fcolorbox{red!70!black}{white}{
      \begin{minipage}[t]{5.25cm}
        {\tiny\textbf{\#8}}\\
        \noindent\hbox to 5.25cm{\includegraphics[width=1.10cm,height=1.8cm,keepaspectratio]{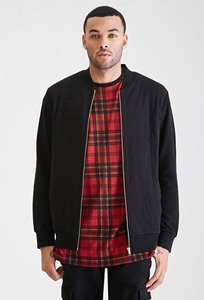}\hss\includegraphics[width=1.10cm,height=1.8cm,keepaspectratio]{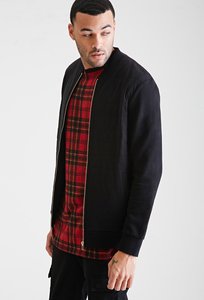}\hss\includegraphics[width=1.10cm,height=1.8cm,keepaspectratio]{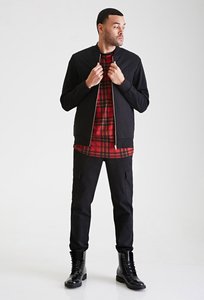}\hss\includegraphics[width=1.10cm,height=1.8cm,keepaspectratio]{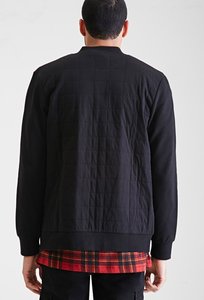}\hss\phantom{\rule{1.10cm}{1.8cm}}\hss}
      \end{minipage}
    }
  \end{minipage}
\hfill
  \begin{minipage}[t]{5.55cm}
    \fcolorbox{red!70!black}{white}{
      \begin{minipage}[t]{5.25cm}
        {\tiny\textbf{\#9}}\\
        \noindent\hbox to 5.25cm{\includegraphics[width=1.10cm,height=1.8cm,keepaspectratio]{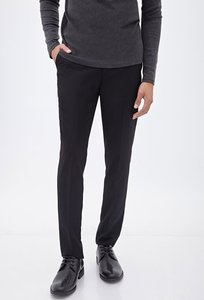}\hss\includegraphics[width=1.10cm,height=1.8cm,keepaspectratio]{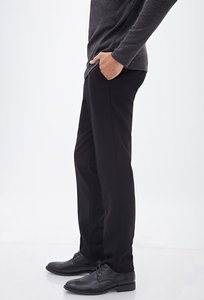}\hss\includegraphics[width=1.10cm,height=1.8cm,keepaspectratio]{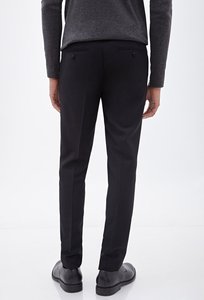}\hss\includegraphics[width=1.10cm,height=1.8cm,keepaspectratio]{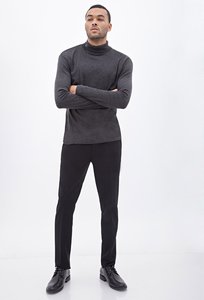}\hss\phantom{\rule{1.10cm}{1.8cm}}\hss}
      \end{minipage}
    }
  \end{minipage}
\par\vspace{2pt}
\noindent
  \begin{minipage}[t]{5.55cm}
    \fcolorbox{red!70!black}{white}{
      \begin{minipage}[t]{5.25cm}
        {\tiny\textbf{\#10}}\\
        \noindent\hbox to 5.25cm{\includegraphics[width=1.10cm,height=1.8cm,keepaspectratio]{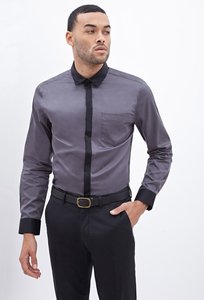}\hss\includegraphics[width=1.10cm,height=1.8cm,keepaspectratio]{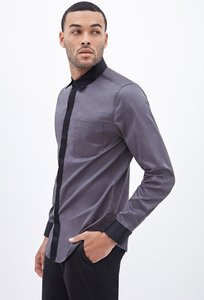}\hss\includegraphics[width=1.10cm,height=1.8cm,keepaspectratio]{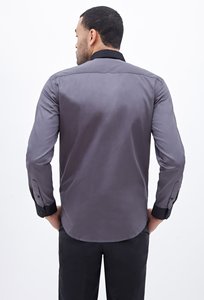}\hss\includegraphics[width=1.10cm,height=1.8cm,keepaspectratio]{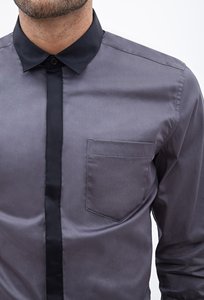}\hss\phantom{\rule{1.10cm}{1.8cm}}\hss}
      \end{minipage}
    }
  \end{minipage}
\hfill
  \begin{minipage}[t]{5.55cm}\end{minipage}
\hfill
  \begin{minipage}[t]{5.55cm}\end{minipage}
\par\vspace{2pt}
\noindent\rule{\linewidth}{0.4pt}
\par\vspace{2pt}
% -- doubao --
{\small\textbf{Doubao-E-V}}\quad{\small Rank~4}\\
\noindent
  \begin{minipage}[t]{5.55cm}
    \fcolorbox{red!70!black}{white}{
      \begin{minipage}[t]{5.25cm}
        {\tiny\textbf{\#1}}\\
        \noindent\hbox to 5.25cm{\includegraphics[width=1.10cm,height=1.8cm,keepaspectratio]{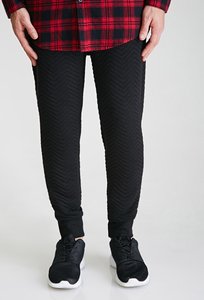}\hss\includegraphics[width=1.10cm,height=1.8cm,keepaspectratio]{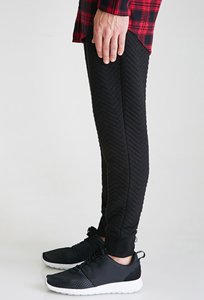}\hss\includegraphics[width=1.10cm,height=1.8cm,keepaspectratio]{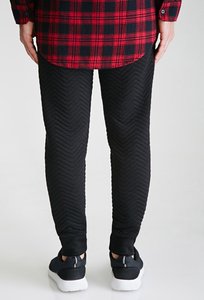}\hss\includegraphics[width=1.10cm,height=1.8cm,keepaspectratio]{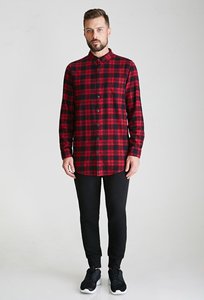}\hss\phantom{\rule{1.10cm}{1.8cm}}\hss}
      \end{minipage}
    }
  \end{minipage}
\hfill
  \begin{minipage}[t]{5.55cm}
    \fcolorbox{red!70!black}{white}{
      \begin{minipage}[t]{5.25cm}
        {\tiny\textbf{\#2}}\\
        \noindent\hbox to 5.25cm{\includegraphics[width=1.10cm,height=1.8cm,keepaspectratio]{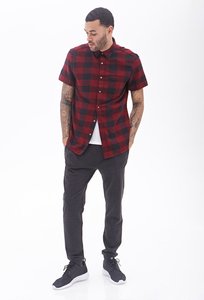}\hss\includegraphics[width=1.10cm,height=1.8cm,keepaspectratio]{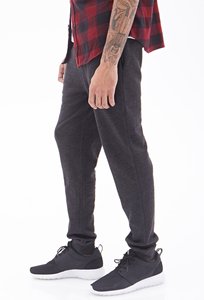}\hss\includegraphics[width=1.10cm,height=1.8cm,keepaspectratio]{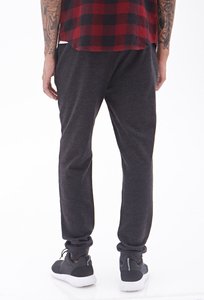}\hss\includegraphics[width=1.10cm,height=1.8cm,keepaspectratio]{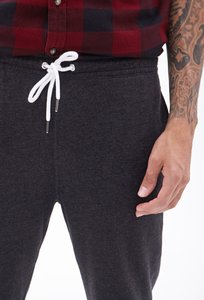}\hss\phantom{\rule{1.10cm}{1.8cm}}\hss}
      \end{minipage}
    }
  \end{minipage}
\hfill
  \begin{minipage}[t]{5.55cm}
    \fcolorbox{red!70!black}{white}{
      \begin{minipage}[t]{5.25cm}
        {\tiny\textbf{\#3}}\\
        \noindent\hbox to 5.25cm{\includegraphics[width=1.10cm,height=1.8cm,keepaspectratio]{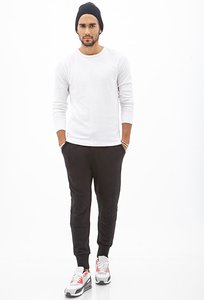}\hss\includegraphics[width=1.10cm,height=1.8cm,keepaspectratio]{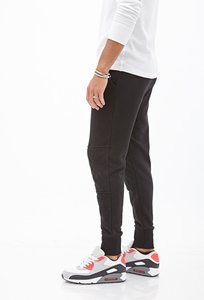}\hss\includegraphics[width=1.10cm,height=1.8cm,keepaspectratio]{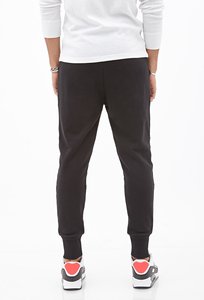}\hss\includegraphics[width=1.10cm,height=1.8cm,keepaspectratio]{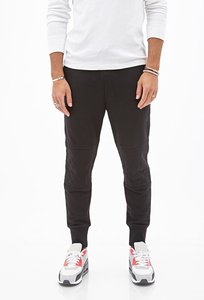}\hss\phantom{\rule{1.10cm}{1.8cm}}\hss}
      \end{minipage}
    }
  \end{minipage}
\par\vspace{2pt}
\noindent
  \begin{minipage}[t]{5.55cm}
    \fcolorbox{green!60!black}{white}{
      \begin{minipage}[t]{5.25cm}
        {\tiny\textbf{\#4}}\\
        \noindent\hbox to 5.25cm{\includegraphics[width=1.10cm,height=1.8cm,keepaspectratio]{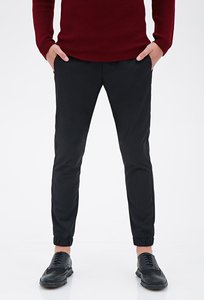}\hss\includegraphics[width=1.10cm,height=1.8cm,keepaspectratio]{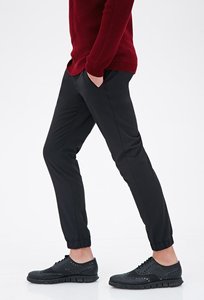}\hss\includegraphics[width=1.10cm,height=1.8cm,keepaspectratio]{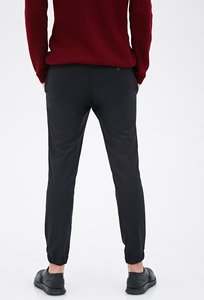}\hss\includegraphics[width=1.10cm,height=1.8cm,keepaspectratio]{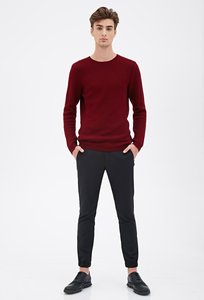}\hss\phantom{\rule{1.10cm}{1.8cm}}\hss}
      \end{minipage}
    }
  \end{minipage}
\hfill
  \begin{minipage}[t]{5.55cm}
    \fcolorbox{red!70!black}{white}{
      \begin{minipage}[t]{5.25cm}
        {\tiny\textbf{\#5}}\\
        \noindent\hbox to 5.25cm{\includegraphics[width=1.10cm,height=1.8cm,keepaspectratio]{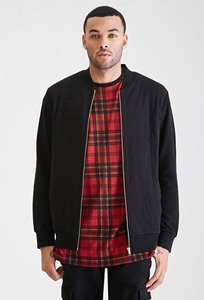}\hss\includegraphics[width=1.10cm,height=1.8cm,keepaspectratio]{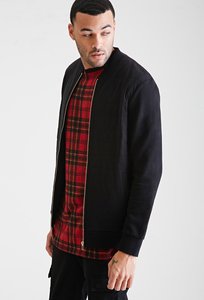}\hss\includegraphics[width=1.10cm,height=1.8cm,keepaspectratio]{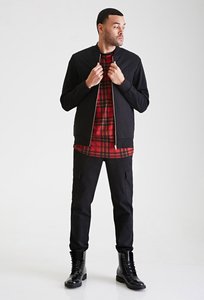}\hss\includegraphics[width=1.10cm,height=1.8cm,keepaspectratio]{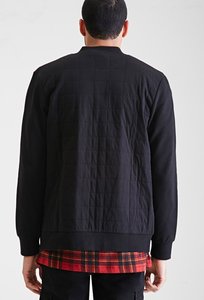}\hss\phantom{\rule{1.10cm}{1.8cm}}\hss}
      \end{minipage}
    }
  \end{minipage}
\hfill
  \begin{minipage}[t]{5.55cm}
    \fcolorbox{red!70!black}{white}{
      \begin{minipage}[t]{5.25cm}
        {\tiny\textbf{\#6}}\\
        \noindent\hbox to 5.25cm{\includegraphics[width=1.10cm,height=1.8cm,keepaspectratio]{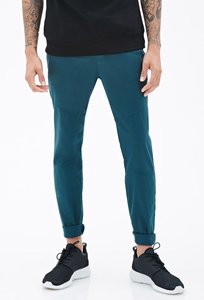}\hss\includegraphics[width=1.10cm,height=1.8cm,keepaspectratio]{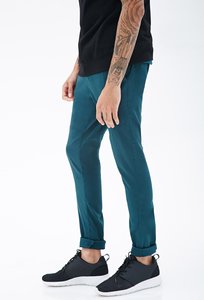}\hss\includegraphics[width=1.10cm,height=1.8cm,keepaspectratio]{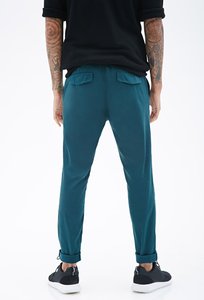}\hss\includegraphics[width=1.10cm,height=1.8cm,keepaspectratio]{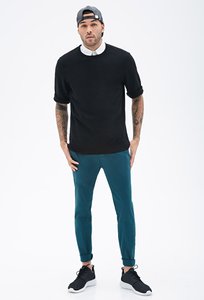}\hss\phantom{\rule{1.10cm}{1.8cm}}\hss}
      \end{minipage}
    }
  \end{minipage}
\par\vspace{2pt}
\noindent
  \begin{minipage}[t]{5.55cm}
    \fcolorbox{red!70!black}{white}{
      \begin{minipage}[t]{5.25cm}
        {\tiny\textbf{\#7}}\\
        \noindent\hbox to 5.25cm{\includegraphics[width=1.10cm,height=1.8cm,keepaspectratio]{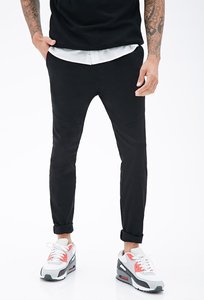}\hss\includegraphics[width=1.10cm,height=1.8cm,keepaspectratio]{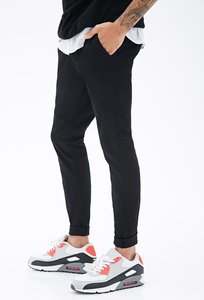}\hss\includegraphics[width=1.10cm,height=1.8cm,keepaspectratio]{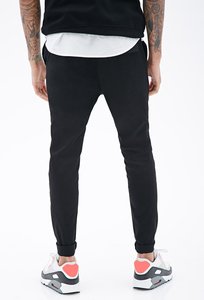}\hss\includegraphics[width=1.10cm,height=1.8cm,keepaspectratio]{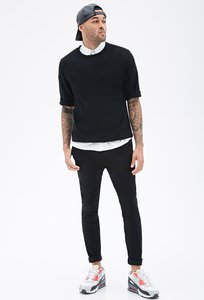}\hss\phantom{\rule{1.10cm}{1.8cm}}\hss}
      \end{minipage}
    }
  \end{minipage}
\hfill
  \begin{minipage}[t]{5.55cm}
    \fcolorbox{red!70!black}{white}{
      \begin{minipage}[t]{5.25cm}
        {\tiny\textbf{\#8}}\\
        \noindent\hbox to 5.25cm{\includegraphics[width=1.10cm,height=1.8cm,keepaspectratio]{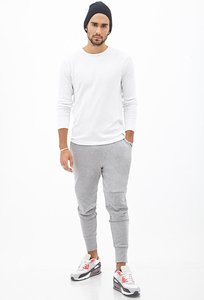}\hss\includegraphics[width=1.10cm,height=1.8cm,keepaspectratio]{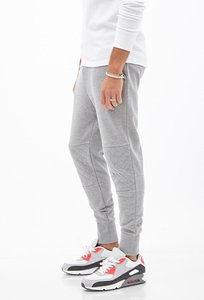}\hss\includegraphics[width=1.10cm,height=1.8cm,keepaspectratio]{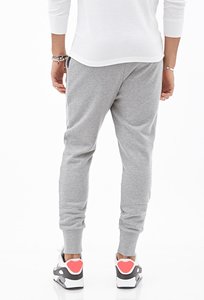}\hss\includegraphics[width=1.10cm,height=1.8cm,keepaspectratio]{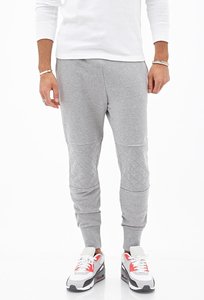}\hss\phantom{\rule{1.10cm}{1.8cm}}\hss}
      \end{minipage}
    }
  \end{minipage}
\hfill
  \begin{minipage}[t]{5.55cm}
    \fcolorbox{red!70!black}{white}{
      \begin{minipage}[t]{5.25cm}
        {\tiny\textbf{\#9}}\\
        \noindent\hbox to 5.25cm{\includegraphics[width=1.10cm,height=1.8cm,keepaspectratio]{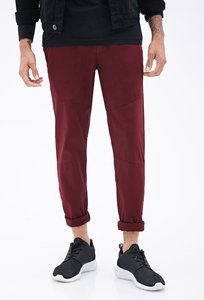}\hss\includegraphics[width=1.10cm,height=1.8cm,keepaspectratio]{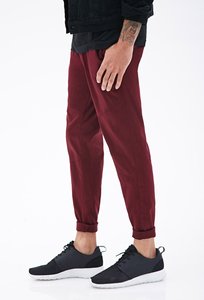}\hss\includegraphics[width=1.10cm,height=1.8cm,keepaspectratio]{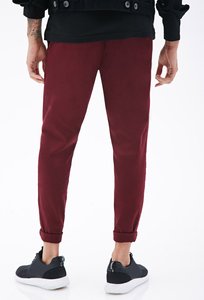}\hss\includegraphics[width=1.10cm,height=1.8cm,keepaspectratio]{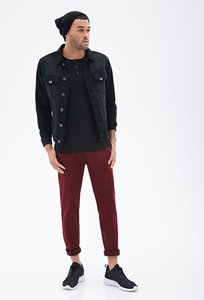}\hss\phantom{\rule{1.10cm}{1.8cm}}\hss}
      \end{minipage}
    }
  \end{minipage}
\par\vspace{2pt}
\noindent
  \begin{minipage}[t]{5.55cm}
    \fcolorbox{red!70!black}{white}{
      \begin{minipage}[t]{5.25cm}
        {\tiny\textbf{\#10}}\\
        \noindent\hbox to 5.25cm{\includegraphics[width=1.10cm,height=1.8cm,keepaspectratio]{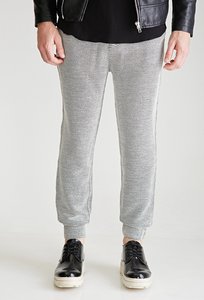}\hss\includegraphics[width=1.10cm,height=1.8cm,keepaspectratio]{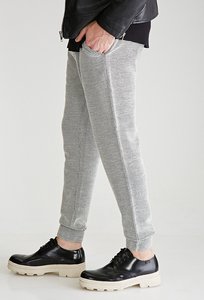}\hss\includegraphics[width=1.10cm,height=1.8cm,keepaspectratio]{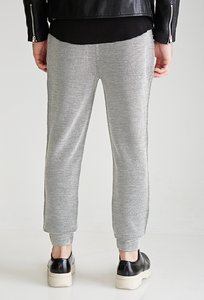}\hss\includegraphics[width=1.10cm,height=1.8cm,keepaspectratio]{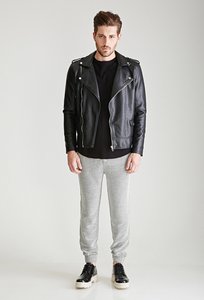}\hss\phantom{\rule{1.10cm}{1.8cm}}\hss}
      \end{minipage}
    }
  \end{minipage}
\hfill
  \begin{minipage}[t]{5.55cm}\end{minipage}
\hfill
  \begin{minipage}[t]{5.55cm}\end{minipage}
\par\vspace{2pt}
\noindent\rule{\linewidth}{0.4pt}
\par\vspace{20pt}

\par\vspace{16pt}
\noindent\textbf{\large Example 3}
\par\vspace{4pt}
\noindent\rule{\linewidth}{1.2pt}
\par\vspace{4pt}
% ── Case 3: short::deepfashion::1773 ──
\noindent\hfill%
  \begin{minipage}[t]{6.0cm}
    \noindent\hbox to 6.0cm{\includegraphics[width=1.20cm,height=2.0cm,keepaspectratio]{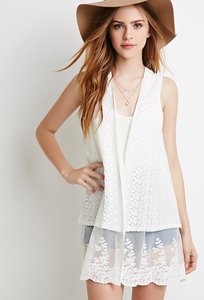}\hss\includegraphics[width=1.20cm,height=2.0cm,keepaspectratio]{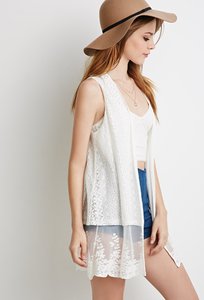}\hss\includegraphics[width=1.20cm,height=2.0cm,keepaspectratio]{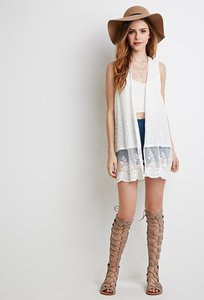}\hss\includegraphics[width=1.20cm,height=2.0cm,keepaspectratio]{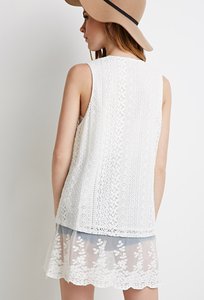}\hss\phantom{\rule{1.20cm}{2.0cm}}\hss}
    \par\vspace{1pt}
    {\scriptsize\textbf{}}
    \par\vspace{0pt}
    \parbox[t]{6.0cm}{\tiny\raggedright White sleeveless open-front vest featuring a crochet lace bodice and sheer floral-embroidered hem. Relaxed mid-thigh length with scalloped edges and wide armholes. Bohemian layering piece with mixed textures, perfect for warm-weather styling.}
  \end{minipage}%
\hfill%
  \begin{minipage}[t]{3.5cm}
    \centering
    \vspace{0.45cm}%
    \parbox{3.5cm}{\centering\tiny Add a center-front tie with gold metal tips and a 2-3 inch fringe trim to the entire hemline, replacing the source's tiered embroidery with a uniform lattice-crochet pattern.}\\[2pt]
    {\normalsize$\longrightarrow$}
  \end{minipage}%
\hfill%
  \begin{minipage}[t]{6.0cm}
    \noindent\hbox to 6.0cm{\includegraphics[width=1.20cm,height=2.0cm,keepaspectratio]{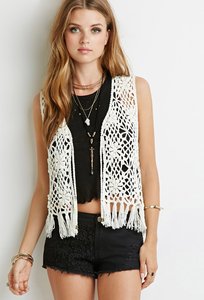}\hss\includegraphics[width=1.20cm,height=2.0cm,keepaspectratio]{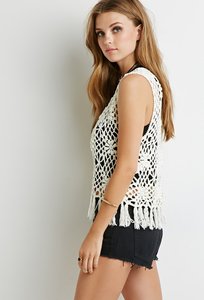}\hss\includegraphics[width=1.20cm,height=2.0cm,keepaspectratio]{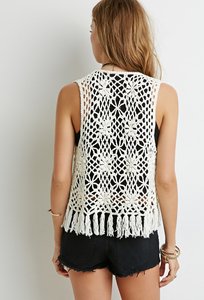}\hss\includegraphics[width=1.20cm,height=2.0cm,keepaspectratio]{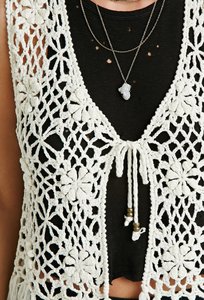}\hss\phantom{\rule{1.20cm}{2.0cm}}\hss}
    \par\vspace{1pt}
    {\scriptsize\textbf{Ground Truth}}
    \par\vspace{0pt}
    \parbox[t]{6.0cm}{\tiny\raggedright White sleeveless crochet vest featuring floral lattice pattern, deep V-neck, open front with tie closure and gold metal tips, and fringe tassel hem. Bohemian layering piece with relaxed fit, perfect for festival or beach wear.}
  \end{minipage}%
\hfill
\par\vspace{4pt}
\par\vspace{4pt}
\noindent\rule{\linewidth}{0.4pt}
% -- mt_align --
{\small\textbf{\textbf{Ours}}}\quad{\small Rank~1}\\
\noindent
  \begin{minipage}[t]{5.55cm}
    \fcolorbox{green!60!black}{white}{
      \begin{minipage}[t]{5.25cm}
        {\tiny\textbf{\#1}}\\
        \noindent\hbox to 5.25cm{\includegraphics[width=1.10cm,height=1.8cm,keepaspectratio]{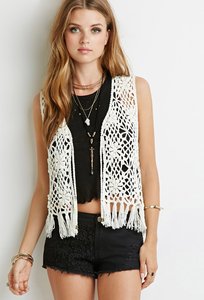}\hss\includegraphics[width=1.10cm,height=1.8cm,keepaspectratio]{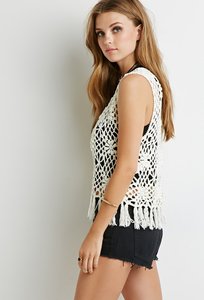}\hss\includegraphics[width=1.10cm,height=1.8cm,keepaspectratio]{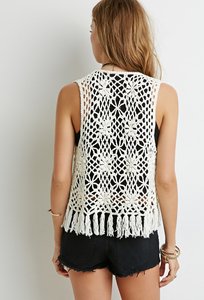}\hss\includegraphics[width=1.10cm,height=1.8cm,keepaspectratio]{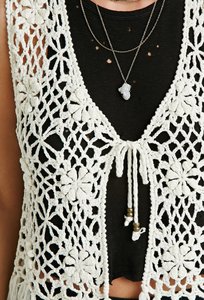}\hss\phantom{\rule{1.10cm}{1.8cm}}\hss}
      \end{minipage}
    }
  \end{minipage}
\hfill
  \begin{minipage}[t]{5.55cm}
    \fcolorbox{red!70!black}{white}{
      \begin{minipage}[t]{5.25cm}
        {\tiny\textbf{\#2}}\\
        \noindent\hbox to 5.25cm{\includegraphics[width=1.10cm,height=1.8cm,keepaspectratio]{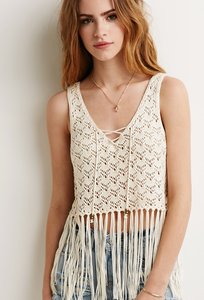}\hss\includegraphics[width=1.10cm,height=1.8cm,keepaspectratio]{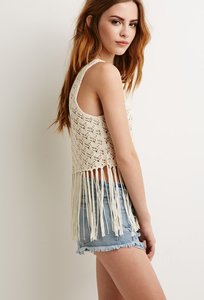}\hss\includegraphics[width=1.10cm,height=1.8cm,keepaspectratio]{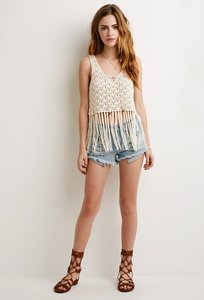}\hss\includegraphics[width=1.10cm,height=1.8cm,keepaspectratio]{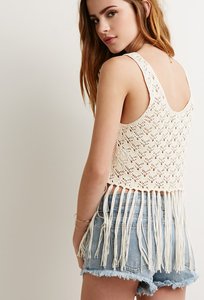}\hss\phantom{\rule{1.10cm}{1.8cm}}\hss}
      \end{minipage}
    }
  \end{minipage}
\hfill
  \begin{minipage}[t]{5.55cm}
    \fcolorbox{red!70!black}{white}{
      \begin{minipage}[t]{5.25cm}
        {\tiny\textbf{\#3}}\\
        \noindent\hbox to 5.25cm{\includegraphics[width=1.10cm,height=1.8cm,keepaspectratio]{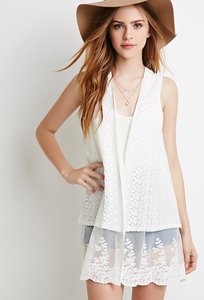}\hss\includegraphics[width=1.10cm,height=1.8cm,keepaspectratio]{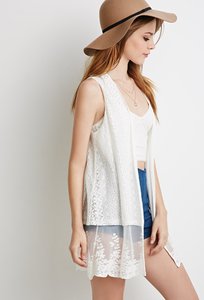}\hss\includegraphics[width=1.10cm,height=1.8cm,keepaspectratio]{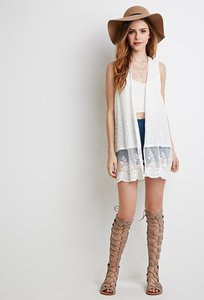}\hss\includegraphics[width=1.10cm,height=1.8cm,keepaspectratio]{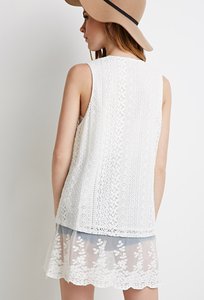}\hss\phantom{\rule{1.10cm}{1.8cm}}\hss}
      \end{minipage}
    }
  \end{minipage}
\par\vspace{2pt}
\noindent
  \begin{minipage}[t]{5.55cm}
    \fcolorbox{red!70!black}{white}{
      \begin{minipage}[t]{5.25cm}
        {\tiny\textbf{\#4}}\\
        \noindent\hbox to 5.25cm{\includegraphics[width=1.10cm,height=1.8cm,keepaspectratio]{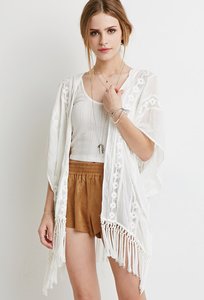}\hss\includegraphics[width=1.10cm,height=1.8cm,keepaspectratio]{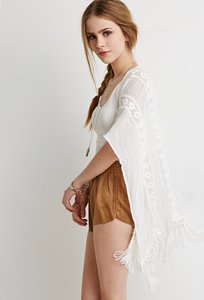}\hss\includegraphics[width=1.10cm,height=1.8cm,keepaspectratio]{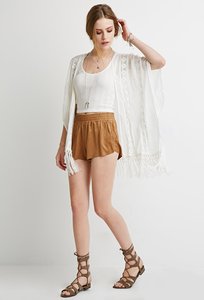}\hss\includegraphics[width=1.10cm,height=1.8cm,keepaspectratio]{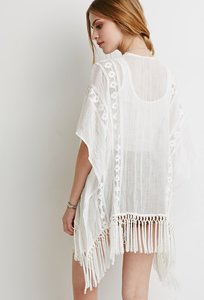}\hss\phantom{\rule{1.10cm}{1.8cm}}\hss}
      \end{minipage}
    }
  \end{minipage}
\hfill
  \begin{minipage}[t]{5.55cm}
    \fcolorbox{red!70!black}{white}{
      \begin{minipage}[t]{5.25cm}
        {\tiny\textbf{\#5}}\\
        \noindent\hbox to 5.25cm{\includegraphics[width=1.10cm,height=1.8cm,keepaspectratio]{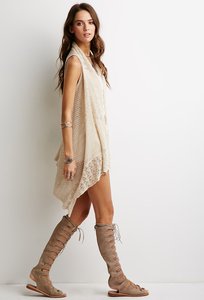}\hss\includegraphics[width=1.10cm,height=1.8cm,keepaspectratio]{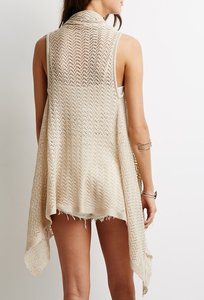}\hss\includegraphics[width=1.10cm,height=1.8cm,keepaspectratio]{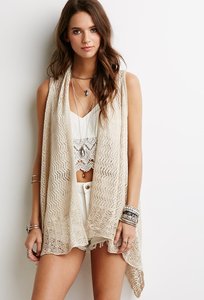}\hss\phantom{\rule{1.10cm}{1.8cm}}\hss\phantom{\rule{1.10cm}{1.8cm}}\hss}
      \end{minipage}
    }
  \end{minipage}
\hfill
  \begin{minipage}[t]{5.55cm}
    \fcolorbox{red!70!black}{white}{
      \begin{minipage}[t]{5.25cm}
        {\tiny\textbf{\#6}}\\
        \noindent\hbox to 5.25cm{\includegraphics[width=1.10cm,height=1.8cm,keepaspectratio]{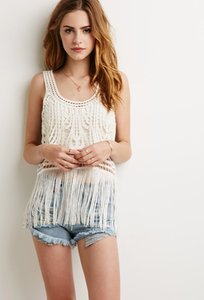}\hss\includegraphics[width=1.10cm,height=1.8cm,keepaspectratio]{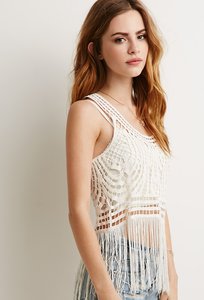}\hss\includegraphics[width=1.10cm,height=1.8cm,keepaspectratio]{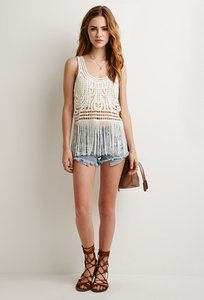}\hss\includegraphics[width=1.10cm,height=1.8cm,keepaspectratio]{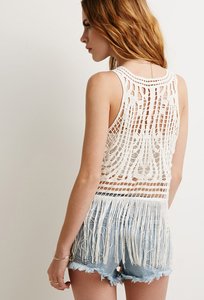}\hss\phantom{\rule{1.10cm}{1.8cm}}\hss}
      \end{minipage}
    }
  \end{minipage}
\par\vspace{2pt}
\noindent
  \begin{minipage}[t]{5.55cm}
    \fcolorbox{red!70!black}{white}{
      \begin{minipage}[t]{5.25cm}
        {\tiny\textbf{\#7}}\\
        \noindent\hbox to 5.25cm{\includegraphics[width=1.10cm,height=1.8cm,keepaspectratio]{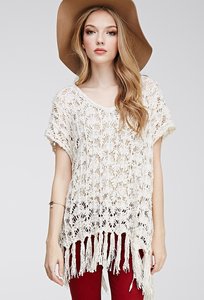}\hss\includegraphics[width=1.10cm,height=1.8cm,keepaspectratio]{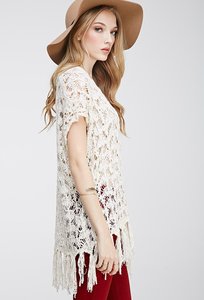}\hss\includegraphics[width=1.10cm,height=1.8cm,keepaspectratio]{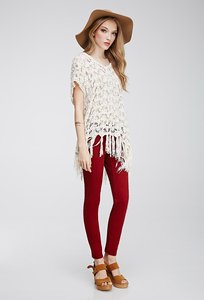}\hss\includegraphics[width=1.10cm,height=1.8cm,keepaspectratio]{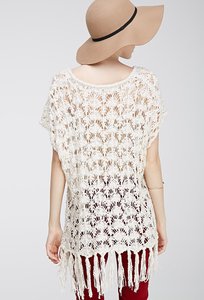}\hss\phantom{\rule{1.10cm}{1.8cm}}\hss}
      \end{minipage}
    }
  \end{minipage}
\hfill
  \begin{minipage}[t]{5.55cm}
    \fcolorbox{red!70!black}{white}{
      \begin{minipage}[t]{5.25cm}
        {\tiny\textbf{\#8}}\\
        \noindent\hbox to 5.25cm{\includegraphics[width=1.10cm,height=1.8cm,keepaspectratio]{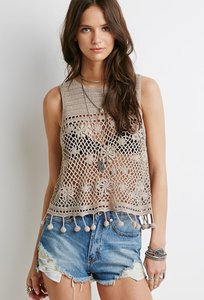}\hss\includegraphics[width=1.10cm,height=1.8cm,keepaspectratio]{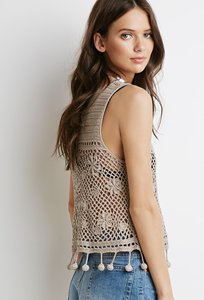}\hss\includegraphics[width=1.10cm,height=1.8cm,keepaspectratio]{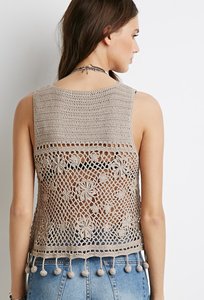}\hss\includegraphics[width=1.10cm,height=1.8cm,keepaspectratio]{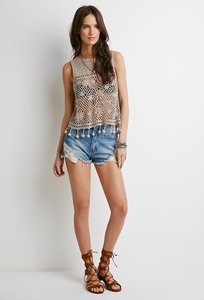}\hss\includegraphics[width=1.10cm,height=1.8cm,keepaspectratio]{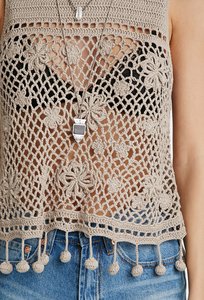}\hss}
      \end{minipage}
    }
  \end{minipage}
\hfill
  \begin{minipage}[t]{5.55cm}
    \fcolorbox{red!70!black}{white}{
      \begin{minipage}[t]{5.25cm}
        {\tiny\textbf{\#9}}\\
        \noindent\hbox to 5.25cm{\includegraphics[width=1.10cm,height=1.8cm,keepaspectratio]{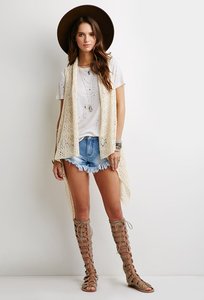}\hss\includegraphics[width=1.10cm,height=1.8cm,keepaspectratio]{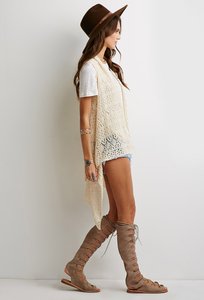}\hss\includegraphics[width=1.10cm,height=1.8cm,keepaspectratio]{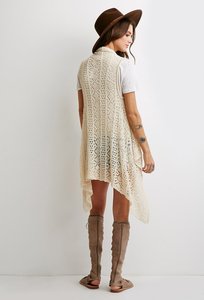}\hss\includegraphics[width=1.10cm,height=1.8cm,keepaspectratio]{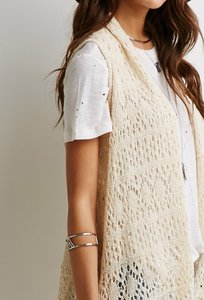}\hss\phantom{\rule{1.10cm}{1.8cm}}\hss}
      \end{minipage}
    }
  \end{minipage}
\par\vspace{2pt}
\noindent
  \begin{minipage}[t]{5.55cm}
    \fcolorbox{red!70!black}{white}{
      \begin{minipage}[t]{5.25cm}
        {\tiny\textbf{\#10}}\\
        \noindent\hbox to 5.25cm{\includegraphics[width=1.10cm,height=1.8cm,keepaspectratio]{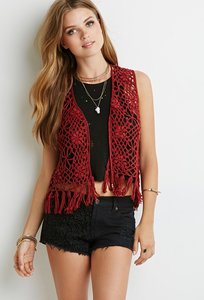}\hss\includegraphics[width=1.10cm,height=1.8cm,keepaspectratio]{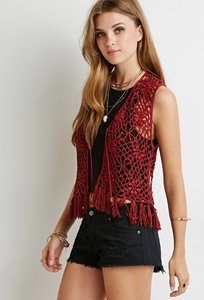}\hss\includegraphics[width=1.10cm,height=1.8cm,keepaspectratio]{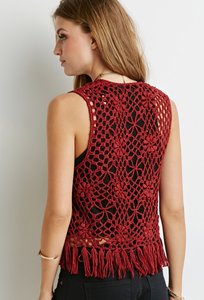}\hss\includegraphics[width=1.10cm,height=1.8cm,keepaspectratio]{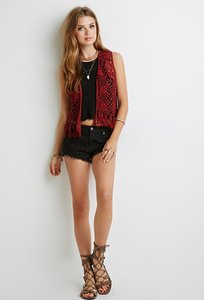}\hss\includegraphics[width=1.10cm,height=1.8cm,keepaspectratio]{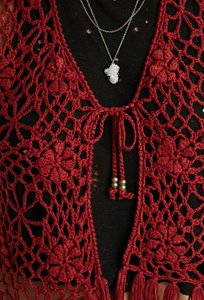}\hss}
      \end{minipage}
    }
  \end{minipage}
\hfill
  \begin{minipage}[t]{5.55cm}\end{minipage}
\hfill
  \begin{minipage}[t]{5.55cm}\end{minipage}
\par\vspace{2pt}
\noindent\rule{\linewidth}{0.4pt}
\par\vspace{2pt}
% -- qwen3_vl_2b --
{\small\textbf{Qwen3-VL-2B}}\quad{\small Rank~4}\\
\noindent
  \begin{minipage}[t]{5.55cm}
    \fcolorbox{red!70!black}{white}{
      \begin{minipage}[t]{5.25cm}
        {\tiny\textbf{\#1}}\\
        \noindent\hbox to 5.25cm{\includegraphics[width=1.10cm,height=1.8cm,keepaspectratio]{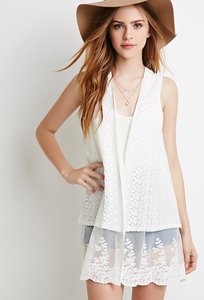}\hss\includegraphics[width=1.10cm,height=1.8cm,keepaspectratio]{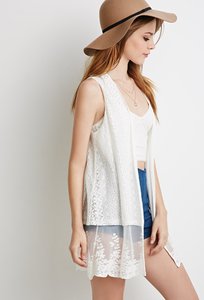}\hss\includegraphics[width=1.10cm,height=1.8cm,keepaspectratio]{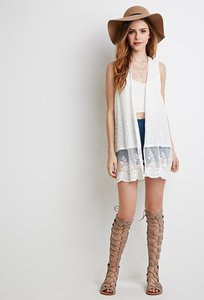}\hss\includegraphics[width=1.10cm,height=1.8cm,keepaspectratio]{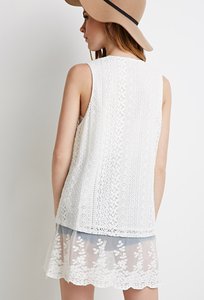}\hss\phantom{\rule{1.10cm}{1.8cm}}\hss}
      \end{minipage}
    }
  \end{minipage}
\hfill
  \begin{minipage}[t]{5.55cm}
    \fcolorbox{red!70!black}{white}{
      \begin{minipage}[t]{5.25cm}
        {\tiny\textbf{\#2}}\\
        \noindent\hbox to 5.25cm{\includegraphics[width=1.10cm,height=1.8cm,keepaspectratio]{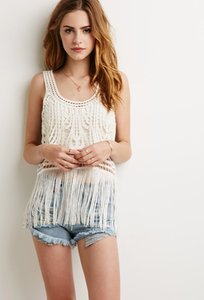}\hss\includegraphics[width=1.10cm,height=1.8cm,keepaspectratio]{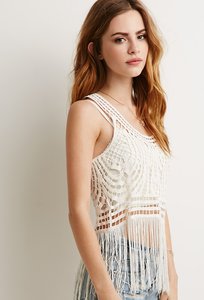}\hss\includegraphics[width=1.10cm,height=1.8cm,keepaspectratio]{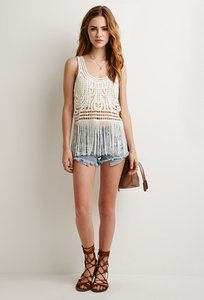}\hss\includegraphics[width=1.10cm,height=1.8cm,keepaspectratio]{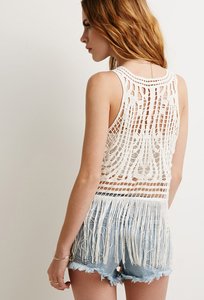}\hss\phantom{\rule{1.10cm}{1.8cm}}\hss}
      \end{minipage}
    }
  \end{minipage}
\hfill
  \begin{minipage}[t]{5.55cm}
    \fcolorbox{red!70!black}{white}{
      \begin{minipage}[t]{5.25cm}
        {\tiny\textbf{\#3}}\\
        \noindent\hbox to 5.25cm{\includegraphics[width=1.10cm,height=1.8cm,keepaspectratio]{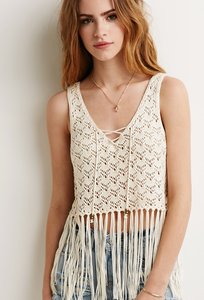}\hss\includegraphics[width=1.10cm,height=1.8cm,keepaspectratio]{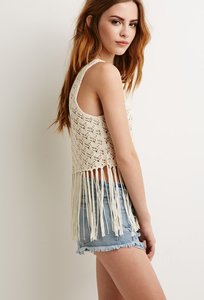}\hss\includegraphics[width=1.10cm,height=1.8cm,keepaspectratio]{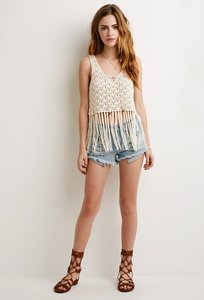}\hss\includegraphics[width=1.10cm,height=1.8cm,keepaspectratio]{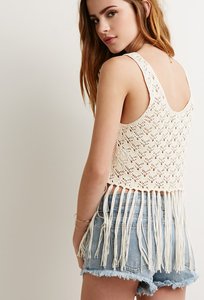}\hss\phantom{\rule{1.10cm}{1.8cm}}\hss}
      \end{minipage}
    }
  \end{minipage}
\par\vspace{2pt}
\noindent
  \begin{minipage}[t]{5.55cm}
    \fcolorbox{green!60!black}{white}{
      \begin{minipage}[t]{5.25cm}
        {\tiny\textbf{\#4}}\\
        \noindent\hbox to 5.25cm{\includegraphics[width=1.10cm,height=1.8cm,keepaspectratio]{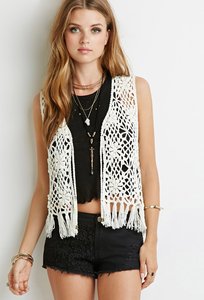}\hss\includegraphics[width=1.10cm,height=1.8cm,keepaspectratio]{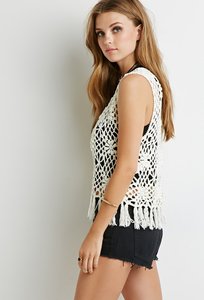}\hss\includegraphics[width=1.10cm,height=1.8cm,keepaspectratio]{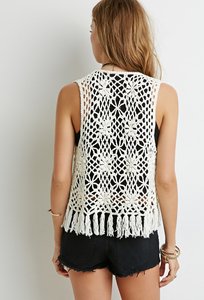}\hss\includegraphics[width=1.10cm,height=1.8cm,keepaspectratio]{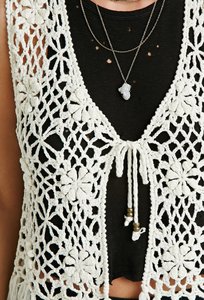}\hss\phantom{\rule{1.10cm}{1.8cm}}\hss}
      \end{minipage}
    }
  \end{minipage}
\hfill
  \begin{minipage}[t]{5.55cm}
    \fcolorbox{red!70!black}{white}{
      \begin{minipage}[t]{5.25cm}
        {\tiny\textbf{\#5}}\\
        \noindent\hbox to 5.25cm{\includegraphics[width=1.10cm,height=1.8cm,keepaspectratio]{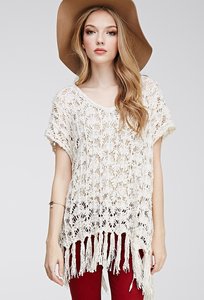}\hss\includegraphics[width=1.10cm,height=1.8cm,keepaspectratio]{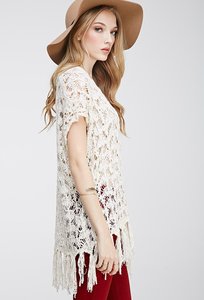}\hss\includegraphics[width=1.10cm,height=1.8cm,keepaspectratio]{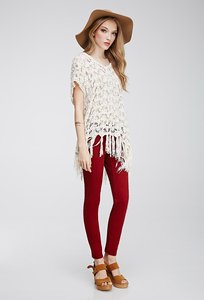}\hss\includegraphics[width=1.10cm,height=1.8cm,keepaspectratio]{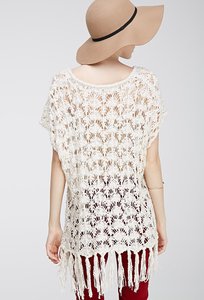}\hss\phantom{\rule{1.10cm}{1.8cm}}\hss}
      \end{minipage}
    }
  \end{minipage}
\hfill
  \begin{minipage}[t]{5.55cm}
    \fcolorbox{red!70!black}{white}{
      \begin{minipage}[t]{5.25cm}
        {\tiny\textbf{\#6}}\\
        \noindent\hbox to 5.25cm{\includegraphics[width=1.10cm,height=1.8cm,keepaspectratio]{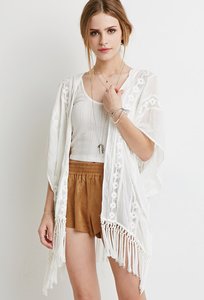}\hss\includegraphics[width=1.10cm,height=1.8cm,keepaspectratio]{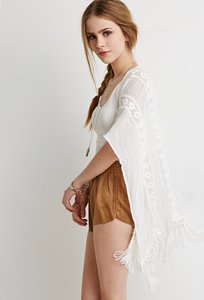}\hss\includegraphics[width=1.10cm,height=1.8cm,keepaspectratio]{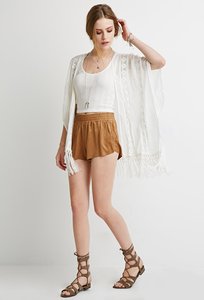}\hss\includegraphics[width=1.10cm,height=1.8cm,keepaspectratio]{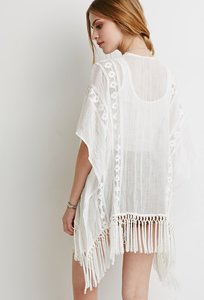}\hss\phantom{\rule{1.10cm}{1.8cm}}\hss}
      \end{minipage}
    }
  \end{minipage}
\par\vspace{2pt}
\noindent
  \begin{minipage}[t]{5.55cm}
    \fcolorbox{red!70!black}{white}{
      \begin{minipage}[t]{5.25cm}
        {\tiny\textbf{\#7}}\\
        \noindent\hbox to 5.25cm{\includegraphics[width=1.10cm,height=1.8cm,keepaspectratio]{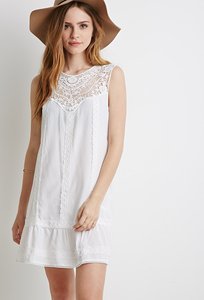}\hss\includegraphics[width=1.10cm,height=1.8cm,keepaspectratio]{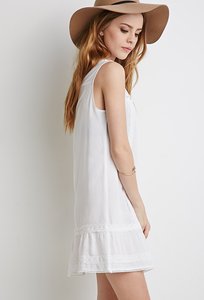}\hss\includegraphics[width=1.10cm,height=1.8cm,keepaspectratio]{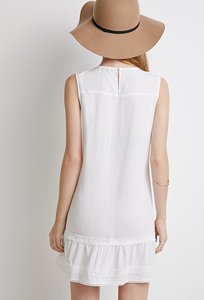}\hss\includegraphics[width=1.10cm,height=1.8cm,keepaspectratio]{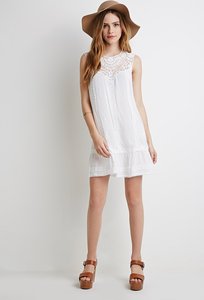}\hss\includegraphics[width=1.10cm,height=1.8cm,keepaspectratio]{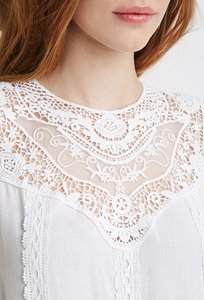}\hss}
      \end{minipage}
    }
  \end{minipage}
\hfill
  \begin{minipage}[t]{5.55cm}
    \fcolorbox{red!70!black}{white}{
      \begin{minipage}[t]{5.25cm}
        {\tiny\textbf{\#8}}\\
        \noindent\hbox to 5.25cm{\includegraphics[width=1.10cm,height=1.8cm,keepaspectratio]{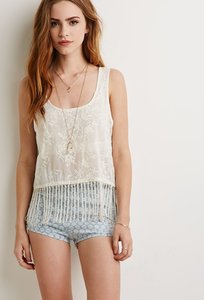}\hss\includegraphics[width=1.10cm,height=1.8cm,keepaspectratio]{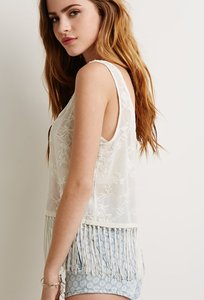}\hss\includegraphics[width=1.10cm,height=1.8cm,keepaspectratio]{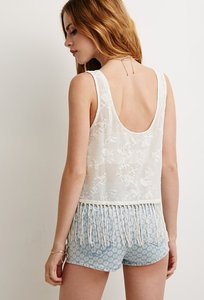}\hss\phantom{\rule{1.10cm}{1.8cm}}\hss\phantom{\rule{1.10cm}{1.8cm}}\hss}
      \end{minipage}
    }
  \end{minipage}
\hfill
  \begin{minipage}[t]{5.55cm}
    \fcolorbox{red!70!black}{white}{
      \begin{minipage}[t]{5.25cm}
        {\tiny\textbf{\#9}}\\
        \noindent\hbox to 5.25cm{\includegraphics[width=1.10cm,height=1.8cm,keepaspectratio]{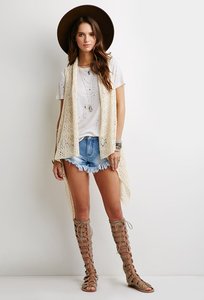}\hss\includegraphics[width=1.10cm,height=1.8cm,keepaspectratio]{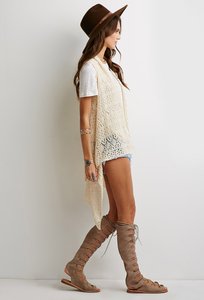}\hss\includegraphics[width=1.10cm,height=1.8cm,keepaspectratio]{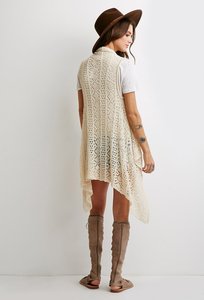}\hss\includegraphics[width=1.10cm,height=1.8cm,keepaspectratio]{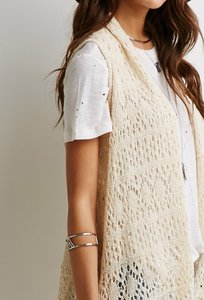}\hss\phantom{\rule{1.10cm}{1.8cm}}\hss}
      \end{minipage}
    }
  \end{minipage}
\par\vspace{2pt}
\noindent
  \begin{minipage}[t]{5.55cm}
    \fcolorbox{red!70!black}{white}{
      \begin{minipage}[t]{5.25cm}
        {\tiny\textbf{\#10}}\\
        \noindent\hbox to 5.25cm{\includegraphics[width=1.10cm,height=1.8cm,keepaspectratio]{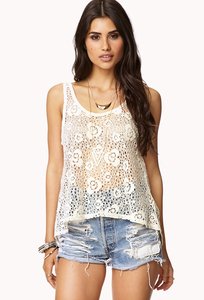}\hss\includegraphics[width=1.10cm,height=1.8cm,keepaspectratio]{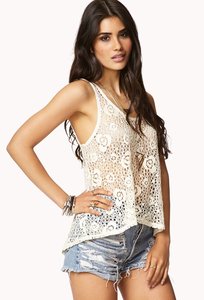}\hss\includegraphics[width=1.10cm,height=1.8cm,keepaspectratio]{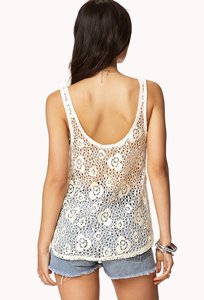}\hss\includegraphics[width=1.10cm,height=1.8cm,keepaspectratio]{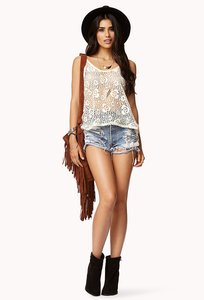}\hss\phantom{\rule{1.10cm}{1.8cm}}\hss}
      \end{minipage}
    }
  \end{minipage}
\hfill
  \begin{minipage}[t]{5.55cm}\end{minipage}
\hfill
  \begin{minipage}[t]{5.55cm}\end{minipage}
\par\vspace{2pt}
\noindent\rule{\linewidth}{0.4pt}
\par\vspace{2pt}
% -- qwen3_vl_8b --
{\small\textbf{Qwen3-VL-8B}}\quad{\small Rank~2}\\
\noindent
  \begin{minipage}[t]{5.55cm}
    \fcolorbox{red!70!black}{white}{
      \begin{minipage}[t]{5.25cm}
        {\tiny\textbf{\#1}}\\
        \noindent\hbox to 5.25cm{\includegraphics[width=1.10cm,height=1.8cm,keepaspectratio]{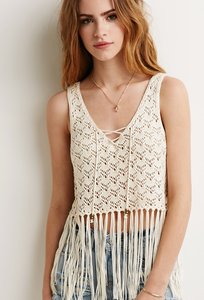}\hss\includegraphics[width=1.10cm,height=1.8cm,keepaspectratio]{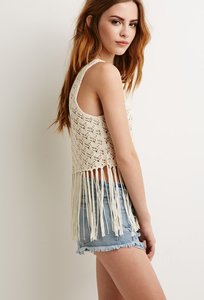}\hss\includegraphics[width=1.10cm,height=1.8cm,keepaspectratio]{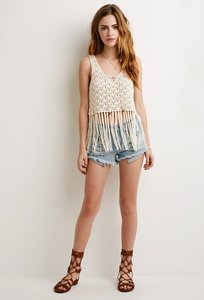}\hss\includegraphics[width=1.10cm,height=1.8cm,keepaspectratio]{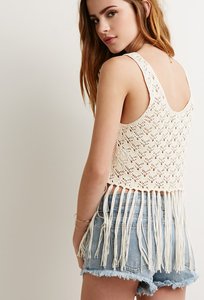}\hss\phantom{\rule{1.10cm}{1.8cm}}\hss}
      \end{minipage}
    }
  \end{minipage}
\hfill
  \begin{minipage}[t]{5.55cm}
    \fcolorbox{green!60!black}{white}{
      \begin{minipage}[t]{5.25cm}
        {\tiny\textbf{\#2}}\\
        \noindent\hbox to 5.25cm{\includegraphics[width=1.10cm,height=1.8cm,keepaspectratio]{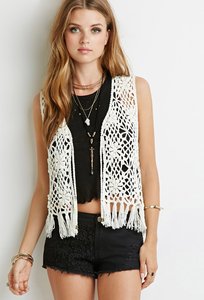}\hss\includegraphics[width=1.10cm,height=1.8cm,keepaspectratio]{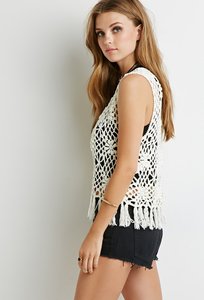}\hss\includegraphics[width=1.10cm,height=1.8cm,keepaspectratio]{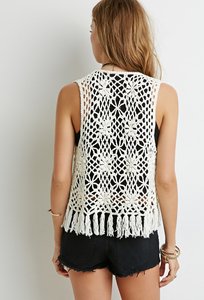}\hss\includegraphics[width=1.10cm,height=1.8cm,keepaspectratio]{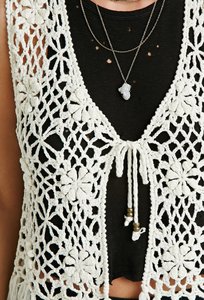}\hss\phantom{\rule{1.10cm}{1.8cm}}\hss}
      \end{minipage}
    }
  \end{minipage}
\hfill
  \begin{minipage}[t]{5.55cm}
    \fcolorbox{red!70!black}{white}{
      \begin{minipage}[t]{5.25cm}
        {\tiny\textbf{\#3}}\\
        \noindent\hbox to 5.25cm{\includegraphics[width=1.10cm,height=1.8cm,keepaspectratio]{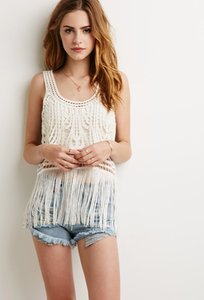}\hss\includegraphics[width=1.10cm,height=1.8cm,keepaspectratio]{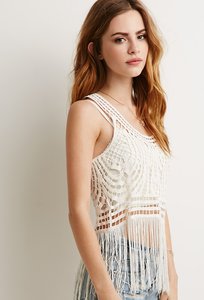}\hss\includegraphics[width=1.10cm,height=1.8cm,keepaspectratio]{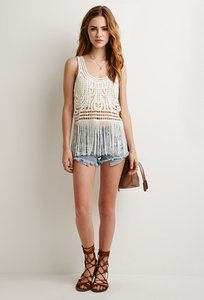}\hss\includegraphics[width=1.10cm,height=1.8cm,keepaspectratio]{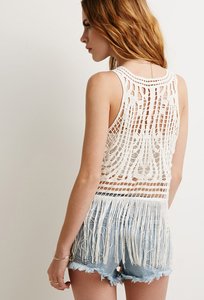}\hss\phantom{\rule{1.10cm}{1.8cm}}\hss}
      \end{minipage}
    }
  \end{minipage}
\par\vspace{2pt}
\noindent
  \begin{minipage}[t]{5.55cm}
    \fcolorbox{red!70!black}{white}{
      \begin{minipage}[t]{5.25cm}
        {\tiny\textbf{\#4}}\\
        \noindent\hbox to 5.25cm{\includegraphics[width=1.10cm,height=1.8cm,keepaspectratio]{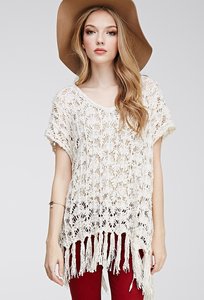}\hss\includegraphics[width=1.10cm,height=1.8cm,keepaspectratio]{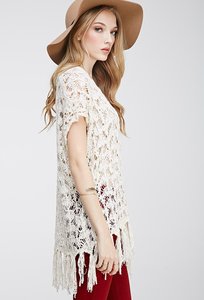}\hss\includegraphics[width=1.10cm,height=1.8cm,keepaspectratio]{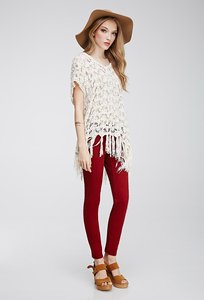}\hss\includegraphics[width=1.10cm,height=1.8cm,keepaspectratio]{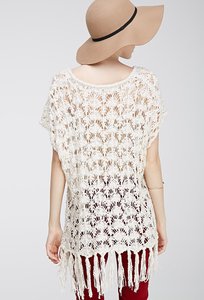}\hss\phantom{\rule{1.10cm}{1.8cm}}\hss}
      \end{minipage}
    }
  \end{minipage}
\hfill
  \begin{minipage}[t]{5.55cm}
    \fcolorbox{red!70!black}{white}{
      \begin{minipage}[t]{5.25cm}
        {\tiny\textbf{\#5}}\\
        \noindent\hbox to 5.25cm{\includegraphics[width=1.10cm,height=1.8cm,keepaspectratio]{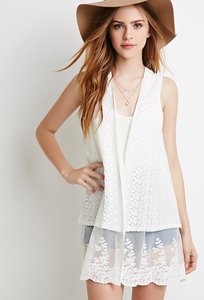}\hss\includegraphics[width=1.10cm,height=1.8cm,keepaspectratio]{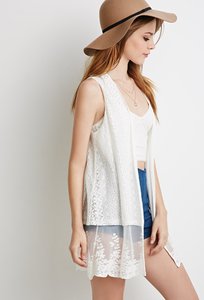}\hss\includegraphics[width=1.10cm,height=1.8cm,keepaspectratio]{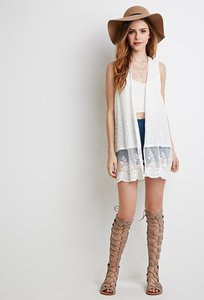}\hss\includegraphics[width=1.10cm,height=1.8cm,keepaspectratio]{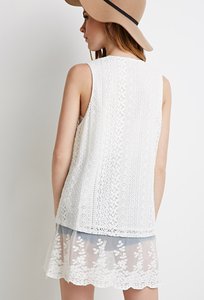}\hss\phantom{\rule{1.10cm}{1.8cm}}\hss}
      \end{minipage}
    }
  \end{minipage}
\hfill
  \begin{minipage}[t]{5.55cm}
    \fcolorbox{red!70!black}{white}{
      \begin{minipage}[t]{5.25cm}
        {\tiny\textbf{\#6}}\\
        \noindent\hbox to 5.25cm{\includegraphics[width=1.10cm,height=1.8cm,keepaspectratio]{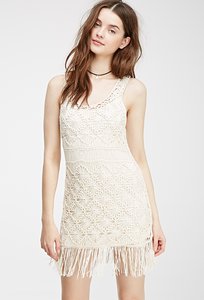}\hss\includegraphics[width=1.10cm,height=1.8cm,keepaspectratio]{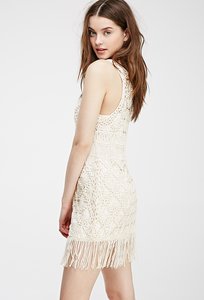}\hss\includegraphics[width=1.10cm,height=1.8cm,keepaspectratio]{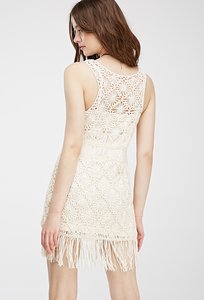}\hss\phantom{\rule{1.10cm}{1.8cm}}\hss\phantom{\rule{1.10cm}{1.8cm}}\hss}
      \end{minipage}
    }
  \end{minipage}
\par\vspace{2pt}
\noindent
  \begin{minipage}[t]{5.55cm}
    \fcolorbox{red!70!black}{white}{
      \begin{minipage}[t]{5.25cm}
        {\tiny\textbf{\#7}}\\
        \noindent\hbox to 5.25cm{\includegraphics[width=1.10cm,height=1.8cm,keepaspectratio]{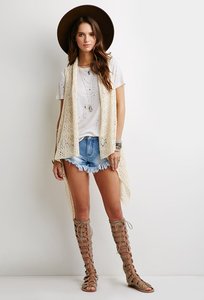}\hss\includegraphics[width=1.10cm,height=1.8cm,keepaspectratio]{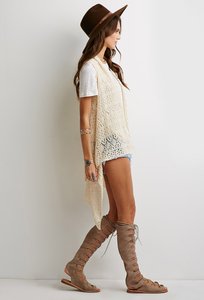}\hss\includegraphics[width=1.10cm,height=1.8cm,keepaspectratio]{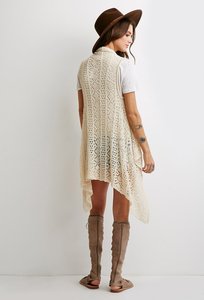}\hss\includegraphics[width=1.10cm,height=1.8cm,keepaspectratio]{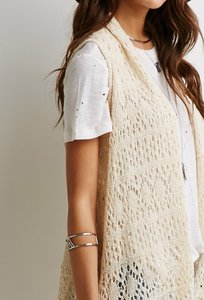}\hss\phantom{\rule{1.10cm}{1.8cm}}\hss}
      \end{minipage}
    }
  \end{minipage}
\hfill
  \begin{minipage}[t]{5.55cm}
    \fcolorbox{red!70!black}{white}{
      \begin{minipage}[t]{5.25cm}
        {\tiny\textbf{\#8}}\\
        \noindent\hbox to 5.25cm{\includegraphics[width=1.10cm,height=1.8cm,keepaspectratio]{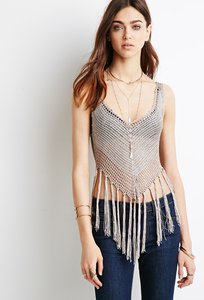}\hss\includegraphics[width=1.10cm,height=1.8cm,keepaspectratio]{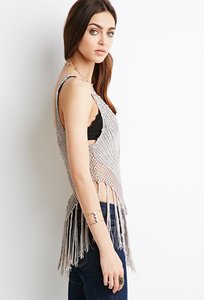}\hss\includegraphics[width=1.10cm,height=1.8cm,keepaspectratio]{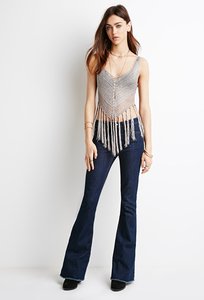}\hss\includegraphics[width=1.10cm,height=1.8cm,keepaspectratio]{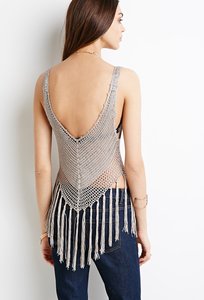}\hss\phantom{\rule{1.10cm}{1.8cm}}\hss}
      \end{minipage}
    }
  \end{minipage}
\hfill
  \begin{minipage}[t]{5.55cm}
    \fcolorbox{red!70!black}{white}{
      \begin{minipage}[t]{5.25cm}
        {\tiny\textbf{\#9}}\\
        \noindent\hbox to 5.25cm{\includegraphics[width=1.10cm,height=1.8cm,keepaspectratio]{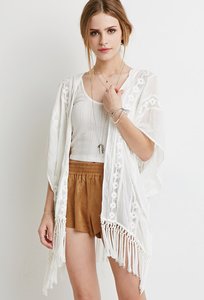}\hss\includegraphics[width=1.10cm,height=1.8cm,keepaspectratio]{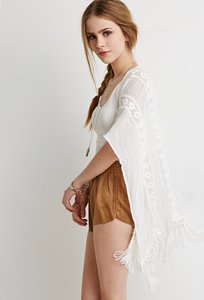}\hss\includegraphics[width=1.10cm,height=1.8cm,keepaspectratio]{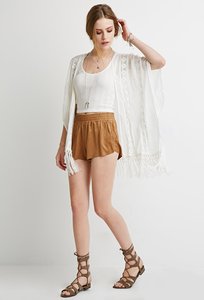}\hss\includegraphics[width=1.10cm,height=1.8cm,keepaspectratio]{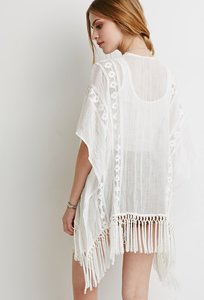}\hss\phantom{\rule{1.10cm}{1.8cm}}\hss}
      \end{minipage}
    }
  \end{minipage}
\par\vspace{2pt}
\noindent
  \begin{minipage}[t]{5.55cm}
    \fcolorbox{red!70!black}{white}{
      \begin{minipage}[t]{5.25cm}
        {\tiny\textbf{\#10}}\\
        \noindent\hbox to 5.25cm{\includegraphics[width=1.10cm,height=1.8cm,keepaspectratio]{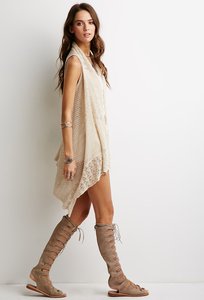}\hss\includegraphics[width=1.10cm,height=1.8cm,keepaspectratio]{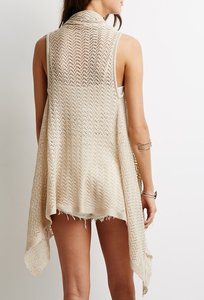}\hss\includegraphics[width=1.10cm,height=1.8cm,keepaspectratio]{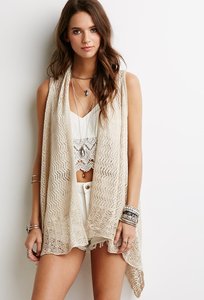}\hss\phantom{\rule{1.10cm}{1.8cm}}\hss\phantom{\rule{1.10cm}{1.8cm}}\hss}
      \end{minipage}
    }
  \end{minipage}
\hfill
  \begin{minipage}[t]{5.55cm}\end{minipage}
\hfill
  \begin{minipage}[t]{5.55cm}\end{minipage}
\par\vspace{2pt}
\noindent\rule{\linewidth}{0.4pt}
\par\vspace{2pt}
% -- reznembed --
{\small\textbf{RezNEmbed}}\quad{\small Rank~4}\\
\noindent
  \begin{minipage}[t]{5.55cm}
    \fcolorbox{red!70!black}{white}{
      \begin{minipage}[t]{5.25cm}
        {\tiny\textbf{\#1}}\\
        \noindent\hbox to 5.25cm{\includegraphics[width=1.10cm,height=1.8cm,keepaspectratio]{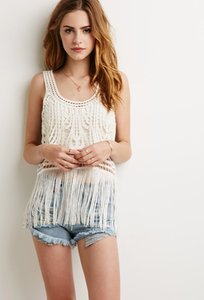}\hss\includegraphics[width=1.10cm,height=1.8cm,keepaspectratio]{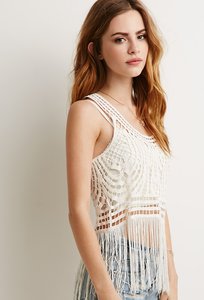}\hss\includegraphics[width=1.10cm,height=1.8cm,keepaspectratio]{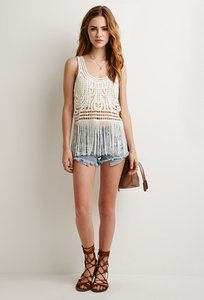}\hss\includegraphics[width=1.10cm,height=1.8cm,keepaspectratio]{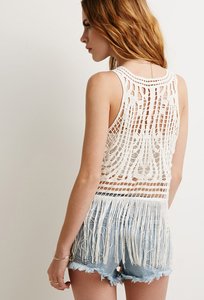}\hss\phantom{\rule{1.10cm}{1.8cm}}\hss}
      \end{minipage}
    }
  \end{minipage}
\hfill
  \begin{minipage}[t]{5.55cm}
    \fcolorbox{red!70!black}{white}{
      \begin{minipage}[t]{5.25cm}
        {\tiny\textbf{\#2}}\\
        \noindent\hbox to 5.25cm{\includegraphics[width=1.10cm,height=1.8cm,keepaspectratio]{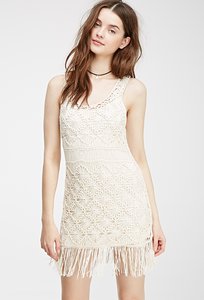}\hss\includegraphics[width=1.10cm,height=1.8cm,keepaspectratio]{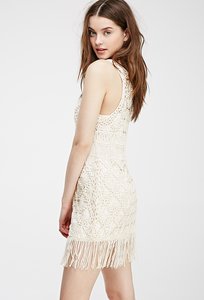}\hss\includegraphics[width=1.10cm,height=1.8cm,keepaspectratio]{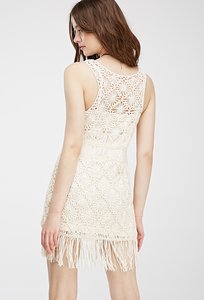}\hss\phantom{\rule{1.10cm}{1.8cm}}\hss\phantom{\rule{1.10cm}{1.8cm}}\hss}
      \end{minipage}
    }
  \end{minipage}
\hfill
  \begin{minipage}[t]{5.55cm}
    \fcolorbox{red!70!black}{white}{
      \begin{minipage}[t]{5.25cm}
        {\tiny\textbf{\#3}}\\
        \noindent\hbox to 5.25cm{\includegraphics[width=1.10cm,height=1.8cm,keepaspectratio]{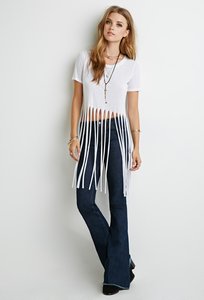}\hss\includegraphics[width=1.10cm,height=1.8cm,keepaspectratio]{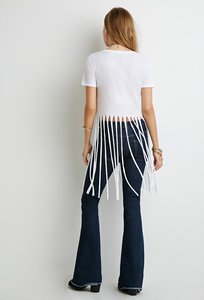}\hss\includegraphics[width=1.10cm,height=1.8cm,keepaspectratio]{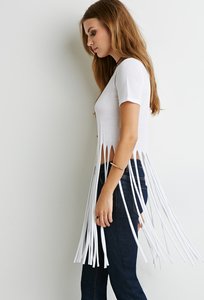}\hss\phantom{\rule{1.10cm}{1.8cm}}\hss\phantom{\rule{1.10cm}{1.8cm}}\hss}
      \end{minipage}
    }
  \end{minipage}
\par\vspace{2pt}
\noindent
  \begin{minipage}[t]{5.55cm}
    \fcolorbox{green!60!black}{white}{
      \begin{minipage}[t]{5.25cm}
        {\tiny\textbf{\#4}}\\
        \noindent\hbox to 5.25cm{\includegraphics[width=1.10cm,height=1.8cm,keepaspectratio]{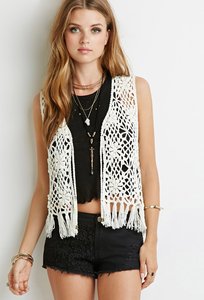}\hss\includegraphics[width=1.10cm,height=1.8cm,keepaspectratio]{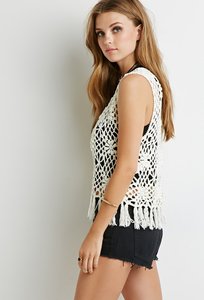}\hss\includegraphics[width=1.10cm,height=1.8cm,keepaspectratio]{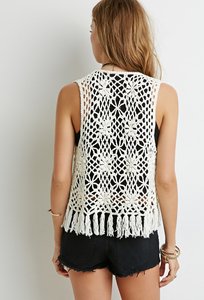}\hss\includegraphics[width=1.10cm,height=1.8cm,keepaspectratio]{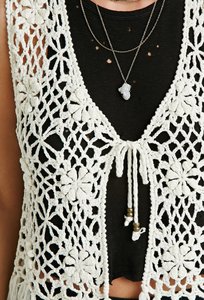}\hss\phantom{\rule{1.10cm}{1.8cm}}\hss}
      \end{minipage}
    }
  \end{minipage}
\hfill
  \begin{minipage}[t]{5.55cm}
    \fcolorbox{red!70!black}{white}{
      \begin{minipage}[t]{5.25cm}
        {\tiny\textbf{\#5}}\\
        \noindent\hbox to 5.25cm{\includegraphics[width=1.10cm,height=1.8cm,keepaspectratio]{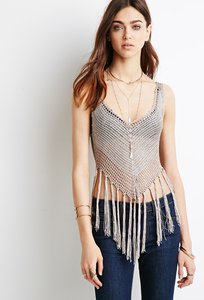}\hss\includegraphics[width=1.10cm,height=1.8cm,keepaspectratio]{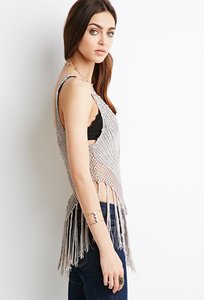}\hss\includegraphics[width=1.10cm,height=1.8cm,keepaspectratio]{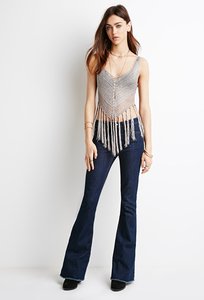}\hss\includegraphics[width=1.10cm,height=1.8cm,keepaspectratio]{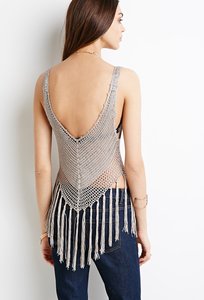}\hss\phantom{\rule{1.10cm}{1.8cm}}\hss}
      \end{minipage}
    }
  \end{minipage}
\hfill
  \begin{minipage}[t]{5.55cm}
    \fcolorbox{red!70!black}{white}{
      \begin{minipage}[t]{5.25cm}
        {\tiny\textbf{\#6}}\\
        \noindent\hbox to 5.25cm{\includegraphics[width=1.10cm,height=1.8cm,keepaspectratio]{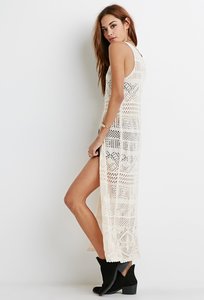}\hss\includegraphics[width=1.10cm,height=1.8cm,keepaspectratio]{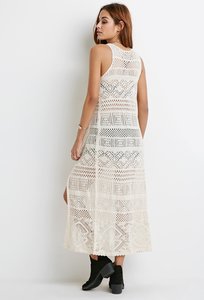}\hss\includegraphics[width=1.10cm,height=1.8cm,keepaspectratio]{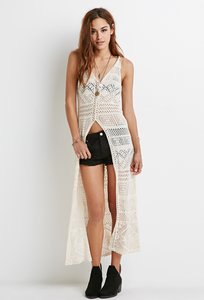}\hss\phantom{\rule{1.10cm}{1.8cm}}\hss\phantom{\rule{1.10cm}{1.8cm}}\hss}
      \end{minipage}
    }
  \end{minipage}
\par\vspace{2pt}
\noindent
  \begin{minipage}[t]{5.55cm}
    \fcolorbox{red!70!black}{white}{
      \begin{minipage}[t]{5.25cm}
        {\tiny\textbf{\#7}}\\
        \noindent\hbox to 5.25cm{\includegraphics[width=1.10cm,height=1.8cm,keepaspectratio]{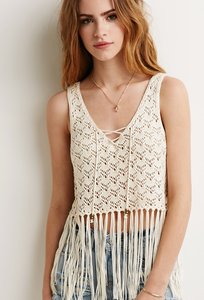}\hss\includegraphics[width=1.10cm,height=1.8cm,keepaspectratio]{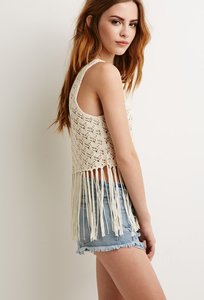}\hss\includegraphics[width=1.10cm,height=1.8cm,keepaspectratio]{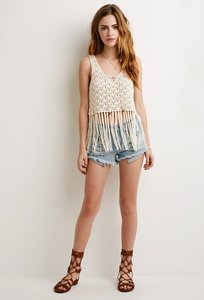}\hss\includegraphics[width=1.10cm,height=1.8cm,keepaspectratio]{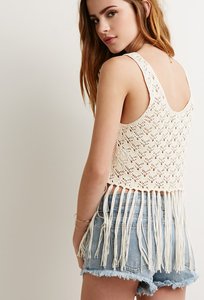}\hss\phantom{\rule{1.10cm}{1.8cm}}\hss}
      \end{minipage}
    }
  \end{minipage}
\hfill
  \begin{minipage}[t]{5.55cm}
    \fcolorbox{red!70!black}{white}{
      \begin{minipage}[t]{5.25cm}
        {\tiny\textbf{\#8}}\\
        \noindent\hbox to 5.25cm{\includegraphics[width=1.10cm,height=1.8cm,keepaspectratio]{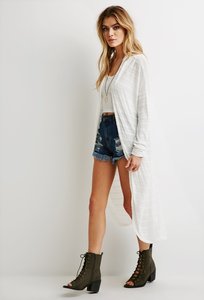}\hss\includegraphics[width=1.10cm,height=1.8cm,keepaspectratio]{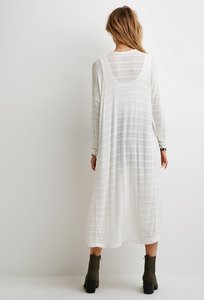}\hss\includegraphics[width=1.10cm,height=1.8cm,keepaspectratio]{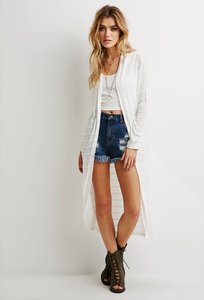}\hss\phantom{\rule{1.10cm}{1.8cm}}\hss\phantom{\rule{1.10cm}{1.8cm}}\hss}
      \end{minipage}
    }
  \end{minipage}
\hfill
  \begin{minipage}[t]{5.55cm}
    \fcolorbox{red!70!black}{white}{
      \begin{minipage}[t]{5.25cm}
        {\tiny\textbf{\#9}}\\
        \noindent\hbox to 5.25cm{\includegraphics[width=1.10cm,height=1.8cm,keepaspectratio]{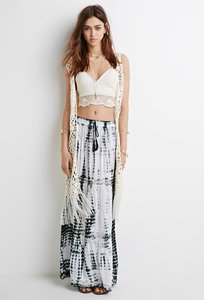}\hss\includegraphics[width=1.10cm,height=1.8cm,keepaspectratio]{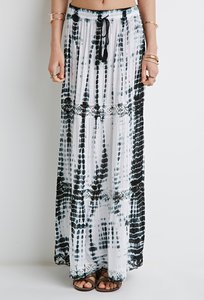}\hss\includegraphics[width=1.10cm,height=1.8cm,keepaspectratio]{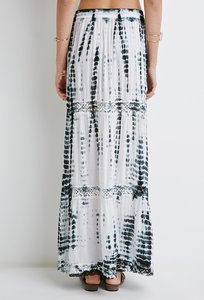}\hss\includegraphics[width=1.10cm,height=1.8cm,keepaspectratio]{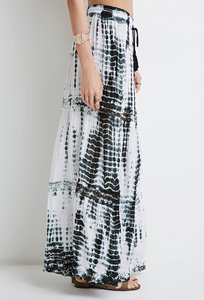}\hss\phantom{\rule{1.10cm}{1.8cm}}\hss}
      \end{minipage}
    }
  \end{minipage}
\par\vspace{2pt}
\noindent
  \begin{minipage}[t]{5.55cm}
    \fcolorbox{red!70!black}{white}{
      \begin{minipage}[t]{5.25cm}
        {\tiny\textbf{\#10}}\\
        \noindent\hbox to 5.25cm{\includegraphics[width=1.10cm,height=1.8cm,keepaspectratio]{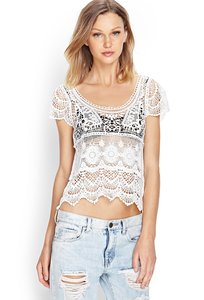}\hss\includegraphics[width=1.10cm,height=1.8cm,keepaspectratio]{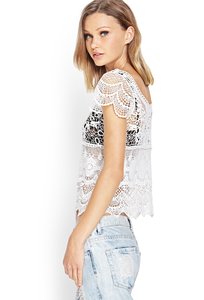}\hss\includegraphics[width=1.10cm,height=1.8cm,keepaspectratio]{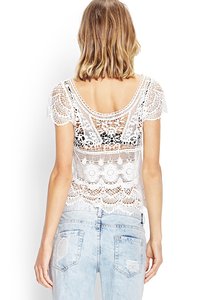}\hss\includegraphics[width=1.10cm,height=1.8cm,keepaspectratio]{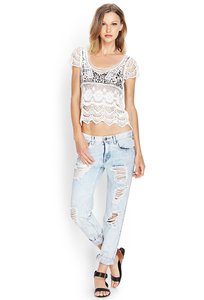}\hss\phantom{\rule{1.10cm}{1.8cm}}\hss}
      \end{minipage}
    }
  \end{minipage}
\hfill
  \begin{minipage}[t]{5.55cm}\end{minipage}
\hfill
  \begin{minipage}[t]{5.55cm}\end{minipage}
\par\vspace{2pt}
\noindent\rule{\linewidth}{0.4pt}
\par\vspace{2pt}
% -- doubao --
{\small\textbf{Doubao-E-V}}\quad{\small Rank~5}\\
\noindent
  \begin{minipage}[t]{5.55cm}
    \fcolorbox{red!70!black}{white}{
      \begin{minipage}[t]{5.25cm}
        {\tiny\textbf{\#1}}\\
        \noindent\hbox to 5.25cm{\includegraphics[width=1.10cm,height=1.8cm,keepaspectratio]{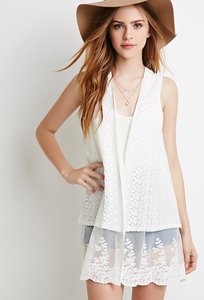}\hss\includegraphics[width=1.10cm,height=1.8cm,keepaspectratio]{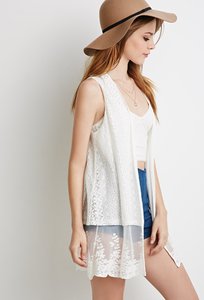}\hss\includegraphics[width=1.10cm,height=1.8cm,keepaspectratio]{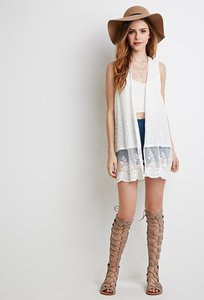}\hss\includegraphics[width=1.10cm,height=1.8cm,keepaspectratio]{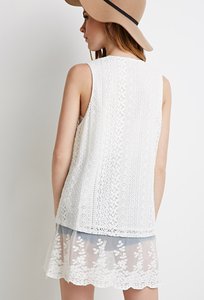}\hss\phantom{\rule{1.10cm}{1.8cm}}\hss}
      \end{minipage}
    }
  \end{minipage}
\hfill
  \begin{minipage}[t]{5.55cm}
    \fcolorbox{red!70!black}{white}{
      \begin{minipage}[t]{5.25cm}
        {\tiny\textbf{\#2}}\\
        \noindent\hbox to 5.25cm{\includegraphics[width=1.10cm,height=1.8cm,keepaspectratio]{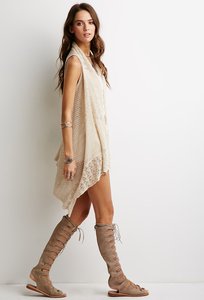}\hss\includegraphics[width=1.10cm,height=1.8cm,keepaspectratio]{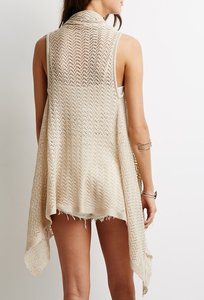}\hss\includegraphics[width=1.10cm,height=1.8cm,keepaspectratio]{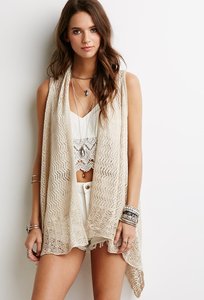}\hss\phantom{\rule{1.10cm}{1.8cm}}\hss\phantom{\rule{1.10cm}{1.8cm}}\hss}
      \end{minipage}
    }
  \end{minipage}
\hfill
  \begin{minipage}[t]{5.55cm}
    \fcolorbox{red!70!black}{white}{
      \begin{minipage}[t]{5.25cm}
        {\tiny\textbf{\#3}}\\
        \noindent\hbox to 5.25cm{\includegraphics[width=1.10cm,height=1.8cm,keepaspectratio]{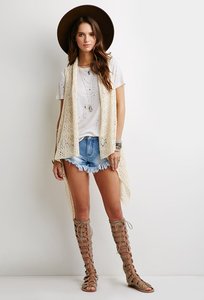}\hss\includegraphics[width=1.10cm,height=1.8cm,keepaspectratio]{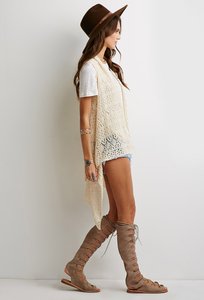}\hss\includegraphics[width=1.10cm,height=1.8cm,keepaspectratio]{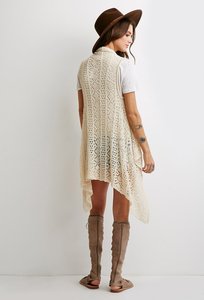}\hss\includegraphics[width=1.10cm,height=1.8cm,keepaspectratio]{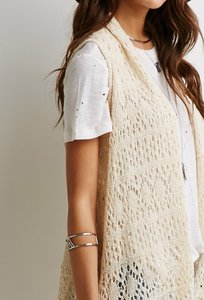}\hss\phantom{\rule{1.10cm}{1.8cm}}\hss}
      \end{minipage}
    }
  \end{minipage}
\par\vspace{2pt}
\noindent
  \begin{minipage}[t]{5.55cm}
    \fcolorbox{red!70!black}{white}{
      \begin{minipage}[t]{5.25cm}
        {\tiny\textbf{\#4}}\\
        \noindent\hbox to 5.25cm{\includegraphics[width=1.10cm,height=1.8cm,keepaspectratio]{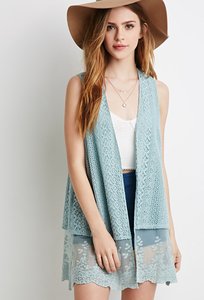}\hss\includegraphics[width=1.10cm,height=1.8cm,keepaspectratio]{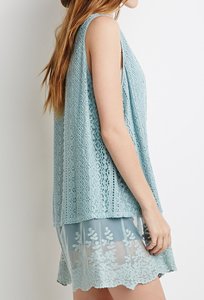}\hss\includegraphics[width=1.10cm,height=1.8cm,keepaspectratio]{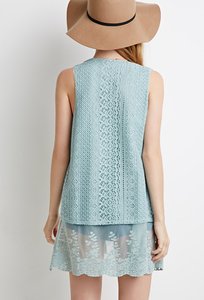}\hss\phantom{\rule{1.10cm}{1.8cm}}\hss\phantom{\rule{1.10cm}{1.8cm}}\hss}
      \end{minipage}
    }
  \end{minipage}
\hfill
  \begin{minipage}[t]{5.55cm}
    \fcolorbox{green!60!black}{white}{
      \begin{minipage}[t]{5.25cm}
        {\tiny\textbf{\#5}}\\
        \noindent\hbox to 5.25cm{\includegraphics[width=1.10cm,height=1.8cm,keepaspectratio]{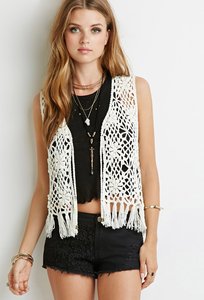}\hss\includegraphics[width=1.10cm,height=1.8cm,keepaspectratio]{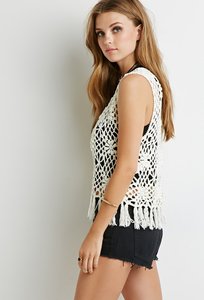}\hss\includegraphics[width=1.10cm,height=1.8cm,keepaspectratio]{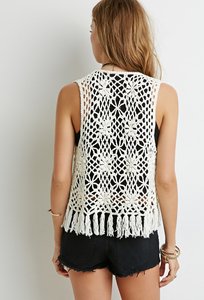}\hss\includegraphics[width=1.10cm,height=1.8cm,keepaspectratio]{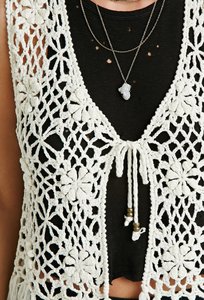}\hss\phantom{\rule{1.10cm}{1.8cm}}\hss}
      \end{minipage}
    }
  \end{minipage}
\hfill
  \begin{minipage}[t]{5.55cm}
    \fcolorbox{red!70!black}{white}{
      \begin{minipage}[t]{5.25cm}
        {\tiny\textbf{\#6}}\\
        \noindent\hbox to 5.25cm{\includegraphics[width=1.10cm,height=1.8cm,keepaspectratio]{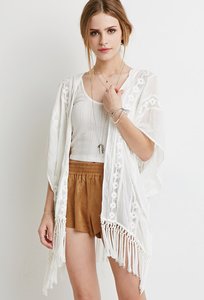}\hss\includegraphics[width=1.10cm,height=1.8cm,keepaspectratio]{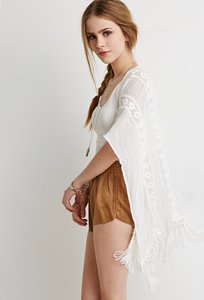}\hss\includegraphics[width=1.10cm,height=1.8cm,keepaspectratio]{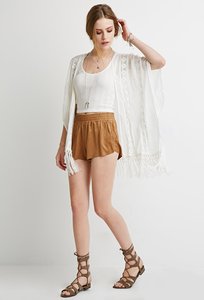}\hss\includegraphics[width=1.10cm,height=1.8cm,keepaspectratio]{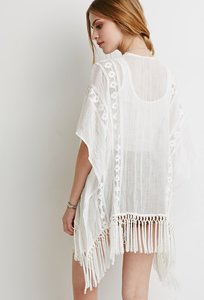}\hss\phantom{\rule{1.10cm}{1.8cm}}\hss}
      \end{minipage}
    }
  \end{minipage}
\par\vspace{2pt}
\noindent
  \begin{minipage}[t]{5.55cm}
    \fcolorbox{red!70!black}{white}{
      \begin{minipage}[t]{5.25cm}
        {\tiny\textbf{\#7}}\\
        \noindent\hbox to 5.25cm{\includegraphics[width=1.10cm,height=1.8cm,keepaspectratio]{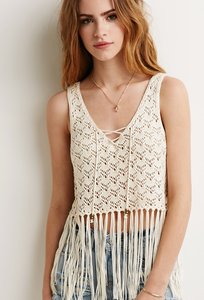}\hss\includegraphics[width=1.10cm,height=1.8cm,keepaspectratio]{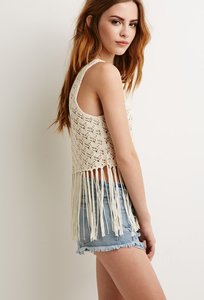}\hss\includegraphics[width=1.10cm,height=1.8cm,keepaspectratio]{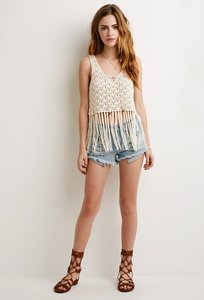}\hss\includegraphics[width=1.10cm,height=1.8cm,keepaspectratio]{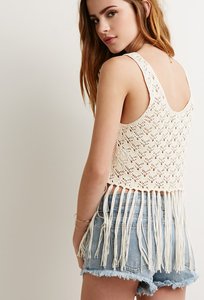}\hss\phantom{\rule{1.10cm}{1.8cm}}\hss}
      \end{minipage}
    }
  \end{minipage}
\hfill
  \begin{minipage}[t]{5.55cm}
    \fcolorbox{red!70!black}{white}{
      \begin{minipage}[t]{5.25cm}
        {\tiny\textbf{\#8}}\\
        \noindent\hbox to 5.25cm{\includegraphics[width=1.10cm,height=1.8cm,keepaspectratio]{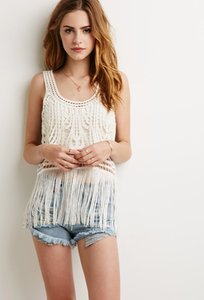}\hss\includegraphics[width=1.10cm,height=1.8cm,keepaspectratio]{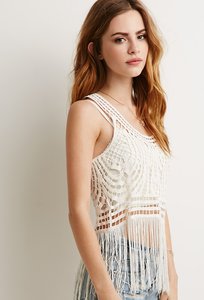}\hss\includegraphics[width=1.10cm,height=1.8cm,keepaspectratio]{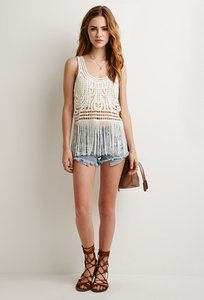}\hss\includegraphics[width=1.10cm,height=1.8cm,keepaspectratio]{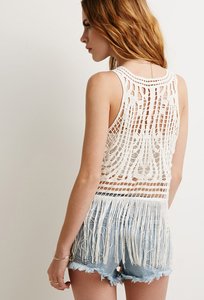}\hss\phantom{\rule{1.10cm}{1.8cm}}\hss}
      \end{minipage}
    }
  \end{minipage}
\hfill
  \begin{minipage}[t]{5.55cm}
    \fcolorbox{red!70!black}{white}{
      \begin{minipage}[t]{5.25cm}
        {\tiny\textbf{\#9}}\\
        \noindent\hbox to 5.25cm{\includegraphics[width=1.10cm,height=1.8cm,keepaspectratio]{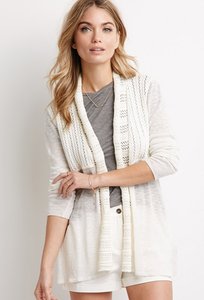}\hss\includegraphics[width=1.10cm,height=1.8cm,keepaspectratio]{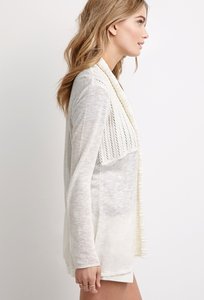}\hss\includegraphics[width=1.10cm,height=1.8cm,keepaspectratio]{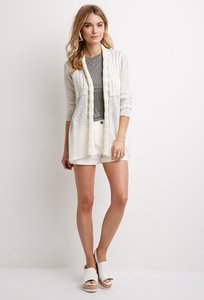}\hss\includegraphics[width=1.10cm,height=1.8cm,keepaspectratio]{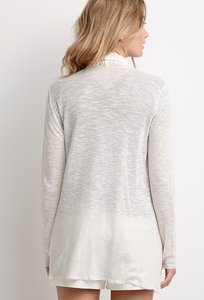}\hss\phantom{\rule{1.10cm}{1.8cm}}\hss}
      \end{minipage}
    }
  \end{minipage}
\par\vspace{2pt}
\noindent
  \begin{minipage}[t]{5.55cm}
    \fcolorbox{red!70!black}{white}{
      \begin{minipage}[t]{5.25cm}
        {\tiny\textbf{\#10}}\\
        \noindent\hbox to 5.25cm{\includegraphics[width=1.10cm,height=1.8cm,keepaspectratio]{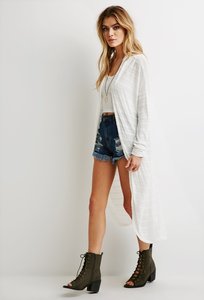}\hss\includegraphics[width=1.10cm,height=1.8cm,keepaspectratio]{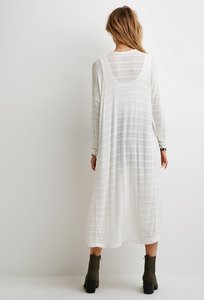}\hss\includegraphics[width=1.10cm,height=1.8cm,keepaspectratio]{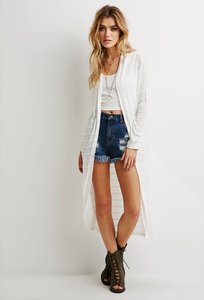}\hss\phantom{\rule{1.10cm}{1.8cm}}\hss\phantom{\rule{1.10cm}{1.8cm}}\hss}
      \end{minipage}
    }
  \end{minipage}
\hfill
  \begin{minipage}[t]{5.55cm}\end{minipage}
\hfill
  \begin{minipage}[t]{5.55cm}\end{minipage}
\par\vspace{2pt}
\noindent\rule{\linewidth}{0.4pt}
\par\vspace{20pt}

\par\vspace{16pt}
\noindent\textbf{\large Example 4}
\par\vspace{4pt}
\noindent\rule{\linewidth}{1.2pt}
\par\vspace{4pt}
% ── Case 4: short::deepfashion::2554 ──
\noindent\hfill%
  \begin{minipage}[t]{6.0cm}
    \noindent\hbox to 6.0cm{\includegraphics[width=1.20cm,height=2.0cm,keepaspectratio]{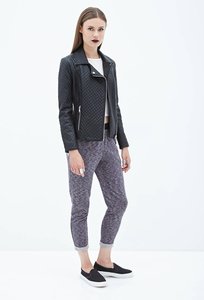}\hss\includegraphics[width=1.20cm,height=2.0cm,keepaspectratio]{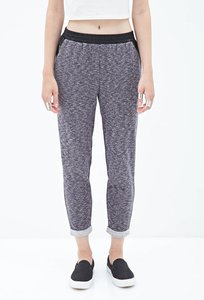}\hss\includegraphics[width=1.20cm,height=2.0cm,keepaspectratio]{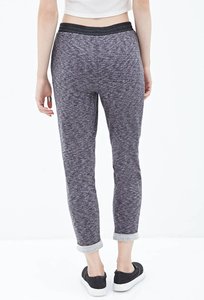}\hss\includegraphics[width=1.20cm,height=2.0cm,keepaspectratio]{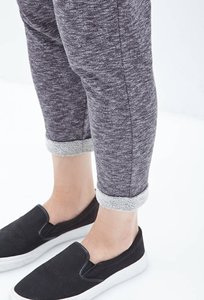}\hss\includegraphics[width=1.20cm,height=2.0cm,keepaspectratio]{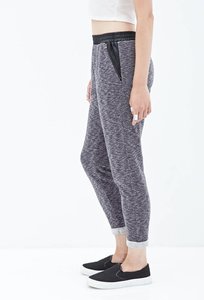}\hss}
    \par\vspace{1pt}
    {\scriptsize\textbf{}}
    \par\vspace{0pt}
    \parbox[t]{6.0cm}{\tiny\raggedright Heather gray marled jogger pants featuring a black elastic waistband, zippered side pockets with black trim, and rolled cuffs revealing a light gray interior. Designed with a relaxed tapered fit, soft heathered knit fabric, and cropped ankle length for versatile casual or athletic wear.}
  \end{minipage}%
\hfill%
  \begin{minipage}[t]{3.5cm}
    \centering
    \vspace{0.45cm}%
    \parbox{3.5cm}{\centering\tiny Replace standard side pockets with black faux leather-trimmed zippered pockets and update the waistband to a ribbed, quilted texture.}\\[2pt]
    {\normalsize$\longrightarrow$}
  \end{minipage}%
\hfill%
  \begin{minipage}[t]{6.0cm}
    \noindent\hbox to 6.0cm{\includegraphics[width=1.20cm,height=2.0cm,keepaspectratio]{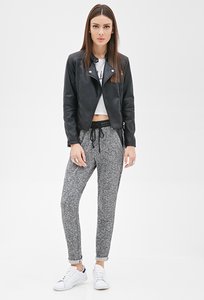}\hss\includegraphics[width=1.20cm,height=2.0cm,keepaspectratio]{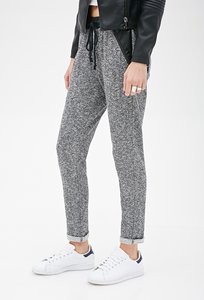}\hss\includegraphics[width=1.20cm,height=2.0cm,keepaspectratio]{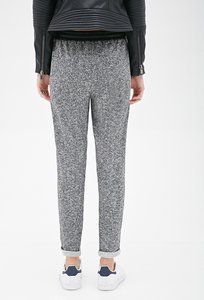}\hss\includegraphics[width=1.20cm,height=2.0cm,keepaspectratio]{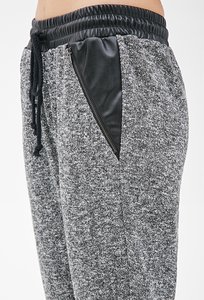}\hss\phantom{\rule{1.20cm}{2.0cm}}\hss}
    \par\vspace{1pt}
    {\scriptsize\textbf{Ground Truth}}
    \par\vspace{0pt}
    \parbox[t]{6.0cm}{\tiny\raggedright Gray marled knit jogger pants featuring black faux leather-trimmed diagonal zip side pockets, quilted elastic waistband with drawstring, and cuffed hems. Relaxed tapered fit with no back pockets. Heathered gray and white textured fabric with edgy moto-inspired details.}
  \end{minipage}%
\hfill
\par\vspace{4pt}
\par\vspace{4pt}
\noindent\rule{\linewidth}{0.4pt}
% -- mt_align --
{\small\textbf{\textbf{Ours}}}\quad{\small Rank~1}\\
\noindent
  \begin{minipage}[t]{5.55cm}
    \fcolorbox{green!60!black}{white}{
      \begin{minipage}[t]{5.25cm}
        {\tiny\textbf{\#1}}\\
        \noindent\hbox to 5.25cm{\includegraphics[width=1.10cm,height=1.8cm,keepaspectratio]{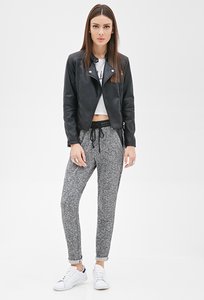}\hss\includegraphics[width=1.10cm,height=1.8cm,keepaspectratio]{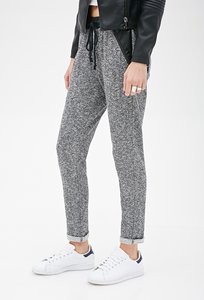}\hss\includegraphics[width=1.10cm,height=1.8cm,keepaspectratio]{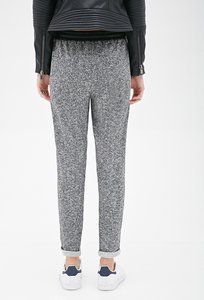}\hss\includegraphics[width=1.10cm,height=1.8cm,keepaspectratio]{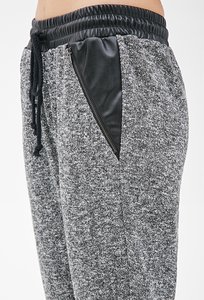}\hss\phantom{\rule{1.10cm}{1.8cm}}\hss}
      \end{minipage}
    }
  \end{minipage}
\hfill
  \begin{minipage}[t]{5.55cm}
    \fcolorbox{red!70!black}{white}{
      \begin{minipage}[t]{5.25cm}
        {\tiny\textbf{\#2}}\\
        \noindent\hbox to 5.25cm{\includegraphics[width=1.10cm,height=1.8cm,keepaspectratio]{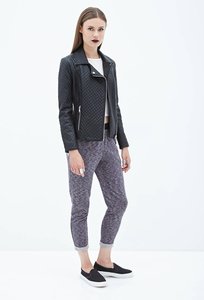}\hss\includegraphics[width=1.10cm,height=1.8cm,keepaspectratio]{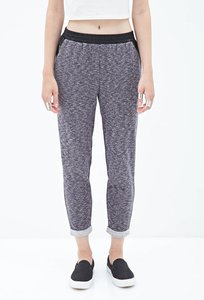}\hss\includegraphics[width=1.10cm,height=1.8cm,keepaspectratio]{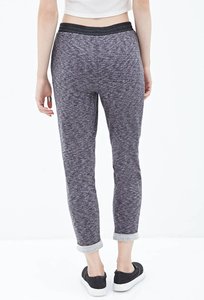}\hss\includegraphics[width=1.10cm,height=1.8cm,keepaspectratio]{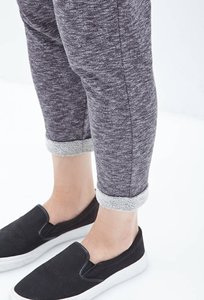}\hss\includegraphics[width=1.10cm,height=1.8cm,keepaspectratio]{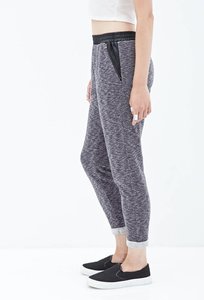}\hss}
      \end{minipage}
    }
  \end{minipage}
\hfill
  \begin{minipage}[t]{5.55cm}
    \fcolorbox{red!70!black}{white}{
      \begin{minipage}[t]{5.25cm}
        {\tiny\textbf{\#3}}\\
        \noindent\hbox to 5.25cm{\includegraphics[width=1.10cm,height=1.8cm,keepaspectratio]{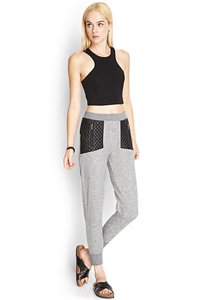}\hss\includegraphics[width=1.10cm,height=1.8cm,keepaspectratio]{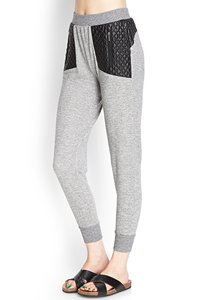}\hss\includegraphics[width=1.10cm,height=1.8cm,keepaspectratio]{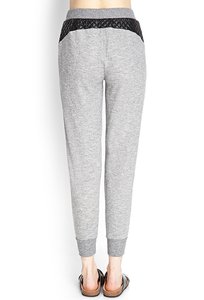}\hss\includegraphics[width=1.10cm,height=1.8cm,keepaspectratio]{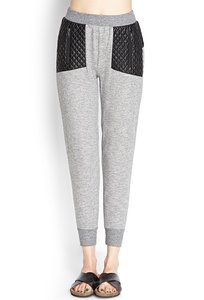}\hss\phantom{\rule{1.10cm}{1.8cm}}\hss}
      \end{minipage}
    }
  \end{minipage}
\par\vspace{2pt}
\noindent
  \begin{minipage}[t]{5.55cm}
    \fcolorbox{red!70!black}{white}{
      \begin{minipage}[t]{5.25cm}
        {\tiny\textbf{\#4}}\\
        \noindent\hbox to 5.25cm{\includegraphics[width=1.10cm,height=1.8cm,keepaspectratio]{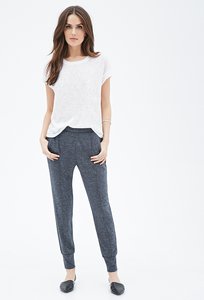}\hss\includegraphics[width=1.10cm,height=1.8cm,keepaspectratio]{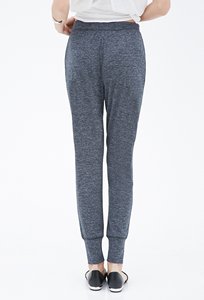}\hss\includegraphics[width=1.10cm,height=1.8cm,keepaspectratio]{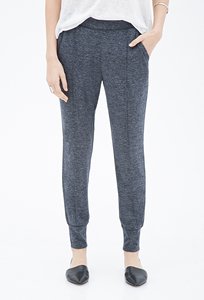}\hss\includegraphics[width=1.10cm,height=1.8cm,keepaspectratio]{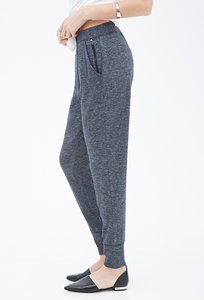}\hss\phantom{\rule{1.10cm}{1.8cm}}\hss}
      \end{minipage}
    }
  \end{minipage}
\hfill
  \begin{minipage}[t]{5.55cm}
    \fcolorbox{red!70!black}{white}{
      \begin{minipage}[t]{5.25cm}
        {\tiny\textbf{\#5}}\\
        \noindent\hbox to 5.25cm{\includegraphics[width=1.10cm,height=1.8cm,keepaspectratio]{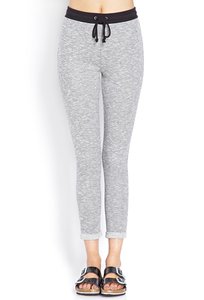}\hss\includegraphics[width=1.10cm,height=1.8cm,keepaspectratio]{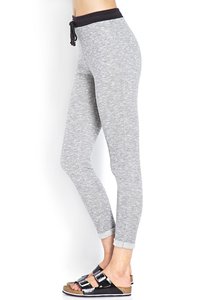}\hss\includegraphics[width=1.10cm,height=1.8cm,keepaspectratio]{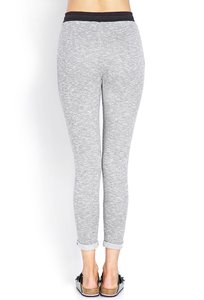}\hss\includegraphics[width=1.10cm,height=1.8cm,keepaspectratio]{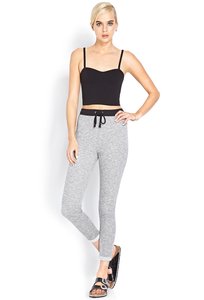}\hss\phantom{\rule{1.10cm}{1.8cm}}\hss}
      \end{minipage}
    }
  \end{minipage}
\hfill
  \begin{minipage}[t]{5.55cm}
    \fcolorbox{red!70!black}{white}{
      \begin{minipage}[t]{5.25cm}
        {\tiny\textbf{\#6}}\\
        \noindent\hbox to 5.25cm{\includegraphics[width=1.10cm,height=1.8cm,keepaspectratio]{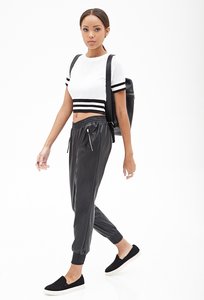}\hss\includegraphics[width=1.10cm,height=1.8cm,keepaspectratio]{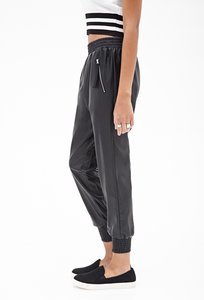}\hss\includegraphics[width=1.10cm,height=1.8cm,keepaspectratio]{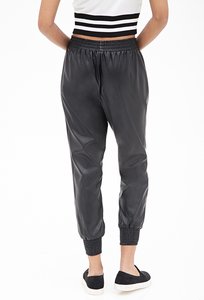}\hss\includegraphics[width=1.10cm,height=1.8cm,keepaspectratio]{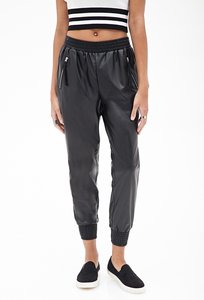}\hss\phantom{\rule{1.10cm}{1.8cm}}\hss}
      \end{minipage}
    }
  \end{minipage}
\par\vspace{2pt}
\noindent
  \begin{minipage}[t]{5.55cm}
    \fcolorbox{red!70!black}{white}{
      \begin{minipage}[t]{5.25cm}
        {\tiny\textbf{\#7}}\\
        \noindent\hbox to 5.25cm{\includegraphics[width=1.10cm,height=1.8cm,keepaspectratio]{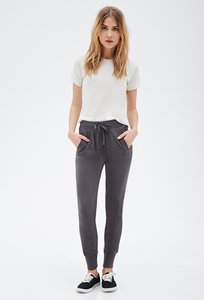}\hss\includegraphics[width=1.10cm,height=1.8cm,keepaspectratio]{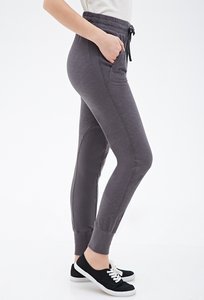}\hss\includegraphics[width=1.10cm,height=1.8cm,keepaspectratio]{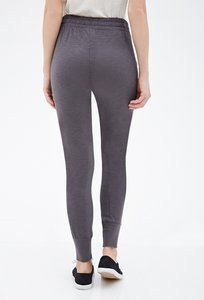}\hss\includegraphics[width=1.10cm,height=1.8cm,keepaspectratio]{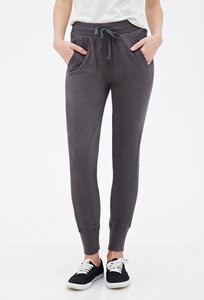}\hss\includegraphics[width=1.10cm,height=1.8cm,keepaspectratio]{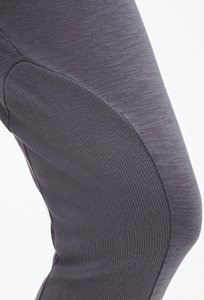}\hss}
      \end{minipage}
    }
  \end{minipage}
\hfill
  \begin{minipage}[t]{5.55cm}
    \fcolorbox{red!70!black}{white}{
      \begin{minipage}[t]{5.25cm}
        {\tiny\textbf{\#8}}\\
        \noindent\hbox to 5.25cm{\includegraphics[width=1.10cm,height=1.8cm,keepaspectratio]{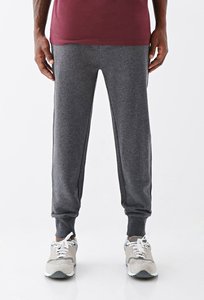}\hss\includegraphics[width=1.10cm,height=1.8cm,keepaspectratio]{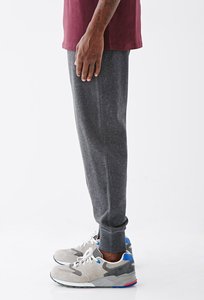}\hss\includegraphics[width=1.10cm,height=1.8cm,keepaspectratio]{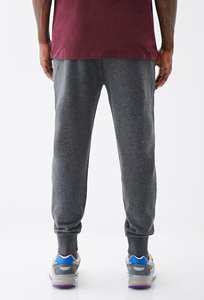}\hss\includegraphics[width=1.10cm,height=1.8cm,keepaspectratio]{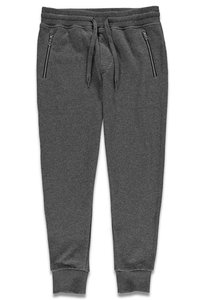}\hss\includegraphics[width=1.10cm,height=1.8cm,keepaspectratio]{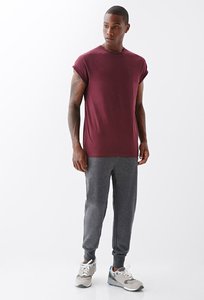}\hss}
      \end{minipage}
    }
  \end{minipage}
\hfill
  \begin{minipage}[t]{5.55cm}
    \fcolorbox{red!70!black}{white}{
      \begin{minipage}[t]{5.25cm}
        {\tiny\textbf{\#9}}\\
        \noindent\hbox to 5.25cm{\includegraphics[width=1.10cm,height=1.8cm,keepaspectratio]{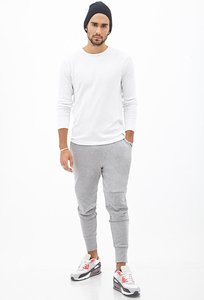}\hss\includegraphics[width=1.10cm,height=1.8cm,keepaspectratio]{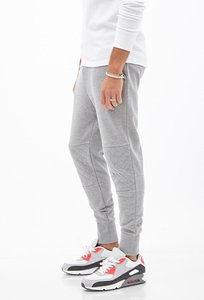}\hss\includegraphics[width=1.10cm,height=1.8cm,keepaspectratio]{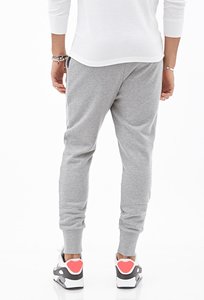}\hss\includegraphics[width=1.10cm,height=1.8cm,keepaspectratio]{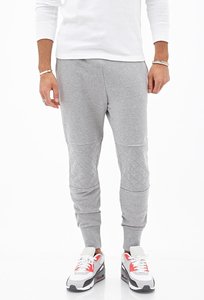}\hss\phantom{\rule{1.10cm}{1.8cm}}\hss}
      \end{minipage}
    }
  \end{minipage}
\par\vspace{2pt}
\noindent
  \begin{minipage}[t]{5.55cm}
    \fcolorbox{red!70!black}{white}{
      \begin{minipage}[t]{5.25cm}
        {\tiny\textbf{\#10}}\\
        \noindent\hbox to 5.25cm{\includegraphics[width=1.10cm,height=1.8cm,keepaspectratio]{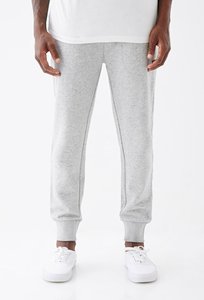}\hss\includegraphics[width=1.10cm,height=1.8cm,keepaspectratio]{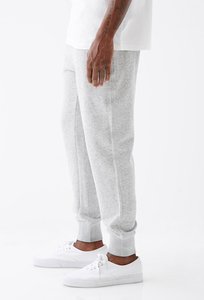}\hss\includegraphics[width=1.10cm,height=1.8cm,keepaspectratio]{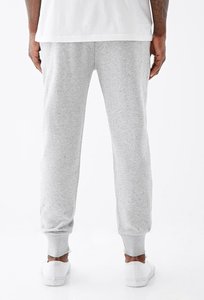}\hss\includegraphics[width=1.10cm,height=1.8cm,keepaspectratio]{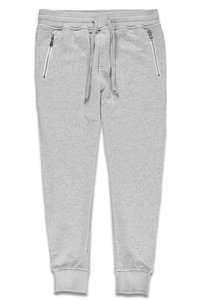}\hss\includegraphics[width=1.10cm,height=1.8cm,keepaspectratio]{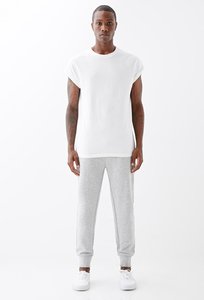}\hss}
      \end{minipage}
    }
  \end{minipage}
\hfill
  \begin{minipage}[t]{5.55cm}\end{minipage}
\hfill
  \begin{minipage}[t]{5.55cm}\end{minipage}
\par\vspace{2pt}
\noindent\rule{\linewidth}{0.4pt}
\par\vspace{2pt}
% -- qwen3_vl_2b --
{\small\textbf{Qwen3-VL-2B}}\quad{\small Rank~3}\\
\noindent
  \begin{minipage}[t]{5.55cm}
    \fcolorbox{red!70!black}{white}{
      \begin{minipage}[t]{5.25cm}
        {\tiny\textbf{\#1}}\\
        \noindent\hbox to 5.25cm{\includegraphics[width=1.10cm,height=1.8cm,keepaspectratio]{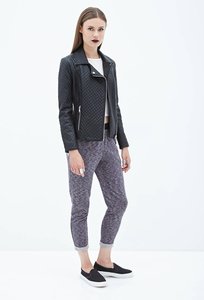}\hss\includegraphics[width=1.10cm,height=1.8cm,keepaspectratio]{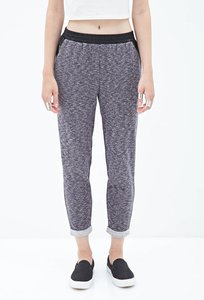}\hss\includegraphics[width=1.10cm,height=1.8cm,keepaspectratio]{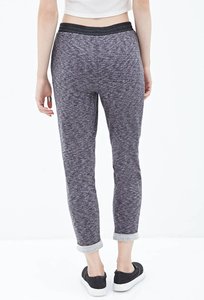}\hss\includegraphics[width=1.10cm,height=1.8cm,keepaspectratio]{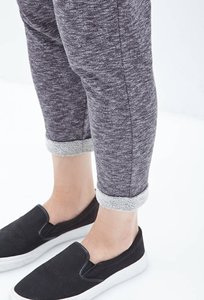}\hss\includegraphics[width=1.10cm,height=1.8cm,keepaspectratio]{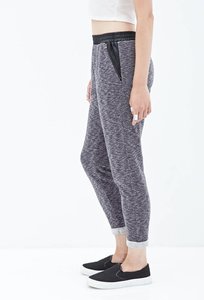}\hss}
      \end{minipage}
    }
  \end{minipage}
\hfill
  \begin{minipage}[t]{5.55cm}
    \fcolorbox{red!70!black}{white}{
      \begin{minipage}[t]{5.25cm}
        {\tiny\textbf{\#2}}\\
        \noindent\hbox to 5.25cm{\includegraphics[width=1.10cm,height=1.8cm,keepaspectratio]{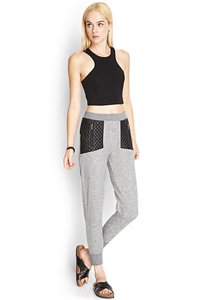}\hss\includegraphics[width=1.10cm,height=1.8cm,keepaspectratio]{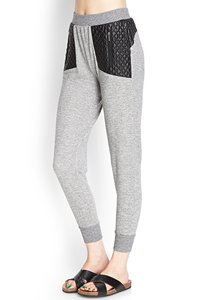}\hss\includegraphics[width=1.10cm,height=1.8cm,keepaspectratio]{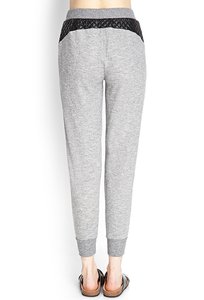}\hss\includegraphics[width=1.10cm,height=1.8cm,keepaspectratio]{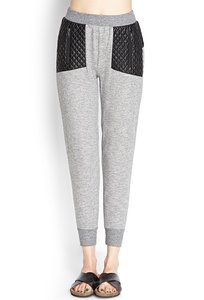}\hss\phantom{\rule{1.10cm}{1.8cm}}\hss}
      \end{minipage}
    }
  \end{minipage}
\hfill
  \begin{minipage}[t]{5.55cm}
    \fcolorbox{green!60!black}{white}{
      \begin{minipage}[t]{5.25cm}
        {\tiny\textbf{\#3}}\\
        \noindent\hbox to 5.25cm{\includegraphics[width=1.10cm,height=1.8cm,keepaspectratio]{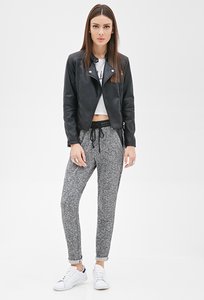}\hss\includegraphics[width=1.10cm,height=1.8cm,keepaspectratio]{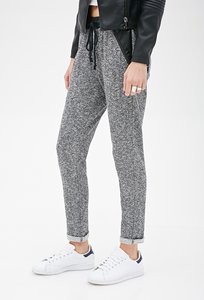}\hss\includegraphics[width=1.10cm,height=1.8cm,keepaspectratio]{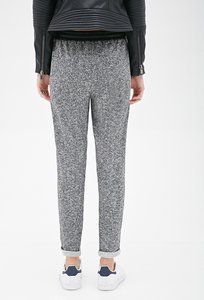}\hss\includegraphics[width=1.10cm,height=1.8cm,keepaspectratio]{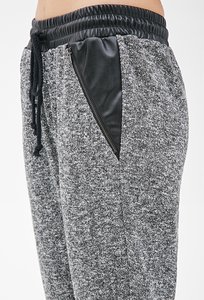}\hss\phantom{\rule{1.10cm}{1.8cm}}\hss}
      \end{minipage}
    }
  \end{minipage}
\par\vspace{2pt}
\noindent
  \begin{minipage}[t]{5.55cm}
    \fcolorbox{red!70!black}{white}{
      \begin{minipage}[t]{5.25cm}
        {\tiny\textbf{\#4}}\\
        \noindent\hbox to 5.25cm{\includegraphics[width=1.10cm,height=1.8cm,keepaspectratio]{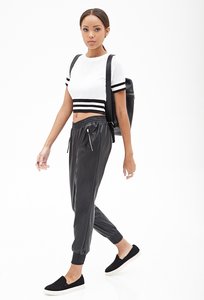}\hss\includegraphics[width=1.10cm,height=1.8cm,keepaspectratio]{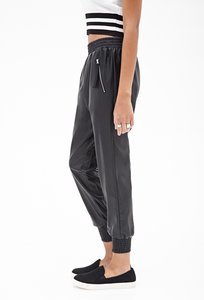}\hss\includegraphics[width=1.10cm,height=1.8cm,keepaspectratio]{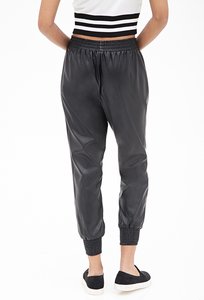}\hss\includegraphics[width=1.10cm,height=1.8cm,keepaspectratio]{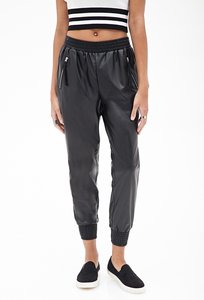}\hss\phantom{\rule{1.10cm}{1.8cm}}\hss}
      \end{minipage}
    }
  \end{minipage}
\hfill
  \begin{minipage}[t]{5.55cm}
    \fcolorbox{red!70!black}{white}{
      \begin{minipage}[t]{5.25cm}
        {\tiny\textbf{\#5}}\\
        \noindent\hbox to 5.25cm{\includegraphics[width=1.10cm,height=1.8cm,keepaspectratio]{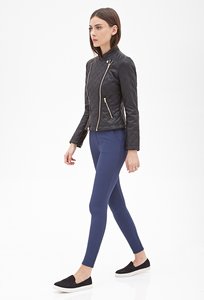}\hss\includegraphics[width=1.10cm,height=1.8cm,keepaspectratio]{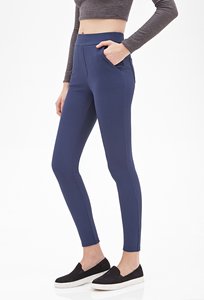}\hss\includegraphics[width=1.10cm,height=1.8cm,keepaspectratio]{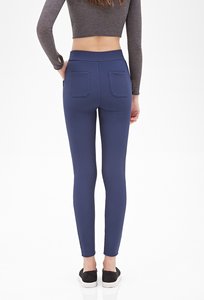}\hss\includegraphics[width=1.10cm,height=1.8cm,keepaspectratio]{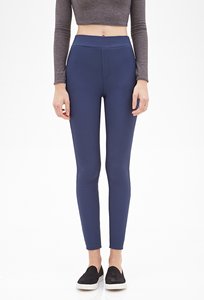}\hss\phantom{\rule{1.10cm}{1.8cm}}\hss}
      \end{minipage}
    }
  \end{minipage}
\hfill
  \begin{minipage}[t]{5.55cm}
    \fcolorbox{red!70!black}{white}{
      \begin{minipage}[t]{5.25cm}
        {\tiny\textbf{\#6}}\\
        \noindent\hbox to 5.25cm{\includegraphics[width=1.10cm,height=1.8cm,keepaspectratio]{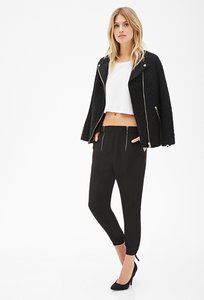}\hss\includegraphics[width=1.10cm,height=1.8cm,keepaspectratio]{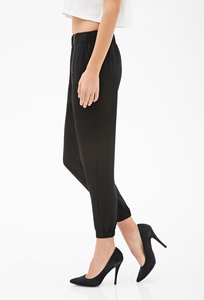}\hss\includegraphics[width=1.10cm,height=1.8cm,keepaspectratio]{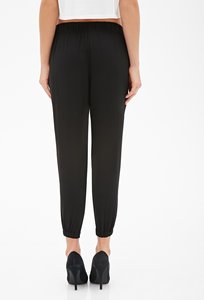}\hss\includegraphics[width=1.10cm,height=1.8cm,keepaspectratio]{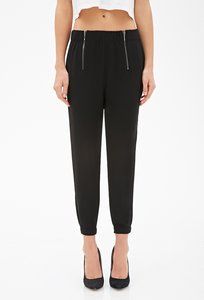}\hss\phantom{\rule{1.10cm}{1.8cm}}\hss}
      \end{minipage}
    }
  \end{minipage}
\par\vspace{2pt}
\noindent
  \begin{minipage}[t]{5.55cm}
    \fcolorbox{red!70!black}{white}{
      \begin{minipage}[t]{5.25cm}
        {\tiny\textbf{\#7}}\\
        \noindent\hbox to 5.25cm{\includegraphics[width=1.10cm,height=1.8cm,keepaspectratio]{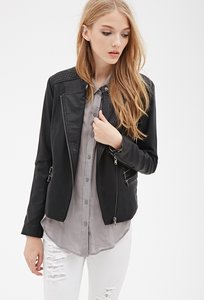}\hss\includegraphics[width=1.10cm,height=1.8cm,keepaspectratio]{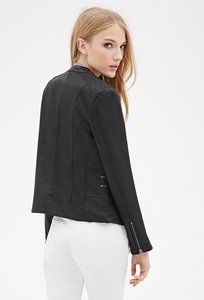}\hss\includegraphics[width=1.10cm,height=1.8cm,keepaspectratio]{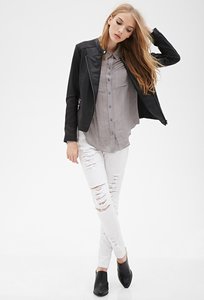}\hss\includegraphics[width=1.10cm,height=1.8cm,keepaspectratio]{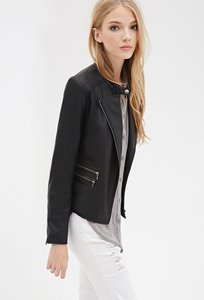}\hss\phantom{\rule{1.10cm}{1.8cm}}\hss}
      \end{minipage}
    }
  \end{minipage}
\hfill
  \begin{minipage}[t]{5.55cm}
    \fcolorbox{red!70!black}{white}{
      \begin{minipage}[t]{5.25cm}
        {\tiny\textbf{\#8}}\\
        \noindent\hbox to 5.25cm{\includegraphics[width=1.10cm,height=1.8cm,keepaspectratio]{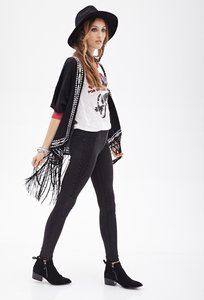}\hss\includegraphics[width=1.10cm,height=1.8cm,keepaspectratio]{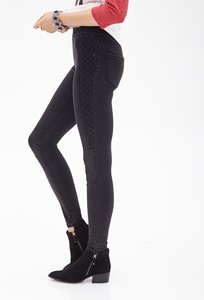}\hss\includegraphics[width=1.10cm,height=1.8cm,keepaspectratio]{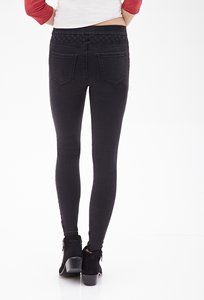}\hss\includegraphics[width=1.10cm,height=1.8cm,keepaspectratio]{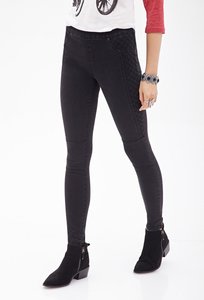}\hss\phantom{\rule{1.10cm}{1.8cm}}\hss}
      \end{minipage}
    }
  \end{minipage}
\hfill
  \begin{minipage}[t]{5.55cm}
    \fcolorbox{red!70!black}{white}{
      \begin{minipage}[t]{5.25cm}
        {\tiny\textbf{\#9}}\\
        \noindent\hbox to 5.25cm{\includegraphics[width=1.10cm,height=1.8cm,keepaspectratio]{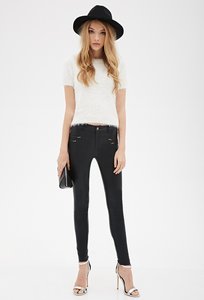}\hss\includegraphics[width=1.10cm,height=1.8cm,keepaspectratio]{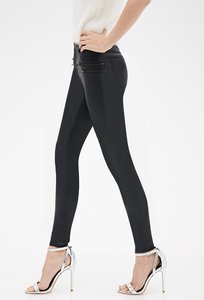}\hss\includegraphics[width=1.10cm,height=1.8cm,keepaspectratio]{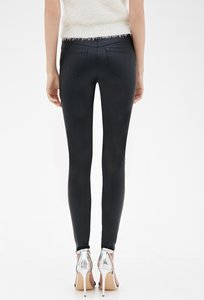}\hss\includegraphics[width=1.10cm,height=1.8cm,keepaspectratio]{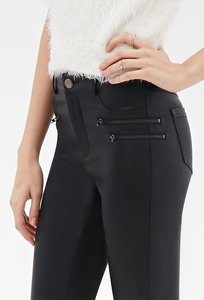}\hss\phantom{\rule{1.10cm}{1.8cm}}\hss}
      \end{minipage}
    }
  \end{minipage}
\par\vspace{2pt}
\noindent
  \begin{minipage}[t]{5.55cm}
    \fcolorbox{red!70!black}{white}{
      \begin{minipage}[t]{5.25cm}
        {\tiny\textbf{\#10}}\\
        \noindent\hbox to 5.25cm{\includegraphics[width=1.10cm,height=1.8cm,keepaspectratio]{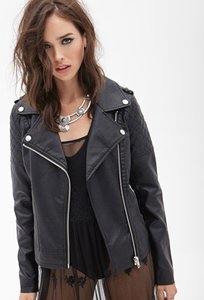}\hss\includegraphics[width=1.10cm,height=1.8cm,keepaspectratio]{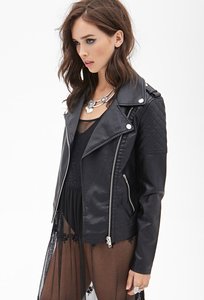}\hss\includegraphics[width=1.10cm,height=1.8cm,keepaspectratio]{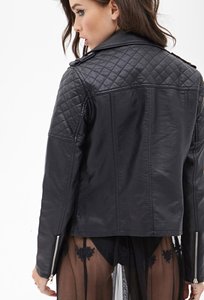}\hss\includegraphics[width=1.10cm,height=1.8cm,keepaspectratio]{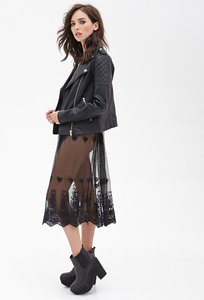}\hss\phantom{\rule{1.10cm}{1.8cm}}\hss}
      \end{minipage}
    }
  \end{minipage}
\hfill
  \begin{minipage}[t]{5.55cm}\end{minipage}
\hfill
  \begin{minipage}[t]{5.55cm}\end{minipage}
\par\vspace{2pt}
\noindent\rule{\linewidth}{0.4pt}
\par\vspace{2pt}
% -- qwen3_vl_8b --
{\small\textbf{Qwen3-VL-8B}}\quad{\small Rank~3}\\
\noindent
  \begin{minipage}[t]{5.55cm}
    \fcolorbox{red!70!black}{white}{
      \begin{minipage}[t]{5.25cm}
        {\tiny\textbf{\#1}}\\
        \noindent\hbox to 5.25cm{\includegraphics[width=1.10cm,height=1.8cm,keepaspectratio]{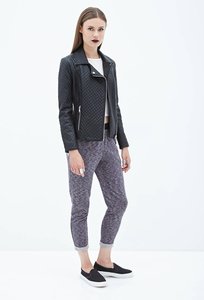}\hss\includegraphics[width=1.10cm,height=1.8cm,keepaspectratio]{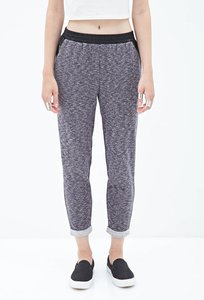}\hss\includegraphics[width=1.10cm,height=1.8cm,keepaspectratio]{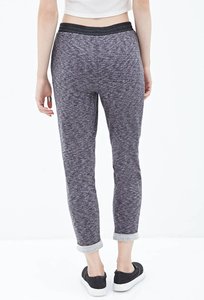}\hss\includegraphics[width=1.10cm,height=1.8cm,keepaspectratio]{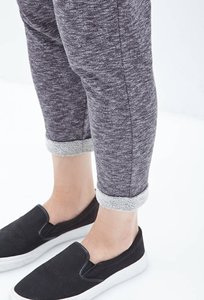}\hss\includegraphics[width=1.10cm,height=1.8cm,keepaspectratio]{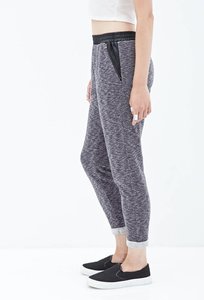}\hss}
      \end{minipage}
    }
  \end{minipage}
\hfill
  \begin{minipage}[t]{5.55cm}
    \fcolorbox{red!70!black}{white}{
      \begin{minipage}[t]{5.25cm}
        {\tiny\textbf{\#2}}\\
        \noindent\hbox to 5.25cm{\includegraphics[width=1.10cm,height=1.8cm,keepaspectratio]{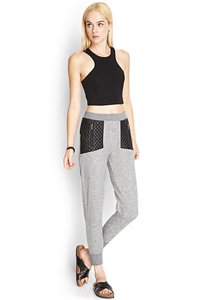}\hss\includegraphics[width=1.10cm,height=1.8cm,keepaspectratio]{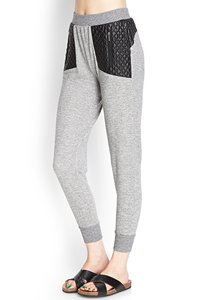}\hss\includegraphics[width=1.10cm,height=1.8cm,keepaspectratio]{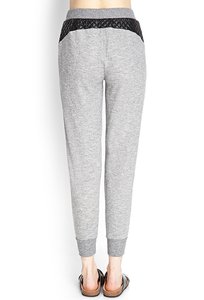}\hss\includegraphics[width=1.10cm,height=1.8cm,keepaspectratio]{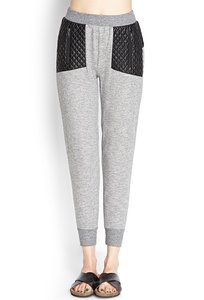}\hss\phantom{\rule{1.10cm}{1.8cm}}\hss}
      \end{minipage}
    }
  \end{minipage}
\hfill
  \begin{minipage}[t]{5.55cm}
    \fcolorbox{green!60!black}{white}{
      \begin{minipage}[t]{5.25cm}
        {\tiny\textbf{\#3}}\\
        \noindent\hbox to 5.25cm{\includegraphics[width=1.10cm,height=1.8cm,keepaspectratio]{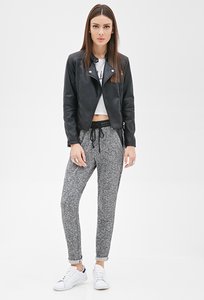}\hss\includegraphics[width=1.10cm,height=1.8cm,keepaspectratio]{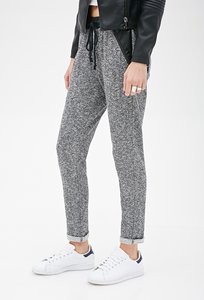}\hss\includegraphics[width=1.10cm,height=1.8cm,keepaspectratio]{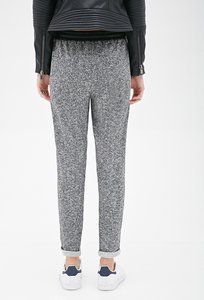}\hss\includegraphics[width=1.10cm,height=1.8cm,keepaspectratio]{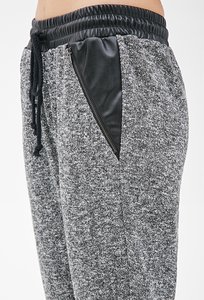}\hss\phantom{\rule{1.10cm}{1.8cm}}\hss}
      \end{minipage}
    }
  \end{minipage}
\par\vspace{2pt}
\noindent
  \begin{minipage}[t]{5.55cm}
    \fcolorbox{red!70!black}{white}{
      \begin{minipage}[t]{5.25cm}
        {\tiny\textbf{\#4}}\\
        \noindent\hbox to 5.25cm{\includegraphics[width=1.10cm,height=1.8cm,keepaspectratio]{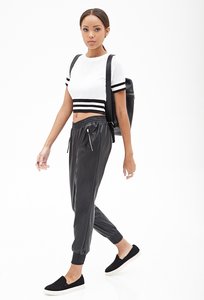}\hss\includegraphics[width=1.10cm,height=1.8cm,keepaspectratio]{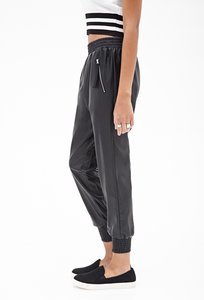}\hss\includegraphics[width=1.10cm,height=1.8cm,keepaspectratio]{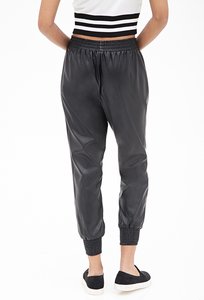}\hss\includegraphics[width=1.10cm,height=1.8cm,keepaspectratio]{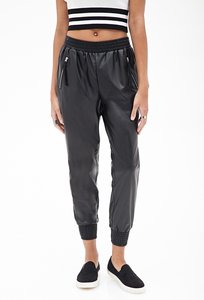}\hss\phantom{\rule{1.10cm}{1.8cm}}\hss}
      \end{minipage}
    }
  \end{minipage}
\hfill
  \begin{minipage}[t]{5.55cm}
    \fcolorbox{red!70!black}{white}{
      \begin{minipage}[t]{5.25cm}
        {\tiny\textbf{\#5}}\\
        \noindent\hbox to 5.25cm{\includegraphics[width=1.10cm,height=1.8cm,keepaspectratio]{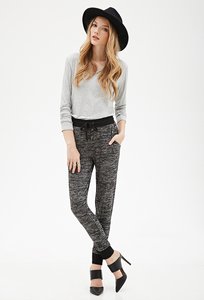}\hss\includegraphics[width=1.10cm,height=1.8cm,keepaspectratio]{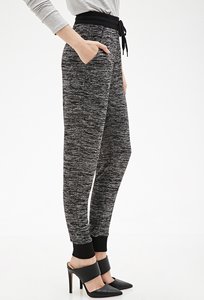}\hss\includegraphics[width=1.10cm,height=1.8cm,keepaspectratio]{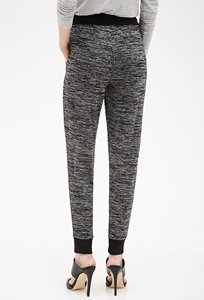}\hss\includegraphics[width=1.10cm,height=1.8cm,keepaspectratio]{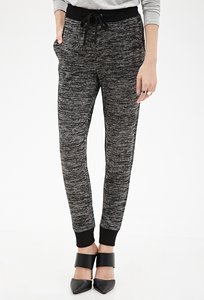}\hss\phantom{\rule{1.10cm}{1.8cm}}\hss}
      \end{minipage}
    }
  \end{minipage}
\hfill
  \begin{minipage}[t]{5.55cm}
    \fcolorbox{red!70!black}{white}{
      \begin{minipage}[t]{5.25cm}
        {\tiny\textbf{\#6}}\\
        \noindent\hbox to 5.25cm{\includegraphics[width=1.10cm,height=1.8cm,keepaspectratio]{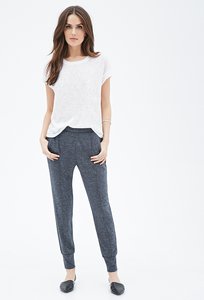}\hss\includegraphics[width=1.10cm,height=1.8cm,keepaspectratio]{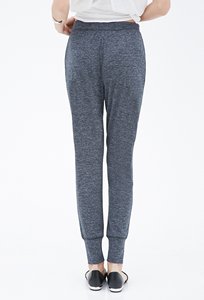}\hss\includegraphics[width=1.10cm,height=1.8cm,keepaspectratio]{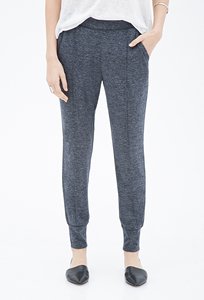}\hss\includegraphics[width=1.10cm,height=1.8cm,keepaspectratio]{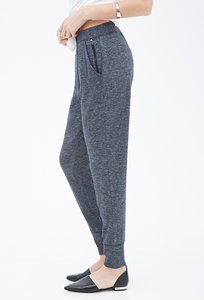}\hss\phantom{\rule{1.10cm}{1.8cm}}\hss}
      \end{minipage}
    }
  \end{minipage}
\par\vspace{2pt}
\noindent
  \begin{minipage}[t]{5.55cm}
    \fcolorbox{red!70!black}{white}{
      \begin{minipage}[t]{5.25cm}
        {\tiny\textbf{\#7}}\\
        \noindent\hbox to 5.25cm{\includegraphics[width=1.10cm,height=1.8cm,keepaspectratio]{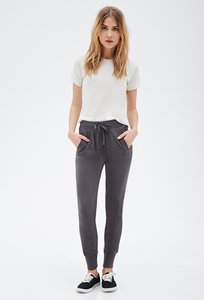}\hss\includegraphics[width=1.10cm,height=1.8cm,keepaspectratio]{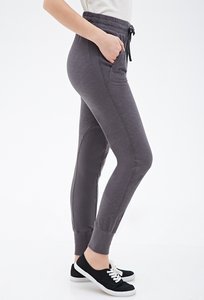}\hss\includegraphics[width=1.10cm,height=1.8cm,keepaspectratio]{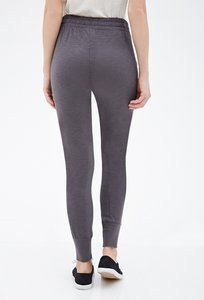}\hss\includegraphics[width=1.10cm,height=1.8cm,keepaspectratio]{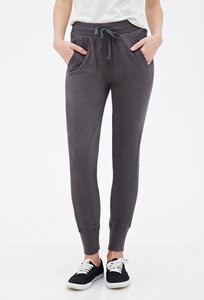}\hss\includegraphics[width=1.10cm,height=1.8cm,keepaspectratio]{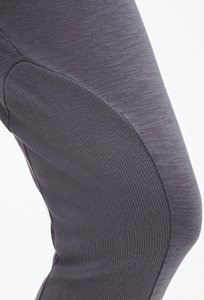}\hss}
      \end{minipage}
    }
  \end{minipage}
\hfill
  \begin{minipage}[t]{5.55cm}
    \fcolorbox{red!70!black}{white}{
      \begin{minipage}[t]{5.25cm}
        {\tiny\textbf{\#8}}\\
        \noindent\hbox to 5.25cm{\includegraphics[width=1.10cm,height=1.8cm,keepaspectratio]{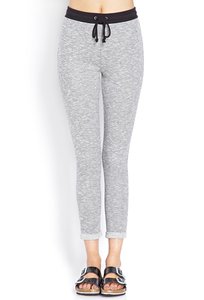}\hss\includegraphics[width=1.10cm,height=1.8cm,keepaspectratio]{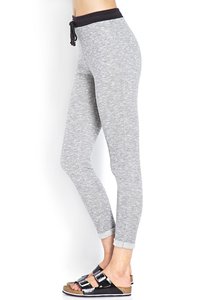}\hss\includegraphics[width=1.10cm,height=1.8cm,keepaspectratio]{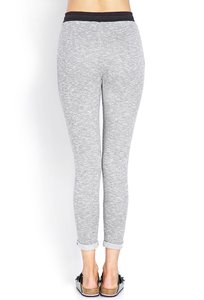}\hss\includegraphics[width=1.10cm,height=1.8cm,keepaspectratio]{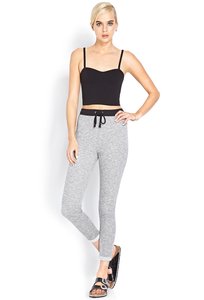}\hss\phantom{\rule{1.10cm}{1.8cm}}\hss}
      \end{minipage}
    }
  \end{minipage}
\hfill
  \begin{minipage}[t]{5.55cm}
    \fcolorbox{red!70!black}{white}{
      \begin{minipage}[t]{5.25cm}
        {\tiny\textbf{\#9}}\\
        \noindent\hbox to 5.25cm{\includegraphics[width=1.10cm,height=1.8cm,keepaspectratio]{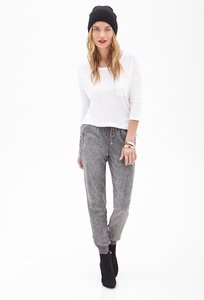}\hss\includegraphics[width=1.10cm,height=1.8cm,keepaspectratio]{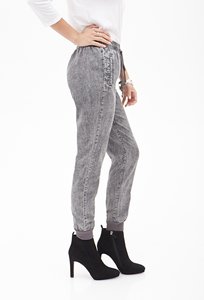}\hss\includegraphics[width=1.10cm,height=1.8cm,keepaspectratio]{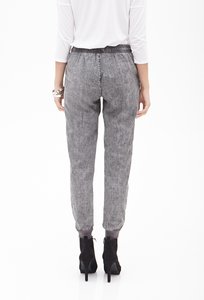}\hss\includegraphics[width=1.10cm,height=1.8cm,keepaspectratio]{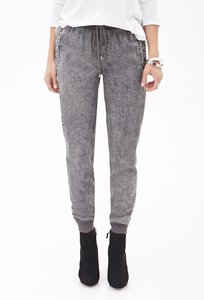}\hss\phantom{\rule{1.10cm}{1.8cm}}\hss}
      \end{minipage}
    }
  \end{minipage}
\par\vspace{2pt}
\noindent
  \begin{minipage}[t]{5.55cm}
    \fcolorbox{red!70!black}{white}{
      \begin{minipage}[t]{5.25cm}
        {\tiny\textbf{\#10}}\\
        \noindent\hbox to 5.25cm{\includegraphics[width=1.10cm,height=1.8cm,keepaspectratio]{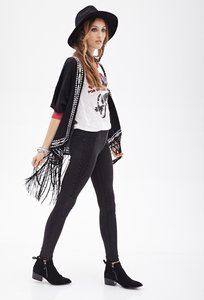}\hss\includegraphics[width=1.10cm,height=1.8cm,keepaspectratio]{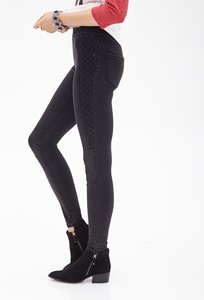}\hss\includegraphics[width=1.10cm,height=1.8cm,keepaspectratio]{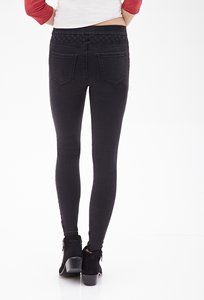}\hss\includegraphics[width=1.10cm,height=1.8cm,keepaspectratio]{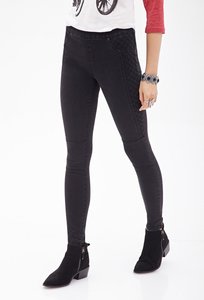}\hss\phantom{\rule{1.10cm}{1.8cm}}\hss}
      \end{minipage}
    }
  \end{minipage}
\hfill
  \begin{minipage}[t]{5.55cm}\end{minipage}
\hfill
  \begin{minipage}[t]{5.55cm}\end{minipage}
\par\vspace{2pt}
\noindent\rule{\linewidth}{0.4pt}
\par\vspace{2pt}
% -- reznembed --
{\small\textbf{RezNEmbed}}\quad{\small Rank~2}\\
\noindent
  \begin{minipage}[t]{5.55cm}
    \fcolorbox{red!70!black}{white}{
      \begin{minipage}[t]{5.25cm}
        {\tiny\textbf{\#1}}\\
        \noindent\hbox to 5.25cm{\includegraphics[width=1.10cm,height=1.8cm,keepaspectratio]{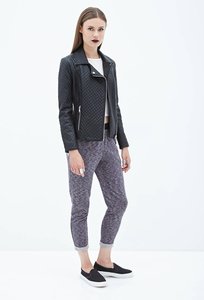}\hss\includegraphics[width=1.10cm,height=1.8cm,keepaspectratio]{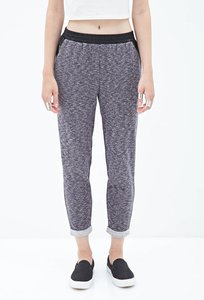}\hss\includegraphics[width=1.10cm,height=1.8cm,keepaspectratio]{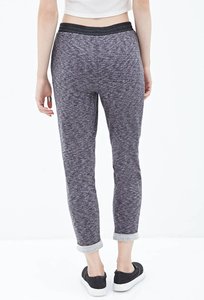}\hss\includegraphics[width=1.10cm,height=1.8cm,keepaspectratio]{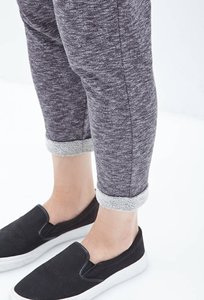}\hss\includegraphics[width=1.10cm,height=1.8cm,keepaspectratio]{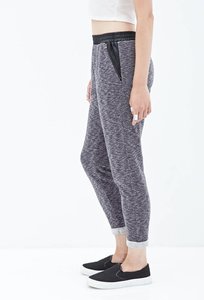}\hss}
      \end{minipage}
    }
  \end{minipage}
\hfill
  \begin{minipage}[t]{5.55cm}
    \fcolorbox{green!60!black}{white}{
      \begin{minipage}[t]{5.25cm}
        {\tiny\textbf{\#2}}\\
        \noindent\hbox to 5.25cm{\includegraphics[width=1.10cm,height=1.8cm,keepaspectratio]{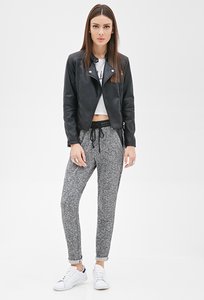}\hss\includegraphics[width=1.10cm,height=1.8cm,keepaspectratio]{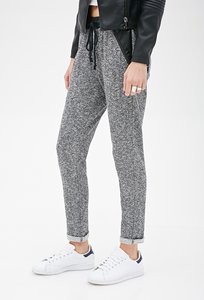}\hss\includegraphics[width=1.10cm,height=1.8cm,keepaspectratio]{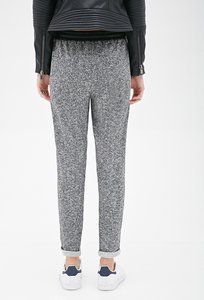}\hss\includegraphics[width=1.10cm,height=1.8cm,keepaspectratio]{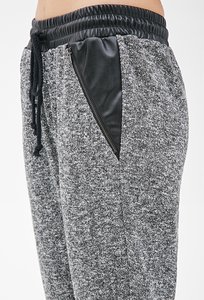}\hss\phantom{\rule{1.10cm}{1.8cm}}\hss}
      \end{minipage}
    }
  \end{minipage}
\hfill
  \begin{minipage}[t]{5.55cm}
    \fcolorbox{red!70!black}{white}{
      \begin{minipage}[t]{5.25cm}
        {\tiny\textbf{\#3}}\\
        \noindent\hbox to 5.25cm{\includegraphics[width=1.10cm,height=1.8cm,keepaspectratio]{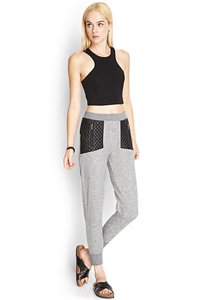}\hss\includegraphics[width=1.10cm,height=1.8cm,keepaspectratio]{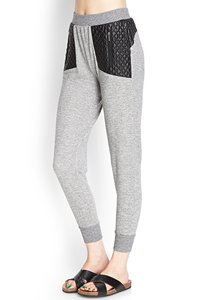}\hss\includegraphics[width=1.10cm,height=1.8cm,keepaspectratio]{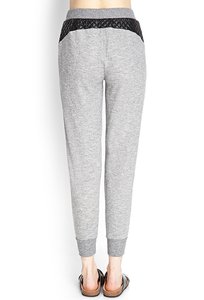}\hss\includegraphics[width=1.10cm,height=1.8cm,keepaspectratio]{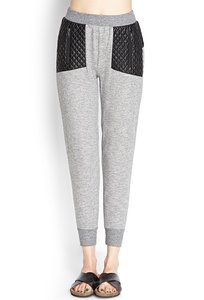}\hss\phantom{\rule{1.10cm}{1.8cm}}\hss}
      \end{minipage}
    }
  \end{minipage}
\par\vspace{2pt}
\noindent
  \begin{minipage}[t]{5.55cm}
    \fcolorbox{red!70!black}{white}{
      \begin{minipage}[t]{5.25cm}
        {\tiny\textbf{\#4}}\\
        \noindent\hbox to 5.25cm{\includegraphics[width=1.10cm,height=1.8cm,keepaspectratio]{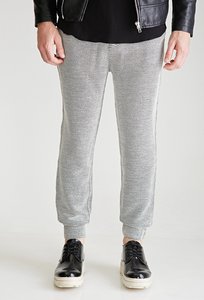}\hss\includegraphics[width=1.10cm,height=1.8cm,keepaspectratio]{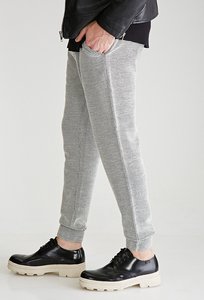}\hss\includegraphics[width=1.10cm,height=1.8cm,keepaspectratio]{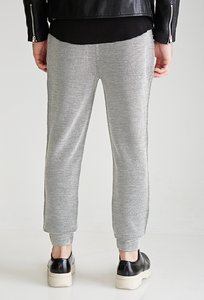}\hss\includegraphics[width=1.10cm,height=1.8cm,keepaspectratio]{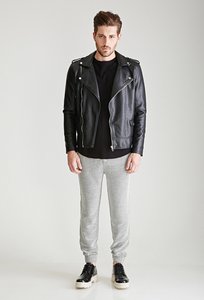}\hss\phantom{\rule{1.10cm}{1.8cm}}\hss}
      \end{minipage}
    }
  \end{minipage}
\hfill
  \begin{minipage}[t]{5.55cm}
    \fcolorbox{red!70!black}{white}{
      \begin{minipage}[t]{5.25cm}
        {\tiny\textbf{\#5}}\\
        \noindent\hbox to 5.25cm{\includegraphics[width=1.10cm,height=1.8cm,keepaspectratio]{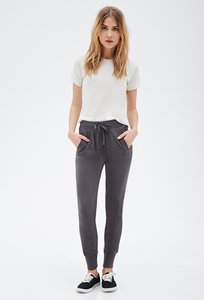}\hss\includegraphics[width=1.10cm,height=1.8cm,keepaspectratio]{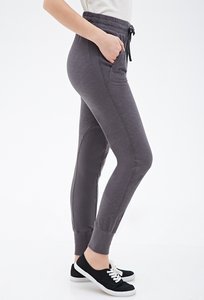}\hss\includegraphics[width=1.10cm,height=1.8cm,keepaspectratio]{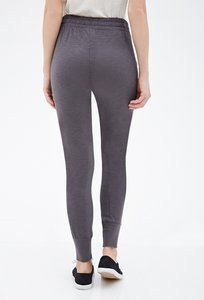}\hss\includegraphics[width=1.10cm,height=1.8cm,keepaspectratio]{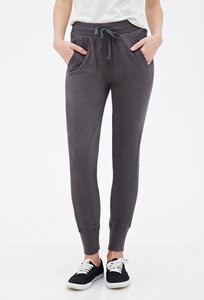}\hss\includegraphics[width=1.10cm,height=1.8cm,keepaspectratio]{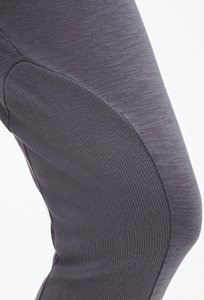}\hss}
      \end{minipage}
    }
  \end{minipage}
\hfill
  \begin{minipage}[t]{5.55cm}
    \fcolorbox{red!70!black}{white}{
      \begin{minipage}[t]{5.25cm}
        {\tiny\textbf{\#6}}\\
        \noindent\hbox to 5.25cm{\includegraphics[width=1.10cm,height=1.8cm,keepaspectratio]{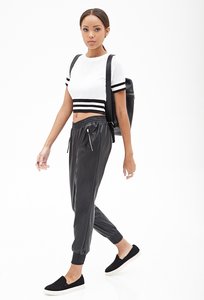}\hss\includegraphics[width=1.10cm,height=1.8cm,keepaspectratio]{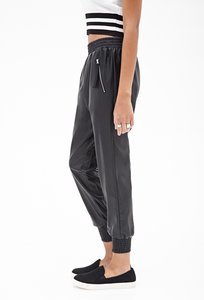}\hss\includegraphics[width=1.10cm,height=1.8cm,keepaspectratio]{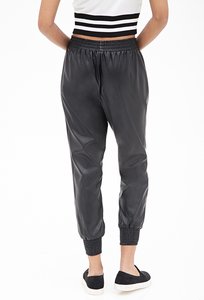}\hss\includegraphics[width=1.10cm,height=1.8cm,keepaspectratio]{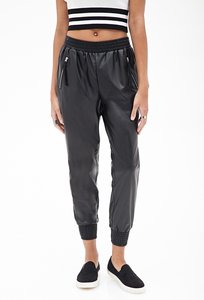}\hss\phantom{\rule{1.10cm}{1.8cm}}\hss}
      \end{minipage}
    }
  \end{minipage}
\par\vspace{2pt}
\noindent
  \begin{minipage}[t]{5.55cm}
    \fcolorbox{red!70!black}{white}{
      \begin{minipage}[t]{5.25cm}
        {\tiny\textbf{\#7}}\\
        \noindent\hbox to 5.25cm{\includegraphics[width=1.10cm,height=1.8cm,keepaspectratio]{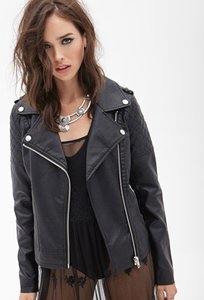}\hss\includegraphics[width=1.10cm,height=1.8cm,keepaspectratio]{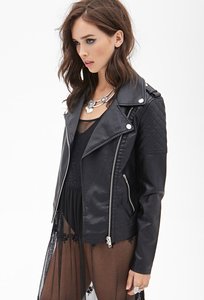}\hss\includegraphics[width=1.10cm,height=1.8cm,keepaspectratio]{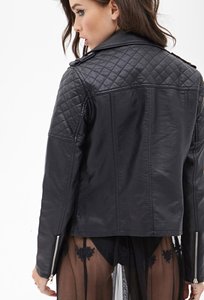}\hss\includegraphics[width=1.10cm,height=1.8cm,keepaspectratio]{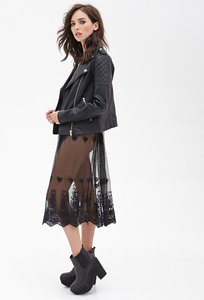}\hss\phantom{\rule{1.10cm}{1.8cm}}\hss}
      \end{minipage}
    }
  \end{minipage}
\hfill
  \begin{minipage}[t]{5.55cm}
    \fcolorbox{red!70!black}{white}{
      \begin{minipage}[t]{5.25cm}
        {\tiny\textbf{\#8}}\\
        \noindent\hbox to 5.25cm{\includegraphics[width=1.10cm,height=1.8cm,keepaspectratio]{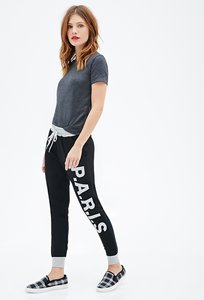}\hss\includegraphics[width=1.10cm,height=1.8cm,keepaspectratio]{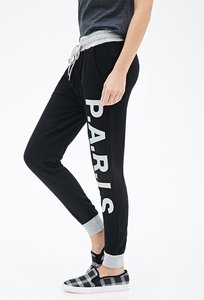}\hss\includegraphics[width=1.10cm,height=1.8cm,keepaspectratio]{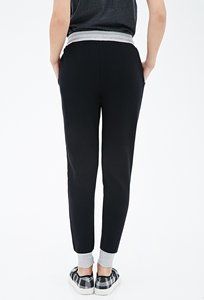}\hss\includegraphics[width=1.10cm,height=1.8cm,keepaspectratio]{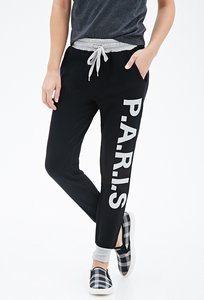}\hss\phantom{\rule{1.10cm}{1.8cm}}\hss}
      \end{minipage}
    }
  \end{minipage}
\hfill
  \begin{minipage}[t]{5.55cm}
    \fcolorbox{red!70!black}{white}{
      \begin{minipage}[t]{5.25cm}
        {\tiny\textbf{\#9}}\\
        \noindent\hbox to 5.25cm{\includegraphics[width=1.10cm,height=1.8cm,keepaspectratio]{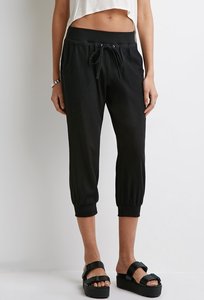}\hss\includegraphics[width=1.10cm,height=1.8cm,keepaspectratio]{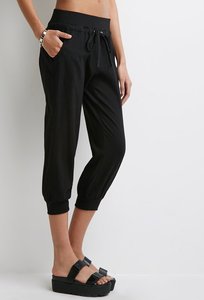}\hss\includegraphics[width=1.10cm,height=1.8cm,keepaspectratio]{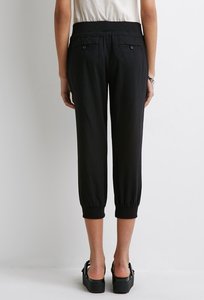}\hss\includegraphics[width=1.10cm,height=1.8cm,keepaspectratio]{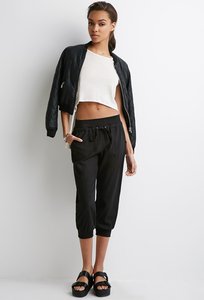}\hss\phantom{\rule{1.10cm}{1.8cm}}\hss}
      \end{minipage}
    }
  \end{minipage}
\par\vspace{2pt}
\noindent
  \begin{minipage}[t]{5.55cm}
    \fcolorbox{red!70!black}{white}{
      \begin{minipage}[t]{5.25cm}
        {\tiny\textbf{\#10}}\\
        \noindent\hbox to 5.25cm{\includegraphics[width=1.10cm,height=1.8cm,keepaspectratio]{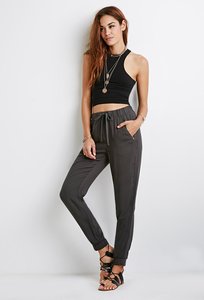}\hss\includegraphics[width=1.10cm,height=1.8cm,keepaspectratio]{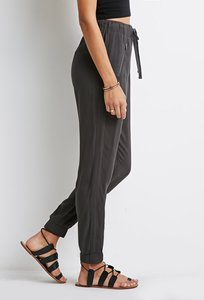}\hss\includegraphics[width=1.10cm,height=1.8cm,keepaspectratio]{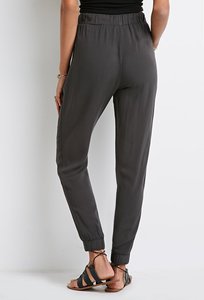}\hss\includegraphics[width=1.10cm,height=1.8cm,keepaspectratio]{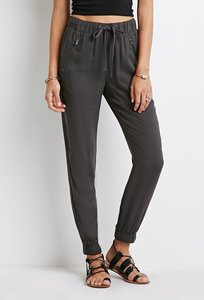}\hss\phantom{\rule{1.10cm}{1.8cm}}\hss}
      \end{minipage}
    }
  \end{minipage}
\hfill
  \begin{minipage}[t]{5.55cm}\end{minipage}
\hfill
  \begin{minipage}[t]{5.55cm}\end{minipage}
\par\vspace{2pt}
\noindent\rule{\linewidth}{0.4pt}
\par\vspace{2pt}
% -- doubao --
{\small\textbf{Doubao-E-V}}\quad{\small Rank~2}\\
\noindent
  \begin{minipage}[t]{5.55cm}
    \fcolorbox{red!70!black}{white}{
      \begin{minipage}[t]{5.25cm}
        {\tiny\textbf{\#1}}\\
        \noindent\hbox to 5.25cm{\includegraphics[width=1.10cm,height=1.8cm,keepaspectratio]{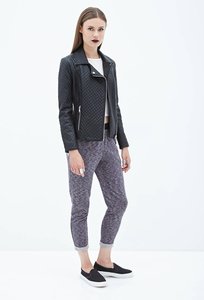}\hss\includegraphics[width=1.10cm,height=1.8cm,keepaspectratio]{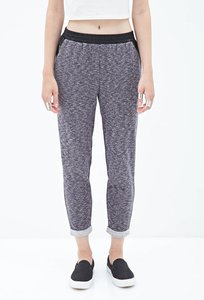}\hss\includegraphics[width=1.10cm,height=1.8cm,keepaspectratio]{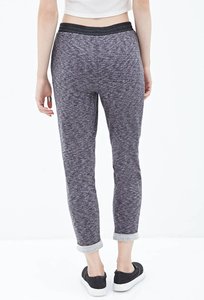}\hss\includegraphics[width=1.10cm,height=1.8cm,keepaspectratio]{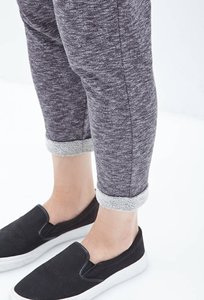}\hss\includegraphics[width=1.10cm,height=1.8cm,keepaspectratio]{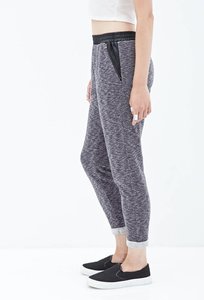}\hss}
      \end{minipage}
    }
  \end{minipage}
\hfill
  \begin{minipage}[t]{5.55cm}
    \fcolorbox{green!60!black}{white}{
      \begin{minipage}[t]{5.25cm}
        {\tiny\textbf{\#2}}\\
        \noindent\hbox to 5.25cm{\includegraphics[width=1.10cm,height=1.8cm,keepaspectratio]{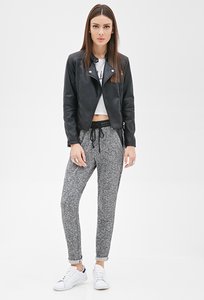}\hss\includegraphics[width=1.10cm,height=1.8cm,keepaspectratio]{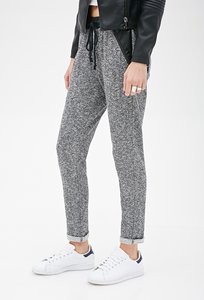}\hss\includegraphics[width=1.10cm,height=1.8cm,keepaspectratio]{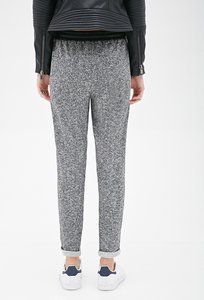}\hss\includegraphics[width=1.10cm,height=1.8cm,keepaspectratio]{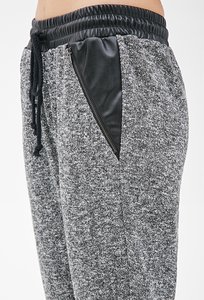}\hss\phantom{\rule{1.10cm}{1.8cm}}\hss}
      \end{minipage}
    }
  \end{minipage}
\hfill
  \begin{minipage}[t]{5.55cm}
    \fcolorbox{red!70!black}{white}{
      \begin{minipage}[t]{5.25cm}
        {\tiny\textbf{\#3}}\\
        \noindent\hbox to 5.25cm{\includegraphics[width=1.10cm,height=1.8cm,keepaspectratio]{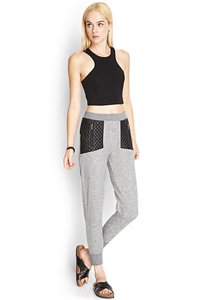}\hss\includegraphics[width=1.10cm,height=1.8cm,keepaspectratio]{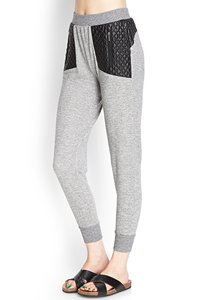}\hss\includegraphics[width=1.10cm,height=1.8cm,keepaspectratio]{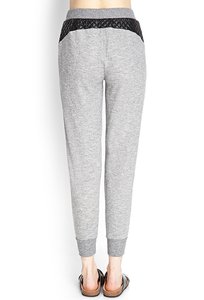}\hss\includegraphics[width=1.10cm,height=1.8cm,keepaspectratio]{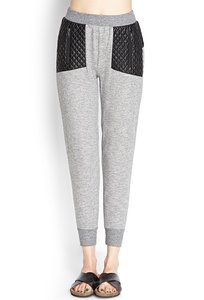}\hss\phantom{\rule{1.10cm}{1.8cm}}\hss}
      \end{minipage}
    }
  \end{minipage}
\par\vspace{2pt}
\noindent
  \begin{minipage}[t]{5.55cm}
    \fcolorbox{red!70!black}{white}{
      \begin{minipage}[t]{5.25cm}
        {\tiny\textbf{\#4}}\\
        \noindent\hbox to 5.25cm{\includegraphics[width=1.10cm,height=1.8cm,keepaspectratio]{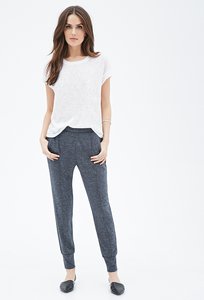}\hss\includegraphics[width=1.10cm,height=1.8cm,keepaspectratio]{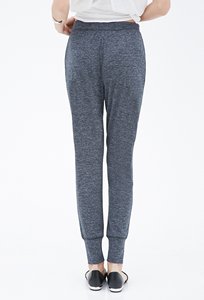}\hss\includegraphics[width=1.10cm,height=1.8cm,keepaspectratio]{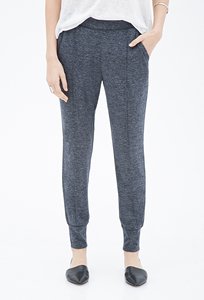}\hss\includegraphics[width=1.10cm,height=1.8cm,keepaspectratio]{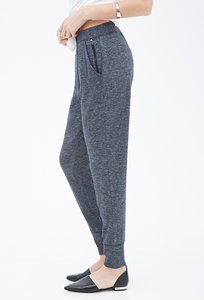}\hss\phantom{\rule{1.10cm}{1.8cm}}\hss}
      \end{minipage}
    }
  \end{minipage}
\hfill
  \begin{minipage}[t]{5.55cm}
    \fcolorbox{red!70!black}{white}{
      \begin{minipage}[t]{5.25cm}
        {\tiny\textbf{\#5}}\\
        \noindent\hbox to 5.25cm{\includegraphics[width=1.10cm,height=1.8cm,keepaspectratio]{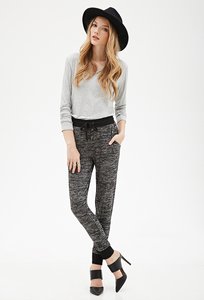}\hss\includegraphics[width=1.10cm,height=1.8cm,keepaspectratio]{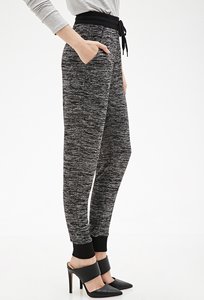}\hss\includegraphics[width=1.10cm,height=1.8cm,keepaspectratio]{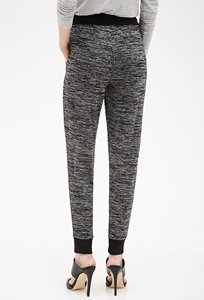}\hss\includegraphics[width=1.10cm,height=1.8cm,keepaspectratio]{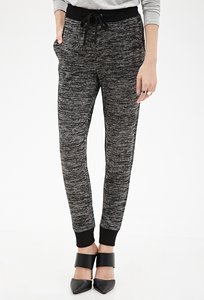}\hss\phantom{\rule{1.10cm}{1.8cm}}\hss}
      \end{minipage}
    }
  \end{minipage}
\hfill
  \begin{minipage}[t]{5.55cm}
    \fcolorbox{red!70!black}{white}{
      \begin{minipage}[t]{5.25cm}
        {\tiny\textbf{\#6}}\\
        \noindent\hbox to 5.25cm{\includegraphics[width=1.10cm,height=1.8cm,keepaspectratio]{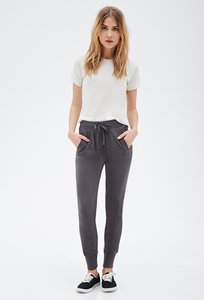}\hss\includegraphics[width=1.10cm,height=1.8cm,keepaspectratio]{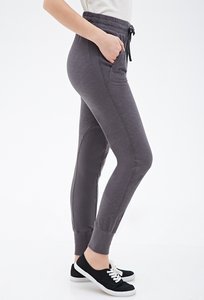}\hss\includegraphics[width=1.10cm,height=1.8cm,keepaspectratio]{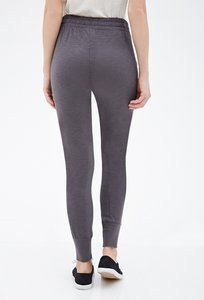}\hss\includegraphics[width=1.10cm,height=1.8cm,keepaspectratio]{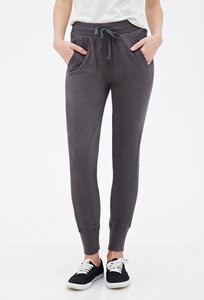}\hss\includegraphics[width=1.10cm,height=1.8cm,keepaspectratio]{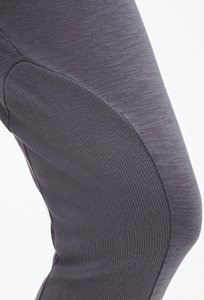}\hss}
      \end{minipage}
    }
  \end{minipage}
\par\vspace{2pt}
\noindent
  \begin{minipage}[t]{5.55cm}
    \fcolorbox{red!70!black}{white}{
      \begin{minipage}[t]{5.25cm}
        {\tiny\textbf{\#7}}\\
        \noindent\hbox to 5.25cm{\includegraphics[width=1.10cm,height=1.8cm,keepaspectratio]{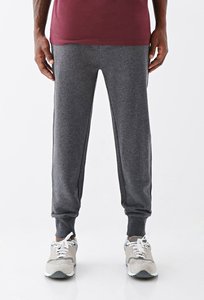}\hss\includegraphics[width=1.10cm,height=1.8cm,keepaspectratio]{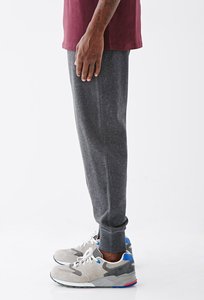}\hss\includegraphics[width=1.10cm,height=1.8cm,keepaspectratio]{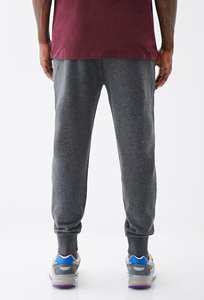}\hss\includegraphics[width=1.10cm,height=1.8cm,keepaspectratio]{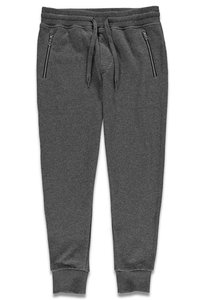}\hss\includegraphics[width=1.10cm,height=1.8cm,keepaspectratio]{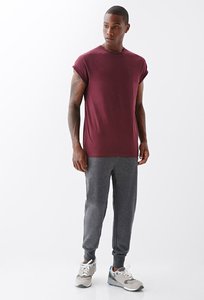}\hss}
      \end{minipage}
    }
  \end{minipage}
\hfill
  \begin{minipage}[t]{5.55cm}
    \fcolorbox{red!70!black}{white}{
      \begin{minipage}[t]{5.25cm}
        {\tiny\textbf{\#8}}\\
        \noindent\hbox to 5.25cm{\includegraphics[width=1.10cm,height=1.8cm,keepaspectratio]{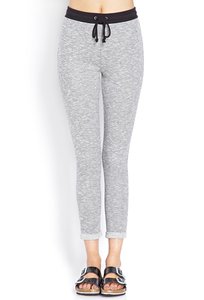}\hss\includegraphics[width=1.10cm,height=1.8cm,keepaspectratio]{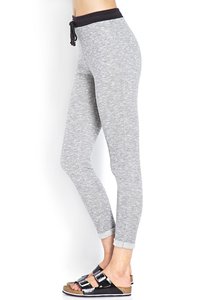}\hss\includegraphics[width=1.10cm,height=1.8cm,keepaspectratio]{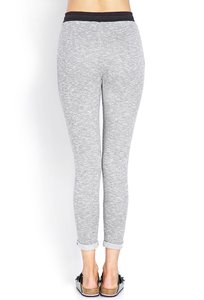}\hss\includegraphics[width=1.10cm,height=1.8cm,keepaspectratio]{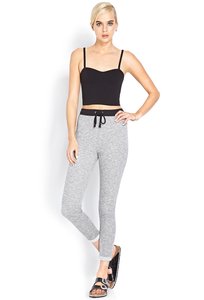}\hss\phantom{\rule{1.10cm}{1.8cm}}\hss}
      \end{minipage}
    }
  \end{minipage}
\hfill
  \begin{minipage}[t]{5.55cm}
    \fcolorbox{red!70!black}{white}{
      \begin{minipage}[t]{5.25cm}
        {\tiny\textbf{\#9}}\\
        \noindent\hbox to 5.25cm{\includegraphics[width=1.10cm,height=1.8cm,keepaspectratio]{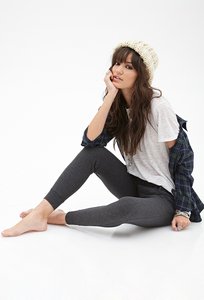}\hss\includegraphics[width=1.10cm,height=1.8cm,keepaspectratio]{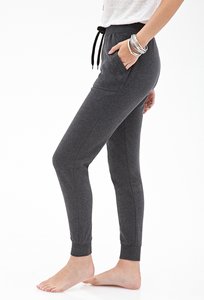}\hss\includegraphics[width=1.10cm,height=1.8cm,keepaspectratio]{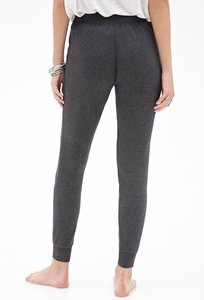}\hss\includegraphics[width=1.10cm,height=1.8cm,keepaspectratio]{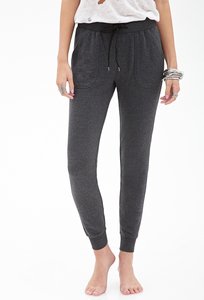}\hss\phantom{\rule{1.10cm}{1.8cm}}\hss}
      \end{minipage}
    }
  \end{minipage}
\par\vspace{2pt}
\noindent
  \begin{minipage}[t]{5.55cm}
    \fcolorbox{red!70!black}{white}{
      \begin{minipage}[t]{5.25cm}
        {\tiny\textbf{\#10}}\\
        \noindent\hbox to 5.25cm{\includegraphics[width=1.10cm,height=1.8cm,keepaspectratio]{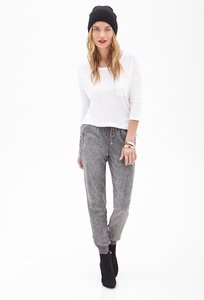}\hss\includegraphics[width=1.10cm,height=1.8cm,keepaspectratio]{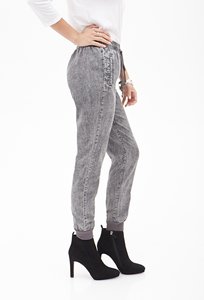}\hss\includegraphics[width=1.10cm,height=1.8cm,keepaspectratio]{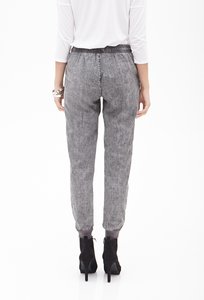}\hss\includegraphics[width=1.10cm,height=1.8cm,keepaspectratio]{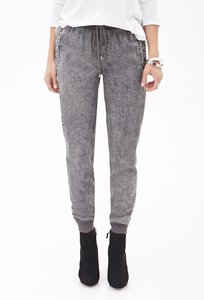}\hss\phantom{\rule{1.10cm}{1.8cm}}\hss}
      \end{minipage}
    }
  \end{minipage}
\hfill
  \begin{minipage}[t]{5.55cm}\end{minipage}
\hfill
  \begin{minipage}[t]{5.55cm}\end{minipage}
\par\vspace{2pt}
\noindent\rule{\linewidth}{0.4pt}
\par\vspace{20pt}

\par\vspace{16pt}
\noindent\textbf{\large Example 5}
\par\vspace{4pt}
\noindent\rule{\linewidth}{1.2pt}
\par\vspace{4pt}
% ── Case 5: short::deepfashion::5122 ──
\noindent\hfill%
  \begin{minipage}[t]{6.0cm}
    \noindent\hbox to 6.0cm{\includegraphics[width=1.20cm,height=2.0cm,keepaspectratio]{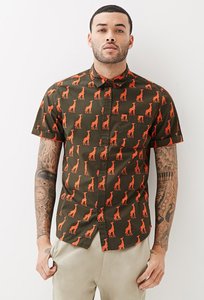}\hss\includegraphics[width=1.20cm,height=2.0cm,keepaspectratio]{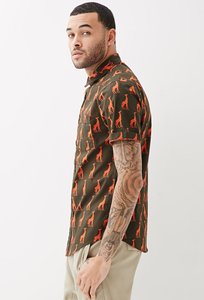}\hss\includegraphics[width=1.20cm,height=2.0cm,keepaspectratio]{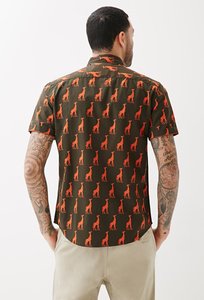}\hss\includegraphics[width=1.20cm,height=2.0cm,keepaspectratio]{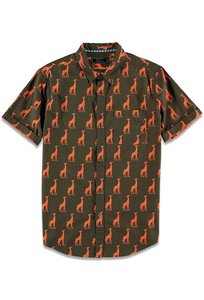}\hss\includegraphics[width=1.20cm,height=2.0cm,keepaspectratio]{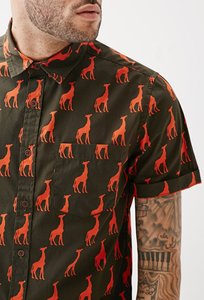}\hss}
    \par\vspace{1pt}
    {\scriptsize\textbf{}}
    \par\vspace{0pt}
    \parbox[t]{6.0cm}{\tiny\raggedright Men's short-sleeve button-up shirt featuring a bold all-over orange giraffe print on a dark olive green background. Designed with a classic point collar revealing a contrasting geometric lining, a single left chest pocket, and a full button front closure. The short sleeves have cuffed hems, and the curved shirt tail hem offers a relaxed silhouette. Regular fit, lightweight woven fabric.}
  \end{minipage}%
\hfill%
  \begin{minipage}[t]{3.5cm}
    \centering
    \vspace{0.45cm}%
    \parbox{3.5cm}{\centering\tiny Replace olive giraffe print with blue-based colorful abstract geometric pattern and update placket buttons from dark to white.}\\[2pt]
    {\normalsize$\longrightarrow$}
  \end{minipage}%
\hfill%
  \begin{minipage}[t]{6.0cm}
    \noindent\hbox to 6.0cm{\includegraphics[width=1.20cm,height=2.0cm,keepaspectratio]{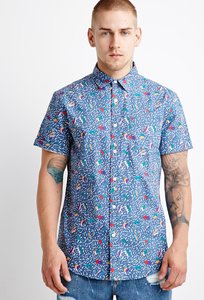}\hss\includegraphics[width=1.20cm,height=2.0cm,keepaspectratio]{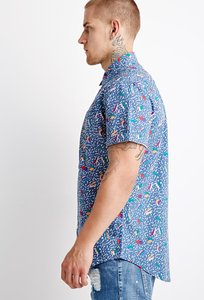}\hss\includegraphics[width=1.20cm,height=2.0cm,keepaspectratio]{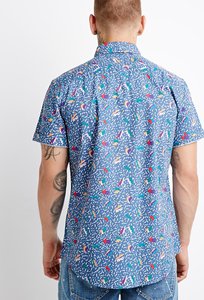}\hss\includegraphics[width=1.20cm,height=2.0cm,keepaspectratio]{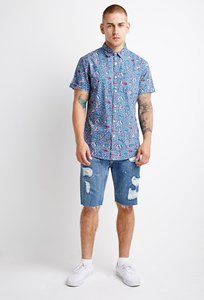}\hss\includegraphics[width=1.20cm,height=2.0cm,keepaspectratio]{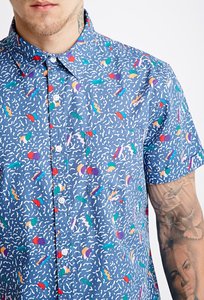}\hss}
    \par\vspace{1pt}
    {\scriptsize\textbf{Ground Truth}}
    \par\vspace{0pt}
    \parbox[t]{6.0cm}{\tiny\raggedright Men's short-sleeve button-up shirt featuring a vibrant blue base with retro 90s-inspired abstract print of white squiggles and colorful geometric shapes. Classic point collar, full button front with white buttons, single left chest pocket, and regular fit. Lightweight woven fabric, casual summer style.}
  \end{minipage}%
\hfill
\par\vspace{4pt}
\par\vspace{4pt}
\noindent\rule{\linewidth}{0.4pt}
% -- mt_align --
{\small\textbf{\textbf{Ours}}}\quad{\small Rank~1}\\
\noindent
  \begin{minipage}[t]{5.55cm}
    \fcolorbox{green!60!black}{white}{
      \begin{minipage}[t]{5.25cm}
        {\tiny\textbf{\#1}}\\
        \noindent\hbox to 5.25cm{\includegraphics[width=1.10cm,height=1.8cm,keepaspectratio]{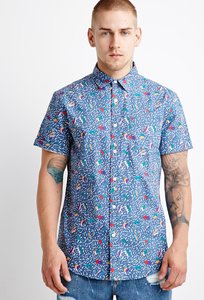}\hss\includegraphics[width=1.10cm,height=1.8cm,keepaspectratio]{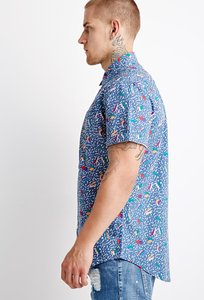}\hss\includegraphics[width=1.10cm,height=1.8cm,keepaspectratio]{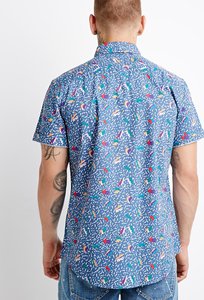}\hss\includegraphics[width=1.10cm,height=1.8cm,keepaspectratio]{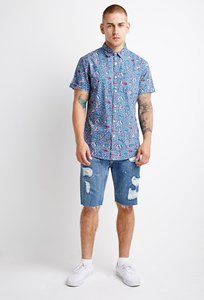}\hss\includegraphics[width=1.10cm,height=1.8cm,keepaspectratio]{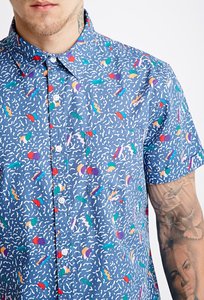}\hss}
      \end{minipage}
    }
  \end{minipage}
\hfill
  \begin{minipage}[t]{5.55cm}
    \fcolorbox{red!70!black}{white}{
      \begin{minipage}[t]{5.25cm}
        {\tiny\textbf{\#2}}\\
        \noindent\hbox to 5.25cm{\includegraphics[width=1.10cm,height=1.8cm,keepaspectratio]{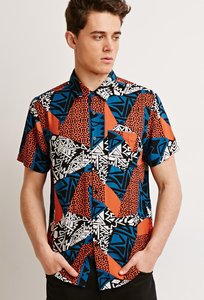}\hss\includegraphics[width=1.10cm,height=1.8cm,keepaspectratio]{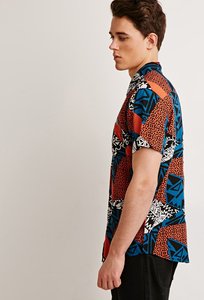}\hss\includegraphics[width=1.10cm,height=1.8cm,keepaspectratio]{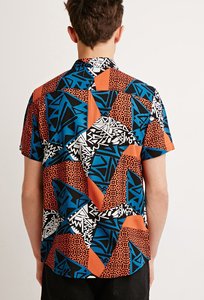}\hss\includegraphics[width=1.10cm,height=1.8cm,keepaspectratio]{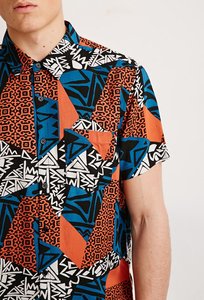}\hss\phantom{\rule{1.10cm}{1.8cm}}\hss}
      \end{minipage}
    }
  \end{minipage}
\hfill
  \begin{minipage}[t]{5.55cm}
    \fcolorbox{red!70!black}{white}{
      \begin{minipage}[t]{5.25cm}
        {\tiny\textbf{\#3}}\\
        \noindent\hbox to 5.25cm{\includegraphics[width=1.10cm,height=1.8cm,keepaspectratio]{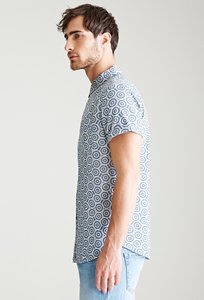}\hss\includegraphics[width=1.10cm,height=1.8cm,keepaspectratio]{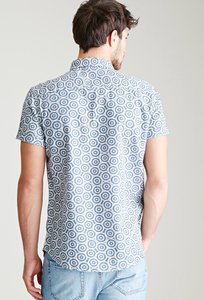}\hss\includegraphics[width=1.10cm,height=1.8cm,keepaspectratio]{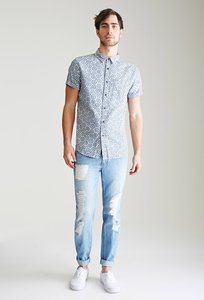}\hss\includegraphics[width=1.10cm,height=1.8cm,keepaspectratio]{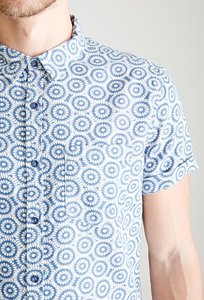}\hss\phantom{\rule{1.10cm}{1.8cm}}\hss}
      \end{minipage}
    }
  \end{minipage}
\par\vspace{2pt}
\noindent
  \begin{minipage}[t]{5.55cm}
    \fcolorbox{red!70!black}{white}{
      \begin{minipage}[t]{5.25cm}
        {\tiny\textbf{\#4}}\\
        \noindent\hbox to 5.25cm{\includegraphics[width=1.10cm,height=1.8cm,keepaspectratio]{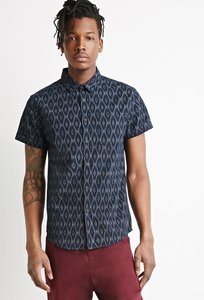}\hss\includegraphics[width=1.10cm,height=1.8cm,keepaspectratio]{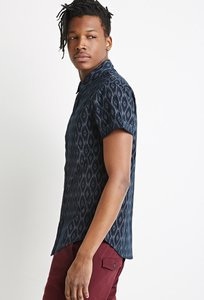}\hss\includegraphics[width=1.10cm,height=1.8cm,keepaspectratio]{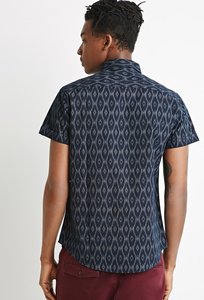}\hss\includegraphics[width=1.10cm,height=1.8cm,keepaspectratio]{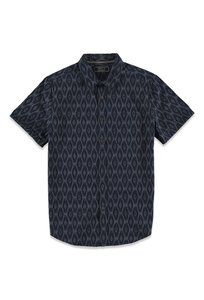}\hss\includegraphics[width=1.10cm,height=1.8cm,keepaspectratio]{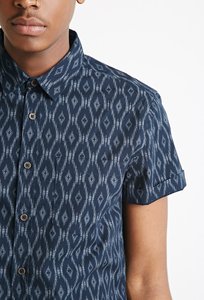}\hss}
      \end{minipage}
    }
  \end{minipage}
\hfill
  \begin{minipage}[t]{5.55cm}
    \fcolorbox{red!70!black}{white}{
      \begin{minipage}[t]{5.25cm}
        {\tiny\textbf{\#5}}\\
        \noindent\hbox to 5.25cm{\includegraphics[width=1.10cm,height=1.8cm,keepaspectratio]{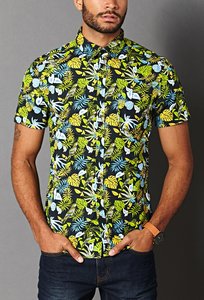}\hss\includegraphics[width=1.10cm,height=1.8cm,keepaspectratio]{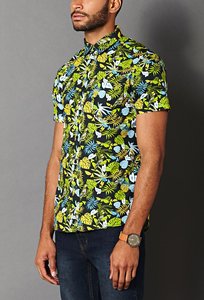}\hss\includegraphics[width=1.10cm,height=1.8cm,keepaspectratio]{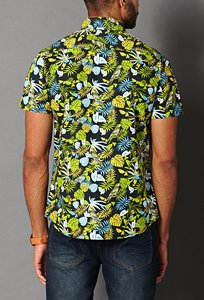}\hss\includegraphics[width=1.10cm,height=1.8cm,keepaspectratio]{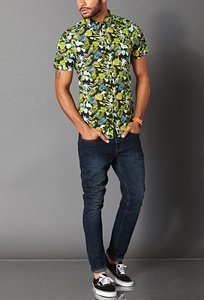}\hss\includegraphics[width=1.10cm,height=1.8cm,keepaspectratio]{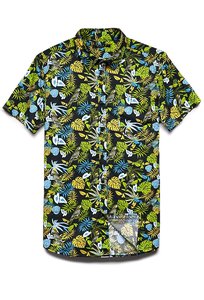}\hss}
      \end{minipage}
    }
  \end{minipage}
\hfill
  \begin{minipage}[t]{5.55cm}
    \fcolorbox{red!70!black}{white}{
      \begin{minipage}[t]{5.25cm}
        {\tiny\textbf{\#6}}\\
        \noindent\hbox to 5.25cm{\includegraphics[width=1.10cm,height=1.8cm,keepaspectratio]{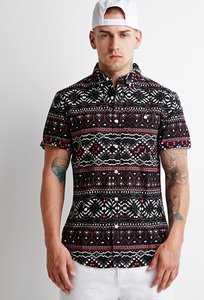}\hss\includegraphics[width=1.10cm,height=1.8cm,keepaspectratio]{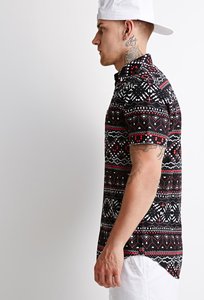}\hss\includegraphics[width=1.10cm,height=1.8cm,keepaspectratio]{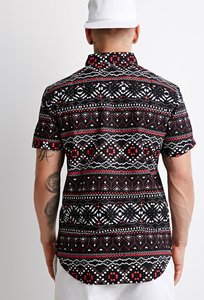}\hss\includegraphics[width=1.10cm,height=1.8cm,keepaspectratio]{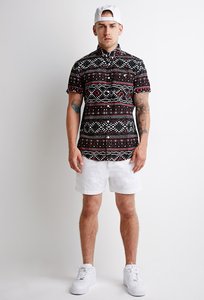}\hss\includegraphics[width=1.10cm,height=1.8cm,keepaspectratio]{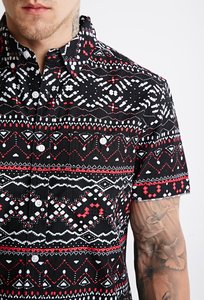}\hss}
      \end{minipage}
    }
  \end{minipage}
\par\vspace{2pt}
\noindent
  \begin{minipage}[t]{5.55cm}
    \fcolorbox{red!70!black}{white}{
      \begin{minipage}[t]{5.25cm}
        {\tiny\textbf{\#7}}\\
        \noindent\hbox to 5.25cm{\includegraphics[width=1.10cm,height=1.8cm,keepaspectratio]{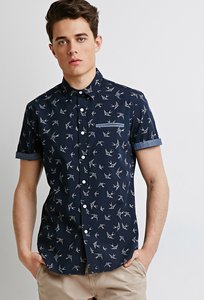}\hss\includegraphics[width=1.10cm,height=1.8cm,keepaspectratio]{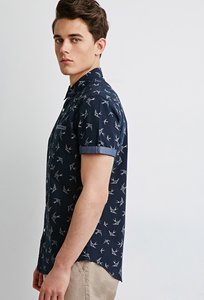}\hss\includegraphics[width=1.10cm,height=1.8cm,keepaspectratio]{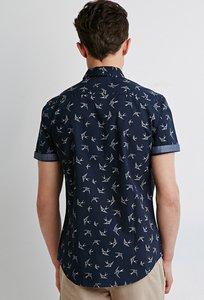}\hss\includegraphics[width=1.10cm,height=1.8cm,keepaspectratio]{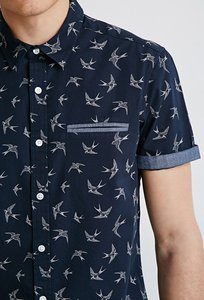}\hss\phantom{\rule{1.10cm}{1.8cm}}\hss}
      \end{minipage}
    }
  \end{minipage}
\hfill
  \begin{minipage}[t]{5.55cm}
    \fcolorbox{red!70!black}{white}{
      \begin{minipage}[t]{5.25cm}
        {\tiny\textbf{\#8}}\\
        \noindent\hbox to 5.25cm{\includegraphics[width=1.10cm,height=1.8cm,keepaspectratio]{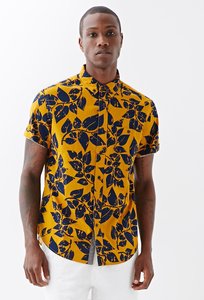}\hss\includegraphics[width=1.10cm,height=1.8cm,keepaspectratio]{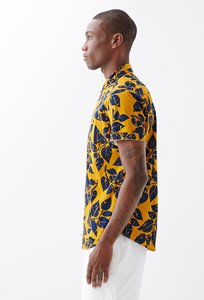}\hss\includegraphics[width=1.10cm,height=1.8cm,keepaspectratio]{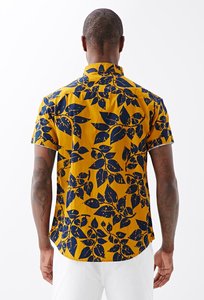}\hss\includegraphics[width=1.10cm,height=1.8cm,keepaspectratio]{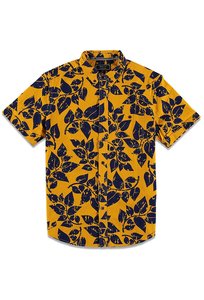}\hss\includegraphics[width=1.10cm,height=1.8cm,keepaspectratio]{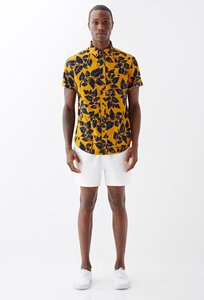}\hss}
      \end{minipage}
    }
  \end{minipage}
\hfill
  \begin{minipage}[t]{5.55cm}
    \fcolorbox{red!70!black}{white}{
      \begin{minipage}[t]{5.25cm}
        {\tiny\textbf{\#9}}\\
        \noindent\hbox to 5.25cm{\includegraphics[width=1.10cm,height=1.8cm,keepaspectratio]{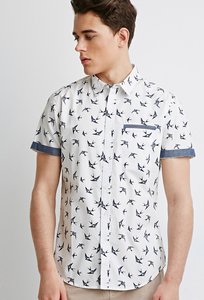}\hss\includegraphics[width=1.10cm,height=1.8cm,keepaspectratio]{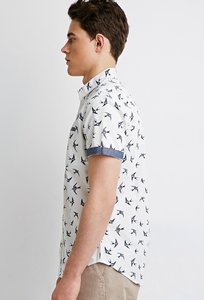}\hss\includegraphics[width=1.10cm,height=1.8cm,keepaspectratio]{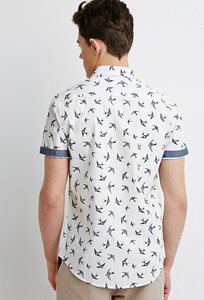}\hss\includegraphics[width=1.10cm,height=1.8cm,keepaspectratio]{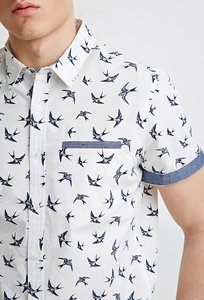}\hss\phantom{\rule{1.10cm}{1.8cm}}\hss}
      \end{minipage}
    }
  \end{minipage}
\par\vspace{2pt}
\noindent
  \begin{minipage}[t]{5.55cm}
    \fcolorbox{red!70!black}{white}{
      \begin{minipage}[t]{5.25cm}
        {\tiny\textbf{\#10}}\\
        \noindent\hbox to 5.25cm{\includegraphics[width=1.10cm,height=1.8cm,keepaspectratio]{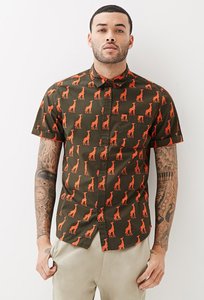}\hss\includegraphics[width=1.10cm,height=1.8cm,keepaspectratio]{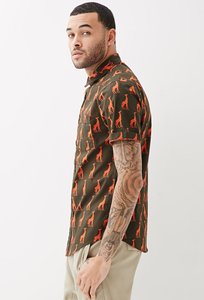}\hss\includegraphics[width=1.10cm,height=1.8cm,keepaspectratio]{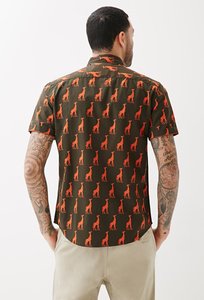}\hss\includegraphics[width=1.10cm,height=1.8cm,keepaspectratio]{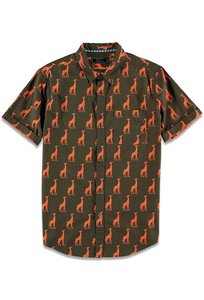}\hss\includegraphics[width=1.10cm,height=1.8cm,keepaspectratio]{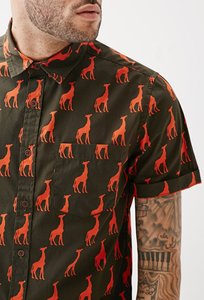}\hss}
      \end{minipage}
    }
  \end{minipage}
\hfill
  \begin{minipage}[t]{5.55cm}\end{minipage}
\hfill
  \begin{minipage}[t]{5.55cm}\end{minipage}
\par\vspace{2pt}
\noindent\rule{\linewidth}{0.4pt}
\par\vspace{2pt}
% -- qwen3_vl_2b --
{\small\textbf{Qwen3-VL-2B}}\quad{\small Rank~2}\\
\noindent
  \begin{minipage}[t]{5.55cm}
    \fcolorbox{red!70!black}{white}{
      \begin{minipage}[t]{5.25cm}
        {\tiny\textbf{\#1}}\\
        \noindent\hbox to 5.25cm{\includegraphics[width=1.10cm,height=1.8cm,keepaspectratio]{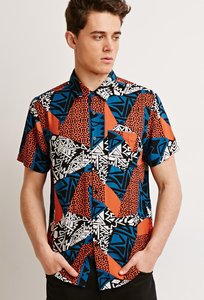}\hss\includegraphics[width=1.10cm,height=1.8cm,keepaspectratio]{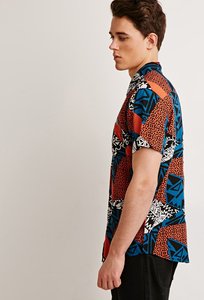}\hss\includegraphics[width=1.10cm,height=1.8cm,keepaspectratio]{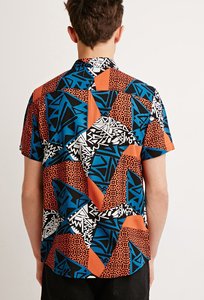}\hss\includegraphics[width=1.10cm,height=1.8cm,keepaspectratio]{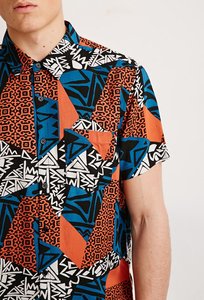}\hss\phantom{\rule{1.10cm}{1.8cm}}\hss}
      \end{minipage}
    }
  \end{minipage}
\hfill
  \begin{minipage}[t]{5.55cm}
    \fcolorbox{green!60!black}{white}{
      \begin{minipage}[t]{5.25cm}
        {\tiny\textbf{\#2}}\\
        \noindent\hbox to 5.25cm{\includegraphics[width=1.10cm,height=1.8cm,keepaspectratio]{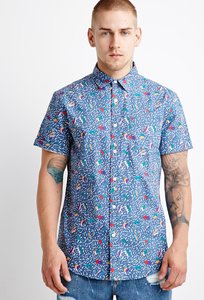}\hss\includegraphics[width=1.10cm,height=1.8cm,keepaspectratio]{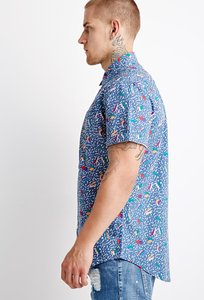}\hss\includegraphics[width=1.10cm,height=1.8cm,keepaspectratio]{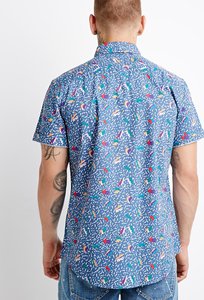}\hss\includegraphics[width=1.10cm,height=1.8cm,keepaspectratio]{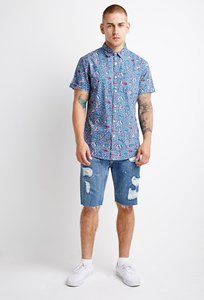}\hss\includegraphics[width=1.10cm,height=1.8cm,keepaspectratio]{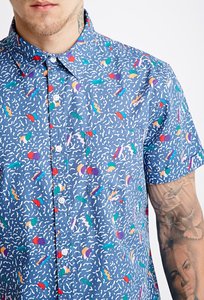}\hss}
      \end{minipage}
    }
  \end{minipage}
\hfill
  \begin{minipage}[t]{5.55cm}
    \fcolorbox{red!70!black}{white}{
      \begin{minipage}[t]{5.25cm}
        {\tiny\textbf{\#3}}\\
        \noindent\hbox to 5.25cm{\includegraphics[width=1.10cm,height=1.8cm,keepaspectratio]{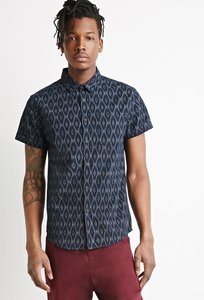}\hss\includegraphics[width=1.10cm,height=1.8cm,keepaspectratio]{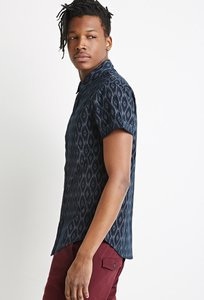}\hss\includegraphics[width=1.10cm,height=1.8cm,keepaspectratio]{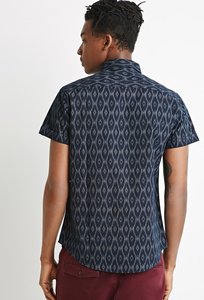}\hss\includegraphics[width=1.10cm,height=1.8cm,keepaspectratio]{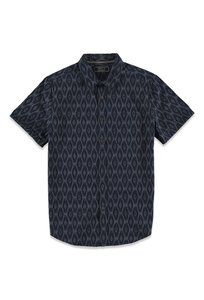}\hss\includegraphics[width=1.10cm,height=1.8cm,keepaspectratio]{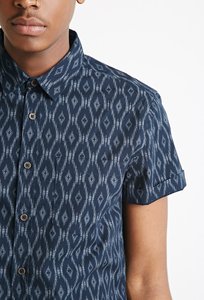}\hss}
      \end{minipage}
    }
  \end{minipage}
\par\vspace{2pt}
\noindent
  \begin{minipage}[t]{5.55cm}
    \fcolorbox{red!70!black}{white}{
      \begin{minipage}[t]{5.25cm}
        {\tiny\textbf{\#4}}\\
        \noindent\hbox to 5.25cm{\includegraphics[width=1.10cm,height=1.8cm,keepaspectratio]{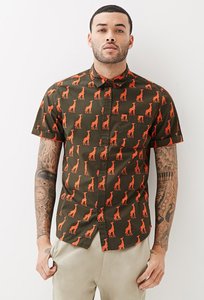}\hss\includegraphics[width=1.10cm,height=1.8cm,keepaspectratio]{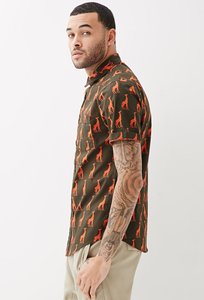}\hss\includegraphics[width=1.10cm,height=1.8cm,keepaspectratio]{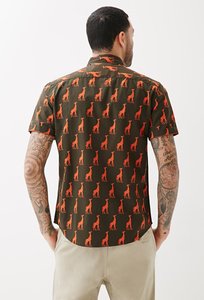}\hss\includegraphics[width=1.10cm,height=1.8cm,keepaspectratio]{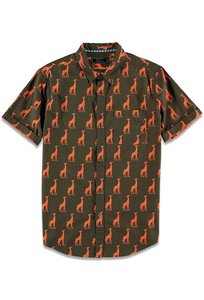}\hss\includegraphics[width=1.10cm,height=1.8cm,keepaspectratio]{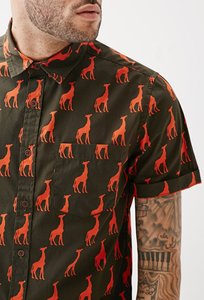}\hss}
      \end{minipage}
    }
  \end{minipage}
\hfill
  \begin{minipage}[t]{5.55cm}
    \fcolorbox{red!70!black}{white}{
      \begin{minipage}[t]{5.25cm}
        {\tiny\textbf{\#5}}\\
        \noindent\hbox to 5.25cm{\includegraphics[width=1.10cm,height=1.8cm,keepaspectratio]{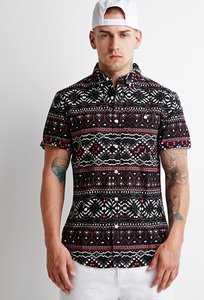}\hss\includegraphics[width=1.10cm,height=1.8cm,keepaspectratio]{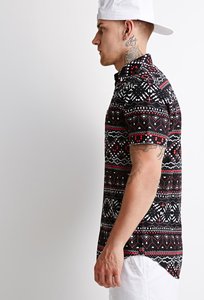}\hss\includegraphics[width=1.10cm,height=1.8cm,keepaspectratio]{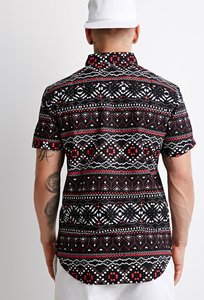}\hss\includegraphics[width=1.10cm,height=1.8cm,keepaspectratio]{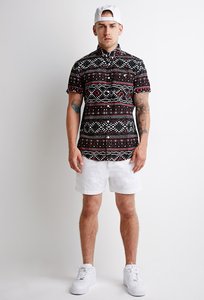}\hss\includegraphics[width=1.10cm,height=1.8cm,keepaspectratio]{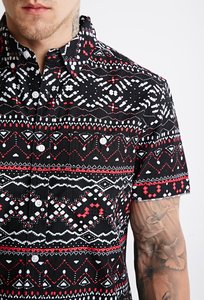}\hss}
      \end{minipage}
    }
  \end{minipage}
\hfill
  \begin{minipage}[t]{5.55cm}
    \fcolorbox{red!70!black}{white}{
      \begin{minipage}[t]{5.25cm}
        {\tiny\textbf{\#6}}\\
        \noindent\hbox to 5.25cm{\includegraphics[width=1.10cm,height=1.8cm,keepaspectratio]{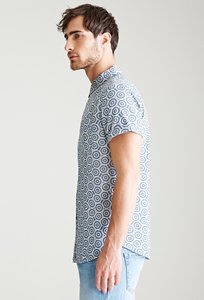}\hss\includegraphics[width=1.10cm,height=1.8cm,keepaspectratio]{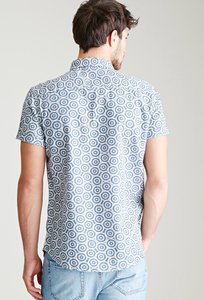}\hss\includegraphics[width=1.10cm,height=1.8cm,keepaspectratio]{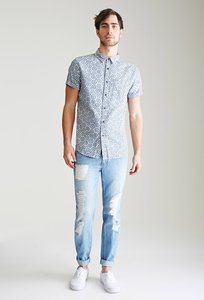}\hss\includegraphics[width=1.10cm,height=1.8cm,keepaspectratio]{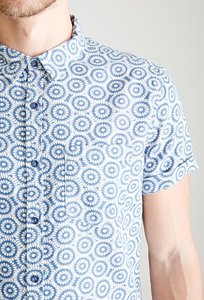}\hss\phantom{\rule{1.10cm}{1.8cm}}\hss}
      \end{minipage}
    }
  \end{minipage}
\par\vspace{2pt}
\noindent
  \begin{minipage}[t]{5.55cm}
    \fcolorbox{red!70!black}{white}{
      \begin{minipage}[t]{5.25cm}
        {\tiny\textbf{\#7}}\\
        \noindent\hbox to 5.25cm{\includegraphics[width=1.10cm,height=1.8cm,keepaspectratio]{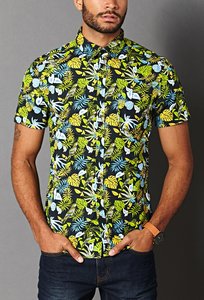}\hss\includegraphics[width=1.10cm,height=1.8cm,keepaspectratio]{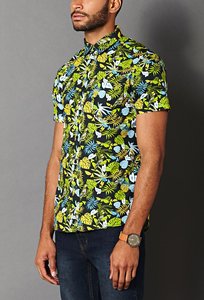}\hss\includegraphics[width=1.10cm,height=1.8cm,keepaspectratio]{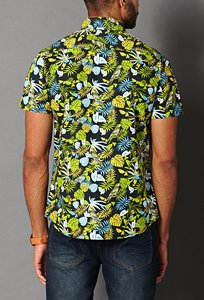}\hss\includegraphics[width=1.10cm,height=1.8cm,keepaspectratio]{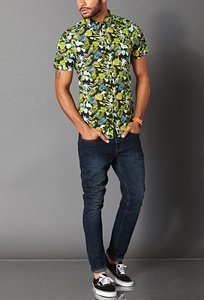}\hss\includegraphics[width=1.10cm,height=1.8cm,keepaspectratio]{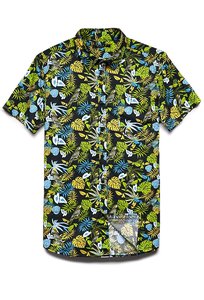}\hss}
      \end{minipage}
    }
  \end{minipage}
\hfill
  \begin{minipage}[t]{5.55cm}
    \fcolorbox{red!70!black}{white}{
      \begin{minipage}[t]{5.25cm}
        {\tiny\textbf{\#8}}\\
        \noindent\hbox to 5.25cm{\includegraphics[width=1.10cm,height=1.8cm,keepaspectratio]{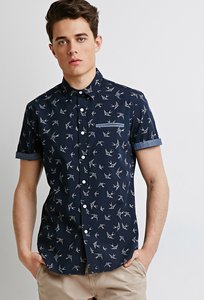}\hss\includegraphics[width=1.10cm,height=1.8cm,keepaspectratio]{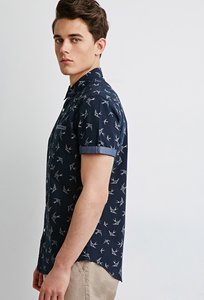}\hss\includegraphics[width=1.10cm,height=1.8cm,keepaspectratio]{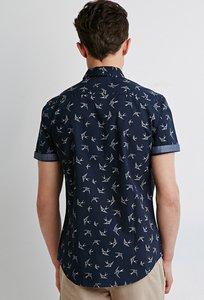}\hss\includegraphics[width=1.10cm,height=1.8cm,keepaspectratio]{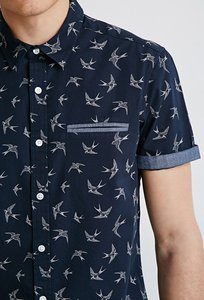}\hss\phantom{\rule{1.10cm}{1.8cm}}\hss}
      \end{minipage}
    }
  \end{minipage}
\hfill
  \begin{minipage}[t]{5.55cm}
    \fcolorbox{red!70!black}{white}{
      \begin{minipage}[t]{5.25cm}
        {\tiny\textbf{\#9}}\\
        \noindent\hbox to 5.25cm{\includegraphics[width=1.10cm,height=1.8cm,keepaspectratio]{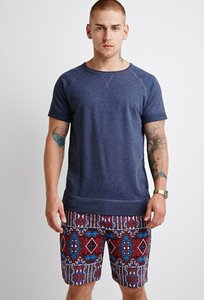}\hss\includegraphics[width=1.10cm,height=1.8cm,keepaspectratio]{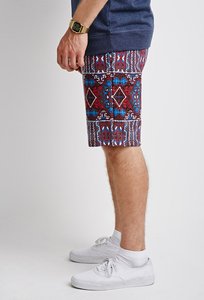}\hss\includegraphics[width=1.10cm,height=1.8cm,keepaspectratio]{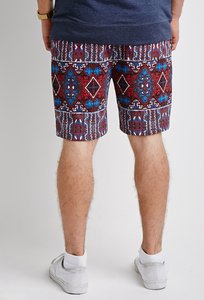}\hss\includegraphics[width=1.10cm,height=1.8cm,keepaspectratio]{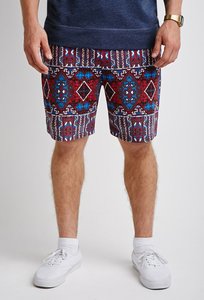}\hss\phantom{\rule{1.10cm}{1.8cm}}\hss}
      \end{minipage}
    }
  \end{minipage}
\par\vspace{2pt}
\noindent
  \begin{minipage}[t]{5.55cm}
    \fcolorbox{red!70!black}{white}{
      \begin{minipage}[t]{5.25cm}
        {\tiny\textbf{\#10}}\\
        \noindent\hbox to 5.25cm{\includegraphics[width=1.10cm,height=1.8cm,keepaspectratio]{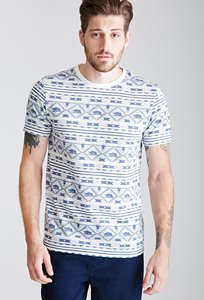}\hss\includegraphics[width=1.10cm,height=1.8cm,keepaspectratio]{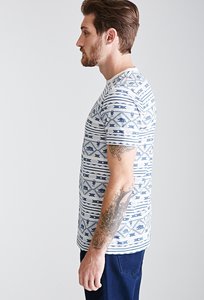}\hss\includegraphics[width=1.10cm,height=1.8cm,keepaspectratio]{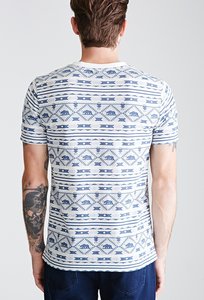}\hss\includegraphics[width=1.10cm,height=1.8cm,keepaspectratio]{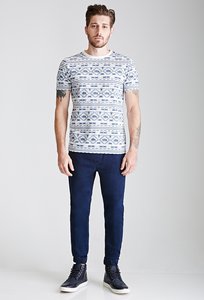}\hss\includegraphics[width=1.10cm,height=1.8cm,keepaspectratio]{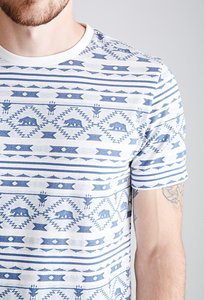}\hss}
      \end{minipage}
    }
  \end{minipage}
\hfill
  \begin{minipage}[t]{5.55cm}\end{minipage}
\hfill
  \begin{minipage}[t]{5.55cm}\end{minipage}
\par\vspace{2pt}
\noindent\rule{\linewidth}{0.4pt}
\par\vspace{2pt}
% -- qwen3_vl_8b --
{\small\textbf{Qwen3-VL-8B}}\quad{\small Rank~2}\\
\noindent
  \begin{minipage}[t]{5.55cm}
    \fcolorbox{red!70!black}{white}{
      \begin{minipage}[t]{5.25cm}
        {\tiny\textbf{\#1}}\\
        \noindent\hbox to 5.25cm{\includegraphics[width=1.10cm,height=1.8cm,keepaspectratio]{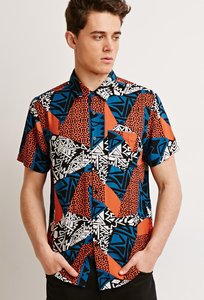}\hss\includegraphics[width=1.10cm,height=1.8cm,keepaspectratio]{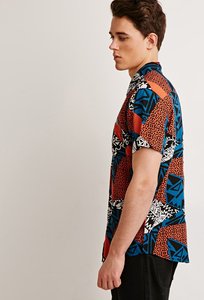}\hss\includegraphics[width=1.10cm,height=1.8cm,keepaspectratio]{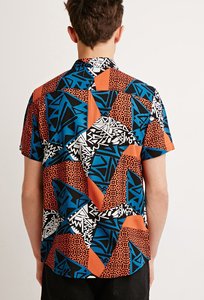}\hss\includegraphics[width=1.10cm,height=1.8cm,keepaspectratio]{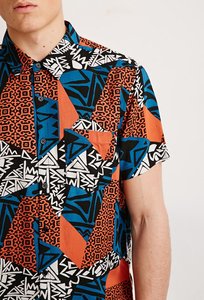}\hss\phantom{\rule{1.10cm}{1.8cm}}\hss}
      \end{minipage}
    }
  \end{minipage}
\hfill
  \begin{minipage}[t]{5.55cm}
    \fcolorbox{green!60!black}{white}{
      \begin{minipage}[t]{5.25cm}
        {\tiny\textbf{\#2}}\\
        \noindent\hbox to 5.25cm{\includegraphics[width=1.10cm,height=1.8cm,keepaspectratio]{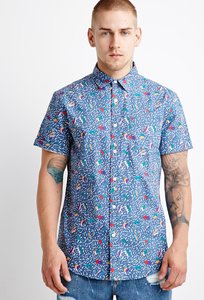}\hss\includegraphics[width=1.10cm,height=1.8cm,keepaspectratio]{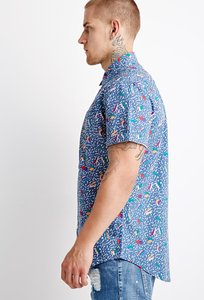}\hss\includegraphics[width=1.10cm,height=1.8cm,keepaspectratio]{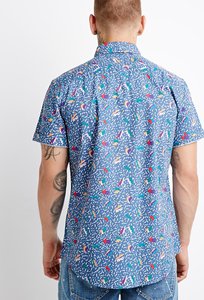}\hss\includegraphics[width=1.10cm,height=1.8cm,keepaspectratio]{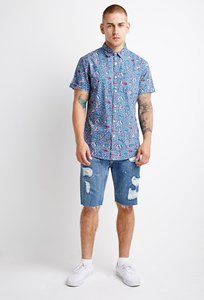}\hss\includegraphics[width=1.10cm,height=1.8cm,keepaspectratio]{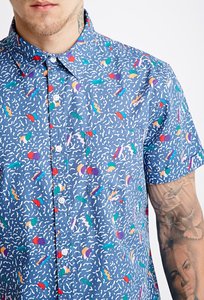}\hss}
      \end{minipage}
    }
  \end{minipage}
\hfill
  \begin{minipage}[t]{5.55cm}
    \fcolorbox{red!70!black}{white}{
      \begin{minipage}[t]{5.25cm}
        {\tiny\textbf{\#3}}\\
        \noindent\hbox to 5.25cm{\includegraphics[width=1.10cm,height=1.8cm,keepaspectratio]{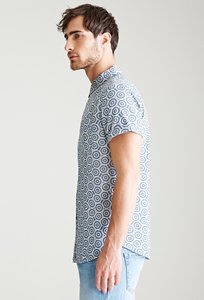}\hss\includegraphics[width=1.10cm,height=1.8cm,keepaspectratio]{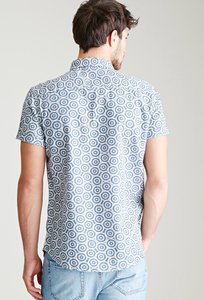}\hss\includegraphics[width=1.10cm,height=1.8cm,keepaspectratio]{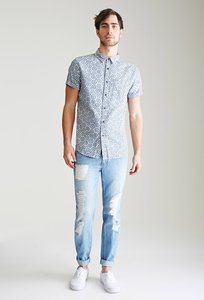}\hss\includegraphics[width=1.10cm,height=1.8cm,keepaspectratio]{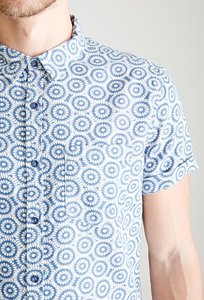}\hss\phantom{\rule{1.10cm}{1.8cm}}\hss}
      \end{minipage}
    }
  \end{minipage}
\par\vspace{2pt}
\noindent
  \begin{minipage}[t]{5.55cm}
    \fcolorbox{red!70!black}{white}{
      \begin{minipage}[t]{5.25cm}
        {\tiny\textbf{\#4}}\\
        \noindent\hbox to 5.25cm{\includegraphics[width=1.10cm,height=1.8cm,keepaspectratio]{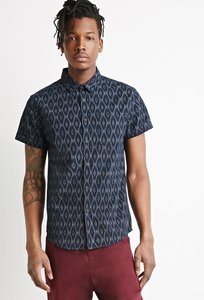}\hss\includegraphics[width=1.10cm,height=1.8cm,keepaspectratio]{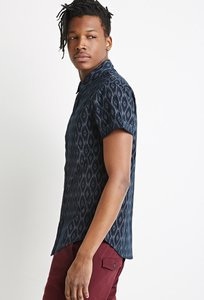}\hss\includegraphics[width=1.10cm,height=1.8cm,keepaspectratio]{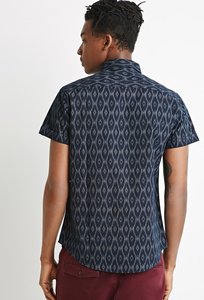}\hss\includegraphics[width=1.10cm,height=1.8cm,keepaspectratio]{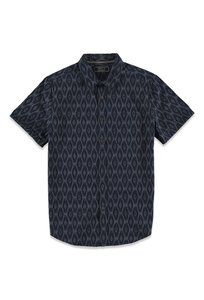}\hss\includegraphics[width=1.10cm,height=1.8cm,keepaspectratio]{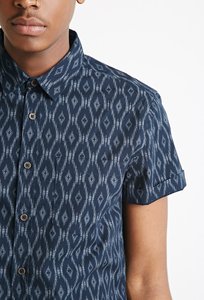}\hss}
      \end{minipage}
    }
  \end{minipage}
\hfill
  \begin{minipage}[t]{5.55cm}
    \fcolorbox{red!70!black}{white}{
      \begin{minipage}[t]{5.25cm}
        {\tiny\textbf{\#5}}\\
        \noindent\hbox to 5.25cm{\includegraphics[width=1.10cm,height=1.8cm,keepaspectratio]{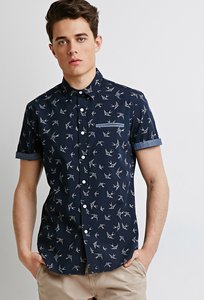}\hss\includegraphics[width=1.10cm,height=1.8cm,keepaspectratio]{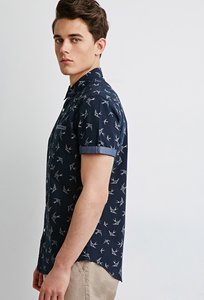}\hss\includegraphics[width=1.10cm,height=1.8cm,keepaspectratio]{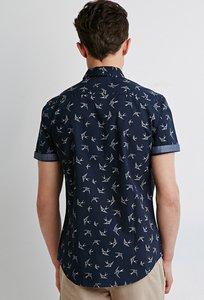}\hss\includegraphics[width=1.10cm,height=1.8cm,keepaspectratio]{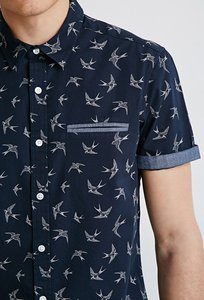}\hss\phantom{\rule{1.10cm}{1.8cm}}\hss}
      \end{minipage}
    }
  \end{minipage}
\hfill
  \begin{minipage}[t]{5.55cm}
    \fcolorbox{red!70!black}{white}{
      \begin{minipage}[t]{5.25cm}
        {\tiny\textbf{\#6}}\\
        \noindent\hbox to 5.25cm{\includegraphics[width=1.10cm,height=1.8cm,keepaspectratio]{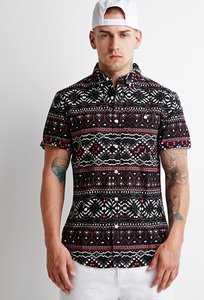}\hss\includegraphics[width=1.10cm,height=1.8cm,keepaspectratio]{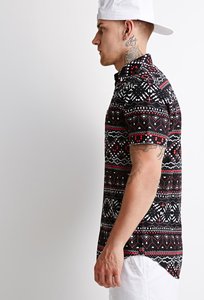}\hss\includegraphics[width=1.10cm,height=1.8cm,keepaspectratio]{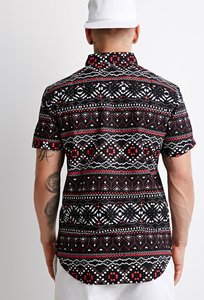}\hss\includegraphics[width=1.10cm,height=1.8cm,keepaspectratio]{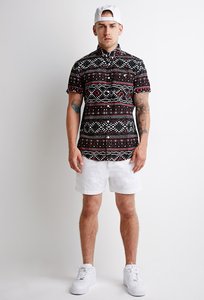}\hss\includegraphics[width=1.10cm,height=1.8cm,keepaspectratio]{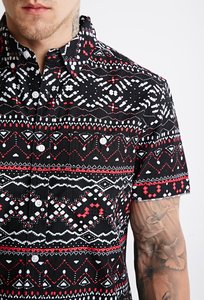}\hss}
      \end{minipage}
    }
  \end{minipage}
\par\vspace{2pt}
\noindent
  \begin{minipage}[t]{5.55cm}
    \fcolorbox{red!70!black}{white}{
      \begin{minipage}[t]{5.25cm}
        {\tiny\textbf{\#7}}\\
        \noindent\hbox to 5.25cm{\includegraphics[width=1.10cm,height=1.8cm,keepaspectratio]{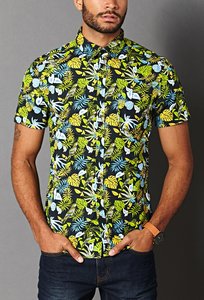}\hss\includegraphics[width=1.10cm,height=1.8cm,keepaspectratio]{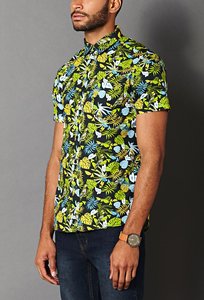}\hss\includegraphics[width=1.10cm,height=1.8cm,keepaspectratio]{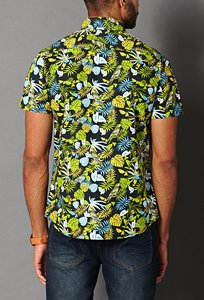}\hss\includegraphics[width=1.10cm,height=1.8cm,keepaspectratio]{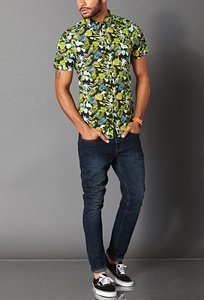}\hss\includegraphics[width=1.10cm,height=1.8cm,keepaspectratio]{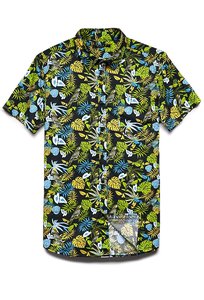}\hss}
      \end{minipage}
    }
  \end{minipage}
\hfill
  \begin{minipage}[t]{5.55cm}
    \fcolorbox{red!70!black}{white}{
      \begin{minipage}[t]{5.25cm}
        {\tiny\textbf{\#8}}\\
        \noindent\hbox to 5.25cm{\includegraphics[width=1.10cm,height=1.8cm,keepaspectratio]{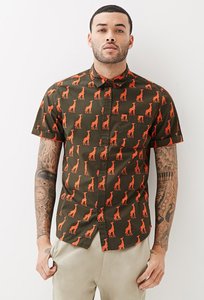}\hss\includegraphics[width=1.10cm,height=1.8cm,keepaspectratio]{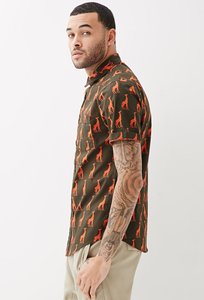}\hss\includegraphics[width=1.10cm,height=1.8cm,keepaspectratio]{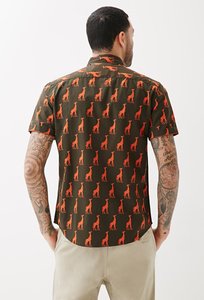}\hss\includegraphics[width=1.10cm,height=1.8cm,keepaspectratio]{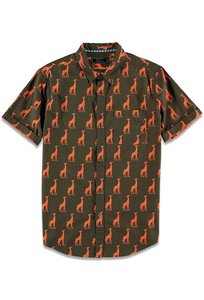}\hss\includegraphics[width=1.10cm,height=1.8cm,keepaspectratio]{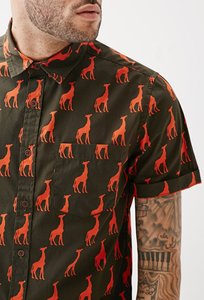}\hss}
      \end{minipage}
    }
  \end{minipage}
\hfill
  \begin{minipage}[t]{5.55cm}
    \fcolorbox{red!70!black}{white}{
      \begin{minipage}[t]{5.25cm}
        {\tiny\textbf{\#9}}\\
        \noindent\hbox to 5.25cm{\includegraphics[width=1.10cm,height=1.8cm,keepaspectratio]{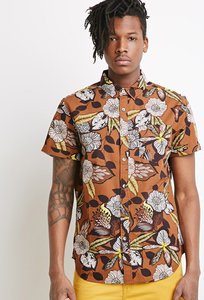}\hss\includegraphics[width=1.10cm,height=1.8cm,keepaspectratio]{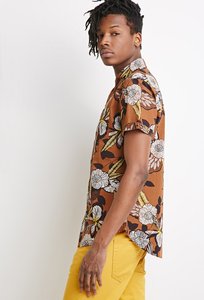}\hss\includegraphics[width=1.10cm,height=1.8cm,keepaspectratio]{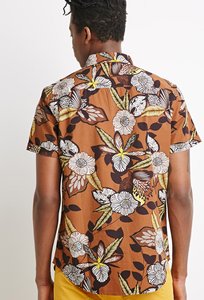}\hss\includegraphics[width=1.10cm,height=1.8cm,keepaspectratio]{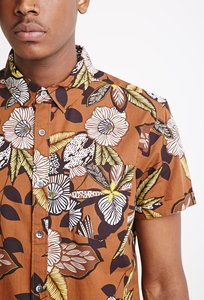}\hss\phantom{\rule{1.10cm}{1.8cm}}\hss}
      \end{minipage}
    }
  \end{minipage}
\par\vspace{2pt}
\noindent
  \begin{minipage}[t]{5.55cm}
    \fcolorbox{red!70!black}{white}{
      \begin{minipage}[t]{5.25cm}
        {\tiny\textbf{\#10}}\\
        \noindent\hbox to 5.25cm{\includegraphics[width=1.10cm,height=1.8cm,keepaspectratio]{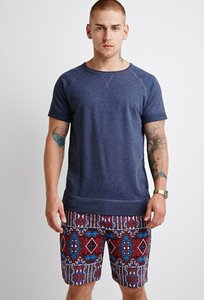}\hss\includegraphics[width=1.10cm,height=1.8cm,keepaspectratio]{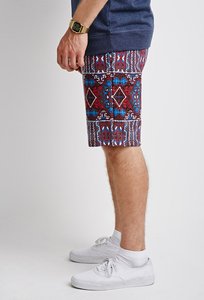}\hss\includegraphics[width=1.10cm,height=1.8cm,keepaspectratio]{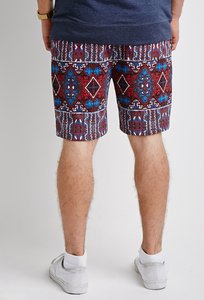}\hss\includegraphics[width=1.10cm,height=1.8cm,keepaspectratio]{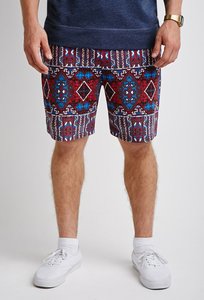}\hss\phantom{\rule{1.10cm}{1.8cm}}\hss}
      \end{minipage}
    }
  \end{minipage}
\hfill
  \begin{minipage}[t]{5.55cm}\end{minipage}
\hfill
  \begin{minipage}[t]{5.55cm}\end{minipage}
\par\vspace{2pt}
\noindent\rule{\linewidth}{0.4pt}
\par\vspace{2pt}
% -- reznembed --
{\small\textbf{RezNEmbed}}\quad{\small Rank~2}\\
\noindent
  \begin{minipage}[t]{5.55cm}
    \fcolorbox{red!70!black}{white}{
      \begin{minipage}[t]{5.25cm}
        {\tiny\textbf{\#1}}\\
        \noindent\hbox to 5.25cm{\includegraphics[width=1.10cm,height=1.8cm,keepaspectratio]{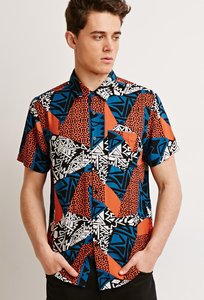}\hss\includegraphics[width=1.10cm,height=1.8cm,keepaspectratio]{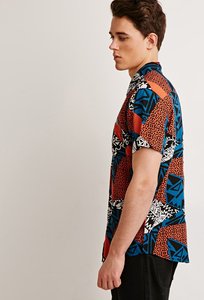}\hss\includegraphics[width=1.10cm,height=1.8cm,keepaspectratio]{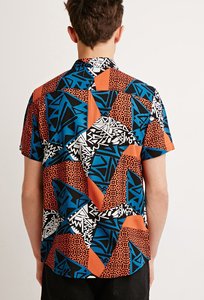}\hss\includegraphics[width=1.10cm,height=1.8cm,keepaspectratio]{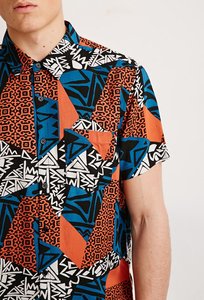}\hss\phantom{\rule{1.10cm}{1.8cm}}\hss}
      \end{minipage}
    }
  \end{minipage}
\hfill
  \begin{minipage}[t]{5.55cm}
    \fcolorbox{green!60!black}{white}{
      \begin{minipage}[t]{5.25cm}
        {\tiny\textbf{\#2}}\\
        \noindent\hbox to 5.25cm{\includegraphics[width=1.10cm,height=1.8cm,keepaspectratio]{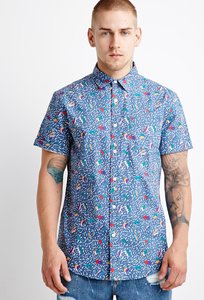}\hss\includegraphics[width=1.10cm,height=1.8cm,keepaspectratio]{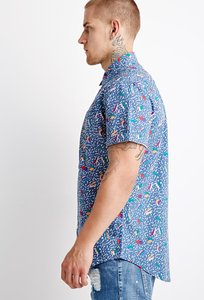}\hss\includegraphics[width=1.10cm,height=1.8cm,keepaspectratio]{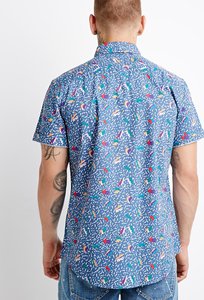}\hss\includegraphics[width=1.10cm,height=1.8cm,keepaspectratio]{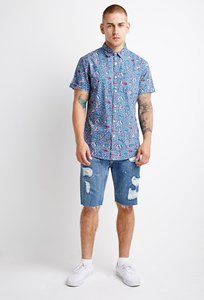}\hss\includegraphics[width=1.10cm,height=1.8cm,keepaspectratio]{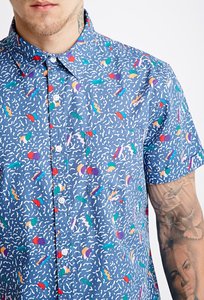}\hss}
      \end{minipage}
    }
  \end{minipage}
\hfill
  \begin{minipage}[t]{5.55cm}
    \fcolorbox{red!70!black}{white}{
      \begin{minipage}[t]{5.25cm}
        {\tiny\textbf{\#3}}\\
        \noindent\hbox to 5.25cm{\includegraphics[width=1.10cm,height=1.8cm,keepaspectratio]{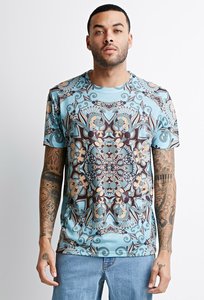}\hss\includegraphics[width=1.10cm,height=1.8cm,keepaspectratio]{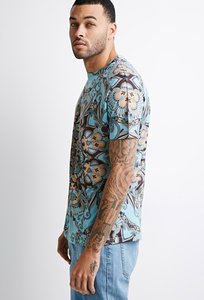}\hss\includegraphics[width=1.10cm,height=1.8cm,keepaspectratio]{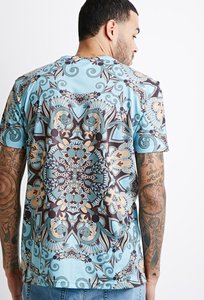}\hss\includegraphics[width=1.10cm,height=1.8cm,keepaspectratio]{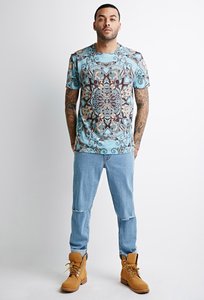}\hss\includegraphics[width=1.10cm,height=1.8cm,keepaspectratio]{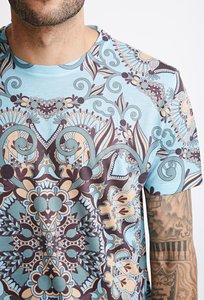}\hss}
      \end{minipage}
    }
  \end{minipage}
\par\vspace{2pt}
\noindent
  \begin{minipage}[t]{5.55cm}
    \fcolorbox{red!70!black}{white}{
      \begin{minipage}[t]{5.25cm}
        {\tiny\textbf{\#4}}\\
        \noindent\hbox to 5.25cm{\includegraphics[width=1.10cm,height=1.8cm,keepaspectratio]{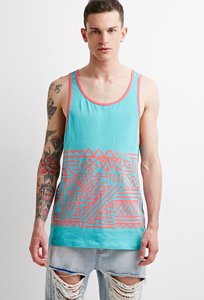}\hss\includegraphics[width=1.10cm,height=1.8cm,keepaspectratio]{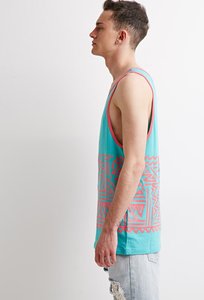}\hss\includegraphics[width=1.10cm,height=1.8cm,keepaspectratio]{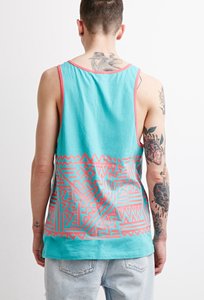}\hss\includegraphics[width=1.10cm,height=1.8cm,keepaspectratio]{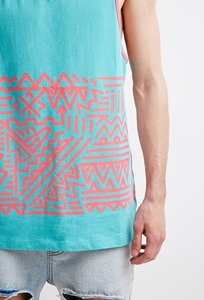}\hss\phantom{\rule{1.10cm}{1.8cm}}\hss}
      \end{minipage}
    }
  \end{minipage}
\hfill
  \begin{minipage}[t]{5.55cm}
    \fcolorbox{red!70!black}{white}{
      \begin{minipage}[t]{5.25cm}
        {\tiny\textbf{\#5}}\\
        \noindent\hbox to 5.25cm{\includegraphics[width=1.10cm,height=1.8cm,keepaspectratio]{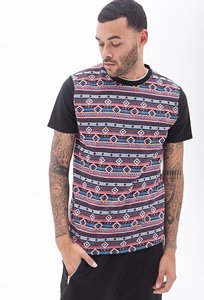}\hss\includegraphics[width=1.10cm,height=1.8cm,keepaspectratio]{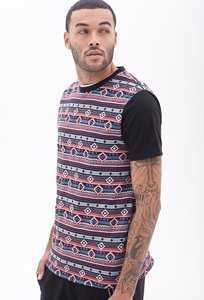}\hss\includegraphics[width=1.10cm,height=1.8cm,keepaspectratio]{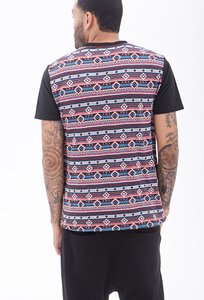}\hss\includegraphics[width=1.10cm,height=1.8cm,keepaspectratio]{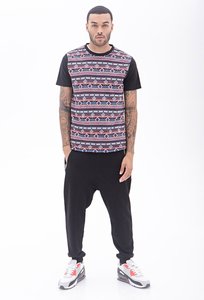}\hss\includegraphics[width=1.10cm,height=1.8cm,keepaspectratio]{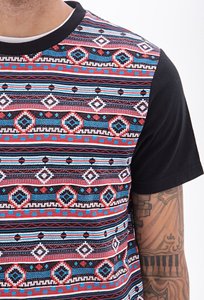}\hss}
      \end{minipage}
    }
  \end{minipage}
\hfill
  \begin{minipage}[t]{5.55cm}
    \fcolorbox{red!70!black}{white}{
      \begin{minipage}[t]{5.25cm}
        {\tiny\textbf{\#6}}\\
        \noindent\hbox to 5.25cm{\includegraphics[width=1.10cm,height=1.8cm,keepaspectratio]{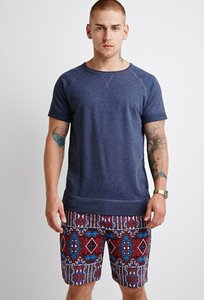}\hss\includegraphics[width=1.10cm,height=1.8cm,keepaspectratio]{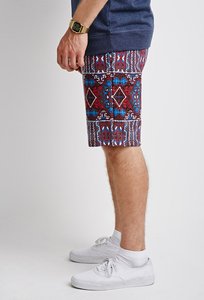}\hss\includegraphics[width=1.10cm,height=1.8cm,keepaspectratio]{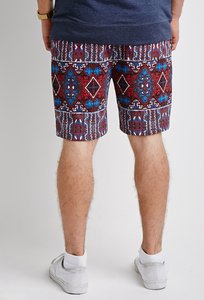}\hss\includegraphics[width=1.10cm,height=1.8cm,keepaspectratio]{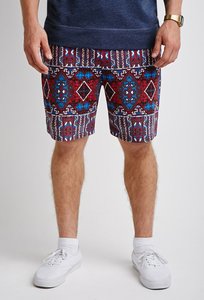}\hss\phantom{\rule{1.10cm}{1.8cm}}\hss}
      \end{minipage}
    }
  \end{minipage}
\par\vspace{2pt}
\noindent
  \begin{minipage}[t]{5.55cm}
    \fcolorbox{red!70!black}{white}{
      \begin{minipage}[t]{5.25cm}
        {\tiny\textbf{\#7}}\\
        \noindent\hbox to 5.25cm{\includegraphics[width=1.10cm,height=1.8cm,keepaspectratio]{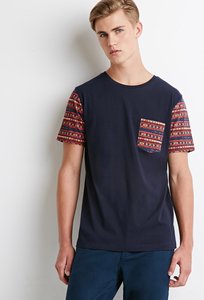}\hss\includegraphics[width=1.10cm,height=1.8cm,keepaspectratio]{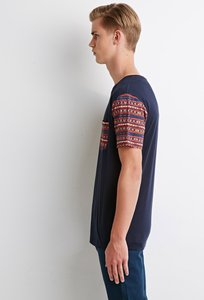}\hss\includegraphics[width=1.10cm,height=1.8cm,keepaspectratio]{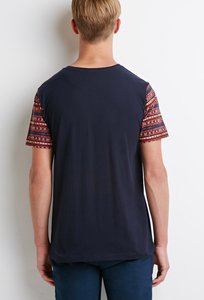}\hss\includegraphics[width=1.10cm,height=1.8cm,keepaspectratio]{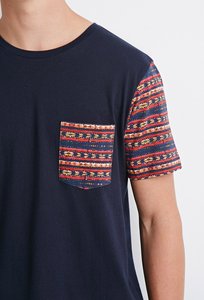}\hss\phantom{\rule{1.10cm}{1.8cm}}\hss}
      \end{minipage}
    }
  \end{minipage}
\hfill
  \begin{minipage}[t]{5.55cm}
    \fcolorbox{red!70!black}{white}{
      \begin{minipage}[t]{5.25cm}
        {\tiny\textbf{\#8}}\\
        \noindent\hbox to 5.25cm{\includegraphics[width=1.10cm,height=1.8cm,keepaspectratio]{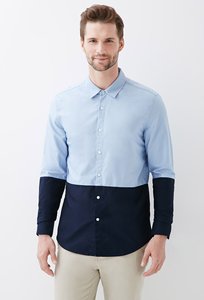}\hss\includegraphics[width=1.10cm,height=1.8cm,keepaspectratio]{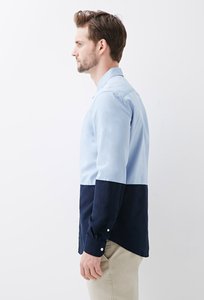}\hss\includegraphics[width=1.10cm,height=1.8cm,keepaspectratio]{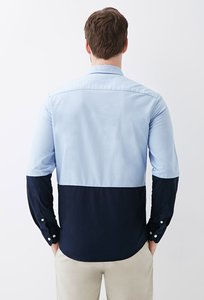}\hss\includegraphics[width=1.10cm,height=1.8cm,keepaspectratio]{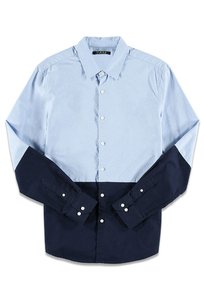}\hss\includegraphics[width=1.10cm,height=1.8cm,keepaspectratio]{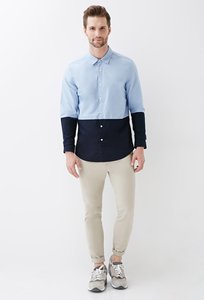}\hss}
      \end{minipage}
    }
  \end{minipage}
\hfill
  \begin{minipage}[t]{5.55cm}
    \fcolorbox{red!70!black}{white}{
      \begin{minipage}[t]{5.25cm}
        {\tiny\textbf{\#9}}\\
        \noindent\hbox to 5.25cm{\includegraphics[width=1.10cm,height=1.8cm,keepaspectratio]{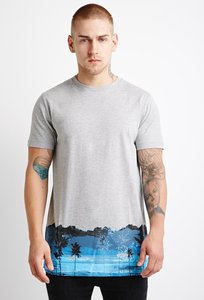}\hss\includegraphics[width=1.10cm,height=1.8cm,keepaspectratio]{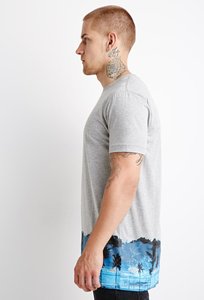}\hss\includegraphics[width=1.10cm,height=1.8cm,keepaspectratio]{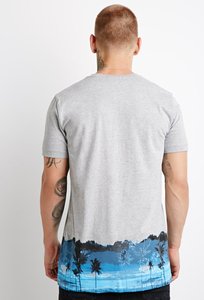}\hss\includegraphics[width=1.10cm,height=1.8cm,keepaspectratio]{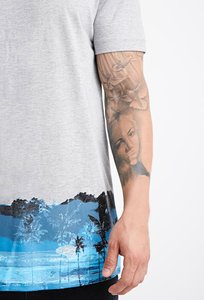}\hss\phantom{\rule{1.10cm}{1.8cm}}\hss}
      \end{minipage}
    }
  \end{minipage}
\par\vspace{2pt}
\noindent
  \begin{minipage}[t]{5.55cm}
    \fcolorbox{red!70!black}{white}{
      \begin{minipage}[t]{5.25cm}
        {\tiny\textbf{\#10}}\\
        \noindent\hbox to 5.25cm{\includegraphics[width=1.10cm,height=1.8cm,keepaspectratio]{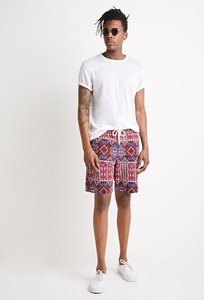}\hss\includegraphics[width=1.10cm,height=1.8cm,keepaspectratio]{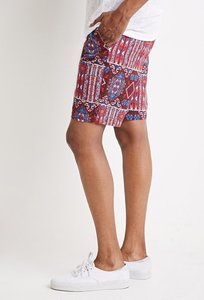}\hss\includegraphics[width=1.10cm,height=1.8cm,keepaspectratio]{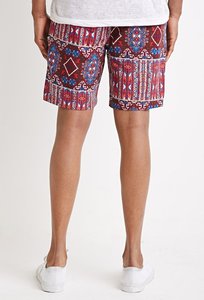}\hss\includegraphics[width=1.10cm,height=1.8cm,keepaspectratio]{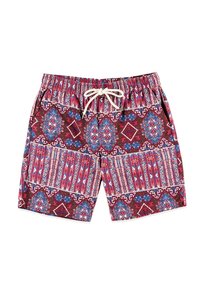}\hss\includegraphics[width=1.10cm,height=1.8cm,keepaspectratio]{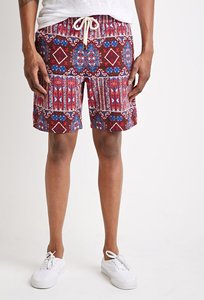}\hss}
      \end{minipage}
    }
  \end{minipage}
\hfill
  \begin{minipage}[t]{5.55cm}\end{minipage}
\hfill
  \begin{minipage}[t]{5.55cm}\end{minipage}
\par\vspace{2pt}
\noindent\rule{\linewidth}{0.4pt}
\par\vspace{2pt}
% -- doubao --
{\small\textbf{Doubao-E-V}}\quad{\small Rank~3}\\
\noindent
  \begin{minipage}[t]{5.55cm}
    \fcolorbox{red!70!black}{white}{
      \begin{minipage}[t]{5.25cm}
        {\tiny\textbf{\#1}}\\
        \noindent\hbox to 5.25cm{\includegraphics[width=1.10cm,height=1.8cm,keepaspectratio]{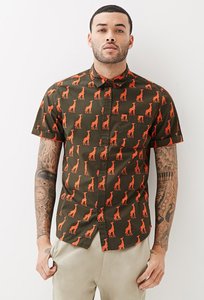}\hss\includegraphics[width=1.10cm,height=1.8cm,keepaspectratio]{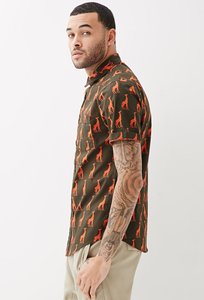}\hss\includegraphics[width=1.10cm,height=1.8cm,keepaspectratio]{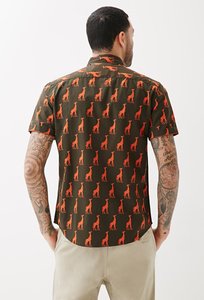}\hss\includegraphics[width=1.10cm,height=1.8cm,keepaspectratio]{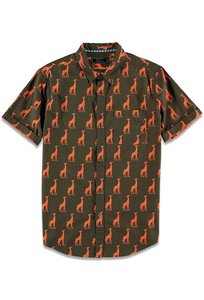}\hss\includegraphics[width=1.10cm,height=1.8cm,keepaspectratio]{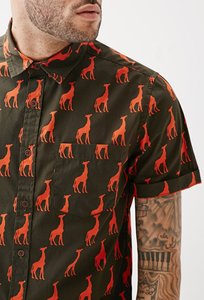}\hss}
      \end{minipage}
    }
  \end{minipage}
\hfill
  \begin{minipage}[t]{5.55cm}
    \fcolorbox{red!70!black}{white}{
      \begin{minipage}[t]{5.25cm}
        {\tiny\textbf{\#2}}\\
        \noindent\hbox to 5.25cm{\includegraphics[width=1.10cm,height=1.8cm,keepaspectratio]{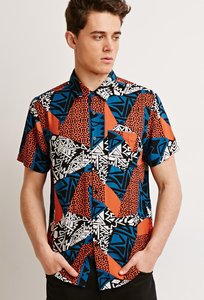}\hss\includegraphics[width=1.10cm,height=1.8cm,keepaspectratio]{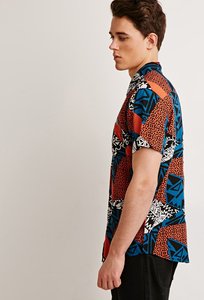}\hss\includegraphics[width=1.10cm,height=1.8cm,keepaspectratio]{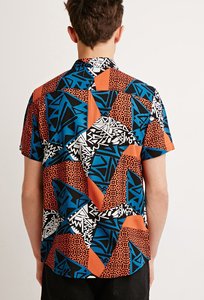}\hss\includegraphics[width=1.10cm,height=1.8cm,keepaspectratio]{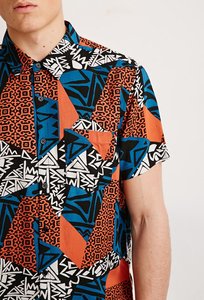}\hss\phantom{\rule{1.10cm}{1.8cm}}\hss}
      \end{minipage}
    }
  \end{minipage}
\hfill
  \begin{minipage}[t]{5.55cm}
    \fcolorbox{green!60!black}{white}{
      \begin{minipage}[t]{5.25cm}
        {\tiny\textbf{\#3}}\\
        \noindent\hbox to 5.25cm{\includegraphics[width=1.10cm,height=1.8cm,keepaspectratio]{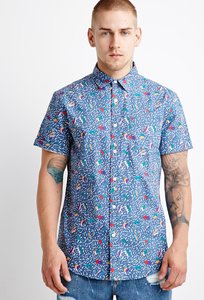}\hss\includegraphics[width=1.10cm,height=1.8cm,keepaspectratio]{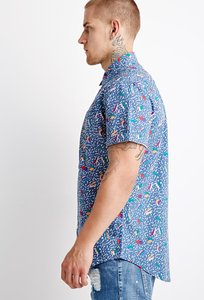}\hss\includegraphics[width=1.10cm,height=1.8cm,keepaspectratio]{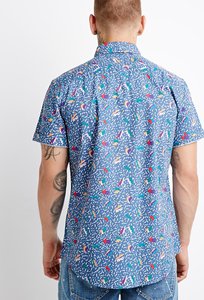}\hss\includegraphics[width=1.10cm,height=1.8cm,keepaspectratio]{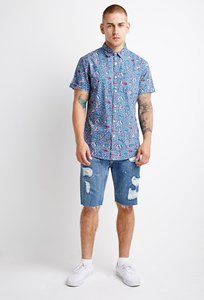}\hss\includegraphics[width=1.10cm,height=1.8cm,keepaspectratio]{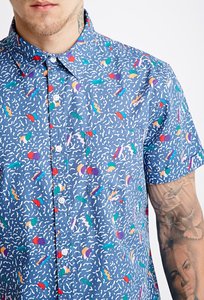}\hss}
      \end{minipage}
    }
  \end{minipage}
\par\vspace{2pt}
\noindent
  \begin{minipage}[t]{5.55cm}
    \fcolorbox{red!70!black}{white}{
      \begin{minipage}[t]{5.25cm}
        {\tiny\textbf{\#4}}\\
        \noindent\hbox to 5.25cm{\includegraphics[width=1.10cm,height=1.8cm,keepaspectratio]{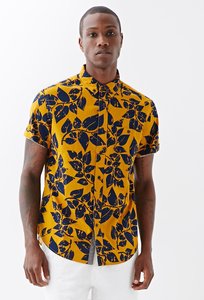}\hss\includegraphics[width=1.10cm,height=1.8cm,keepaspectratio]{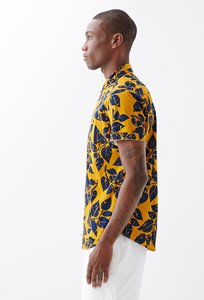}\hss\includegraphics[width=1.10cm,height=1.8cm,keepaspectratio]{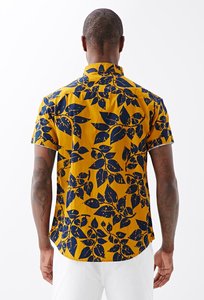}\hss\includegraphics[width=1.10cm,height=1.8cm,keepaspectratio]{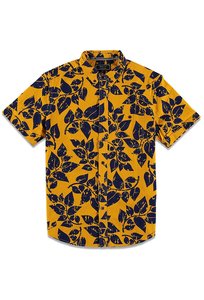}\hss\includegraphics[width=1.10cm,height=1.8cm,keepaspectratio]{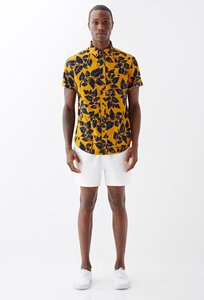}\hss}
      \end{minipage}
    }
  \end{minipage}
\hfill
  \begin{minipage}[t]{5.55cm}
    \fcolorbox{red!70!black}{white}{
      \begin{minipage}[t]{5.25cm}
        {\tiny\textbf{\#5}}\\
        \noindent\hbox to 5.25cm{\includegraphics[width=1.10cm,height=1.8cm,keepaspectratio]{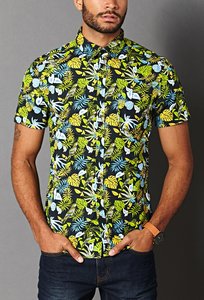}\hss\includegraphics[width=1.10cm,height=1.8cm,keepaspectratio]{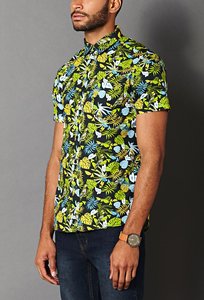}\hss\includegraphics[width=1.10cm,height=1.8cm,keepaspectratio]{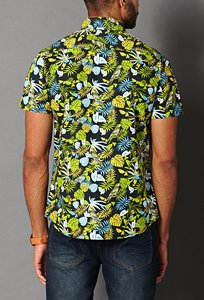}\hss\includegraphics[width=1.10cm,height=1.8cm,keepaspectratio]{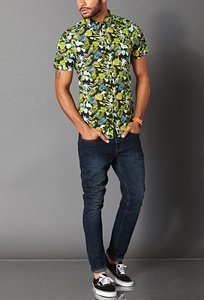}\hss\includegraphics[width=1.10cm,height=1.8cm,keepaspectratio]{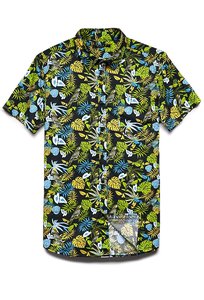}\hss}
      \end{minipage}
    }
  \end{minipage}
\hfill
  \begin{minipage}[t]{5.55cm}
    \fcolorbox{red!70!black}{white}{
      \begin{minipage}[t]{5.25cm}
        {\tiny\textbf{\#6}}\\
        \noindent\hbox to 5.25cm{\includegraphics[width=1.10cm,height=1.8cm,keepaspectratio]{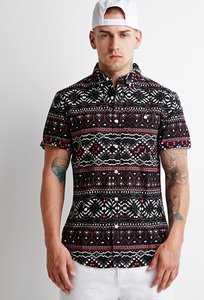}\hss\includegraphics[width=1.10cm,height=1.8cm,keepaspectratio]{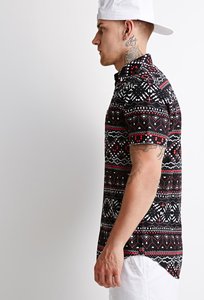}\hss\includegraphics[width=1.10cm,height=1.8cm,keepaspectratio]{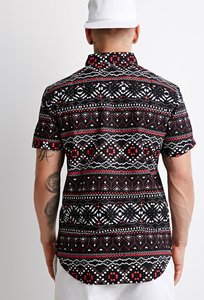}\hss\includegraphics[width=1.10cm,height=1.8cm,keepaspectratio]{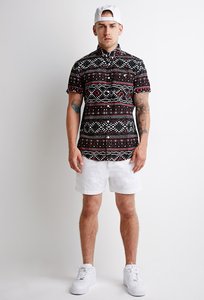}\hss\includegraphics[width=1.10cm,height=1.8cm,keepaspectratio]{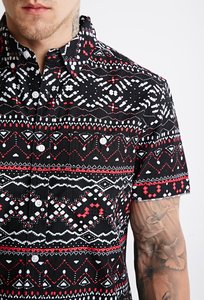}\hss}
      \end{minipage}
    }
  \end{minipage}
\par\vspace{2pt}
\noindent
  \begin{minipage}[t]{5.55cm}
    \fcolorbox{red!70!black}{white}{
      \begin{minipage}[t]{5.25cm}
        {\tiny\textbf{\#7}}\\
        \noindent\hbox to 5.25cm{\includegraphics[width=1.10cm,height=1.8cm,keepaspectratio]{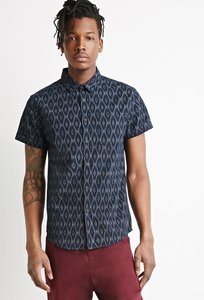}\hss\includegraphics[width=1.10cm,height=1.8cm,keepaspectratio]{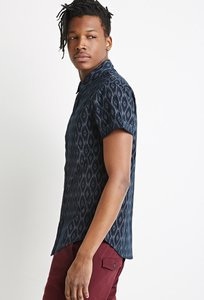}\hss\includegraphics[width=1.10cm,height=1.8cm,keepaspectratio]{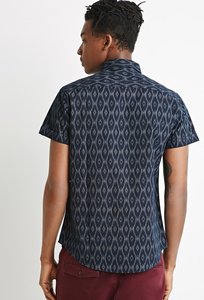}\hss\includegraphics[width=1.10cm,height=1.8cm,keepaspectratio]{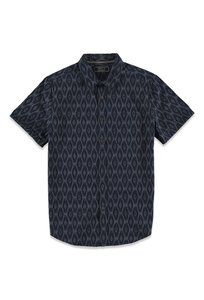}\hss\includegraphics[width=1.10cm,height=1.8cm,keepaspectratio]{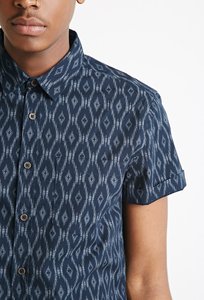}\hss}
      \end{minipage}
    }
  \end{minipage}
\hfill
  \begin{minipage}[t]{5.55cm}
    \fcolorbox{red!70!black}{white}{
      \begin{minipage}[t]{5.25cm}
        {\tiny\textbf{\#8}}\\
        \noindent\hbox to 5.25cm{\includegraphics[width=1.10cm,height=1.8cm,keepaspectratio]{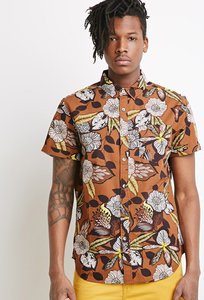}\hss\includegraphics[width=1.10cm,height=1.8cm,keepaspectratio]{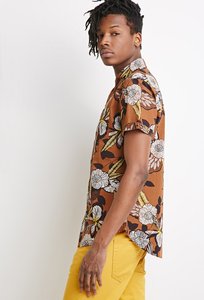}\hss\includegraphics[width=1.10cm,height=1.8cm,keepaspectratio]{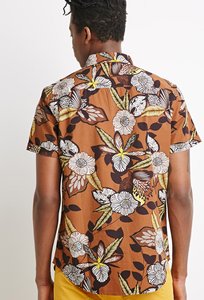}\hss\includegraphics[width=1.10cm,height=1.8cm,keepaspectratio]{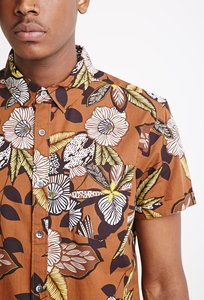}\hss\phantom{\rule{1.10cm}{1.8cm}}\hss}
      \end{minipage}
    }
  \end{minipage}
\hfill
  \begin{minipage}[t]{5.55cm}
    \fcolorbox{red!70!black}{white}{
      \begin{minipage}[t]{5.25cm}
        {\tiny\textbf{\#9}}\\
        \noindent\hbox to 5.25cm{\includegraphics[width=1.10cm,height=1.8cm,keepaspectratio]{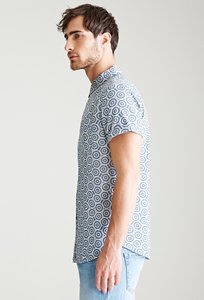}\hss\includegraphics[width=1.10cm,height=1.8cm,keepaspectratio]{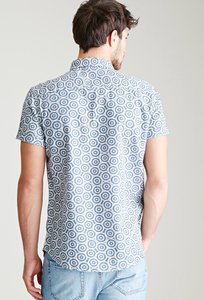}\hss\includegraphics[width=1.10cm,height=1.8cm,keepaspectratio]{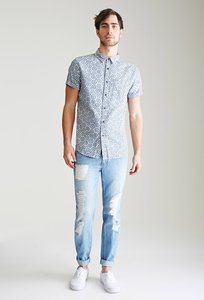}\hss\includegraphics[width=1.10cm,height=1.8cm,keepaspectratio]{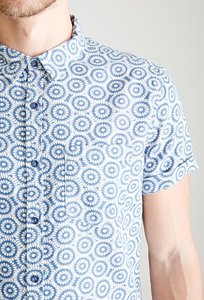}\hss\phantom{\rule{1.10cm}{1.8cm}}\hss}
      \end{minipage}
    }
  \end{minipage}
\par\vspace{2pt}
\noindent
  \begin{minipage}[t]{5.55cm}
    \fcolorbox{red!70!black}{white}{
      \begin{minipage}[t]{5.25cm}
        {\tiny\textbf{\#10}}\\
        \noindent\hbox to 5.25cm{\includegraphics[width=1.10cm,height=1.8cm,keepaspectratio]{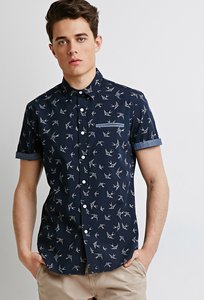}\hss\includegraphics[width=1.10cm,height=1.8cm,keepaspectratio]{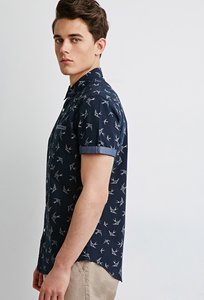}\hss\includegraphics[width=1.10cm,height=1.8cm,keepaspectratio]{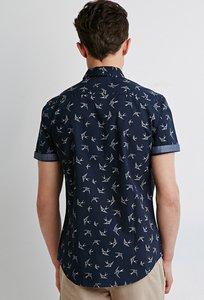}\hss\includegraphics[width=1.10cm,height=1.8cm,keepaspectratio]{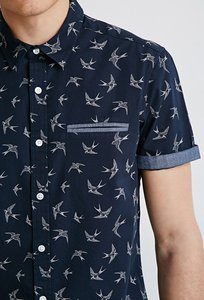}\hss\phantom{\rule{1.10cm}{1.8cm}}\hss}
      \end{minipage}
    }
  \end{minipage}
\hfill
  \begin{minipage}[t]{5.55cm}\end{minipage}
\hfill
  \begin{minipage}[t]{5.55cm}\end{minipage}
\par\vspace{2pt}
\noindent\rule{\linewidth}{0.4pt}
\par\vspace{20pt}

\par\vspace{16pt}
\noindent\textbf{\large Example 6}
\par\vspace{4pt}
\noindent\rule{\linewidth}{1.2pt}
\par\vspace{4pt}
% ── Case 6: short::deepfashion::4259 ──
\noindent\hfill%
  \begin{minipage}[t]{6.0cm}
    \noindent\hbox to 6.0cm{\includegraphics[width=1.20cm,height=2.0cm,keepaspectratio]{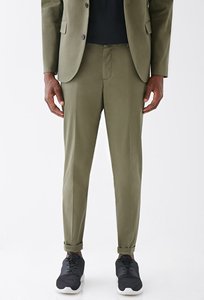}\hss\includegraphics[width=1.20cm,height=2.0cm,keepaspectratio]{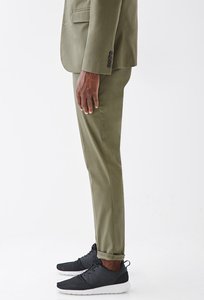}\hss\includegraphics[width=1.20cm,height=2.0cm,keepaspectratio]{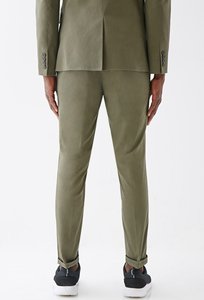}\hss\includegraphics[width=1.20cm,height=2.0cm,keepaspectratio]{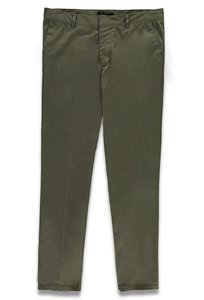}\hss\includegraphics[width=1.20cm,height=2.0cm,keepaspectratio]{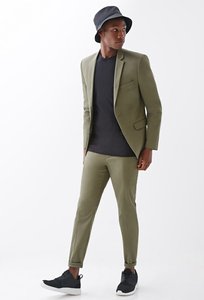}\hss}
    \par\vspace{1pt}
    {\scriptsize\textbf{}}
    \par\vspace{0pt}
    \parbox[t]{6.0cm}{\tiny\raggedright Olive green slim-fit dress pants with sharp center creases, zip fly, belt loops, and cuffed hems. Crafted from smooth suiting fabric with a subtle sheen, featuring a modern tapered silhouette perfect for contemporary formal or smart-casual wear.}
  \end{minipage}%
\hfill%
  \begin{minipage}[t]{3.5cm}
    \centering
    \vspace{0.45cm}%
    \parbox{3.5cm}{\centering\tiny Front: Add articulated horizontal knee panel seaming. Back: Replace clean rear with two symmetrical flap-style back pockets. Color changed from olive to teal technical fabric.}\\[2pt]
    {\normalsize$\longrightarrow$}
  \end{minipage}%
\hfill%
  \begin{minipage}[t]{6.0cm}
    \noindent\hbox to 6.0cm{\includegraphics[width=1.20cm,height=2.0cm,keepaspectratio]{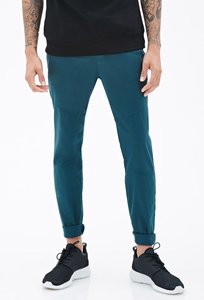}\hss\includegraphics[width=1.20cm,height=2.0cm,keepaspectratio]{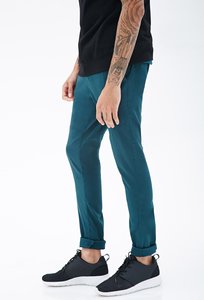}\hss\includegraphics[width=1.20cm,height=2.0cm,keepaspectratio]{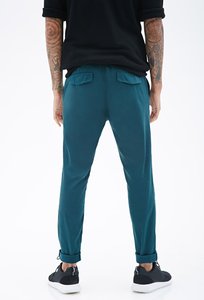}\hss\includegraphics[width=1.20cm,height=2.0cm,keepaspectratio]{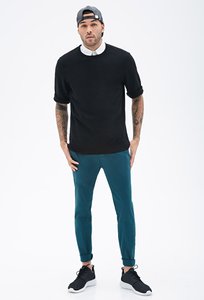}\hss\phantom{\rule{1.20cm}{2.0cm}}\hss}
    \par\vspace{1pt}
    {\scriptsize\textbf{Ground Truth}}
    \par\vspace{0pt}
    \parbox[t]{6.0cm}{\tiny\raggedright Teal slim-fit pants with articulated knee panels, cuffed hems, and clean pocket styling. Contemporary smart-casual design featuring a tapered silhouette and technical fabric construction.}
  \end{minipage}%
\hfill
\par\vspace{4pt}
\par\vspace{4pt}
\noindent\rule{\linewidth}{0.4pt}
% -- mt_align --
{\small\textbf{\textbf{Ours}}}\quad{\small Rank~1}\\
\noindent
  \begin{minipage}[t]{5.55cm}
    \fcolorbox{green!60!black}{white}{
      \begin{minipage}[t]{5.25cm}
        {\tiny\textbf{\#1}}\\
        \noindent\hbox to 5.25cm{\includegraphics[width=1.10cm,height=1.8cm,keepaspectratio]{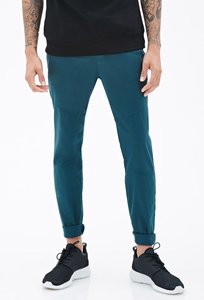}\hss\includegraphics[width=1.10cm,height=1.8cm,keepaspectratio]{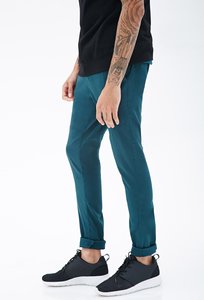}\hss\includegraphics[width=1.10cm,height=1.8cm,keepaspectratio]{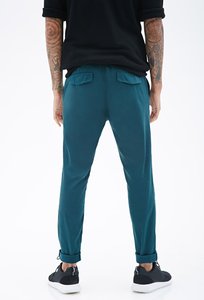}\hss\includegraphics[width=1.10cm,height=1.8cm,keepaspectratio]{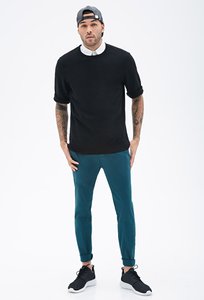}\hss\phantom{\rule{1.10cm}{1.8cm}}\hss}
      \end{minipage}
    }
  \end{minipage}
\hfill
  \begin{minipage}[t]{5.55cm}
    \fcolorbox{red!70!black}{white}{
      \begin{minipage}[t]{5.25cm}
        {\tiny\textbf{\#2}}\\
        \noindent\hbox to 5.25cm{\includegraphics[width=1.10cm,height=1.8cm,keepaspectratio]{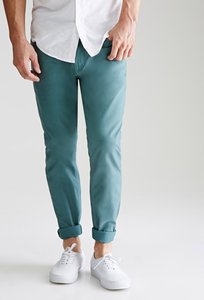}\hss\includegraphics[width=1.10cm,height=1.8cm,keepaspectratio]{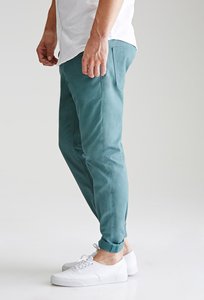}\hss\includegraphics[width=1.10cm,height=1.8cm,keepaspectratio]{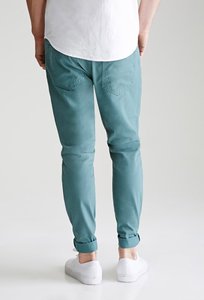}\hss\includegraphics[width=1.10cm,height=1.8cm,keepaspectratio]{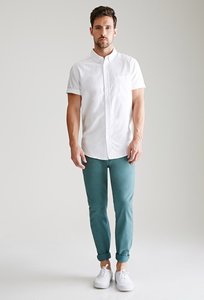}\hss\phantom{\rule{1.10cm}{1.8cm}}\hss}
      \end{minipage}
    }
  \end{minipage}
\hfill
  \begin{minipage}[t]{5.55cm}
    \fcolorbox{red!70!black}{white}{
      \begin{minipage}[t]{5.25cm}
        {\tiny\textbf{\#3}}\\
        \noindent\hbox to 5.25cm{\includegraphics[width=1.10cm,height=1.8cm,keepaspectratio]{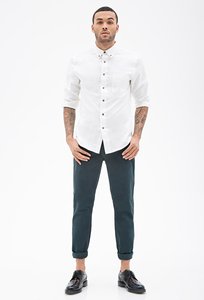}\hss\includegraphics[width=1.10cm,height=1.8cm,keepaspectratio]{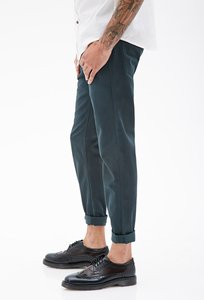}\hss\includegraphics[width=1.10cm,height=1.8cm,keepaspectratio]{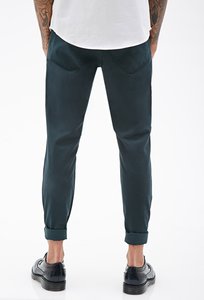}\hss\includegraphics[width=1.10cm,height=1.8cm,keepaspectratio]{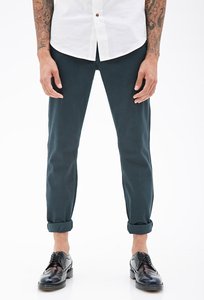}\hss\phantom{\rule{1.10cm}{1.8cm}}\hss}
      \end{minipage}
    }
  \end{minipage}
\par\vspace{2pt}
\noindent
  \begin{minipage}[t]{5.55cm}
    \fcolorbox{red!70!black}{white}{
      \begin{minipage}[t]{5.25cm}
        {\tiny\textbf{\#4}}\\
        \noindent\hbox to 5.25cm{\includegraphics[width=1.10cm,height=1.8cm,keepaspectratio]{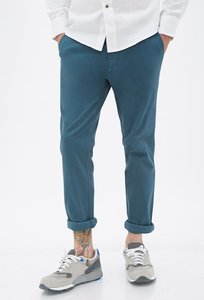}\hss\includegraphics[width=1.10cm,height=1.8cm,keepaspectratio]{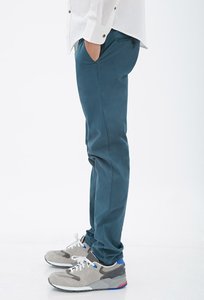}\hss\includegraphics[width=1.10cm,height=1.8cm,keepaspectratio]{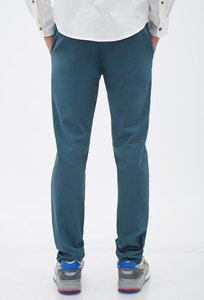}\hss\includegraphics[width=1.10cm,height=1.8cm,keepaspectratio]{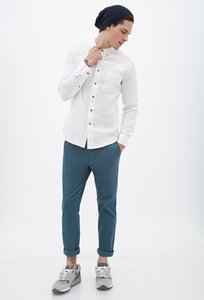}\hss\phantom{\rule{1.10cm}{1.8cm}}\hss}
      \end{minipage}
    }
  \end{minipage}
\hfill
  \begin{minipage}[t]{5.55cm}
    \fcolorbox{red!70!black}{white}{
      \begin{minipage}[t]{5.25cm}
        {\tiny\textbf{\#5}}\\
        \noindent\hbox to 5.25cm{\includegraphics[width=1.10cm,height=1.8cm,keepaspectratio]{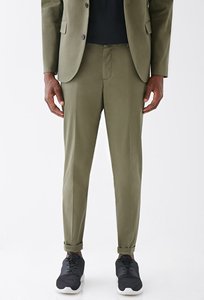}\hss\includegraphics[width=1.10cm,height=1.8cm,keepaspectratio]{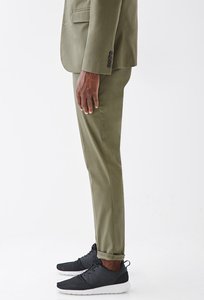}\hss\includegraphics[width=1.10cm,height=1.8cm,keepaspectratio]{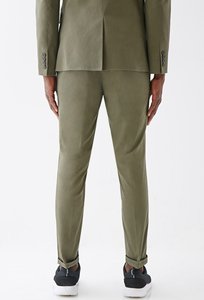}\hss\includegraphics[width=1.10cm,height=1.8cm,keepaspectratio]{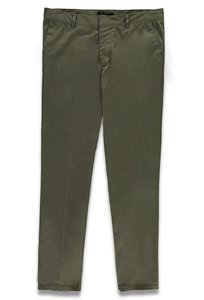}\hss\includegraphics[width=1.10cm,height=1.8cm,keepaspectratio]{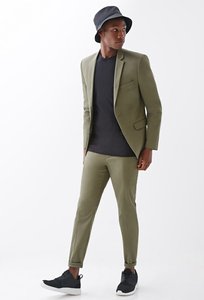}\hss}
      \end{minipage}
    }
  \end{minipage}
\hfill
  \begin{minipage}[t]{5.55cm}
    \fcolorbox{red!70!black}{white}{
      \begin{minipage}[t]{5.25cm}
        {\tiny\textbf{\#6}}\\
        \noindent\hbox to 5.25cm{\includegraphics[width=1.10cm,height=1.8cm,keepaspectratio]{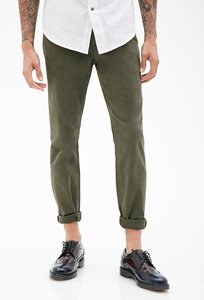}\hss\includegraphics[width=1.10cm,height=1.8cm,keepaspectratio]{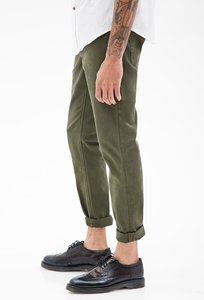}\hss\includegraphics[width=1.10cm,height=1.8cm,keepaspectratio]{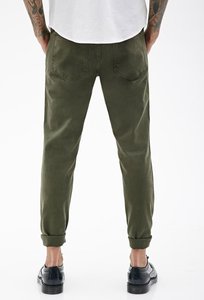}\hss\includegraphics[width=1.10cm,height=1.8cm,keepaspectratio]{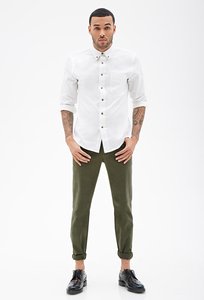}\hss\phantom{\rule{1.10cm}{1.8cm}}\hss}
      \end{minipage}
    }
  \end{minipage}
\par\vspace{2pt}
\noindent
  \begin{minipage}[t]{5.55cm}
    \fcolorbox{red!70!black}{white}{
      \begin{minipage}[t]{5.25cm}
        {\tiny\textbf{\#7}}\\
        \noindent\hbox to 5.25cm{\includegraphics[width=1.10cm,height=1.8cm,keepaspectratio]{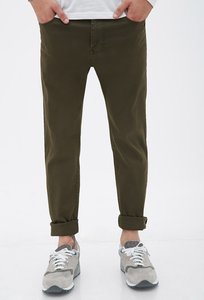}\hss\includegraphics[width=1.10cm,height=1.8cm,keepaspectratio]{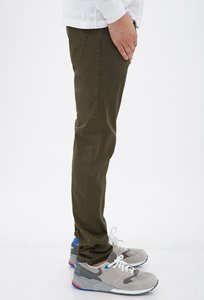}\hss\includegraphics[width=1.10cm,height=1.8cm,keepaspectratio]{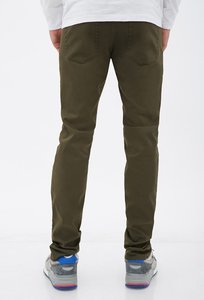}\hss\includegraphics[width=1.10cm,height=1.8cm,keepaspectratio]{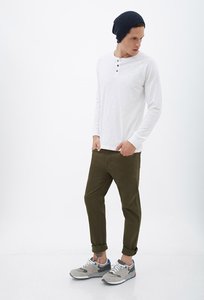}\hss\phantom{\rule{1.10cm}{1.8cm}}\hss}
      \end{minipage}
    }
  \end{minipage}
\hfill
  \begin{minipage}[t]{5.55cm}
    \fcolorbox{red!70!black}{white}{
      \begin{minipage}[t]{5.25cm}
        {\tiny\textbf{\#8}}\\
        \noindent\hbox to 5.25cm{\includegraphics[width=1.10cm,height=1.8cm,keepaspectratio]{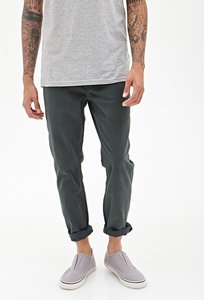}\hss\includegraphics[width=1.10cm,height=1.8cm,keepaspectratio]{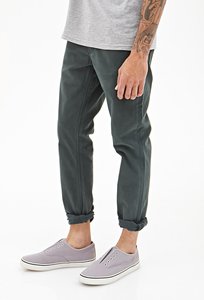}\hss\includegraphics[width=1.10cm,height=1.8cm,keepaspectratio]{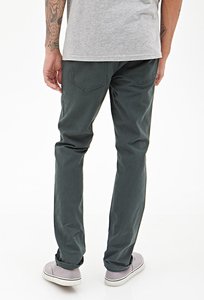}\hss\includegraphics[width=1.10cm,height=1.8cm,keepaspectratio]{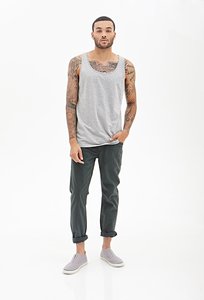}\hss\phantom{\rule{1.10cm}{1.8cm}}\hss}
      \end{minipage}
    }
  \end{minipage}
\hfill
  \begin{minipage}[t]{5.55cm}
    \fcolorbox{red!70!black}{white}{
      \begin{minipage}[t]{5.25cm}
        {\tiny\textbf{\#9}}\\
        \noindent\hbox to 5.25cm{\includegraphics[width=1.10cm,height=1.8cm,keepaspectratio]{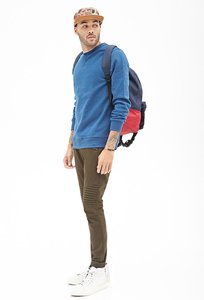}\hss\includegraphics[width=1.10cm,height=1.8cm,keepaspectratio]{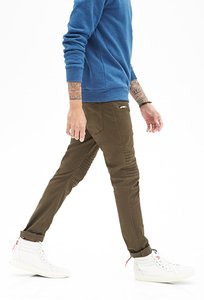}\hss\includegraphics[width=1.10cm,height=1.8cm,keepaspectratio]{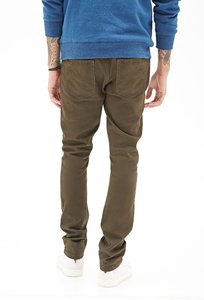}\hss\includegraphics[width=1.10cm,height=1.8cm,keepaspectratio]{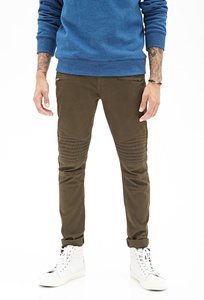}\hss\phantom{\rule{1.10cm}{1.8cm}}\hss}
      \end{minipage}
    }
  \end{minipage}
\par\vspace{2pt}
\noindent
  \begin{minipage}[t]{5.55cm}
    \fcolorbox{red!70!black}{white}{
      \begin{minipage}[t]{5.25cm}
        {\tiny\textbf{\#10}}\\
        \noindent\hbox to 5.25cm{\includegraphics[width=1.10cm,height=1.8cm,keepaspectratio]{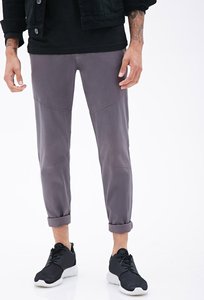}\hss\includegraphics[width=1.10cm,height=1.8cm,keepaspectratio]{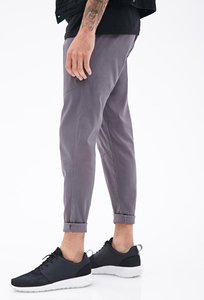}\hss\includegraphics[width=1.10cm,height=1.8cm,keepaspectratio]{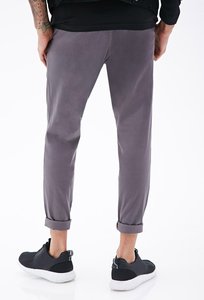}\hss\includegraphics[width=1.10cm,height=1.8cm,keepaspectratio]{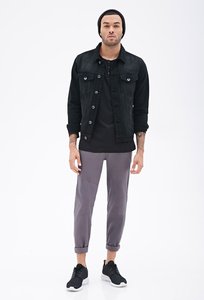}\hss\phantom{\rule{1.10cm}{1.8cm}}\hss}
      \end{minipage}
    }
  \end{minipage}
\hfill
  \begin{minipage}[t]{5.55cm}\end{minipage}
\hfill
  \begin{minipage}[t]{5.55cm}\end{minipage}
\par\vspace{2pt}
\noindent\rule{\linewidth}{0.4pt}
\par\vspace{2pt}
% -- qwen3_vl_2b --
{\small\textbf{Qwen3-VL-2B}}\quad{\small Rank~3}\\
\noindent
  \begin{minipage}[t]{5.55cm}
    \fcolorbox{red!70!black}{white}{
      \begin{minipage}[t]{5.25cm}
        {\tiny\textbf{\#1}}\\
        \noindent\hbox to 5.25cm{\includegraphics[width=1.10cm,height=1.8cm,keepaspectratio]{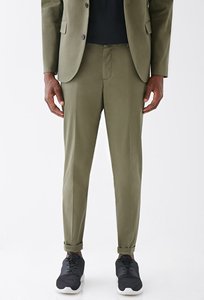}\hss\includegraphics[width=1.10cm,height=1.8cm,keepaspectratio]{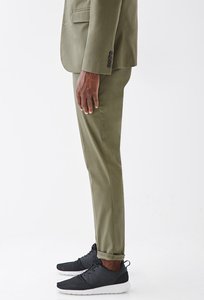}\hss\includegraphics[width=1.10cm,height=1.8cm,keepaspectratio]{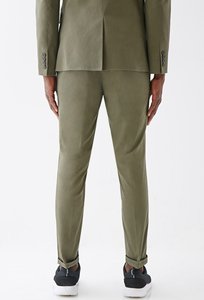}\hss\includegraphics[width=1.10cm,height=1.8cm,keepaspectratio]{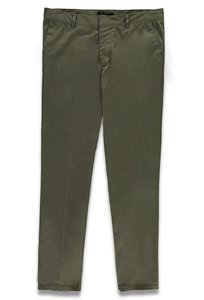}\hss\includegraphics[width=1.10cm,height=1.8cm,keepaspectratio]{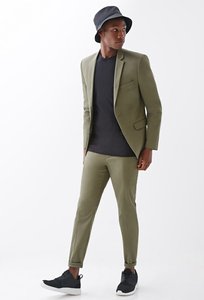}\hss}
      \end{minipage}
    }
  \end{minipage}
\hfill
  \begin{minipage}[t]{5.55cm}
    \fcolorbox{red!70!black}{white}{
      \begin{minipage}[t]{5.25cm}
        {\tiny\textbf{\#2}}\\
        \noindent\hbox to 5.25cm{\includegraphics[width=1.10cm,height=1.8cm,keepaspectratio]{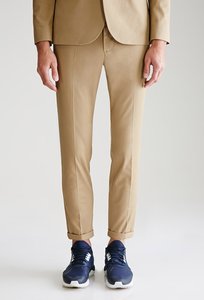}\hss\includegraphics[width=1.10cm,height=1.8cm,keepaspectratio]{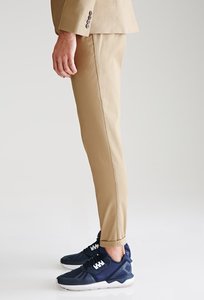}\hss\includegraphics[width=1.10cm,height=1.8cm,keepaspectratio]{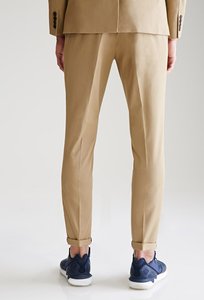}\hss\includegraphics[width=1.10cm,height=1.8cm,keepaspectratio]{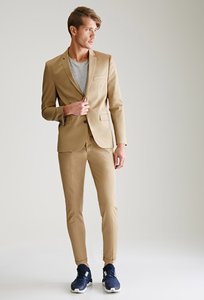}\hss\phantom{\rule{1.10cm}{1.8cm}}\hss}
      \end{minipage}
    }
  \end{minipage}
\hfill
  \begin{minipage}[t]{5.55cm}
    \fcolorbox{green!60!black}{white}{
      \begin{minipage}[t]{5.25cm}
        {\tiny\textbf{\#3}}\\
        \noindent\hbox to 5.25cm{\includegraphics[width=1.10cm,height=1.8cm,keepaspectratio]{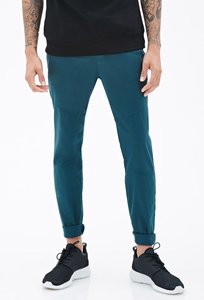}\hss\includegraphics[width=1.10cm,height=1.8cm,keepaspectratio]{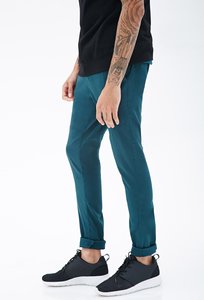}\hss\includegraphics[width=1.10cm,height=1.8cm,keepaspectratio]{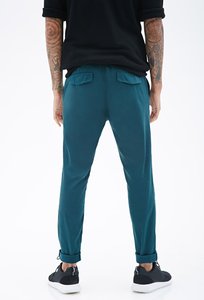}\hss\includegraphics[width=1.10cm,height=1.8cm,keepaspectratio]{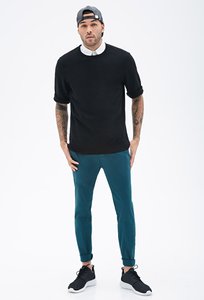}\hss\phantom{\rule{1.10cm}{1.8cm}}\hss}
      \end{minipage}
    }
  \end{minipage}
\par\vspace{2pt}
\noindent
  \begin{minipage}[t]{5.55cm}
    \fcolorbox{red!70!black}{white}{
      \begin{minipage}[t]{5.25cm}
        {\tiny\textbf{\#4}}\\
        \noindent\hbox to 5.25cm{\includegraphics[width=1.10cm,height=1.8cm,keepaspectratio]{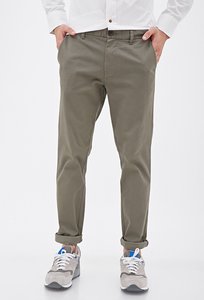}\hss\includegraphics[width=1.10cm,height=1.8cm,keepaspectratio]{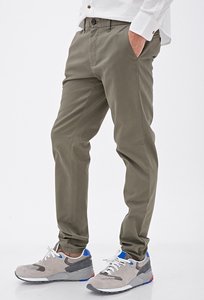}\hss\includegraphics[width=1.10cm,height=1.8cm,keepaspectratio]{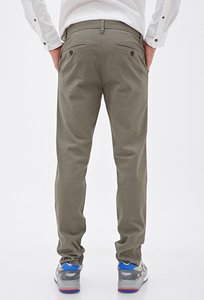}\hss\includegraphics[width=1.10cm,height=1.8cm,keepaspectratio]{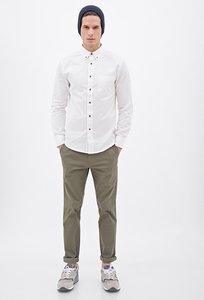}\hss\phantom{\rule{1.10cm}{1.8cm}}\hss}
      \end{minipage}
    }
  \end{minipage}
\hfill
  \begin{minipage}[t]{5.55cm}
    \fcolorbox{red!70!black}{white}{
      \begin{minipage}[t]{5.25cm}
        {\tiny\textbf{\#5}}\\
        \noindent\hbox to 5.25cm{\includegraphics[width=1.10cm,height=1.8cm,keepaspectratio]{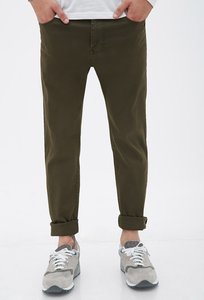}\hss\includegraphics[width=1.10cm,height=1.8cm,keepaspectratio]{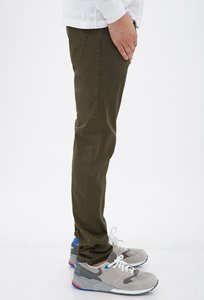}\hss\includegraphics[width=1.10cm,height=1.8cm,keepaspectratio]{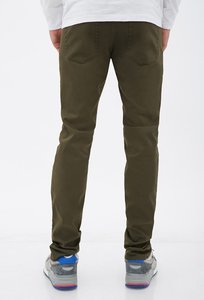}\hss\includegraphics[width=1.10cm,height=1.8cm,keepaspectratio]{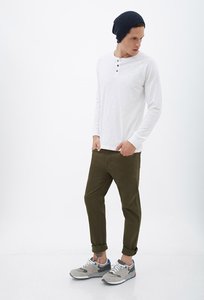}\hss\phantom{\rule{1.10cm}{1.8cm}}\hss}
      \end{minipage}
    }
  \end{minipage}
\hfill
  \begin{minipage}[t]{5.55cm}
    \fcolorbox{red!70!black}{white}{
      \begin{minipage}[t]{5.25cm}
        {\tiny\textbf{\#6}}\\
        \noindent\hbox to 5.25cm{\includegraphics[width=1.10cm,height=1.8cm,keepaspectratio]{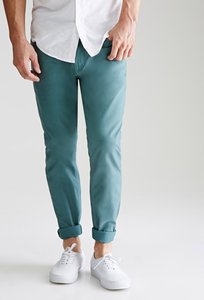}\hss\includegraphics[width=1.10cm,height=1.8cm,keepaspectratio]{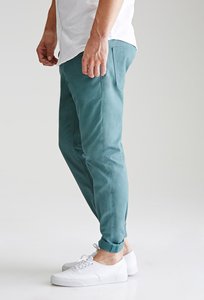}\hss\includegraphics[width=1.10cm,height=1.8cm,keepaspectratio]{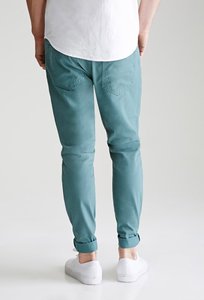}\hss\includegraphics[width=1.10cm,height=1.8cm,keepaspectratio]{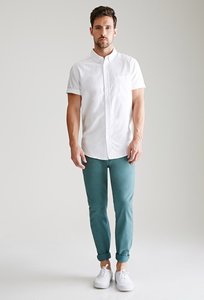}\hss\phantom{\rule{1.10cm}{1.8cm}}\hss}
      \end{minipage}
    }
  \end{minipage}
\par\vspace{2pt}
\noindent
  \begin{minipage}[t]{5.55cm}
    \fcolorbox{red!70!black}{white}{
      \begin{minipage}[t]{5.25cm}
        {\tiny\textbf{\#7}}\\
        \noindent\hbox to 5.25cm{\includegraphics[width=1.10cm,height=1.8cm,keepaspectratio]{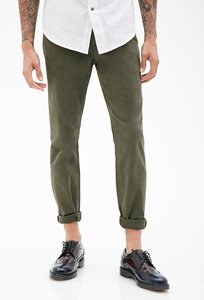}\hss\includegraphics[width=1.10cm,height=1.8cm,keepaspectratio]{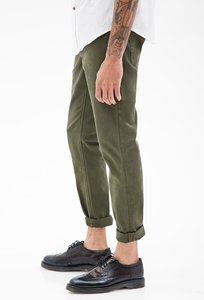}\hss\includegraphics[width=1.10cm,height=1.8cm,keepaspectratio]{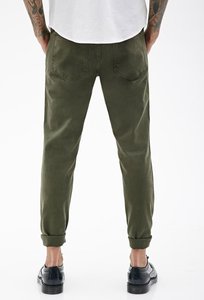}\hss\includegraphics[width=1.10cm,height=1.8cm,keepaspectratio]{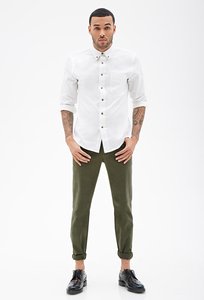}\hss\phantom{\rule{1.10cm}{1.8cm}}\hss}
      \end{minipage}
    }
  \end{minipage}
\hfill
  \begin{minipage}[t]{5.55cm}
    \fcolorbox{red!70!black}{white}{
      \begin{minipage}[t]{5.25cm}
        {\tiny\textbf{\#8}}\\
        \noindent\hbox to 5.25cm{\includegraphics[width=1.10cm,height=1.8cm,keepaspectratio]{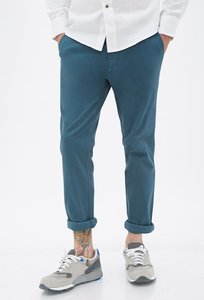}\hss\includegraphics[width=1.10cm,height=1.8cm,keepaspectratio]{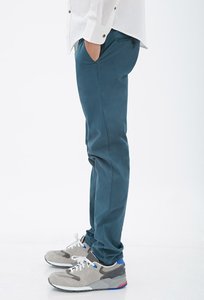}\hss\includegraphics[width=1.10cm,height=1.8cm,keepaspectratio]{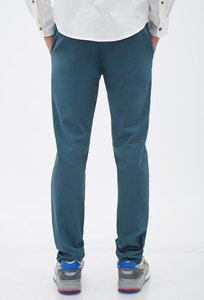}\hss\includegraphics[width=1.10cm,height=1.8cm,keepaspectratio]{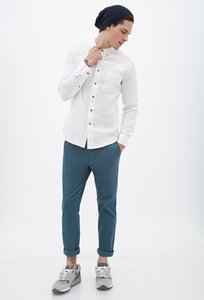}\hss\phantom{\rule{1.10cm}{1.8cm}}\hss}
      \end{minipage}
    }
  \end{minipage}
\hfill
  \begin{minipage}[t]{5.55cm}
    \fcolorbox{red!70!black}{white}{
      \begin{minipage}[t]{5.25cm}
        {\tiny\textbf{\#9}}\\
        \noindent\hbox to 5.25cm{\includegraphics[width=1.10cm,height=1.8cm,keepaspectratio]{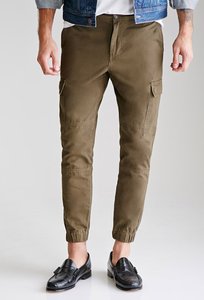}\hss\includegraphics[width=1.10cm,height=1.8cm,keepaspectratio]{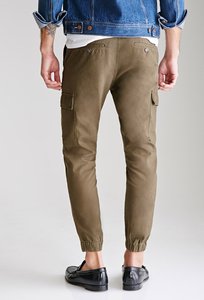}\hss\includegraphics[width=1.10cm,height=1.8cm,keepaspectratio]{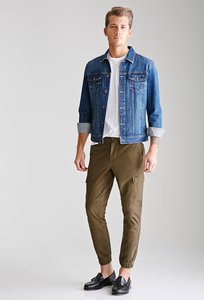}\hss\includegraphics[width=1.10cm,height=1.8cm,keepaspectratio]{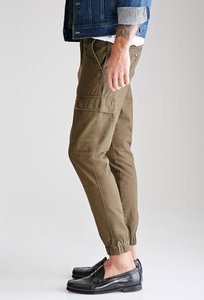}\hss\phantom{\rule{1.10cm}{1.8cm}}\hss}
      \end{minipage}
    }
  \end{minipage}
\par\vspace{2pt}
\noindent
  \begin{minipage}[t]{5.55cm}
    \fcolorbox{red!70!black}{white}{
      \begin{minipage}[t]{5.25cm}
        {\tiny\textbf{\#10}}\\
        \noindent\hbox to 5.25cm{\includegraphics[width=1.10cm,height=1.8cm,keepaspectratio]{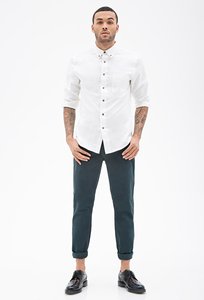}\hss\includegraphics[width=1.10cm,height=1.8cm,keepaspectratio]{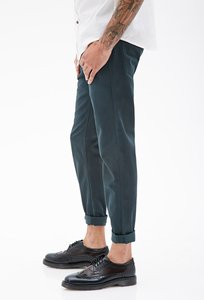}\hss\includegraphics[width=1.10cm,height=1.8cm,keepaspectratio]{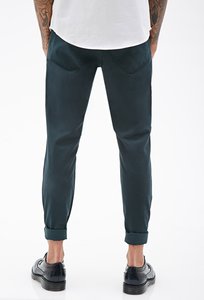}\hss\includegraphics[width=1.10cm,height=1.8cm,keepaspectratio]{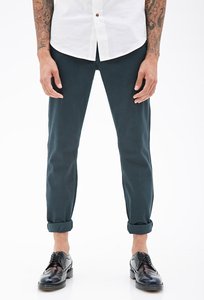}\hss\phantom{\rule{1.10cm}{1.8cm}}\hss}
      \end{minipage}
    }
  \end{minipage}
\hfill
  \begin{minipage}[t]{5.55cm}\end{minipage}
\hfill
  \begin{minipage}[t]{5.55cm}\end{minipage}
\par\vspace{2pt}
\noindent\rule{\linewidth}{0.4pt}
\par\vspace{2pt}
% -- qwen3_vl_8b --
{\small\textbf{Qwen3-VL-8B}}\quad{\small Rank~1}\\
\noindent
  \begin{minipage}[t]{5.55cm}
    \fcolorbox{green!60!black}{white}{
      \begin{minipage}[t]{5.25cm}
        {\tiny\textbf{\#1}}\\
        \noindent\hbox to 5.25cm{\includegraphics[width=1.10cm,height=1.8cm,keepaspectratio]{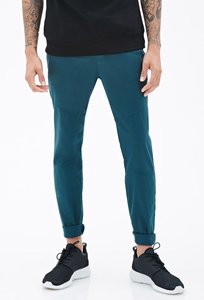}\hss\includegraphics[width=1.10cm,height=1.8cm,keepaspectratio]{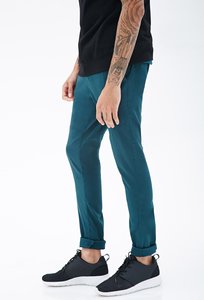}\hss\includegraphics[width=1.10cm,height=1.8cm,keepaspectratio]{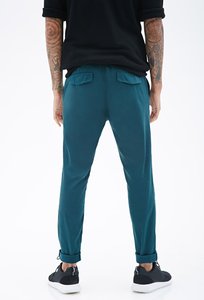}\hss\includegraphics[width=1.10cm,height=1.8cm,keepaspectratio]{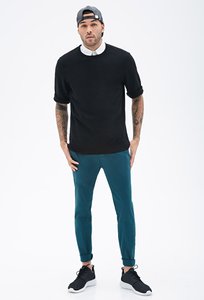}\hss\phantom{\rule{1.10cm}{1.8cm}}\hss}
      \end{minipage}
    }
  \end{minipage}
\hfill
  \begin{minipage}[t]{5.55cm}
    \fcolorbox{red!70!black}{white}{
      \begin{minipage}[t]{5.25cm}
        {\tiny\textbf{\#2}}\\
        \noindent\hbox to 5.25cm{\includegraphics[width=1.10cm,height=1.8cm,keepaspectratio]{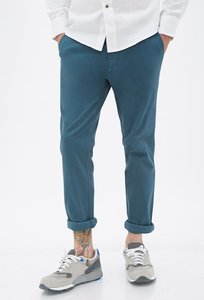}\hss\includegraphics[width=1.10cm,height=1.8cm,keepaspectratio]{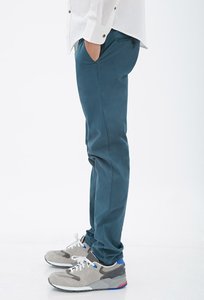}\hss\includegraphics[width=1.10cm,height=1.8cm,keepaspectratio]{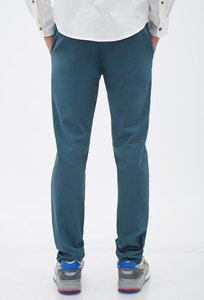}\hss\includegraphics[width=1.10cm,height=1.8cm,keepaspectratio]{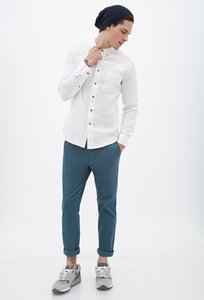}\hss\phantom{\rule{1.10cm}{1.8cm}}\hss}
      \end{minipage}
    }
  \end{minipage}
\hfill
  \begin{minipage}[t]{5.55cm}
    \fcolorbox{red!70!black}{white}{
      \begin{minipage}[t]{5.25cm}
        {\tiny\textbf{\#3}}\\
        \noindent\hbox to 5.25cm{\includegraphics[width=1.10cm,height=1.8cm,keepaspectratio]{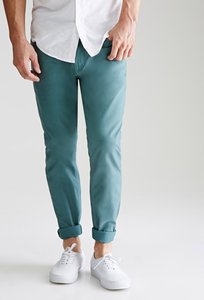}\hss\includegraphics[width=1.10cm,height=1.8cm,keepaspectratio]{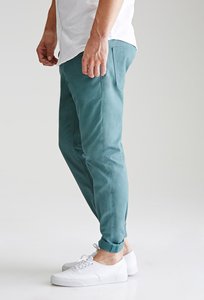}\hss\includegraphics[width=1.10cm,height=1.8cm,keepaspectratio]{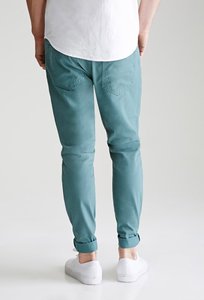}\hss\includegraphics[width=1.10cm,height=1.8cm,keepaspectratio]{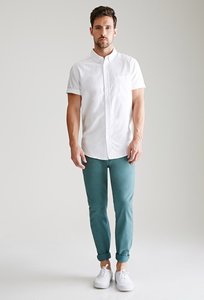}\hss\phantom{\rule{1.10cm}{1.8cm}}\hss}
      \end{minipage}
    }
  \end{minipage}
\par\vspace{2pt}
\noindent
  \begin{minipage}[t]{5.55cm}
    \fcolorbox{red!70!black}{white}{
      \begin{minipage}[t]{5.25cm}
        {\tiny\textbf{\#4}}\\
        \noindent\hbox to 5.25cm{\includegraphics[width=1.10cm,height=1.8cm,keepaspectratio]{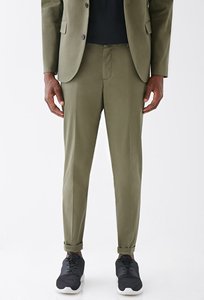}\hss\includegraphics[width=1.10cm,height=1.8cm,keepaspectratio]{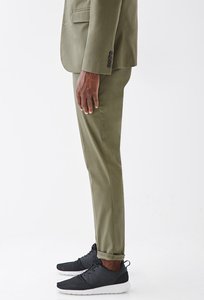}\hss\includegraphics[width=1.10cm,height=1.8cm,keepaspectratio]{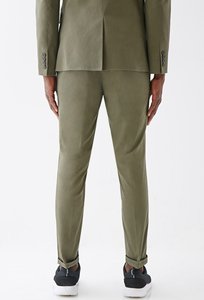}\hss\includegraphics[width=1.10cm,height=1.8cm,keepaspectratio]{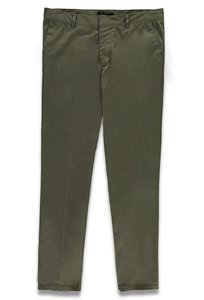}\hss\includegraphics[width=1.10cm,height=1.8cm,keepaspectratio]{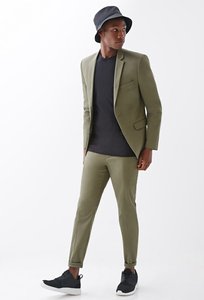}\hss}
      \end{minipage}
    }
  \end{minipage}
\hfill
  \begin{minipage}[t]{5.55cm}
    \fcolorbox{red!70!black}{white}{
      \begin{minipage}[t]{5.25cm}
        {\tiny\textbf{\#5}}\\
        \noindent\hbox to 5.25cm{\includegraphics[width=1.10cm,height=1.8cm,keepaspectratio]{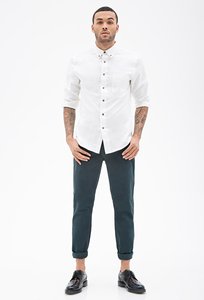}\hss\includegraphics[width=1.10cm,height=1.8cm,keepaspectratio]{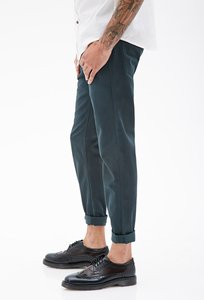}\hss\includegraphics[width=1.10cm,height=1.8cm,keepaspectratio]{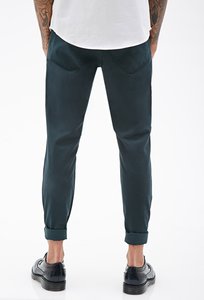}\hss\includegraphics[width=1.10cm,height=1.8cm,keepaspectratio]{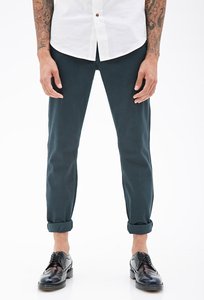}\hss\phantom{\rule{1.10cm}{1.8cm}}\hss}
      \end{minipage}
    }
  \end{minipage}
\hfill
  \begin{minipage}[t]{5.55cm}
    \fcolorbox{red!70!black}{white}{
      \begin{minipage}[t]{5.25cm}
        {\tiny\textbf{\#6}}\\
        \noindent\hbox to 5.25cm{\includegraphics[width=1.10cm,height=1.8cm,keepaspectratio]{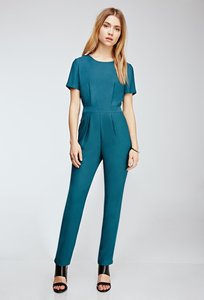}\hss\includegraphics[width=1.10cm,height=1.8cm,keepaspectratio]{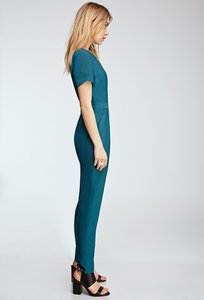}\hss\includegraphics[width=1.10cm,height=1.8cm,keepaspectratio]{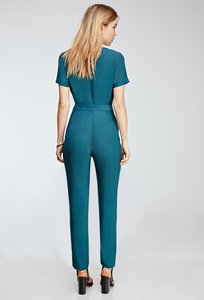}\hss\phantom{\rule{1.10cm}{1.8cm}}\hss\phantom{\rule{1.10cm}{1.8cm}}\hss}
      \end{minipage}
    }
  \end{minipage}
\par\vspace{2pt}
\noindent
  \begin{minipage}[t]{5.55cm}
    \fcolorbox{red!70!black}{white}{
      \begin{minipage}[t]{5.25cm}
        {\tiny\textbf{\#7}}\\
        \noindent\hbox to 5.25cm{\includegraphics[width=1.10cm,height=1.8cm,keepaspectratio]{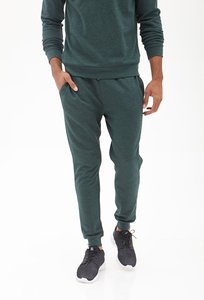}\hss\includegraphics[width=1.10cm,height=1.8cm,keepaspectratio]{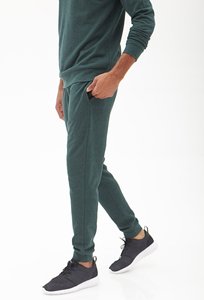}\hss\includegraphics[width=1.10cm,height=1.8cm,keepaspectratio]{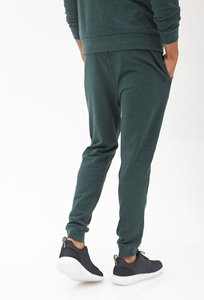}\hss\includegraphics[width=1.10cm,height=1.8cm,keepaspectratio]{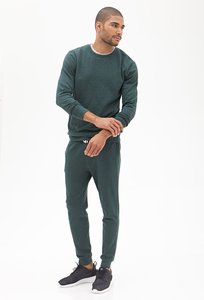}\hss\includegraphics[width=1.10cm,height=1.8cm,keepaspectratio]{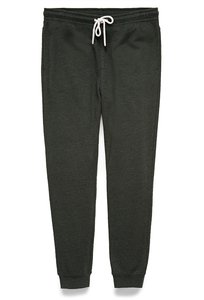}\hss}
      \end{minipage}
    }
  \end{minipage}
\hfill
  \begin{minipage}[t]{5.55cm}
    \fcolorbox{red!70!black}{white}{
      \begin{minipage}[t]{5.25cm}
        {\tiny\textbf{\#8}}\\
        \noindent\hbox to 5.25cm{\includegraphics[width=1.10cm,height=1.8cm,keepaspectratio]{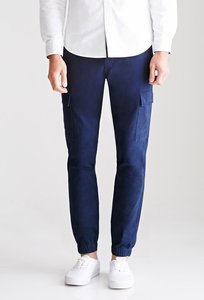}\hss\includegraphics[width=1.10cm,height=1.8cm,keepaspectratio]{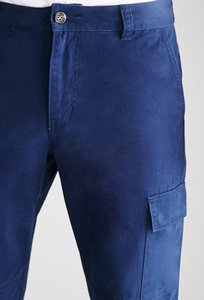}\hss\includegraphics[width=1.10cm,height=1.8cm,keepaspectratio]{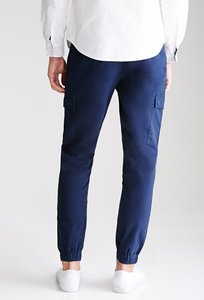}\hss\includegraphics[width=1.10cm,height=1.8cm,keepaspectratio]{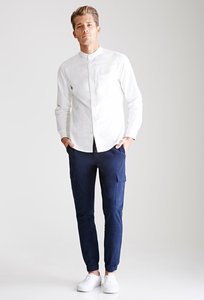}\hss\includegraphics[width=1.10cm,height=1.8cm,keepaspectratio]{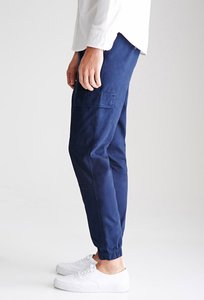}\hss}
      \end{minipage}
    }
  \end{minipage}
\hfill
  \begin{minipage}[t]{5.55cm}
    \fcolorbox{red!70!black}{white}{
      \begin{minipage}[t]{5.25cm}
        {\tiny\textbf{\#9}}\\
        \noindent\hbox to 5.25cm{\includegraphics[width=1.10cm,height=1.8cm,keepaspectratio]{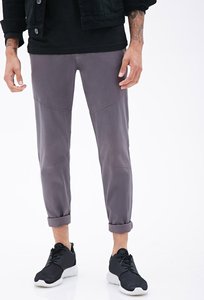}\hss\includegraphics[width=1.10cm,height=1.8cm,keepaspectratio]{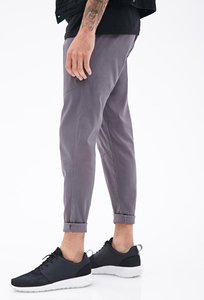}\hss\includegraphics[width=1.10cm,height=1.8cm,keepaspectratio]{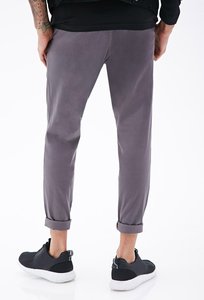}\hss\includegraphics[width=1.10cm,height=1.8cm,keepaspectratio]{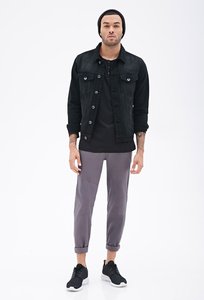}\hss\phantom{\rule{1.10cm}{1.8cm}}\hss}
      \end{minipage}
    }
  \end{minipage}
\par\vspace{2pt}
\noindent
  \begin{minipage}[t]{5.55cm}
    \fcolorbox{red!70!black}{white}{
      \begin{minipage}[t]{5.25cm}
        {\tiny\textbf{\#10}}\\
        \noindent\hbox to 5.25cm{\includegraphics[width=1.10cm,height=1.8cm,keepaspectratio]{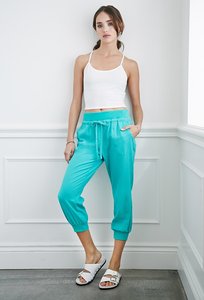}\hss\includegraphics[width=1.10cm,height=1.8cm,keepaspectratio]{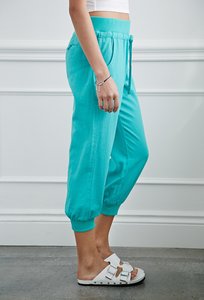}\hss\includegraphics[width=1.10cm,height=1.8cm,keepaspectratio]{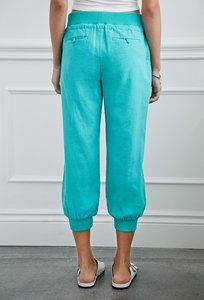}\hss\includegraphics[width=1.10cm,height=1.8cm,keepaspectratio]{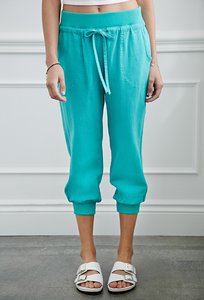}\hss\phantom{\rule{1.10cm}{1.8cm}}\hss}
      \end{minipage}
    }
  \end{minipage}
\hfill
  \begin{minipage}[t]{5.55cm}\end{minipage}
\hfill
  \begin{minipage}[t]{5.55cm}\end{minipage}
\par\vspace{2pt}
\noindent\rule{\linewidth}{0.4pt}
\par\vspace{2pt}
% -- reznembed --
{\small\textbf{RezNEmbed}}\quad{\small Rank~4}\\
\noindent
  \begin{minipage}[t]{5.55cm}
    \fcolorbox{red!70!black}{white}{
      \begin{minipage}[t]{5.25cm}
        {\tiny\textbf{\#1}}\\
        \noindent\hbox to 5.25cm{\includegraphics[width=1.10cm,height=1.8cm,keepaspectratio]{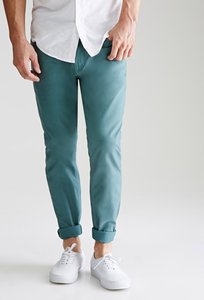}\hss\includegraphics[width=1.10cm,height=1.8cm,keepaspectratio]{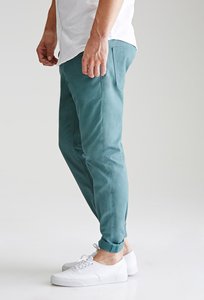}\hss\includegraphics[width=1.10cm,height=1.8cm,keepaspectratio]{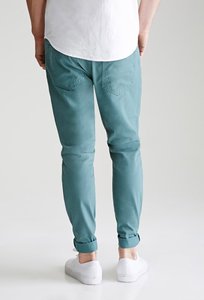}\hss\includegraphics[width=1.10cm,height=1.8cm,keepaspectratio]{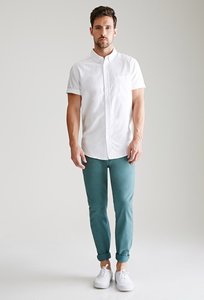}\hss\phantom{\rule{1.10cm}{1.8cm}}\hss}
      \end{minipage}
    }
  \end{minipage}
\hfill
  \begin{minipage}[t]{5.55cm}
    \fcolorbox{red!70!black}{white}{
      \begin{minipage}[t]{5.25cm}
        {\tiny\textbf{\#2}}\\
        \noindent\hbox to 5.25cm{\includegraphics[width=1.10cm,height=1.8cm,keepaspectratio]{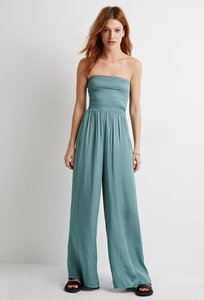}\hss\includegraphics[width=1.10cm,height=1.8cm,keepaspectratio]{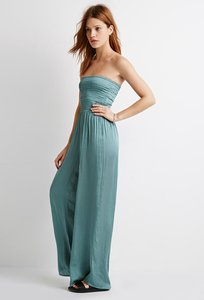}\hss\includegraphics[width=1.10cm,height=1.8cm,keepaspectratio]{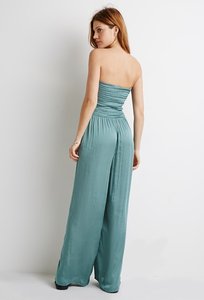}\hss\includegraphics[width=1.10cm,height=1.8cm,keepaspectratio]{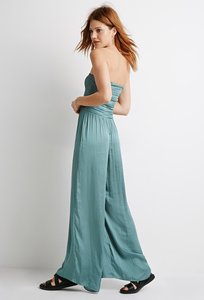}\hss\phantom{\rule{1.10cm}{1.8cm}}\hss}
      \end{minipage}
    }
  \end{minipage}
\hfill
  \begin{minipage}[t]{5.55cm}
    \fcolorbox{red!70!black}{white}{
      \begin{minipage}[t]{5.25cm}
        {\tiny\textbf{\#3}}\\
        \noindent\hbox to 5.25cm{\includegraphics[width=1.10cm,height=1.8cm,keepaspectratio]{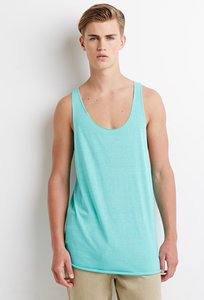}\hss\includegraphics[width=1.10cm,height=1.8cm,keepaspectratio]{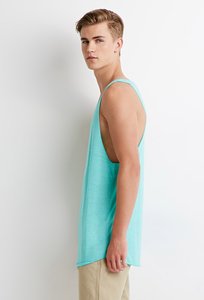}\hss\includegraphics[width=1.10cm,height=1.8cm,keepaspectratio]{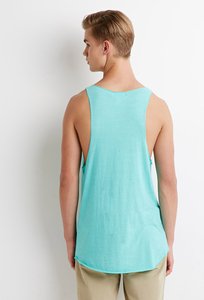}\hss\phantom{\rule{1.10cm}{1.8cm}}\hss\phantom{\rule{1.10cm}{1.8cm}}\hss}
      \end{minipage}
    }
  \end{minipage}
\par\vspace{2pt}
\noindent
  \begin{minipage}[t]{5.55cm}
    \fcolorbox{green!60!black}{white}{
      \begin{minipage}[t]{5.25cm}
        {\tiny\textbf{\#4}}\\
        \noindent\hbox to 5.25cm{\includegraphics[width=1.10cm,height=1.8cm,keepaspectratio]{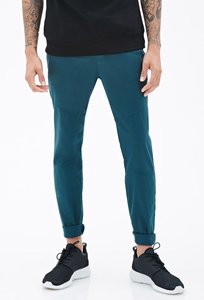}\hss\includegraphics[width=1.10cm,height=1.8cm,keepaspectratio]{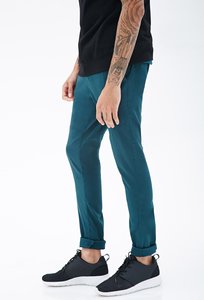}\hss\includegraphics[width=1.10cm,height=1.8cm,keepaspectratio]{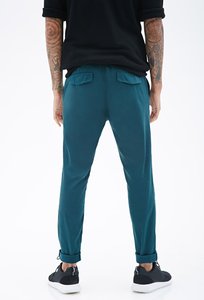}\hss\includegraphics[width=1.10cm,height=1.8cm,keepaspectratio]{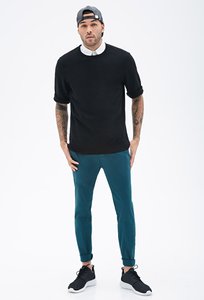}\hss\phantom{\rule{1.10cm}{1.8cm}}\hss}
      \end{minipage}
    }
  \end{minipage}
\hfill
  \begin{minipage}[t]{5.55cm}
    \fcolorbox{red!70!black}{white}{
      \begin{minipage}[t]{5.25cm}
        {\tiny\textbf{\#5}}\\
        \noindent\hbox to 5.25cm{\includegraphics[width=1.10cm,height=1.8cm,keepaspectratio]{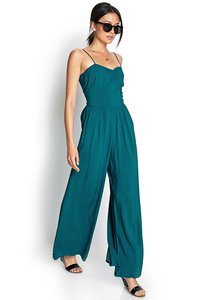}\hss\includegraphics[width=1.10cm,height=1.8cm,keepaspectratio]{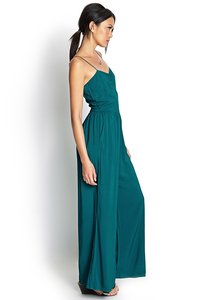}\hss\includegraphics[width=1.10cm,height=1.8cm,keepaspectratio]{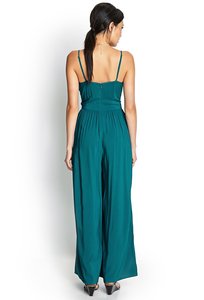}\hss\includegraphics[width=1.10cm,height=1.8cm,keepaspectratio]{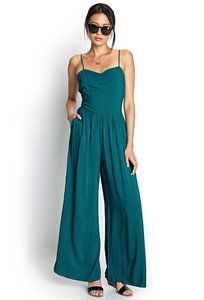}\hss\phantom{\rule{1.10cm}{1.8cm}}\hss}
      \end{minipage}
    }
  \end{minipage}
\hfill
  \begin{minipage}[t]{5.55cm}
    \fcolorbox{red!70!black}{white}{
      \begin{minipage}[t]{5.25cm}
        {\tiny\textbf{\#6}}\\
        \noindent\hbox to 5.25cm{\includegraphics[width=1.10cm,height=1.8cm,keepaspectratio]{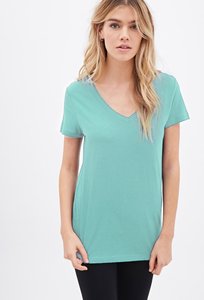}\hss\includegraphics[width=1.10cm,height=1.8cm,keepaspectratio]{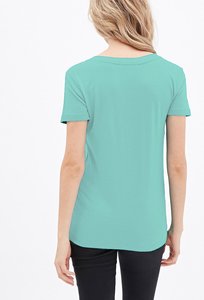}\hss\includegraphics[width=1.10cm,height=1.8cm,keepaspectratio]{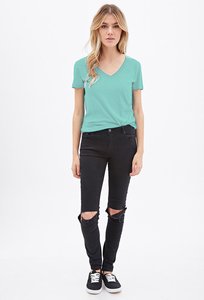}\hss\includegraphics[width=1.10cm,height=1.8cm,keepaspectratio]{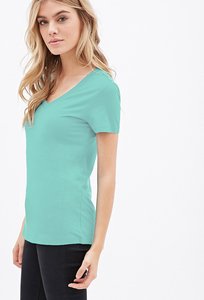}\hss\phantom{\rule{1.10cm}{1.8cm}}\hss}
      \end{minipage}
    }
  \end{minipage}
\par\vspace{2pt}
\noindent
  \begin{minipage}[t]{5.55cm}
    \fcolorbox{red!70!black}{white}{
      \begin{minipage}[t]{5.25cm}
        {\tiny\textbf{\#7}}\\
        \noindent\hbox to 5.25cm{\includegraphics[width=1.10cm,height=1.8cm,keepaspectratio]{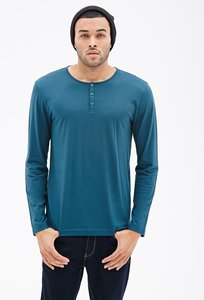}\hss\includegraphics[width=1.10cm,height=1.8cm,keepaspectratio]{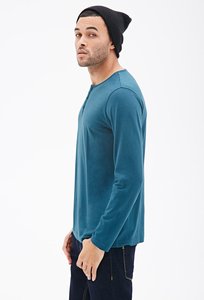}\hss\includegraphics[width=1.10cm,height=1.8cm,keepaspectratio]{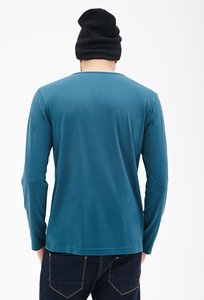}\hss\includegraphics[width=1.10cm,height=1.8cm,keepaspectratio]{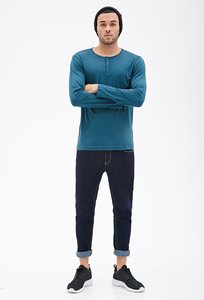}\hss\includegraphics[width=1.10cm,height=1.8cm,keepaspectratio]{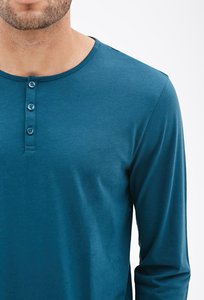}\hss}
      \end{minipage}
    }
  \end{minipage}
\hfill
  \begin{minipage}[t]{5.55cm}
    \fcolorbox{red!70!black}{white}{
      \begin{minipage}[t]{5.25cm}
        {\tiny\textbf{\#8}}\\
        \noindent\hbox to 5.25cm{\includegraphics[width=1.10cm,height=1.8cm,keepaspectratio]{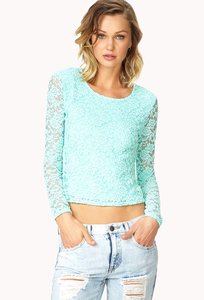}\hss\includegraphics[width=1.10cm,height=1.8cm,keepaspectratio]{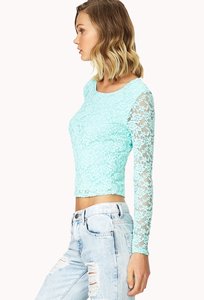}\hss\includegraphics[width=1.10cm,height=1.8cm,keepaspectratio]{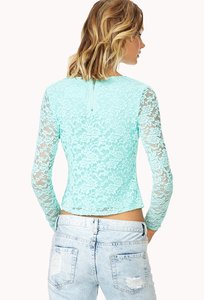}\hss\includegraphics[width=1.10cm,height=1.8cm,keepaspectratio]{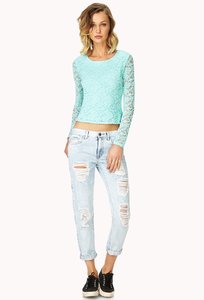}\hss\phantom{\rule{1.10cm}{1.8cm}}\hss}
      \end{minipage}
    }
  \end{minipage}
\hfill
  \begin{minipage}[t]{5.55cm}
    \fcolorbox{red!70!black}{white}{
      \begin{minipage}[t]{5.25cm}
        {\tiny\textbf{\#9}}\\
        \noindent\hbox to 5.25cm{\includegraphics[width=1.10cm,height=1.8cm,keepaspectratio]{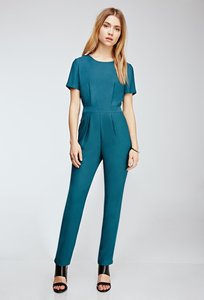}\hss\includegraphics[width=1.10cm,height=1.8cm,keepaspectratio]{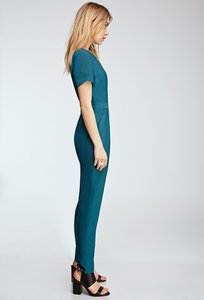}\hss\includegraphics[width=1.10cm,height=1.8cm,keepaspectratio]{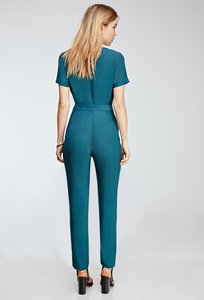}\hss\phantom{\rule{1.10cm}{1.8cm}}\hss\phantom{\rule{1.10cm}{1.8cm}}\hss}
      \end{minipage}
    }
  \end{minipage}
\par\vspace{2pt}
\noindent
  \begin{minipage}[t]{5.55cm}
    \fcolorbox{red!70!black}{white}{
      \begin{minipage}[t]{5.25cm}
        {\tiny\textbf{\#10}}\\
        \noindent\hbox to 5.25cm{\includegraphics[width=1.10cm,height=1.8cm,keepaspectratio]{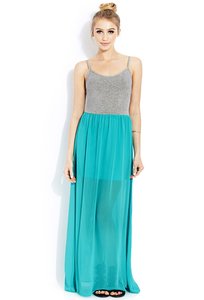}\hss\includegraphics[width=1.10cm,height=1.8cm,keepaspectratio]{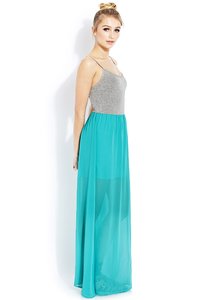}\hss\includegraphics[width=1.10cm,height=1.8cm,keepaspectratio]{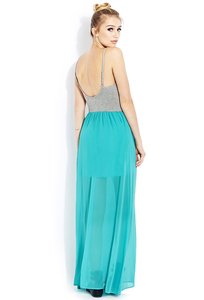}\hss\includegraphics[width=1.10cm,height=1.8cm,keepaspectratio]{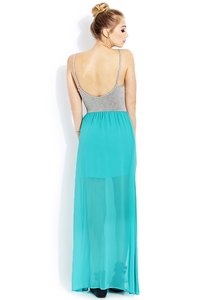}\hss\includegraphics[width=1.10cm,height=1.8cm,keepaspectratio]{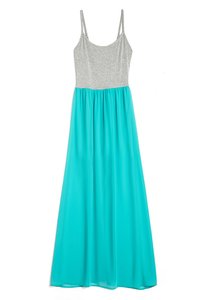}\hss}
      \end{minipage}
    }
  \end{minipage}
\hfill
  \begin{minipage}[t]{5.55cm}\end{minipage}
\hfill
  \begin{minipage}[t]{5.55cm}\end{minipage}
\par\vspace{2pt}
\noindent\rule{\linewidth}{0.4pt}
\par\vspace{2pt}
% -- doubao --
{\small\textbf{Doubao-E-V}}\quad{\small Rank~5}\\
\noindent
  \begin{minipage}[t]{5.55cm}
    \fcolorbox{red!70!black}{white}{
      \begin{minipage}[t]{5.25cm}
        {\tiny\textbf{\#1}}\\
        \noindent\hbox to 5.25cm{\includegraphics[width=1.10cm,height=1.8cm,keepaspectratio]{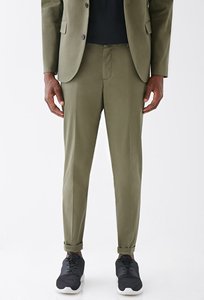}\hss\includegraphics[width=1.10cm,height=1.8cm,keepaspectratio]{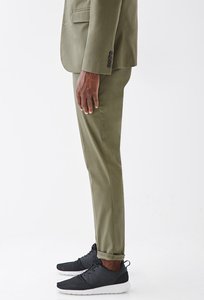}\hss\includegraphics[width=1.10cm,height=1.8cm,keepaspectratio]{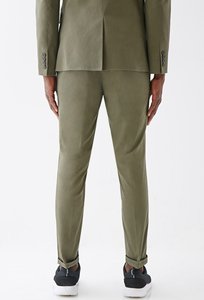}\hss\includegraphics[width=1.10cm,height=1.8cm,keepaspectratio]{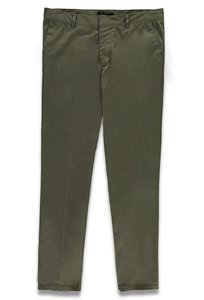}\hss\includegraphics[width=1.10cm,height=1.8cm,keepaspectratio]{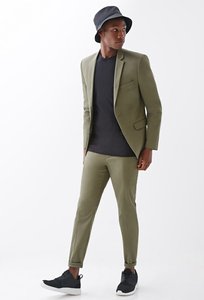}\hss}
      \end{minipage}
    }
  \end{minipage}
\hfill
  \begin{minipage}[t]{5.55cm}
    \fcolorbox{red!70!black}{white}{
      \begin{minipage}[t]{5.25cm}
        {\tiny\textbf{\#2}}\\
        \noindent\hbox to 5.25cm{\includegraphics[width=1.10cm,height=1.8cm,keepaspectratio]{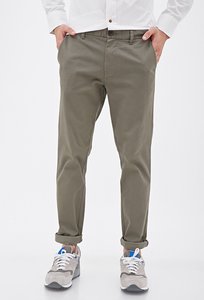}\hss\includegraphics[width=1.10cm,height=1.8cm,keepaspectratio]{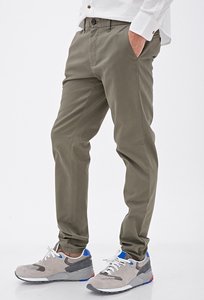}\hss\includegraphics[width=1.10cm,height=1.8cm,keepaspectratio]{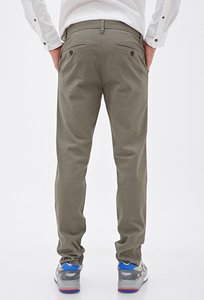}\hss\includegraphics[width=1.10cm,height=1.8cm,keepaspectratio]{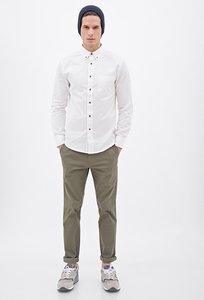}\hss\phantom{\rule{1.10cm}{1.8cm}}\hss}
      \end{minipage}
    }
  \end{minipage}
\hfill
  \begin{minipage}[t]{5.55cm}
    \fcolorbox{red!70!black}{white}{
      \begin{minipage}[t]{5.25cm}
        {\tiny\textbf{\#3}}\\
        \noindent\hbox to 5.25cm{\includegraphics[width=1.10cm,height=1.8cm,keepaspectratio]{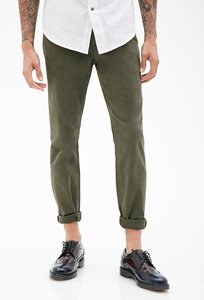}\hss\includegraphics[width=1.10cm,height=1.8cm,keepaspectratio]{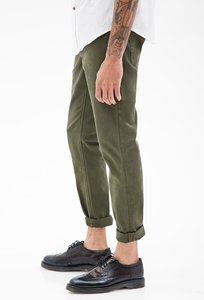}\hss\includegraphics[width=1.10cm,height=1.8cm,keepaspectratio]{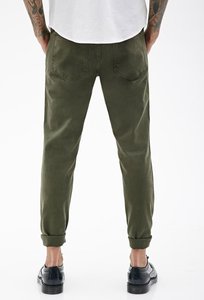}\hss\includegraphics[width=1.10cm,height=1.8cm,keepaspectratio]{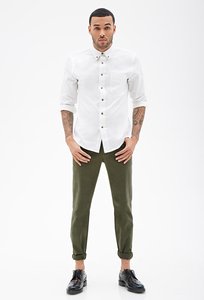}\hss\phantom{\rule{1.10cm}{1.8cm}}\hss}
      \end{minipage}
    }
  \end{minipage}
\par\vspace{2pt}
\noindent
  \begin{minipage}[t]{5.55cm}
    \fcolorbox{red!70!black}{white}{
      \begin{minipage}[t]{5.25cm}
        {\tiny\textbf{\#4}}\\
        \noindent\hbox to 5.25cm{\includegraphics[width=1.10cm,height=1.8cm,keepaspectratio]{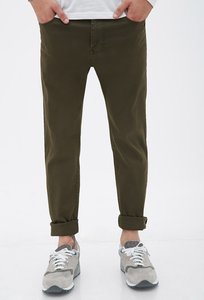}\hss\includegraphics[width=1.10cm,height=1.8cm,keepaspectratio]{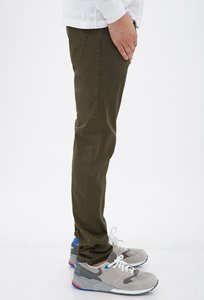}\hss\includegraphics[width=1.10cm,height=1.8cm,keepaspectratio]{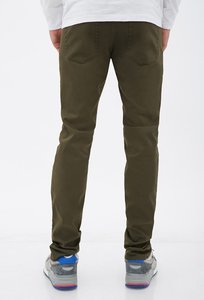}\hss\includegraphics[width=1.10cm,height=1.8cm,keepaspectratio]{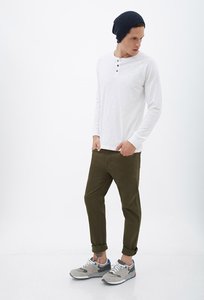}\hss\phantom{\rule{1.10cm}{1.8cm}}\hss}
      \end{minipage}
    }
  \end{minipage}
\hfill
  \begin{minipage}[t]{5.55cm}
    \fcolorbox{green!60!black}{white}{
      \begin{minipage}[t]{5.25cm}
        {\tiny\textbf{\#5}}\\
        \noindent\hbox to 5.25cm{\includegraphics[width=1.10cm,height=1.8cm,keepaspectratio]{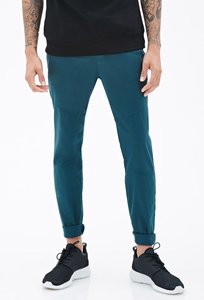}\hss\includegraphics[width=1.10cm,height=1.8cm,keepaspectratio]{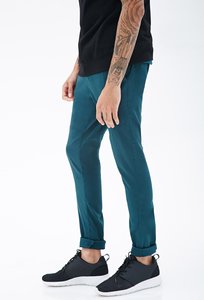}\hss\includegraphics[width=1.10cm,height=1.8cm,keepaspectratio]{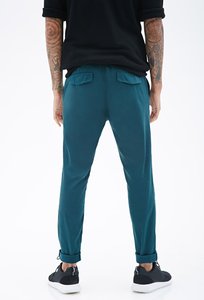}\hss\includegraphics[width=1.10cm,height=1.8cm,keepaspectratio]{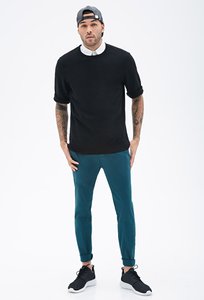}\hss\phantom{\rule{1.10cm}{1.8cm}}\hss}
      \end{minipage}
    }
  \end{minipage}
\hfill
  \begin{minipage}[t]{5.55cm}
    \fcolorbox{red!70!black}{white}{
      \begin{minipage}[t]{5.25cm}
        {\tiny\textbf{\#6}}\\
        \noindent\hbox to 5.25cm{\includegraphics[width=1.10cm,height=1.8cm,keepaspectratio]{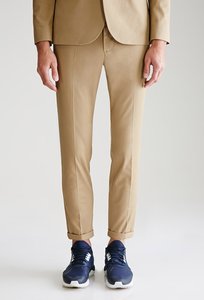}\hss\includegraphics[width=1.10cm,height=1.8cm,keepaspectratio]{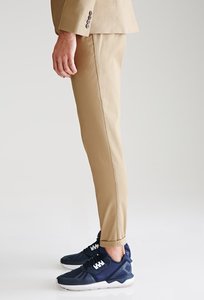}\hss\includegraphics[width=1.10cm,height=1.8cm,keepaspectratio]{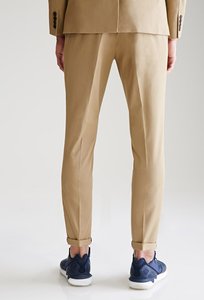}\hss\includegraphics[width=1.10cm,height=1.8cm,keepaspectratio]{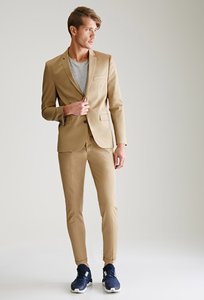}\hss\phantom{\rule{1.10cm}{1.8cm}}\hss}
      \end{minipage}
    }
  \end{minipage}
\par\vspace{2pt}
\noindent
  \begin{minipage}[t]{5.55cm}
    \fcolorbox{red!70!black}{white}{
      \begin{minipage}[t]{5.25cm}
        {\tiny\textbf{\#7}}\\
        \noindent\hbox to 5.25cm{\includegraphics[width=1.10cm,height=1.8cm,keepaspectratio]{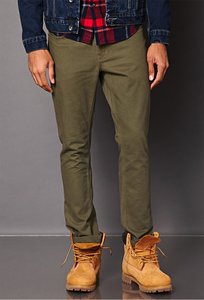}\hss\includegraphics[width=1.10cm,height=1.8cm,keepaspectratio]{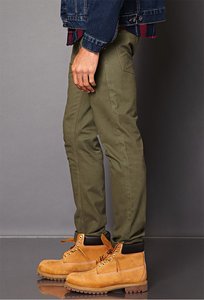}\hss\includegraphics[width=1.10cm,height=1.8cm,keepaspectratio]{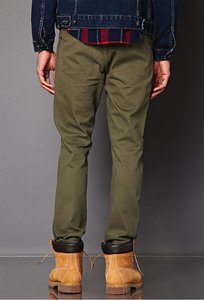}\hss\includegraphics[width=1.10cm,height=1.8cm,keepaspectratio]{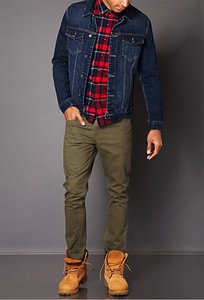}\hss\includegraphics[width=1.10cm,height=1.8cm,keepaspectratio]{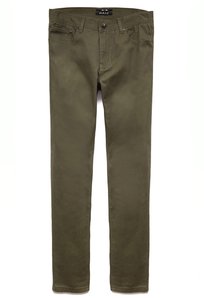}\hss}
      \end{minipage}
    }
  \end{minipage}
\hfill
  \begin{minipage}[t]{5.55cm}
    \fcolorbox{red!70!black}{white}{
      \begin{minipage}[t]{5.25cm}
        {\tiny\textbf{\#8}}\\
        \noindent\hbox to 5.25cm{\includegraphics[width=1.10cm,height=1.8cm,keepaspectratio]{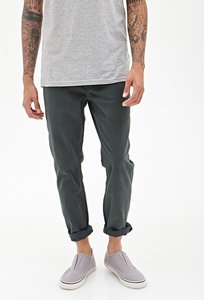}\hss\includegraphics[width=1.10cm,height=1.8cm,keepaspectratio]{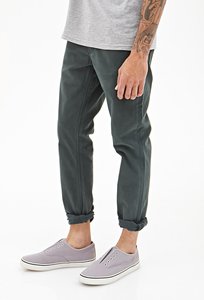}\hss\includegraphics[width=1.10cm,height=1.8cm,keepaspectratio]{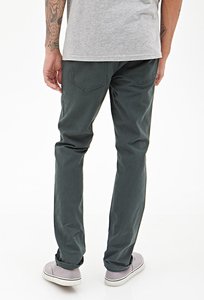}\hss\includegraphics[width=1.10cm,height=1.8cm,keepaspectratio]{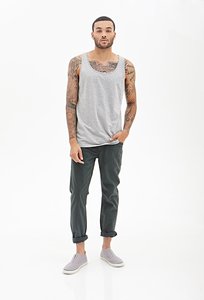}\hss\phantom{\rule{1.10cm}{1.8cm}}\hss}
      \end{minipage}
    }
  \end{minipage}
\hfill
  \begin{minipage}[t]{5.55cm}
    \fcolorbox{red!70!black}{white}{
      \begin{minipage}[t]{5.25cm}
        {\tiny\textbf{\#9}}\\
        \noindent\hbox to 5.25cm{\includegraphics[width=1.10cm,height=1.8cm,keepaspectratio]{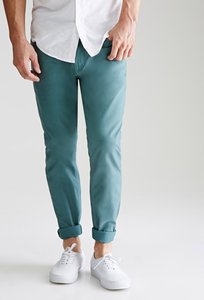}\hss\includegraphics[width=1.10cm,height=1.8cm,keepaspectratio]{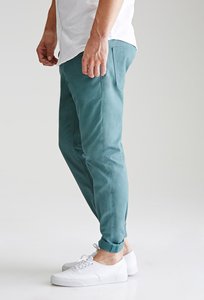}\hss\includegraphics[width=1.10cm,height=1.8cm,keepaspectratio]{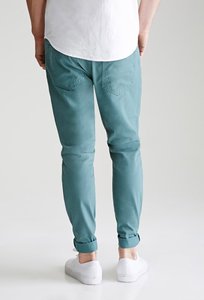}\hss\includegraphics[width=1.10cm,height=1.8cm,keepaspectratio]{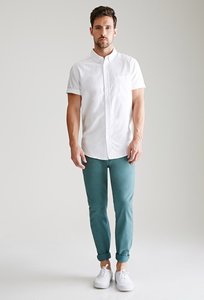}\hss\phantom{\rule{1.10cm}{1.8cm}}\hss}
      \end{minipage}
    }
  \end{minipage}
\par\vspace{2pt}
\noindent
  \begin{minipage}[t]{5.55cm}
    \fcolorbox{red!70!black}{white}{
      \begin{minipage}[t]{5.25cm}
        {\tiny\textbf{\#10}}\\
        \noindent\hbox to 5.25cm{\includegraphics[width=1.10cm,height=1.8cm,keepaspectratio]{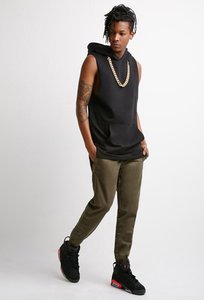}\hss\includegraphics[width=1.10cm,height=1.8cm,keepaspectratio]{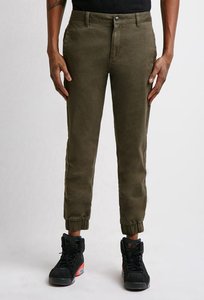}\hss\includegraphics[width=1.10cm,height=1.8cm,keepaspectratio]{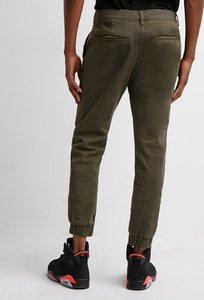}\hss\includegraphics[width=1.10cm,height=1.8cm,keepaspectratio]{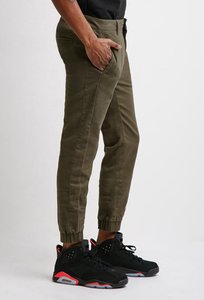}\hss\phantom{\rule{1.10cm}{1.8cm}}\hss}
      \end{minipage}
    }
  \end{minipage}
\hfill
  \begin{minipage}[t]{5.55cm}\end{minipage}
\hfill
  \begin{minipage}[t]{5.55cm}\end{minipage}
\par\vspace{2pt}
\noindent\rule{\linewidth}{0.4pt}
\par\vspace{20pt}

\par\vspace{16pt}
\noindent\textbf{\large Example 7}
\par\vspace{4pt}
\noindent\rule{\linewidth}{1.2pt}
\par\vspace{4pt}
% ── Case 7: short::deepfashion::3923 ──
\noindent\hfill%
  \begin{minipage}[t]{6.0cm}
    \noindent\hbox to 6.0cm{\includegraphics[width=1.20cm,height=2.0cm,keepaspectratio]{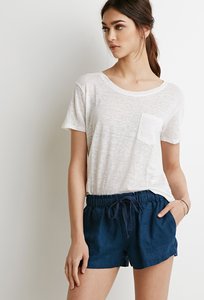}\hss\includegraphics[width=1.20cm,height=2.0cm,keepaspectratio]{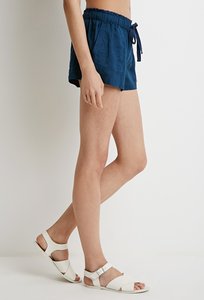}\hss\includegraphics[width=1.20cm,height=2.0cm,keepaspectratio]{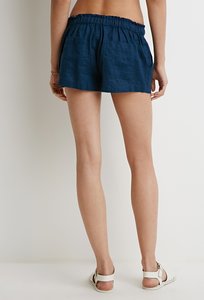}\hss\includegraphics[width=1.20cm,height=2.0cm,keepaspectratio]{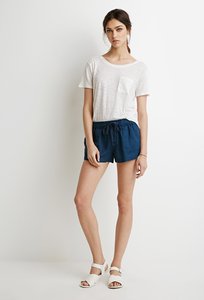}\hss\includegraphics[width=1.20cm,height=2.0cm,keepaspectratio]{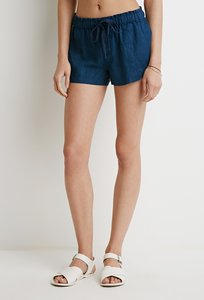}\hss}
    \par\vspace{1pt}
    {\scriptsize\textbf{}}
    \par\vspace{0pt}
    \parbox[t]{6.0cm}{\tiny\raggedright Navy blue linen-blend shorts with wide elastic waistband and functional drawstring tie. Features side pockets and dual back patch pockets. Relaxed fit with short inseam and clean-finished hem. Lightweight, breathable textured fabric perfect for casual summer and beach wear.}
  \end{minipage}%
\hfill%
  \begin{minipage}[t]{3.5cm}
    \centering
    \vspace{0.45cm}%
    \parbox{3.5cm}{\centering\tiny Change to black knit fabric with cuffed hems and front slash pockets; replace back patch pockets with horizontal welt pockets.}\\[2pt]
    {\normalsize$\longrightarrow$}
  \end{minipage}%
\hfill%
  \begin{minipage}[t]{6.0cm}
    \noindent\hbox to 6.0cm{\includegraphics[width=1.20cm,height=2.0cm,keepaspectratio]{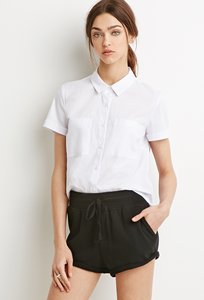}\hss\includegraphics[width=1.20cm,height=2.0cm,keepaspectratio]{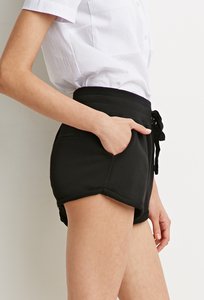}\hss\includegraphics[width=1.20cm,height=2.0cm,keepaspectratio]{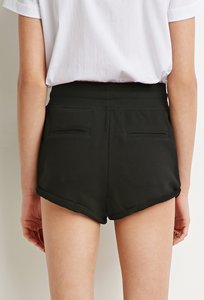}\hss\includegraphics[width=1.20cm,height=2.0cm,keepaspectratio]{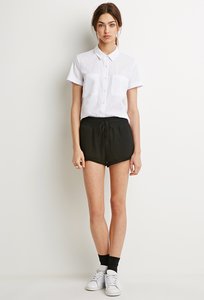}\hss\includegraphics[width=1.20cm,height=2.0cm,keepaspectratio]{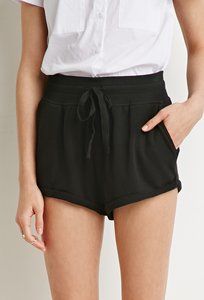}\hss}
    \par\vspace{1pt}
    {\scriptsize\textbf{Ground Truth}}
    \par\vspace{0pt}
    \parbox[t]{6.0cm}{\tiny\raggedright Black relaxed-fit knit shorts featuring a wide elastic drawstring waist, side slash pockets, back welt pockets, and cuffed hems. Constructed from soft, mid-weight fabric with a comfortable high-rise fit and sporty-casual aesthetic perfect for everyday loungewear.}
  \end{minipage}%
\hfill
\par\vspace{4pt}
\par\vspace{4pt}
\noindent\rule{\linewidth}{0.4pt}
% -- mt_align --
{\small\textbf{\textbf{Ours}}}\quad{\small Rank~2}\\
\noindent
  \begin{minipage}[t]{5.55cm}
    \fcolorbox{red!70!black}{white}{
      \begin{minipage}[t]{5.25cm}
        {\tiny\textbf{\#1}}\\
        \noindent\hbox to 5.25cm{\includegraphics[width=1.10cm,height=1.8cm,keepaspectratio]{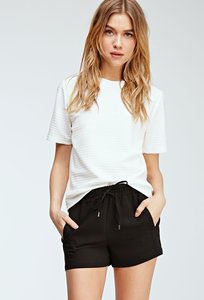}\hss\includegraphics[width=1.10cm,height=1.8cm,keepaspectratio]{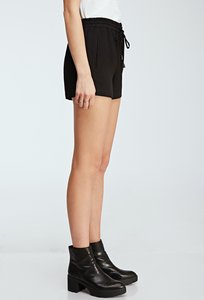}\hss\includegraphics[width=1.10cm,height=1.8cm,keepaspectratio]{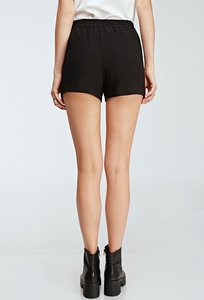}\hss\includegraphics[width=1.10cm,height=1.8cm,keepaspectratio]{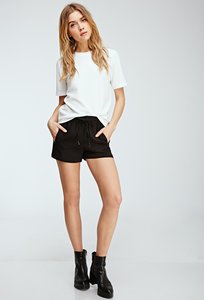}\hss\includegraphics[width=1.10cm,height=1.8cm,keepaspectratio]{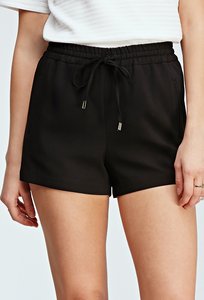}\hss}
      \end{minipage}
    }
  \end{minipage}
\hfill
  \begin{minipage}[t]{5.55cm}
    \fcolorbox{green!60!black}{white}{
      \begin{minipage}[t]{5.25cm}
        {\tiny\textbf{\#2}}\\
        \noindent\hbox to 5.25cm{\includegraphics[width=1.10cm,height=1.8cm,keepaspectratio]{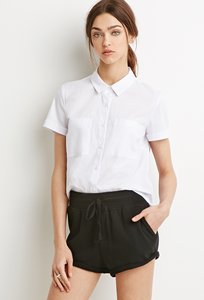}\hss\includegraphics[width=1.10cm,height=1.8cm,keepaspectratio]{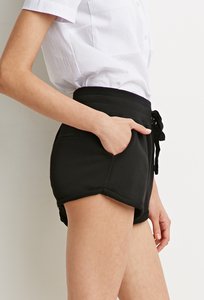}\hss\includegraphics[width=1.10cm,height=1.8cm,keepaspectratio]{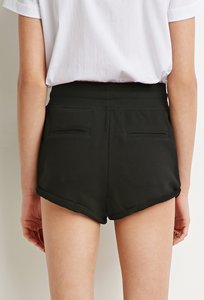}\hss\includegraphics[width=1.10cm,height=1.8cm,keepaspectratio]{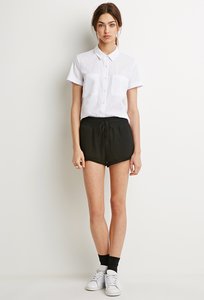}\hss\includegraphics[width=1.10cm,height=1.8cm,keepaspectratio]{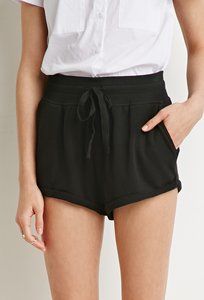}\hss}
      \end{minipage}
    }
  \end{minipage}
\hfill
  \begin{minipage}[t]{5.55cm}
    \fcolorbox{red!70!black}{white}{
      \begin{minipage}[t]{5.25cm}
        {\tiny\textbf{\#3}}\\
        \noindent\hbox to 5.25cm{\includegraphics[width=1.10cm,height=1.8cm,keepaspectratio]{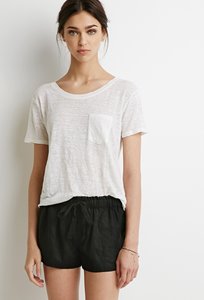}\hss\includegraphics[width=1.10cm,height=1.8cm,keepaspectratio]{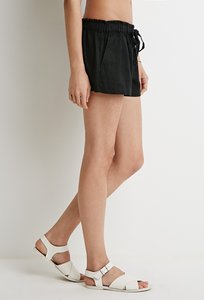}\hss\includegraphics[width=1.10cm,height=1.8cm,keepaspectratio]{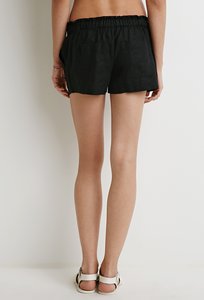}\hss\includegraphics[width=1.10cm,height=1.8cm,keepaspectratio]{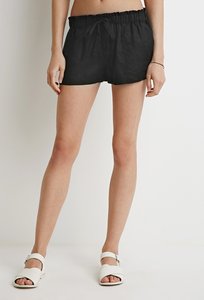}\hss\phantom{\rule{1.10cm}{1.8cm}}\hss}
      \end{minipage}
    }
  \end{minipage}
\par\vspace{2pt}
\noindent
  \begin{minipage}[t]{5.55cm}
    \fcolorbox{red!70!black}{white}{
      \begin{minipage}[t]{5.25cm}
        {\tiny\textbf{\#4}}\\
        \noindent\hbox to 5.25cm{\includegraphics[width=1.10cm,height=1.8cm,keepaspectratio]{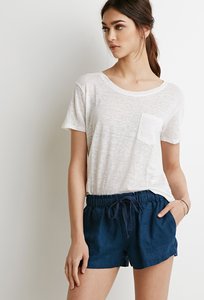}\hss\includegraphics[width=1.10cm,height=1.8cm,keepaspectratio]{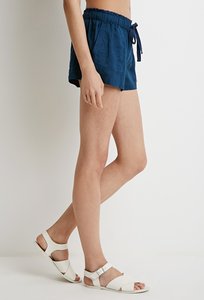}\hss\includegraphics[width=1.10cm,height=1.8cm,keepaspectratio]{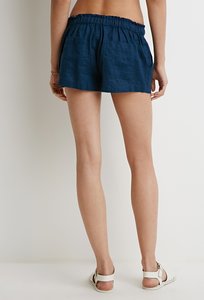}\hss\includegraphics[width=1.10cm,height=1.8cm,keepaspectratio]{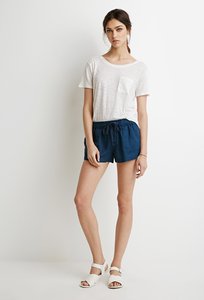}\hss\includegraphics[width=1.10cm,height=1.8cm,keepaspectratio]{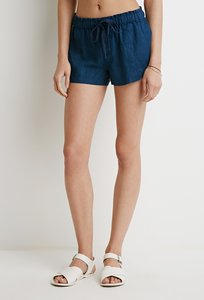}\hss}
      \end{minipage}
    }
  \end{minipage}
\hfill
  \begin{minipage}[t]{5.55cm}
    \fcolorbox{red!70!black}{white}{
      \begin{minipage}[t]{5.25cm}
        {\tiny\textbf{\#5}}\\
        \noindent\hbox to 5.25cm{\includegraphics[width=1.10cm,height=1.8cm,keepaspectratio]{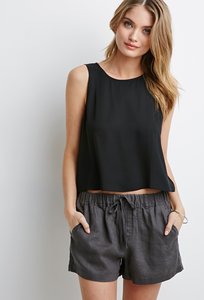}\hss\includegraphics[width=1.10cm,height=1.8cm,keepaspectratio]{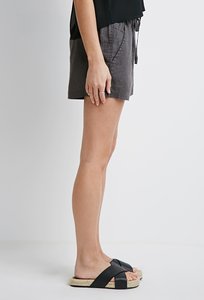}\hss\includegraphics[width=1.10cm,height=1.8cm,keepaspectratio]{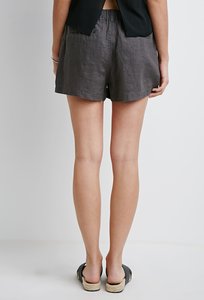}\hss\includegraphics[width=1.10cm,height=1.8cm,keepaspectratio]{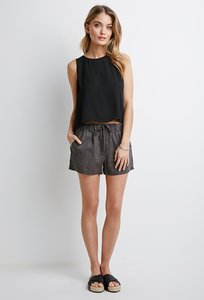}\hss\includegraphics[width=1.10cm,height=1.8cm,keepaspectratio]{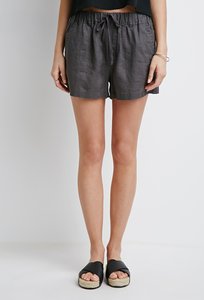}\hss}
      \end{minipage}
    }
  \end{minipage}
\hfill
  \begin{minipage}[t]{5.55cm}
    \fcolorbox{red!70!black}{white}{
      \begin{minipage}[t]{5.25cm}
        {\tiny\textbf{\#6}}\\
        \noindent\hbox to 5.25cm{\includegraphics[width=1.10cm,height=1.8cm,keepaspectratio]{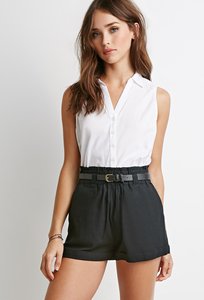}\hss\includegraphics[width=1.10cm,height=1.8cm,keepaspectratio]{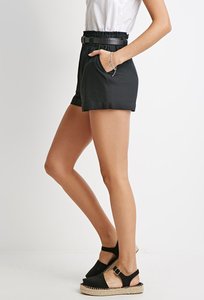}\hss\includegraphics[width=1.10cm,height=1.8cm,keepaspectratio]{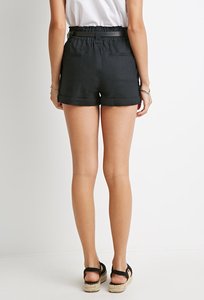}\hss\includegraphics[width=1.10cm,height=1.8cm,keepaspectratio]{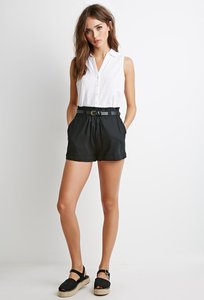}\hss\includegraphics[width=1.10cm,height=1.8cm,keepaspectratio]{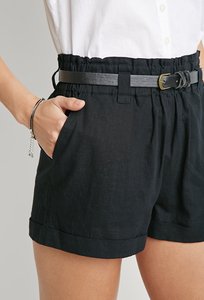}\hss}
      \end{minipage}
    }
  \end{minipage}
\par\vspace{2pt}
\noindent
  \begin{minipage}[t]{5.55cm}
    \fcolorbox{red!70!black}{white}{
      \begin{minipage}[t]{5.25cm}
        {\tiny\textbf{\#7}}\\
        \noindent\hbox to 5.25cm{\includegraphics[width=1.10cm,height=1.8cm,keepaspectratio]{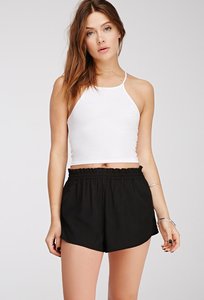}\hss\includegraphics[width=1.10cm,height=1.8cm,keepaspectratio]{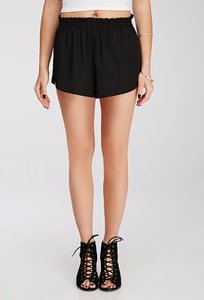}\hss\includegraphics[width=1.10cm,height=1.8cm,keepaspectratio]{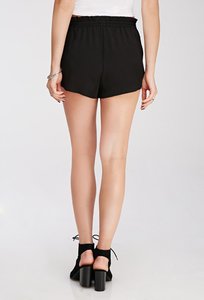}\hss\includegraphics[width=1.10cm,height=1.8cm,keepaspectratio]{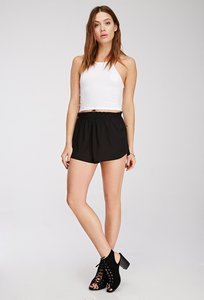}\hss\includegraphics[width=1.10cm,height=1.8cm,keepaspectratio]{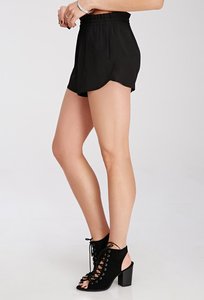}\hss}
      \end{minipage}
    }
  \end{minipage}
\hfill
  \begin{minipage}[t]{5.55cm}
    \fcolorbox{red!70!black}{white}{
      \begin{minipage}[t]{5.25cm}
        {\tiny\textbf{\#8}}\\
        \noindent\hbox to 5.25cm{\includegraphics[width=1.10cm,height=1.8cm,keepaspectratio]{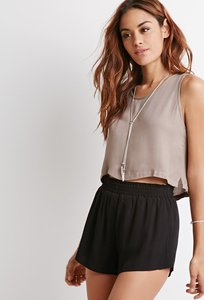}\hss\includegraphics[width=1.10cm,height=1.8cm,keepaspectratio]{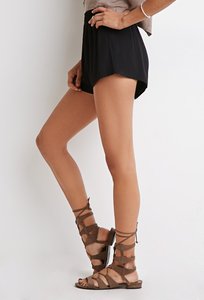}\hss\includegraphics[width=1.10cm,height=1.8cm,keepaspectratio]{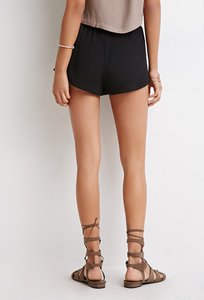}\hss\includegraphics[width=1.10cm,height=1.8cm,keepaspectratio]{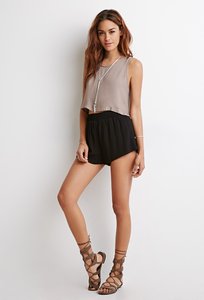}\hss\includegraphics[width=1.10cm,height=1.8cm,keepaspectratio]{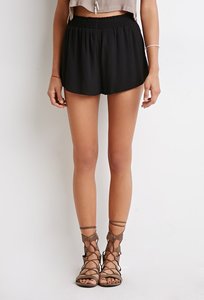}\hss}
      \end{minipage}
    }
  \end{minipage}
\hfill
  \begin{minipage}[t]{5.55cm}
    \fcolorbox{red!70!black}{white}{
      \begin{minipage}[t]{5.25cm}
        {\tiny\textbf{\#9}}\\
        \noindent\hbox to 5.25cm{\includegraphics[width=1.10cm,height=1.8cm,keepaspectratio]{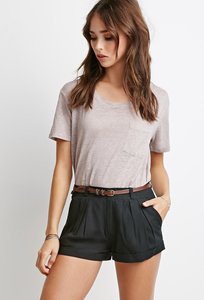}\hss\includegraphics[width=1.10cm,height=1.8cm,keepaspectratio]{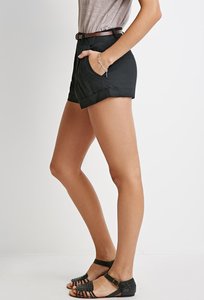}\hss\includegraphics[width=1.10cm,height=1.8cm,keepaspectratio]{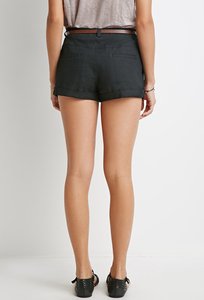}\hss\includegraphics[width=1.10cm,height=1.8cm,keepaspectratio]{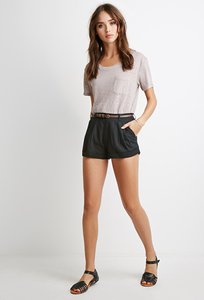}\hss\includegraphics[width=1.10cm,height=1.8cm,keepaspectratio]{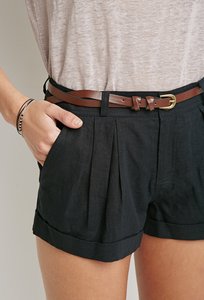}\hss}
      \end{minipage}
    }
  \end{minipage}
\par\vspace{2pt}
\noindent
  \begin{minipage}[t]{5.55cm}
    \fcolorbox{red!70!black}{white}{
      \begin{minipage}[t]{5.25cm}
        {\tiny\textbf{\#10}}\\
        \noindent\hbox to 5.25cm{\includegraphics[width=1.10cm,height=1.8cm,keepaspectratio]{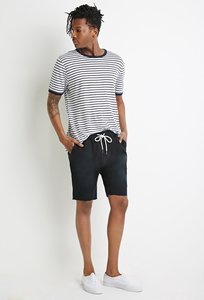}\hss\includegraphics[width=1.10cm,height=1.8cm,keepaspectratio]{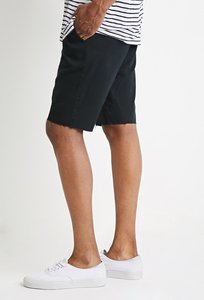}\hss\includegraphics[width=1.10cm,height=1.8cm,keepaspectratio]{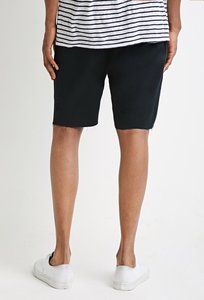}\hss\includegraphics[width=1.10cm,height=1.8cm,keepaspectratio]{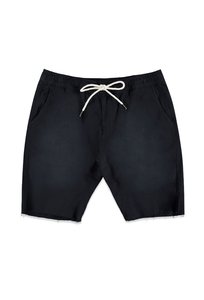}\hss\includegraphics[width=1.10cm,height=1.8cm,keepaspectratio]{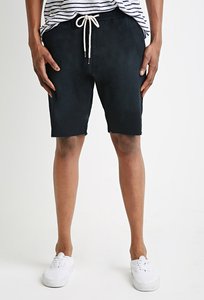}\hss}
      \end{minipage}
    }
  \end{minipage}
\hfill
  \begin{minipage}[t]{5.55cm}\end{minipage}
\hfill
  \begin{minipage}[t]{5.55cm}\end{minipage}
\par\vspace{2pt}
\noindent\rule{\linewidth}{0.4pt}
\par\vspace{2pt}
% -- qwen3_vl_2b --
{\small\textbf{Qwen3-VL-2B}}\quad{\small Rank~5}\\
\noindent
  \begin{minipage}[t]{5.55cm}
    \fcolorbox{red!70!black}{white}{
      \begin{minipage}[t]{5.25cm}
        {\tiny\textbf{\#1}}\\
        \noindent\hbox to 5.25cm{\includegraphics[width=1.10cm,height=1.8cm,keepaspectratio]{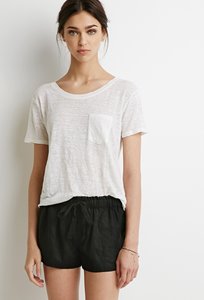}\hss\includegraphics[width=1.10cm,height=1.8cm,keepaspectratio]{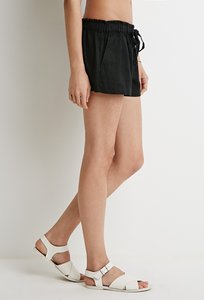}\hss\includegraphics[width=1.10cm,height=1.8cm,keepaspectratio]{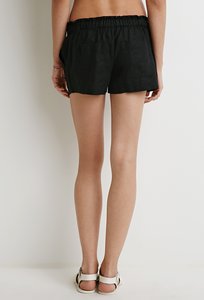}\hss\includegraphics[width=1.10cm,height=1.8cm,keepaspectratio]{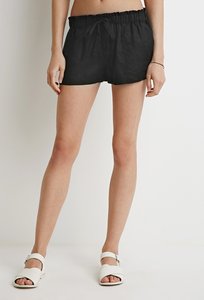}\hss\phantom{\rule{1.10cm}{1.8cm}}\hss}
      \end{minipage}
    }
  \end{minipage}
\hfill
  \begin{minipage}[t]{5.55cm}
    \fcolorbox{red!70!black}{white}{
      \begin{minipage}[t]{5.25cm}
        {\tiny\textbf{\#2}}\\
        \noindent\hbox to 5.25cm{\includegraphics[width=1.10cm,height=1.8cm,keepaspectratio]{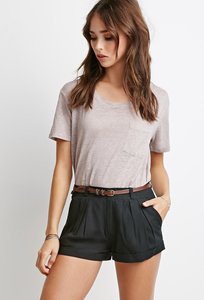}\hss\includegraphics[width=1.10cm,height=1.8cm,keepaspectratio]{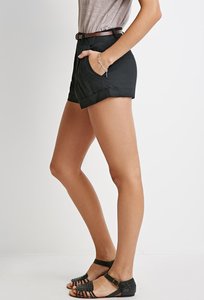}\hss\includegraphics[width=1.10cm,height=1.8cm,keepaspectratio]{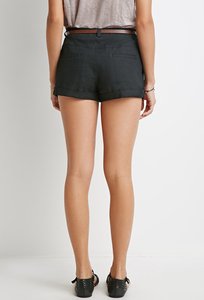}\hss\includegraphics[width=1.10cm,height=1.8cm,keepaspectratio]{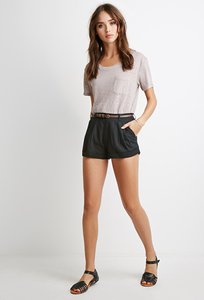}\hss\includegraphics[width=1.10cm,height=1.8cm,keepaspectratio]{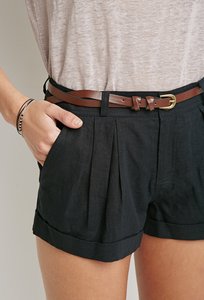}\hss}
      \end{minipage}
    }
  \end{minipage}
\hfill
  \begin{minipage}[t]{5.55cm}
    \fcolorbox{red!70!black}{white}{
      \begin{minipage}[t]{5.25cm}
        {\tiny\textbf{\#3}}\\
        \noindent\hbox to 5.25cm{\includegraphics[width=1.10cm,height=1.8cm,keepaspectratio]{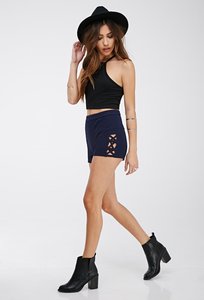}\hss\includegraphics[width=1.10cm,height=1.8cm,keepaspectratio]{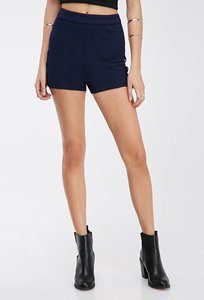}\hss\includegraphics[width=1.10cm,height=1.8cm,keepaspectratio]{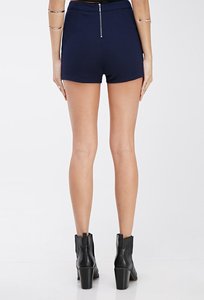}\hss\includegraphics[width=1.10cm,height=1.8cm,keepaspectratio]{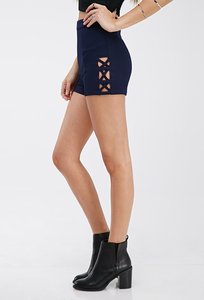}\hss\includegraphics[width=1.10cm,height=1.8cm,keepaspectratio]{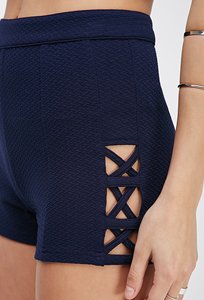}\hss}
      \end{minipage}
    }
  \end{minipage}
\par\vspace{2pt}
\noindent
  \begin{minipage}[t]{5.55cm}
    \fcolorbox{red!70!black}{white}{
      \begin{minipage}[t]{5.25cm}
        {\tiny\textbf{\#4}}\\
        \noindent\hbox to 5.25cm{\includegraphics[width=1.10cm,height=1.8cm,keepaspectratio]{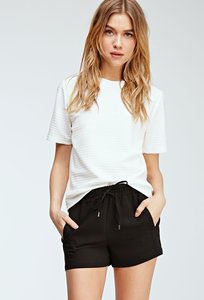}\hss\includegraphics[width=1.10cm,height=1.8cm,keepaspectratio]{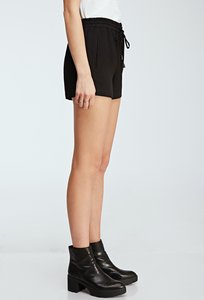}\hss\includegraphics[width=1.10cm,height=1.8cm,keepaspectratio]{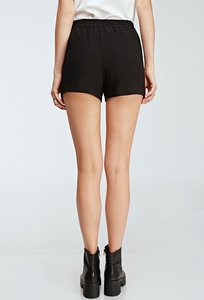}\hss\includegraphics[width=1.10cm,height=1.8cm,keepaspectratio]{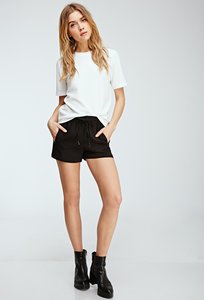}\hss\includegraphics[width=1.10cm,height=1.8cm,keepaspectratio]{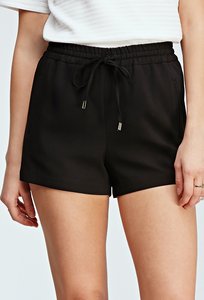}\hss}
      \end{minipage}
    }
  \end{minipage}
\hfill
  \begin{minipage}[t]{5.55cm}
    \fcolorbox{green!60!black}{white}{
      \begin{minipage}[t]{5.25cm}
        {\tiny\textbf{\#5}}\\
        \noindent\hbox to 5.25cm{\includegraphics[width=1.10cm,height=1.8cm,keepaspectratio]{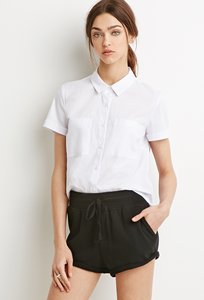}\hss\includegraphics[width=1.10cm,height=1.8cm,keepaspectratio]{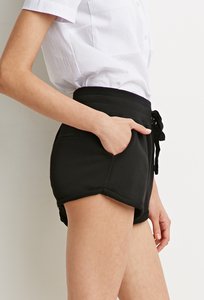}\hss\includegraphics[width=1.10cm,height=1.8cm,keepaspectratio]{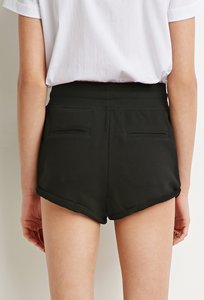}\hss\includegraphics[width=1.10cm,height=1.8cm,keepaspectratio]{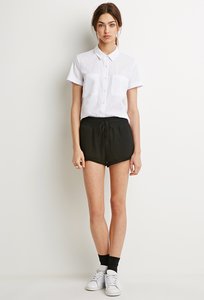}\hss\includegraphics[width=1.10cm,height=1.8cm,keepaspectratio]{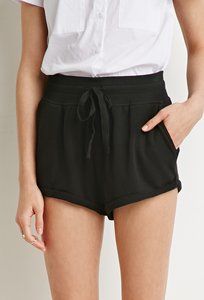}\hss}
      \end{minipage}
    }
  \end{minipage}
\hfill
  \begin{minipage}[t]{5.55cm}
    \fcolorbox{red!70!black}{white}{
      \begin{minipage}[t]{5.25cm}
        {\tiny\textbf{\#6}}\\
        \noindent\hbox to 5.25cm{\includegraphics[width=1.10cm,height=1.8cm,keepaspectratio]{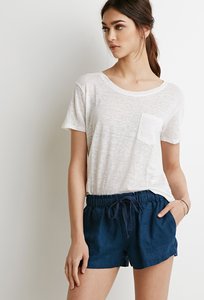}\hss\includegraphics[width=1.10cm,height=1.8cm,keepaspectratio]{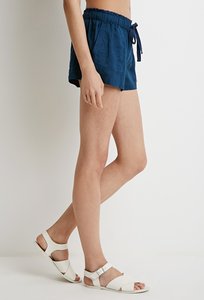}\hss\includegraphics[width=1.10cm,height=1.8cm,keepaspectratio]{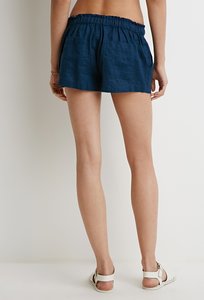}\hss\includegraphics[width=1.10cm,height=1.8cm,keepaspectratio]{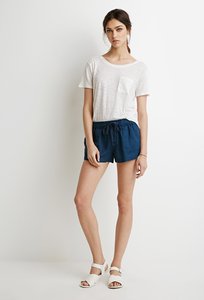}\hss\includegraphics[width=1.10cm,height=1.8cm,keepaspectratio]{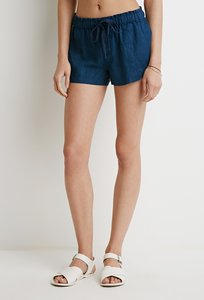}\hss}
      \end{minipage}
    }
  \end{minipage}
\par\vspace{2pt}
\noindent
  \begin{minipage}[t]{5.55cm}
    \fcolorbox{red!70!black}{white}{
      \begin{minipage}[t]{5.25cm}
        {\tiny\textbf{\#7}}\\
        \noindent\hbox to 5.25cm{\includegraphics[width=1.10cm,height=1.8cm,keepaspectratio]{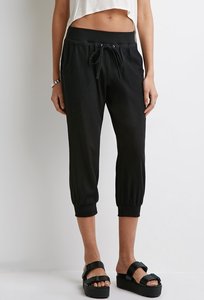}\hss\includegraphics[width=1.10cm,height=1.8cm,keepaspectratio]{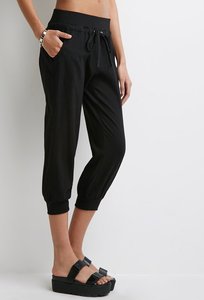}\hss\includegraphics[width=1.10cm,height=1.8cm,keepaspectratio]{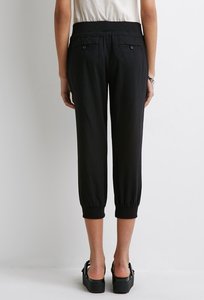}\hss\includegraphics[width=1.10cm,height=1.8cm,keepaspectratio]{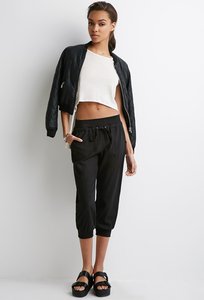}\hss\phantom{\rule{1.10cm}{1.8cm}}\hss}
      \end{minipage}
    }
  \end{minipage}
\hfill
  \begin{minipage}[t]{5.55cm}
    \fcolorbox{red!70!black}{white}{
      \begin{minipage}[t]{5.25cm}
        {\tiny\textbf{\#8}}\\
        \noindent\hbox to 5.25cm{\includegraphics[width=1.10cm,height=1.8cm,keepaspectratio]{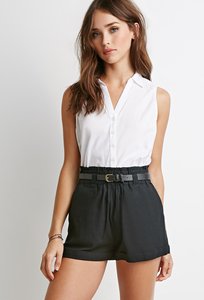}\hss\includegraphics[width=1.10cm,height=1.8cm,keepaspectratio]{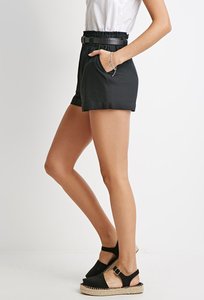}\hss\includegraphics[width=1.10cm,height=1.8cm,keepaspectratio]{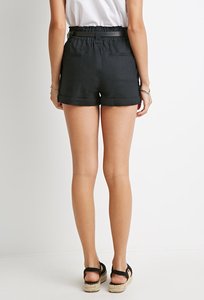}\hss\includegraphics[width=1.10cm,height=1.8cm,keepaspectratio]{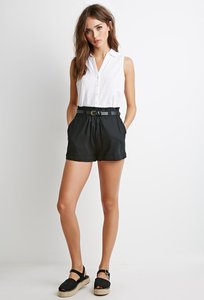}\hss\includegraphics[width=1.10cm,height=1.8cm,keepaspectratio]{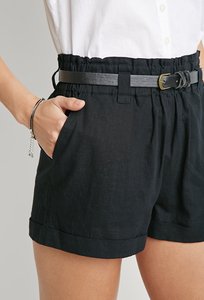}\hss}
      \end{minipage}
    }
  \end{minipage}
\hfill
  \begin{minipage}[t]{5.55cm}
    \fcolorbox{red!70!black}{white}{
      \begin{minipage}[t]{5.25cm}
        {\tiny\textbf{\#9}}\\
        \noindent\hbox to 5.25cm{\includegraphics[width=1.10cm,height=1.8cm,keepaspectratio]{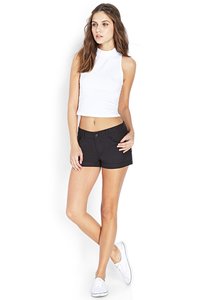}\hss\includegraphics[width=1.10cm,height=1.8cm,keepaspectratio]{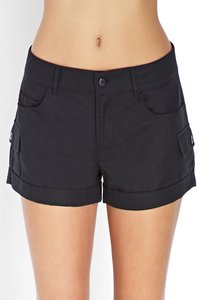}\hss\phantom{\rule{1.10cm}{1.8cm}}\hss\phantom{\rule{1.10cm}{1.8cm}}\hss\phantom{\rule{1.10cm}{1.8cm}}\hss}
      \end{minipage}
    }
  \end{minipage}
\par\vspace{2pt}
\noindent
  \begin{minipage}[t]{5.55cm}
    \fcolorbox{red!70!black}{white}{
      \begin{minipage}[t]{5.25cm}
        {\tiny\textbf{\#10}}\\
        \noindent\hbox to 5.25cm{\includegraphics[width=1.10cm,height=1.8cm,keepaspectratio]{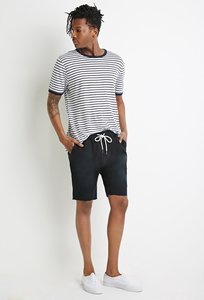}\hss\includegraphics[width=1.10cm,height=1.8cm,keepaspectratio]{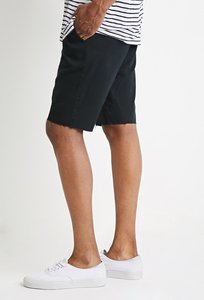}\hss\includegraphics[width=1.10cm,height=1.8cm,keepaspectratio]{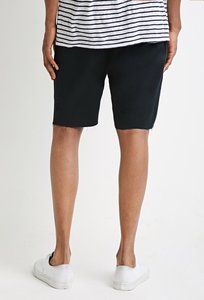}\hss\includegraphics[width=1.10cm,height=1.8cm,keepaspectratio]{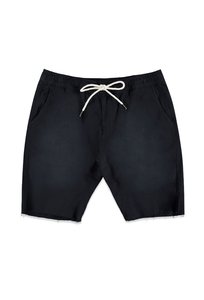}\hss\includegraphics[width=1.10cm,height=1.8cm,keepaspectratio]{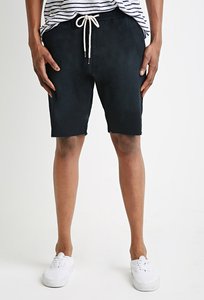}\hss}
      \end{minipage}
    }
  \end{minipage}
\hfill
  \begin{minipage}[t]{5.55cm}\end{minipage}
\hfill
  \begin{minipage}[t]{5.55cm}\end{minipage}
\par\vspace{2pt}
\noindent\rule{\linewidth}{0.4pt}
\par\vspace{2pt}
% -- qwen3_vl_8b --
{\small\textbf{Qwen3-VL-8B}}\quad{\small Rank~3}\\
\noindent
  \begin{minipage}[t]{5.55cm}
    \fcolorbox{red!70!black}{white}{
      \begin{minipage}[t]{5.25cm}
        {\tiny\textbf{\#1}}\\
        \noindent\hbox to 5.25cm{\includegraphics[width=1.10cm,height=1.8cm,keepaspectratio]{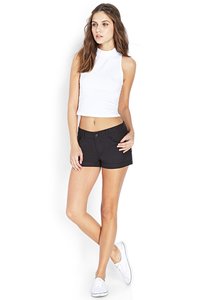}\hss\includegraphics[width=1.10cm,height=1.8cm,keepaspectratio]{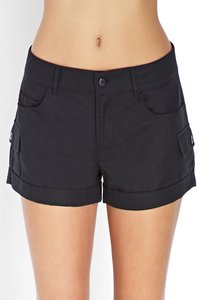}\hss\phantom{\rule{1.10cm}{1.8cm}}\hss\phantom{\rule{1.10cm}{1.8cm}}\hss\phantom{\rule{1.10cm}{1.8cm}}\hss}
      \end{minipage}
    }
  \end{minipage}
\hfill
  \begin{minipage}[t]{5.55cm}
    \fcolorbox{red!70!black}{white}{
      \begin{minipage}[t]{5.25cm}
        {\tiny\textbf{\#2}}\\
        \noindent\hbox to 5.25cm{\includegraphics[width=1.10cm,height=1.8cm,keepaspectratio]{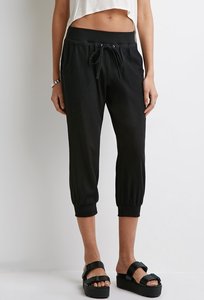}\hss\includegraphics[width=1.10cm,height=1.8cm,keepaspectratio]{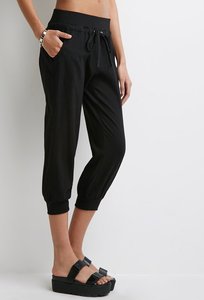}\hss\includegraphics[width=1.10cm,height=1.8cm,keepaspectratio]{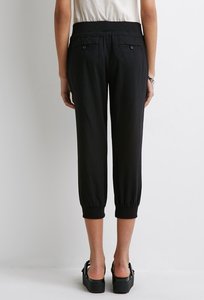}\hss\includegraphics[width=1.10cm,height=1.8cm,keepaspectratio]{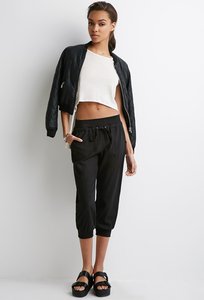}\hss\phantom{\rule{1.10cm}{1.8cm}}\hss}
      \end{minipage}
    }
  \end{minipage}
\hfill
  \begin{minipage}[t]{5.55cm}
    \fcolorbox{green!60!black}{white}{
      \begin{minipage}[t]{5.25cm}
        {\tiny\textbf{\#3}}\\
        \noindent\hbox to 5.25cm{\includegraphics[width=1.10cm,height=1.8cm,keepaspectratio]{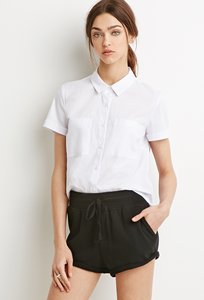}\hss\includegraphics[width=1.10cm,height=1.8cm,keepaspectratio]{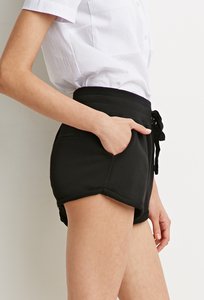}\hss\includegraphics[width=1.10cm,height=1.8cm,keepaspectratio]{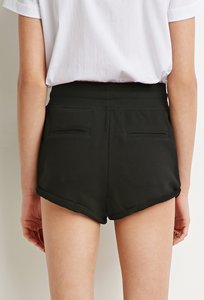}\hss\includegraphics[width=1.10cm,height=1.8cm,keepaspectratio]{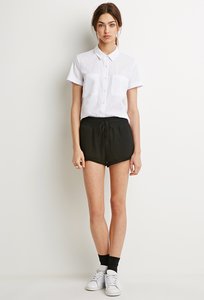}\hss\includegraphics[width=1.10cm,height=1.8cm,keepaspectratio]{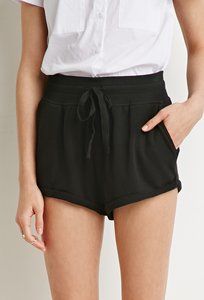}\hss}
      \end{minipage}
    }
  \end{minipage}
\par\vspace{2pt}
\noindent
  \begin{minipage}[t]{5.55cm}
    \fcolorbox{red!70!black}{white}{
      \begin{minipage}[t]{5.25cm}
        {\tiny\textbf{\#4}}\\
        \noindent\hbox to 5.25cm{\includegraphics[width=1.10cm,height=1.8cm,keepaspectratio]{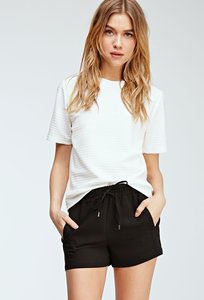}\hss\includegraphics[width=1.10cm,height=1.8cm,keepaspectratio]{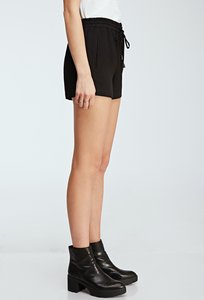}\hss\includegraphics[width=1.10cm,height=1.8cm,keepaspectratio]{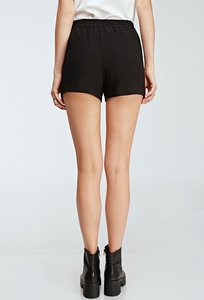}\hss\includegraphics[width=1.10cm,height=1.8cm,keepaspectratio]{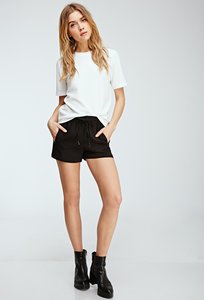}\hss\includegraphics[width=1.10cm,height=1.8cm,keepaspectratio]{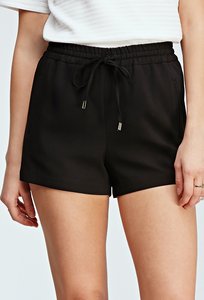}\hss}
      \end{minipage}
    }
  \end{minipage}
\hfill
  \begin{minipage}[t]{5.55cm}
    \fcolorbox{red!70!black}{white}{
      \begin{minipage}[t]{5.25cm}
        {\tiny\textbf{\#5}}\\
        \noindent\hbox to 5.25cm{\includegraphics[width=1.10cm,height=1.8cm,keepaspectratio]{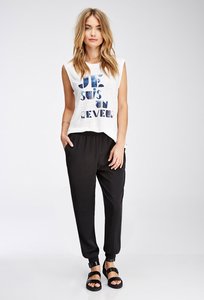}\hss\includegraphics[width=1.10cm,height=1.8cm,keepaspectratio]{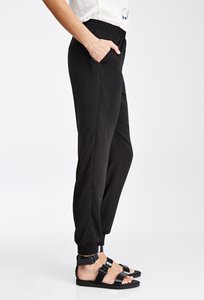}\hss\includegraphics[width=1.10cm,height=1.8cm,keepaspectratio]{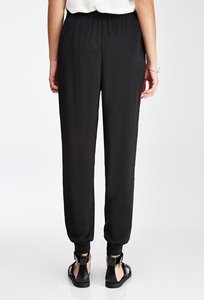}\hss\includegraphics[width=1.10cm,height=1.8cm,keepaspectratio]{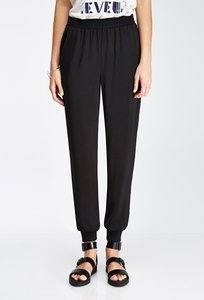}\hss\phantom{\rule{1.10cm}{1.8cm}}\hss}
      \end{minipage}
    }
  \end{minipage}
\hfill
  \begin{minipage}[t]{5.55cm}
    \fcolorbox{red!70!black}{white}{
      \begin{minipage}[t]{5.25cm}
        {\tiny\textbf{\#6}}\\
        \noindent\hbox to 5.25cm{\includegraphics[width=1.10cm,height=1.8cm,keepaspectratio]{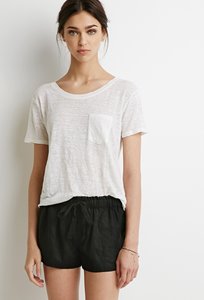}\hss\includegraphics[width=1.10cm,height=1.8cm,keepaspectratio]{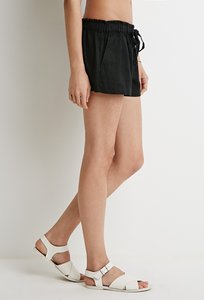}\hss\includegraphics[width=1.10cm,height=1.8cm,keepaspectratio]{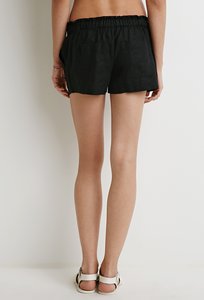}\hss\includegraphics[width=1.10cm,height=1.8cm,keepaspectratio]{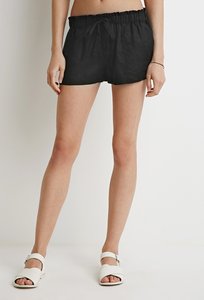}\hss\phantom{\rule{1.10cm}{1.8cm}}\hss}
      \end{minipage}
    }
  \end{minipage}
\par\vspace{2pt}
\noindent
  \begin{minipage}[t]{5.55cm}
    \fcolorbox{red!70!black}{white}{
      \begin{minipage}[t]{5.25cm}
        {\tiny\textbf{\#7}}\\
        \noindent\hbox to 5.25cm{\includegraphics[width=1.10cm,height=1.8cm,keepaspectratio]{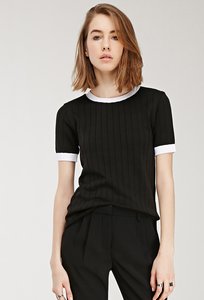}\hss\includegraphics[width=1.10cm,height=1.8cm,keepaspectratio]{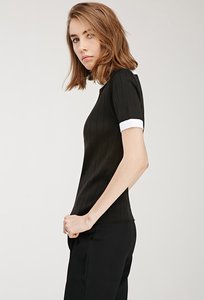}\hss\includegraphics[width=1.10cm,height=1.8cm,keepaspectratio]{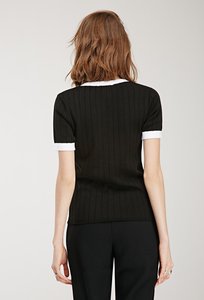}\hss\includegraphics[width=1.10cm,height=1.8cm,keepaspectratio]{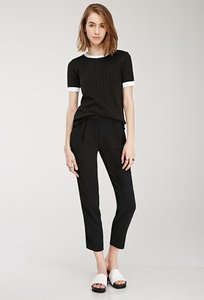}\hss\includegraphics[width=1.10cm,height=1.8cm,keepaspectratio]{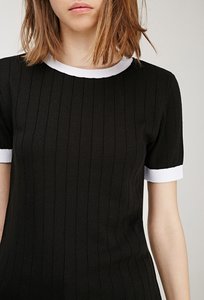}\hss}
      \end{minipage}
    }
  \end{minipage}
\hfill
  \begin{minipage}[t]{5.55cm}
    \fcolorbox{red!70!black}{white}{
      \begin{minipage}[t]{5.25cm}
        {\tiny\textbf{\#8}}\\
        \noindent\hbox to 5.25cm{\includegraphics[width=1.10cm,height=1.8cm,keepaspectratio]{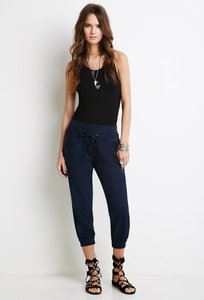}\hss\includegraphics[width=1.10cm,height=1.8cm,keepaspectratio]{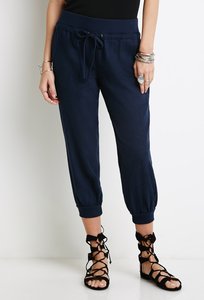}\hss\includegraphics[width=1.10cm,height=1.8cm,keepaspectratio]{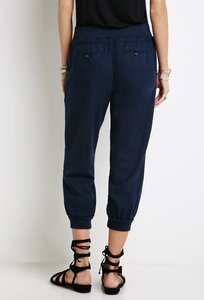}\hss\includegraphics[width=1.10cm,height=1.8cm,keepaspectratio]{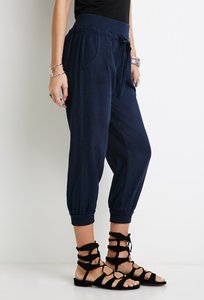}\hss\phantom{\rule{1.10cm}{1.8cm}}\hss}
      \end{minipage}
    }
  \end{minipage}
\hfill
  \begin{minipage}[t]{5.55cm}
    \fcolorbox{red!70!black}{white}{
      \begin{minipage}[t]{5.25cm}
        {\tiny\textbf{\#9}}\\
        \noindent\hbox to 5.25cm{\includegraphics[width=1.10cm,height=1.8cm,keepaspectratio]{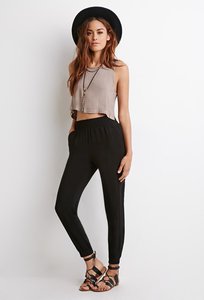}\hss\includegraphics[width=1.10cm,height=1.8cm,keepaspectratio]{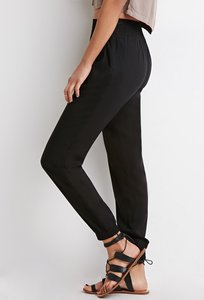}\hss\includegraphics[width=1.10cm,height=1.8cm,keepaspectratio]{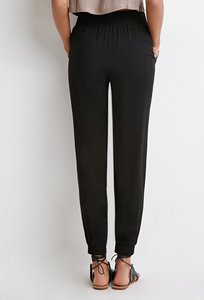}\hss\includegraphics[width=1.10cm,height=1.8cm,keepaspectratio]{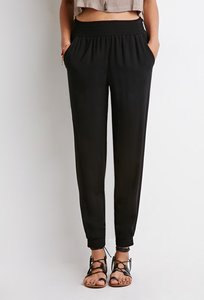}\hss\phantom{\rule{1.10cm}{1.8cm}}\hss}
      \end{minipage}
    }
  \end{minipage}
\par\vspace{2pt}
\noindent
  \begin{minipage}[t]{5.55cm}
    \fcolorbox{red!70!black}{white}{
      \begin{minipage}[t]{5.25cm}
        {\tiny\textbf{\#10}}\\
        \noindent\hbox to 5.25cm{\includegraphics[width=1.10cm,height=1.8cm,keepaspectratio]{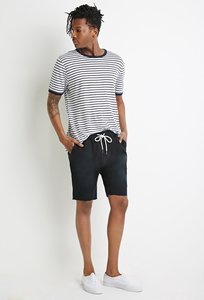}\hss\includegraphics[width=1.10cm,height=1.8cm,keepaspectratio]{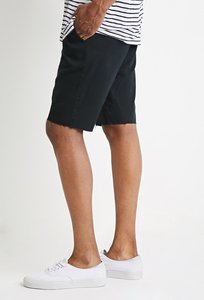}\hss\includegraphics[width=1.10cm,height=1.8cm,keepaspectratio]{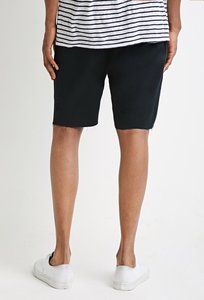}\hss\includegraphics[width=1.10cm,height=1.8cm,keepaspectratio]{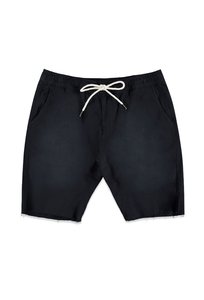}\hss\includegraphics[width=1.10cm,height=1.8cm,keepaspectratio]{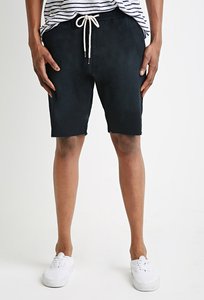}\hss}
      \end{minipage}
    }
  \end{minipage}
\hfill
  \begin{minipage}[t]{5.55cm}\end{minipage}
\hfill
  \begin{minipage}[t]{5.55cm}\end{minipage}
\par\vspace{2pt}
\noindent\rule{\linewidth}{0.4pt}
\par\vspace{2pt}
% -- reznembed --
{\small\textbf{RezNEmbed}}\quad{\small Not in Top-10}\\
\noindent
  \begin{minipage}[t]{5.55cm}
    \fcolorbox{red!70!black}{white}{
      \begin{minipage}[t]{5.25cm}
        {\tiny\textbf{\#1}}\\
        \noindent\hbox to 5.25cm{\includegraphics[width=1.10cm,height=1.8cm,keepaspectratio]{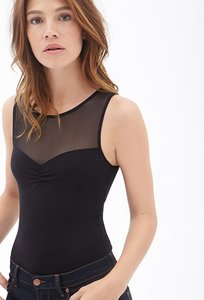}\hss\includegraphics[width=1.10cm,height=1.8cm,keepaspectratio]{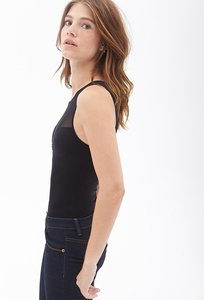}\hss\includegraphics[width=1.10cm,height=1.8cm,keepaspectratio]{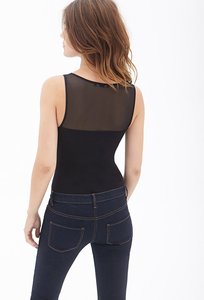}\hss\includegraphics[width=1.10cm,height=1.8cm,keepaspectratio]{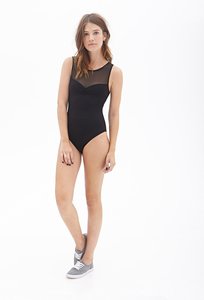}\hss\phantom{\rule{1.10cm}{1.8cm}}\hss}
      \end{minipage}
    }
  \end{minipage}
\hfill
  \begin{minipage}[t]{5.55cm}
    \fcolorbox{red!70!black}{white}{
      \begin{minipage}[t]{5.25cm}
        {\tiny\textbf{\#2}}\\
        \noindent\hbox to 5.25cm{\includegraphics[width=1.10cm,height=1.8cm,keepaspectratio]{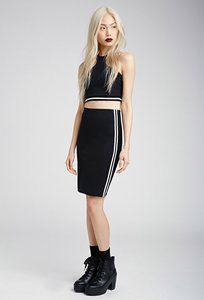}\hss\includegraphics[width=1.10cm,height=1.8cm,keepaspectratio]{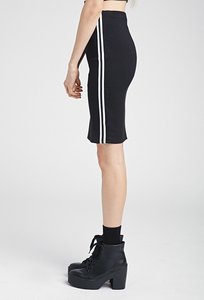}\hss\includegraphics[width=1.10cm,height=1.8cm,keepaspectratio]{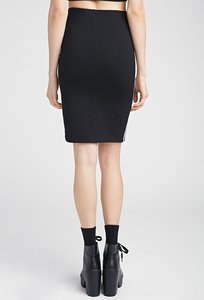}\hss\includegraphics[width=1.10cm,height=1.8cm,keepaspectratio]{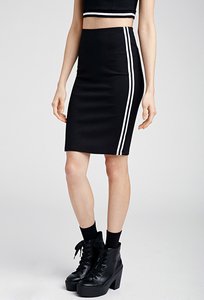}\hss\phantom{\rule{1.10cm}{1.8cm}}\hss}
      \end{minipage}
    }
  \end{minipage}
\hfill
  \begin{minipage}[t]{5.55cm}
    \fcolorbox{red!70!black}{white}{
      \begin{minipage}[t]{5.25cm}
        {\tiny\textbf{\#3}}\\
        \noindent\hbox to 5.25cm{\includegraphics[width=1.10cm,height=1.8cm,keepaspectratio]{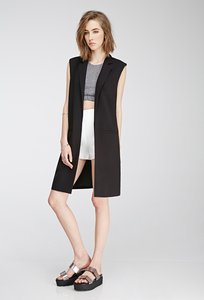}\hss\includegraphics[width=1.10cm,height=1.8cm,keepaspectratio]{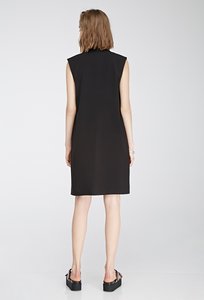}\hss\includegraphics[width=1.10cm,height=1.8cm,keepaspectratio]{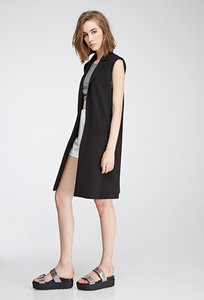}\hss\phantom{\rule{1.10cm}{1.8cm}}\hss\phantom{\rule{1.10cm}{1.8cm}}\hss}
      \end{minipage}
    }
  \end{minipage}
\par\vspace{2pt}
\noindent
  \begin{minipage}[t]{5.55cm}
    \fcolorbox{red!70!black}{white}{
      \begin{minipage}[t]{5.25cm}
        {\tiny\textbf{\#4}}\\
        \noindent\hbox to 5.25cm{\includegraphics[width=1.10cm,height=1.8cm,keepaspectratio]{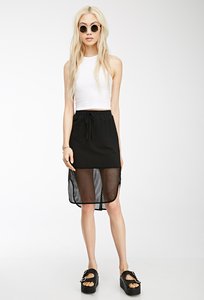}\hss\includegraphics[width=1.10cm,height=1.8cm,keepaspectratio]{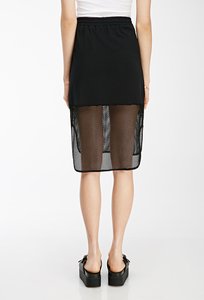}\hss\includegraphics[width=1.10cm,height=1.8cm,keepaspectratio]{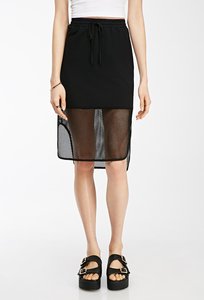}\hss\includegraphics[width=1.10cm,height=1.8cm,keepaspectratio]{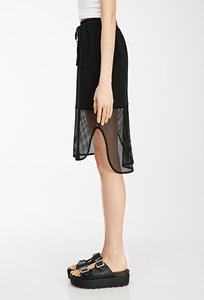}\hss\phantom{\rule{1.10cm}{1.8cm}}\hss}
      \end{minipage}
    }
  \end{minipage}
\hfill
  \begin{minipage}[t]{5.55cm}
    \fcolorbox{red!70!black}{white}{
      \begin{minipage}[t]{5.25cm}
        {\tiny\textbf{\#5}}\\
        \noindent\hbox to 5.25cm{\includegraphics[width=1.10cm,height=1.8cm,keepaspectratio]{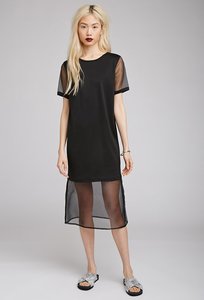}\hss\includegraphics[width=1.10cm,height=1.8cm,keepaspectratio]{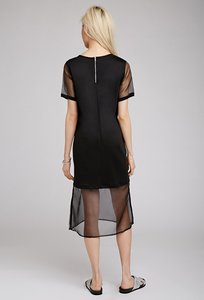}\hss\includegraphics[width=1.10cm,height=1.8cm,keepaspectratio]{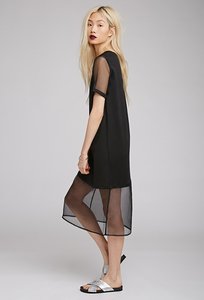}\hss\phantom{\rule{1.10cm}{1.8cm}}\hss\phantom{\rule{1.10cm}{1.8cm}}\hss}
      \end{minipage}
    }
  \end{minipage}
\hfill
  \begin{minipage}[t]{5.55cm}
    \fcolorbox{red!70!black}{white}{
      \begin{minipage}[t]{5.25cm}
        {\tiny\textbf{\#6}}\\
        \noindent\hbox to 5.25cm{\includegraphics[width=1.10cm,height=1.8cm,keepaspectratio]{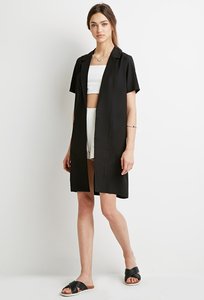}\hss\includegraphics[width=1.10cm,height=1.8cm,keepaspectratio]{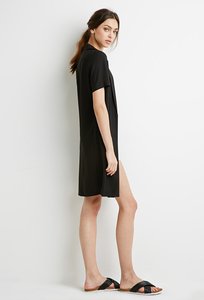}\hss\includegraphics[width=1.10cm,height=1.8cm,keepaspectratio]{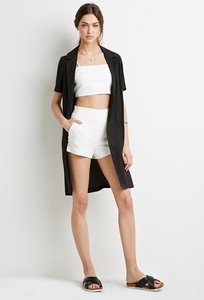}\hss\includegraphics[width=1.10cm,height=1.8cm,keepaspectratio]{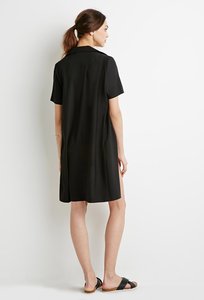}\hss\phantom{\rule{1.10cm}{1.8cm}}\hss}
      \end{minipage}
    }
  \end{minipage}
\par\vspace{2pt}
\noindent
  \begin{minipage}[t]{5.55cm}
    \fcolorbox{red!70!black}{white}{
      \begin{minipage}[t]{5.25cm}
        {\tiny\textbf{\#7}}\\
        \noindent\hbox to 5.25cm{\includegraphics[width=1.10cm,height=1.8cm,keepaspectratio]{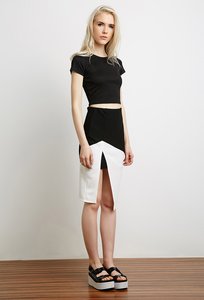}\hss\includegraphics[width=1.10cm,height=1.8cm,keepaspectratio]{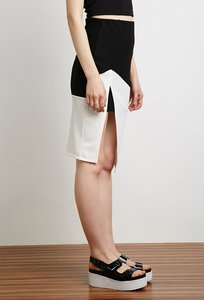}\hss\includegraphics[width=1.10cm,height=1.8cm,keepaspectratio]{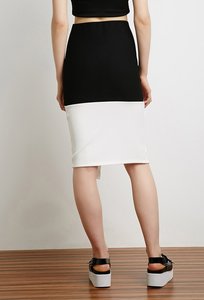}\hss\includegraphics[width=1.10cm,height=1.8cm,keepaspectratio]{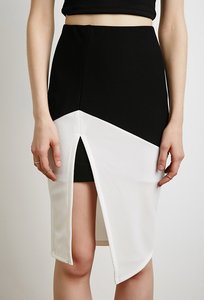}\hss\phantom{\rule{1.10cm}{1.8cm}}\hss}
      \end{minipage}
    }
  \end{minipage}
\hfill
  \begin{minipage}[t]{5.55cm}
    \fcolorbox{red!70!black}{white}{
      \begin{minipage}[t]{5.25cm}
        {\tiny\textbf{\#8}}\\
        \noindent\hbox to 5.25cm{\includegraphics[width=1.10cm,height=1.8cm,keepaspectratio]{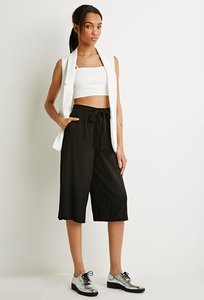}\hss\includegraphics[width=1.10cm,height=1.8cm,keepaspectratio]{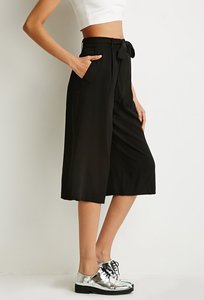}\hss\includegraphics[width=1.10cm,height=1.8cm,keepaspectratio]{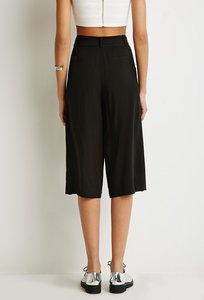}\hss\includegraphics[width=1.10cm,height=1.8cm,keepaspectratio]{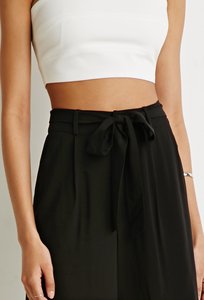}\hss\phantom{\rule{1.10cm}{1.8cm}}\hss}
      \end{minipage}
    }
  \end{minipage}
\hfill
  \begin{minipage}[t]{5.55cm}
    \fcolorbox{red!70!black}{white}{
      \begin{minipage}[t]{5.25cm}
        {\tiny\textbf{\#9}}\\
        \noindent\hbox to 5.25cm{\includegraphics[width=1.10cm,height=1.8cm,keepaspectratio]{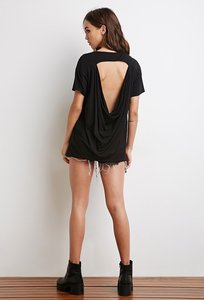}\hss\includegraphics[width=1.10cm,height=1.8cm,keepaspectratio]{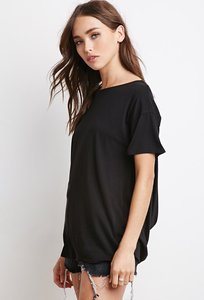}\hss\includegraphics[width=1.10cm,height=1.8cm,keepaspectratio]{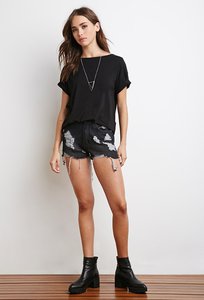}\hss\phantom{\rule{1.10cm}{1.8cm}}\hss\phantom{\rule{1.10cm}{1.8cm}}\hss}
      \end{minipage}
    }
  \end{minipage}
\par\vspace{2pt}
\noindent
  \begin{minipage}[t]{5.55cm}
    \fcolorbox{red!70!black}{white}{
      \begin{minipage}[t]{5.25cm}
        {\tiny\textbf{\#10}}\\
        \noindent\hbox to 5.25cm{\includegraphics[width=1.10cm,height=1.8cm,keepaspectratio]{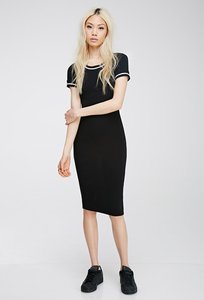}\hss\includegraphics[width=1.10cm,height=1.8cm,keepaspectratio]{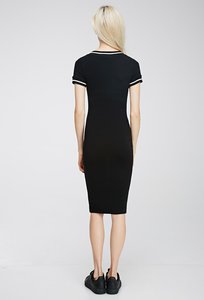}\hss\includegraphics[width=1.10cm,height=1.8cm,keepaspectratio]{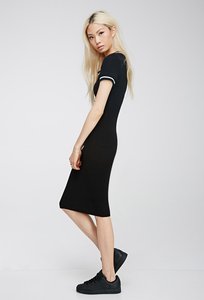}\hss\phantom{\rule{1.10cm}{1.8cm}}\hss\phantom{\rule{1.10cm}{1.8cm}}\hss}
      \end{minipage}
    }
  \end{minipage}
\hfill
  \begin{minipage}[t]{5.55cm}\end{minipage}
\hfill
  \begin{minipage}[t]{5.55cm}\end{minipage}
\par\vspace{2pt}
\noindent\rule{\linewidth}{0.4pt}
\par\vspace{2pt}
% -- doubao --
{\small\textbf{Doubao-E-V}}\quad{\small Rank~4}\\
\noindent
  \begin{minipage}[t]{5.55cm}
    \fcolorbox{red!70!black}{white}{
      \begin{minipage}[t]{5.25cm}
        {\tiny\textbf{\#1}}\\
        \noindent\hbox to 5.25cm{\includegraphics[width=1.10cm,height=1.8cm,keepaspectratio]{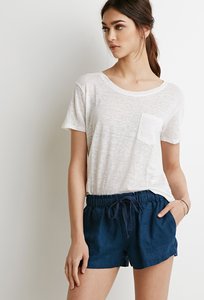}\hss\includegraphics[width=1.10cm,height=1.8cm,keepaspectratio]{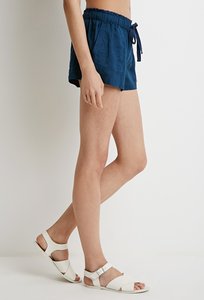}\hss\includegraphics[width=1.10cm,height=1.8cm,keepaspectratio]{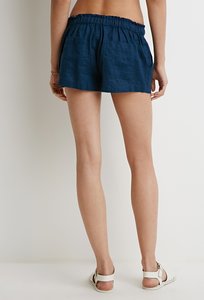}\hss\includegraphics[width=1.10cm,height=1.8cm,keepaspectratio]{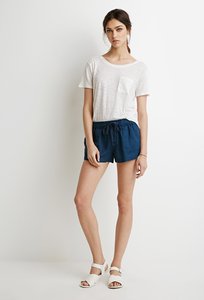}\hss\includegraphics[width=1.10cm,height=1.8cm,keepaspectratio]{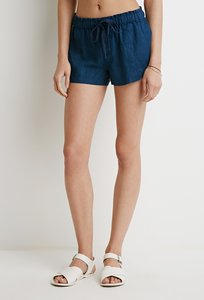}\hss}
      \end{minipage}
    }
  \end{minipage}
\hfill
  \begin{minipage}[t]{5.55cm}
    \fcolorbox{red!70!black}{white}{
      \begin{minipage}[t]{5.25cm}
        {\tiny\textbf{\#2}}\\
        \noindent\hbox to 5.25cm{\includegraphics[width=1.10cm,height=1.8cm,keepaspectratio]{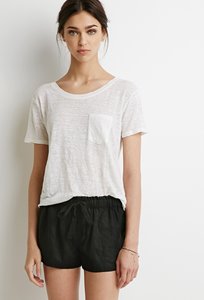}\hss\includegraphics[width=1.10cm,height=1.8cm,keepaspectratio]{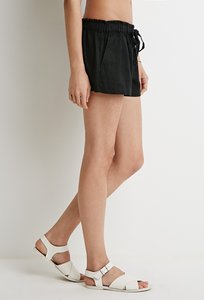}\hss\includegraphics[width=1.10cm,height=1.8cm,keepaspectratio]{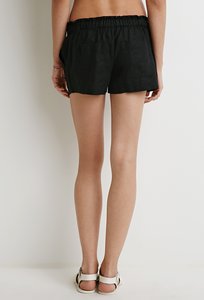}\hss\includegraphics[width=1.10cm,height=1.8cm,keepaspectratio]{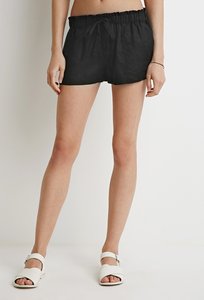}\hss\phantom{\rule{1.10cm}{1.8cm}}\hss}
      \end{minipage}
    }
  \end{minipage}
\hfill
  \begin{minipage}[t]{5.55cm}
    \fcolorbox{red!70!black}{white}{
      \begin{minipage}[t]{5.25cm}
        {\tiny\textbf{\#3}}\\
        \noindent\hbox to 5.25cm{\includegraphics[width=1.10cm,height=1.8cm,keepaspectratio]{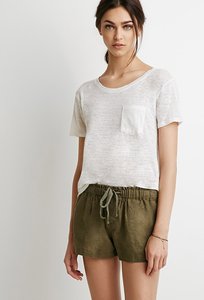}\hss\includegraphics[width=1.10cm,height=1.8cm,keepaspectratio]{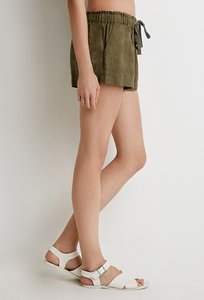}\hss\includegraphics[width=1.10cm,height=1.8cm,keepaspectratio]{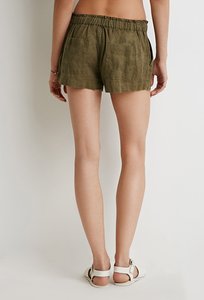}\hss\includegraphics[width=1.10cm,height=1.8cm,keepaspectratio]{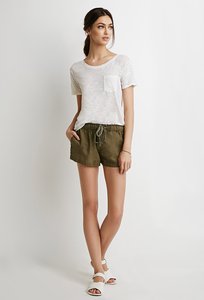}\hss\includegraphics[width=1.10cm,height=1.8cm,keepaspectratio]{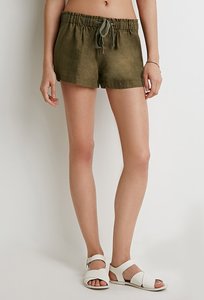}\hss}
      \end{minipage}
    }
  \end{minipage}
\par\vspace{2pt}
\noindent
  \begin{minipage}[t]{5.55cm}
    \fcolorbox{green!60!black}{white}{
      \begin{minipage}[t]{5.25cm}
        {\tiny\textbf{\#4}}\\
        \noindent\hbox to 5.25cm{\includegraphics[width=1.10cm,height=1.8cm,keepaspectratio]{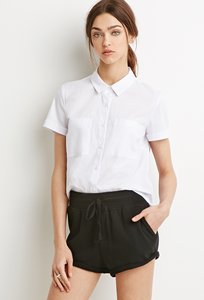}\hss\includegraphics[width=1.10cm,height=1.8cm,keepaspectratio]{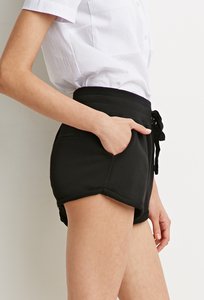}\hss\includegraphics[width=1.10cm,height=1.8cm,keepaspectratio]{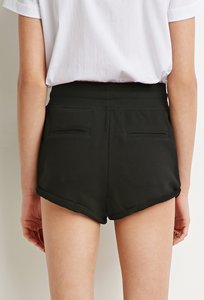}\hss\includegraphics[width=1.10cm,height=1.8cm,keepaspectratio]{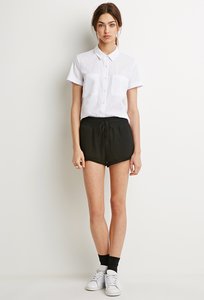}\hss\includegraphics[width=1.10cm,height=1.8cm,keepaspectratio]{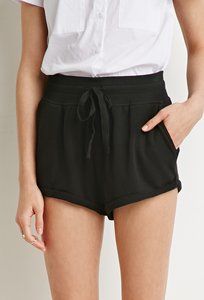}\hss}
      \end{minipage}
    }
  \end{minipage}
\hfill
  \begin{minipage}[t]{5.55cm}
    \fcolorbox{red!70!black}{white}{
      \begin{minipage}[t]{5.25cm}
        {\tiny\textbf{\#5}}\\
        \noindent\hbox to 5.25cm{\includegraphics[width=1.10cm,height=1.8cm,keepaspectratio]{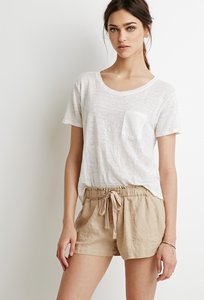}\hss\includegraphics[width=1.10cm,height=1.8cm,keepaspectratio]{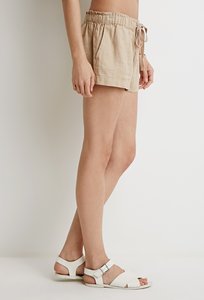}\hss\includegraphics[width=1.10cm,height=1.8cm,keepaspectratio]{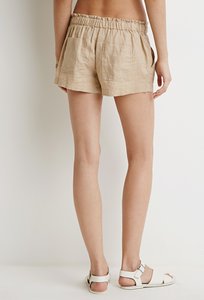}\hss\includegraphics[width=1.10cm,height=1.8cm,keepaspectratio]{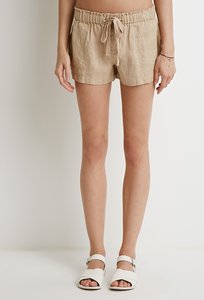}\hss\phantom{\rule{1.10cm}{1.8cm}}\hss}
      \end{minipage}
    }
  \end{minipage}
\hfill
  \begin{minipage}[t]{5.55cm}
    \fcolorbox{red!70!black}{white}{
      \begin{minipage}[t]{5.25cm}
        {\tiny\textbf{\#6}}\\
        \noindent\hbox to 5.25cm{\includegraphics[width=1.10cm,height=1.8cm,keepaspectratio]{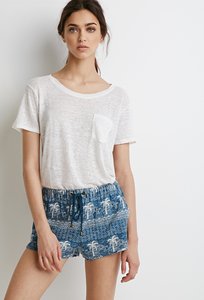}\hss\includegraphics[width=1.10cm,height=1.8cm,keepaspectratio]{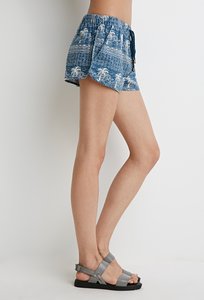}\hss\includegraphics[width=1.10cm,height=1.8cm,keepaspectratio]{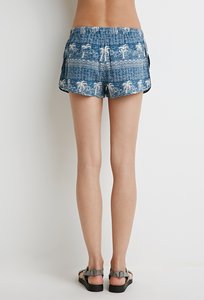}\hss\includegraphics[width=1.10cm,height=1.8cm,keepaspectratio]{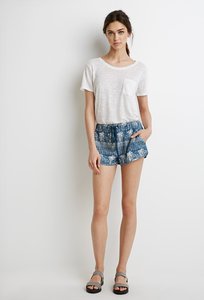}\hss\includegraphics[width=1.10cm,height=1.8cm,keepaspectratio]{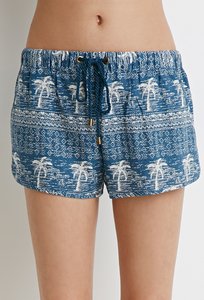}\hss}
      \end{minipage}
    }
  \end{minipage}
\par\vspace{2pt}
\noindent
  \begin{minipage}[t]{5.55cm}
    \fcolorbox{red!70!black}{white}{
      \begin{minipage}[t]{5.25cm}
        {\tiny\textbf{\#7}}\\
        \noindent\hbox to 5.25cm{\includegraphics[width=1.10cm,height=1.8cm,keepaspectratio]{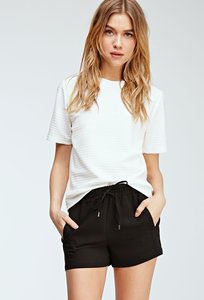}\hss\includegraphics[width=1.10cm,height=1.8cm,keepaspectratio]{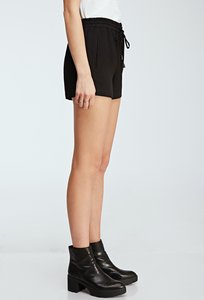}\hss\includegraphics[width=1.10cm,height=1.8cm,keepaspectratio]{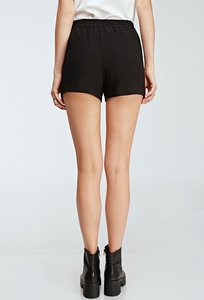}\hss\includegraphics[width=1.10cm,height=1.8cm,keepaspectratio]{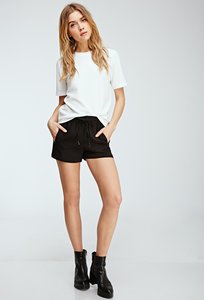}\hss\includegraphics[width=1.10cm,height=1.8cm,keepaspectratio]{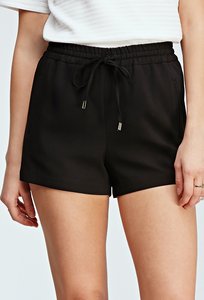}\hss}
      \end{minipage}
    }
  \end{minipage}
\hfill
  \begin{minipage}[t]{5.55cm}
    \fcolorbox{red!70!black}{white}{
      \begin{minipage}[t]{5.25cm}
        {\tiny\textbf{\#8}}\\
        \noindent\hbox to 5.25cm{\includegraphics[width=1.10cm,height=1.8cm,keepaspectratio]{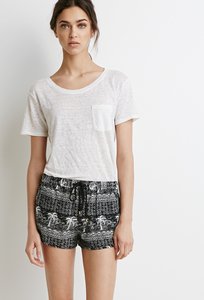}\hss\includegraphics[width=1.10cm,height=1.8cm,keepaspectratio]{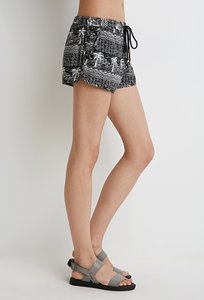}\hss\includegraphics[width=1.10cm,height=1.8cm,keepaspectratio]{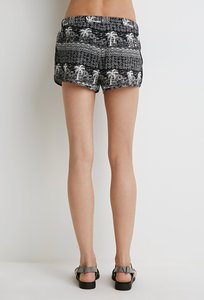}\hss\includegraphics[width=1.10cm,height=1.8cm,keepaspectratio]{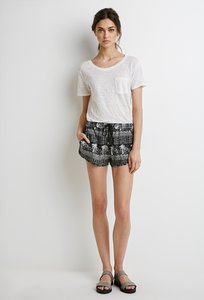}\hss\includegraphics[width=1.10cm,height=1.8cm,keepaspectratio]{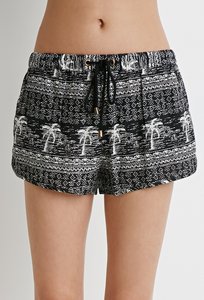}\hss}
      \end{minipage}
    }
  \end{minipage}
\hfill
  \begin{minipage}[t]{5.55cm}
    \fcolorbox{red!70!black}{white}{
      \begin{minipage}[t]{5.25cm}
        {\tiny\textbf{\#9}}\\
        \noindent\hbox to 5.25cm{\includegraphics[width=1.10cm,height=1.8cm,keepaspectratio]{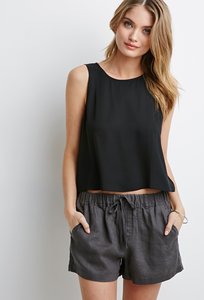}\hss\includegraphics[width=1.10cm,height=1.8cm,keepaspectratio]{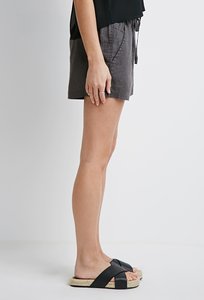}\hss\includegraphics[width=1.10cm,height=1.8cm,keepaspectratio]{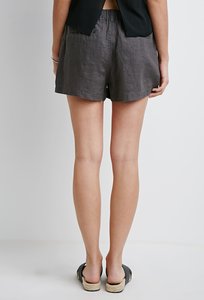}\hss\includegraphics[width=1.10cm,height=1.8cm,keepaspectratio]{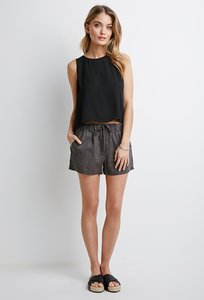}\hss\includegraphics[width=1.10cm,height=1.8cm,keepaspectratio]{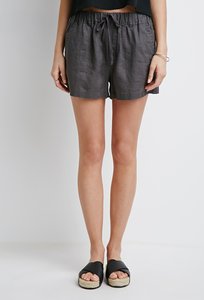}\hss}
      \end{minipage}
    }
  \end{minipage}
\par\vspace{2pt}
\noindent
  \begin{minipage}[t]{5.55cm}
    \fcolorbox{red!70!black}{white}{
      \begin{minipage}[t]{5.25cm}
        {\tiny\textbf{\#10}}\\
        \noindent\hbox to 5.25cm{\includegraphics[width=1.10cm,height=1.8cm,keepaspectratio]{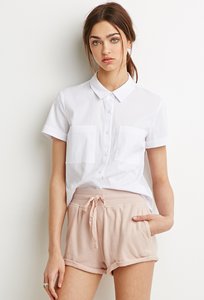}\hss\includegraphics[width=1.10cm,height=1.8cm,keepaspectratio]{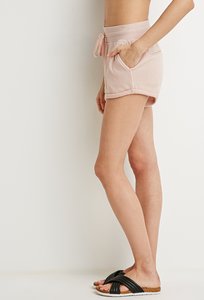}\hss\includegraphics[width=1.10cm,height=1.8cm,keepaspectratio]{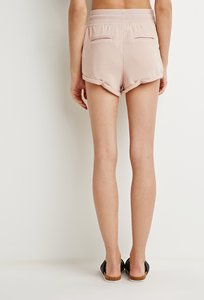}\hss\includegraphics[width=1.10cm,height=1.8cm,keepaspectratio]{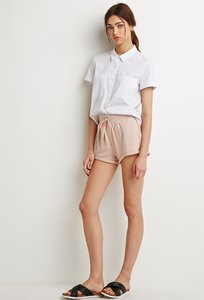}\hss\includegraphics[width=1.10cm,height=1.8cm,keepaspectratio]{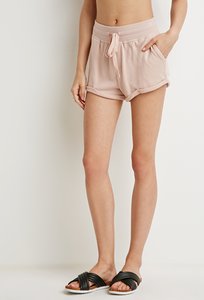}\hss}
      \end{minipage}
    }
  \end{minipage}
\hfill
  \begin{minipage}[t]{5.55cm}\end{minipage}
\hfill
  \begin{minipage}[t]{5.55cm}\end{minipage}
\par\vspace{2pt}
\noindent\rule{\linewidth}{0.4pt}
\par\vspace{20pt}

\par\vspace{16pt}
\noindent\textbf{\large Example 8}
\par\vspace{4pt}
\noindent\rule{\linewidth}{1.2pt}
\par\vspace{4pt}
% ── Case 8: short::deepfashion::743 ──
\noindent\hfill%
  \begin{minipage}[t]{6.0cm}
    \noindent\hbox to 6.0cm{\includegraphics[width=1.20cm,height=2.0cm,keepaspectratio]{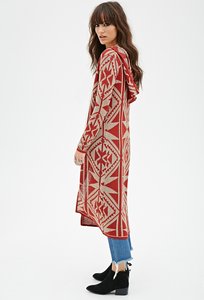}\hss\includegraphics[width=1.20cm,height=2.0cm,keepaspectratio]{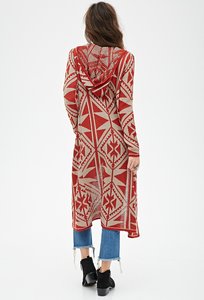}\hss\includegraphics[width=1.20cm,height=2.0cm,keepaspectratio]{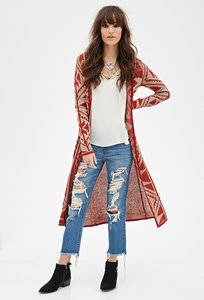}\hss\phantom{\rule{1.20cm}{2.0cm}}\hss\phantom{\rule{1.20cm}{2.0cm}}\hss}
    \par\vspace{1pt}
    {\scriptsize\textbf{}}
    \par\vspace{0pt}
    \parbox[t]{6.0cm}{\tiny\raggedright Longline hooded cardigan featuring a bold red and cream geometric tribal pattern with open front design, long sleeves, and midi length. Lightweight knit construction with Southwestern-inspired diamond motifs, symmetrical pattern placement, and relaxed bohemian silhouette perfect for casual layering.}
  \end{minipage}%
\hfill%
  \begin{minipage}[t]{3.5cm}
    \centering
    \vspace{0.45cm}%
    \parbox{3.5cm}{\centering\tiny Change color to black/white; add asymmetrical fringe hem to the front and back; modify geometric pattern to include horizontal striped bands.}\\[2pt]
    {\normalsize$\longrightarrow$}
  \end{minipage}%
\hfill%
  \begin{minipage}[t]{6.0cm}
    \noindent\hbox to 6.0cm{\includegraphics[width=1.20cm,height=2.0cm,keepaspectratio]{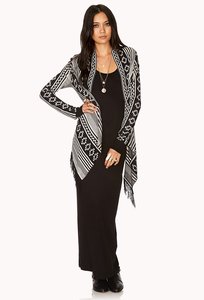}\hss\includegraphics[width=1.20cm,height=2.0cm,keepaspectratio]{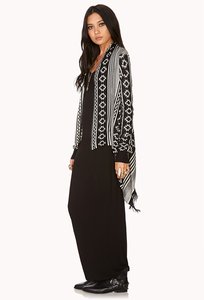}\hss\includegraphics[width=1.20cm,height=2.0cm,keepaspectratio]{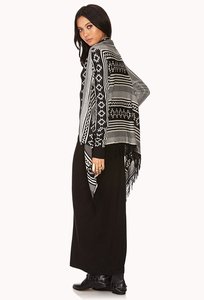}\hss\includegraphics[width=1.20cm,height=2.0cm,keepaspectratio]{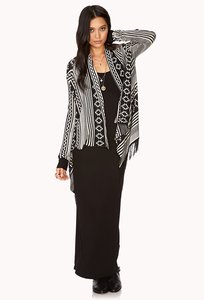}\hss\phantom{\rule{1.20cm}{2.0cm}}\hss}
    \par\vspace{1pt}
    {\scriptsize\textbf{Ground Truth}}
    \par\vspace{0pt}
    \parbox[t]{6.0cm}{\tiny\raggedright Black and white geometric open-front cardigan featuring tribal-inspired zigzag and diamond patterns, long sleeves with contrasting black ribbed cuffs, and fringe tassel trim along the asymmetrical hem. Relaxed, draped silhouette perfect for layering.}
  \end{minipage}%
\hfill
\par\vspace{4pt}
\par\vspace{4pt}
\noindent\rule{\linewidth}{0.4pt}
% -- mt_align --
{\small\textbf{\textbf{Ours}}}\quad{\small Rank~2}\\
\noindent
  \begin{minipage}[t]{5.55cm}
    \fcolorbox{red!70!black}{white}{
      \begin{minipage}[t]{5.25cm}
        {\tiny\textbf{\#1}}\\
        \noindent\hbox to 5.25cm{\includegraphics[width=1.10cm,height=1.8cm,keepaspectratio]{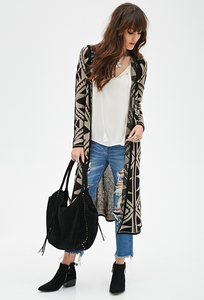}\hss\includegraphics[width=1.10cm,height=1.8cm,keepaspectratio]{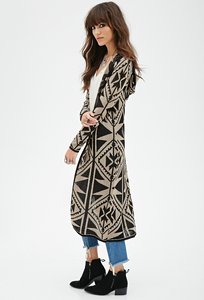}\hss\includegraphics[width=1.10cm,height=1.8cm,keepaspectratio]{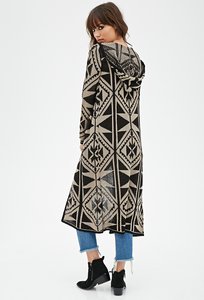}\hss\includegraphics[width=1.10cm,height=1.8cm,keepaspectratio]{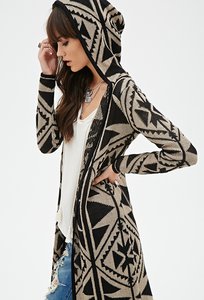}\hss\phantom{\rule{1.10cm}{1.8cm}}\hss}
      \end{minipage}
    }
  \end{minipage}
\hfill
  \begin{minipage}[t]{5.55cm}
    \fcolorbox{green!60!black}{white}{
      \begin{minipage}[t]{5.25cm}
        {\tiny\textbf{\#2}}\\
        \noindent\hbox to 5.25cm{\includegraphics[width=1.10cm,height=1.8cm,keepaspectratio]{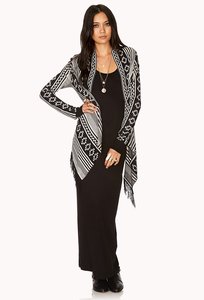}\hss\includegraphics[width=1.10cm,height=1.8cm,keepaspectratio]{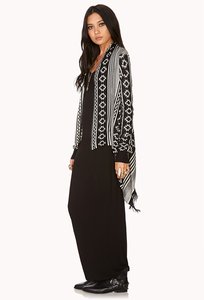}\hss\includegraphics[width=1.10cm,height=1.8cm,keepaspectratio]{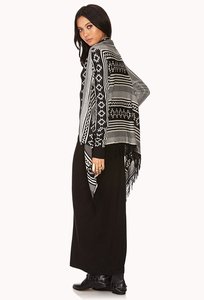}\hss\includegraphics[width=1.10cm,height=1.8cm,keepaspectratio]{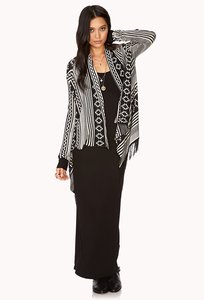}\hss\phantom{\rule{1.10cm}{1.8cm}}\hss}
      \end{minipage}
    }
  \end{minipage}
\hfill
  \begin{minipage}[t]{5.55cm}
    \fcolorbox{red!70!black}{white}{
      \begin{minipage}[t]{5.25cm}
        {\tiny\textbf{\#3}}\\
        \noindent\hbox to 5.25cm{\includegraphics[width=1.10cm,height=1.8cm,keepaspectratio]{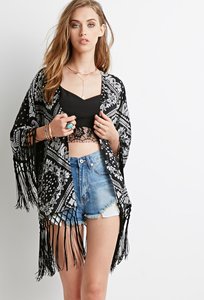}\hss\includegraphics[width=1.10cm,height=1.8cm,keepaspectratio]{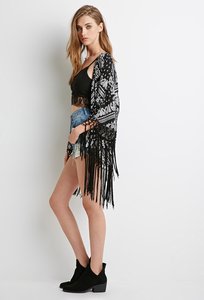}\hss\includegraphics[width=1.10cm,height=1.8cm,keepaspectratio]{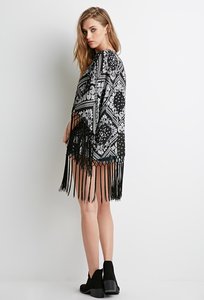}\hss\phantom{\rule{1.10cm}{1.8cm}}\hss\phantom{\rule{1.10cm}{1.8cm}}\hss}
      \end{minipage}
    }
  \end{minipage}
\par\vspace{2pt}
\noindent
  \begin{minipage}[t]{5.55cm}
    \fcolorbox{red!70!black}{white}{
      \begin{minipage}[t]{5.25cm}
        {\tiny\textbf{\#4}}\\
        \noindent\hbox to 5.25cm{\includegraphics[width=1.10cm,height=1.8cm,keepaspectratio]{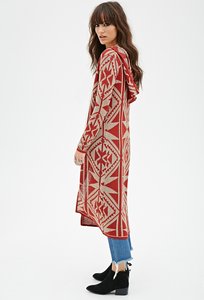}\hss\includegraphics[width=1.10cm,height=1.8cm,keepaspectratio]{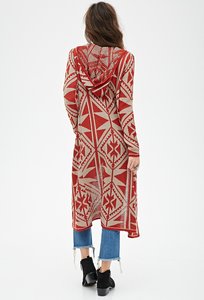}\hss\includegraphics[width=1.10cm,height=1.8cm,keepaspectratio]{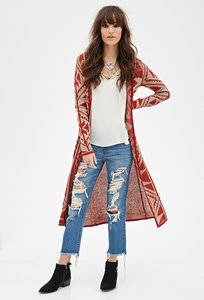}\hss\phantom{\rule{1.10cm}{1.8cm}}\hss\phantom{\rule{1.10cm}{1.8cm}}\hss}
      \end{minipage}
    }
  \end{minipage}
\hfill
  \begin{minipage}[t]{5.55cm}
    \fcolorbox{red!70!black}{white}{
      \begin{minipage}[t]{5.25cm}
        {\tiny\textbf{\#5}}\\
        \noindent\hbox to 5.25cm{\includegraphics[width=1.10cm,height=1.8cm,keepaspectratio]{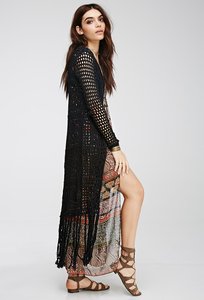}\hss\includegraphics[width=1.10cm,height=1.8cm,keepaspectratio]{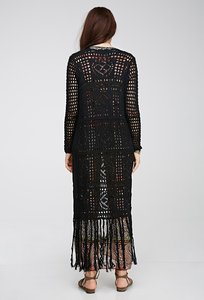}\hss\includegraphics[width=1.10cm,height=1.8cm,keepaspectratio]{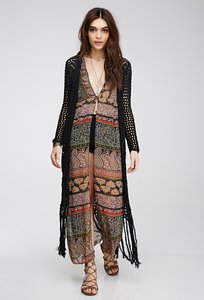}\hss\phantom{\rule{1.10cm}{1.8cm}}\hss\phantom{\rule{1.10cm}{1.8cm}}\hss}
      \end{minipage}
    }
  \end{minipage}
\hfill
  \begin{minipage}[t]{5.55cm}
    \fcolorbox{red!70!black}{white}{
      \begin{minipage}[t]{5.25cm}
        {\tiny\textbf{\#6}}\\
        \noindent\hbox to 5.25cm{\includegraphics[width=1.10cm,height=1.8cm,keepaspectratio]{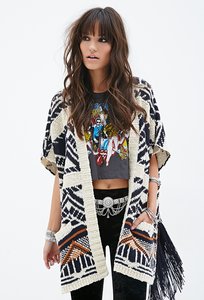}\hss\includegraphics[width=1.10cm,height=1.8cm,keepaspectratio]{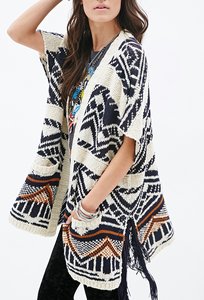}\hss\includegraphics[width=1.10cm,height=1.8cm,keepaspectratio]{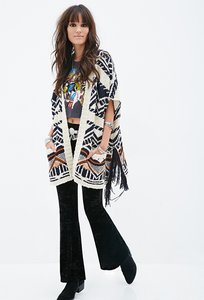}\hss\includegraphics[width=1.10cm,height=1.8cm,keepaspectratio]{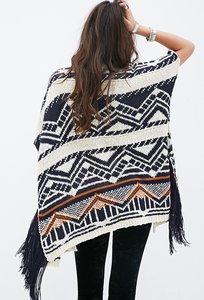}\hss\phantom{\rule{1.10cm}{1.8cm}}\hss}
      \end{minipage}
    }
  \end{minipage}
\par\vspace{2pt}
\noindent
  \begin{minipage}[t]{5.55cm}
    \fcolorbox{red!70!black}{white}{
      \begin{minipage}[t]{5.25cm}
        {\tiny\textbf{\#7}}\\
        \noindent\hbox to 5.25cm{\includegraphics[width=1.10cm,height=1.8cm,keepaspectratio]{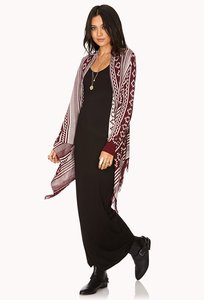}\hss\includegraphics[width=1.10cm,height=1.8cm,keepaspectratio]{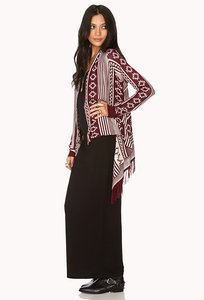}\hss\includegraphics[width=1.10cm,height=1.8cm,keepaspectratio]{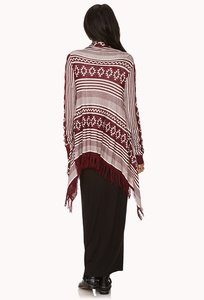}\hss\includegraphics[width=1.10cm,height=1.8cm,keepaspectratio]{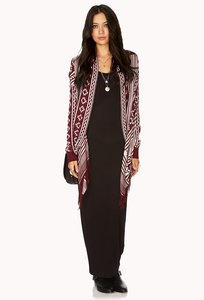}\hss\phantom{\rule{1.10cm}{1.8cm}}\hss}
      \end{minipage}
    }
  \end{minipage}
\hfill
  \begin{minipage}[t]{5.55cm}
    \fcolorbox{red!70!black}{white}{
      \begin{minipage}[t]{5.25cm}
        {\tiny\textbf{\#8}}\\
        \noindent\hbox to 5.25cm{\includegraphics[width=1.10cm,height=1.8cm,keepaspectratio]{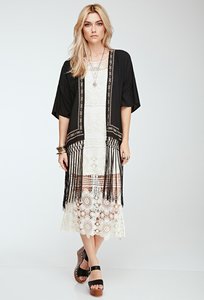}\hss\includegraphics[width=1.10cm,height=1.8cm,keepaspectratio]{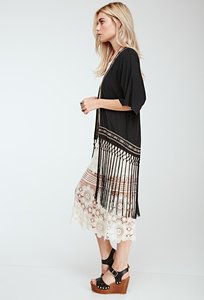}\hss\includegraphics[width=1.10cm,height=1.8cm,keepaspectratio]{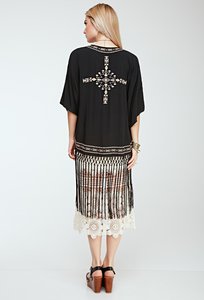}\hss\includegraphics[width=1.10cm,height=1.8cm,keepaspectratio]{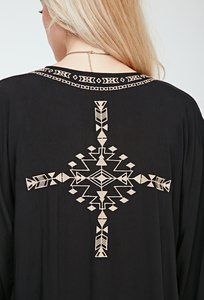}\hss\phantom{\rule{1.10cm}{1.8cm}}\hss}
      \end{minipage}
    }
  \end{minipage}
\hfill
  \begin{minipage}[t]{5.55cm}
    \fcolorbox{red!70!black}{white}{
      \begin{minipage}[t]{5.25cm}
        {\tiny\textbf{\#9}}\\
        \noindent\hbox to 5.25cm{\includegraphics[width=1.10cm,height=1.8cm,keepaspectratio]{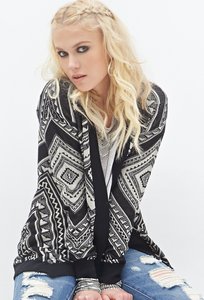}\hss\includegraphics[width=1.10cm,height=1.8cm,keepaspectratio]{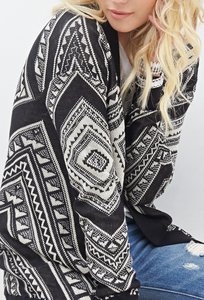}\hss\includegraphics[width=1.10cm,height=1.8cm,keepaspectratio]{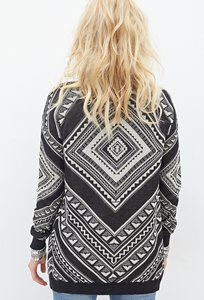}\hss\includegraphics[width=1.10cm,height=1.8cm,keepaspectratio]{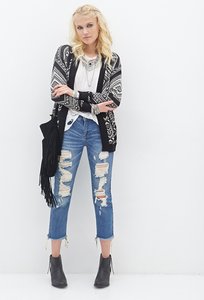}\hss\phantom{\rule{1.10cm}{1.8cm}}\hss}
      \end{minipage}
    }
  \end{minipage}
\par\vspace{2pt}
\noindent
  \begin{minipage}[t]{5.55cm}
    \fcolorbox{red!70!black}{white}{
      \begin{minipage}[t]{5.25cm}
        {\tiny\textbf{\#10}}\\
        \noindent\hbox to 5.25cm{\includegraphics[width=1.10cm,height=1.8cm,keepaspectratio]{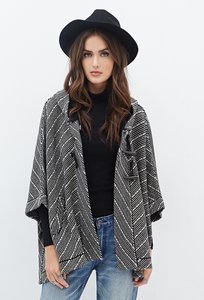}\hss\includegraphics[width=1.10cm,height=1.8cm,keepaspectratio]{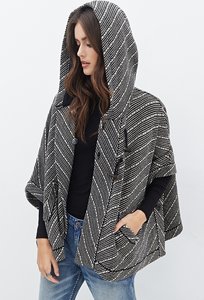}\hss\includegraphics[width=1.10cm,height=1.8cm,keepaspectratio]{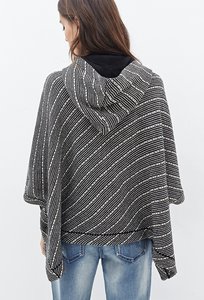}\hss\includegraphics[width=1.10cm,height=1.8cm,keepaspectratio]{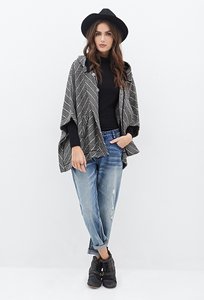}\hss\phantom{\rule{1.10cm}{1.8cm}}\hss}
      \end{minipage}
    }
  \end{minipage}
\hfill
  \begin{minipage}[t]{5.55cm}\end{minipage}
\hfill
  \begin{minipage}[t]{5.55cm}\end{minipage}
\par\vspace{2pt}
\noindent\rule{\linewidth}{0.4pt}
\par\vspace{2pt}
% -- qwen3_vl_2b --
{\small\textbf{Qwen3-VL-2B}}\quad{\small Rank~3}\\
\noindent
  \begin{minipage}[t]{5.55cm}
    \fcolorbox{red!70!black}{white}{
      \begin{minipage}[t]{5.25cm}
        {\tiny\textbf{\#1}}\\
        \noindent\hbox to 5.25cm{\includegraphics[width=1.10cm,height=1.8cm,keepaspectratio]{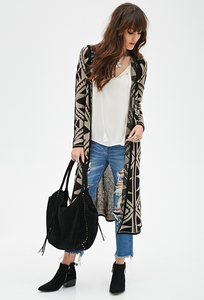}\hss\includegraphics[width=1.10cm,height=1.8cm,keepaspectratio]{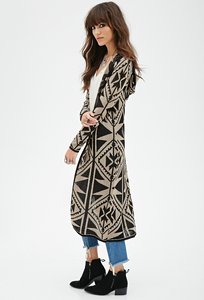}\hss\includegraphics[width=1.10cm,height=1.8cm,keepaspectratio]{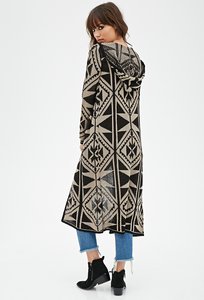}\hss\includegraphics[width=1.10cm,height=1.8cm,keepaspectratio]{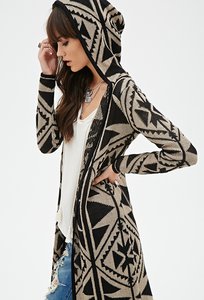}\hss\phantom{\rule{1.10cm}{1.8cm}}\hss}
      \end{minipage}
    }
  \end{minipage}
\hfill
  \begin{minipage}[t]{5.55cm}
    \fcolorbox{red!70!black}{white}{
      \begin{minipage}[t]{5.25cm}
        {\tiny\textbf{\#2}}\\
        \noindent\hbox to 5.25cm{\includegraphics[width=1.10cm,height=1.8cm,keepaspectratio]{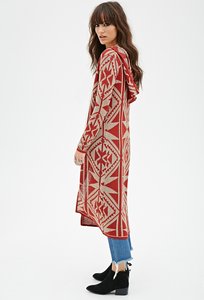}\hss\includegraphics[width=1.10cm,height=1.8cm,keepaspectratio]{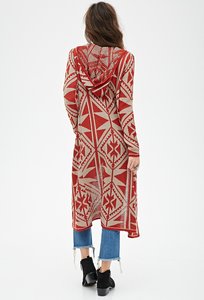}\hss\includegraphics[width=1.10cm,height=1.8cm,keepaspectratio]{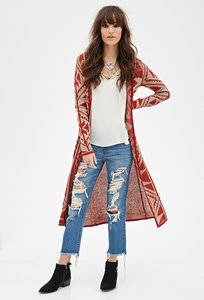}\hss\phantom{\rule{1.10cm}{1.8cm}}\hss\phantom{\rule{1.10cm}{1.8cm}}\hss}
      \end{minipage}
    }
  \end{minipage}
\hfill
  \begin{minipage}[t]{5.55cm}
    \fcolorbox{green!60!black}{white}{
      \begin{minipage}[t]{5.25cm}
        {\tiny\textbf{\#3}}\\
        \noindent\hbox to 5.25cm{\includegraphics[width=1.10cm,height=1.8cm,keepaspectratio]{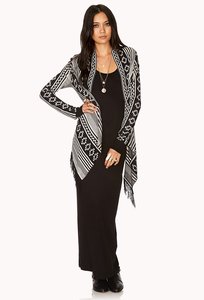}\hss\includegraphics[width=1.10cm,height=1.8cm,keepaspectratio]{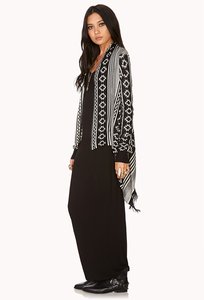}\hss\includegraphics[width=1.10cm,height=1.8cm,keepaspectratio]{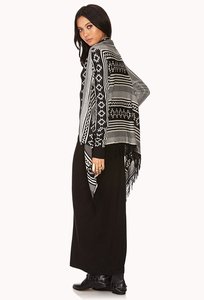}\hss\includegraphics[width=1.10cm,height=1.8cm,keepaspectratio]{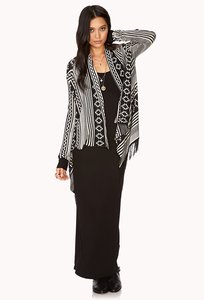}\hss\phantom{\rule{1.10cm}{1.8cm}}\hss}
      \end{minipage}
    }
  \end{minipage}
\par\vspace{2pt}
\noindent
  \begin{minipage}[t]{5.55cm}
    \fcolorbox{red!70!black}{white}{
      \begin{minipage}[t]{5.25cm}
        {\tiny\textbf{\#4}}\\
        \noindent\hbox to 5.25cm{\includegraphics[width=1.10cm,height=1.8cm,keepaspectratio]{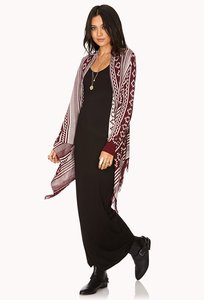}\hss\includegraphics[width=1.10cm,height=1.8cm,keepaspectratio]{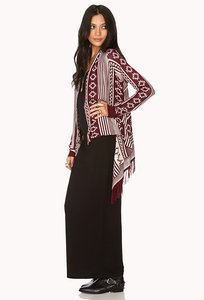}\hss\includegraphics[width=1.10cm,height=1.8cm,keepaspectratio]{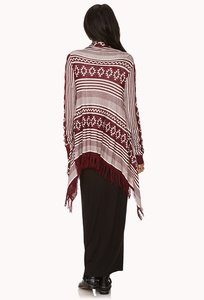}\hss\includegraphics[width=1.10cm,height=1.8cm,keepaspectratio]{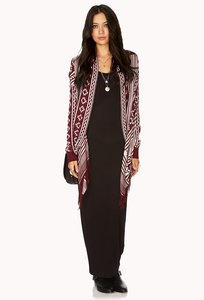}\hss\phantom{\rule{1.10cm}{1.8cm}}\hss}
      \end{minipage}
    }
  \end{minipage}
\hfill
  \begin{minipage}[t]{5.55cm}
    \fcolorbox{red!70!black}{white}{
      \begin{minipage}[t]{5.25cm}
        {\tiny\textbf{\#5}}\\
        \noindent\hbox to 5.25cm{\includegraphics[width=1.10cm,height=1.8cm,keepaspectratio]{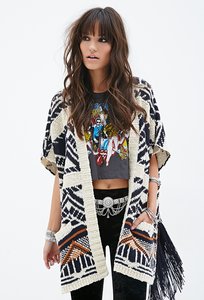}\hss\includegraphics[width=1.10cm,height=1.8cm,keepaspectratio]{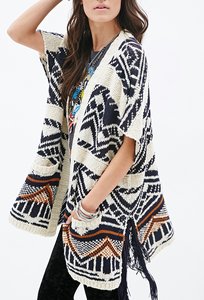}\hss\includegraphics[width=1.10cm,height=1.8cm,keepaspectratio]{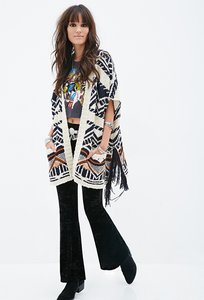}\hss\includegraphics[width=1.10cm,height=1.8cm,keepaspectratio]{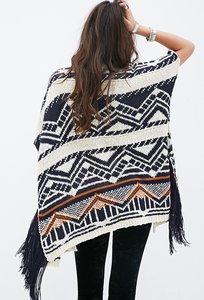}\hss\phantom{\rule{1.10cm}{1.8cm}}\hss}
      \end{minipage}
    }
  \end{minipage}
\hfill
  \begin{minipage}[t]{5.55cm}
    \fcolorbox{red!70!black}{white}{
      \begin{minipage}[t]{5.25cm}
        {\tiny\textbf{\#6}}\\
        \noindent\hbox to 5.25cm{\includegraphics[width=1.10cm,height=1.8cm,keepaspectratio]{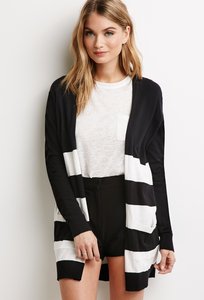}\hss\includegraphics[width=1.10cm,height=1.8cm,keepaspectratio]{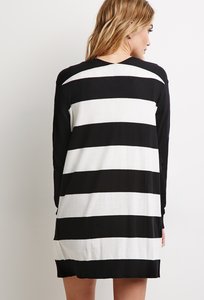}\hss\includegraphics[width=1.10cm,height=1.8cm,keepaspectratio]{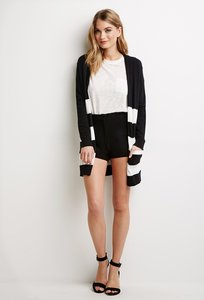}\hss\includegraphics[width=1.10cm,height=1.8cm,keepaspectratio]{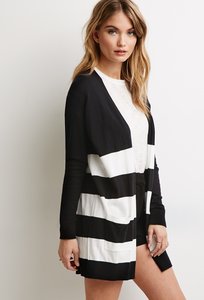}\hss\phantom{\rule{1.10cm}{1.8cm}}\hss}
      \end{minipage}
    }
  \end{minipage}
\par\vspace{2pt}
\noindent
  \begin{minipage}[t]{5.55cm}
    \fcolorbox{red!70!black}{white}{
      \begin{minipage}[t]{5.25cm}
        {\tiny\textbf{\#7}}\\
        \noindent\hbox to 5.25cm{\includegraphics[width=1.10cm,height=1.8cm,keepaspectratio]{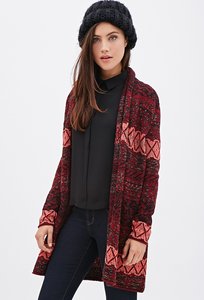}\hss\includegraphics[width=1.10cm,height=1.8cm,keepaspectratio]{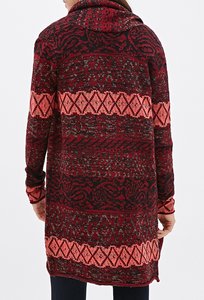}\hss\includegraphics[width=1.10cm,height=1.8cm,keepaspectratio]{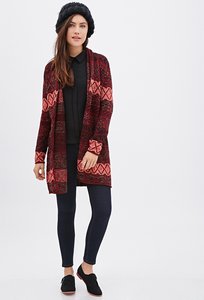}\hss\includegraphics[width=1.10cm,height=1.8cm,keepaspectratio]{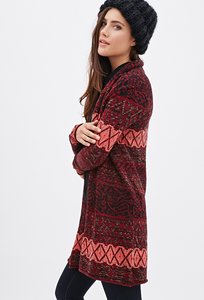}\hss\phantom{\rule{1.10cm}{1.8cm}}\hss}
      \end{minipage}
    }
  \end{minipage}
\hfill
  \begin{minipage}[t]{5.55cm}
    \fcolorbox{red!70!black}{white}{
      \begin{minipage}[t]{5.25cm}
        {\tiny\textbf{\#8}}\\
        \noindent\hbox to 5.25cm{\includegraphics[width=1.10cm,height=1.8cm,keepaspectratio]{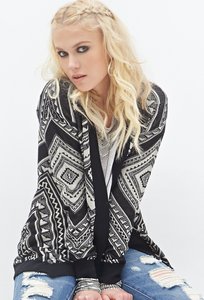}\hss\includegraphics[width=1.10cm,height=1.8cm,keepaspectratio]{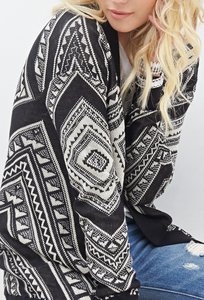}\hss\includegraphics[width=1.10cm,height=1.8cm,keepaspectratio]{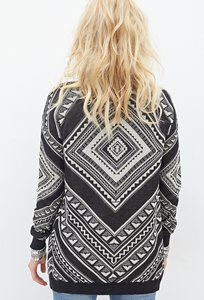}\hss\includegraphics[width=1.10cm,height=1.8cm,keepaspectratio]{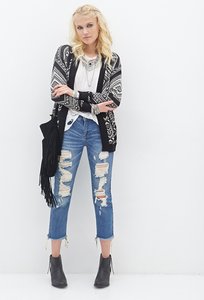}\hss\phantom{\rule{1.10cm}{1.8cm}}\hss}
      \end{minipage}
    }
  \end{minipage}
\hfill
  \begin{minipage}[t]{5.55cm}
    \fcolorbox{red!70!black}{white}{
      \begin{minipage}[t]{5.25cm}
        {\tiny\textbf{\#9}}\\
        \noindent\hbox to 5.25cm{\includegraphics[width=1.10cm,height=1.8cm,keepaspectratio]{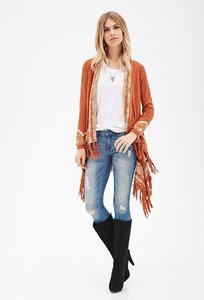}\hss\includegraphics[width=1.10cm,height=1.8cm,keepaspectratio]{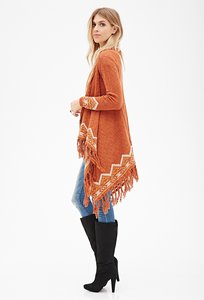}\hss\includegraphics[width=1.10cm,height=1.8cm,keepaspectratio]{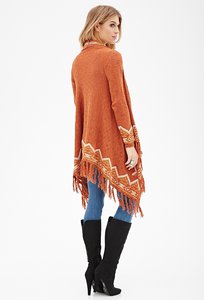}\hss\phantom{\rule{1.10cm}{1.8cm}}\hss\phantom{\rule{1.10cm}{1.8cm}}\hss}
      \end{minipage}
    }
  \end{minipage}
\par\vspace{2pt}
\noindent
  \begin{minipage}[t]{5.55cm}
    \fcolorbox{red!70!black}{white}{
      \begin{minipage}[t]{5.25cm}
        {\tiny\textbf{\#10}}\\
        \noindent\hbox to 5.25cm{\includegraphics[width=1.10cm,height=1.8cm,keepaspectratio]{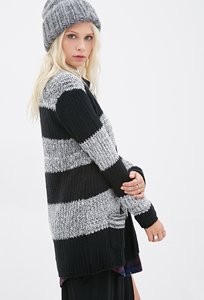}\hss\includegraphics[width=1.10cm,height=1.8cm,keepaspectratio]{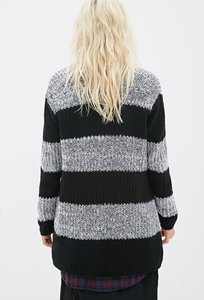}\hss\includegraphics[width=1.10cm,height=1.8cm,keepaspectratio]{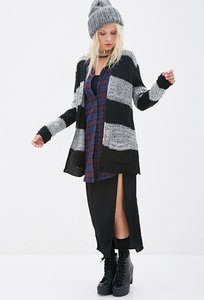}\hss\phantom{\rule{1.10cm}{1.8cm}}\hss\phantom{\rule{1.10cm}{1.8cm}}\hss}
      \end{minipage}
    }
  \end{minipage}
\hfill
  \begin{minipage}[t]{5.55cm}\end{minipage}
\hfill
  \begin{minipage}[t]{5.55cm}\end{minipage}
\par\vspace{2pt}
\noindent\rule{\linewidth}{0.4pt}
\par\vspace{2pt}
% -- qwen3_vl_8b --
{\small\textbf{Qwen3-VL-8B}}\quad{\small Rank~5}\\
\noindent
  \begin{minipage}[t]{5.55cm}
    \fcolorbox{red!70!black}{white}{
      \begin{minipage}[t]{5.25cm}
        {\tiny\textbf{\#1}}\\
        \noindent\hbox to 5.25cm{\includegraphics[width=1.10cm,height=1.8cm,keepaspectratio]{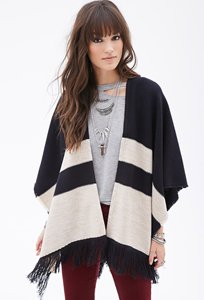}\hss\includegraphics[width=1.10cm,height=1.8cm,keepaspectratio]{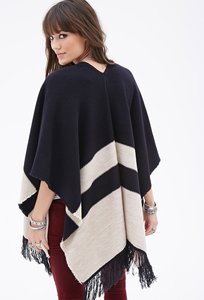}\hss\includegraphics[width=1.10cm,height=1.8cm,keepaspectratio]{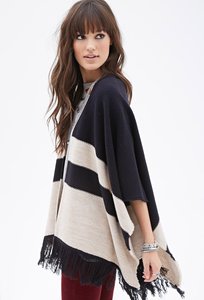}\hss\phantom{\rule{1.10cm}{1.8cm}}\hss\phantom{\rule{1.10cm}{1.8cm}}\hss}
      \end{minipage}
    }
  \end{minipage}
\hfill
  \begin{minipage}[t]{5.55cm}
    \fcolorbox{red!70!black}{white}{
      \begin{minipage}[t]{5.25cm}
        {\tiny\textbf{\#2}}\\
        \noindent\hbox to 5.25cm{\includegraphics[width=1.10cm,height=1.8cm,keepaspectratio]{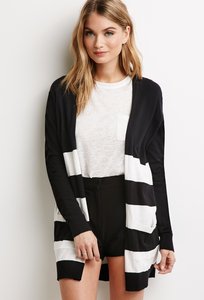}\hss\includegraphics[width=1.10cm,height=1.8cm,keepaspectratio]{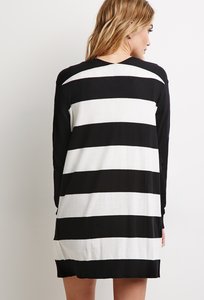}\hss\includegraphics[width=1.10cm,height=1.8cm,keepaspectratio]{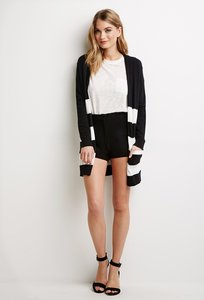}\hss\includegraphics[width=1.10cm,height=1.8cm,keepaspectratio]{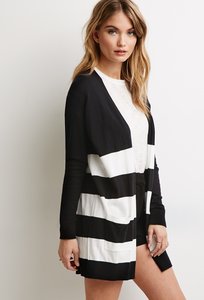}\hss\phantom{\rule{1.10cm}{1.8cm}}\hss}
      \end{minipage}
    }
  \end{minipage}
\hfill
  \begin{minipage}[t]{5.55cm}
    \fcolorbox{red!70!black}{white}{
      \begin{minipage}[t]{5.25cm}
        {\tiny\textbf{\#3}}\\
        \noindent\hbox to 5.25cm{\includegraphics[width=1.10cm,height=1.8cm,keepaspectratio]{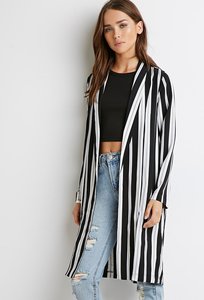}\hss\includegraphics[width=1.10cm,height=1.8cm,keepaspectratio]{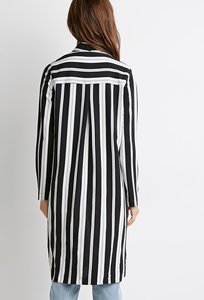}\hss\includegraphics[width=1.10cm,height=1.8cm,keepaspectratio]{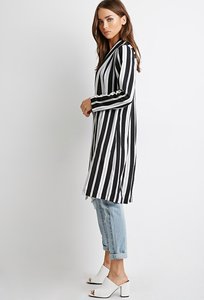}\hss\phantom{\rule{1.10cm}{1.8cm}}\hss\phantom{\rule{1.10cm}{1.8cm}}\hss}
      \end{minipage}
    }
  \end{minipage}
\par\vspace{2pt}
\noindent
  \begin{minipage}[t]{5.55cm}
    \fcolorbox{red!70!black}{white}{
      \begin{minipage}[t]{5.25cm}
        {\tiny\textbf{\#4}}\\
        \noindent\hbox to 5.25cm{\includegraphics[width=1.10cm,height=1.8cm,keepaspectratio]{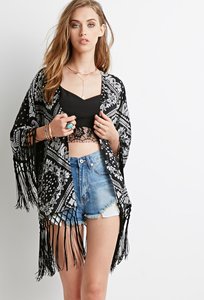}\hss\includegraphics[width=1.10cm,height=1.8cm,keepaspectratio]{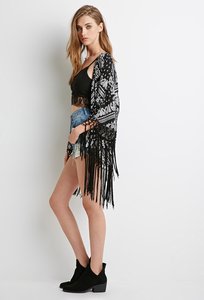}\hss\includegraphics[width=1.10cm,height=1.8cm,keepaspectratio]{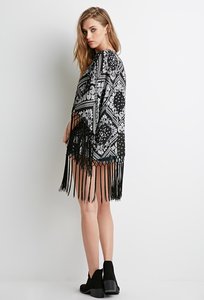}\hss\phantom{\rule{1.10cm}{1.8cm}}\hss\phantom{\rule{1.10cm}{1.8cm}}\hss}
      \end{minipage}
    }
  \end{minipage}
\hfill
  \begin{minipage}[t]{5.55cm}
    \fcolorbox{green!60!black}{white}{
      \begin{minipage}[t]{5.25cm}
        {\tiny\textbf{\#5}}\\
        \noindent\hbox to 5.25cm{\includegraphics[width=1.10cm,height=1.8cm,keepaspectratio]{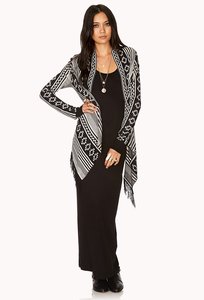}\hss\includegraphics[width=1.10cm,height=1.8cm,keepaspectratio]{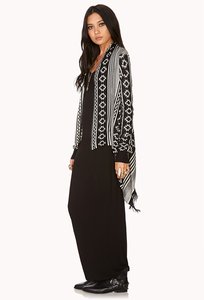}\hss\includegraphics[width=1.10cm,height=1.8cm,keepaspectratio]{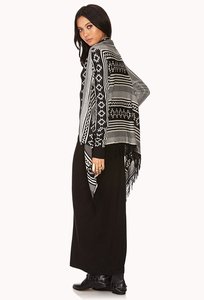}\hss\includegraphics[width=1.10cm,height=1.8cm,keepaspectratio]{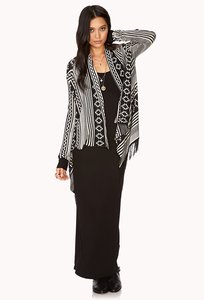}\hss\phantom{\rule{1.10cm}{1.8cm}}\hss}
      \end{minipage}
    }
  \end{minipage}
\hfill
  \begin{minipage}[t]{5.55cm}
    \fcolorbox{red!70!black}{white}{
      \begin{minipage}[t]{5.25cm}
        {\tiny\textbf{\#6}}\\
        \noindent\hbox to 5.25cm{\includegraphics[width=1.10cm,height=1.8cm,keepaspectratio]{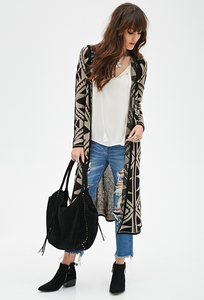}\hss\includegraphics[width=1.10cm,height=1.8cm,keepaspectratio]{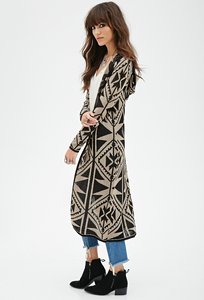}\hss\includegraphics[width=1.10cm,height=1.8cm,keepaspectratio]{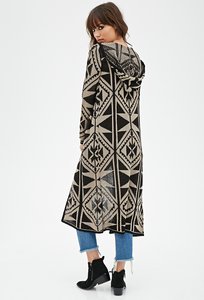}\hss\includegraphics[width=1.10cm,height=1.8cm,keepaspectratio]{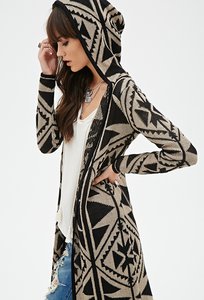}\hss\phantom{\rule{1.10cm}{1.8cm}}\hss}
      \end{minipage}
    }
  \end{minipage}
\par\vspace{2pt}
\noindent
  \begin{minipage}[t]{5.55cm}
    \fcolorbox{red!70!black}{white}{
      \begin{minipage}[t]{5.25cm}
        {\tiny\textbf{\#7}}\\
        \noindent\hbox to 5.25cm{\includegraphics[width=1.10cm,height=1.8cm,keepaspectratio]{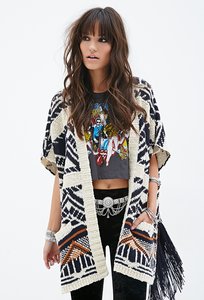}\hss\includegraphics[width=1.10cm,height=1.8cm,keepaspectratio]{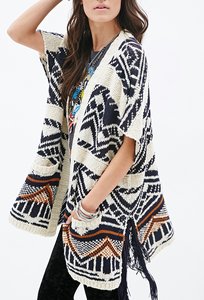}\hss\includegraphics[width=1.10cm,height=1.8cm,keepaspectratio]{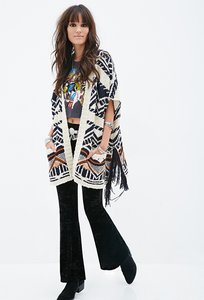}\hss\includegraphics[width=1.10cm,height=1.8cm,keepaspectratio]{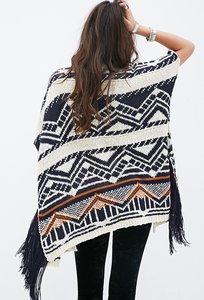}\hss\phantom{\rule{1.10cm}{1.8cm}}\hss}
      \end{minipage}
    }
  \end{minipage}
\hfill
  \begin{minipage}[t]{5.55cm}
    \fcolorbox{red!70!black}{white}{
      \begin{minipage}[t]{5.25cm}
        {\tiny\textbf{\#8}}\\
        \noindent\hbox to 5.25cm{\includegraphics[width=1.10cm,height=1.8cm,keepaspectratio]{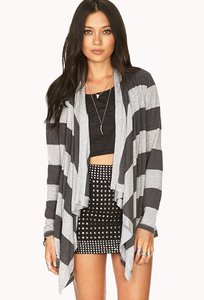}\hss\includegraphics[width=1.10cm,height=1.8cm,keepaspectratio]{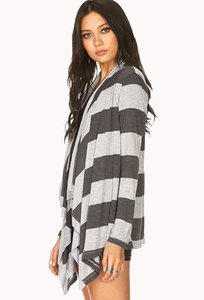}\hss\includegraphics[width=1.10cm,height=1.8cm,keepaspectratio]{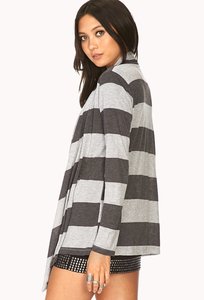}\hss\includegraphics[width=1.10cm,height=1.8cm,keepaspectratio]{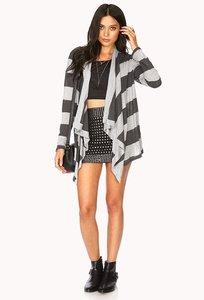}\hss\phantom{\rule{1.10cm}{1.8cm}}\hss}
      \end{minipage}
    }
  \end{minipage}
\hfill
  \begin{minipage}[t]{5.55cm}
    \fcolorbox{red!70!black}{white}{
      \begin{minipage}[t]{5.25cm}
        {\tiny\textbf{\#9}}\\
        \noindent\hbox to 5.25cm{\includegraphics[width=1.10cm,height=1.8cm,keepaspectratio]{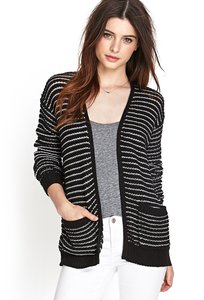}\hss\includegraphics[width=1.10cm,height=1.8cm,keepaspectratio]{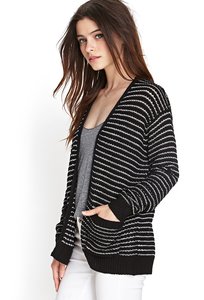}\hss\includegraphics[width=1.10cm,height=1.8cm,keepaspectratio]{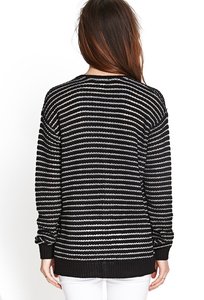}\hss\includegraphics[width=1.10cm,height=1.8cm,keepaspectratio]{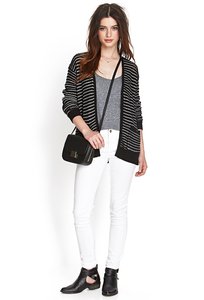}\hss\phantom{\rule{1.10cm}{1.8cm}}\hss}
      \end{minipage}
    }
  \end{minipage}
\par\vspace{2pt}
\noindent
  \begin{minipage}[t]{5.55cm}
    \fcolorbox{red!70!black}{white}{
      \begin{minipage}[t]{5.25cm}
        {\tiny\textbf{\#10}}\\
        \noindent\hbox to 5.25cm{\includegraphics[width=1.10cm,height=1.8cm,keepaspectratio]{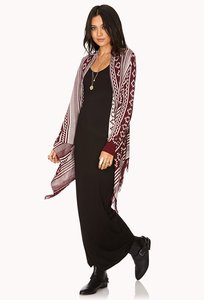}\hss\includegraphics[width=1.10cm,height=1.8cm,keepaspectratio]{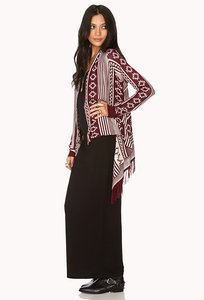}\hss\includegraphics[width=1.10cm,height=1.8cm,keepaspectratio]{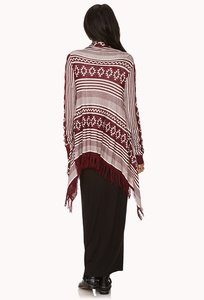}\hss\includegraphics[width=1.10cm,height=1.8cm,keepaspectratio]{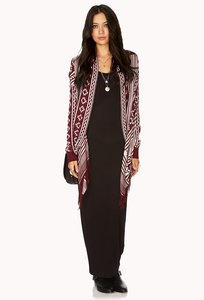}\hss\phantom{\rule{1.10cm}{1.8cm}}\hss}
      \end{minipage}
    }
  \end{minipage}
\hfill
  \begin{minipage}[t]{5.55cm}\end{minipage}
\hfill
  \begin{minipage}[t]{5.55cm}\end{minipage}
\par\vspace{2pt}
\noindent\rule{\linewidth}{0.4pt}
\par\vspace{2pt}
% -- reznembed --
{\small\textbf{RezNEmbed}}\quad{\small Not in Top-10}\\
\noindent
  \begin{minipage}[t]{5.55cm}
    \fcolorbox{red!70!black}{white}{
      \begin{minipage}[t]{5.25cm}
        {\tiny\textbf{\#1}}\\
        \noindent\hbox to 5.25cm{\includegraphics[width=1.10cm,height=1.8cm,keepaspectratio]{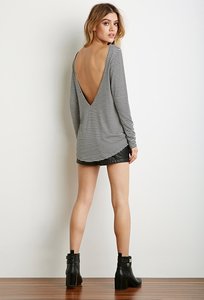}\hss\includegraphics[width=1.10cm,height=1.8cm,keepaspectratio]{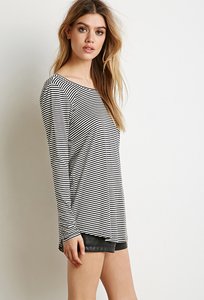}\hss\includegraphics[width=1.10cm,height=1.8cm,keepaspectratio]{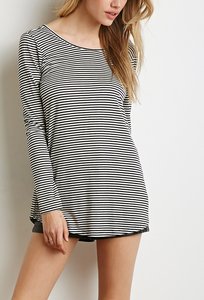}\hss\phantom{\rule{1.10cm}{1.8cm}}\hss\phantom{\rule{1.10cm}{1.8cm}}\hss}
      \end{minipage}
    }
  \end{minipage}
\hfill
  \begin{minipage}[t]{5.55cm}
    \fcolorbox{red!70!black}{white}{
      \begin{minipage}[t]{5.25cm}
        {\tiny\textbf{\#2}}\\
        \noindent\hbox to 5.25cm{\includegraphics[width=1.10cm,height=1.8cm,keepaspectratio]{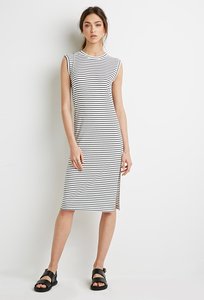}\hss\includegraphics[width=1.10cm,height=1.8cm,keepaspectratio]{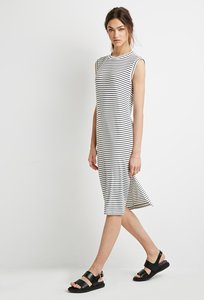}\hss\includegraphics[width=1.10cm,height=1.8cm,keepaspectratio]{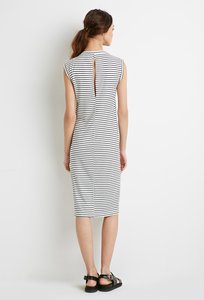}\hss\includegraphics[width=1.10cm,height=1.8cm,keepaspectratio]{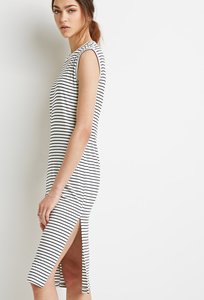}\hss\phantom{\rule{1.10cm}{1.8cm}}\hss}
      \end{minipage}
    }
  \end{minipage}
\hfill
  \begin{minipage}[t]{5.55cm}
    \fcolorbox{red!70!black}{white}{
      \begin{minipage}[t]{5.25cm}
        {\tiny\textbf{\#3}}\\
        \noindent\hbox to 5.25cm{\includegraphics[width=1.10cm,height=1.8cm,keepaspectratio]{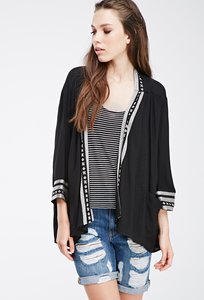}\hss\includegraphics[width=1.10cm,height=1.8cm,keepaspectratio]{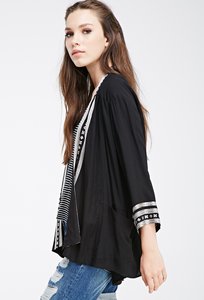}\hss\includegraphics[width=1.10cm,height=1.8cm,keepaspectratio]{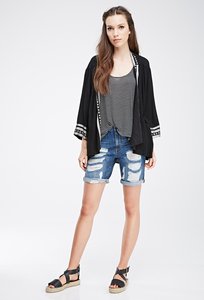}\hss\includegraphics[width=1.10cm,height=1.8cm,keepaspectratio]{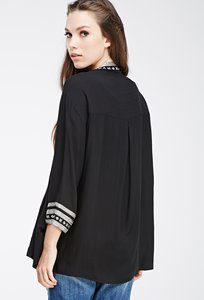}\hss\phantom{\rule{1.10cm}{1.8cm}}\hss}
      \end{minipage}
    }
  \end{minipage}
\par\vspace{2pt}
\noindent
  \begin{minipage}[t]{5.55cm}
    \fcolorbox{red!70!black}{white}{
      \begin{minipage}[t]{5.25cm}
        {\tiny\textbf{\#4}}\\
        \noindent\hbox to 5.25cm{\includegraphics[width=1.10cm,height=1.8cm,keepaspectratio]{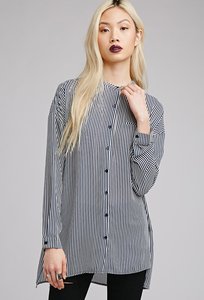}\hss\includegraphics[width=1.10cm,height=1.8cm,keepaspectratio]{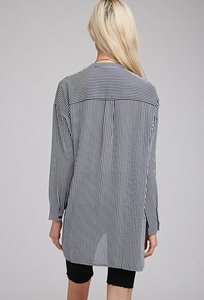}\hss\includegraphics[width=1.10cm,height=1.8cm,keepaspectratio]{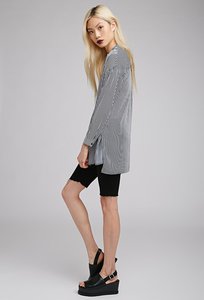}\hss\includegraphics[width=1.10cm,height=1.8cm,keepaspectratio]{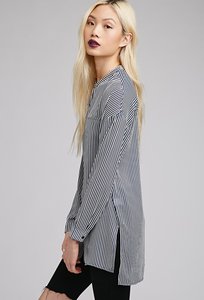}\hss\phantom{\rule{1.10cm}{1.8cm}}\hss}
      \end{minipage}
    }
  \end{minipage}
\hfill
  \begin{minipage}[t]{5.55cm}
    \fcolorbox{red!70!black}{white}{
      \begin{minipage}[t]{5.25cm}
        {\tiny\textbf{\#5}}\\
        \noindent\hbox to 5.25cm{\includegraphics[width=1.10cm,height=1.8cm,keepaspectratio]{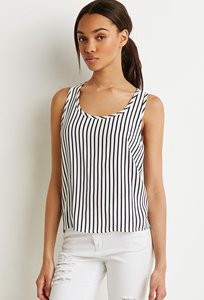}\hss\includegraphics[width=1.10cm,height=1.8cm,keepaspectratio]{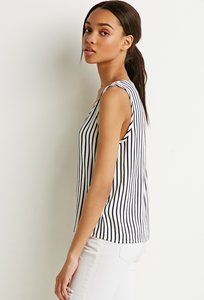}\hss\includegraphics[width=1.10cm,height=1.8cm,keepaspectratio]{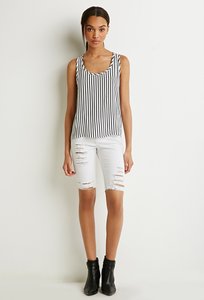}\hss\includegraphics[width=1.10cm,height=1.8cm,keepaspectratio]{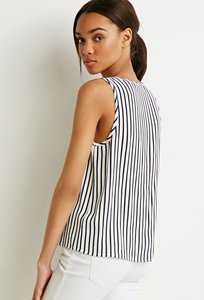}\hss\phantom{\rule{1.10cm}{1.8cm}}\hss}
      \end{minipage}
    }
  \end{minipage}
\hfill
  \begin{minipage}[t]{5.55cm}
    \fcolorbox{red!70!black}{white}{
      \begin{minipage}[t]{5.25cm}
        {\tiny\textbf{\#6}}\\
        \noindent\hbox to 5.25cm{\includegraphics[width=1.10cm,height=1.8cm,keepaspectratio]{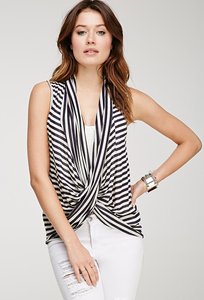}\hss\includegraphics[width=1.10cm,height=1.8cm,keepaspectratio]{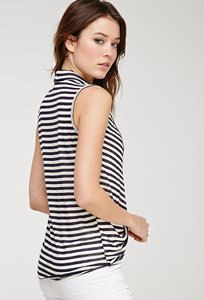}\hss\includegraphics[width=1.10cm,height=1.8cm,keepaspectratio]{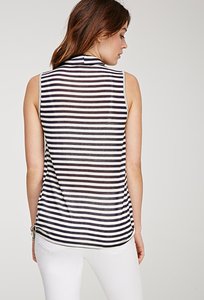}\hss\includegraphics[width=1.10cm,height=1.8cm,keepaspectratio]{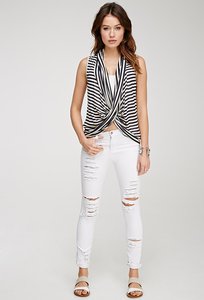}\hss\phantom{\rule{1.10cm}{1.8cm}}\hss}
      \end{minipage}
    }
  \end{minipage}
\par\vspace{2pt}
\noindent
  \begin{minipage}[t]{5.55cm}
    \fcolorbox{red!70!black}{white}{
      \begin{minipage}[t]{5.25cm}
        {\tiny\textbf{\#7}}\\
        \noindent\hbox to 5.25cm{\includegraphics[width=1.10cm,height=1.8cm,keepaspectratio]{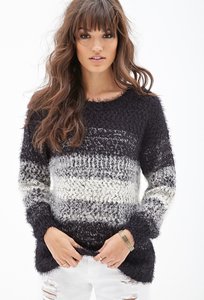}\hss\includegraphics[width=1.10cm,height=1.8cm,keepaspectratio]{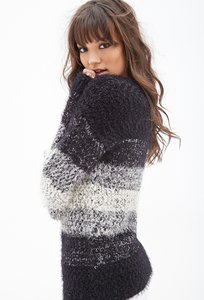}\hss\includegraphics[width=1.10cm,height=1.8cm,keepaspectratio]{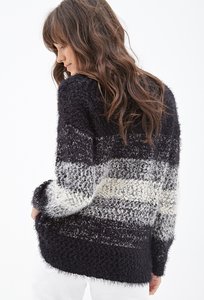}\hss\includegraphics[width=1.10cm,height=1.8cm,keepaspectratio]{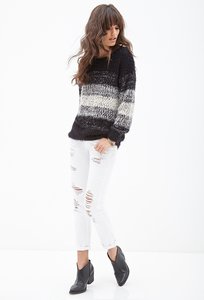}\hss\phantom{\rule{1.10cm}{1.8cm}}\hss}
      \end{minipage}
    }
  \end{minipage}
\hfill
  \begin{minipage}[t]{5.55cm}
    \fcolorbox{red!70!black}{white}{
      \begin{minipage}[t]{5.25cm}
        {\tiny\textbf{\#8}}\\
        \noindent\hbox to 5.25cm{\includegraphics[width=1.10cm,height=1.8cm,keepaspectratio]{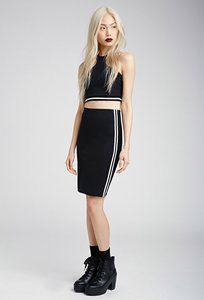}\hss\includegraphics[width=1.10cm,height=1.8cm,keepaspectratio]{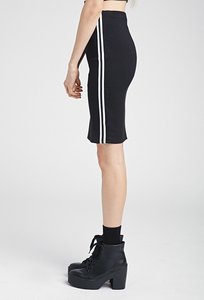}\hss\includegraphics[width=1.10cm,height=1.8cm,keepaspectratio]{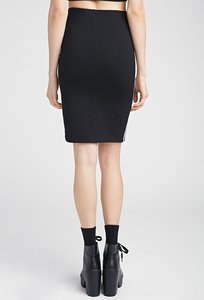}\hss\includegraphics[width=1.10cm,height=1.8cm,keepaspectratio]{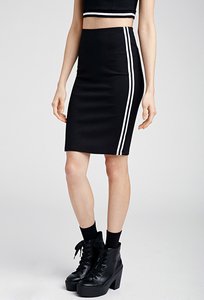}\hss\phantom{\rule{1.10cm}{1.8cm}}\hss}
      \end{minipage}
    }
  \end{minipage}
\hfill
  \begin{minipage}[t]{5.55cm}
    \fcolorbox{red!70!black}{white}{
      \begin{minipage}[t]{5.25cm}
        {\tiny\textbf{\#9}}\\
        \noindent\hbox to 5.25cm{\includegraphics[width=1.10cm,height=1.8cm,keepaspectratio]{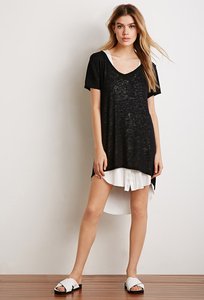}\hss\includegraphics[width=1.10cm,height=1.8cm,keepaspectratio]{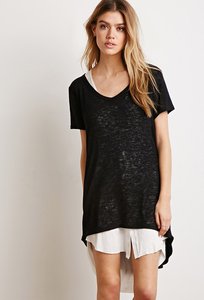}\hss\includegraphics[width=1.10cm,height=1.8cm,keepaspectratio]{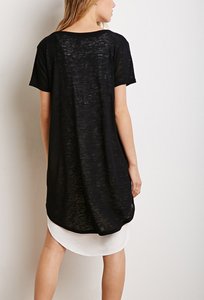}\hss\includegraphics[width=1.10cm,height=1.8cm,keepaspectratio]{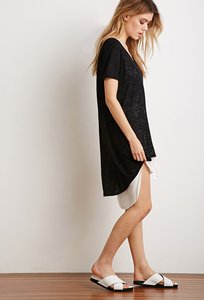}\hss\phantom{\rule{1.10cm}{1.8cm}}\hss}
      \end{minipage}
    }
  \end{minipage}
\par\vspace{2pt}
\noindent
  \begin{minipage}[t]{5.55cm}
    \fcolorbox{red!70!black}{white}{
      \begin{minipage}[t]{5.25cm}
        {\tiny\textbf{\#10}}\\
        \noindent\hbox to 5.25cm{\includegraphics[width=1.10cm,height=1.8cm,keepaspectratio]{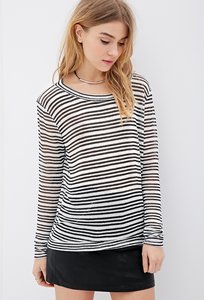}\hss\includegraphics[width=1.10cm,height=1.8cm,keepaspectratio]{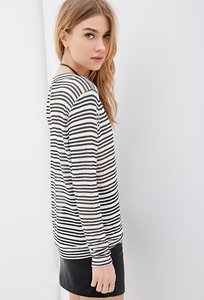}\hss\includegraphics[width=1.10cm,height=1.8cm,keepaspectratio]{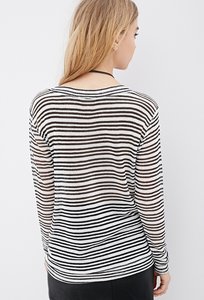}\hss\includegraphics[width=1.10cm,height=1.8cm,keepaspectratio]{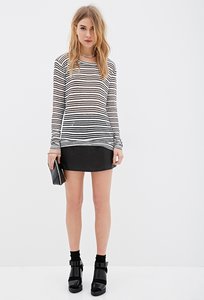}\hss\phantom{\rule{1.10cm}{1.8cm}}\hss}
      \end{minipage}
    }
  \end{minipage}
\hfill
  \begin{minipage}[t]{5.55cm}\end{minipage}
\hfill
  \begin{minipage}[t]{5.55cm}\end{minipage}
\par\vspace{2pt}
\noindent\rule{\linewidth}{0.4pt}
\par\vspace{2pt}
% -- doubao --
{\small\textbf{Doubao-E-V}}\quad{\small Rank~3}\\
\noindent
  \begin{minipage}[t]{5.55cm}
    \fcolorbox{red!70!black}{white}{
      \begin{minipage}[t]{5.25cm}
        {\tiny\textbf{\#1}}\\
        \noindent\hbox to 5.25cm{\includegraphics[width=1.10cm,height=1.8cm,keepaspectratio]{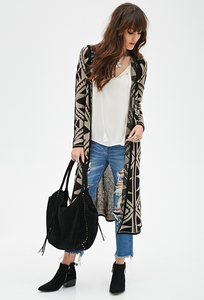}\hss\includegraphics[width=1.10cm,height=1.8cm,keepaspectratio]{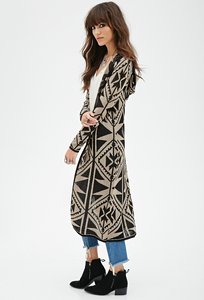}\hss\includegraphics[width=1.10cm,height=1.8cm,keepaspectratio]{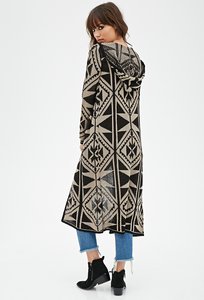}\hss\includegraphics[width=1.10cm,height=1.8cm,keepaspectratio]{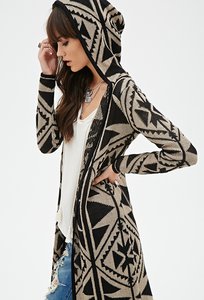}\hss\phantom{\rule{1.10cm}{1.8cm}}\hss}
      \end{minipage}
    }
  \end{minipage}
\hfill
  \begin{minipage}[t]{5.55cm}
    \fcolorbox{red!70!black}{white}{
      \begin{minipage}[t]{5.25cm}
        {\tiny\textbf{\#2}}\\
        \noindent\hbox to 5.25cm{\includegraphics[width=1.10cm,height=1.8cm,keepaspectratio]{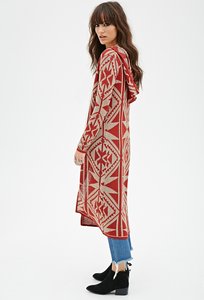}\hss\includegraphics[width=1.10cm,height=1.8cm,keepaspectratio]{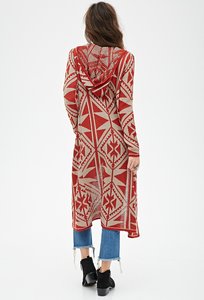}\hss\includegraphics[width=1.10cm,height=1.8cm,keepaspectratio]{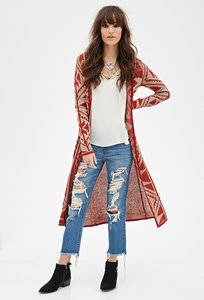}\hss\phantom{\rule{1.10cm}{1.8cm}}\hss\phantom{\rule{1.10cm}{1.8cm}}\hss}
      \end{minipage}
    }
  \end{minipage}
\hfill
  \begin{minipage}[t]{5.55cm}
    \fcolorbox{green!60!black}{white}{
      \begin{minipage}[t]{5.25cm}
        {\tiny\textbf{\#3}}\\
        \noindent\hbox to 5.25cm{\includegraphics[width=1.10cm,height=1.8cm,keepaspectratio]{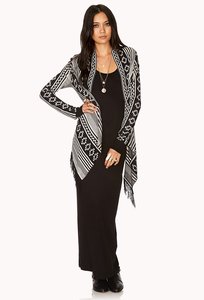}\hss\includegraphics[width=1.10cm,height=1.8cm,keepaspectratio]{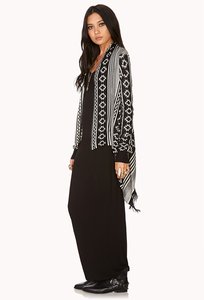}\hss\includegraphics[width=1.10cm,height=1.8cm,keepaspectratio]{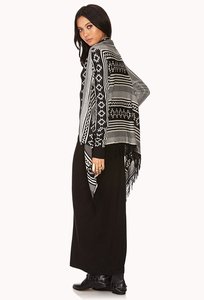}\hss\includegraphics[width=1.10cm,height=1.8cm,keepaspectratio]{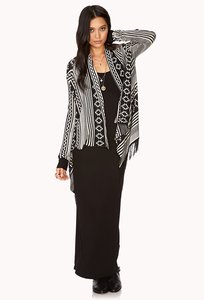}\hss\phantom{\rule{1.10cm}{1.8cm}}\hss}
      \end{minipage}
    }
  \end{minipage}
\par\vspace{2pt}
\noindent
  \begin{minipage}[t]{5.55cm}
    \fcolorbox{red!70!black}{white}{
      \begin{minipage}[t]{5.25cm}
        {\tiny\textbf{\#4}}\\
        \noindent\hbox to 5.25cm{\includegraphics[width=1.10cm,height=1.8cm,keepaspectratio]{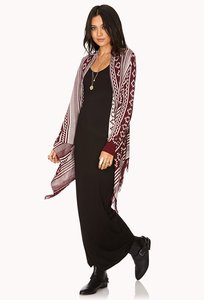}\hss\includegraphics[width=1.10cm,height=1.8cm,keepaspectratio]{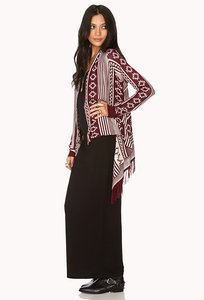}\hss\includegraphics[width=1.10cm,height=1.8cm,keepaspectratio]{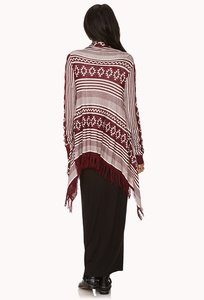}\hss\includegraphics[width=1.10cm,height=1.8cm,keepaspectratio]{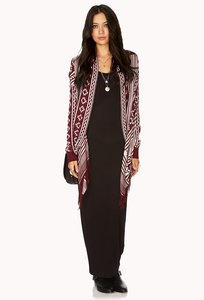}\hss\phantom{\rule{1.10cm}{1.8cm}}\hss}
      \end{minipage}
    }
  \end{minipage}
\hfill
  \begin{minipage}[t]{5.55cm}
    \fcolorbox{red!70!black}{white}{
      \begin{minipage}[t]{5.25cm}
        {\tiny\textbf{\#5}}\\
        \noindent\hbox to 5.25cm{\includegraphics[width=1.10cm,height=1.8cm,keepaspectratio]{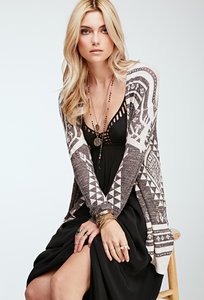}\hss\includegraphics[width=1.10cm,height=1.8cm,keepaspectratio]{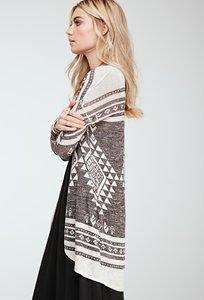}\hss\includegraphics[width=1.10cm,height=1.8cm,keepaspectratio]{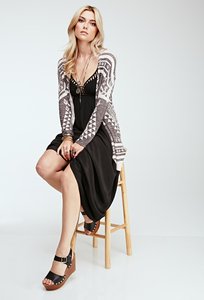}\hss\includegraphics[width=1.10cm,height=1.8cm,keepaspectratio]{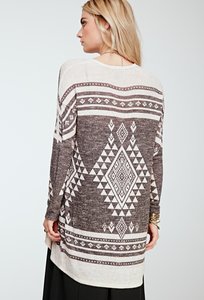}\hss\phantom{\rule{1.10cm}{1.8cm}}\hss}
      \end{minipage}
    }
  \end{minipage}
\hfill
  \begin{minipage}[t]{5.55cm}
    \fcolorbox{red!70!black}{white}{
      \begin{minipage}[t]{5.25cm}
        {\tiny\textbf{\#6}}\\
        \noindent\hbox to 5.25cm{\includegraphics[width=1.10cm,height=1.8cm,keepaspectratio]{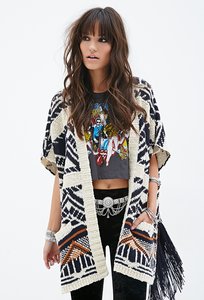}\hss\includegraphics[width=1.10cm,height=1.8cm,keepaspectratio]{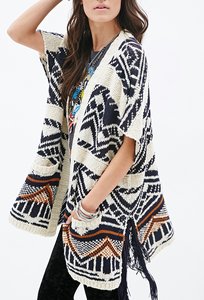}\hss\includegraphics[width=1.10cm,height=1.8cm,keepaspectratio]{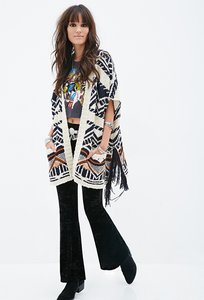}\hss\includegraphics[width=1.10cm,height=1.8cm,keepaspectratio]{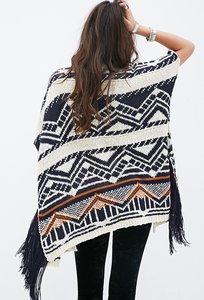}\hss\phantom{\rule{1.10cm}{1.8cm}}\hss}
      \end{minipage}
    }
  \end{minipage}
\par\vspace{2pt}
\noindent
  \begin{minipage}[t]{5.55cm}
    \fcolorbox{red!70!black}{white}{
      \begin{minipage}[t]{5.25cm}
        {\tiny\textbf{\#7}}\\
        \noindent\hbox to 5.25cm{\includegraphics[width=1.10cm,height=1.8cm,keepaspectratio]{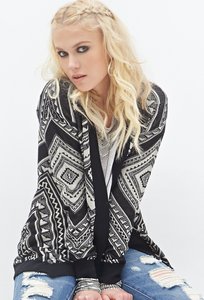}\hss\includegraphics[width=1.10cm,height=1.8cm,keepaspectratio]{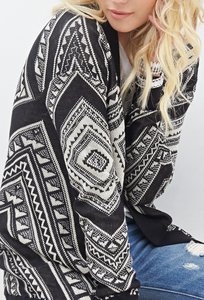}\hss\includegraphics[width=1.10cm,height=1.8cm,keepaspectratio]{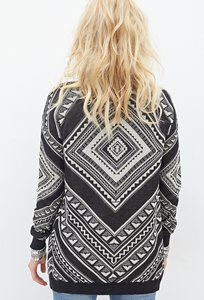}\hss\includegraphics[width=1.10cm,height=1.8cm,keepaspectratio]{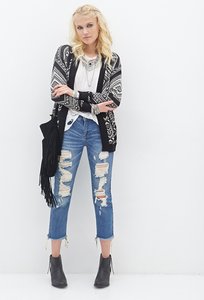}\hss\phantom{\rule{1.10cm}{1.8cm}}\hss}
      \end{minipage}
    }
  \end{minipage}
\hfill
  \begin{minipage}[t]{5.55cm}
    \fcolorbox{red!70!black}{white}{
      \begin{minipage}[t]{5.25cm}
        {\tiny\textbf{\#8}}\\
        \noindent\hbox to 5.25cm{\includegraphics[width=1.10cm,height=1.8cm,keepaspectratio]{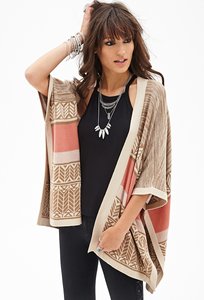}\hss\includegraphics[width=1.10cm,height=1.8cm,keepaspectratio]{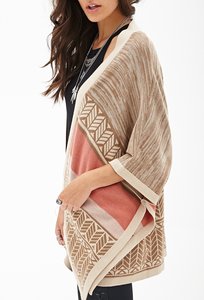}\hss\includegraphics[width=1.10cm,height=1.8cm,keepaspectratio]{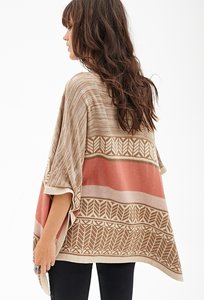}\hss\includegraphics[width=1.10cm,height=1.8cm,keepaspectratio]{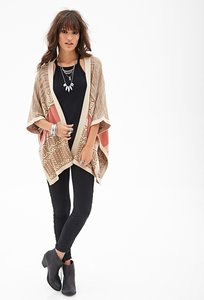}\hss\phantom{\rule{1.10cm}{1.8cm}}\hss}
      \end{minipage}
    }
  \end{minipage}
\hfill
  \begin{minipage}[t]{5.55cm}
    \fcolorbox{red!70!black}{white}{
      \begin{minipage}[t]{5.25cm}
        {\tiny\textbf{\#9}}\\
        \noindent\hbox to 5.25cm{\includegraphics[width=1.10cm,height=1.8cm,keepaspectratio]{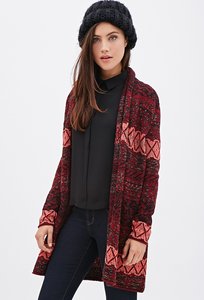}\hss\includegraphics[width=1.10cm,height=1.8cm,keepaspectratio]{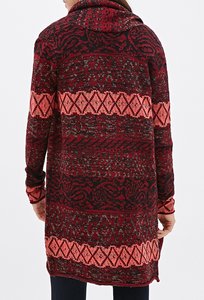}\hss\includegraphics[width=1.10cm,height=1.8cm,keepaspectratio]{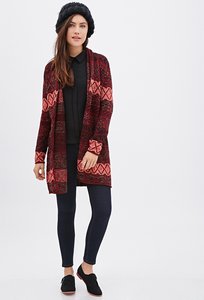}\hss\includegraphics[width=1.10cm,height=1.8cm,keepaspectratio]{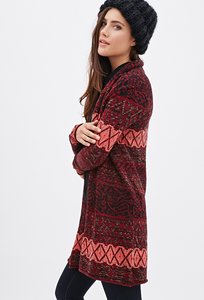}\hss\phantom{\rule{1.10cm}{1.8cm}}\hss}
      \end{minipage}
    }
  \end{minipage}
\par\vspace{2pt}
\noindent
  \begin{minipage}[t]{5.55cm}
    \fcolorbox{red!70!black}{white}{
      \begin{minipage}[t]{5.25cm}
        {\tiny\textbf{\#10}}\\
        \noindent\hbox to 5.25cm{\includegraphics[width=1.10cm,height=1.8cm,keepaspectratio]{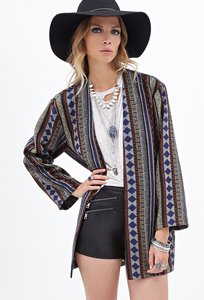}\hss\includegraphics[width=1.10cm,height=1.8cm,keepaspectratio]{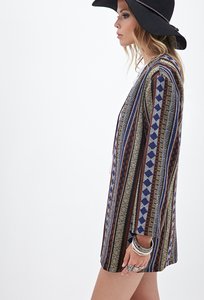}\hss\includegraphics[width=1.10cm,height=1.8cm,keepaspectratio]{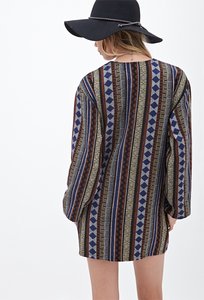}\hss\phantom{\rule{1.10cm}{1.8cm}}\hss\phantom{\rule{1.10cm}{1.8cm}}\hss}
      \end{minipage}
    }
  \end{minipage}
\hfill
  \begin{minipage}[t]{5.55cm}\end{minipage}
\hfill
  \begin{minipage}[t]{5.55cm}\end{minipage}
\par\vspace{2pt}
\noindent\rule{\linewidth}{0.4pt}
\par\vspace{20pt}

\par\vspace{16pt}
\noindent\textbf{\large Example 9}
\par\vspace{4pt}
\noindent\rule{\linewidth}{1.2pt}
\par\vspace{4pt}
% ── Case 9: short::deepfashion::805 ──
\noindent\hfill%
  \begin{minipage}[t]{6.0cm}
    \noindent\hbox to 6.0cm{\includegraphics[width=1.20cm,height=2.0cm,keepaspectratio]{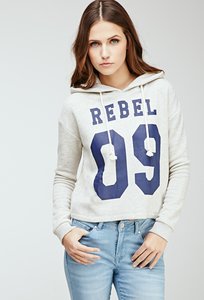}\hss\includegraphics[width=1.20cm,height=2.0cm,keepaspectratio]{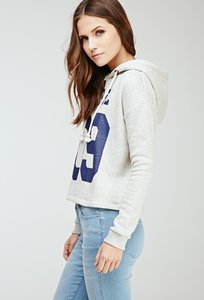}\hss\includegraphics[width=1.20cm,height=2.0cm,keepaspectratio]{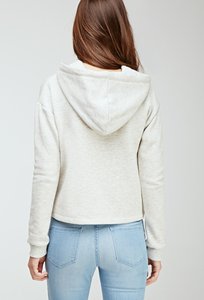}\hss\includegraphics[width=1.20cm,height=2.0cm,keepaspectratio]{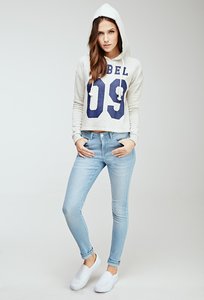}\hss\includegraphics[width=1.20cm,height=2.0cm,keepaspectratio]{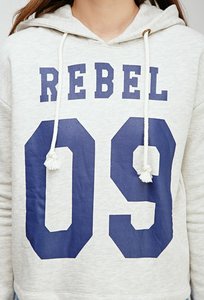}\hss}
    \par\vspace{1pt}
    {\scriptsize\textbf{}}
    \par\vspace{0pt}
    \parbox[t]{6.0cm}{\tiny\raggedright Heather gray cropped hoodie featuring blue 'REBEL 09' collegiate graphic print, drawstring hood with white tassel ties, long sleeves with ribbed cuffs, and a relaxed boxy fit. Plain back, no pockets, waist-length hem. Casual athletic style.}
  \end{minipage}%
\hfill%
  \begin{minipage}[t]{3.5cm}
    \centering
    \vspace{0.45cm}%
    \parbox{3.5cm}{\centering\tiny Replace chest graphic with a front kangaroo pocket, change fabric to marled cranberry red, and add a contrasting solid red lining to the hood interior.}\\[2pt]
    {\normalsize$\longrightarrow$}
  \end{minipage}%
\hfill%
  \begin{minipage}[t]{6.0cm}
    \noindent\hbox to 6.0cm{\includegraphics[width=1.20cm,height=2.0cm,keepaspectratio]{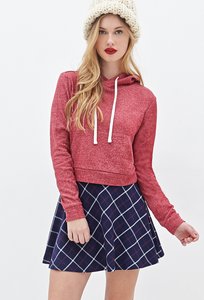}\hss\includegraphics[width=1.20cm,height=2.0cm,keepaspectratio]{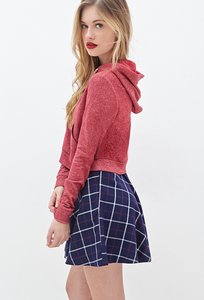}\hss\includegraphics[width=1.20cm,height=2.0cm,keepaspectratio]{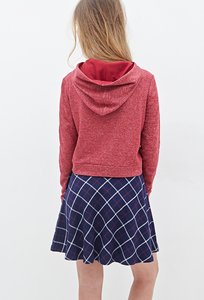}\hss\includegraphics[width=1.20cm,height=2.0cm,keepaspectratio]{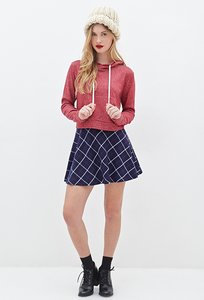}\hss\phantom{\rule{1.20cm}{2.0cm}}\hss}
    \par\vspace{1pt}
    {\scriptsize\textbf{Ground Truth}}
    \par\vspace{0pt}
    \parbox[t]{6.0cm}{\tiny\raggedright A heathered cranberry cropped hoodie featuring a drawstring hood with white strings, a front kangaroo pocket, long sleeves with ribbed cuffs, and a ribbed cropped hem. The marled knit fabric offers a relaxed fit, making it a versatile casual layering piece perfect for pairing with high-waisted skirts or jeans.}
  \end{minipage}%
\hfill
\par\vspace{4pt}
\par\vspace{4pt}
\noindent\rule{\linewidth}{0.4pt}
% -- mt_align --
{\small\textbf{\textbf{Ours}}}\quad{\small Rank~1}\\
\noindent
  \begin{minipage}[t]{5.55cm}
    \fcolorbox{green!60!black}{white}{
      \begin{minipage}[t]{5.25cm}
        {\tiny\textbf{\#1}}\\
        \noindent\hbox to 5.25cm{\includegraphics[width=1.10cm,height=1.8cm,keepaspectratio]{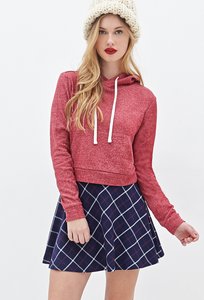}\hss\includegraphics[width=1.10cm,height=1.8cm,keepaspectratio]{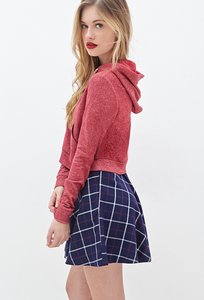}\hss\includegraphics[width=1.10cm,height=1.8cm,keepaspectratio]{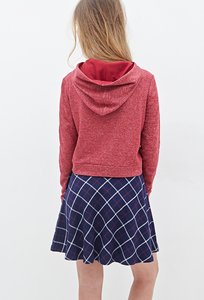}\hss\includegraphics[width=1.10cm,height=1.8cm,keepaspectratio]{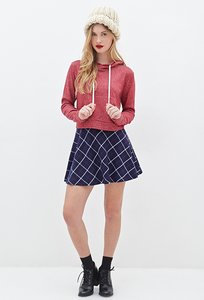}\hss\phantom{\rule{1.10cm}{1.8cm}}\hss}
      \end{minipage}
    }
  \end{minipage}
\hfill
  \begin{minipage}[t]{5.55cm}
    \fcolorbox{red!70!black}{white}{
      \begin{minipage}[t]{5.25cm}
        {\tiny\textbf{\#2}}\\
        \noindent\hbox to 5.25cm{\includegraphics[width=1.10cm,height=1.8cm,keepaspectratio]{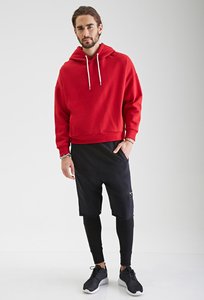}\hss\includegraphics[width=1.10cm,height=1.8cm,keepaspectratio]{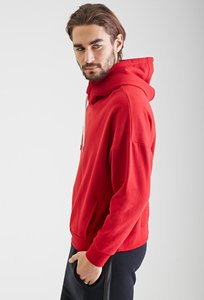}\hss\includegraphics[width=1.10cm,height=1.8cm,keepaspectratio]{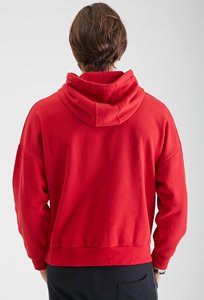}\hss\includegraphics[width=1.10cm,height=1.8cm,keepaspectratio]{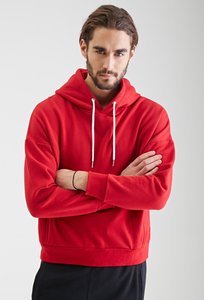}\hss\includegraphics[width=1.10cm,height=1.8cm,keepaspectratio]{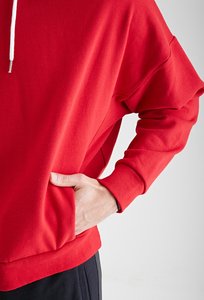}\hss}
      \end{minipage}
    }
  \end{minipage}
\hfill
  \begin{minipage}[t]{5.55cm}
    \fcolorbox{red!70!black}{white}{
      \begin{minipage}[t]{5.25cm}
        {\tiny\textbf{\#3}}\\
        \noindent\hbox to 5.25cm{\includegraphics[width=1.10cm,height=1.8cm,keepaspectratio]{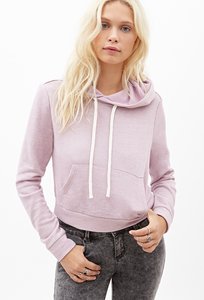}\hss\includegraphics[width=1.10cm,height=1.8cm,keepaspectratio]{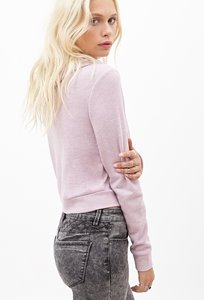}\hss\includegraphics[width=1.10cm,height=1.8cm,keepaspectratio]{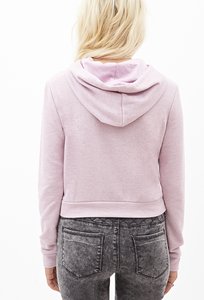}\hss\includegraphics[width=1.10cm,height=1.8cm,keepaspectratio]{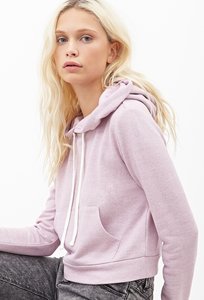}\hss\phantom{\rule{1.10cm}{1.8cm}}\hss}
      \end{minipage}
    }
  \end{minipage}
\par\vspace{2pt}
\noindent
  \begin{minipage}[t]{5.55cm}
    \fcolorbox{red!70!black}{white}{
      \begin{minipage}[t]{5.25cm}
        {\tiny\textbf{\#4}}\\
        \noindent\hbox to 5.25cm{\includegraphics[width=1.10cm,height=1.8cm,keepaspectratio]{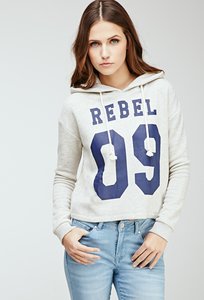}\hss\includegraphics[width=1.10cm,height=1.8cm,keepaspectratio]{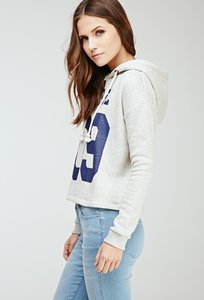}\hss\includegraphics[width=1.10cm,height=1.8cm,keepaspectratio]{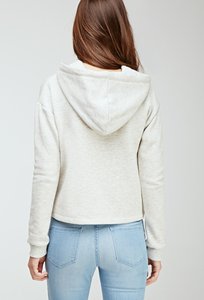}\hss\includegraphics[width=1.10cm,height=1.8cm,keepaspectratio]{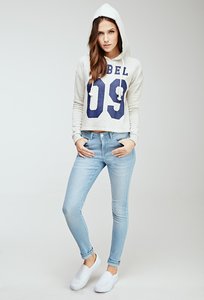}\hss\includegraphics[width=1.10cm,height=1.8cm,keepaspectratio]{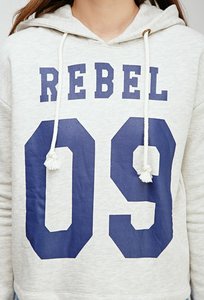}\hss}
      \end{minipage}
    }
  \end{minipage}
\hfill
  \begin{minipage}[t]{5.55cm}
    \fcolorbox{red!70!black}{white}{
      \begin{minipage}[t]{5.25cm}
        {\tiny\textbf{\#5}}\\
        \noindent\hbox to 5.25cm{\includegraphics[width=1.10cm,height=1.8cm,keepaspectratio]{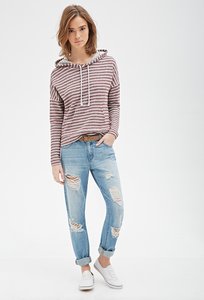}\hss\includegraphics[width=1.10cm,height=1.8cm,keepaspectratio]{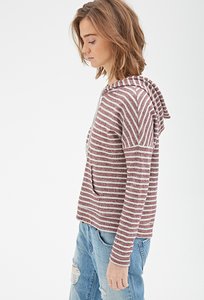}\hss\includegraphics[width=1.10cm,height=1.8cm,keepaspectratio]{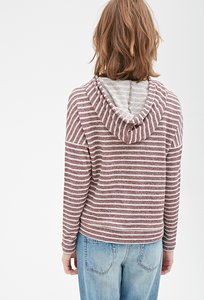}\hss\includegraphics[width=1.10cm,height=1.8cm,keepaspectratio]{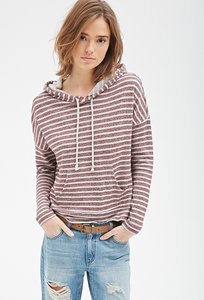}\hss\includegraphics[width=1.10cm,height=1.8cm,keepaspectratio]{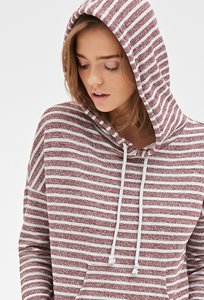}\hss}
      \end{minipage}
    }
  \end{minipage}
\hfill
  \begin{minipage}[t]{5.55cm}
    \fcolorbox{red!70!black}{white}{
      \begin{minipage}[t]{5.25cm}
        {\tiny\textbf{\#6}}\\
        \noindent\hbox to 5.25cm{\includegraphics[width=1.10cm,height=1.8cm,keepaspectratio]{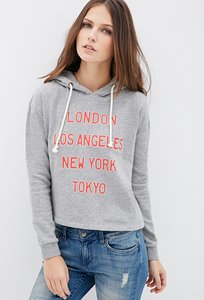}\hss\includegraphics[width=1.10cm,height=1.8cm,keepaspectratio]{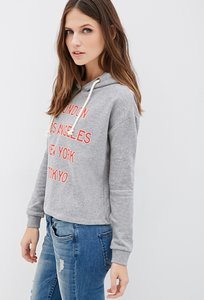}\hss\includegraphics[width=1.10cm,height=1.8cm,keepaspectratio]{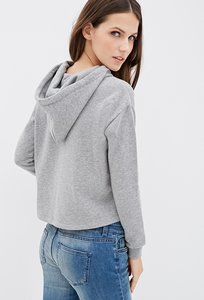}\hss\includegraphics[width=1.10cm,height=1.8cm,keepaspectratio]{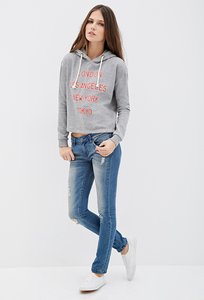}\hss\includegraphics[width=1.10cm,height=1.8cm,keepaspectratio]{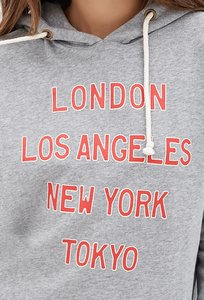}\hss}
      \end{minipage}
    }
  \end{minipage}
\par\vspace{2pt}
\noindent
  \begin{minipage}[t]{5.55cm}
    \fcolorbox{red!70!black}{white}{
      \begin{minipage}[t]{5.25cm}
        {\tiny\textbf{\#7}}\\
        \noindent\hbox to 5.25cm{\includegraphics[width=1.10cm,height=1.8cm,keepaspectratio]{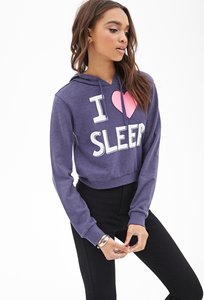}\hss\includegraphics[width=1.10cm,height=1.8cm,keepaspectratio]{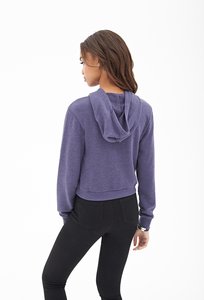}\hss\includegraphics[width=1.10cm,height=1.8cm,keepaspectratio]{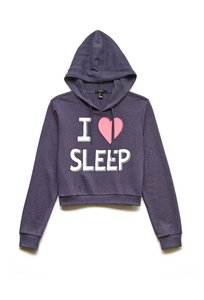}\hss\includegraphics[width=1.10cm,height=1.8cm,keepaspectratio]{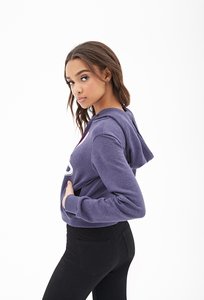}\hss\phantom{\rule{1.10cm}{1.8cm}}\hss}
      \end{minipage}
    }
  \end{minipage}
\hfill
  \begin{minipage}[t]{5.55cm}
    \fcolorbox{red!70!black}{white}{
      \begin{minipage}[t]{5.25cm}
        {\tiny\textbf{\#8}}\\
        \noindent\hbox to 5.25cm{\includegraphics[width=1.10cm,height=1.8cm,keepaspectratio]{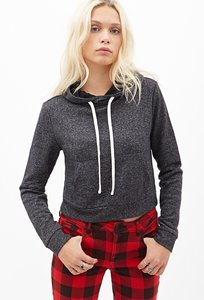}\hss\includegraphics[width=1.10cm,height=1.8cm,keepaspectratio]{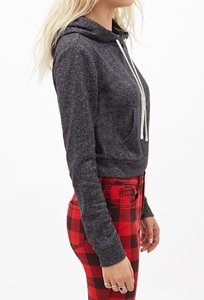}\hss\includegraphics[width=1.10cm,height=1.8cm,keepaspectratio]{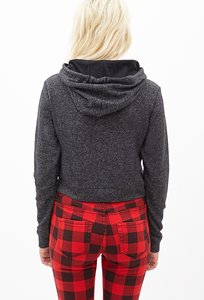}\hss\includegraphics[width=1.10cm,height=1.8cm,keepaspectratio]{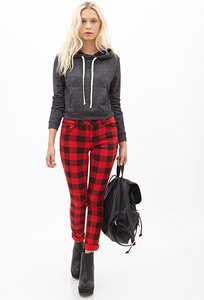}\hss\phantom{\rule{1.10cm}{1.8cm}}\hss}
      \end{minipage}
    }
  \end{minipage}
\hfill
  \begin{minipage}[t]{5.55cm}
    \fcolorbox{red!70!black}{white}{
      \begin{minipage}[t]{5.25cm}
        {\tiny\textbf{\#9}}\\
        \noindent\hbox to 5.25cm{\includegraphics[width=1.10cm,height=1.8cm,keepaspectratio]{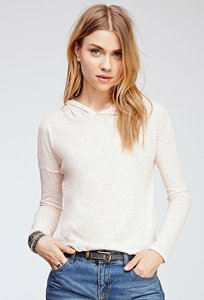}\hss\includegraphics[width=1.10cm,height=1.8cm,keepaspectratio]{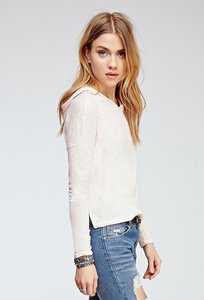}\hss\includegraphics[width=1.10cm,height=1.8cm,keepaspectratio]{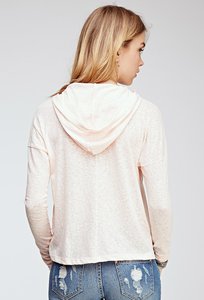}\hss\includegraphics[width=1.10cm,height=1.8cm,keepaspectratio]{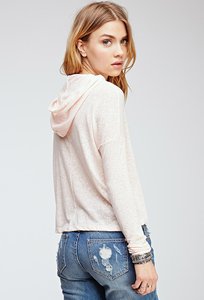}\hss\phantom{\rule{1.10cm}{1.8cm}}\hss}
      \end{minipage}
    }
  \end{minipage}
\par\vspace{2pt}
\noindent
  \begin{minipage}[t]{5.55cm}
    \fcolorbox{red!70!black}{white}{
      \begin{minipage}[t]{5.25cm}
        {\tiny\textbf{\#10}}\\
        \noindent\hbox to 5.25cm{\includegraphics[width=1.10cm,height=1.8cm,keepaspectratio]{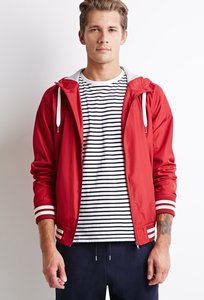}\hss\includegraphics[width=1.10cm,height=1.8cm,keepaspectratio]{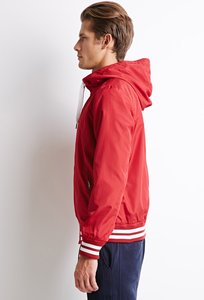}\hss\includegraphics[width=1.10cm,height=1.8cm,keepaspectratio]{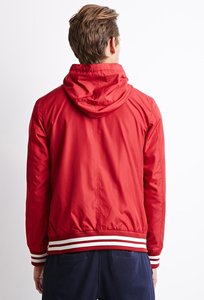}\hss\includegraphics[width=1.10cm,height=1.8cm,keepaspectratio]{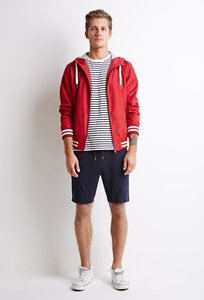}\hss\includegraphics[width=1.10cm,height=1.8cm,keepaspectratio]{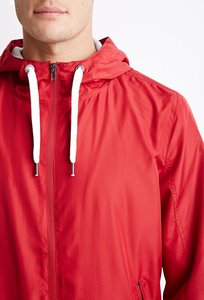}\hss}
      \end{minipage}
    }
  \end{minipage}
\hfill
  \begin{minipage}[t]{5.55cm}\end{minipage}
\hfill
  \begin{minipage}[t]{5.55cm}\end{minipage}
\par\vspace{2pt}
\noindent\rule{\linewidth}{0.4pt}
\par\vspace{2pt}
% -- qwen3_vl_2b --
{\small\textbf{Qwen3-VL-2B}}\quad{\small Rank~2}\\
\noindent
  \begin{minipage}[t]{5.55cm}
    \fcolorbox{red!70!black}{white}{
      \begin{minipage}[t]{5.25cm}
        {\tiny\textbf{\#1}}\\
        \noindent\hbox to 5.25cm{\includegraphics[width=1.10cm,height=1.8cm,keepaspectratio]{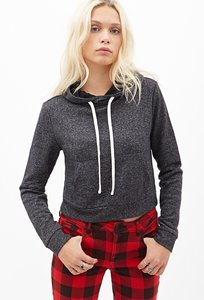}\hss\includegraphics[width=1.10cm,height=1.8cm,keepaspectratio]{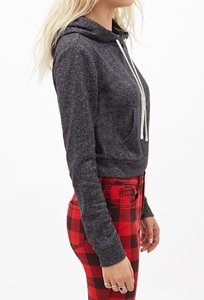}\hss\includegraphics[width=1.10cm,height=1.8cm,keepaspectratio]{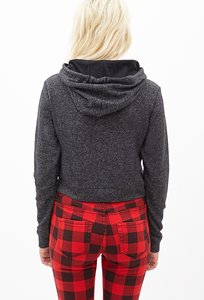}\hss\includegraphics[width=1.10cm,height=1.8cm,keepaspectratio]{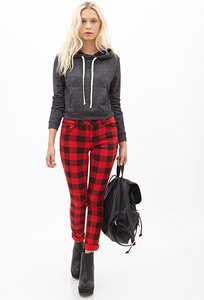}\hss\phantom{\rule{1.10cm}{1.8cm}}\hss}
      \end{minipage}
    }
  \end{minipage}
\hfill
  \begin{minipage}[t]{5.55cm}
    \fcolorbox{green!60!black}{white}{
      \begin{minipage}[t]{5.25cm}
        {\tiny\textbf{\#2}}\\
        \noindent\hbox to 5.25cm{\includegraphics[width=1.10cm,height=1.8cm,keepaspectratio]{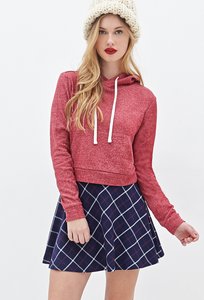}\hss\includegraphics[width=1.10cm,height=1.8cm,keepaspectratio]{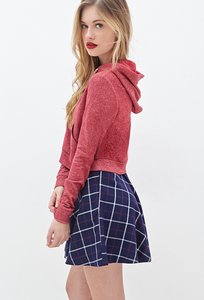}\hss\includegraphics[width=1.10cm,height=1.8cm,keepaspectratio]{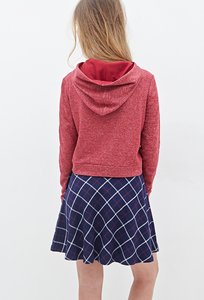}\hss\includegraphics[width=1.10cm,height=1.8cm,keepaspectratio]{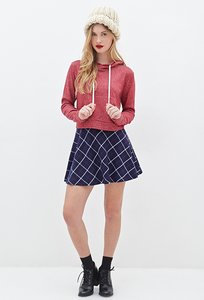}\hss\phantom{\rule{1.10cm}{1.8cm}}\hss}
      \end{minipage}
    }
  \end{minipage}
\hfill
  \begin{minipage}[t]{5.55cm}
    \fcolorbox{red!70!black}{white}{
      \begin{minipage}[t]{5.25cm}
        {\tiny\textbf{\#3}}\\
        \noindent\hbox to 5.25cm{\includegraphics[width=1.10cm,height=1.8cm,keepaspectratio]{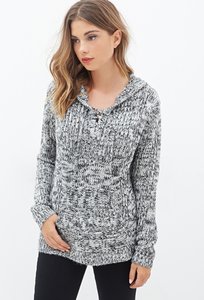}\hss\includegraphics[width=1.10cm,height=1.8cm,keepaspectratio]{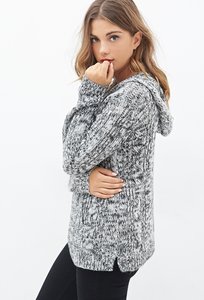}\hss\includegraphics[width=1.10cm,height=1.8cm,keepaspectratio]{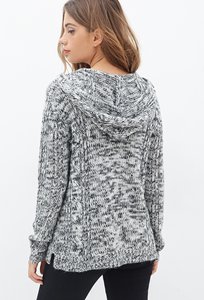}\hss\includegraphics[width=1.10cm,height=1.8cm,keepaspectratio]{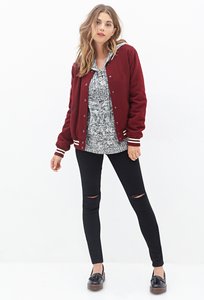}\hss\phantom{\rule{1.10cm}{1.8cm}}\hss}
      \end{minipage}
    }
  \end{minipage}
\par\vspace{2pt}
\noindent
  \begin{minipage}[t]{5.55cm}
    \fcolorbox{red!70!black}{white}{
      \begin{minipage}[t]{5.25cm}
        {\tiny\textbf{\#4}}\\
        \noindent\hbox to 5.25cm{\includegraphics[width=1.10cm,height=1.8cm,keepaspectratio]{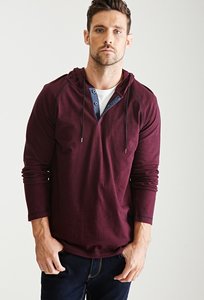}\hss\includegraphics[width=1.10cm,height=1.8cm,keepaspectratio]{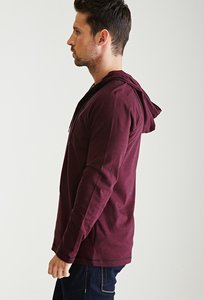}\hss\includegraphics[width=1.10cm,height=1.8cm,keepaspectratio]{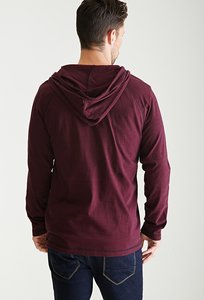}\hss\includegraphics[width=1.10cm,height=1.8cm,keepaspectratio]{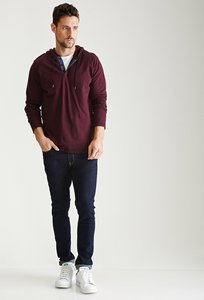}\hss\includegraphics[width=1.10cm,height=1.8cm,keepaspectratio]{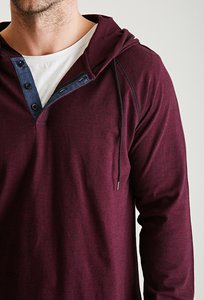}\hss}
      \end{minipage}
    }
  \end{minipage}
\hfill
  \begin{minipage}[t]{5.55cm}
    \fcolorbox{red!70!black}{white}{
      \begin{minipage}[t]{5.25cm}
        {\tiny\textbf{\#5}}\\
        \noindent\hbox to 5.25cm{\includegraphics[width=1.10cm,height=1.8cm,keepaspectratio]{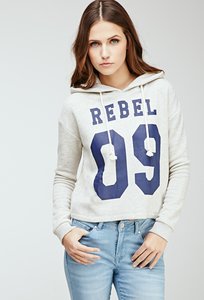}\hss\includegraphics[width=1.10cm,height=1.8cm,keepaspectratio]{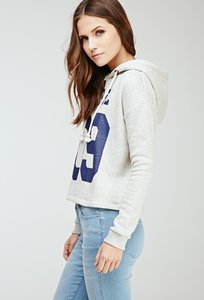}\hss\includegraphics[width=1.10cm,height=1.8cm,keepaspectratio]{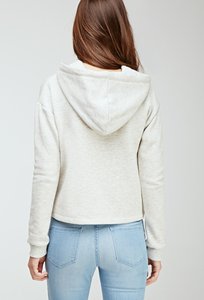}\hss\includegraphics[width=1.10cm,height=1.8cm,keepaspectratio]{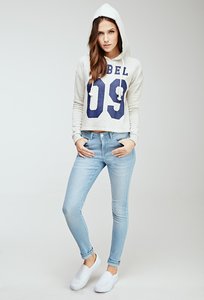}\hss\includegraphics[width=1.10cm,height=1.8cm,keepaspectratio]{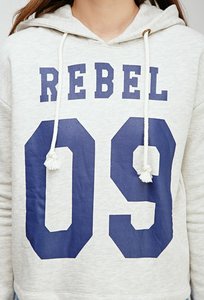}\hss}
      \end{minipage}
    }
  \end{minipage}
\hfill
  \begin{minipage}[t]{5.55cm}
    \fcolorbox{red!70!black}{white}{
      \begin{minipage}[t]{5.25cm}
        {\tiny\textbf{\#6}}\\
        \noindent\hbox to 5.25cm{\includegraphics[width=1.10cm,height=1.8cm,keepaspectratio]{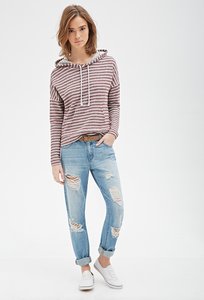}\hss\includegraphics[width=1.10cm,height=1.8cm,keepaspectratio]{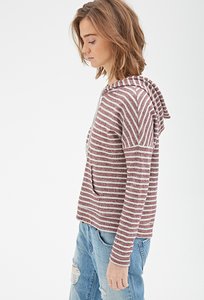}\hss\includegraphics[width=1.10cm,height=1.8cm,keepaspectratio]{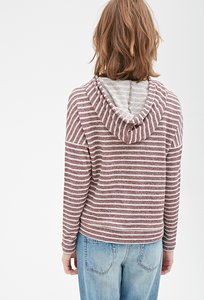}\hss\includegraphics[width=1.10cm,height=1.8cm,keepaspectratio]{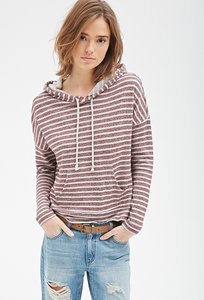}\hss\includegraphics[width=1.10cm,height=1.8cm,keepaspectratio]{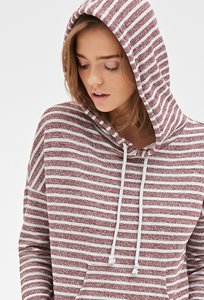}\hss}
      \end{minipage}
    }
  \end{minipage}
\par\vspace{2pt}
\noindent
  \begin{minipage}[t]{5.55cm}
    \fcolorbox{red!70!black}{white}{
      \begin{minipage}[t]{5.25cm}
        {\tiny\textbf{\#7}}\\
        \noindent\hbox to 5.25cm{\includegraphics[width=1.10cm,height=1.8cm,keepaspectratio]{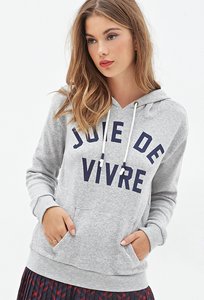}\hss\includegraphics[width=1.10cm,height=1.8cm,keepaspectratio]{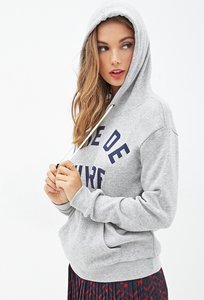}\hss\includegraphics[width=1.10cm,height=1.8cm,keepaspectratio]{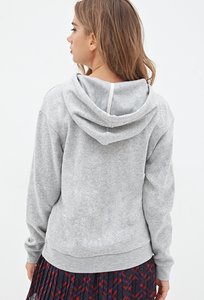}\hss\includegraphics[width=1.10cm,height=1.8cm,keepaspectratio]{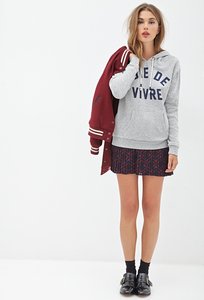}\hss\phantom{\rule{1.10cm}{1.8cm}}\hss}
      \end{minipage}
    }
  \end{minipage}
\hfill
  \begin{minipage}[t]{5.55cm}
    \fcolorbox{red!70!black}{white}{
      \begin{minipage}[t]{5.25cm}
        {\tiny\textbf{\#8}}\\
        \noindent\hbox to 5.25cm{\includegraphics[width=1.10cm,height=1.8cm,keepaspectratio]{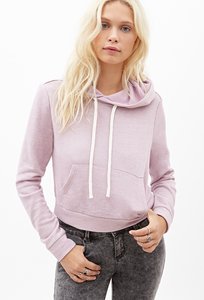}\hss\includegraphics[width=1.10cm,height=1.8cm,keepaspectratio]{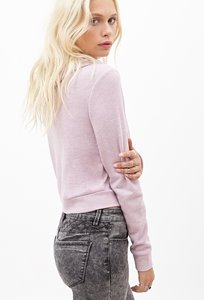}\hss\includegraphics[width=1.10cm,height=1.8cm,keepaspectratio]{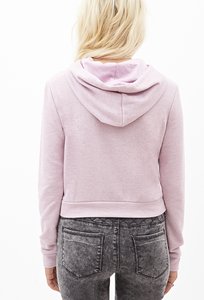}\hss\includegraphics[width=1.10cm,height=1.8cm,keepaspectratio]{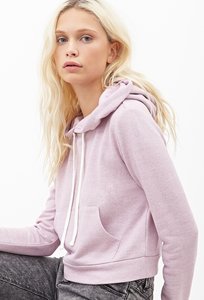}\hss\phantom{\rule{1.10cm}{1.8cm}}\hss}
      \end{minipage}
    }
  \end{minipage}
\hfill
  \begin{minipage}[t]{5.55cm}
    \fcolorbox{red!70!black}{white}{
      \begin{minipage}[t]{5.25cm}
        {\tiny\textbf{\#9}}\\
        \noindent\hbox to 5.25cm{\includegraphics[width=1.10cm,height=1.8cm,keepaspectratio]{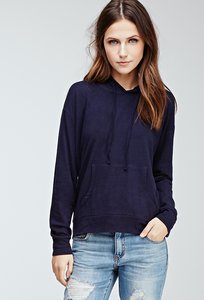}\hss\includegraphics[width=1.10cm,height=1.8cm,keepaspectratio]{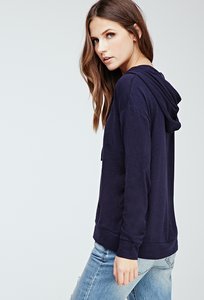}\hss\includegraphics[width=1.10cm,height=1.8cm,keepaspectratio]{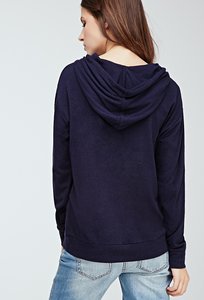}\hss\includegraphics[width=1.10cm,height=1.8cm,keepaspectratio]{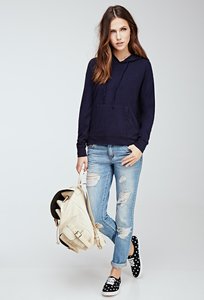}\hss\includegraphics[width=1.10cm,height=1.8cm,keepaspectratio]{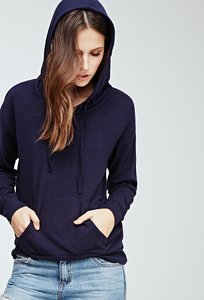}\hss}
      \end{minipage}
    }
  \end{minipage}
\par\vspace{2pt}
\noindent
  \begin{minipage}[t]{5.55cm}
    \fcolorbox{red!70!black}{white}{
      \begin{minipage}[t]{5.25cm}
        {\tiny\textbf{\#10}}\\
        \noindent\hbox to 5.25cm{\includegraphics[width=1.10cm,height=1.8cm,keepaspectratio]{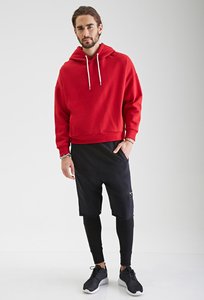}\hss\includegraphics[width=1.10cm,height=1.8cm,keepaspectratio]{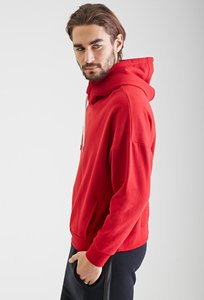}\hss\includegraphics[width=1.10cm,height=1.8cm,keepaspectratio]{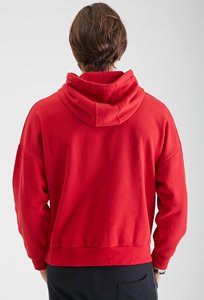}\hss\includegraphics[width=1.10cm,height=1.8cm,keepaspectratio]{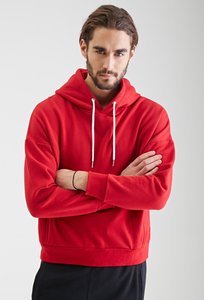}\hss\includegraphics[width=1.10cm,height=1.8cm,keepaspectratio]{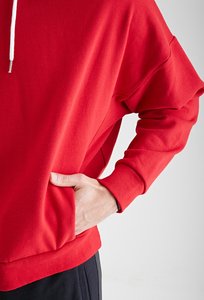}\hss}
      \end{minipage}
    }
  \end{minipage}
\hfill
  \begin{minipage}[t]{5.55cm}\end{minipage}
\hfill
  \begin{minipage}[t]{5.55cm}\end{minipage}
\par\vspace{2pt}
\noindent\rule{\linewidth}{0.4pt}
\par\vspace{2pt}
% -- qwen3_vl_8b --
{\small\textbf{Qwen3-VL-8B}}\quad{\small Rank~4}\\
\noindent
  \begin{minipage}[t]{5.55cm}
    \fcolorbox{red!70!black}{white}{
      \begin{minipage}[t]{5.25cm}
        {\tiny\textbf{\#1}}\\
        \noindent\hbox to 5.25cm{\includegraphics[width=1.10cm,height=1.8cm,keepaspectratio]{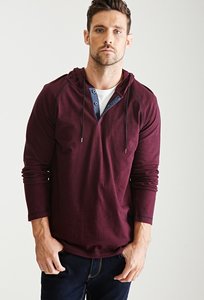}\hss\includegraphics[width=1.10cm,height=1.8cm,keepaspectratio]{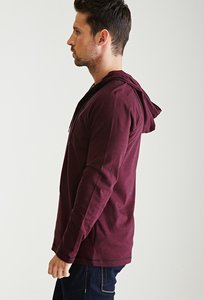}\hss\includegraphics[width=1.10cm,height=1.8cm,keepaspectratio]{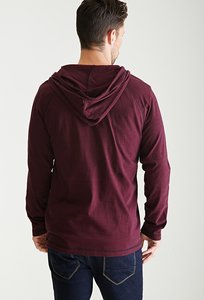}\hss\includegraphics[width=1.10cm,height=1.8cm,keepaspectratio]{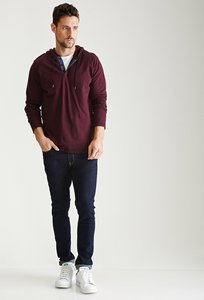}\hss\includegraphics[width=1.10cm,height=1.8cm,keepaspectratio]{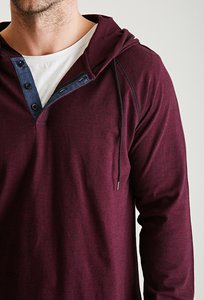}\hss}
      \end{minipage}
    }
  \end{minipage}
\hfill
  \begin{minipage}[t]{5.55cm}
    \fcolorbox{red!70!black}{white}{
      \begin{minipage}[t]{5.25cm}
        {\tiny\textbf{\#2}}\\
        \noindent\hbox to 5.25cm{\includegraphics[width=1.10cm,height=1.8cm,keepaspectratio]{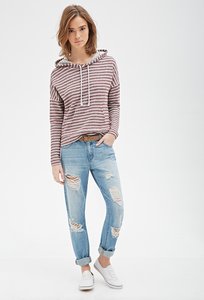}\hss\includegraphics[width=1.10cm,height=1.8cm,keepaspectratio]{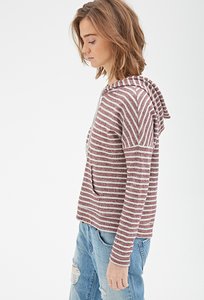}\hss\includegraphics[width=1.10cm,height=1.8cm,keepaspectratio]{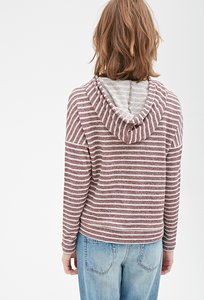}\hss\includegraphics[width=1.10cm,height=1.8cm,keepaspectratio]{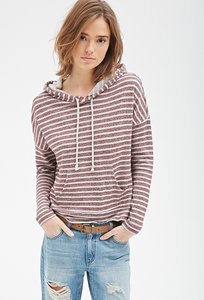}\hss\includegraphics[width=1.10cm,height=1.8cm,keepaspectratio]{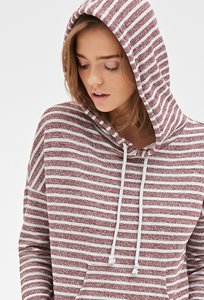}\hss}
      \end{minipage}
    }
  \end{minipage}
\hfill
  \begin{minipage}[t]{5.55cm}
    \fcolorbox{red!70!black}{white}{
      \begin{minipage}[t]{5.25cm}
        {\tiny\textbf{\#3}}\\
        \noindent\hbox to 5.25cm{\includegraphics[width=1.10cm,height=1.8cm,keepaspectratio]{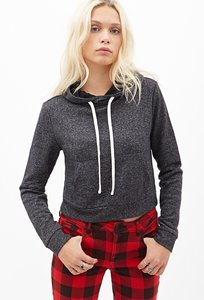}\hss\includegraphics[width=1.10cm,height=1.8cm,keepaspectratio]{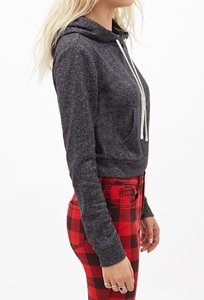}\hss\includegraphics[width=1.10cm,height=1.8cm,keepaspectratio]{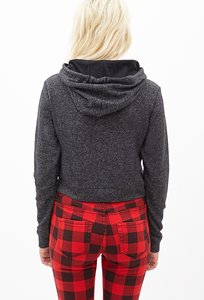}\hss\includegraphics[width=1.10cm,height=1.8cm,keepaspectratio]{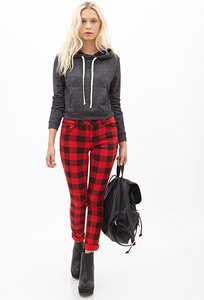}\hss\phantom{\rule{1.10cm}{1.8cm}}\hss}
      \end{minipage}
    }
  \end{minipage}
\par\vspace{2pt}
\noindent
  \begin{minipage}[t]{5.55cm}
    \fcolorbox{green!60!black}{white}{
      \begin{minipage}[t]{5.25cm}
        {\tiny\textbf{\#4}}\\
        \noindent\hbox to 5.25cm{\includegraphics[width=1.10cm,height=1.8cm,keepaspectratio]{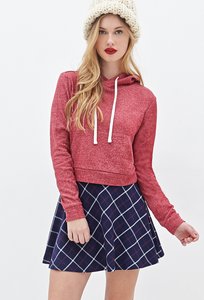}\hss\includegraphics[width=1.10cm,height=1.8cm,keepaspectratio]{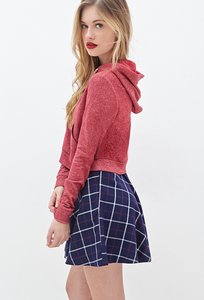}\hss\includegraphics[width=1.10cm,height=1.8cm,keepaspectratio]{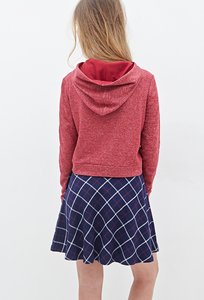}\hss\includegraphics[width=1.10cm,height=1.8cm,keepaspectratio]{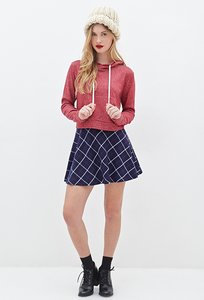}\hss\phantom{\rule{1.10cm}{1.8cm}}\hss}
      \end{minipage}
    }
  \end{minipage}
\hfill
  \begin{minipage}[t]{5.55cm}
    \fcolorbox{red!70!black}{white}{
      \begin{minipage}[t]{5.25cm}
        {\tiny\textbf{\#5}}\\
        \noindent\hbox to 5.25cm{\includegraphics[width=1.10cm,height=1.8cm,keepaspectratio]{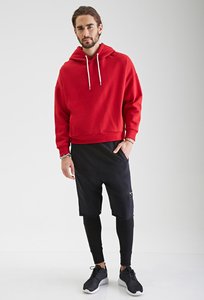}\hss\includegraphics[width=1.10cm,height=1.8cm,keepaspectratio]{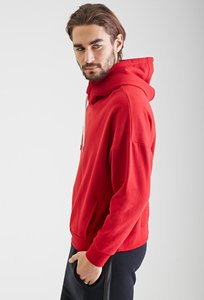}\hss\includegraphics[width=1.10cm,height=1.8cm,keepaspectratio]{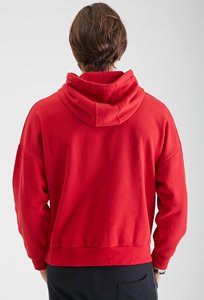}\hss\includegraphics[width=1.10cm,height=1.8cm,keepaspectratio]{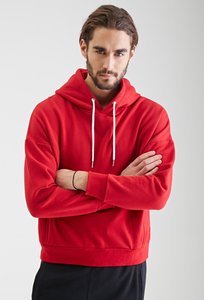}\hss\includegraphics[width=1.10cm,height=1.8cm,keepaspectratio]{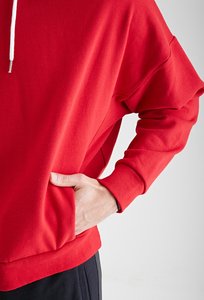}\hss}
      \end{minipage}
    }
  \end{minipage}
\hfill
  \begin{minipage}[t]{5.55cm}
    \fcolorbox{red!70!black}{white}{
      \begin{minipage}[t]{5.25cm}
        {\tiny\textbf{\#6}}\\
        \noindent\hbox to 5.25cm{\includegraphics[width=1.10cm,height=1.8cm,keepaspectratio]{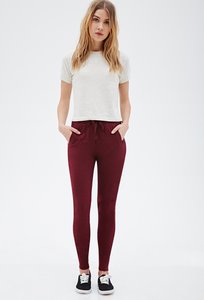}\hss\includegraphics[width=1.10cm,height=1.8cm,keepaspectratio]{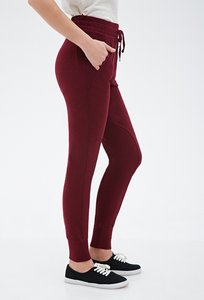}\hss\includegraphics[width=1.10cm,height=1.8cm,keepaspectratio]{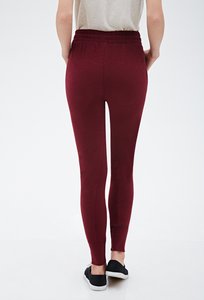}\hss\includegraphics[width=1.10cm,height=1.8cm,keepaspectratio]{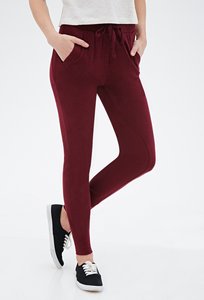}\hss\phantom{\rule{1.10cm}{1.8cm}}\hss}
      \end{minipage}
    }
  \end{minipage}
\par\vspace{2pt}
\noindent
  \begin{minipage}[t]{5.55cm}
    \fcolorbox{red!70!black}{white}{
      \begin{minipage}[t]{5.25cm}
        {\tiny\textbf{\#7}}\\
        \noindent\hbox to 5.25cm{\includegraphics[width=1.10cm,height=1.8cm,keepaspectratio]{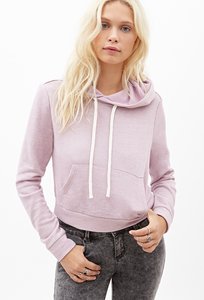}\hss\includegraphics[width=1.10cm,height=1.8cm,keepaspectratio]{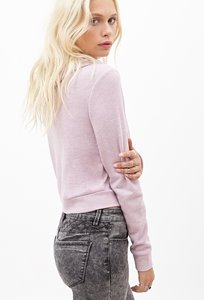}\hss\includegraphics[width=1.10cm,height=1.8cm,keepaspectratio]{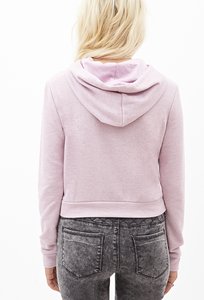}\hss\includegraphics[width=1.10cm,height=1.8cm,keepaspectratio]{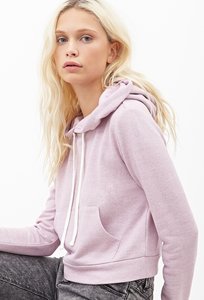}\hss\phantom{\rule{1.10cm}{1.8cm}}\hss}
      \end{minipage}
    }
  \end{minipage}
\hfill
  \begin{minipage}[t]{5.55cm}
    \fcolorbox{red!70!black}{white}{
      \begin{minipage}[t]{5.25cm}
        {\tiny\textbf{\#8}}\\
        \noindent\hbox to 5.25cm{\includegraphics[width=1.10cm,height=1.8cm,keepaspectratio]{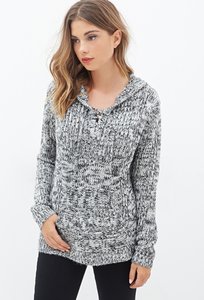}\hss\includegraphics[width=1.10cm,height=1.8cm,keepaspectratio]{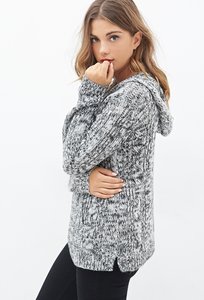}\hss\includegraphics[width=1.10cm,height=1.8cm,keepaspectratio]{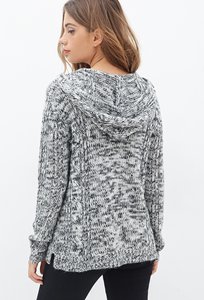}\hss\includegraphics[width=1.10cm,height=1.8cm,keepaspectratio]{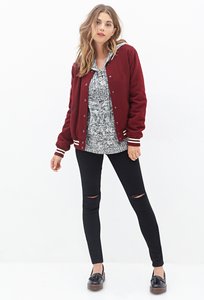}\hss\phantom{\rule{1.10cm}{1.8cm}}\hss}
      \end{minipage}
    }
  \end{minipage}
\hfill
  \begin{minipage}[t]{5.55cm}
    \fcolorbox{red!70!black}{white}{
      \begin{minipage}[t]{5.25cm}
        {\tiny\textbf{\#9}}\\
        \noindent\hbox to 5.25cm{\includegraphics[width=1.10cm,height=1.8cm,keepaspectratio]{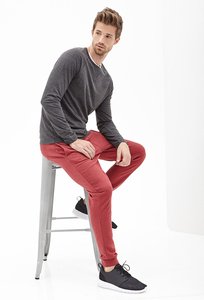}\hss\includegraphics[width=1.10cm,height=1.8cm,keepaspectratio]{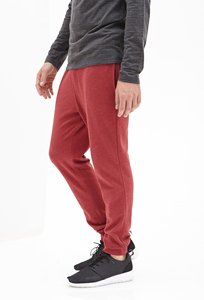}\hss\includegraphics[width=1.10cm,height=1.8cm,keepaspectratio]{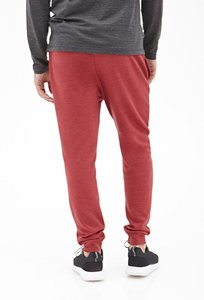}\hss\includegraphics[width=1.10cm,height=1.8cm,keepaspectratio]{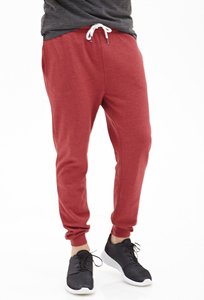}\hss\phantom{\rule{1.10cm}{1.8cm}}\hss}
      \end{minipage}
    }
  \end{minipage}
\par\vspace{2pt}
\noindent
  \begin{minipage}[t]{5.55cm}
    \fcolorbox{red!70!black}{white}{
      \begin{minipage}[t]{5.25cm}
        {\tiny\textbf{\#10}}\\
        \noindent\hbox to 5.25cm{\includegraphics[width=1.10cm,height=1.8cm,keepaspectratio]{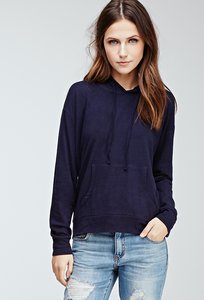}\hss\includegraphics[width=1.10cm,height=1.8cm,keepaspectratio]{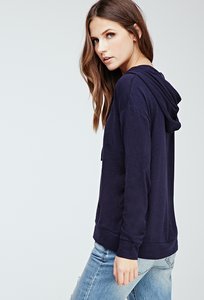}\hss\includegraphics[width=1.10cm,height=1.8cm,keepaspectratio]{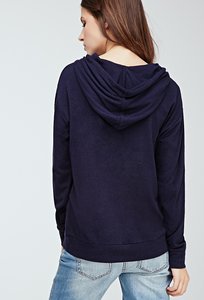}\hss\includegraphics[width=1.10cm,height=1.8cm,keepaspectratio]{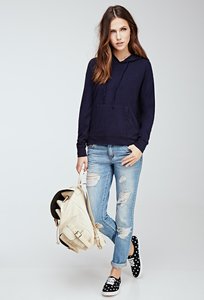}\hss\includegraphics[width=1.10cm,height=1.8cm,keepaspectratio]{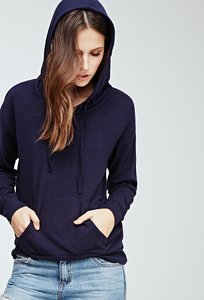}\hss}
      \end{minipage}
    }
  \end{minipage}
\hfill
  \begin{minipage}[t]{5.55cm}\end{minipage}
\hfill
  \begin{minipage}[t]{5.55cm}\end{minipage}
\par\vspace{2pt}
\noindent\rule{\linewidth}{0.4pt}
\par\vspace{2pt}
% -- reznembed --
{\small\textbf{RezNEmbed}}\quad{\small Not in Top-10}\\
\noindent
  \begin{minipage}[t]{5.55cm}
    \fcolorbox{red!70!black}{white}{
      \begin{minipage}[t]{5.25cm}
        {\tiny\textbf{\#1}}\\
        \noindent\hbox to 5.25cm{\includegraphics[width=1.10cm,height=1.8cm,keepaspectratio]{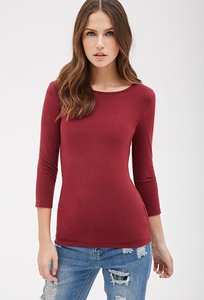}\hss\includegraphics[width=1.10cm,height=1.8cm,keepaspectratio]{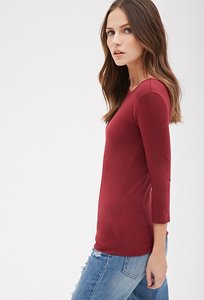}\hss\includegraphics[width=1.10cm,height=1.8cm,keepaspectratio]{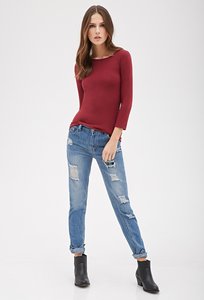}\hss\includegraphics[width=1.10cm,height=1.8cm,keepaspectratio]{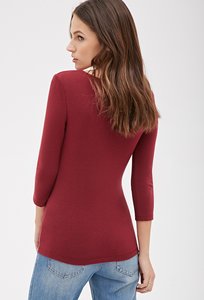}\hss\phantom{\rule{1.10cm}{1.8cm}}\hss}
      \end{minipage}
    }
  \end{minipage}
\hfill
  \begin{minipage}[t]{5.55cm}
    \fcolorbox{red!70!black}{white}{
      \begin{minipage}[t]{5.25cm}
        {\tiny\textbf{\#2}}\\
        \noindent\hbox to 5.25cm{\includegraphics[width=1.10cm,height=1.8cm,keepaspectratio]{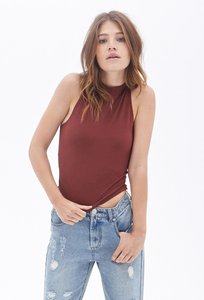}\hss\includegraphics[width=1.10cm,height=1.8cm,keepaspectratio]{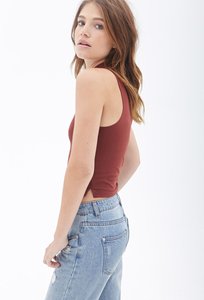}\hss\includegraphics[width=1.10cm,height=1.8cm,keepaspectratio]{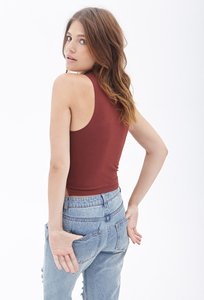}\hss\includegraphics[width=1.10cm,height=1.8cm,keepaspectratio]{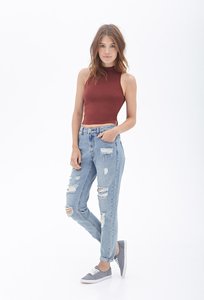}\hss\phantom{\rule{1.10cm}{1.8cm}}\hss}
      \end{minipage}
    }
  \end{minipage}
\hfill
  \begin{minipage}[t]{5.55cm}
    \fcolorbox{red!70!black}{white}{
      \begin{minipage}[t]{5.25cm}
        {\tiny\textbf{\#3}}\\
        \noindent\hbox to 5.25cm{\includegraphics[width=1.10cm,height=1.8cm,keepaspectratio]{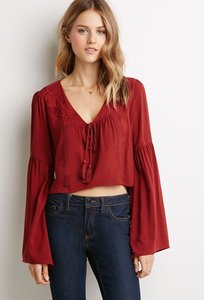}\hss\includegraphics[width=1.10cm,height=1.8cm,keepaspectratio]{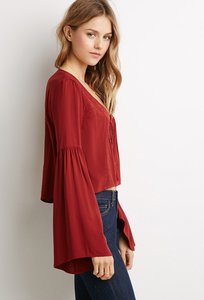}\hss\includegraphics[width=1.10cm,height=1.8cm,keepaspectratio]{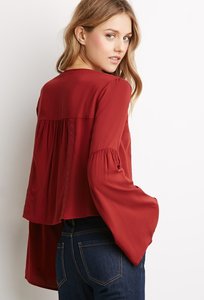}\hss\phantom{\rule{1.10cm}{1.8cm}}\hss\phantom{\rule{1.10cm}{1.8cm}}\hss}
      \end{minipage}
    }
  \end{minipage}
\par\vspace{2pt}
\noindent
  \begin{minipage}[t]{5.55cm}
    \fcolorbox{red!70!black}{white}{
      \begin{minipage}[t]{5.25cm}
        {\tiny\textbf{\#4}}\\
        \noindent\hbox to 5.25cm{\includegraphics[width=1.10cm,height=1.8cm,keepaspectratio]{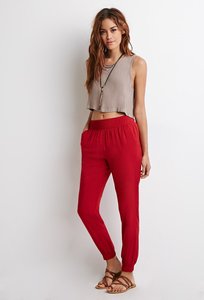}\hss\includegraphics[width=1.10cm,height=1.8cm,keepaspectratio]{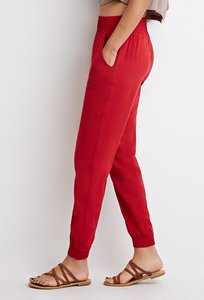}\hss\includegraphics[width=1.10cm,height=1.8cm,keepaspectratio]{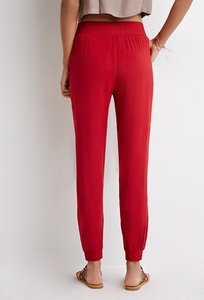}\hss\includegraphics[width=1.10cm,height=1.8cm,keepaspectratio]{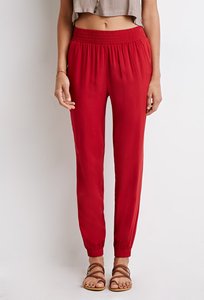}\hss\phantom{\rule{1.10cm}{1.8cm}}\hss}
      \end{minipage}
    }
  \end{minipage}
\hfill
  \begin{minipage}[t]{5.55cm}
    \fcolorbox{red!70!black}{white}{
      \begin{minipage}[t]{5.25cm}
        {\tiny\textbf{\#5}}\\
        \noindent\hbox to 5.25cm{\includegraphics[width=1.10cm,height=1.8cm,keepaspectratio]{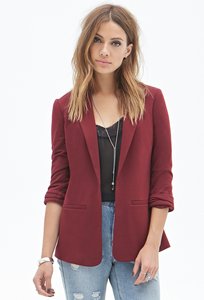}\hss\includegraphics[width=1.10cm,height=1.8cm,keepaspectratio]{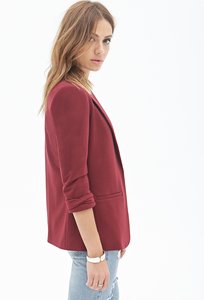}\hss\includegraphics[width=1.10cm,height=1.8cm,keepaspectratio]{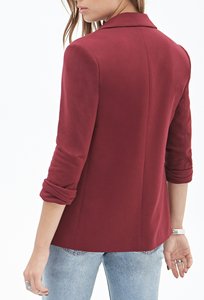}\hss\includegraphics[width=1.10cm,height=1.8cm,keepaspectratio]{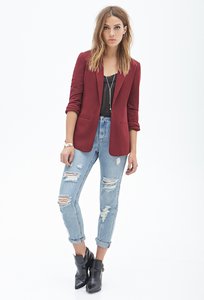}\hss\phantom{\rule{1.10cm}{1.8cm}}\hss}
      \end{minipage}
    }
  \end{minipage}
\hfill
  \begin{minipage}[t]{5.55cm}
    \fcolorbox{red!70!black}{white}{
      \begin{minipage}[t]{5.25cm}
        {\tiny\textbf{\#6}}\\
        \noindent\hbox to 5.25cm{\includegraphics[width=1.10cm,height=1.8cm,keepaspectratio]{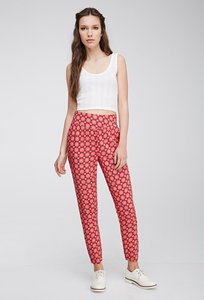}\hss\includegraphics[width=1.10cm,height=1.8cm,keepaspectratio]{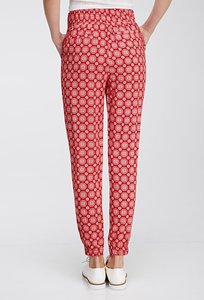}\hss\includegraphics[width=1.10cm,height=1.8cm,keepaspectratio]{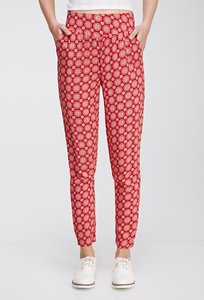}\hss\includegraphics[width=1.10cm,height=1.8cm,keepaspectratio]{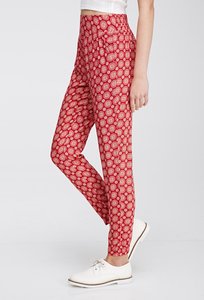}\hss\phantom{\rule{1.10cm}{1.8cm}}\hss}
      \end{minipage}
    }
  \end{minipage}
\par\vspace{2pt}
\noindent
  \begin{minipage}[t]{5.55cm}
    \fcolorbox{red!70!black}{white}{
      \begin{minipage}[t]{5.25cm}
        {\tiny\textbf{\#7}}\\
        \noindent\hbox to 5.25cm{\includegraphics[width=1.10cm,height=1.8cm,keepaspectratio]{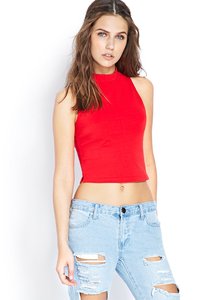}\hss\includegraphics[width=1.10cm,height=1.8cm,keepaspectratio]{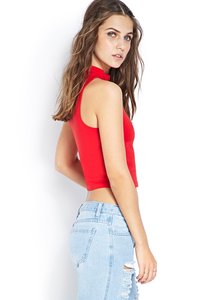}\hss\includegraphics[width=1.10cm,height=1.8cm,keepaspectratio]{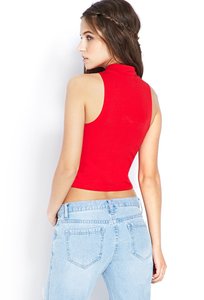}\hss\includegraphics[width=1.10cm,height=1.8cm,keepaspectratio]{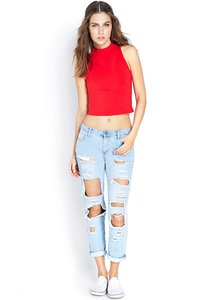}\hss\phantom{\rule{1.10cm}{1.8cm}}\hss}
      \end{minipage}
    }
  \end{minipage}
\hfill
  \begin{minipage}[t]{5.55cm}
    \fcolorbox{red!70!black}{white}{
      \begin{minipage}[t]{5.25cm}
        {\tiny\textbf{\#8}}\\
        \noindent\hbox to 5.25cm{\includegraphics[width=1.10cm,height=1.8cm,keepaspectratio]{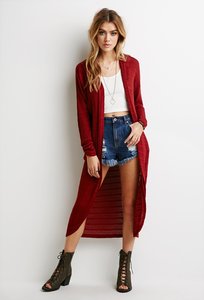}\hss\includegraphics[width=1.10cm,height=1.8cm,keepaspectratio]{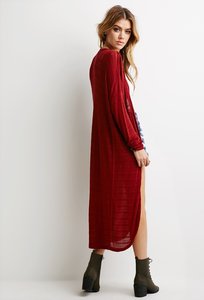}\hss\includegraphics[width=1.10cm,height=1.8cm,keepaspectratio]{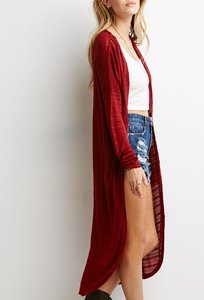}\hss\phantom{\rule{1.10cm}{1.8cm}}\hss\phantom{\rule{1.10cm}{1.8cm}}\hss}
      \end{minipage}
    }
  \end{minipage}
\hfill
  \begin{minipage}[t]{5.55cm}
    \fcolorbox{red!70!black}{white}{
      \begin{minipage}[t]{5.25cm}
        {\tiny\textbf{\#9}}\\
        \noindent\hbox to 5.25cm{\includegraphics[width=1.10cm,height=1.8cm,keepaspectratio]{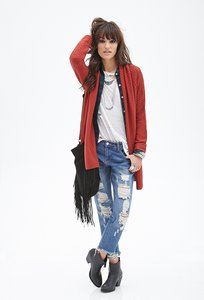}\hss\includegraphics[width=1.10cm,height=1.8cm,keepaspectratio]{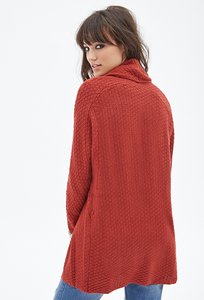}\hss\includegraphics[width=1.10cm,height=1.8cm,keepaspectratio]{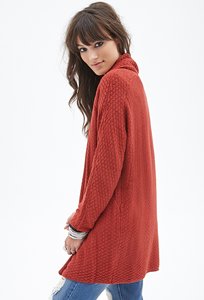}\hss\phantom{\rule{1.10cm}{1.8cm}}\hss\phantom{\rule{1.10cm}{1.8cm}}\hss}
      \end{minipage}
    }
  \end{minipage}
\par\vspace{2pt}
\noindent
  \begin{minipage}[t]{5.55cm}
    \fcolorbox{red!70!black}{white}{
      \begin{minipage}[t]{5.25cm}
        {\tiny\textbf{\#10}}\\
        \noindent\hbox to 5.25cm{\includegraphics[width=1.10cm,height=1.8cm,keepaspectratio]{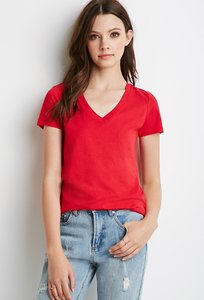}\hss\includegraphics[width=1.10cm,height=1.8cm,keepaspectratio]{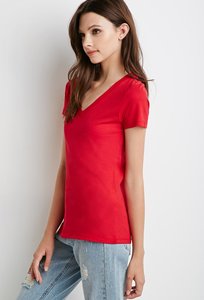}\hss\includegraphics[width=1.10cm,height=1.8cm,keepaspectratio]{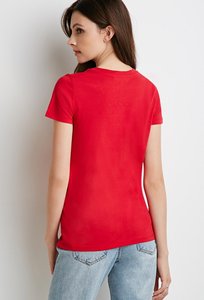}\hss\includegraphics[width=1.10cm,height=1.8cm,keepaspectratio]{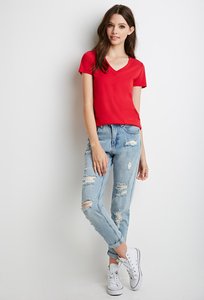}\hss\phantom{\rule{1.10cm}{1.8cm}}\hss}
      \end{minipage}
    }
  \end{minipage}
\hfill
  \begin{minipage}[t]{5.55cm}\end{minipage}
\hfill
  \begin{minipage}[t]{5.55cm}\end{minipage}
\par\vspace{2pt}
\noindent\rule{\linewidth}{0.4pt}
\par\vspace{2pt}
% -- doubao --
{\small\textbf{Doubao-E-V}}\quad{\small Rank~2}\\
\noindent
  \begin{minipage}[t]{5.55cm}
    \fcolorbox{red!70!black}{white}{
      \begin{minipage}[t]{5.25cm}
        {\tiny\textbf{\#1}}\\
        \noindent\hbox to 5.25cm{\includegraphics[width=1.10cm,height=1.8cm,keepaspectratio]{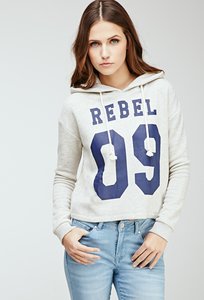}\hss\includegraphics[width=1.10cm,height=1.8cm,keepaspectratio]{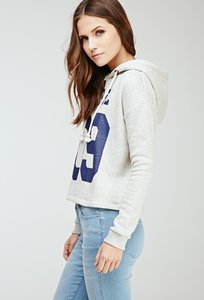}\hss\includegraphics[width=1.10cm,height=1.8cm,keepaspectratio]{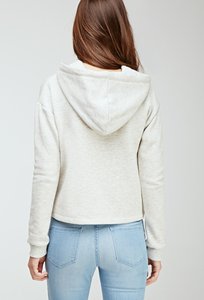}\hss\includegraphics[width=1.10cm,height=1.8cm,keepaspectratio]{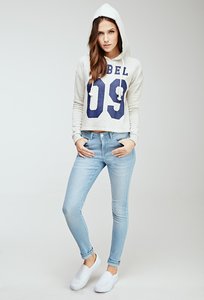}\hss\includegraphics[width=1.10cm,height=1.8cm,keepaspectratio]{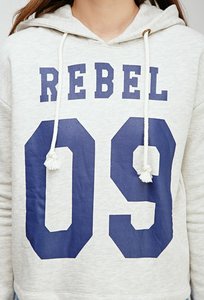}\hss}
      \end{minipage}
    }
  \end{minipage}
\hfill
  \begin{minipage}[t]{5.55cm}
    \fcolorbox{green!60!black}{white}{
      \begin{minipage}[t]{5.25cm}
        {\tiny\textbf{\#2}}\\
        \noindent\hbox to 5.25cm{\includegraphics[width=1.10cm,height=1.8cm,keepaspectratio]{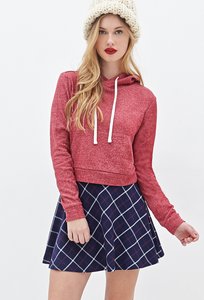}\hss\includegraphics[width=1.10cm,height=1.8cm,keepaspectratio]{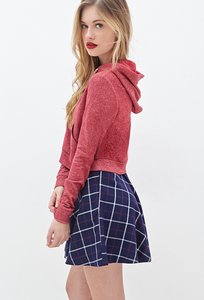}\hss\includegraphics[width=1.10cm,height=1.8cm,keepaspectratio]{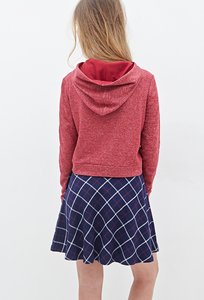}\hss\includegraphics[width=1.10cm,height=1.8cm,keepaspectratio]{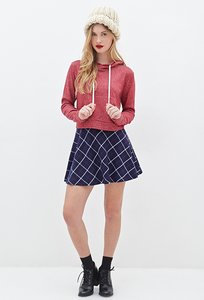}\hss\phantom{\rule{1.10cm}{1.8cm}}\hss}
      \end{minipage}
    }
  \end{minipage}
\hfill
  \begin{minipage}[t]{5.55cm}
    \fcolorbox{red!70!black}{white}{
      \begin{minipage}[t]{5.25cm}
        {\tiny\textbf{\#3}}\\
        \noindent\hbox to 5.25cm{\includegraphics[width=1.10cm,height=1.8cm,keepaspectratio]{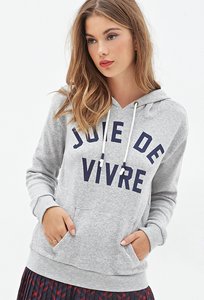}\hss\includegraphics[width=1.10cm,height=1.8cm,keepaspectratio]{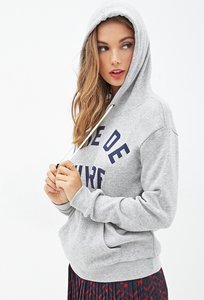}\hss\includegraphics[width=1.10cm,height=1.8cm,keepaspectratio]{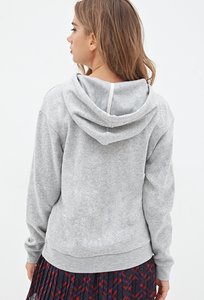}\hss\includegraphics[width=1.10cm,height=1.8cm,keepaspectratio]{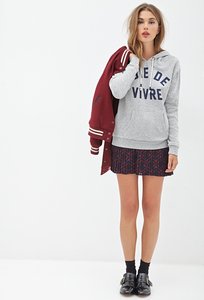}\hss\phantom{\rule{1.10cm}{1.8cm}}\hss}
      \end{minipage}
    }
  \end{minipage}
\par\vspace{2pt}
\noindent
  \begin{minipage}[t]{5.55cm}
    \fcolorbox{red!70!black}{white}{
      \begin{minipage}[t]{5.25cm}
        {\tiny\textbf{\#4}}\\
        \noindent\hbox to 5.25cm{\includegraphics[width=1.10cm,height=1.8cm,keepaspectratio]{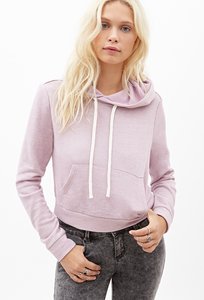}\hss\includegraphics[width=1.10cm,height=1.8cm,keepaspectratio]{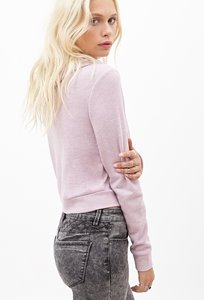}\hss\includegraphics[width=1.10cm,height=1.8cm,keepaspectratio]{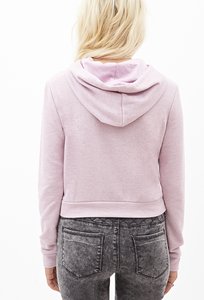}\hss\includegraphics[width=1.10cm,height=1.8cm,keepaspectratio]{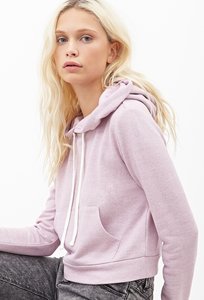}\hss\phantom{\rule{1.10cm}{1.8cm}}\hss}
      \end{minipage}
    }
  \end{minipage}
\hfill
  \begin{minipage}[t]{5.55cm}
    \fcolorbox{red!70!black}{white}{
      \begin{minipage}[t]{5.25cm}
        {\tiny\textbf{\#5}}\\
        \noindent\hbox to 5.25cm{\includegraphics[width=1.10cm,height=1.8cm,keepaspectratio]{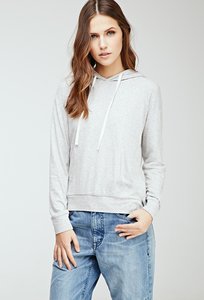}\hss\includegraphics[width=1.10cm,height=1.8cm,keepaspectratio]{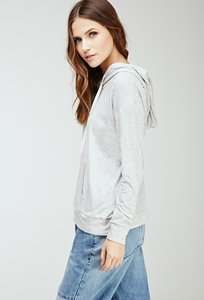}\hss\includegraphics[width=1.10cm,height=1.8cm,keepaspectratio]{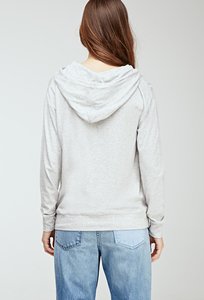}\hss\phantom{\rule{1.10cm}{1.8cm}}\hss\phantom{\rule{1.10cm}{1.8cm}}\hss}
      \end{minipage}
    }
  \end{minipage}
\hfill
  \begin{minipage}[t]{5.55cm}
    \fcolorbox{red!70!black}{white}{
      \begin{minipage}[t]{5.25cm}
        {\tiny\textbf{\#6}}\\
        \noindent\hbox to 5.25cm{\includegraphics[width=1.10cm,height=1.8cm,keepaspectratio]{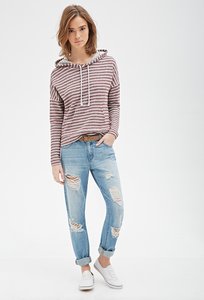}\hss\includegraphics[width=1.10cm,height=1.8cm,keepaspectratio]{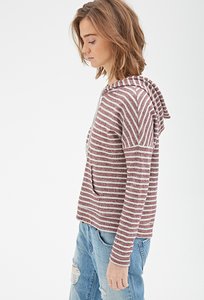}\hss\includegraphics[width=1.10cm,height=1.8cm,keepaspectratio]{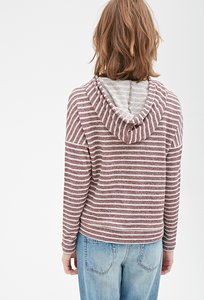}\hss\includegraphics[width=1.10cm,height=1.8cm,keepaspectratio]{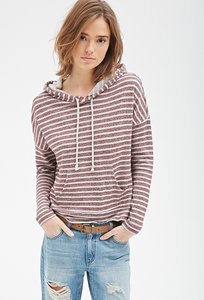}\hss\includegraphics[width=1.10cm,height=1.8cm,keepaspectratio]{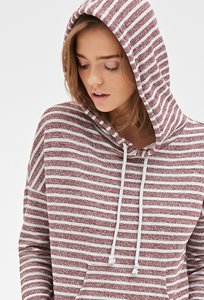}\hss}
      \end{minipage}
    }
  \end{minipage}
\par\vspace{2pt}
\noindent
  \begin{minipage}[t]{5.55cm}
    \fcolorbox{red!70!black}{white}{
      \begin{minipage}[t]{5.25cm}
        {\tiny\textbf{\#7}}\\
        \noindent\hbox to 5.25cm{\includegraphics[width=1.10cm,height=1.8cm,keepaspectratio]{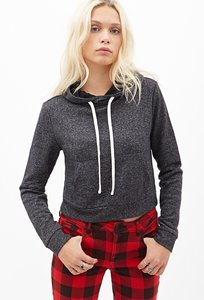}\hss\includegraphics[width=1.10cm,height=1.8cm,keepaspectratio]{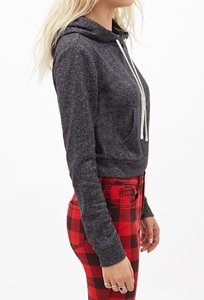}\hss\includegraphics[width=1.10cm,height=1.8cm,keepaspectratio]{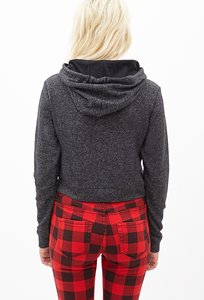}\hss\includegraphics[width=1.10cm,height=1.8cm,keepaspectratio]{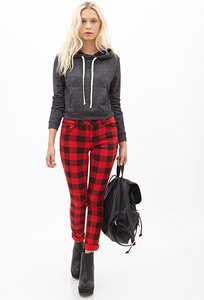}\hss\phantom{\rule{1.10cm}{1.8cm}}\hss}
      \end{minipage}
    }
  \end{minipage}
\hfill
  \begin{minipage}[t]{5.55cm}
    \fcolorbox{red!70!black}{white}{
      \begin{minipage}[t]{5.25cm}
        {\tiny\textbf{\#8}}\\
        \noindent\hbox to 5.25cm{\includegraphics[width=1.10cm,height=1.8cm,keepaspectratio]{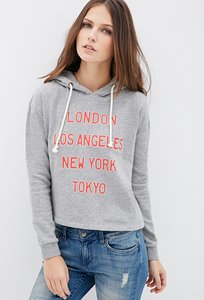}\hss\includegraphics[width=1.10cm,height=1.8cm,keepaspectratio]{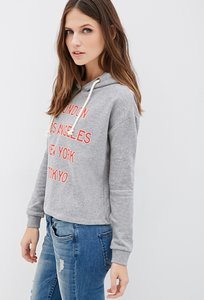}\hss\includegraphics[width=1.10cm,height=1.8cm,keepaspectratio]{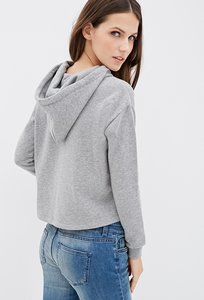}\hss\includegraphics[width=1.10cm,height=1.8cm,keepaspectratio]{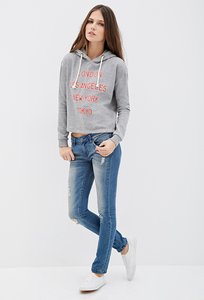}\hss\includegraphics[width=1.10cm,height=1.8cm,keepaspectratio]{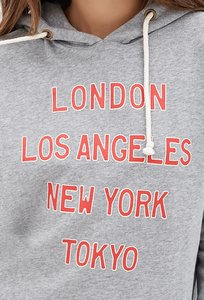}\hss}
      \end{minipage}
    }
  \end{minipage}
\hfill
  \begin{minipage}[t]{5.55cm}
    \fcolorbox{red!70!black}{white}{
      \begin{minipage}[t]{5.25cm}
        {\tiny\textbf{\#9}}\\
        \noindent\hbox to 5.25cm{\includegraphics[width=1.10cm,height=1.8cm,keepaspectratio]{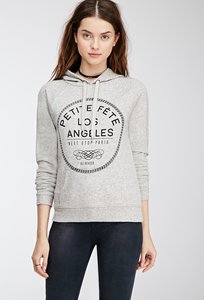}\hss\includegraphics[width=1.10cm,height=1.8cm,keepaspectratio]{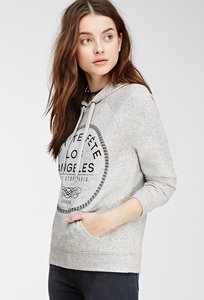}\hss\includegraphics[width=1.10cm,height=1.8cm,keepaspectratio]{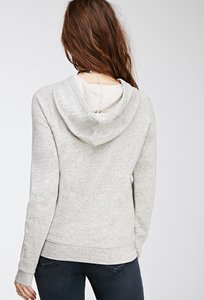}\hss\includegraphics[width=1.10cm,height=1.8cm,keepaspectratio]{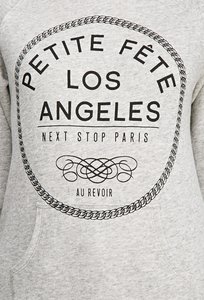}\hss\phantom{\rule{1.10cm}{1.8cm}}\hss}
      \end{minipage}
    }
  \end{minipage}
\par\vspace{2pt}
\noindent
  \begin{minipage}[t]{5.55cm}
    \fcolorbox{red!70!black}{white}{
      \begin{minipage}[t]{5.25cm}
        {\tiny\textbf{\#10}}\\
        \noindent\hbox to 5.25cm{\includegraphics[width=1.10cm,height=1.8cm,keepaspectratio]{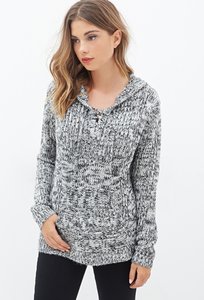}\hss\includegraphics[width=1.10cm,height=1.8cm,keepaspectratio]{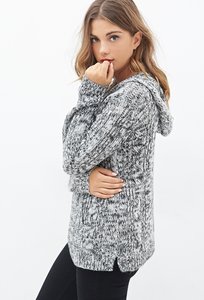}\hss\includegraphics[width=1.10cm,height=1.8cm,keepaspectratio]{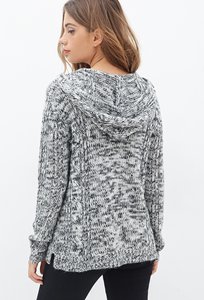}\hss\includegraphics[width=1.10cm,height=1.8cm,keepaspectratio]{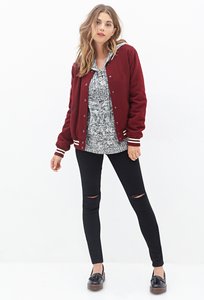}\hss\phantom{\rule{1.10cm}{1.8cm}}\hss}
      \end{minipage}
    }
  \end{minipage}
\hfill
  \begin{minipage}[t]{5.55cm}\end{minipage}
\hfill
  \begin{minipage}[t]{5.55cm}\end{minipage}
\par\vspace{2pt}
% \noindent\rule{\linewidth}{0.4pt}
\par\vspace{20pt}

\end{document}